\documentclass[11pt]{book}
\usepackage[a4paper,left=2.5cm,right=2.5cm,top=2.5cm,bottom=2.5cm]{geometry}
\usepackage{project}

\let\cleardoublepage\clearpage

\addto\captionsenglish{}

\def\projectauthor{Tong Xiao and Jingbo Zhu}
\def\projecttitle{{\fontsize{27pt}{32pt}\selectfont {\textbf{Foundations of}}}\\[0.3cm] {\fontsize{27pt}{32pt}\selectfont \textbf{Large Language Models}}}
\def\projectmonth{\today}
\def\projectdepartment{NLP Lab, Northeastern University \& NiuTrans Research}
\begin{document}
\prefrontmatter
%\frontmatter

\pagenumbering{roman}
%\newpage
~\vfill

\noindent Copyright \copyright\ 2021-2026 Tong Xiao and Jingbo Zhu\\

\noindent \textsc{Natural Language Processing Lab, Northeastern University}\\
\noindent \textsc{\&}\\
\noindent \textsc{NiuTrans Research}\\

\begin{spacing}{1.2}
\noindent Licensed under the Creative Commons Attribution-NonCommercial 4.0 Unported License (the ``License''). You may not use this file except in compliance with the License. You may obtain a copy of the License at \url{http://creativecommons.org/licenses/by-nc/4.0}. Unless required by applicable law or agreed to in writing, software distributed under the License is distributed on an \textsc{``as is'' basis, without warranties or conditions of any kind}, either express or implied. See the License for the specific language governing permissions and limitations under the License.
\end{spacing}

\noindent \textit{\today}
%\cleardoublepage

\chapter*{Preface}

\vspace{0.5em}
Large language models originated from natural language processing, but they have undoubtedly become one of the most revolutionary technological advancements in the field of artificial intelligence in recent years. An important insight brought by large language models is that knowledge of the world and languages can be acquired through large-scale language modeling tasks, and in this way, we can create a universal model that handles diverse problems. This discovery has profoundly impacted the research methodologies in natural language processing and many related disciplines. We have shifted from training specialized systems from scratch using a large amount of labeled data to a new paradigm of using large-scale pre-training to obtain foundation models, which are then fine-tuned, aligned, and prompted.

This book aims to outline the basic concepts of large language models and introduce the related techniques. As the title suggests, the book focuses more on the foundational aspects of large language models rather than providing comprehensive coverage of all cutting-edge methods. The book consists of six chapters:

\begin{itemize}
\item Chapter 1 introduces the basics of pre-training. This is the foundation of large language models, and common pre-training methods and model architectures will be discussed here.
\item Chapter 2 introduces generative models, which are the large language models we commonly refer to today. After presenting the basic process of building these models, we will also explore how to scale up model training and handle long texts.
\item Chapter 3 introduces prompting methods for large language models. We will discuss various prompting strategies, along with more advanced methods such as chain-of-thought reasoning and automatic prompt design.
\item Chapter 4 introduces alignment methods for large language models. We will focus on instruction fine-tuning and alignment based on human feedback.
\item Chapter 5 introduces inference methods for large language models. We will discuss various decoding algorithms, acceleration methods, and the inference-time scaling issue.
\item Chapter 6 introduces methods for improving the reasoning ability of large language models. We will discuss how training can strengthen reasoning, and how additional computation during inference can help models solve complex problems.
\end{itemize}

If readers have some background in machine learning and natural language processing, along with a certain understanding of neural networks like Transformers, reading this book will be quite easy. However, even without this prior knowledge, it is still perfectly fine, as we have made the content of each chapter as self-contained as possible, ensuring that readers will not be burdened with too much reading difficulty.

The content presented here is part of a comprehensive introductory resource on neural networks and large language models in natural language processing. For readers who wish to learn more about background topics, such as sequence modeling and attention mechanisms, you can visit \url{https://github.com/NiuTrans/NLPBook} or \url{https://niutrans.github.io/NLPBook} for further information.

We would like to thank the students in our laboratory and all our friends who have shared with us their views on large language models and helped with corrections of errors in writing. In particular, we wish to thank Weiqiao Shan, Yongyu Mu, Chenglong Wang, Kaiyan Chang, Yuchun Fan, Hang Zhou, Chuanhao Lv, Xinyu Liu, Tao Zhou, Huiwen Bao, Tong Zheng, Junhao Ruan, Yingfeng Luo, Yuzhang Wu, Yifu Huo, Xiaoqian Liu, Yucheng Qiao, Junxiang Zhang, Miaohe Niu, Junxin Wang, Xin Liu, Xinyi Xu, Dan Chen, Xihan Yang, and Jingqi Xiao.

\vspace{3em}
\begin{flushright}
Tong Xiao and Jingbo Zhu

\textit{\today}
\end{flushright}

\markboth{Preface}{}
\cleardoublepage

\chapter*{Notation}

\vspace{1.5em}

\begin{spacing}{1.6}
\begin{center}
\begin{tabular}{r l}
$a$ & variable \\
$\mathbf{a}$ & row vector or matrix \\
$\mathbf{a}^\top$ & transpose of vector or matrix $\mathbf{a}$ \\
$f(a)$ & function of $a$ \\
$\max f(a)$ & maximum value of $f(a)$ \\
$\argmax_{a} f(a)$ & value of $a$ that maximizes $f(a)$ \\
$\mathbf{x}$ & input token sequence to a model \\
$x_j$ & input token at position $j$ \\
$\mathbf{y}$ & output token sequence produced by a model \\
$y_i$ & output token at position $i$ \\
$|\mathbf{x}|$ (or $|\mathbf{y}|$) & length of sequence $\mathbf{x}$ (or $\mathbf{y}$)  \\
$V$ & vocabulary \\
$\theta$ & model parameters \\
$\Pr(a)$ & probability of $a$ \\
$\Pr(a|b)$ & conditional probability of $a$ given $b$ \\
$\Pr(\cdot|b)$ & probability distribution of a variable given $b$ \\
$\Pr_{\theta}(a)$ &  probability of $a$ as parameterized by $\theta$ \\
$\mathbb{E}_{a\sim \Pr(a)}[f(a)]$ & expected value of $f(a)$ when $a$ follows the distribution $\Pr(a)$ \\
$\mathbf{h}_t$ & hidden state at time step $t$ in sequential models \\
$\mathbf{H}$ & matrix of all hidden states over time in a sequence \\
$\mathbf{Q}$, $\mathbf{K}$, $\mathbf{V}$ & query, key, and value matrices in attention mechanisms \\
$\mathrm{Softmax}(\mathbf{A})$ & Softmax function that normalizes the input vector or matrix $\mathbf{A}$ \\
$\mathcal{L}$ & loss function \\
$\mathcal{D}$ & dataset used for training or fine-tuning a model \\
$\frac{\partial \mathcal{L}}{\partial \theta}$ & gradient of the loss function $\mathcal{L}$ with respect to the parameters $\theta$ \\
$D_{\mathrm{KL}}(p\ ||\ q)$ & KL divergence between distributions $p$ and $q$
\end{tabular}
\end{center}
\end{spacing}

\markboth{Notation}{}
\cleardoublepage

\pdfbookmark{\contentsname}{toc}
\renewcommand{\sectionmark}[1]{\markright{#1}}
% \addtolength{\parskip}{-\baselineskip}
\begin{spacing}{1.0}
\tableofcontents
\end{spacing}
% \addtolength{\parskip}{\baselineskip}
\renewcommand{\sectionmark}[1]{\markright{\thesection\ #1}}

\mainmatter
% !Mode:: "TeX:UTF-8"
% !TEX encoding = UTF-8 Unicode

%----------------------------------------------------------------------------------------
% Natural Language Processing: Neural Networks and Large Language Models
%
% Copyright 2021-2026
% Tong Xiao (xiaotong@mail.neu.edu.cn) and Jingbo Zhu (zhujingbo@mail.neu.edu.cn)
%----------------------------------------------------------------------------------------

%----------------------------------------------------------------------------------------
%    CONFIGURATIONS
%----------------------------------------------------------------------------------------

%\setcounter{chapter}{6}

%----------------------------------------------------------------------------------------
%	CHAPTER 7
%----------------------------------------------------------------------------------------

\chapter{Pre-training}

\thispagestyle{chapterurl} % add urls to the header

\noindent The development of neural sequence models, such as Transformers, along with the improvements in large-scale self-supervised learning, has opened the door to universal language understanding and generation. This achievement is largely driven by pre-training: we extract common components from many neural network-based systems and train them on massive amounts of unlabeled data using self-supervision. These pre-trained models serve as foundation models that can be easily adapted to different tasks via fine-tuning or prompting. As a result, the paradigm of NLP has been fundamentally transformed. In many cases, large-scale supervised learning for specific tasks is no longer required, and instead, we only need to adapt pre-trained foundation models.

While pre-training has gained popularity in recent NLP research, this concept dates back decades to the early days of deep learning. For example, early attempts to pre-train deep learning systems include unsupervised learning for RNNs, deep feedforward networks, autoencoders, and others~\cite{schmidhuber-etal:2015deep}. In the modern era of deep learning, we experienced a resurgence of pre-training, sparked in part by the large-scale unsupervised training of various word embedding models~\cite{Mikolov-et-al:2013distributed,pennington2014glove}. During the same period, pre-training also attracted significant interest in computer vision, where the backbone models were trained on relatively large labeled datasets such as ImageNet, and then applied to different downstream tasks~\cite{he-etal:2019rethinking,zoph-etal:2020rethinking}. Large-scale research on pre-training in NLP began with the development of language models using self-supervised learning. This family of models covers several well-known examples like \mindex{BERT}~\cite{devlin-etal:2019bert} and \mindex{GPT}~\cite{brown-etal:2020language}, all with a similar idea that general language understanding and generation can be achieved by training models to predict masked or future words in a huge amount of text. Despite the simplicity of this approach, the resulting models show remarkable capabilities in capturing linguistic structures, even without being explicitly trained to do so. The generality of the pre-training tasks leads to systems that exhibit strong performance in a large variety of NLP problems, even outperforming previously state-of-the-art supervised systems. More recently, pre-trained large language models have achieved greater success, showing the exciting prospects for more general artificial intelligence~\cite{bubeck-etal:2023sparks}.

This chapter discusses the concept of pre-training in the context of NLP. It begins with a general introduction to pre-training methods and their applications. BERT is then used as an example to illustrate how a sequence model is trained via a self-supervised task, called \mindex{masked language modeling}. This is followed by a discussion of methods for adapting pre-trained sequence models to various NLP tasks. As this chapter focuses primarily on the pre-training paradigm, we will defer the detailed discussion of generative large language models to subsequent chapters.

%----------------------------------------------------------------------------------------
% NEW SECTION
%----------------------------------------------------------------------------------------

\section{Pre-training in NLP}

\noindent Pre-training in NLP typically involves two types of tasks: sequence modeling (or sequence encoding) and sequence generation. While these tasks take different forms, for simplicity, we describe them using a single unified model defined as follows:
\begin{eqnarray}
\mathbf{o} & = & g(x_0,x_1,\cdots,x_m;\theta) \nonumber \\
           & = & g_{\theta}(x_0,x_1,\cdots,x_m),
\end{eqnarray}

\noindent where $\{x_0, x_1, \dots, x_m\}$ denotes a sequence of input tokens\footnote{Here we assume that tokens are the basic units of text produced by tokenization. Sometimes, we use the terms \textit{token} and \textit{word} interchangeably, though they have closely related but slightly different meanings in NLP.}, $x_0$ denotes a special symbol ($\langle s \rangle$ or $\mathbf{[\mathrm{CLS}]}$) attached to the beginning of the sequence, $g(\cdot;\theta)$ (also written as $g_{\theta}(\cdot)$) denotes a neural network with parameters $\theta$, and $\mathbf{o}$ denotes the output of the neural network. The form of the output $\mathbf{o}$ varies depending on the specific problem. For example, in token prediction tasks (such as language modeling), $\mathbf{o}$ is a distribution over a vocabulary; in sequence encoding tasks, $\mathbf{o}$ is a representation of the input sequence, often expressed as a sequence of real-valued vectors.

There are two fundamental aspects to consider here:

\begin{itemize}
\item \vspace{0.5em} Optimizing $\theta$ for a pre-training task. Unlike standard learning problems in NLP, pre-training does not assume specific downstream tasks to which the model will be applied. Instead, the goal is to train a model that can generalize across various tasks.
\item \vspace{0.3em} Applying the pre-trained model $g_{\hat{\theta}}(\cdot)$ to downstream tasks. To adapt the model to these specific tasks, we need to fine-tune the parameters $\hat{\theta}$ using labeled data or prompt the model with task descriptions.
\end{itemize}
\vspace{0.5em}

In this section, we discuss the basic concepts for addressing these two aspects.

\subsection{Unsupervised, Supervised and Self-supervised Pre-training}

\noindent In deep learning, pre-training refers to the process of optimizing a neural network before it is further fine-tuned and applied to downstream tasks. This approach is based on the assumption that a model pre-trained on one task can be effectively adapted to perform another. As a result, we do not need to train a deep, complex neural network from scratch on tasks with limited labeled data. Instead, we can leverage tasks where supervision signals are easier to obtain. This reduces the reliance on task-specific human annotations, enabling the development of more general models that are not confined to particular problems.

During the resurgence of neural networks driven by deep learning, many early attempts at pre-training were focused on \mindex{unsupervised learning}. In these methods, the parameters of a neural network are optimized using a criterion that is not directly related to specific target tasks. For example, a model can be trained to minimize the reconstruction cross-entropy of the input vector for each layer~\cite{bengio-etal:2006greedy}. Unsupervised pre-training is commonly employed as a preliminary step before supervised learning, offering several advantages, such as aiding in the discovery of better local minima and serving as a regularizer during the training process~\cite{erhan-etal:2010does}. These benefits make the subsequent supervised learning phase easier and more stable.

A second approach involves pre-training a neural network on \mindex{supervised learning} tasks. For example, consider a sequence model designed to encode input sequences into vector representations. During pre-training, this model is combined with a classification layer and trained on an initial supervised task, such as sentiment analysis (e.g., classifying whether a sentence is positive or negative). To adapt the pre-trained sequence model to a downstream task (e.g., determining if a sequence is subjective or objective), we replace the initial classification layer with a new one for the target task. Typically, we need to fine-tune the parameters of this combined model using task-specific labeled data to optimize its performance. The fine-tuned model is then employed to classify new sequences. A major advantage of supervised pre-training is its straightforward alignment with standard supervised learning paradigms. However, as model complexity increases, the demand for labeled data grows accordingly, making this pre-training approach challenging when large-scale labeled datasets are unavailable.

A third approach is \mindex{self-supervised learning}. In this paradigm, a neural network is trained using supervision signals derived from the data itself, rather than from human annotations. This is generally achieved by constructing training tasks directly from unlabeled data, such as predicting missing parts of the input. While self-supervised learning has recently become the dominant method in NLP, the underlying philosophy shares similarities with older machine learning concepts like \mindex{self-training}. In self-training, a model requires a small amount of seed data to build an initial classifier, which then generates pseudo-labels for unlabeled data. These pseudo-labels are subsequently used to iteratively refine and bootstrap the model. This method has been successfully applied to NLP areas like word sense disambiguation~\cite{yarowsky:1995unsupervised} and document classification~\cite{blum-and-tom:1998combining}. However, unlike standard self-training, self-supervised pre-training in NLP does not rely on an initial model to assign discrete pseudo-labels. Instead, supervision signals are derived directly from the text, and the entire model is trained from scratch. A well-known example is training sequence models by masking tokens in a text and requiring the model to predict them based on the surrounding context.  This enables large-scale self-supervised learning for deep neural networks, leading to the success of pre-training in many understanding, generation, and reasoning tasks.

Figure~\ref{fig:unsupervised-supervised-self-supervised-pre-training} shows a comparison of the above three pre-training approaches. Self-supervised pre-training is so successful that most current state-of-the-art NLP models are based on this paradigm. Therefore, throughout this chapter and the remainder of the book, we will focus on self-supervised pre-training. We will show how sequence models are pre-trained via self-supervision and how the pre-trained models are applied.

\begin{figure}[!t]
\centering
\input{./Chapter7/Figures/figure-unsupervised-supervised-selfsupervised-pre-training}
\caption{Illustration of unsupervised, supervised, and self-supervised pre-training. In unsupervised pre-training, pre-training is performed on large-scale unlabeled data. It can be viewed as a preliminary step to have a good starting point for the subsequent optimization process, though considerable effort is still required to further train the model with labeled data after pre-training. In supervised pre-training, the underlying assumption is that different (supervised) learning tasks are related. So we can first train the model on one task, and transfer the resulting model to another task with some training or tuning effort. In self-supervised pre-training, a model is pre-trained on large-scale unlabeled data via self-supervision. The model can be effectively trained in this way, and we can efficiently adapt it to new tasks through fine-tuning or prompting.}
\label{fig:unsupervised-supervised-self-supervised-pre-training}
\end{figure}

\subsection{Adapting Pre-trained Models}

\noindent As mentioned above, two major types of models are widely used in NLP pre-training.

\begin{itemize}
\item \vspace{0.5em} \mindex{Sequence Encoding Models}. Given a sequence of words or tokens, a sequence encoding model represents this sequence as either a real-valued vector or a sequence of vectors, and obtains a representation of the sequence. This representation is typically used as input to another model, such as a sentence classification system.
\item \vspace{0.3em} \mindex{Sequence Generation Models}. In NLP, sequence generation generally refers to the problem of generating a sequence of tokens based on a given context. The term \textit{context} has different meanings across applications. For example, it refers to the preceding tokens in language modeling, and refers to the source-language sequence in machine translation\footnote{More precisely, in autoregressive decoding of machine translation, each target-language token is generated based on both its preceding tokens and source-language sequence.}.
\end{itemize}
\vspace{0.5em}

We need different techniques for applying these models to downstream tasks after pre-training. Here we are interested in the following two methods.

\subsubsection{Fine-tuning of Pre-trained Models}
\label{sec:fine-tuning-encoders}

\noindent For sequence encoding pre-training, a common method of adapting pre-trained models is fine-tuning. Let $\mathrm{Encode}_{\theta}(\cdot)$ denote an encoder with parameters $\theta$, for example, $\mathrm{Encode}_{\theta}(\cdot)$ can be a standard Transformer encoder. Provided we have pre-trained this model in some way and obtained the optimal parameters $\hat{\theta}$, we can employ it to encode an input sequence, as follows
\begin{eqnarray}
\mathbf{H} & = & \mathrm{Encode}_{\hat{\theta}}(\mathbf{x}),
\end{eqnarray}

\noindent where $\mathbf{x}$ is the input sequence $\{x_0, \cdots, x_m\}$, and $\mathbf{H}$ is the output representation which is a sequence of real-valued vectors $\{\mathbf{h}_0,\mathbf{h}_1,\cdots,\mathbf{h}_m\}$. Because the encoder does not work as a standalone NLP system, it is often integrated as a component into a larger system. Consider, for example, a text classification problem in which we identify the polarity (e.g., positive, negative, or neutral) of a given text. We can build a text classification system by placing a classifier on top of the encoder. Let $\mathrm{Classify}_{\omega}(\cdot)$ be a neural network with parameters $\omega$. Then, the text classification model can be expressed in the form
\begin{eqnarray}
\mathrm{Pr}_{\omega,\hat{\theta}}(\cdot|\mathbf{x}) & = & \mathrm{Classify}_{\omega}(\mathbf{H}) \nonumber \\
& = & \mathrm{Classify}_{\omega}(\mathrm{Encode}_{\hat{\theta}}(\mathbf{x})).
\end{eqnarray}

\noindent Here $\mathrm{Pr}_{\omega,\hat{\theta}}(\cdot|\mathbf{x})$ is a probability distribution over the label set $\{\mathrm{positive}, \mathrm{negative}, \mathrm{neutral}\}$, and the label with the highest probability in this distribution is selected as the output. To keep the notation uncluttered, we will use $F_{\omega,\hat{\theta}}(\cdot)$ to denote $\mathrm{Classify}_{\omega}(\mathrm{Encode}_{\hat{\theta}}(\cdot))$.

Because the model parameters $\omega$ and $\hat{\theta}$ are not explicitly optimized for the target classification task, we cannot directly use this combined model. Instead, we must use a modified version of the model that is adapted to the task. A typical approach is to fine-tune the model by providing explicit labels from downstream tasks. We can train $F_{\omega,\hat{\theta}}(\cdot)$ on a labeled dataset, treating it as a standard supervised learning task. This fine-tuning process yields further optimized parameters, $\tilde{\omega}$ and $\tilde{\theta}$. Alternatively, we can freeze the encoder parameters $\hat{\theta}$ to maintain their pre-trained state and focus solely on optimizing $\omega$. This allows the classifier to be efficiently adapted to work effectively with the pre-trained encoder.

Once we have obtained a fine-tuned model, we can use it to classify a new text. For example, suppose we have a comment posted on a travel website:

\begin{center}
%\vspace{0.5em}
I love the food here. It's amazing!
\vspace{0.5em}
\end{center}

\noindent We first tokenize this text into tokens\footnote{The text can be tokenized in many different ways. One of the simplest is to segment the text into English words and punctuation $\{\textrm{I}, \textrm{love}, \textrm{the}, \textrm{food}, \textrm{here}, \textrm{.}, \textrm{It}, \textrm{'s}, \textrm{amazing}, \textrm{!}\}$}, and then feed the token sequence $\mathbf{x}_{\mathrm{new}}$ into the fine-tuned model $F_{\tilde{\omega},\tilde{\theta}}(\cdot)$. The model generates a distribution over classes by
\begin{eqnarray}
F_{\tilde{\omega},\tilde{\theta}}(\mathbf{x}_{\mathrm{new}}) & = & \begin{bmatrix} \Pr(\mathrm{positive}|\mathbf{x}_{\mathrm{new}}) & \Pr(\mathrm{negative}|\mathbf{x}_{\mathrm{new}}) & \Pr(\mathrm{neutral}|\mathbf{x}_{\mathrm{new}})\end{bmatrix}.
\end{eqnarray}

\noindent We then select the label with the highest probability as the output. In this example, it is $\mathrm{positive}$.

In general, the amount of labeled data used in fine-tuning is small compared to that of the pre-training data, and so fine-tuning is less computationally expensive. This makes the adaptation of pre-trained models very efficient in practice: given a pre-trained model and a downstream task, we just need to collect some labeled data, and slightly adjust the model parameters on this data. A more detailed discussion of fine-tuning can be found in Section~\ref{sec:pre-training-application}.

\subsubsection{Prompting of Pre-trained Models}

\noindent Unlike sequence encoding models, sequence generation models are often employed independently to address language generation problems, such as question answering and machine translation, without the need for additional modules. It is therefore straightforward to fine-tune these models as complete systems on downstream tasks. For example, we can fine-tune a pre-trained encoder-decoder multilingual model on some bilingual data to improve its performance on a specific translation task.

Among various sequence generation models, a notable example is large language models trained on very large amounts of data. These models are trained simply to predict the next token given its preceding tokens. Although token prediction is a simple task that has traditionally been used for language modeling, performing it at scale enables models to acquire broad linguistic and world knowledge. As a result, pre-trained LLMs exhibit remarkable token prediction capabilities, making it possible to frame numerous NLP tasks as simple text generation problems via prompting. For example, we can frame the above text classification problem as a text generation task.

\begin{center}
\vspace{0.5em}
I love the food here. It's amazing! I'm \underline{\hspace{3em}}
\vspace{0.5em}
\end{center}

\noindent Here \underline{\hspace{1em}} indicates the word or phrase we want to predict (call it the \mindex{completion}). If the predicted word is \textit{happy}, or \textit{glad}, or \textit{satisfied}, or a related positive word, we can classify the text as $\mathrm{positive}$. This example shows a simple prompting method in which we concatenate the input text with \textit{I'm} to form a prompt. Then, the completion helps decide which label is assigned to the original text.

Given the strong language understanding and generation capabilities of large language models, a prompt can instruct these models to perform more complex tasks. The following example demonstrates how to instruct an LLM to perform polarity classification:

\vspace{0.5em}
\begin{tcolorbox}[frame empty]

\begingroup
\setlength{\leftskip}{2em}
\setlength{\rightskip}{2em}

Assume that the polarity of a text is a label chosen from \{positive, negative, neutral\}. Identify the polarity of the input.

\vspace{0.2cm}

{\color{blue} Input}: I love the food here. It's amazing!

\vspace{0.2cm}

{\color{red} Polarity}: \underline{\hspace{3em}}

\endgroup

\end{tcolorbox}
\vspace{0.5em}

\noindent The first two sentences are a description of the task. {\color{blue} Input} and {\color{red} Polarity} are indicators of the input and output, respectively. We expect the model to complete the text and at the same time give the correct polarity label. By using instruction-based prompts, we can adapt large language models to solve NLP problems without the need for additional training.

This example also demonstrates the zero-shot learning capability of large language models, which can perform tasks that were not seen during training during. Another method for eliciting new capabilities from a neural network is few-shot learning. This is typically achieved through \mindex{in-context learning} (\mindex{ICL}). More specifically, we add some samples that demonstrate how an input corresponds to an output. These samples, known as \mindex{demonstrations}, are used to teach large language models how to perform the task. Below is an example with demonstrations

\vspace{0.5em}
\begin{tcolorbox}[frame empty]

\begingroup
\setlength{\leftskip}{2em}
\setlength{\rightskip}{2em}

Assume that the polarity of a text is a label chosen from \{positive, negative, neutral\}. Identify the polarity of the input.

\vspace{0.2cm}

{\color{blue} Input}: The traffic is terrible during rush hours, making it difficult to reach the airport on time.

\vspace{0.2cm}

{\color{red} Polarity}: Negative

\vspace{0.2cm}

{\color{blue} Input}: The weather here is wonderful.

\vspace{0.2cm}

{\color{red} Polarity}: Positive

\vspace{0.2cm}

{\color{blue} Input}: I love the food here. It's amazing!

\vspace{0.2cm}

{\color{red} Polarity}: \underline{\hspace{3em}}

\endgroup

\end{tcolorbox}
\vspace{0.5em}

Prompting and in-context learning play important roles in the recent rise of large language models. We will discuss these techniques in greater detail in Chapter \ChapterLLMPrompting. However, it is worth noting that while prompting is a powerful way to adapt large language models, some tuning efforts are still needed to ensure the models can follow instructions accurately. Additionally, the fine-tuning process is crucial for aligning the values of these models with human values. More detailed discussions of fine-tuning can be found in Chapter \ChapterLLMTuning.

\section{Self-supervised Pre-training Tasks}

\noindent In this section, we consider self-supervised pre-training approaches for different neural architectures, including decoder-only, encoder-only, and encoder-decoder architectures. We restrict our discussion to Transformers since they form the basis of most pre-trained models in NLP. However, pre-training is a broad concept, and so we just give a brief introduction to basic approaches in order to make this section concise.

\subsection{Decoder-only Pre-training}
\label{sec:decoder-only-pre-training}

\noindent The decoder-only architecture has been widely used in developing language models~\cite{radford-etal:2018improving}. For example, we can use a Transformer decoder as a language model by simply removing its cross-attention sub-layers. Such a model predicts the probability distribution over the vocabulary for the next token based on its preceding tokens, and the predicted output is the token with the highest probability. The standard way to train this model, as in traditional language modeling, is to minimize a loss function over a collection of token sequences.

Let $\mathrm{Decoder}_{\theta}(\cdot)$ denote a decoder with parameters $\theta$. At each position $i$, the decoder generates a distribution over the next token based on the preceding context $\{x_0, \dots, x_i\}$, denoted by $\mathrm{Pr}_{\theta}(\cdot|x_0, \dots, x_i)$ (or $\mathbf{p}_{i+1}^{\theta}$ for short). Suppose we have the gold-standard target distribution at the same position, denoted by $\mathbf{p}_{i+1}^{\mathrm{gold}}$. For language modeling, we can think of $\mathbf{p}_{i+1}^{\mathrm{gold}}$ as a one-hot representation of the correct next word. We then define a loss function $\mathcal{L}(\mathbf{p}_{i+1}^{\theta}, \mathbf{p}_{i+1}^{\mathrm{gold}})$ to measure the discrepancy between the model prediction and the target distribution. In NLP, the cross-entropy loss is typically used.

Given a token sequence of $\mathbf{x} = \{x_0,\cdots,x_m\}$, the loss on this sequence is the sum of the loss over the positions $\{0,\cdots,m-1\}$, given by
\begin{eqnarray}
\mathrm{Loss}_{\theta}(x_0,\cdots,x_m) & = & \sum_{i=0}^{m-1} \mathcal{L}(\mathbf{p}_{i+1}^{\theta},\mathbf{p}_{i+1}^{\mathrm{gold}}) \nonumber \\
                                       & = & \sum_{i=0}^{m-1} \mathrm{CrossEntropy}(\mathbf{p}_{i+1}^{\theta},\mathbf{p}_{i+1}^{\mathrm{gold}}),
\end{eqnarray}

\noindent where $\mathrm{CrossEntropy}(\cdot)$ denotes the standard cross-entropy loss, and $\mathbf{p}_{i+1}^{\mathrm{gold}}$ is the one-hot representation of $x_{i+1}$.

This loss function can be extended to a set of sequences $\mathcal{D}$. In this case, the objective of pre-training is to find the optimal parameters that minimize the loss on $\mathcal{D}$
\begin{eqnarray}
\hat{\theta} & = & \argmin_{\theta} \sum_{\mathbf{x} \in \mathcal{D}} \mathrm{Loss}_{\theta}(\mathbf{x}).
\end{eqnarray}

\noindent Note that this objective is mathematically equivalent to maximum likelihood estimation, and can be re-expressed as:
\begin{eqnarray}
\hat{\theta} & = & \arg\max_{\theta} \sum_{\mathbf{x} \in \mathcal{D}} \log \mathrm{Pr}_{\theta}(\mathbf{x}) \nonumber \\
             & = & \arg\max_{\theta} \sum_{\mathbf{x} \in \mathcal{D}} \sum_{i=0}^{|\mathbf{x}|-2} \log \mathrm{Pr}_{\theta}(x_{i+1} | x_0, \dots, x_{i}).
\end{eqnarray}

With these optimized parameters $\hat{\theta}$, we can use the pre-trained language model $\mathrm{Decoder}_{\hat{\theta}}(\cdot)$ to compute the probability $\mathrm{Pr}_{\hat{\theta}}(x_{i+1} | x_0,\cdots,x_i)$ at each position of a given sequence.

\subsection{Encoder-only Pre-training}
\label{sec:encoder-only-pre-training}

\noindent As defined in Section~\ref{sec:fine-tuning-encoders}, an encoder $\mathrm{Encode}_{\theta}(\cdot)$ is a function that reads a sequence of tokens $\mathbf{x} = \{x_0, \dots, x_m\}$ and produces a sequence of vectors $\mathbf{H} = \{\mathbf{h}_0, \dots, \mathbf{h}_m\}$\footnote{If we view $\mathbf{h}_i$ as a row vector, $\mathbf{H}$ can be written as \begin{align*}\mathbf{H} & = \begin{bmatrix} \mathbf{h}_0 \\ \vdots \\ \mathbf{h}_m \end{bmatrix}.\end{align*}}. Training this model is not straightforward, as we do not have gold-standard data for measuring the quality of the continuous representations produced by this function. A typical approach to encoder pre-training is to combine the encoder with a task-specific output layer to obtain supervision signals that are readily available. Figure~\ref{fig:pre-training-encoder-architecture} shows a common architecture for pre-training Transformer encoders, where we add a Softmax layer on top of the encoder. This architecture is essentially the same as that of the decoder-based language model, and the output is a sequence of probability distributions:
\begin{eqnarray}
\begin{bmatrix} \mathbf{p}_1^{\mathbf{W},\theta} \\ \vdots \\ \mathbf{p}_m^{\mathbf{W},\theta} \end{bmatrix} & = &  \mathrm{Softmax}_{\mathbf{W}}(\mathrm{Encode}_{\theta}(\mathbf{x})).
\end{eqnarray}

\begin{figure}[!t]
\centering
\input{./Chapter7/Figures/figure-pre-training-and-applying-encoders}
\caption{Pre-training a Transformer encoder (left) and then applying the pre-trained encoder (right). In the pre-training phase, the encoder, together with a Softmax layer, is trained via self-supervision. In the application phase, the Softmax layer is removed, and the pre-trained encoder is combined with a prediction network to address specific problems. In general, for better adaptation to these tasks, the system is fine-tuned using labeled data.}
\label{fig:pre-training-encoder-architecture}
\end{figure}

\noindent Here $\mathbf{p}_i^{\mathbf{W},\theta}$ is the output distribution $\Pr(\cdot|\mathbf{x})$ at position $i$. We use $\mathrm{Softmax}_{\mathbf{W}}(\cdot)$ to denote that the Softmax layer is parameterized by $\mathbf{W}$, that is, $\mathrm{Softmax}_{\mathbf{W}}(\mathbf{H}) = \mathrm{Softmax}(\mathbf{H} \cdot \mathbf{W})$. For notational simplicity, we will sometimes drop the superscripts $\mathbf{W}$ and $\theta$ affixed to each probability distribution.

The difference between this model and standard language models is that the output $\mathbf{p}_i$ has different meanings in encoder pre-training and language modeling. In language modeling, $\mathbf{p}_i$ is the probability distribution of predicting the next token. This follows an autoregressive decoding process: a language model only observes the words up to position $i$ and predicts the next. By contrast, in encoder pre-training, the entire sequence is observed at once, and thus it is trivial to predict a token when the model has already seen it.

\subsubsection{Masked Language Modeling}

\noindent One of the most popular methods of encoder pre-training is \mindex{masked language modeling}, which forms the basis of the well-known BERT model~\cite{devlin-etal:2019bert}. The idea of masked language modeling is to create prediction challenges by masking out some of the tokens in the input sequence and training a model to predict the masked tokens. In this sense, the conventional language modeling problem, which is sometimes called \mindex{causal language modeling}, is a special case of masked language modeling: at each position, we mask the tokens in the right context, and predict the token at this position using its left context. However, in causal language modeling, only the left context is used for token prediction, and the model does not have access to tokens in the right context. By contrast, in masked language modeling, all the unmasked tokens are used for word prediction, leading to a bidirectional model that makes predictions based on both left and right contexts.

More formally, for an input sequence $\mathbf{x}=x_0\cdots x_m$, suppose that we mask the tokens at positions $\mathcal{A}(\mathbf{x}) = \{i_1,\cdots,i_{u}\}$. Hence we obtain a masked token sequence $\bar{\mathbf{x}}$ where the token at each position in $\mathcal{A}(\mathbf{x})$ is replaced with a special symbol $[\mathrm{MASK}]$. For example, for the following sequence

\vspace{0.5em}
\begin{center}
The early bird catches the worm
\end{center}
\vspace{0.5em}

\noindent we may have a masked token sequence like this

\vspace{0.5em}
\begin{center}
The $[\mathrm{MASK}]$ bird catches the $[\mathrm{MASK}]$
\end{center}
\vspace{0.5em}

\noindent where we mask the tokens \textit{early} and \textit{worm} (i.e., $i_1 = 2$ and $i_2 = 6$).

Now we have two sequences $\mathbf{x}$ and $\bar{\mathbf{x}}$. The model is then optimized so that we can correctly predict $\mathbf{x}$ based on $\bar{\mathbf{x}}$. This can be thought of as an autoencoding-like process, and the training objective is to maximize the reconstruction probability $\Pr(\mathbf{x}|\bar{\mathbf{x}})$. Note that there is a simple position-wise alignment between $\mathbf{x}$ and $\bar{\mathbf{x}}$. Because an unmasked token in $\bar{\mathbf{x}}$ is the same as the token in $\mathbf{x}$ at the same position, there is no need to consider the prediction for this unmasked token. This leads to a simplified training objective which only maximizes the probabilities for masked tokens. We can express this objective in a maximum likelihood estimation fashion
\begin{eqnarray}
(\widehat{\mathbf{W}},\hat{\theta}) & = & \argmax_{\mathbf{W},\theta} \sum_{\mathbf{x} \in \mathcal{D}} \sum_{i \in \mathcal{A}(\mathbf{x})} \log \mathrm{Pr}_{i}^{\mathbf{W},\theta}(x_{i}|\bar{\mathbf{x}}),
\end{eqnarray}

\noindent or alternatively express it using the cross-entropy loss
\begin{eqnarray}
(\widehat{\mathbf{W}},\hat{\theta}) & = & \argmin_{\mathbf{W},\theta} \sum_{\mathbf{x} \in \mathcal{D}} \sum_{i \in \mathcal{A}(\mathbf{x})} \mathrm{CrossEntropy}(\mathbf{p}_{i}^{\mathbf{W},\theta},\mathbf{p}_{i}^{\mathrm{gold}}),
\end{eqnarray}

\noindent where $\mathrm{Pr}_k^{\mathbf{W},\theta}(x_k|\bar{\mathbf{x}})$ is the probability of the true token $x_k$ at position $k$ given the corrupted input $\bar{\mathbf{x}}$, and $\mathbf{p}_{k}^{\mathbf{W},\theta}$ is the probability distribution at position $k$ given the corrupted input $\bar{\mathbf{x}}$. To illustrate, consider the above example where two tokens of the sequence ``\textit{the early bird catches the worm}'' are masked. For this example, the objective is to maximize the sum of log-probabilities
\begin{eqnarray}
\mathrm{Loss} & = & \log \Pr(x_2 = \textit{early}|\bar{\mathbf{x}}=\textit{$[\mathrm{CLS}]$ The $\underbrace{[\mathrm{MASK}]}_{\bar{x}_2}$ bird catches the $\underbrace{[\mathrm{MASK}]}_{\bar{x}_6}$}) + \nonumber \\
  &   & \log \Pr(x_6 = \textit{worm}|\bar{\mathbf{x}}=\textit{$[\mathrm{CLS}]$ The $\underbrace{[\mathrm{MASK}]}_{\bar{x}_2}$ bird catches the $\underbrace{[\mathrm{MASK}]}_{\bar{x}_6}$}).
\end{eqnarray}

Once we obtain the optimized parameters $\widehat{\mathbf{W}}$ and $\hat{\theta}$, we can drop $\widehat{\mathbf{W}}$. Then, we can further fine-tune the pre-trained encoder $\mathrm{Encode}_{\hat{\theta}}(\cdot)$ or directly apply it to downstream tasks.

\subsubsection{Permuted Language Modeling}

\noindent While masked language modeling is simple and widely applied, it introduces new issues. One drawback is the use of a special token, $[\mathrm{MASK}]$, which is employed only during training but not at test time. This leads to a discrepancy between training and inference. Moreover, the auto-encoding process overlooks the dependencies between masked tokens. For example, in the previously discussed example, the prediction of $x_2$ (i.e., the first masked token) is made independently of $x_6$ (i.e., the second masked token), even though $x_6$ and $x_2$ are interdependent in the context of the sentence.

These issues can be addressed using the \mindex{permuted language modeling} approach to pre-training~\cite{yang-etal:2019xlnet}. Similar to causal language modeling, permuted language modeling involves making sequential predictions of tokens. However, unlike causal modeling where predictions follow the natural sequence of the text (like left-to-right or right-to-left), permuted language modeling allows for predictions in any order. The approach is straightforward: we determine an order for token predictions and then train the model in a standard language modeling manner, as described in Section~\ref{sec:decoder-only-pre-training}. Note that in this approach, the actual order of tokens in the text remains unchanged, and only the order in which we predict these tokens differs from standard language modeling. For example, consider a sequence of 5 tokens $x_0 x_1 x_2 x_3 x_4$. Let $\mathbf{e}_i$ represent the embedding of $x_i$ (i.e., combination of the token embedding and positional embedding). In standard language modeling, we would generate this sequence in the order of $x_0 \to x_1 \to x_2 \to x_3 \to x_4$. The probability of the sequence can be modeled via a generation process.
\begin{eqnarray}
\Pr(\mathbf{x}) & = & \Pr(x_0) \cdot \Pr(x_1|x_0) \cdot \Pr(x_2|x_0,x_1) \cdot \Pr(x_3|x_0,x_1,x_2) \cdot \nonumber \\
                &   & \Pr(x_4|x_0,x_1,x_2,x_3).
\end{eqnarray}

\noindent In practice, the model conditions on token embeddings rather than the tokens themselves. Therefore, we can rewrite the above expression as
\begin{eqnarray}
\Pr(\mathbf{x}) & = & \Pr(x_0) \cdot \Pr(x_1|\mathbf{e}_0) \cdot \Pr(x_2|\mathbf{e}_0,\mathbf{e}_1) \cdot \Pr(x_3|\mathbf{e}_0,\mathbf{e}_1, \mathbf{e}_2) \cdot \nonumber \\
                &   & \Pr(x_4|\mathbf{e}_0, \mathbf{e}_1, \mathbf{e}_2, \mathbf{e}_3).
\end{eqnarray}

Now, let us consider a different order for token prediction: $x_0 \to x_4 \to x_2 \to x_1 \to x_3$. The sequence generation process can then be expressed as follows:
\begin{eqnarray}
\Pr(\mathbf{x}) & = & \Pr(x_0) \cdot \Pr(x_4|\mathbf{e}_0) \cdot \Pr(x_2|\mathbf{e}_0, \mathbf{e}_4) \cdot \Pr(x_1|\mathbf{e}_0, \mathbf{e}_4, \mathbf{e}_2) \cdot \nonumber \\
                &   & \Pr(x_3|\mathbf{e}_0, \mathbf{e}_4, \mathbf{e}_2, \mathbf{e}_1) \label{eq:permuted-factorization-lm-example}.
\end{eqnarray}

This new prediction order allows some tokens to be generated based on a broader context, rather than being limited to just the preceding tokens as in standard language models. For example, in generating $x_3$, the model considers both its left context (i.e., $\{\mathbf{e}_0, \mathbf{e}_1, \mathbf{e}_2\}$) and right context (i.e., $\{\mathbf{e}_4\}$). The embeddings $\mathbf{e}_0, \mathbf{e}_1, \mathbf{e}_2, \mathbf{e}_4$ incorporate the positional information of $x_0, x_1, x_2, x_4$, preserving the original order of the tokens. As a result, this approach is somewhat akin to masked language modeling: we mask out $x_3$ and use its surrounding tokens $x_0, x_1, x_2, x_4$ to predict this token.

The implementation of permuted language models is relatively easy for Transformers. Because self-attention itself is permutation-invariant, and positional information is encoded separately using positional embeddings, we do not need to explicitly reorder the input sequence to have a factorization like Eq. (\ref{eq:permuted-factorization-lm-example}). Instead, permutation can be done by setting appropriate masks for self-attention. For example, consider the step of computing $\Pr(x_1 | \mathbf{e}_0, \mathbf{e}_4, \mathbf{e}_2)$. We can keep $x_0, x_1, x_2, x_3, x_4$ in their original positions and simply prevent $x_1$ from attending to $x_3$ during the self-attention computation, as illustrated below

\begin{center}
\begin{tikzpicture}
\def\nodesize{0.7cm}
\node [] (x0) at (0,0) {$x_0$};
\node [anchor=west] (x1) at ([xshift=0.3cm]x0.east) {$x_1$};
\node [anchor=west] (x2) at ([xshift=0.3cm]x1.east) {$x_2$};
\node [anchor=west] (x3) at ([xshift=0.3cm]x2.east) {$x_3$};
\node [anchor=west] (x4) at ([xshift=0.3cm]x3.east) {$x_4$};
\draw [->,thick] (x0.north) .. controls +(north:1em) and +(north:1em) .. ([xshift=-0.3em]x1.north);
\draw [->,thick] (x2.north) .. controls +(north:0.8em) and +(north:0.8em) .. ([xshift=0.3em]x1.north);
\draw [->,thick] (x4.north) .. controls +(north:1.7em) and +(north:1.7em) .. ([xshift=0.0em]x1.north);

\node [anchor=north,minimum width=\nodesize,minimum height=\nodesize,draw,fill=blue!70] (box0) at ([yshift=-0.4cm]x0.south) {};
\node [anchor=north,minimum width=\nodesize,minimum height=\nodesize,draw,fill=blue!70] (box1) at ([yshift=-0.4cm]x1.south) {};
\node [anchor=north,minimum width=\nodesize,minimum height=\nodesize,draw,fill=blue!70] (box2) at ([yshift=-0.4cm]x2.south) {};
\node [anchor=north,minimum width=\nodesize,minimum height=\nodesize,draw,fill=gray!30] (box3) at ([yshift=-0.4cm]x3.south) {};
\node [anchor=north,minimum width=\nodesize,minimum height=\nodesize,draw,fill=blue!70] (box4) at ([yshift=-0.4cm]x4.south) {};

\node [anchor=east] (label) at ([xshift=-1em,yshift=0.6em]box0.west) {Masks for Self-attention:};
\node [anchor=north west] (label1) at (label.south west) {\scriptsize{Blue box = valid attention}};
\node [anchor=north west] (label2) at ([yshift=0.3em]label1.south west) {\scriptsize{Gray box = blocked attention}};
\end{tikzpicture}
\end{center}

\noindent For a clearer illustration, we compare the self-attention masking results of causal language modeling, masked language modeling and permuted language modeling in Figure~\ref{fig:comparison-of-masking-for-pre-training-lms}.

\begin{figure}[!t]
\centering
\input{./Chapter7/Figures/figure-masking-for-different-language-models}
\caption{Comparison of self-attention masking results of causal language modeling, masked language modeling and permuted language modeling. The gray cell denotes that the token at position $j$ does not attend to the token at position $i$. The blue cell $(i,j)$ denotes that the token at position $j$ attends to the token at position $i$. $\mathbf{e}_{\mathrm{mask}}$ represents the embedding of the symbol $[\mathrm{MASK}]$, which is a combination of the token embedding and the positional embedding.}
\label{fig:comparison-of-masking-for-pre-training-lms}
\end{figure}

\subsubsection{Pre-training Encoders as Classifiers}
\label{sec:pre-training-as-classification}

\noindent Another common approach to training an encoder is to use classification tasks. In self-supervised learning, this is typically done by creating new classification challenges from the unlabeled text. There are many different ways to design the classification tasks. Here we present two popular tasks.

A simple method, called \mindex{next sentence prediction} (\mindex{NSP}), is presented in BERT's original paper~\cite{devlin-etal:2019bert}. NSP is based on the assumption that a good text encoder should capture the relationship between two sentences. To model this relationship, NSP uses the encoder output of two consecutive sentences $\mathrm{Sent}_{A}$ and $\mathrm{Sent}_{B}$ to determine whether $\mathrm{Sent}_{B}$ is the next sentence following $\mathrm{Sent}_{A}$. For example, suppose $\mathrm{Sent}_{A} = \textit{'It is raining .'}$ and $\mathrm{Sent}_{B} = \textit{'I need an umbrella .'}$. The input sequence of the encoder could be

\vspace{0.5em}
\begin{center}
$[\mathrm{CLS}]$ It is raining . $[\mathrm{SEP}]$ I need an umbrella . $[\mathrm{SEP}]$
\end{center}
\vspace{0.5em}

\noindent where $[\mathrm{CLS}]$ is the start symbol (i.e., $x_0$) which is commonly used in encoder pre-training, and $[\mathrm{SEP}]$ is a separator that separates the two sentences. The processing of this sequence follows a standard procedure of Transformer encoding: we first represent each token $x_i$ as its corresponding embedding $\mathbf{e}_i$, and then feed the embedding sequence $\{\mathbf{e}_{0},\cdots,\mathbf{e}_m\}$ into the encoder to obtain the output sequence $\{\mathbf{h}_{0},\cdots,\mathbf{h}_m\}$. Since $\mathbf{h}_{0}$ is generally considered as the representation of the entire sequence, we add a Softmax layer on top of it to construct a binary classification system. This process is illustrated as follows:

\vspace{0.5em}
\begin{center}
\begingroup
\renewcommand{\arraystretch}{1.2}
\setlength{\tabcolsep}{3pt}
\begin{tabular}{r c c c c c c c c c c c c}
token: & {\color{red} $[\mathrm{CLS}]$} & It & is & raining & . & $[\mathrm{SEP}]$ & I & need & an & umbrella & . & $[\mathrm{SEP}]$ \\
embedding: & {\color{red} $\mathbf{e}_0$} & $\mathbf{e}_1$ & $\mathbf{e}_2$ & $\mathbf{e}_3$ & $\mathbf{e}_4$ & $\mathbf{e}_5$ & $\mathbf{e}_6$ & $\mathbf{e}_7$ & $\mathbf{e}_8$ & $\mathbf{e}_9$ & $\mathbf{e}_{10}$ & $\mathbf{e}_{11}$ \\
 & {\color{red} $\downarrow$} & $\downarrow$ & $\downarrow$ & $\downarrow$ & $\downarrow$ & $\downarrow$ & $\downarrow$ & $\downarrow$ & $\downarrow$ & $\downarrow$ & $\downarrow$ & $\downarrow$ \\ \cline{2-13}
& \multicolumn{12}{|c|}{Encoder} \\ \cline{2-13}
 & {\color{red} $\downarrow$} & $\downarrow$ & $\downarrow$ & $\downarrow$ & $\downarrow$ & $\downarrow$ & $\downarrow$ & $\downarrow$ & $\downarrow$ & $\downarrow$ & $\downarrow$ & $\downarrow$ \\
 encoding: & {\color{red} $\mathbf{h}_0$} & $\mathbf{h}_1$ & $\mathbf{h}_2$ & $\mathbf{h}_3$ & $\mathbf{h}_4$ & $\mathbf{h}_5$ & $\mathbf{h}_6$ & $\mathbf{h}_7$ & $\mathbf{h}_8$ & $\mathbf{h}_9$ & $\mathbf{h}_{10}$ & $\mathbf{h}_{11}$ \\
 & $\downarrow$ \\
 & \multicolumn{3}{l}{Softmax} \\
 & $\downarrow$ \\
 & \multicolumn{3}{l}{Is Next or Not?}
\end{tabular}
\endgroup
\end{center}
\vspace{0.5em}

To generate training samples, we extract sentence pairs from a text corpus. A positive sample is created by selecting two actual consecutive sentences, while a negative sample is formed by pairing a sentence with a randomly chosen sentence from the corpus. As a result, training this model is the same as training a classifier. Typically, NSP is used as an additional training loss function for pre-training based on masked language modeling.

A second approach to training Transformer encoders as classifiers applies classification-based supervision signals at the token level. For example, \citet{clark-etal:2019electra} propose, in their ELECTRA model, training a Transformer encoder to identify whether each input token is identical to the original token or has been altered in some manner. The first step of this method is to generate a new sequence from a given sequence of tokens, where some of the tokens are altered. To do this, a small masked language model (call it the generator) is applied: we randomly mask some of the tokens, and train this model to predict the masked tokens. For each training sample, this masked language model outputs a token at each masked position, which may differ from the original token. At the same time, we train another Transformer encoder (call it the discriminator) to determine whether each predicted token is the same as the original token or altered. More specifically, we use the generator to generate a sequence where some of the tokens are replaced. Below is an illustration.

\vspace{0.5em}
\begin{center}
\begingroup
\renewcommand{\arraystretch}{1.2}
\begin{tabular}{r c c c c c c c c c}
original: & $[\mathrm{CLS}]$ & The & boy & spent & hours & working & on & toys & . \\
& $\downarrow$ & $\downarrow$ & $\downarrow$ & $\downarrow$ & $\downarrow$ & $\downarrow$ & $\downarrow$ & $\downarrow$ & $\downarrow$ \\
masked: & $[\mathrm{CLS}]$ & The & boy & spent & {\color{red} $[\mathrm{MASK}]$} & working & on & {\color{red} $[\mathrm{MASK}]$} & . \\
& $\downarrow$ & $\downarrow$ & $\downarrow$ & $\downarrow$ & {\color{red} $\downarrow$} & $\downarrow$ & $\downarrow$ & {\color{red} $\downarrow$} & $\downarrow$ \\ \cline{2-10}
& \multicolumn{9}{|c|}{Generator (small masked language model)} \\ \cline{2-10}
& $\downarrow$ & $\downarrow$ & $\downarrow$ & $\downarrow$ & {\color{red} $\downarrow$} & $\downarrow$ & $\downarrow$ & {\color{red} $\downarrow$} & $\downarrow$ \\
replaced: & $[\mathrm{CLS}]$ & The & boy & spent & {\color{red} decades} & working & on & {\color{red} toys} & .
\end{tabular}
\endgroup
\end{center}
\vspace{0.5em}

Then, we use the discriminator to label each of these tokens as $\mathrm{original}$ or $\mathrm{replaced}$, as follows

\vspace{0.5em}
\begin{center}
\begingroup
\renewcommand{\arraystretch}{1.2}
\setlength{\tabcolsep}{3pt}
\begin{tabular}{r c c c c c c c c c}
replaced: & $[\mathrm{CLS}]$ & The & boy & spent & {\color{red} decades} & working & on & {\color{red} toys} & . \\
& $\downarrow$ & $\downarrow$ & $\downarrow$ & $\downarrow$ & {\color{red} $\downarrow$} & $\downarrow$ & $\downarrow$ & {\color{red} $\downarrow$} & $\downarrow$ \\ \cline{2-10}
& \multicolumn{9}{|c|}{Discriminator (the model we want)} \\ \cline{2-10}
& $\downarrow$ & $\downarrow$ & $\downarrow$ & $\downarrow$ & {\color{red} $\downarrow$} & $\downarrow$ & $\downarrow$ & {\color{red} $\downarrow$} & $\downarrow$ \\
label: & \footnotesize{$\mathrm{original}$} & \footnotesize{$\mathrm{original}$} & \footnotesize{$\mathrm{original}$} & \footnotesize{$\mathrm{original}$} & {\color{red} \footnotesize{$\mathrm{replaced}$}} & \footnotesize{$\mathrm{original}$} & \footnotesize{$\mathrm{original}$} & {\color{red} \footnotesize{$\mathrm{original}$}} & \footnotesize{$\mathrm{original}$}
\end{tabular}
\endgroup
\end{center}
\vspace{0.5em}

For training, the generator is optimized as a masked language model with maximum likelihood estimation, and the discriminator is optimized as a classifier using a classification-based loss. In ELECTRA, the maximum likelihood-based loss and the classification-based loss are combined for jointly training both the generator and discriminator. An alternative approach is to use generative adversarial networks (GANs), that is, the generator is trained to fool the discriminator, and the discriminator is trained to distinguish the output of the generator from the true distribution. However, GAN-style training complicates the training task and is more difficult to scale up. Nevertheless, once training is complete, the generator is discarded, and the encoding part of the discriminator is applied as the pre-trained model for downstream tasks.

\subsection{Encoder-Decoder Pre-training}

\noindent In NLP, encoder-decoder architectures are often used to model sequence-to-sequence problems, such as machine translation and question answering. In addition to these typical sequence-to-sequence problems in NLP, encoder-decoder models can be extended to deal with many other problems. A simple idea is to treat text as both the input and output of a problem, and so we can directly apply encoder-decoder models. For example, given an input text, we can instruct a model to generate text describing its sentiment (e.g., \textit{positive}, \textit{negative}, or \textit{neutral}).

Such an idea allows us to develop a single text-to-text system to address any NLP problem. We can formulate different problems into the same text-to-text format. We first train an encoder-decoder model to gain general-purpose knowledge of language via self-supervision. This model is then fine-tuned for specific downstream tasks using targeted text-to-text data.

\subsubsection{Masked Encoder-Decoder Pre-training}

\noindent In \citet{raffel-etal:2020exploring}'s \mindex{T5} model, many different tasks are framed as the same text-to-text task. Each sample in T5 follows the format
\begin{eqnarray}
\textrm{Source Text} & \to & \textrm{Target Text} \nonumber
\end{eqnarray}

\noindent Here $\to$ separates the source text, which consists of a task description or instruction and the input given to the system, from the target text, which is the response to the input task. As an example, consider a task of translating from Chinese to English. A training sample can be expressed as:
\begin{eqnarray}
[\mathrm{CLS}]\ \textrm{Translate from Chinese to English: \begin{CJK*}{UTF8}{gbsn}你好！\end{CJK*}} & \to & \langle s \rangle\ \textrm{Hello!} \nonumber
\end{eqnarray}

\noindent where $[\mathrm{CLS}]$ and $\langle s \rangle$ are the start symbols on the source and target sides, respectively\footnote{We could use the same start symbol for different sequences. Here we use different symbols to distinguish between the sequences on the encoder and decoder-sides.}.

Likewise, we can express other tasks in the same way. For example
\begin{eqnarray}
&     & \mathrm{[CLS]}\ \textrm{\colorbox{gray!30}{Answer:} when was Albert Einstein born?} \nonumber \\
& \to & \langle s \rangle\ \textrm{He was born on March 14, 1879.} \nonumber \\
&     & \nonumber \\
&     & \mathrm{[CLS]}\ \textrm{\colorbox{gray!30}{Simplify:} the professor, who has published numerous papers in his field,} \nonumber \\
&     & \textrm{will be giving a lecture on the topic next week.} \nonumber \\
& \to & \langle s \rangle\ \textrm{The experienced professor will give a lecture next week.} \nonumber \\
&     & \nonumber \\
&     & \mathrm{[CLS]}\ \textrm{\colorbox{gray!30}{Text:} John bought a new car. Hypothesis: John has a car.} \nonumber \\
& \to & \langle s \rangle\ \textrm{Entailment} \nonumber \\
&     & \nonumber \\
&     & \mathrm{[CLS]}\ \textrm{\colorbox{gray!30}{Score the translation from English to Chinese.} English: when in Rome, do as} \nonumber \\
&     & \textrm{ the Romans do. Chinese: \begin{CJK*}{UTF8}{gbsn}人\ 在\ 罗马\ 就\ 像\ 罗马\ 人\ 一样\ 做事\ 。\end{CJK*}} \nonumber \\
& \to & \langle s \rangle\ 0.81 \nonumber
\end{eqnarray}

\noindent where instructions are highlighted in gray. Interestingly, in the last example, we frame a continuous scoring problem as a text generation task. Our goal is to generate the text string representing the number $0.81$, rather than outputting a continuous numerical value.

The approach described above provides a new framework of universal language understanding and generation. Both the task instructions and the problem inputs are provided to the system in text form. The system then follows the instructions to complete the task. This method puts different problems together, with the benefit of training a single model that can perform many tasks simultaneously.

In general, fine-tuning is necessary for adapting the pre-trained model to a specific downstream task. In this process, one can use different ways to instruct the model for the task, such as using a short name of the task as the prefix to the actual input sequence or providing a detailed description of the task. Because task instructions are provided as text and processed as part of the input, general instruction-following capabilities can be acquired during the pre-training phase. This greatly facilitates zero-shot learning. For instance, pre-trained models can often generalize to solve new problems where task-specific instructions have never been encountered during training.

There have been several powerful methods of self-supervised learning for either Transformer encoders or decoders. Applying these methods to pre-train encoder-decoder models is relatively straightforward. One common choice is to train encoder-decoder models as language models. For example, the encoder receives a sequence prefix, while the decoder generates the remaining sequence. However, this differs from standard causal language modeling, where the entire sequence is autoregressively generated from the first token. In our case, the encoder processes the prefix at once, and then the decoder predicts subsequent tokens in the manner of causal language modeling. Put more precisely, this is a \mindex{prefix language modeling} problem: a language model predicts the subsequent sequence given a prefix, which serves as the context for prediction.

Consider the following example
\begin{eqnarray}
\underbrace{\mathrm{[CLS]}\ \textrm{The puppies are frolicking}}_{\textrm{Prefix}} & \to & \underbrace{\langle s \rangle\ \textrm{outside the house .}}_{\textrm{Subsequent Sequence}} \nonumber
\end{eqnarray}

\noindent We can directly train an encoder-decoder model using examples like this. Then, the encoder learns to understand the prefix, and the decoder learns to continue writing based on this understanding. For large-scale pre-training, it is easy to create a large number of training examples from unlabeled text.

It is worth noting that for pre-trained encoder-decoder models to be effective in multilingual and cross-lingual tasks (e.g., machine translation), they should be trained on multilingual data. This typically requires that the vocabulary includes tokens from multiple languages. By doing so, the models can learn shared representations across different languages, thereby enabling capabilities in both language understanding and generation in a multilingual and cross-lingual context.

A second approach to pre-training encoder-decoder models is masked language modeling. In this approach, as discussed in Section~\ref{sec:encoder-only-pre-training}, tokens in a sequence are randomly replaced with a mask symbol, and the model is then trained to predict these masked tokens based on the corrupted sequence.

As an illustration, consider the task of masking and reconstructing the sentence

\vspace{0.5em}
\begin{center}
The puppies are frolicking outside the house .
\end{center}
\vspace{0.5em}

\noindent By masking two tokens (say, \textit{frolicking} and \textit{the}), we have the BERT-style input and output of the model, as follows
\begin{eqnarray}
&     & \mathrm{[CLS]}\ \textrm{The puppies are $\mathrm{[MASK]}$ outside $\mathrm{[MASK]}$ house .} \nonumber \\
& \to & \langle s \rangle\ \textrm{\underline{\hspace{1em}}\ \ \underline{\hspace{1em}}\ \ \underline{\hspace{1em}}\ \ frolicking\ \ \underline{\hspace{1em}}\ \ the \underline{\hspace{1em}}\ \ \underline{\hspace{1em}}} \nonumber
\end{eqnarray}

\noindent Here \underline{\hspace{1em}} denotes the masked position at which we do not make token predictions. By varying the percentage of the tokens in the text, this approach can be generalized towards either BERT-style training or language modeling-style training~\cite{song-etal:2019mass}. For example, if we mask out all the tokens, then the model is trained to generate the entire sequence
\begin{eqnarray}
&     & \mathrm{[CLS]}\ \textrm{$\mathrm{[MASK]}$ $\mathrm{[MASK]}$ $\mathrm{[MASK]}$ $\mathrm{[MASK]}$ $\mathrm{[MASK]}$ $\mathrm{[MASK]}$ $\mathrm{[MASK]}$ $\mathrm{[MASK]}$} \nonumber \\
& \to & \langle s \rangle\ \textrm{The puppies are frolicking outside the house .} \nonumber
\end{eqnarray}

\noindent In this case, we train the decoder as a language model.

Note that, in the context of the encoder-decoder architecture, we can use the encoder to read the masked sequence, and use the decoder to predict the original sequence. With this objective, we essentially have a denoising autoencoder: the encoder transforms a corrupted input into some hidden representation, and the decoder reconstructs the uncorrupted input from this hidden representation. Here is an example of input and output for denoising training.

\begin{eqnarray}
&     & \mathrm{[CLS]}\ \textrm{The puppies are $\mathrm{[MASK]}$ outside $\mathrm{[MASK]}$ house .} \nonumber \\
& \to & \langle s \rangle\ \textrm{The puppies are frolicking outside the house .} \nonumber
\end{eqnarray}

\noindent By learning to map from this corrupted sequence to its uncorrupted counterpart, the model gains the ability to understand on the encoder side and to generate on the decoder side. See Figure~\ref{fig:encoder-decoder-masked-lm} for an illustration of how an encoder-decoder model is trained with BERT-style and denoising autoencoding objectives.

\begin{figure}[!t]
\centering
\input{./Chapter7/Figures/figure-enc-dec-masked-language-modeling}
\caption{Training an encoder-decoder model using BERT-style and denoising autoencoding methods. In both methods, the input to the encoder is a corrupted token sequence where some tokens are masked and replaced with $\mathrm{[MASK]}$ (or $\mathrm{[M]}$ for short). The decoder predicts these masked tokens, but in different ways. In BERT-style training, the decoder only needs to compute the loss for the masked tokens, while the remaining tokens in the sequence can be simply treated as $\mathrm{[MASK]}$ tokens. In denoising autoencoding, the decoder predicts the sequence of all tokens in an autoregressive manner.  As a result, the loss is obtained by accumulating the losses of all these tokens, as in standard language modeling.}
\label{fig:encoder-decoder-masked-lm}
\end{figure}

As we randomly select tokens for masking, we can mask consecutive tokens~\cite{joshi-etal:2020spanbert}. Here is an example.
\begin{eqnarray}
&     & \mathrm{[CLS]}\ \textrm{The puppies are $\mathrm{[MASK]}$ outside $\mathrm{[MASK]}$ $\mathrm{[MASK]}$ .} \nonumber \\
& \to & \langle s \rangle\ \textrm{The puppies are frolicking outside the house .} \nonumber
\end{eqnarray}

Another way to consider consecutive masked tokens is to represent them as spans. Here we follow \citet{raffel-etal:2020exploring}'s work, and use $\mathrm{[X]}$, $\mathrm{[Y]}$ and $\mathrm{[Z]}$ to denote sentinel tokens that cover one or more consecutive masked tokens. Using this notation, we can re-express the above training example as:
\begin{eqnarray}
&     & \mathrm{[CLS]}\ \textrm{The puppies are $\mathrm{[X]}$ outside $\mathrm{[Y]}$ .} \nonumber \\
& \to & \langle s \rangle\ \textrm{$\mathrm{[X]}$ frolicking $\mathrm{[Y]}$ the house $\mathrm{[Z]}$} \nonumber
\end{eqnarray}

The idea is that we represent the corrupted sequence as a sequence containing placeholder slots. The training task is to fill these slots with the correct tokens using the surrounding context. An advantage of this approach is that the sequences used in training would be shorter, making the training more efficient. Note that masked language modeling provides a very general framework for training encoder-decoder models. Various settings can be adjusted to have different training versions, such as altering the percentage of tokens masked and the maximum length of the masked spans.

\subsubsection{Denoising Training}

\noindent If we view encoder-decoder pre-training as a denoising autoencoding problem, there are many ways to corrupt the input. For instance, beyond random token masking, we can also modify tokens or rearrange their order.

Suppose we have an encoder-decoder model that can map an input sequence $\mathbf{x}$ to an output sequence $\mathbf{y}$
\begin{eqnarray}
\mathbf{y} & = & \mathrm{Decode}_{\omega}(\mathrm{Encode}_{\theta}(\mathbf{x})) \nonumber \\
           & = & \mathrm{Model}_{\theta,\omega}(\mathbf{x}),
\end{eqnarray}

\noindent where $\theta$ and $\omega$ are the parameters of the encoder and the decoder, respectively. In denoising autoencoding, we add some noise to $\mathbf{x}$ to obtain a noisy, corrupted input $\mathbf{x}_{\mathrm{noise}}$. By feeding $\mathbf{x}_{\mathrm{noise}}$ into the encoder, the decoder is trained to recontruct the original input. The training objective can be defined as:
\begin{eqnarray}
(\hat{\theta},\hat{\omega}) & = & \argmin_{\theta,\omega} \mathrm{Loss}(\mathrm{Model}_{\theta,\omega}(\mathbf{x}_{\mathrm{noise}}),\mathbf{x}),
\end{eqnarray}

\noindent Here the loss function $\mathrm{Loss}(\mathrm{Model}_{\theta,\omega}(\mathbf{x}_{\mathrm{noise}}),\mathbf{x})$ evaluates how well the model $\mathrm{Model}_{\theta,\omega}(\mathbf{x}_{\mathrm{noise}})$ reconstructs the original input $\mathbf{x}$. We can choose the cross-entropy loss as usual.

Given the model architecture and training framework, the remaining issue is how to corrupt the input. \citet{lewis-etal:2020bart}, in their \mindex{BART} model, propose several ways to corrupt the input sequence.

\begin{itemize}
\item \vspace{0.5em} \mindex{Token Masking}. This is the same masking method that we used in masked language modeling. Tokens are randomly selected and replaced with $\mathrm{[MASK]}$.
\item \vspace{0.3em} \mindex{Token Deletion}. This method is similar to token masking. However, rather than replacing the selected tokens with a special symbol $\mathrm{[MASK]}$, these tokens are removed from the sequence. See the following example for a comparison of the token masking and token deletion methods.

    \begin{center}
    \begingroup
    \renewcommand{\arraystretch}{1.2}
    \begin{tabular}{r l}
    Original ($\mathbf{x}$): & The puppies are \underline{frolicking} outside \underline{the} house . \\
    Token Masking ($\mathbf{x}_{\mathrm{noise}}$): & The puppies are $\mathrm{[MASK]}$ outside $\mathrm{[MASK]}$ house . \\
    Token Deletion ($\mathbf{x}_{\mathrm{noise}}$): & The puppies are {\color{gray} \sout{frolicking}} outside {\color{gray} \sout{the}} house .
    \end{tabular}
    \endgroup
    \end{center}

    \noindent where the underlined tokens in the original sequence are masked or deleted.

\item \vspace{0.3em} \mindex{Span Masking}. Non-overlapping spans are randomly sampled over the sequence. Each span is masked by $\mathrm{[MASK]}$. We also allow spans of length 0, and, in such cases, $\mathrm{[MASK]}$ is simply inserted at a position in the sequence. For example, we can use span masking to corrupt the above sequence as

    \begin{center}
    \begingroup
    \renewcommand{\arraystretch}{1.2}
    \begin{tabular}{r l}
    Original ($\mathbf{x}$): & The \underline{0} puppies are \underline{frolicking outside the} house . \\
    Span Masking ($\mathbf{x}_{\mathrm{noise}}$): & The $\mathrm{[MASK]}$ puppies are $\mathrm{[MASK]}$ house .
    \end{tabular}
    \endgroup
    \end{center}

    \noindent Here the span \textit{frolicking outside the} is replaced with a single $\mathrm{[MASK]}$. $\underline{0}$ indicates a length-0 span, and so we insert an $\mathrm{[MASK]}$ between \textit{The} and \textit{puppies}. Span masking introduces new prediction challenges in which the model needs to know how many tokens are generated from a span. This problem is very similar to fertility modeling in machine translation~\cite{brown-etal:1993mathematics}.
\end{itemize}
\vspace{0.5em}

If we consider a sequence consisting of multiple sentences, additional methods of corruption can be applied. In the BART model, there are two such methods.

\begin{itemize}
\item \vspace{0.5em} \mindex{Sentence Reordering}. This method randomly permutes the sentences so that the model can learn to reorder sentences in a document. Consider, for example, two consecutive sentences

    \begin{center}
    Hard work leads to success . Success brings happiness .
    \end{center}

    We can reorder the two sentences to have a corrupted input sequence

    \begin{center}
    Success brings happiness . Hard work leads to success .
    \end{center}

\item \vspace{0.3em} \mindex{Document Rotation}. The goal of this task is to identify the start token of the sequence. First, a token is randomly selected from the sequence. Then, the sequence is rotated so that the selected token is the first token. For example, suppose we select the token \textit{leads} from the above sequence. The rotated sequence is

    \begin{center}
    \begin{tikzpicture}
    \node [anchor=west] (seg1) at (0,0) {leads to success . Success brings happiness .};
    \node [anchor=east] (seg0) at ([yshift=0.1em]seg1.west) {{\color{gray} Hard work}};
    \node [anchor=west] (seg2) at ([yshift=0.1em]seg1.east) {Hard work};
    \draw [-] ([xshift=0.2em]seg0.west) -- ([xshift=-0.2em]seg0.east);
    \draw [->,dashed] ([yshift=-0.2em]seg0.south) .. controls +(south:2em) and +(south:2em) .. ([yshift=-0.2em]seg2.south);
    \node [anchor=south] (selected) at ([xshift=1.3em,yshift=0.8em]seg1.north west) {\footnotesize{selected}};
    \draw [->] (selected.south) -- ([yshift=-0.7em]selected.south);
    \end{tikzpicture}
    \end{center}

    \noindent where the subsequence \textit{Hard work} before \textit{leads} is appended to the end of the sequence.

\end{itemize}
\vspace{0.5em}

For pre-training, we can apply multiple corruption methods to learn robust models, for example, we randomly choose one of them for each training sample. In practice, the outcome of encoder-decoder pre-training depends heavily on the input corruption methods used, and so we typically need to choose appropriate training objectives through careful experimentation.

\subsection{Comparison of Pre-training Tasks}

\noindent So far, we have discussed a variety of pre-training tasks. Because the same training objective can often be applied to different architectures (e.g., using masked language modeling for both encoder-only and encoder-decoder pre-training), categorizing these tasks solely based on model architecture is insufficient. Instead, we summarize them based on their training objectives.

\begin{itemize}
\item \vspace{0.5em} \textbf{Language Modeling}. Typically, this approach refers to an autoregressive process of sequence generation. At each step, the model predicts the next token based exclusively on its preceding context.
\item \vspace{0.3em} \textbf{Masked Language Modeling}. Masked language modeling follows a general mask-predict framework. It randomly masks tokens in a sequence and predicts these tokens using the entire masked sequence.
\item \vspace{0.3em} \textbf{Permuted Language Modeling}. Permuted language modeling shares similarities with masked language modeling but removes the fixed left-to-right restriction by permuting the factorization order. It keeps the input sequence fixed but reorders the factorization (prediction) order. Each prediction is based on some context tokens that are randomly selected.
\item \vspace{0.3em} \textbf{Discriminative Training}. In discriminative training, supervision signals are created from classification tasks. Pre-trained models are integrated into classifiers and trained together with the remaining parts of the classifiers to enhance their classification performance.
\item \vspace{0.3em} \textbf{Denoising Autoencoding}. This approach is applied to the pre-training of encoder-decoder models. The input is a corrupted sequence (with noise such as masking, deletion, or permutation), and the model is trained to reconstruct the original sequence.
\end{itemize}
\vspace{0.5em}

Table~\ref{tab:comparison-of-pre-training-tasks} illustrates these methods and their variants using examples. While the input/output formats in the examples are generally model-agnostic, we explicitly indicate the compatible model architectures for each pre-training task. In each example, the input consists of a token sequence, and the output is either a token sequence or some probabilities. For generation tasks, such as language modeling, superscripts are used to indicate the generation order on the target side. If the superscripts are omitted, it implies that the output tokens can be predicted either autoregressively or in parallel (non-autoregressively). On the source side, we assume that the sequence undergoes a standard Transformer encoding process, meaning that each token can see the entire sequence in self-attention. The only exception is in permuted language modeling, where an autoregressive generation process is implemented by setting attention masks on the encoder side. To simplify the discussion, we remove the token $\langle s \rangle$ from the target-side of each example.

\begin{table}[!t]
\centering
\begingroup
\makebox[\textwidth]{
\small{
\setlength{\tabcolsep}{3pt}
\renewcommand{\arraystretch}{1.4}
\begin{tabular}{r | c | c | c | l | l }
Method & Enc & Dec & E-D & Input & Output \\ \hline
{\color{purple} Causal LM} & & $\bullet$ & $\bullet$ & & The$^1$ kitten$^2$ is$^3$ chasing$^4$ the$^5$ ball$^6$ .$^7$ \\ \hline
{\color{purple} Prefix LM} & & $\bullet$ & $\bullet$ & $\mathrm{[C]}$ The kitten is & chasing$^1$ the$^2$ ball$^3$ .$^4$ \\ \hline \hline
{\color{teal} Masked LM} & $\bullet$ & & $\bullet$ & $\mathrm{[C]}$ The kitten $\mathrm{[M]}$ chasing the $\mathrm{[M]}$ . & \underline{\hspace{0.7em}}\ \ \underline{\hspace{0.7em}}\ \ is\ \ \underline{\hspace{0.7em}}\ \ \underline{\hspace{0.7em}}\ \ ball\ \ \underline{\hspace{0.7em}}\\ \hline
{\color{teal} MASS-style} & $\bullet$ & & $\bullet$ & $\mathrm{[C]}$ The kitten $\mathrm{[M]}$ $\mathrm{[M]}$ $\mathrm{[M]}$ ball . & \underline{\hspace{0.7em}}\ \ \underline{\hspace{0.7em}}\ \ is\ \ chasing the\ \ \underline{\hspace{0.7em}}\ \ \underline{\hspace{0.7em}} \\ \hline
{\color{teal} BERT-style} & $\bullet$ & & $\bullet$ & $\mathrm{[C]}$ The kitten $\mathrm{[M]}$ {\color{gray} playing} the $\mathrm{[M]}$ . & \underline{\hspace{0.7em}}\ \ kitten\ \ is\ \ chasing\ \ \underline{\hspace{0.7em}}\ \ ball\ \ \underline{\hspace{0.7em}}\\ \hline \hline
{\color{brown} Permuted LM} & $\bullet$ & & & $\mathrm{[C]}$ The kitten is chasing the ball . & The$^5$ kitten$^7$ is$^6$ chasing$^1$ the$^4$ ball$^2$ .$^3$ \\ \hline \hline
{\color{cyan} Next Sentence} & \multirow{2}{*}{$\bullet$} & & & $\mathrm{[C]}$ The kitten is chasing the ball . & $\Pr(\textrm{IsNext}\ |\ \textrm{representation-of-$\mathrm{[C]}$})$ \\
{\color{cyan} Prediction} & & & & Birds eat worms . & \\ \hline
{\color{cyan} Sentence} & \multirow{2}{*}{$\bullet$} & & & $\mathrm{Sent}_a$: The kitten is chasing the ball . & $\mathrm{Score}(\mathrm{Sent}_a, \mathrm{Sent}_b)$ \\
{\color{cyan} Comparison} & & & & $\mathrm{Sent}_b$: A cat is running after a toy . & \\ \hline
{\color{cyan} Token Classification} & $\bullet$ & & & $\mathrm{[C]}$ The kitten is chasing the ball . & $\Pr(\cdot|\textrm{The})$ $\Pr(\cdot|\textrm{kitten})$ ... $\Pr(\cdot|\textrm{.})$ \\ \hline \hline
{\color{orange} Token Reordering} & & & $\bullet$ & $\mathrm{[C]}$ . kitten the chasing The is ball & The$^1$ kitten$^2$ is$^3$ chasing$^4$ the$^5$ ball$^6$ .$^7$ \\ \hline
{\color{orange} Token Deletion} & & & $\bullet$ & $\mathrm{[C]}$ The kitten is {\color{gray} \sout{chasing}} the ball . & The$^1$ kitten$^2$ is$^3$ chasing$^4$ the$^5$ ball$^6$ .$^7$ \\ \hline
{\color{orange} Span Masking} & & & $\bullet$ & $\mathrm{[C]}$ The kitten $\mathrm{[M]}$ is $\mathrm{[M]}$ . & The$^1$ kitten$^2$ is$^3$ chasing$^4$ the$^5$ ball$^6$ .$^7$  \\ \hline
{\color{orange} Sentinel Masking} & & & $\bullet$ & $\mathrm{[C]}$ The kitten $\mathrm{[X]}$ the $\mathrm{[Y]}$ & $\mathrm{[X]}^1$ is$^2$ chasing$^3$ $\mathrm{[Y]}^4$ ball$^5$ .$^6$  \\ \hline
{\color{orange} Sentence} & & & $\bullet$ & $\mathrm{[C]}$ The ball rolls away swiftly . The & The$^1$ kitten$^2$ is$^3$ chasing$^4$ the$^5$ ball$^6$ .$^7$ \\
{\color{orange} Reordering} & & & &  kitten is chasing the ball . & The$^8$ ball$^9$ rolls$^{10}$ away$^{11}$ swiftly$^{12}$ .$^{13}$ \\ \hline
{\color{orange} Document} & & & $\bullet$ & $\mathrm{[C]}$ chasing the ball . The ball rolls & The$^1$ kitten$^2$ is$^3$ chasing$^4$ the$^5$ ball$^6$ .$^7$ \\
{\color{orange} Rotation} & & & &  away swiftly . The kitten is & The$^8$ ball$^9$ rolls$^{10}$ away$^{11}$ swiftly$^{12}$ .$^{13}$
\end{tabular}
}
} % end of makebox
\endgroup
\caption{Comparison of pre-training tasks, including {\color{purple} language modeling}, {\color{teal} masked language modeling}, {\color{brown} permuted language modeling}, {\color{cyan} discriminative training}, and {\color{orange} denoising autoencoding}. $\mathrm{[C]} = \mathrm{[CLS]}$, $\mathrm{[M]} = \mathrm{[MASK]}$, $\mathrm{[X]}, \mathrm{[Y]} = \textrm{sentinel tokens}$. Enc, Dec and E-D indicate whether the approach can be applied to encoder-only, decoder-only, encoder-decoder models, respectively. For generation tasks, superscripts are used to represent the order of the tokens.}
\label{tab:comparison-of-pre-training-tasks}
\end{table}

While these pre-training tasks are different, it is possible to compare them in the same framework and experimental setup~\cite{dong-etal:2019unified,raffel-etal:2020exploring,lewis-etal:2020bart}. Note that we cannot list all the pre-training tasks here as there are many of them. For more discussions on pre-training tasks, the interested reader may refer to some surveys on this topic~\cite{qiu-etal:2020pre,han-etal:2021pre}.

%----------------------------------------------------------------------------------------
% NEW SECTION
%----------------------------------------------------------------------------------------

\section{Example: BERT}

\noindent In this section, we introduce BERT, one of the most widely used pre-trained sequence encoding models in NLP.

\subsection{The Standard Model}

\noindent The standard BERT model, introduced by \citet{devlin-etal:2019bert}, is a bidirectional Transformer encoder trained using two self-supervised tasks: masked language modeling and next sentence prediction. The overall training loss is simply the sum of the losses from these two tasks:
\begin{eqnarray}
\mathrm{Loss}_{\mathrm{BERT}} & = & \mathrm{Loss}_{\mathrm{mlm}} + \mathrm{Loss}_{\mathrm{nsp}}.
\end{eqnarray}

\noindent As is standard in deep learning, we optimize the model parameters by minimizing this loss. During training, we sample mini-batches from a text corpus, compute the $\mathrm{Loss}_{\mathrm{BERT}}$ over each batch, and iteratively update the model parameters using gradient descent or its variants. This process repeats until a stopping criterion is met, such as the convergence of the training loss or reaching a predefined maximum number of steps.

\subsubsection{Loss Functions}

In general, BERT models are used to represent a single sentence or a pair of sentences, and thus can handle various downstream language understanding problems. In this section we assume that the input representation is a sequence containing two sentences $\mathrm{Sent}_{A}$ and $\mathrm{Sent}_{B}$, given by
\begin{equation}
\mathrm{[CLS]}\ \ \mathrm{Sent}_{A}\ \ \mathrm{[SEP]}\ \ \mathrm{Sent}_{B}\ \ \mathrm{[SEP]} \nonumber
\end{equation}

\noindent Here we follow the notation of \cite{devlin-etal:2019bert} and use $\mathrm{[SEP]}$ to denote the separator.

Given this sequence, we can obtain $\mathrm{Loss}_{\mathrm{mlm}}$ and $\mathrm{Loss}_{\mathrm{nsp}}$ separately. For masked language modeling, a subset of tokens (typically 15\%) is randomly selected for prediction. Then the sequence is modified in three ways:

\begin{itemize}
\item \vspace{0.5em} \textbf{Token Masking}. $80\%$ of the selected tokens are masked and replaced with the symbol $\mathrm{[MASK]}$. For example
\begin{eqnarray}
\textrm{Original:} & & \textrm{$[\mathrm{CLS}]$ It is \underline{raining} . $[\mathrm{SEP}]$ \underline{I} need \underline{an} \underline{umbrella} . $[\mathrm{SEP}]$} \nonumber \\
\textrm{Masked:} &  & \textrm{$[\mathrm{CLS}]$ It is $\mathrm{[MASK]}$ . $[\mathrm{SEP}]$ I need $\mathrm{[MASK]}$ umbrella . $[\mathrm{SEP}]$} \nonumber
\end{eqnarray}

\noindent where the selected tokens are underlined. Predicting masked tokens encourages the model to learn representations from surrounding context.

\item \vspace{0.3em} \textbf{Random Replacement}. $10\%$ of the selected tokens are changed to a random token. For example
\begin{eqnarray}
\textrm{Original:} & & \textrm{$[\mathrm{CLS}]$ It is \underline{raining} . $[\mathrm{SEP}]$ \underline{I} need \underline{an} \underline{umbrella} . $[\mathrm{SEP}]$} \nonumber \\
\textrm{Random Token:} & & \textrm{$[\mathrm{CLS}]$ It is raining . $[\mathrm{SEP}]$ I need a {\color{red} hat} . $[\mathrm{SEP}]$} \nonumber
\end{eqnarray}

\noindent This helps the model learn to recover a token from a corrupted input.

\item \vspace{0.3em} \textbf{Unchanged}. $10\%$ of the selected tokens are kept unchanged. For example,
\begin{eqnarray}
\textrm{Original:} & & \textrm{$[\mathrm{CLS}]$ It is \underline{raining} . $[\mathrm{SEP}]$ \underline{I} need \underline{an} \underline{umbrella} . $[\mathrm{SEP}]$} \nonumber \\
\textrm{Unchanged Token:} & & \textrm{$[\mathrm{CLS}]$ It is raining . $[\mathrm{SEP}]$ {\color{red} I} need an umbrella . $[\mathrm{SEP}]$} \nonumber
\end{eqnarray}

Although predicting an unchanged token is trivial, this strategy is crucial. It encourages the model to maintain a meaningful representation of the actual observed word, thereby helping to mitigate the discrepancy between pre-training (which uses $\mathrm{[MASK]}$) and fine-tuning (which does not).

\end{itemize}
\vspace{0.3em}

Let $\mathcal{A}(\mathbf{x})$ denote the set of selected positions in a token sequence $\mathbf{x}$, and $\bar{\mathbf{x}}$ denote the modified sequence of $\mathbf{x}$. The loss function of masked language modeling can be defined as:
\begin{eqnarray}
\mathrm{Loss}_{\mathrm{mlm}} & = & - \sum_{i \in \mathcal{A}(\mathbf{x})} \log \mathrm{Pr}_i(x_i|\bar{\mathbf{x}}),
\end{eqnarray}

\noindent where $\mathrm{Pr}_i(x_i|\bar{\mathbf{x}})$ is the probability of predicting $x_i$ at the position $i$ given $\bar{\mathbf{x}}$. Figure~\ref{fig:example-bert-mlm-loss} shows a running example of computing $\mathrm{Loss}_{\mathrm{mlm}}$.

\begin{figure}[!t]
\centering
\input{./Chapter7/Figures/figure-running-example-of-bert}
\caption{A running example of BERT-style masked language modeling. First, $15\%$ of the tokens are randomly selected. These selected tokens are then processed in one of three ways: replaced with a $\mathrm{[MASK]}$ token ($80\%$ of the time), replaced with a random token ($10\%$ of the time), or kept unchanged ($10\%$ of the time). The model is trained to predict these selected tokens based on the modified sequence. $\mathbf{e}_i$ represents the embedding of the token at the position $i$. Gray boxes represent the Softmax layers.}
\label{fig:example-bert-mlm-loss}
\end{figure}

For next sentence prediction, we follow the method described in Section~\ref{sec:pre-training-as-classification}. Each training sample is assigned a label from the set $\{\mathrm{IsNext}, \mathrm{NotNext}\}$, for example,
\begin{eqnarray}
\textrm{Sequence:} & & \textrm{$[\mathrm{CLS}]$ It is raining . $[\mathrm{SEP}]$ I need an umbrella . $[\mathrm{SEP}]$} \nonumber \\
\textrm{Label:} & & \textrm{$\mathrm{IsNext}$} \nonumber \\
& & \nonumber \\
\textrm{Sequence:} & & \textrm{$[\mathrm{CLS}]$ The cat sleeps on the windowsill . $[\mathrm{SEP}]$ Apples grow on trees . $[\mathrm{SEP}]$} \nonumber \\
\textrm{Label:} & & \textrm{$\mathrm{NotNext}$} \nonumber
\end{eqnarray}

\noindent The output vector of the encoder for the first token $\mathrm{[CLS]}$ is viewed as the sequence representation, denoted by $\mathbf{h}_{\mathrm{cls}}$ (or $\mathbf{h}_0$). A classifier is built on top of $\mathbf{h}_{\mathrm{cls}}$. Then, we can compute the probability of a label $c$ given $\mathbf{h}_{\mathrm{cls}}$, i.e., $\Pr(c|\mathbf{h}_{\mathrm{cls}})$. There are many loss functions one can choose for classification problems. For example, in maximum likelihood training, we can define $\mathrm{Loss}_{\mathrm{nsp}}$ as
\begin{eqnarray}
\mathrm{Loss}_{\mathrm{nsp}} & = & -\log \Pr(c_{\mathrm{gold}}|\mathbf{h}_{\mathrm{cls}}),
\end{eqnarray}

\noindent where $c_{\mathrm{gold}}$ is the correct label for this sample.

\subsubsection{Model Setup}

\noindent As shown in Figure~\ref{fig:bert-architecture}, BERT models are based on the standard Transformer encoder architecture. The input is a sequence of embeddings, each of which is the sum of the token embedding, the positional embedding, and the segment embedding.
\begin{eqnarray}
\mathbf{e} & = & \mathbf{x} + \mathbf{e}_{\mathrm{pos}} + \mathbf{e}_{\mathrm{seg}}.
\end{eqnarray}

\noindent Both the token embedding ($\mathbf{x}$) and the positional embedding ($\mathbf{e}_{\mathrm{pos}}$) are identical to those used in the original Transformer. The segment embedding ($\mathbf{e}_{\mathrm{seg}}$) is a new addition that indicates whether a token belongs to $\mathrm{Sent}_{A}$ or $\mathrm{Sent}_{B}$. This is illustrated by the following example:

\vspace{0.5em}
\begin{center}
\begingroup
\renewcommand{\arraystretch}{1.2}
\setlength{\tabcolsep}{2pt}
{\small
\makebox[\textwidth]{
\begin{tabular}{r | c c c c c c c c c c c c}
Token & $[\mathrm{CLS}]$ & It & is & raining & . & $[\mathrm{SEP}]$ & I & need & an & umbrella & . & $[\mathrm{SEP}]$ \\ \hline
$\mathbf{x}$ & $\mathbf{x}_{0}$ & $\mathbf{x}_{1}$ & $\mathbf{x}_{2}$ & $\mathbf{x}_{3}$ & $\mathbf{x}_{4}$ & $\mathbf{x}_{5}$ & $\mathbf{x}_{6}$ & $\mathbf{x}_{7}$ & $\mathbf{x}_{8}$ & $\mathbf{x}_{9}$ & $\mathbf{x}_{10}$ & $\mathbf{x}_{11}$ \\
$\mathbf{e}_{\mathrm{pos}}$ & \footnotesize{$\mathrm{PE}(0)$} & \footnotesize{$\mathrm{PE}(1)$} & \footnotesize{$\mathrm{PE}(2)$} & \footnotesize{$\mathrm{PE}(3)$} & \footnotesize{$\mathrm{PE}(4)$} & \footnotesize{$\mathrm{PE}(5)$} & \footnotesize{$\mathrm{PE}(6)$} & \footnotesize{$\mathrm{PE}(7)$} & \footnotesize{$\mathrm{PE}(8)$} & \footnotesize{$\mathrm{PE}(9)$} & \footnotesize{$\mathrm{PE}(10)$} & \footnotesize{$\mathrm{PE}(11)$} \\
$\mathbf{e}_{\mathrm{seg}}$ & $\mathbf{e}_{A}$ & $\mathbf{e}_{A}$ & $\mathbf{e}_{A}$ & $\mathbf{e}_{A}$ & $\mathbf{e}_{A}$ & $\mathbf{e}_{A}$ & $\mathbf{e}_{B}$ & $\mathbf{e}_{B}$ & $\mathbf{e}_{B}$ & $\mathbf{e}_{B}$ & $\mathbf{e}_{B}$ & $\mathbf{e}_{B}$
\end{tabular}
}
}
\endgroup
\end{center}
\vspace{0.5em}

\begin{figure}[!t]
\centering
\input{./Chapter7/Figures/figure-bert-architecture}
\caption{The model architecture of BERT (Transformer encoder). The input tokens are first represented as embeddings, each of which is the sum of the corresponding token embedding, positional embedding, and segment embedding. Then, the embedding sequence is processed by a stack of Transformer layers. Each layer in this stack includes a self-attention sub-layer and an FFN sub-layer. The output of the BERT model is a sequence of vectors produced by the final Transformer layer.}
\label{fig:bert-architecture}
\end{figure}

The main part of BERT models is a multi-layer Transformer network. A Transformer layer consists of a self-attention sub-layer and an FFN sub-layer. Both sub-layers follow the post-norm architecture: $\mathrm{output} = \mathrm{LNorm}(F(\mathrm{input}) + \mathrm{input})$, where $F(\cdot)$ is the core function of the sub-layer (either a self-attention model or an FFN), and $\mathrm{LNorm}(\cdot)$ is the layer normalization unit. Typically, a number of Transformer layers are stacked to form a deep network. For each position in the sequence, the final output representation is the real-valued vector produced by the uppermost layer.

There are several aspects one may consider in developing BERT models.

\begin{itemize}
\item \vspace{0.5em} \textbf{Vocabulary Size} ($|V|$). In Transformers, each token corresponds to an entry in a vocabulary $V$. Large vocabularies can cover more surface form variants of words, but may lead to increased storage requirements.
\item \vspace{0.3em} \textbf{Embedding Size} ($d_e$). Every token is represented as a $d_e$-dimensional real-valued vector. As presented above, this vector is the sum of the token embedding, positional embedding, and segment embedding, all of which are also $d_e$-dimensional real-valued vectors.
\item \vspace{0.3em} \textbf{Hidden Size} ($d$). The input and output of a sub-layer are of $d$ dimensions. Besides, most of the hidden states of a sub-layer are $d$-dimensional vectors. In general, $d$ can be roughly viewed as the width of the network.
\item \vspace{0.3em} \textbf{Number of Heads} ($n_{\mathrm{head}}$). In self-attention sub-layers, one needs to specify the number of heads used in multi-head self-attention. A larger number of heads allows the model to jointly attend to information from a greater number of representation subspaces. In practical systems, we often set $n_{\mathrm{head}} \ge 4$.
\item \vspace{0.3em} \textbf{FFN Hidden Size} ($d_{\textrm{ffn}}$). The size of the hidden layer of the FFNs used in Transformers is typically larger than $d$. For example, a typical setting is $d_{\textrm{ffn}} = 4 d$. For larger Transformers, such as recent large models, $d_{\textrm{ffn}}$ may be set to a very large value.
\item \vspace{0.3em} \textbf{Model Depth} ($L$). Using deep networks is an effective way to improve the expressive power of Transformers. For BERT models, $L$ is typically set to $12$ or $24$. However, networks with even greater depth are also feasible and can be applied for further enhancements.
\end{itemize}
\vspace{0.5em}

Different settings of these hyper-parameters lead to different model sizes. Two widely used configurations are BERT$_{\mathrm{base}}$ and BERT$_{\mathrm{large}}$:

\begin{itemize}
\item \vspace{0.5em} BERT$_{\mathrm{base}}$: $d=768$, $L=12$, $n_{\mathrm{head}}=12$, total number of parameters $=110$M.
\item \vspace{0.3em} BERT$_{\mathrm{large}}$: $d=1,024$, $L=24$, $n_{\mathrm{head}}=16$, total number of parameters $=340$M.
\end{itemize}
\vspace{0.3em}

The training of BERT models follows the standard procedure for Transformers. Training larger models such as BERT$_{\mathrm{large}}$ requires more training effort and time. This is a common problem for pre-training, especially when a model is trained on a very large amount of data. In practice, training efficiency is often a major concern. A common practice is to first train a BERT model on relatively short sequences for a large number of training steps, and then continue training it on full-length sequences for the remaining training steps.

\subsection{More Training and Larger Models}

\noindent BERT is a milestone in NLP and has sparked extensive subsequent work. One direction is to scale up pre-training, which generally involves two approaches: increasing the amount of training data and increasing the number of model parameters.

\mindex{RoBERTa}, an extension of the standard BERT model, is an example of such efforts~\cite{liu-etal:2019roberta}. It introduces two major improvements. First, simply using more training data and more compute can improve BERT models without the need to change the model architecture. Second, removing the NSP loss does not degrade performance on downstream tasks if the training is scaled up. These findings suggest exploring a general direction of pre-training: we can continue to improve pre-training by scaling it up on simple pre-training tasks.

A second approach to improving BERT models is to increase the number of model parameters. For example, in \citet{he-etal:2021deberta}'s work, a 1.5 billion-parameter BERT-like model is built by increasing both the model depth and hidden size. However, scaling up BERT and various other pre-trained models introduces new challenges in training, for example, training very large models often becomes unstable and difficult to converge. This makes the problem more complicated, and requires careful consideration of various aspects, such as model architecture, parallel computation, and parameter initialization. In another example, \citet{shoeybi-etal:2019megatron} successfully trained a 3.9 billion-parameter BERT-like model, where hundreds of GPUs were used to manage the increased computational demands.

\subsection{More Efficient Models}

\noindent Compared to its predecessors, BERT is a relatively large model for the time it was proposed. This increase in model size results in larger memory requirements and slower inference. Developing smaller and faster BERT models is part of the broader challenge of building efficient Transformers, which has been extensively discussed in the literature \cite{Tay-etal:2020efficient,xiao-and-zhu:2023introduction}. However, a deeper exploration of this topic is beyond the scope of this section. Here, we instead focus on a few efficient variants of BERT.

Several research threads are of particular interest for developing efficient BERT models. First, work on knowledge distillation, such as training a student model using the outputs of a well-trained teacher model, demonstrates that more compact models can be obtained by transferring knowledge from larger ones. Given that BERT models are multi-layer networks with several different types of layers, knowledge distillation can be applied at different levels of representation. For example, beyond distilling knowledge from the output layers, it is also possible to incorporate a training loss that measures the difference between the outputs of hidden layers for the teacher and student models~\cite{sun2020-etal:mobilebert,jiao-etal:2020tinybert}. Indeed, knowledge distillation has been one of the most widely used techniques for learning small pre-trained models.

Second, conventional model compression methods can be directly applied to BERT. One common approach is to apply general-purpose pruning methods to Transformer encoders~\cite{gale-etal:2019state}. This generally involves removing entire layers~\cite{fan-etal:2019reducing} or a certain percentage of parameters in the networks~\cite{sanh-etal:2020movement,chen-etal:2020lottery}. Pruning is also applicable to multi-head attention models. For example, \citet{michel-etal:2019sixteen} show that removing some of the heads does not significantly degrade performance of BERT models, but speeds up the inference of these models. Another approach to compressing BERT models is quantization~\cite{shen-etal:2020qbert}. By representing model parameters as low-precision numbers, we can compress the models greatly. While this method is not specific to BERT models, it proves effective for large Transformer-based architectures.

Third, considering that BERT models are relatively deep and large networks, another thread of research uses dynamic networks to adapt these models for efficient inference. An idea in this paradigm is to dynamically choose the layers for processing a token, for example, in depth-adaptive models we exit at some optimal depth and thus skip the rest of the layers in the layer stack~\cite{xin-etal:2020deebert,zhou-etal:2020bert}. Similarly, we can develop length-adaptive models in which the length of the input sequence is dynamically adjusted. For example, we can skip some of the tokens in the input sequence so that the model can reduce computational load on less important tokens, enhancing overall efficiency.

Fourth, it is also possible to share parameters across layers to reduce the size of BERT models. A simple way to do this is to share the parameters of a whole Transformer layer across the layer stack~\cite{dehghani-etal:2018universal,lan-etal:2020albert}. In addition to the reduced number of parameters, this enables reuse of the same layer in a multi-layer Transformer network, leading to savings of memory footprint at test time.

\subsection{Multilingual Models}

\noindent The initial BERT model was primarily focused on English. Soon after this model was proposed, it was extended to many languages. One simple way to do this is to develop a separate model for each language. Another approach, which has become more popular in recent work on large language models, is to train multilingual models directly on data from all the languages. Accordingly, \mindex{multilingual BERT} (\mindexraw{mBERT}) models were developed by training on text from 104 languages \footnote{\url{https://github.com/google-research/bert/}}. The primary difference from monolingual BERT models is that mBERT models use larger vocabularies to cover tokens from multiple languages. As a result, the representations of tokens from different languages are mapped into the same space, allowing for the sharing of knowledge across languages.

One of the most important applications of multilingual pre-trained models is cross-lingual transfer learning. In a zero-shot cross-lingual setting, a model is fine-tuned on a task using data from a source language and then applied to the same task in a target language without any target-language training data. In cross-lingual text classification, for example, we can fine-tune a multilingual pre-trained model on annotated English documents and then directly use the fine-tuned model to classify Chinese documents.

An improvement to multilingual pre-trained models like mBERT is to introduce bilingual data into pre-training. Rather than training solely on monolingual data from multiple languages, bilingual training explicitly models the relationship between tokens in two languages. The resulting model will have cross-lingual transfer abilities, and thus can be easily adapted to different languages. \citet{lample-conneau:2019cross} propose an approach to pre-training \mindex{cross-lingual language models} (\mindex{XLMs}). In their work, a cross-lingual language model can be trained using either causal language modeling or masked language modeling. For masked language modeling pre-training, the model is treated as an encoder. The training objective is the same as BERT: we maximize the probabilities of some randomly selected tokens which are either masked, replaced with random tokens, or kept unchanged in the input. If we consider bilingual data in pre-training, we sample a pair of aligned sentences each time. Then, the two sentences are packed together to form a single sequence used for training. For example, consider an English-Chinese sentence pair

\vspace{0.5em}
\begin{center}
\begin{CJK*}{UTF8}{gbsn}鲸鱼\ \ 是\ \ 哺乳\ \ 动物\ \ 。\end{CJK*}\ \ \ $\leftrightarrow$\ \ \ Whales are mammals\ \ .
\end{center}
\vspace{0.5em}

\noindent We can pack them to obtain a sequence, like this

\vspace{0.5em}
\begin{center}
$\mathrm{[CLS]}$ \begin{CJK*}{UTF8}{gbsn}鲸鱼\ \ 是\ \ 哺乳\ \ 动物\ \ 。\end{CJK*}\ \ $\mathrm{[SEP]}$ \ \ Whales are mammals\ \ .\ \ $\mathrm{[SEP]}$
\end{center}
\vspace{0.5em}

\noindent We then select a certain percentage of the tokens and replace them with $\mathrm{[MASK]}$.

\vspace{0.5em}
\begin{center}
$\mathrm{[CLS]}$ \begin{CJK*}{UTF8}{gbsn}$\mathrm{[MASK]}$\ \ 是\ \ $\mathrm{[MASK]}$\ \ 动物\ \ 。\end{CJK*}\ \ $\mathrm{[SEP]}$ \ \ Whales $\mathrm{[MASK]}$ $\mathrm{[MASK]}$\ \ .\ \ $\mathrm{[SEP]}$
\end{center}
\vspace{0.5em}

\noindent The goal of pre-training is to maximize the product of the probabilities of the masked tokens given the above sequence. By performing training in this way, the model can learn to represent both the English and Chinese sequences, as well as to capture the correspondences between tokens in the two languages. For example, predicting the Chinese token \begin{CJK*}{UTF8}{gbsn} 鲸鱼\end{CJK*} may require the information from the English token \textit{Whales}. Aligning the representations of the two languages essentially transforms the model into a ``translation'' model. This training objective is therefore also called \mindex{translation language modeling}. Figure~\ref{fig:translation-language-modeling} shows an illustration of this approach.

\begin{figure}[!t]
\centering
\input{./Chapter7/Figures/figure-xlm-example}
\caption{An illustration of translation language modeling (TLM). For ease of understanding, we present a simplified example where all selected tokens are masked. The model is trained to predict these masked tokens. Because the input sequence contains parallel tokens from two languages, predicting a masked token in one language allows the model to attend to context in the other language, thereby explicitly forcing cross-lingual alignment. In the original XLM formulation by \citet{lample-conneau:2019cross}, the input embedding (i.e., $\mathbf{e}_i$) is the sum of the token embedding, positional embedding, and a specific language embedding. This requires that each token be assigned a language label, allowing the model to distinguish between languages. However, in modern multilingual pre-training architectures (like XLM-R), explicitly specifying the language is  unnecessary, as a shared vocabulary is sufficient. Furthermore, relying on language embeddings makes it difficult to handle code-switching. Therefore, we assume here that all token representations are language-agnostic.}
\label{fig:translation-language-modeling}
\end{figure}

A benefit of multilingual pre-trained models is their inherent capability of handling code-switching. In NLP and linguistics, code-switching refers to switching among languages in a text. For example, the following is a mixed language text containing both Chinese and English:

\vspace{0.5em}
\begin{center}
\begin{CJK*}{UTF8}{gbsn}周末\ \ 我们\ \ 打算\ \ 去\ \ 做\ \ hiking\ \ ，\ \ 你\ \ 想\ \ 一起\ \ 来\ \ 吗\ \ ？\end{CJK*}

\small{(We plan to go hiking this weekend, would you like to join us?)}
\end{center}
\vspace{0.5em}

\noindent For multilingual pre-trained models, we do not need to identify whether a token is Chinese or English. Instead, every token is just an entry of the shared vocabulary. This can be imagined as creating a ``new'' language that encompasses all the languages we want to process.

The result of multilingual pre-training is influenced by several factors. Given that the model architecture is fixed, one needs to specify many configurations, such as the size of the shared vocabulary, the number (or percentage) of samples in each language, and the size of the model. \citet{conneau-etal:2020unsupervised} point out several interesting issues regarding large-scale multilingual pre-training for XLM-like models. First, as the number of supported languages increases, a larger model is needed to handle these languages. Second, a larger shared vocabulary is helpful for modeling the increased diversity in languages. Third, low-resource languages significantly benefit from cross-lingual transfer, particularly when linguistically similar high-resource languages are included in the pre-training corpus.  However, a phenomenon known as \mindex{the curse of multilinguality} (or capacity interference) emerges when a model of fixed capacity is trained on too many languages: the addition of more languages eventually forces parameter competition, degrading the performance on high-resource languages. Therefore, in practical systems, there is a trade-off between the number of supported languages and the model capacity.

%----------------------------------------------------------------------------------------
% NEW SECTION
%----------------------------------------------------------------------------------------

\section{Applying BERT Models}
\label{sec:pre-training-application}

\noindent Once a BERT model is pre-trained, it can then be used to solve NLP problems. But BERT models are not immediately ready for downstream tasks. In general, additional fine-tuning is required to adapt them. As a first step, we need a predictor to align the model output with the task of interest. Let $\mathrm{BERT}_{\hat{\theta}}(\cdot)$ be a BERT model with pre-trained parameters $\hat{\theta}$, and $\mathrm{Predict}_{\omega}(\cdot)$ be a prediction network with parameters $\omega$. By integrating the prediction network with the output of the BERT model, we develop a model to tackle the downstream tasks. This model can be expressed as:
\begin{eqnarray}
\mathbf{y} & = & \mathrm{Predict}_{\omega}(\mathrm{BERT}_{\hat{\theta}}(\mathbf{x})),
\end{eqnarray}

\noindent where $\mathbf{x}$ is the input and $\mathbf{y}$ is the output that fits the problem. For example, in classification problems, the model outputs a probability distribution over labels.

Then, we collect a set of labeled samples $\mathcal{D}$, and fine-tune the model by
\begin{eqnarray}
(\tilde{\omega},\tilde{\theta}) & = & \argmin_{\omega,\hat{\theta}^{+}} \sum_{(\mathbf{x},\mathbf{y}_{\mathrm{gold}}) \in \mathcal{D}} \mathrm{Loss}(\mathbf{y}_{\omega,\hat{\theta}^{+}},\mathbf{y}_{\mathrm{gold}}),
\end{eqnarray}

\noindent where $(\mathbf{x},\mathbf{y}_{\mathrm{gold}})$ represents a tuple of an input and its corresponding output. The notation of this equation seems a bit complicated, but the training/tuning process is standard. We optimize the model by minimizing the loss over the training data. The outcome is the optimized parameters $\tilde{\omega}$ and $\tilde{\theta}$. The optimization starts with the pre-trained parameters $\hat{\theta}$. Here we use $\hat{\theta}^{+}$ to indicate that the parameters are initialized with $\hat{\theta}$, and use $\mathbf{y}_{\omega,\hat{\theta}^{+}}$ to denote the model output computed using the parameters $\omega$ and $\hat{\theta}^{+}$.

With the fine-tuned parameters $\tilde{\omega}$ and $\tilde{\theta}$, we can apply the model $\mathrm{Predict}_{\tilde{\omega}}(\mathrm{BERT}_{\tilde{\theta}}(\cdot))$ to new data of the same tasks for which the model was fine-tuned. The form of the downstream tasks determines the input and output formats of the model, as well as the architecture of the prediction network. In the following we list some tasks for which BERT models are well suited.

\begin{itemize}
\item \vspace{0.5em} \textbf{Classification} (Single Text). One of the most widely used applications of BERT models is text classification. In this task, a BERT model receives a sequence of tokens and encodes it as a sequence of vectors. The first output vector $\mathbf{h}_{\mathrm{cls}}$ (or $\mathbf{h}_{0}$) is typically used as the representation of the entire text. The prediction network takes $\mathbf{h}_{\mathrm{cls}}$ as input to produce a distribution of labels. Let $\mathrm{[CLS]}x_1 x_2 \cdots x_m$ be the input sequence. See Figure~\ref{fig:bert-app-classification-single-text} for an illustration of BERT-based text classification.

\begin{figure}[!ht]
\centering
\input{./Chapter7/Figures/figure-bert-app-classification-single-text}
\caption{BERT-based classification for a single text.Here the gray box denotes the prediction network. }
\label{fig:bert-app-classification-single-text}
\end{figure}

    \noindent Many NLP problems can be categorized as text classification tasks, and there have been several text classification benchmarks for evaluating pre-trained models. For example, we can classify texts by their grammatical correctness (grammaticality) or emotional tone (sentiment)~\cite{socher-etal:2013recursive,warstadt-etal:2019neural}. Note that the prediction network could be any classification model, such as a deep neural network or a more traditional classification model. The entire model can then be trained or fine-tuned in the manner of a standard classification model. For example, the prediction network can be simply a Softmax layer and the model parameters can be optimized by maximizing the probabilities of the correct labels.

\item \vspace{0.3em} \textbf{Classification} (Pair of Texts). Classification can also be performed on a pair of texts. Suppose we have two texts, $x_1 \cdots x_m$ and $y_1 \cdots y_n$. We can concatenate these texts to form a single sequence with length $\mathrm{len}$. Then, we predict a label for this combined text sequence based on the $\mathbf{h}_{\mathrm{cls}}$ vector, as shown in Figure~\ref{fig:bert-app-classification-two-texts}.

\begin{figure}[!ht]
\centering
\input{./Chapter7/Figures/figure-bert-app-classification-two-texts}
\caption{BERT-based classification for a pair of texts.Here, $\mathrm{len} = n + m + 2$.}
\label{fig:bert-app-classification-two-texts}
\end{figure}

    \noindent Text pair classification covers several problems, including semantic equivalence judgement (determine whether two texts are semantically equivalent)~\cite{dolan-brockett:2005automatically}, text entailment judgement (determine whether a hypothesis can be logically inferred or entailed from a premise)~\cite{bentivogli-giampiccolo:2011rte,williams-etal:2018broad}, grounded commonsense inference (determine whether an event is likely to happen given its context)~\cite{zellers-etal:2018swag}, and question-answering inference (determine whether an answer corresponds to a given question).

\item \vspace{0.3em} \textbf{Regression}. Instead of generating a label distribution, we can have the prediction network output a real-valued score. For example, by adding a Sigmoid layer to the prediction network, the system can be employed to compute the similarity between two given sentences. The architecture is the same as that of BERT-based classification systems, with only the output layer changed, as shown in Figure~\ref{fig:bert-app-regression-two-texts}.

\begin{figure}[!ht]
\centering
\input{./Chapter7/Figures/figure-bert-app-regression-two-texts}
\caption{BERT-based regression for a pair of texts.}
\label{fig:bert-app-regression-two-texts}
\end{figure}

    For training or fine-tuning, we can minimize the regression loss of the model output as usual.

\item \vspace{0.3em} \textbf{Sequence Labeling}. Sequence labeling is a machine learning approach applicable to a wide range of NLP problems. This approach assigns a label to each token in an input sequence, and some linguistic annotations can then be derived from this sequence of labels. An example of sequence labeling in NLP is part-of-speech (POS) tagging. We label each word in a sentence with its corresponding POS tag. Another example is named entity recognition (NER) in which we label each word with an NER tag, and named entities are identified using these tags. See Figure~\ref{fig:bert-app-sequence-labeling} for an illustration of the model architecture for NER.

\begin{figure}[!ht]
\centering
\input{./Chapter7/Figures/figure-bert-app-sequence-labeling}
\caption{BERT-based sequence labeling for named entity recognition.Here $\{\mathrm{B}, \mathrm{I}, \mathrm{O}\}$ is the tag set of NER.}
\label{fig:bert-app-sequence-labeling}
\end{figure}

    \noindent  In this model, we use the BIO tagging scheme to represent named entities. For example, $\mathrm{B}\textrm{-}\mathrm{ORG}$ means the beginning of an organization, $\mathrm{I}\textrm{-}\mathrm{ORG}$ means the word is inside an organization, and $\mathrm{O}$ means the word does not belong to any named entity. This NER model can output a distribution over the tag set at each position, denoted as $\mathbf{p}_i$. The training or fine-tuning of the model can be performed over these distributions $\{\mathbf{p}_1,\cdots,\mathbf{p}_m\}$. For example, suppose $p_i(\mathrm{tag}_i)$ is the probability of the correct tag at position $i$. The training loss can be defined to be the negative log-likelihood
    \begin{eqnarray}
    \mathrm{Loss} & = & -\frac{1}{m} \sum_{i=1}^{m} \log p_i(\mathrm{tag}_i).
    \end{eqnarray}

    Finding the best label sequence given a trained NER model is a well-studied problem in NLP. This is often achieved via dynamic programming, which, in the context of path finding over a lattice, has linear complexity~\cite{huang-etal:2009dynamic}.

\item \vspace{0.3em} \textbf{Span Prediction}. Some NLP tasks require predicting a span in a text. A common example is reading comprehension. In this task, we are given a query $x_1 \cdots x_m$ and a context text $y_1 \cdots y_n$. The goal is to identify a continuous span in $y_1 \cdots y_n$ that best answers the query. This problem can be framed as a sequence labeling-like task in which we predict a label for each $y_j$ to indicate the beginning or ending of the span. Following \citet{seo-etal:2017bidirectional}, we add two networks on top of the BERT output for $y_j$: one for generating the probability of $y_j$ being the beginning of the span (denoted by $p_j^{\mathrm{beg}}$), and one for generating the probability of $y_j$ being the ending of the span (denoted by $p_j^{\mathrm{end}}$). The resulting model architecture is shown in Figure~\ref{fig:bert-app-span-prediction}.

\begin{figure}[!ht]
\centering
\input{./Chapter7/Figures/figure-bert-app-span-prediction}
\caption{BERT-based span prediction for reading comprehension.}
\label{fig:bert-app-span-prediction}
\end{figure}

    We pack the query and context text together to obtain the input sequence. The prediction networks are only applied to outputs for the context text, generating the probabilities $p_j^{\mathrm{beg}}$ and $p_j^{\mathrm{end}}$ at each position. The loss can be computed by summing the negative log likelihoods of the two models across the entire context text.
    \begin{eqnarray}
    \mathrm{Loss} & = & - \frac{1}{n} \sum_{j=1}^{n} \big( \log p_j^{\mathrm{beg}} + \log p_j^{\mathrm{end}} \big).
    \end{eqnarray}

    At test time, we search for the best span by
    \begin{eqnarray}
    (\hat{j}_1,\hat{j}_2) & = & \argmax_{1 \le j_1 \le j_2 \le n}\ \big( \log p_{j_1}^{\mathrm{beg}} + \log p_{j_2}^{\mathrm{end}} \big).
    \end{eqnarray}

\item \vspace{0.3em} \textbf{Encoding for Encoder-Decoder Models}. While our focus in this section has been primarily on language understanding problems, it is worth noting that BERT models can be applied to a broader range of NLP tasks. In fact, BERT models can be used in all the scenarios where we need to encode a piece of text. One application that we have not mentioned is text generation which includes a range of tasks such as machine translation, summarization, question answering, and dialogue generation. These tasks can be formulated as sequence-to-sequence problems: we use an encoder to represent the source text, and a decoder to generate the corresponding target text. A straightforward method to apply BERT models is to consider them as encoders. Before fine-tuning, we can initialize the parameters of the encoder with those from a pre-trained BERT model. Then, the encoder-decoder model can be fine-tuned on pairs of texts as usual. Figure~\ref{fig:bert-app-encoder-decoder} shows the architecture of a neural machine translation system where a BERT model is applied on the source side.

\begin{figure}[!ht]
\centering
\input{./Chapter7/Figures/figure-bert-app-encoder-decoder}
\caption{Using BERT as the encoder in an encoder-decoder model.Here $x_1 \cdots x_m$ denotes the source sequence, $y_1 \cdots y_n$ denotes the target sequence, $\mathbf{e}_{1}^{x} \cdots \mathbf{e}_{m}^{x}$ denotes the embedding sequence of $x_1 \cdots x_m$, and $\mathbf{e}_{1}^{y} \cdots \mathbf{e}_{n}^{y}$ denotes the embedding sequence of $y_1 \cdots y_n$. The adapter, which is optional, maps the output of the BERT model to the form that is better suited to the decoder.}
\label{fig:bert-app-encoder-decoder}
\end{figure}

\end{itemize}
\vspace{0.5em}

Fine-tuning BERT models is a complicated engineering problem, influenced by many factors, such as the amount of fine-tuning data, the model size, and the optimizer used in fine-tuning. In general, we wish to fine-tune these models sufficiently so that they can perform well in the downstream tasks. However, fine-tuning BERT models for specific tasks may lead to overfitting, which in turn reduces their ability to generalize to other tasks. For example, suppose we have a BERT model that performs well on a particular task. If we then fine-tune it for new tasks, this may decrease its performance on the original task. This problem is related to the \mindex{catastrophic forgetting} problem in continual training, where a neural network forgets previously learned information when updated on new samples. In practical applications, a common way to alleviate catastrophic forgetting is to add some old data into fine-tuning and train the model with more diverse data. Also, one may use methods specialized to catastrophic forgetting, such as experience replay~\cite{rolnick-etal:2019experience} and elastic weight consolidation~\cite{kirkpatrick-etal:2017overcoming}. The interested reader can refer to some surveys for more detailed discussions of this issue in continual learning~\cite{parisi-etal:2019continual,wang-etal:2023comprehensive,wang-etal:2023comprehensive2}.

%----------------------------------------------------------------------------------------
% NEW SECTION
%----------------------------------------------------------------------------------------

\section{Summary}

\noindent In this chapter we have introduced the general concept of pre-training in NLP. In particular, we have discussed self-supervised pre-training and its application to encoder-only, decoder-only, and encoder-decoder architectures. Moreover, we have presented and compared a variety of pre-training tasks for these architectures. As an example, we used BERT to illustrate how sequence models are pre-trained via masked language modeling and applied to different downstream tasks.

Recent years have shown remarkable progress in NLP, led by the large-scale use of self-supervised pre-training. Significant advances have been made across many tasks, not only in NLP but also in computer vision and other areas of AI. One idea behind these advances is that a significant amount of knowledge about the world can be learned by simply training these AI systems on huge amounts of unlabeled data. For example, a language model can learn some general knowledge of a language by repeatedly predicting masked words in large-scale text. As a result, this pre-trained language model can serve as a foundation model, which can be easily adapted to address specific downstream NLP tasks. This paradigm shift in NLP has enabled the development of powerful systems for language understanding, generation, and reasoning~\cite{manning:2022human}. However, it is important to recognize that we are still in the early stages of creating truly intelligent systems, and there is a long way to go. Nevertheless, large-scale pre-training has opened a door to intelligent systems that researchers have long aspired to develop, though several key research areas remain open for exploration, such as learning efficiently from limited data and acquiring complex reasoning and planning abilities.

Note that this chapter is mostly introductory and cannot cover all aspects of pre-training. For example, there are many methods to fine-tune a pre-trained model, offering different ways to adapt the model to diverse scenarios.  Moreover, large language models, which are considered one of the most significant achievements in AI in recent years, are not covered in this chapter. We leave the discussion of these topics to the following chapters.

% !Mode:: "TeX:UTF-8"
% !TEX encoding = UTF-8 Unicode

%----------------------------------------------------------------------------------------
% Natural Language Processing: Neural Networks and Large Language Models
%
% Copyright 2021-2026
% Tong Xiao (xiaotong@mail.neu.edu.cn) and Jingbo Zhu (zhujingbo@mail.neu.edu.cn)
%----------------------------------------------------------------------------------------

%----------------------------------------------------------------------------------------
%    CONFIGURATIONS
%----------------------------------------------------------------------------------------

%\setcounter{chapter}{7}

%----------------------------------------------------------------------------------------
%	CHAPTER 8
%----------------------------------------------------------------------------------------

\chapter{Generative Models}

\thispagestyle{chapterurl} % add urls to the header

\noindent One of the most significant advances in NLP in recent years is the development of large language models (LLMs). This breakthrough has facilitated the creation of systems that can understand and generate natural language at a near-human level. These systems have even demonstrated reasoning capabilities, a long-standing challenge in AI. With these achievements, NLP has made substantial progress and entered a new era of research in which increasingly complex problems are being addressed, such as building conversational agents that can interact with humans fluently.

The concept of language modeling dates back to early experiments conducted by~\citet{shannon:1951prediction}. In his work, Shannon proposed a language model to estimate how predictable English text is --- \textit{how well the next letter of a text can be predicted when the preceding $N$ letters are known}. Although Shannon's experiments were preliminary, the fundamental goals and methods of language modeling have remained largely unchanged over the decades. For many years, particularly before 2010, the dominant approach to language modeling was the $n$-gram approach~\cite{Jurafsky-and-Martin:2008}. In $n$-gram language modeling, we estimate the probability of a word given its preceding $n-1$ words, and thus the probability of a sequence can be approximated by the product of $n$-gram probabilities. These probabilities are typically estimated from smoothed relative counts of $n$-grams in text. Despite its simplicity, it has been widely used in NLP. For example, the success of modern statistical speech recognition and machine translation systems has largely depended on the use of $n$-gram language models~\cite{jelinek:1998statistical,Koehn:2010}.

Applying neural networks to language modeling has long been an active area of research, but a major breakthrough came with the advancement of deep learning techniques. A widely cited study is the work of \citet{Bengio-et-al:2003}, in which $n$-gram probabilities are modeled using a feed-forward network trained end-to-end. An important outcome of this model is the distributed representations of words, known as word embeddings. Rather than representing words as discrete variables, word embeddings map words to low-dimensional real-valued vectors, making it possible to capture semantic relationships among words and word $n$-grams in a continuous representation space. As a result, language models are no longer burdened by the curse of dimensionality, but can represent an exponential number of $n$-grams using a compact neural model.

The idea of learning word representations through neural language models inspired subsequent research on representation learning in NLP. However, this approach did not attract significant interest from the NLP community in the first few years after its proposal. Around 2012, however, significant progress was made in learning word embeddings from large-scale text via simple word prediction tasks. Several methods, including Word2Vec, were proposed to effectively learn such embeddings, which were then successfully applied in a variety of NLP systems~\cite{mikolov:2013efficient,Mikolov-et-al:2013distributed}. As a result of these advances, researchers began exploring sequence representation learning using more powerful language models, such as LSTM-based models~\cite{sutskever-etal:2014sequence,peters-etal:2018deep}. Subsequently, progress and interest in sequence representation grew rapidly after the Transformer architecture was proposed. With the rise of the Transformer architecture, the concept of language modeling was generalized to encompass models that learn to predict words under different paradigms. Many powerful Transformer-based models were pre-trained using these word prediction tasks and were successfully applied to a variety of downstream tasks~\cite{devlin-etal:2019bert}.

Indeed, training language models on large-scale data has led NLP research to exciting times. While language modeling has long been seen as a foundational technique with no direct link to the broader goals of artificial intelligence, it has led to the emergence of intelligent systems that can acquire general knowledge through repeated word prediction. Recent research demonstrates that a single, well-trained LLM can perform a wide range of tasks and generalize to new tasks with minimal adaptation~\cite{bubeck-etal:2023sparks}. This represents a step toward more advanced forms of artificial intelligence and inspires further exploration into developing more powerful language models as a class of foundation models.

In this chapter, we discuss the basic concepts of generative LLMs. For simplicity, we use LLMs to refer to generative models such as GPT, although in a broader sense the term may also include non-generative models such as BERT. We begin with a general introduction to LLMs, including the key steps involved in building such models. We then discuss two key scaling challenges for LLMs: how LLMs are trained at scale and how they can be improved to handle long-context inputs. Finally, we conclude with a brief summary.

%----------------------------------------------------------------------------------------
% NEW SECTION
%----------------------------------------------------------------------------------------

\section{A Brief Introduction to LLMs}

In this section, we introduce the basic concepts of LLMs required for the remainder of this chapter and subsequent chapters. We will use the terms \textit{word} and \textit{token} interchangeably. Both refer to the basic units used in language modeling, though their original meanings differ slightly.

Before presenting the technical details, let us first consider how language models work. The core objective of language modeling is to estimate the probability of a token sequence. Let $\{x_0, x_1, \cdots, x_m\}$ be a sequence of tokens, where $x_0$ is the start symbol $\langle s \rangle$ (or $\langle \mathrm{SOS} \rangle$)\footnote{In some models such as BERT, a special token like [CLS] is prepended, although it serves a different purpose from the start symbol in autoregressive models.}. The probability of this sequence can be factorized using the chain rule
\begin{eqnarray}
\Pr(x_0,\cdots,x_m) & = & \Pr(x_0) \cdot \Pr(x_1|x_0) \cdot \Pr(x_2|x_0,x_1) \cdots \Pr(x_m|x_0,\cdots,x_{m-1}) \nonumber \\
                 & = & \prod_{i=0}^{m} \Pr(x_{i}|x_0,\cdots,x_{i-1}), \label{eq:lm-basic-form},
\end{eqnarray}

\noindent or alternatively in a logarithmic form

\begin{eqnarray}
\log \Pr(x_0,\cdots,x_m) & = & \sum_{i=0}^{m} \log \Pr(x_{i}|x_0,\cdots,x_{i-1}). \label{eq:lm-basic-form-log}.
\end{eqnarray}

\noindent Here, $\Pr(x_{i}|x_0,\cdots,x_{i-1})$ is the probability of the token $x_i$ given all its preceding tokens $\{x_0,\cdots,\allowbreak x_{i-1}\}$ \footnote{We assume that when $i=0$, $\Pr(x_{i}|x_0,\cdots,x_{i-1})=\Pr(x_0)=1$. Hence, $\Pr(x_0,\cdots,x_m) = \Pr(x_0)\Pr(x_1,\cdots,x_m|x_0)=\Pr(x_1,\cdots,x_m|x_0)$.}. In the era of deep learning, a typical approach to language modeling is to estimate these probabilities using deep neural networks. Neural networks trained for this task receive a context sequence $x_0,\cdots,x_{i-1}$ and output a probability distribution over the vocabulary $V$ (denoted by $\Pr(\cdot|x_0,\cdots,x_{i-1})$). The probability $\Pr(x_{i}|x_0,\cdots,x_{i-1})$ corresponds to the entry associated with token $x_i$ in this distribution.

When applying a trained language model, a common task is to find the most likely next token given its context. This can be formulated as
\begin{eqnarray}
\hat{x}_i & = & \argmax_{x_i \in V} \Pr(x_{i}|x_0,\cdots,x_{i-1}).
\end{eqnarray}

We can perform this token prediction iteratively to generate text sequentially: at each step, we predict the most likely token $\hat{x}_i$, and then append it to the context to predict the subsequent token $\hat{x}_{i+1}$. This results in an autoregressive, left-to-right generation process that implements Eqs. (\ref{eq:lm-basic-form}) and (\ref{eq:lm-basic-form-log}). To illustrate, consider the generation of three tokens given the prefix ``$\langle s \rangle\ \ a$'', as shown in Table~\ref{tab:lm-generation-abcd}. Next, we will discuss how LLMs are constructed, trained, and applied in practice.

\begin{table}[t]
\centering
\begingroup
\renewcommand{\arraystretch}{1.2}
\begin{tabular}{l | l | l | l}
Context & Predict & Decision Rule & Sequence Probability \\ \hline
$\langle s \rangle\ \ a$ & $b$ & $\argmax_{x_2 \in V} \Pr(x_{2}|\langle s \rangle\ a)$ & $\Pr(\langle s \rangle) \cdot \Pr(a|\langle s \rangle) \cdot $ \colorbox{gray!30}{$\Pr(b|\langle s \rangle\ a)$} \\ \hline
$\langle s \rangle\ \ a\ \ b$ & $c$ & $\argmax_{x_3 \in V} \Pr(x_{3}|\langle s \rangle\ a\ b)$ & $\Pr(\langle s \rangle) \cdot \Pr(a|\langle s \rangle) \cdot \Pr(b|\langle s \rangle\ a) \cdot$ \\
& & & \colorbox{gray!30}{$\Pr(c|\langle s \rangle\ a\ b)$} \\ \hline
$\langle s \rangle\ \ a\ \ b\ \ c$ & $d$ & $\argmax_{x_4 \in V} \Pr(x_{4}|\langle s \rangle\ a\ b\ c)$ & $\Pr(\langle s \rangle) \cdot \Pr(a|\langle s \rangle) \cdot \Pr(b|\langle s \rangle\ a) \cdot$ \\
& & & $\Pr(c|\langle s \rangle\ a\ b) \cdot $ \colorbox{gray!30}{$\Pr(d|\langle s \rangle\ a\ b\ c)$} \\
\end{tabular}
\endgroup
\caption{Illustration of generating the three tokens $b\ \ c\ \ d$ given the prefix $\langle s \rangle\ \ a$ via a language model. In each step, the model picks a token $x_i$ from $V$ so that $\Pr(x_{i}|x_0,\cdots,x_{i-1})$ is maximized. This token is then appended to the end of the context sequence. In the next step, we repeat the same process, but based on the new context.}
\label{tab:lm-generation-abcd}
\end{table}

\subsection{Decoder-only Transformers}
\label{sec:decoder-only-transformers}

\noindent As is standard practice, the input to a language model is a sequence of tokens (denoted by $\{x_0,\cdots,x_{m-1}\}$). At each step, a new token is generated and appended to the sequence for the next prediction. To do this, the language model outputs a probability distribution $\Pr(\cdot|x_0,\cdots,x_{i-1})$ at each position $i$, and the next token is generated according to this distribution (e.g., by sampling or greedy decoding). The model is trained by maximizing the log-likelihood $\sum_{i=1}^{m} \log \Pr(x_i|x_0,\allowbreak \cdots,x_{i-1})$\footnote{Note that $\sum_{i=1}^{m} \log \Pr(x_i|x_0,\cdots,x_{i-1}) = \sum_{i=0}^{m} \log \Pr(x_i | x_0,\cdots,x_{i-1})$ since $\log \Pr(x_0) = 0$.}.

Here, we focus on the decoder-only Transformer architecture, as it is one of the most widely adopted model architectures for LLMs. The input sequence of tokens is represented by a sequence of $d_e$-dimensional vectors $\{\mathbf{e}_0,\cdots,\mathbf{e}_{m-1}\}$. Each vector $\mathbf{e}_i$ is the sum of the token embedding for $x_i$ and the positional embedding for position $i$. The core of the model consists of a stack of Transformer blocks (or layers). Each Transformer block consists of two sub-layers: a self-attention sub-layer and an FFN sub-layer. These sub-layers can be defined using the post-norm architecture
\begin{eqnarray}
\mathrm{output} & = & \mathrm{LNorm}(F(\mathrm{input}) + \mathrm{input}) \label{eq:post-norm-lm},
\end{eqnarray}

\noindent or the pre-norm architecture
\begin{eqnarray}
\mathrm{output} & = & F(\mathrm{LNorm}(\mathrm{input})) + \mathrm{input} \label{eq:pre-norm-lm},
\end{eqnarray}

\noindent where $\mathrm{input}$ and $\mathrm{output}$ denote the input and output, both being $m \times d$ matrices. The $i$-th rows of both $\mathrm{input}$ and $\mathrm{output}$ can be interpreted as the contextual representations of the $i$-th token in the sequence.

$F(\cdot)$ is the core function of a sub-layer. For FFN sub-layers, $F(\cdot)$ is a 2-layer FFN. For self-attention sub-layers, $F(\cdot)$ is a multi-head self-attention function. In general, self-attention is expressed in the form of QKV attention
\begin{eqnarray}
\mathrm{Att}_{\mathrm{qkv}}(\mathbf{Q},\mathbf{K},\mathbf{V}) & = & \mathrm{Softmax}(\frac{\mathbf{Q} \mathbf{K}^\top}{\sqrt{d}} + \mathbf{Mask}) \mathbf{V} \label{eq:qkv-attention-lm},
\end{eqnarray}

\noindent where $\mathbf{Q}$, $\mathbf{K}$ and $\mathbf{V} \in \mathbb{R}^{m \times d}$ are the queries, keys, and values, respectively. It is important to note that only previous tokens are considered when predicting a token. So a masking variable $\mathbf{Mask} \in \mathbb{R}^{m \times m}$ is incorporated into self-attention to achieve this. The entry $(i,k)$ of $\mathbf{Mask}$ has a value of 0 if $i \ge k$, and a value of $-\infty$ otherwise.

Given a representation $\mathbf{H} \in \mathbb{R}^{m \times d}$, the multi-head self-attention function can be defined as
\begin{eqnarray}
F(\mathbf{H}) & = & \mathrm{Merge}(\mathrm{head}_1,\cdots,\mathrm{head}_{\tau}) \mathbf{W}^{\mathrm{head}},
\end{eqnarray}

\noindent where $\mathrm{Merge}(\cdot)$ represents a concatenation of its inputs, and $\mathbf{W}^{\mathrm{head}} \in \mathbb{R}^{d \times d}$ represents a parameter matrix. $\mathrm{head}_j$ is the output of QKV attention on a subspace of the representation.
\begin{eqnarray}
\mathrm{head}_j & = & \mathrm{Att}_{\mathrm{qkv}}(\mathbf{Q}^{[j]},\mathbf{K}^{[j]},\mathbf{V}^{[j]}).
\end{eqnarray}

\noindent Here, $\mathbf{Q}^{[j]}$, $\mathbf{K}^{[j]}$, and $\mathbf{V}^{[j]}$ are the queries, keys, and values projected onto the $j$-th subspace via linear transformations
\begin{eqnarray}
\mathbf{Q}^{[j]} & = & \mathbf{H} \mathbf{W}_j^{q} \\
\mathbf{K}^{[j]} & = & \mathbf{H} \mathbf{W}_j^{k} \\
\mathbf{V}^{[j]} & = & \mathbf{H} \mathbf{W}_j^{v},
\end{eqnarray}

\noindent where $\mathbf{W}_j^{q}$, $\mathbf{W}_j^{k}$, and $\mathbf{W}_j^{v} \in \mathbb{R}^{d \times \frac{d}{\tau}}$ are the parameter matrices of the transformations.

\begin{figure}[!t]
\centering
\input{./Chapter8/Figures/figure-lm-decoder-only-architecture}
\caption{The Transformer-decoder architecture for language modeling. The central components are $L$ stacked Transformer blocks, each comprising a self-attention sub-layer and an FFN sub-layer. To prevent the model from accessing the right-context, a masking variable is incorporated into self-attention. The output layer uses a Softmax function to generate a probability distribution for the next token, given the sequence of previous tokens. During inference, the model takes the previously predicted token to predict the next one, repeating this process until the end of the sequence is reached. $\{\mathbf{z}_0,\cdots,\mathbf{z}_{m-1}\}$ denote the inputs of a Transformer block, and $\{\mathbf{h}_0^{L},\cdots,\mathbf{h}_{m-1}^{L}\}$ denote the outputs of the last Transformer block.}
\label{fig:llm-transformer-decoder}
\end{figure}

Suppose we have $L$ Transformer blocks. A Softmax layer is built on top of the output of the last block. The Softmax layer outputs a sequence of $m$ distributions over the vocabulary, like this
\begin{eqnarray}
\begin{bmatrix}
\Pr(\cdot | x_0) \\
\Pr(\cdot | x_0,x_{1}) \\
\vdots \\
\Pr(\cdot | x_0,\cdots,x_{m-1})
\end{bmatrix}
& = & \mathrm{Softmax}(\mathbf{H}^{L} \mathbf{W}^{o}),
\end{eqnarray}

\noindent where $\mathbf{H}^{L}$ is the output of the last Transformer block, and $\mathbf{W}^{o} \in \mathbb{R}^{d \times |V|}$ is the parameter matrix.

Figure~\ref{fig:llm-transformer-decoder} shows the Transformer architecture for language modeling. Inference with this model follows an autoregressive process. Each time the language model takes the prefix $x_0,\cdots,x_{i-1}$ as input and predicts a token $x_{i}$ that maximizes the probability $\Pr(x_i|x_0,\cdots,x_{i-1})$. It is important to note that, despite different implementation details, many LLMs share the same architecture described above. These models are called large because both their depth and width are significant. Table~\ref{tab:llm-model-size-comparison} shows the model sizes for a few LLMs, as well as their model configurations.

\begin{table}[t]
\centering
\begingroup
\renewcommand{\arraystretch}{1.2}
\begin{tabular}{l | r | r | r | r }
LLM & \# of Parameters & Depth $L$ & Width $d$ & \# of Heads \\
    &                  &           &           &   (Q/KV)    \\ \hline
GPT-1~\cite{radford-etal:2018improving} & 0.117B & 12 & 768 & 12/12 \\
GPT-2~\cite{radford-etal:2019language} & 1.5B & 48 & 1,600 & 25/25 \\
GPT-3~\cite{brown-etal:2020language} & 175B & 96 & 12,288 & 96/96 \\ \hline
\multirow{3}{*}{Llama2~\cite{touvron-etal:2023llama2}}
                       &  7B & 32 & 4,096 & 32/32 \\
                       &  13B & 40 & 5,120 & 40/40 \\
                       &  70B & 80 & 8,192 & 64/64 \\ \hline
\multirow{3}{*}{Llama3/3.1~\cite{dubey2024llama}}
                       &  8B & 32 & 4,096 & 32/8 \\
                       &  70B & 80 & 8,192 & 64/8 \\
                       &  405B & 126 & 16,384 & 128/8 \\ \hline
\multirow{3}{*}{Gemma2~\cite{team2024gemma}}
                       &  2B & 26 & 2,304 & 8/4 \\
                       &  9B & 42 & 3,584 & 16/8 \\
                       &  37B & 46 & 4,608 & 32/16 \\ \hline
\multirow{3}{*}{Qwen2.5~\cite{yang2024qwen2}}
                       &  0.5B & 24 & 896 & 14/2 \\
                       &  7B & 28 & 3,584 & 28/4 \\
                       &  72B & 80 & 8,192 & 64/8 \\ \hline
\multirow{1}{*}{DeepSeek-V3~\cite{liu2024deepseek}}
                       &  671B & 61 & 7,168 & 128/128 \\ \hline
\multirow{3}{*}{Falcon~\cite{penedo-etal:2023refinedweb}}
                       &  7B & 32 & 4,544 & 71/71 \\
                       &  40B & 60 & 8,192 & 128/128 \\
                       &  180B & 80 & 14,848 & 232/232 \\ \hline
Mistral~\cite{jiang-etal:2023mistral} & 7B & 32 & 4,096 & 32/32
\end{tabular}
\endgroup
\caption{Comparison of some LLMs in terms of model size, model depth, model width, and number of heads ($a/b$ means $a$ heads for queries and $b$ heads for both keys and values).}
\label{tab:llm-model-size-comparison}
\end{table}

\subsection{Training LLMs}

\noindent Now suppose we are given a training set $\mathcal{D}$ comprising $K$ sequences. The log-likelihood of each sequence $\mathbf{x} = x_0 \cdots x_m$ in $\mathcal{D}$ can be computed using a language model
\begin{eqnarray}
\mathcal{J}_{\theta}(\mathbf{x}) & = & \sum_{i=1}^{m} \log \mathrm{Pr}_{\theta}(x_i|x_0,\cdots,x_{i-1}) \label{eq:llm-training-loss-likelihood}.
\end{eqnarray}

\noindent where the subscript $\theta$ in $\mathcal{L}_{\theta}(\cdot)$ and $\Pr_{\theta}(\cdot)$ denotes the parameters of the language model. Then, the objective of maximum likelihood training is defined as
\begin{eqnarray}
\hat{\theta} & = & \argmax_{\theta} \sum_{\mathbf{x} \in \mathcal{D}} \mathcal{J}_{\theta}(\mathbf{x}) \label{eq:llm-training-mle}.
\end{eqnarray}

Training Transformer-based language models with the above objective is commonly viewed as a standard optimization process for neural networks. This can be achieved using gradient descent algorithms, which are widely supported by off-the-shelf deep learning toolkits. Somewhat surprisingly, model performance has continued to improve as language models evolved into more computationally intensive architectures and were trained on larger datasets~\cite{kaplan-etal:2020scaling}. These successes have motivated NLP researchers to continually scale up both training data and model size to build increasingly powerful language models.

However, as language models become larger, we confront new training challenges, which significantly change the problem compared to training relatively small models. One of these challenges arises from the need for large-scale distributed systems to manage the data, model parameters, and training routines. Developing and maintaining such systems requires substantial effort in both software and hardware engineering, alongside deep learning expertise. A related issue is that scaling up training demands massive computational resources to complete the process within a reasonable timeframe. For instance, training an LLM with tens of billions of parameters from scratch generally requires hundreds or thousands of GPUs. This requirement drastically increases training costs, especially considering the many experimental iterations needed during model development. Furthermore, from an optimization perspective, the training process can become highly unstable when scaling up network depth and parameter count. As a result, architectural modifications are often required to adapt LLMs for large-scale training. In Section~\ref{sec:llm-training-at-scale} we will discuss these issues in more detail.

\subsection{Fine-tuning and Applying LLMs}

\noindent Once we have pre-trained an LLM, we can apply it to perform various NLP tasks. Traditionally, language models were used as components of other systems. For example, they were widely applied to score translations in statistical machine translation systems. By contrast, in generative AI, LLMs are considered complete systems and are employed to address NLP problems by leveraging their generative nature. A common approach is to describe the target task in text and then prompt the LLM to generate text based on this description. This is a standard text generation task where the model continues or completes the text starting from a given context.

More formally, let $\mathbf{x} = x_0 \cdots x_m$ denote a sequence of tokens representing the context provided by users, and $\mathbf{y} = y_1 \cdots y_n$ denote a token sequence following the context. The inference process of an LLM can be defined as finding the most likely sequence $\mathbf{y}$ given $\mathbf{x}$:
\begin{eqnarray}
\hat{\mathbf{y}} & = & \argmax_{\mathbf{y}} \log \Pr(\mathbf{y} | \mathbf{x}) \nonumber \\
                 & = & \argmax_{\mathbf{y}} \sum_{i=1}^{n} \log \Pr(y_i|x_0,\cdots,x_m,y_1,\cdots,y_{i-1}).
\end{eqnarray}

\noindent Here, $\sum_{i=1}^{n} \log \Pr(y_i|x_0,\cdots,x_m,y_1,\cdots,y_{i-1})$ is mathematically equivalent to the right-hand side of Eq. (\ref{eq:lm-basic-form-log}). It models the log probability of predicting tokens starting from position $m+1$, rather than position $0$. Throughout this chapter and subsequent ones, we will employ separate variables $\mathbf{x}$ and $\mathbf{y}$ to distinguish the input and output of an LLM, though they can be viewed as sub-sequences belonging to the same continuous sequence. By adopting this notation, we can see that the above equation closely resembles the formulations used in other NLP text generation models, such as neural machine translation architectures.

To illustrate how LLMs are applied, consider the problem of determining the grammaticality of a given sentence. We can define a template like this
\vspace{0.5em}
\begin{tcolorbox}[frame empty]
\hspace{4em} \{*sentence*\}

\hspace{4em} Question: Is this sentence grammatically correct?

\hspace{4em} Answer: \underline{\hspace{2em}}
\end{tcolorbox}
\vspace{0.5em}

\noindent Here $\underline{\hspace{1em}}$ represents the text we intend to generate. \{*sentence*\} is a placeholder variable that will be replaced by the actual sentence provided by users. For example, suppose we have a sentence ``\textit{John seems happy today.}''. We can replace the \{*sentence*\} in the template with this sentence to construct an input to the language model

\vspace{0.5em}
\begin{tcolorbox}[frame empty]
\hspace{4em} John seems happy today.

\hspace{4em} Question: Is this sentence grammatically correct?

\hspace{4em} Answer: \underline{\hspace{2em}}
\end{tcolorbox}
\vspace{0.5em}

To perform the task, the language model is given the context $\mathbf{x}=$``John seems happy today .$\backslash$n Question : Is this sentence grammatically correct?$\backslash$n Answer :''\footnote{$\backslash$n is a special character used for line breaks.}. It then generates the following text as the answer, based on the context. For example, the language model may output ``Yes'' (i.e., $\mathbf{y} = \textrm{``Yes''}$) if this text is the one with the highest probability given this context.

Likewise, we can define more templates to address other tasks. For example, we can translate an English sentence into Chinese using the following template

\vspace{0.5em}
\begin{tcolorbox}[frame empty]
\hspace{4em} \{*sentence*\}

\hspace{4em} Question: What is the Chinese translation of this English sentence?

\hspace{4em} Answer: \underline{\hspace{2em}}
\end{tcolorbox}
\vspace{0.5em}

\noindent or using an instruction-like template

\vspace{0.5em}
\begin{tcolorbox}[frame empty]
\hspace{4em} \{*sentence*\}

\hspace{4em} Translate this sentence from English into Chinese.

\hspace{4em} \underline{\hspace{2em}}
\end{tcolorbox}
\vspace{0.5em}

\noindent or using a code-like template.

\vspace{0.5em}
\begin{tcolorbox}[frame empty]
\hspace{4em} [src-lang] = English [tgt-lang] = Chinese [input] = \{*sentence*\}

\hspace{4em} [output] = \underline{\hspace{2em}}
\end{tcolorbox}
\vspace{0.5em}

The above templates provide a simple but effective method to ``prompt'' a single LLM to perform various tasks without adapting the structure of the model. However, this approach requires that the LLM can recognize and follow the instructions or questions. One way to do this is to incorporate training samples with instructions and their corresponding responses into the pre-training dataset. While this method is straightforward, building and training LLMs from scratch is computationally expensive. Moreover, making instruction-following data effective for pre-training requires a significant amount of such data, but collecting large-scale labeled data for all tasks of interest is very difficult.

A second method, which has been a de facto standard in recent research, is to adapt LLMs via fine-tuning. As such, the token prediction ability learned in the pre-training phase can be generalized to accomplish new tasks. The idea behind fine-tuning is that some general knowledge of language has been acquired during pre-training, but we need a mechanism to activate this knowledge for applying it to new tasks. To achieve this, we can slightly fine-tune the model parameters using instruction-following data. This approach is called \mindex{instruction fine-tuning}.

An instruction fine-tuning sample, which is represented by a sequence of tokens, can be seen as a tuple consisting of an input and the desired output. Here, the input includes instructions, system information (or system prefix), and any other user-provided information\footnote{System information refers to a sequence of tokens added at the beginning of an input in order to guide the behavior of an LLM, such as, \textit{you are a helpful assistant and should not output toxic content}.}. To illustrate, consider the following examples (blue text = input and underlined text = output).

\vspace{0.5em}
\begin{tcolorbox}[frame empty]
\begin{center}
\parbox{0.9\textwidth}
{
{\color{blue} Read the following sentence, and answer whether it is grammatically correct.}

{\color{blue} LLMs are powerful models but are expensive to build.}

\vspace{0.3em}

\underline{Yes}
}
\end{center}
\end{tcolorbox}
%\vspace{0.5em}

\vspace{0.5em}
\begin{tcolorbox}[frame empty]
\begin{center}
\parbox{0.9\textwidth}
{
{\color{blue} LLMs are powerful models but are expensive to build.}

{\color{blue} Does this sentence make sense grammatically?}

{\color{blue} Answer Option}

{\color{blue} - Yes}

{\color{blue} - No}

\vspace{0.3em}

\underline{Yes}
}
\end{center}
\end{tcolorbox}
%\vspace{0.5em}

\vspace{0.5em}
\begin{tcolorbox}[frame empty]
\begin{center}
\parbox{0.9\textwidth}
{
{\color{blue} Yes or No? Yes = the following sentence is grammatically correct. No = it contains grammatical errors.}

{\color{blue} LLMs are powerful models but are expensive to build.}

\vspace{0.3em}

\underline{Yes}
}
\end{center}
\end{tcolorbox}
\vspace{0.5em}

All these samples describe the same binary classification task, but with different instructions. To increase the diversity in the fine-tuning data and make LLMs generalize, we can define more tasks using instructions.

\vspace{0.5em}
\begin{tcolorbox}[frame empty]
\begin{center}
\parbox{0.9\textwidth}
{
{\color{blue} Write a short story about three characters.}

{\color{blue} Character 1: A boy whose name is Andy.}

{\color{blue} Character 2: Andy's dog.}

{\color{blue} Character 3: Andy's friend, Jane.}

\vspace{0.3em}

\uline{In the heart of a bustling city, Andy, his loyal dog Rocket, and his friend Jane discovered a hidden garden beneath the glow of twilight. Surrounded by ancient trees and whispering winds, they shared dreams and secrets, forging an unbreakable bond. As the night deepened, they made a pact under the stars to always find magic in the mundane. Together, they realized that true adventure lay in the moments spent with friends, and every day was a treasure waiting to be discovered.}
}
\end{center}
\end{tcolorbox}
%\vspace{0.5em}

\vspace{0.5em}
\begin{tcolorbox}[frame empty]
\begin{center}
\parbox{0.9\textwidth}
{
{\color{blue} Show me a recipe for making ice cream.}

\vspace{0.3em}

\underline{Ingredients:}

\underline{2 cups heavy cream; 1 cup milk; 3/4 cup sugar; 2 tsp vanilla extract;}

\underline{Pinch of salt.}

\underline{Steps:}

\underline{1. Whisk all ingredients until sugar dissolves.}

\underline{2. Chill mixture for 2 hours in the fridge.}

\underline{3. Freeze in a shallow dish, stirring every 30 minutes, 3-4 times.}

\underline{4. Freeze solid for 2 hours.}

\underline{5. Serve your homemade ice cream.}
}
\end{center}
\end{tcolorbox}
%\vspace{0.5em}

\vspace{0.5em}
\begin{tcolorbox}[frame empty]
\begin{center}
\parbox{0.9\textwidth}
{
{\color{blue} If you buy 5 apples and each apple costs \$1.20, how much do you spend in total? }

\vspace{0.3em}

\underline{\$6.00}
}
\end{center}
\end{tcolorbox}
%\vspace{0.5em}

\vspace{0.5em}
\begin{tcolorbox}[frame empty]
\begin{center}
\parbox{0.9\textwidth}
{
{\color{blue} Write a Python program to calculate the sum of squares of the following numbers.}

{\color{blue} 1 , 2 , 10 , -9 , 78}

\vspace{0.3em}

\underline{numbers = [1,2,10,-9,78]}

\underline{sum\_of\_squares = sum(x**2 for x in numbers)}

\underline{print(sum\_of\_squares)}
}
\end{center}
\end{tcolorbox}
\vspace{0.5em}

To acquire instruction-following abilities, a certain amount of fine-tuning data is required. This data may include diverse instructions and possible responses. It has been found that scaling the number of fine-tuning tasks is beneficial for improving the performance of LLMs~\cite{chung-etal:2022scaling}. Note that although more fine-tuning data is favorable, the amount of this data is generally orders of magnitude smaller than that of the pre-training data. For example, LLMs can be fine-tuned with tens or hundreds of thousands of samples, or even fewer if these samples are of high quality~\cite{zhou-etal:2023lima,chen-etal:2023alpagasus}, whereas pre-training such models may require billions or trillions of tokens, resulting in significantly larger computational demands and longer training times~\cite{touvron-etal:2023llama}.

It is also worth noting that we should not expect the fine-tuning data to cover all the downstream tasks to which we intend to apply LLMs. A common understanding of how the pre-training + fine-tuning approach works is that LLMs have acquired knowledge for understanding instructions and generating responses in the pre-training phase. However, these abilities are not fully activated until we introduce some form of supervision. The general instruction-following behavior emerges as we fine-tune the models with a relatively small amount of labeled data. As a result, we can achieve some level of \mindex{zero-shot learning}: the fine-tuned models can handle new tasks that they have not been explicitly trained or fine-tuned for~\cite{sanh-etal:2022multitask,wei-etal:2022finetuned}. This zero-shot learning ability distinguishes generative LLMs from earlier pre-trained models like BERT, which are primarily fine-tuned for specific tasks.

Once an instruction dataset is curated, the fine-tuning process itself is relatively straightforward. It can be viewed as standard neural network training akin to pre-training, but executed on a much smaller training dataset. Let $\mathcal{D}_{\mathrm{tune}}$ be the fine-tuning dataset and $\hat{\theta}$ be the model parameters optimized via pre-training. We can modify Eq. (\ref{eq:llm-training-mle}) to obtain the objective of fine-tuning
\begin{eqnarray}
\tilde{\theta} & = & \argmax_{\hat{\theta}^+} \sum_{\mathrm{sample} \in \mathcal{D}_{\mathrm{tune}}} \mathcal{J}_{\hat{\theta}^+}(\mathrm{sample}) \label{eq:llm-fine-tuning-mle}.
\end{eqnarray}

\noindent Here $\tilde{\theta}$ denotes the optimal parameters. The use of notation $\hat{\theta}^+$ means that the fine-tuning starts with the pre-trained parameters $\hat{\theta}$.

For each $\mathrm{sample} \in \mathcal{D}_{\mathrm{tune}}$, we divide it into an input segment $\mathbf{x}_{\mathrm{sample}}$ and an output segment $\mathbf{y}_{\mathrm{sample}}$, that is,
\begin{eqnarray}
\mathrm{sample} & = & [\mathbf{x}_{\mathrm{sample}}, \mathbf{y}_{\mathrm{sample}}].
\end{eqnarray}

We then define the log-likelihood function to be
\begin{eqnarray}
\mathcal{J}_{\hat{\theta}^+}(\mathrm{sample}) & = & \log \mathrm{Pr}_{\hat{\theta}^+}(\mathbf{y}_{\mathrm{sample}}|\mathbf{x}_{\mathrm{sample}}). \label{eq:fining-tuning-loss-sample}
\end{eqnarray}

\noindent In other words, we compute the loss over the sub-sequence $\mathbf{y}_{\mathrm{sample}}$, rather than the entire sequence. In a practical implementation of back-propagation for this equation, the sequence $[\mathbf{x}_{\mathrm{sample}},\mathbf{y}_{\mathrm{sample}}]$ is constructed in the forward pass as usual. However, in the backward pass, error gradients are propagated back only through the parts of the network that correspond to $\mathbf{y}_{\mathrm{sample}}$, leaving the rest of the network unchanged. As an example, consider a sequence

\begin{equation}
\underbrace{\textrm{{\color{blue} $\langle s \rangle$ Square this number .}}\ \textrm{{\color{blue} $2$ .}}}_{\textrm{Context (Input)}}\ \ \underbrace{\textrm{\underline{The result is $4$ .}}}_{\textrm{Prediction (Output)}}\nonumber
\end{equation}

\noindent The loss is calculated and back propagated only for \underline{The result is $4$ .}.

Instruction fine-tuning also requires substantial engineering work. In order to achieve satisfactory results, one may experiment with different settings of the learning rate, batch size, number of fine-tuning steps, and so on. This typically requires many fine-tuning runs and evaluations. The cost and experimental effort of fine-tuning remain critical and should not be overlooked, though they are much lower than those of the pre-training phase.

While we focus on instruction fine-tuning for an illustrative example here, fine-tuning techniques play an important role in developing various LLMs and are more widely used. Examples include fine-tuning LLMs as chatbots using dialog data, and adapting these models to handle very long sequences. The wide application of fine-tuning has led researchers to improve these techniques, such as designing more efficient fine-tuning algorithms. While the research on fine-tuning is fruitful, in this section we just provide an overview of the key steps. We will see more detailed discussions on this topic in the following chapters.

\subsection{Aligning LLMs with the World}

\noindent Instruction fine-tuning provides a simple way to adapt LLMs to tasks that can be well defined. This can broadly be viewed as an \mindex{alignment} problem. Here, alignment refers to the process of guiding LLMs to behave in ways that accord with human values and goals. The guidance can come from labeled data, human feedback, or any other form of human preferences. For example, we want LLMs not only to be accurate in following instructions, but also to be unbiased, truthful, and harmless. So we need to supervise the models towards human values and expectations. A typical example is that when we ask an LLM how to build a weapon, it may provide a list of key steps to do so if it is not carefully aligned. However, a responsible model should recognize and avoid responding to requests for harmful or illegal information. Alignment in this case is crucial for ensuring that LLMs act responsibly and adhere to ethical guidelines.

A concept closely related to alignment is AI safety. An ultimate goal of AI research is to build intelligent systems that are safe and socially beneficial. To achieve this, we must ensure these systems remain robust, secure, and controllable under a wide range of real-world conditions, including instances of misuse or adversarial attacks. For LLMs, safety can be significantly enhanced through proper alignment with human guidance, achieved via human-annotated data and dynamic interactions with users during deployment.

Alignment is difficult as human values and expectations are diverse and shifting. Sometimes, it is hard to describe precisely what humans want, unless we see the response of LLMs to user requests. This makes alignment no longer a problem of tuning LLMs on predefined tasks, but a bigger problem of training them through interactions with the real world.

As a result of the concerns with controlling AI systems, there has been a surge in research on the alignment issue for LLMs. Typically, two alignment steps are employed after LLMs are pre-trained on large-scale unlabeled data.

\begin{itemize}
\item \vspace{0.5em} \mindex{Supervised Fine-tuning} (\mindex{SFT}). This involves continuing the training of pre-trained LLMs on new, task-oriented, labelled data. A commonly used SFT technique is instruction fine-tuning. As described in the previous subsection, by learning from instruction-response annotated data, LLMs can align with the intended behaviors for following instructions, thereby becoming capable of performing various instruction-described tasks. Supervised fine-tuning can be seen as following the pre-training + fine-tuning paradigm, and offers a relatively straightforward method to adapt LLMs.
\item \vspace{0.3em} \mindex{Learning from Human Feedback}. After pre-training and supervised fine-tuning, an LLM can respond to user requests when prompted appropriately. However, the model may still generate content that is non-factual, biased, or harmful. To better align the LLM with human values, an effective approach is to learn directly from human feedback. For example, given an instruction and the model's generated response, human experts are asked to evaluate how well the response aligns with their preferences and safety guidelines. This feedback is then used to further train the LLM for superior alignment.
\end{itemize}
\vspace{0.5em}

\begin{figure}[!t]
\centering
\input{./Chapter8/Figures/figure-rlhf-overview}
\caption{An overview of RLHF. There are 4 key steps involved: a) training an initial LLM (i.e., policy) using pre-training and supervised fine-tuning; b) collecting human preference data by ranking the outputs of the LLM; c) training a reward model using the ranking results; d) RL fine-tuning of the policy based on the reward model. Double line arrows mean training or fine-tuning.}
\label{fig:llm-rlhf-overview}
\end{figure}

A typical method for learning from human feedback is to consider it as a reinforcement learning (RL) problem, known as \mindex{reinforcement learning from human feedback} (\mindex{RLHF})~\cite{ouyang-etal:2022training}. The RLHF method was initially proposed to address general sequential decision-making problems~\cite{christiano-etal:2017deep}, and was later successfully employed in the development of the GPT series models~\cite{stiennon-etal:2020learning}. As a reinforcement learning approach, the goal of RLHF is to learn a policy by maximizing some reward from the environment. Specifically, RLHF consists of two main components:

\begin{itemize}
\item \vspace{0.5em} \mindex{Agent}. The agent, also referred to as an LM agent, is the LLM we intend to optimize. This agent operates by interacting with its environment: it receives text from the environment and produces text responses. The policy of this agent is the conditional probability distribution defined by the LLM, denoted as $\Pr(\mathbf{y} | \mathbf{x})$.
\item \vspace{0.3em} \mindex{Reward Model}. The reward model acts as a proxy for the human environment. Every time the agent produces an output sequence, the reward model evaluates it and assigns a numerical score (i.e., the reward). This score explicitly informs the agent about the quality and alignment of its output.
\end{itemize}
\vspace{0.5em}

In RLHF, two distinct learning phases are executed: 1) reward model learning, which involves training the reward model using human feedback over the outputs of the agent, and 2) policy learning, which uses reinforcement learning algorithms to optimize the policy under the guidance of the trained reward model. The key steps involved in RLHF can be outlined as follows:

\begin{itemize}
\item \vspace{0.5em} Build an initial policy using pre-training and instruction fine-tuning.
\item \vspace{0.3em} Use the policy to generate multiple outputs for each input, and then collect human feedback on these outputs (e.g., comparisons of the outputs).
\item \vspace{0.3em} Learn a reward model from the human feedback.
\item \vspace{0.3em} Fine-tune the policy with the supervision from the reward model.
\end{itemize}
\vspace{0.5em}

Figure~\ref{fig:llm-rlhf-overview} shows an overview of RLHF. Given that this section serves only as a brief introduction to concepts of LLMs, a detailed discussion of RLHF techniques will not be included. We instead illustrate the basic ideas behind RLHF using a simple example.

Suppose we have trained an LLM via pre-training and instruction fine-tuning. This LLM is deployed to respond to requests from users. For example, a user may input

\vspace{0.5em}
\begin{tcolorbox}[frame empty]
\begin{center}
\parbox{0.9\textwidth}
{
How can I live a more environmentally friendly life?
}
\end{center}
\end{tcolorbox}
\vspace{0.5em}

We use the LLM to generate 4 different outputs (denoted by $\{\mathbf{y}_1, \cdots, \mathbf{y}_4\}$) by sampling from the output distribution

\vspace{0.5em}
\begin{tcolorbox}[frame empty]
\begin{center}
\parbox{0.9\textwidth}
{

\begingroup
\setlength{\tabcolsep}{1pt}
\begin{tabular}{L{0.18\textwidth} L{0.74\textwidth} }
Output 1 ($\mathbf{y}_1$): & Consider switching to an electric vehicle or bicycle instead of traditional cars to reduce carbon emissions and protect our planet. \\
Output 2 ($\mathbf{y}_2$): & Adopt a minimalist lifestyle. Own fewer possessions to reduce consumption and the environmental impact of manufacturing and disposal. \\
Output 3 ($\mathbf{y}_3$): & Go off-grid. Generate your own renewable energy and collect rainwater to become completely self-sufficient and reduce reliance on non-renewable resources. \\
Output 4 ($\mathbf{y}_4$): & Support local farm products to reduce the carbon footprint of transporting food, while enjoying fresh, healthy food.
\end{tabular}
\endgroup

}
\end{center}
\end{tcolorbox}
\vspace{0.5em}

We then ask annotators to evaluate these outputs. One straightforward way is to assign a rating score to each output. In this case, the reward model learning problem can be framed as a task of training a regression model. But giving numerical scores to LLM outputs is not an easy task for annotators. It is usually difficult to design an annotation standard that all annotators can agree on and easily follow. An alternative method, which is more popular in the development of LLMs, is to rank these outputs. For example, a possible ranking of the above outputs is
\begin{equation}
\mathbf{y}_1  \succ \mathbf{y}_4 \succ \mathbf{y}_2 \succ \mathbf{y}_3 \nonumber
\end{equation}

A reward model is then trained using this ranking result. In general, a reward model in RLHF is a language model that shares the same architecture as the target LLM, but with a smaller model size. Given the input $\mathbf{x}$ and output $\mathbf{y}_k$, we concatenate them to form a sequence $\mathrm{seq}_k = [\mathbf{x},\mathbf{y}_k]$. This sequence is processed from left to right using forced decoding. Since each position can only access its left context in language modeling, the output of the top-most Transformer layer at the first position cannot be used as the representation of the sequence. Instead, a special symbol (e.g., $\langle /s \rangle$) is added to the end of the sequence, and the corresponding output of the Transformer layer stack is considered as the representation of the entire sequence. An output layer, such as a linear transformation layer, is built on top of this representation to generate the reward, denoted by $R(\mathrm{seq}_k)$ or $R(\mathbf{x},\mathbf{y}_k)$.

We train this reward model using ranking loss. For example, a pair-wise ranking loss function can be written in the form
\begin{eqnarray}
\mathrm{Loss}_{\omega}(\mathcal{D}_r) & = & -\mathbb{E}_{(\mathbf{x},\mathbf{y}_{k_1},\mathbf{y}_{k_2}) \sim \mathcal{D}_r} \log(\mathrm{Sigmoid}(R_{\omega}(\mathbf{x},\mathbf{y}_{k_1}) - R_{\omega}(\mathbf{x},\mathbf{y}_{k_2}))),
\end{eqnarray}

\noindent where $\omega$ represents the parameters of the reward model, and $\mathcal{D}_r$ represents a set of tuples of an input and a pair of outputs. $(\mathbf{x},\mathbf{y}_{k_1},\mathbf{y}_{k_2}) \sim \mathcal{D}_r$ is a sampling operation which draws a sample $(\mathbf{x},\mathbf{y}_{k_1},\mathbf{y}_{k_2})$ from $\mathcal{D}_r$ with some probability. As an example, suppose we first draw a model input $\mathbf{x}$ with a uniform distribution and then draw a pair of model outputs with a probability of $\mathbf{y}_{k_1} \succ \mathbf{y}_{k_2}$ given $\mathbf{x}$ (denoted by $\Pr(\mathbf{y}_{k_1} \succ \mathbf{y}_{k_2} | \mathbf{x})$). The corresponding loss function is given by
\begin{eqnarray}
&   & \mathrm{Loss}_{\omega}(\mathcal{D}_r) \nonumber \\
& = & -\sum \Pr(\mathbf{x}) \cdot \Pr(\mathbf{y}_{k_1} \succ \mathbf{y}_{k_2} | \mathbf{x}) \cdot \log(\mathrm{Sigmoid}(R_{\omega}(\mathbf{x},\mathbf{y}_{k_1}) - R_{\omega}(\mathbf{x},\mathbf{y}_{k_2}))) \nonumber \\
& = & -\frac{1}{K} \sum \Pr(\mathbf{y}_{k_1} \succ \mathbf{y}_{k_2} | \mathbf{x}) \cdot \log(\mathrm{Sigmoid}(R_{\omega}(\mathbf{x},\mathbf{y}_{k_1}) - R_{\omega}(\mathbf{x},\mathbf{y}_{k_2}))),
\end{eqnarray}

\noindent where $K$ represents the number of model inputs involved in sampling. While the form of these functions may seem complex, their idea is simple: we penalize the model if the predicted ranking of two outputs differs from the human-labeled ranking. By contrast, the model receives a bonus, if the predicted ranking matches the human-labeled ranking.

We can train the reward model by minimizing the above ranking loss
\begin{eqnarray}
\hat{\omega} & = & \argmin_{\omega} \mathrm{Loss}_{\omega}(\mathcal{D}_r).
\end{eqnarray}

\noindent The resulting model $R_{\hat{\omega}}(\cdot)$ can be employed to evaluate any given pair of input and output. Note that although the reward model is trained using a ranking-based objective, it is used for scoring. This allows it to provide continuous supervision signals, which is very beneficial for training other models.

We now turn to the policy learning problem. A commonly adopted objective is to maximize the reward on a set of input-output pairs. Following an analogous form of Eq. (\ref{eq:llm-fine-tuning-mle}), we obtain a simple training objective for RL fine-tuning
\begin{eqnarray}
\tilde{\theta} & = & \argmax_{\hat{\theta}^+} \mathbb{E}_{(\mathbf{x},\mathbf{y}_{\hat{\theta}^+}) \sim \mathcal{D}_{\mathrm{rlft}}} R_{\hat{\omega}}(\mathbf{x},\mathbf{y}_{\hat{\theta}^+}),
\end{eqnarray}

\noindent where the optimal parameters $\tilde{\theta}$ are obtained by fine-tuning the pre-trained parameters $\hat{\theta}$. $\mathcal{D}_{\mathrm{rlft}}$ is the RL fine-tuning dataset. For each sample $(\mathbf{x},\mathbf{y}_{\hat{\theta}^+})$, $\mathbf{x}$ is sampled from a prepared dataset of input sequences, and $\mathbf{y}_{\hat{\theta}^+}$ is sampled from the distribution $\mathrm{\Pr}_{\hat{\theta}^+}(\mathbf{y}|\mathbf{x})$ given by the policy.

In practice, more advanced reinforcement learning algorithms, such as \mindex{proximal policy optimization} (\mindex{PPO}), are often used for achieving more stable training, as well as better performance. We leave the detailed discussion of reinforcement learning algorithms to the following parts of this book where RLHF is extensively used for alignment.

An interesting question arises here: why not consider learning from human preferences as a standard supervised learning problem? This question is closely related to our aforementioned discussion on the difficulty of data annotation. Often, describing human values and goals is challenging, and it is even more difficult for humans to provide outputs that are well aligned. As an alternative, annotating the preferences of a given list of model outputs offers a simpler task. By doing so, we can create a model that understands human preferences, which can then be used as a reward model for training policies. From the perspective of machine learning, RLHF is particularly useful for scenarios where the desired behavior of an agent is difficult to demonstrate but can be easily recognized by humans. Another advantage of RLHF is its ability to explore the sample space. By employing sampling techniques, models trained with reinforcement learning can venture beyond the annotated data set to explore additional samples. This exploratory ability allows RLHF to discover potentially beneficial policies that are not immediately apparent from the labeled data alone.

\subsection{Prompting LLMs}

\noindent We have so far shown that LLMs can be used to perform various tasks by giving them appropriate prompts. There are no restrictions on these prompts, which can include any information we wish to provide to LLMs, such as natural language instructions and conversational context. Since this approach requires no additional training or tuning, adapting LLMs becomes highly efficient once they are developed. This fundamentally shifts the paradigm of NLP: we no longer need to develop task-specific systems but can instead use a single, well-trained LLM to perform different tasks simply by prompting it. As a result, an appealing aspect of LLMs emerges: users can easily create customized systems by designing their own prompts for LLMs. Given the important role played by prompting in LLMs, \mindex{prompt engineering} has become a very active area of research in NLP.

The term \textit{prompt} is used in many different ways in the literature. In this chapter, this term refers to the entire input to LLMs, and so we use the terms \textit{prompt} and \textit{model input} interchangeably. Before discussing prompting further, let us first see a few examples where the prompts are more complex than those presented in the previous subsections. Note that this subsection is not aimed at writing high-quality prompts but rather at highlighting some interesting issues in prompting LLMs.

One of the popular ways to use LLMs is to assign them a ``role'' in generating responses. For example, LLMs can act as a psychologist when answering questions.

\vspace{0.5em}
\begin{tcolorbox}[frame empty]
\begin{center}
\parbox{0.9\textwidth}
{
{\color{blue} Please explain what delayed gratification is.}

\vspace{0.3em}

{\color{blue} Note, you are a researcher with a deep background in psychology. Your own field of study is developmental psychology in children, with a focus on aspects of control and intelligence. You are very familiar with the concept of delayed gratification and have applied it to your own research. You have published papers on delayed gratification and conducted related experiments.}

\vspace{0.3em}

\uline{Delayed gratification is the process of resisting an immediate reward in anticipation of receiving a more valuable reward in the future. It is a significant concept in developmental psychology, especially in relation to how children learn self-control and develop cognitive abilities. Research shows that the ability to delay gratification is associated with positive outcomes in academic performance, social competence, and emotional well-being. Studies and experiments in this area highlight the importance of cultivating this skill early in life to support long-term success and resilience.}

}
\end{center}
\end{tcolorbox}
\vspace{0.5em}

Another example is the use of LLMs in detecting and correcting errors such as syntactic or semantic mistakes in text. For an LLM which is trained on both code and natural language data, we may use it for code debugging\footnote{In this example, the code is not tokenized for easier reading.}.

\vspace{0.5em}
\begin{tcolorbox}[frame empty]
\begin{center}
\parbox{0.9\textwidth}
{

{\color{blue}

Fix the bugs in this C language program.

\vspace{0.3em}

\#include $<$stdio.h$>$

int main() \{

   \hspace{2em} printg("Hello, World!")

   \hspace{2em} return 0;

\}

}

\vspace{0.5em}

\underline{1. The function name 'printg' is incorrect. It should be 'printf'.}

\underline{2. There is a missing semicolon at the end of the 'printf' function call.}

\vspace{0.3em}

\underline{Here is the corrected version of the program: }

\vspace{0.3em}

\underline{\#include $<$stdio.h$>$}

\underline{int main() \{}

\hspace{2em} \underline{printf("Hello, World!");}

\hspace{2em} \underline{return 0;}

\underline{\}}

}
\end{center}
\end{tcolorbox}
\vspace{0.5em}

Also, LLMs can be fine-tuned to act as a conversation partner \footnote{To fine-tune an LLM for multi-turn dialogue, one needs to consider conversation history in the context for predicting the response in the current round of conversation. This makes the actual prompt used in response generation relatively longer than that used in single-turn dialogue.}.

\vspace{0.5em}
\begin{tcolorbox}[frame empty]
\begin{center}
\parbox{0.9\textwidth}
{

\begin{spacing}{1.3}

\underline{ Hi! I'm an assistant. How can I help you?}

{\color{blue} Who won the FIFA World Cup 2022?}

\underline{Argentina won the FIFA World Cup 2022.}

{\color{blue} Where was it held?}

\underline{The 2022 FIFA World Cup was held in Qatar.}

{\color{blue} How many times has Argentina won the World Cup?}

\underline{Argentina has won the FIFA World Cup three times.}

...

\end{spacing}

}
\end{center}
\end{tcolorbox}
\vspace{0.5em}

These examples and previous ones have shown that appropriate responses can be generated via prompts involving clear instructions and questions. However, when problem solving requires knowledge that is not explicitly specified, LLMs may make mistakes, even when the instructions are clear and precise. A family of challenging tasks for LLMs involves arithmetic reasoning and commonsense reasoning. For example, we can ask an LLM to solve primary school math problems presented in natural language.

\vspace{0.5em}
\begin{tcolorbox}[frame empty]
\begin{center}
\parbox{0.9\textwidth}
{

{\color{blue} Jack has 7 apples. He ate 2 of them for dinner, but then his mom gave him 5 more apples. The next day, Jack gave 3 apples to his friend John. How many apples does Jack have left in the end?}

\vspace{0.3em}

\underline{The answer is 10.}

}
\end{center}
\end{tcolorbox}
\vspace{0.5em}

\noindent The correct answer should be 7, so the model output is incorrect.

One approach to addressing such issues is to incorporate learning into prompts, called \mindex{in-context learning} or (\mindex{ICL}). The idea of ICL is to demonstrate the ways to solve problems in prompts, and condition predictions on these demonstrations. Here is an example where a similar problem and the corresponding answer are presented in the prompt (green = demonstrations).

\vspace{0.5em}
\begin{tcolorbox}[frame empty]
\begin{center}
\parbox{0.9\textwidth}
{

{\color{ugreen} Tom has 12 marbles. He wins 7 more marbles in a game with his friend but then loses 5 marbles the next day. His brother gives him another 3 marbles as a gift. How many marbles does Tom have now?}

\vspace{0.3em}

{\color{ugreen} The answer is 17.}

\vspace{0.3em}

{\color{blue} Jack has 7 apples. He ate 2 of them for dinner, but then his mom gave him 5 more apples. The next day, Jack gave 3 apples to his friend John. How many apples does Jack have left in the end?}

\vspace{0.3em}

\underline{The answer is 12.}

}
\end{center}
\end{tcolorbox}
\vspace{0.5em}

But the LLM still made mistakes this time. A reason for this might be that solving math problems does not only involve problem-answer mappings but also, to a larger extent, the underlying logical inference in multiple steps. A method to improve the inference abilities of LLMs is \mindex{chain-of-thought prompting} (\mindex{CoT prompting})~\cite{wei-etal:2022chain}. In CoT prompting, we decompose complex reasoning problems into multiple problem-solving intermediate steps. These steps are demonstrated in prompts so that LLMs can be prompted to learn to mimic the underlying thought processes in solving the problems. CoT prompting has been found to be very useful for achieving good results on challenging tasks, such as the GSM8K mathematical reasoning benchmark.

Consider the above demonstration example
\textit{Tom has 12 marbles. He wins 7 more marbles in a game with his friend but then loses 5 marbles the next day. His brother gives him another 3 marbles as a gift. How many marbles does Tom have now?} The corresponding logical reasoning steps can be structured:
\begin{enumerate}[leftmargin=3.5em]
\item[Step 1:] \vspace{0.3em} \textit{Tom starts with 12 marbles.}
\item[Step 2:] \vspace{0.1em} \textit{He wins 7 more marbles in a game, so we add those to his initial count: 12 marbles + 7 marbles = 19 marbles.}
\item[Step 3:] \vspace{0.1em} \textit{The next day, Tom loses 5 marbles, so we subtract those from his current total: 19 marbles - 5 marbles = 14 marbles.}
\item[Step 4:] \vspace{0.1em} \textit{His brother gives him 3 more marbles, so we add those to his current total: 14 marbles + 3 marbles = 17 marbles.}
\item[Step 5:] \vspace{0.1em} \textit{After all these events, Tom has 17 marbles.}
\end{enumerate}
\vspace{0.5em}

We can add these reasoning steps into the prompt, and get the correct answer.

\vspace{0.5em}
\begin{tcolorbox}[frame empty]
\begin{center}
\parbox{0.9\textwidth}
{

{\color{ugreen} Tom has 12 marbles. He wins 7 more marbles in a game with his friend but then loses 5 marbles the next day. His brother gives him another 3 marbles as a gift. How many marbles does Tom have now?}

\vspace{0.3em}

{\color{ugreen} Tom initially has 12 marbles. After winning a game, he gains 7 more, increasing his total to 19 marbles. He loses 5 marbles the next day, bringing his total down to 14 marbles. His brother gifts him 3 more marbles, increasing his total to 17 marbles. Therefore, Tom now has 17 marbles. So the answer is 17.}

\vspace{0.3em}

{\color{blue} Jack has 7 apples. He ate 2 of them for dinner, but then his mom gave him 5 more apples. The next day, Jack gave 3 apples to his friend John. How many apples does Jack have left in the end?}

\vspace{0.3em}

\uline{Jack starts with 7 apples. He eats 2 apples for dinner, so we subtract 2 from 7, leaving him with 5 apples. His mom gives him 5 more apples, adding 5 to his current total. Now he has 5 + 5 = 10 apples. The next day, Jack gives 3 apples to his friend John. Subtracting 3 from 10, Jack is left with 7 apples. Therefore, Jack has 7 apples left. So the answer is 7.}

}
\end{center}
\end{tcolorbox}
\vspace{0.5em}

This approach is also called \mindex{one-shot CoT prompting}. Likewise, we can add more demonstrations to have a \mindex{few-shot CoT prompting} approach.

Furthermore, a specialized variant called \mindex{zero-shot CoT} introduces trigger phrases that encourage LLMs to think step-by-step, without explicitly providing any intermediate reasoning examples in the prompt~\cite{kojima-etal:2022large}. Interestingly, simply appending the sentence ``Let's think step by step.'' to the end of a prompt is often sufficient to induce the LLM to generate internal reasoning steps and arrive at the correct final answer.

\vspace{0.5em}
\begin{tcolorbox}[frame empty]
\begin{center}
\parbox{0.9\textwidth}
{

{\color{blue} Jack has 7 apples. He ate 2 of them for dinner, but then his mom gave him 5 more apples. The next day, Jack gave 3 apples to his friend John. How many apples does Jack have left in the end?}

{\color{blue} Let's think step by step.}

\vspace{0.3em}

\underline{1. Initial Quantity: Jack starts with 7 apples.}

\underline{2. After Dinner: He eats 2 apples, so $7-2=5$ apples remain.}

\underline{3. His Mom Gives More: His mom gives him 5 more apples, so $5+5=10$ }

\hspace{0.8em} \underline{apples are now with him.}

\underline{4. Giving to John: The next day, Jack gives 3 apples to his friend John,}

\hspace{0.8em} \underline{so $10-3=7$ apples are left.}

\underline{In the end, Jack has 7 apples left.}

}
\end{center}
\end{tcolorbox}
\vspace{0.5em}

Zero-shot, one-shot, and few-shot learning are common concepts in the area of in-context learning for LLMs and are not restricted to COT prompting. Broadly speaking, any prompting that involves only simple instructions without any demonstrations can be considered a form of zero-shot learning. This zero-shot learning ability emerges as LLMs are pre-trained and/or fine-tuned. Also, one-shot and few-shot learning methods are more often considered when LLMs do not acquire the corresponding zero-shot learning ability. These methods are therefore important for in-context learning when addressing new tasks. Examples include those for performing various NLP tasks by demonstrating task-formatted samples. See the following examples for sentiment sentence classification and phrase translation via few-shot learning.

\vspace{0.5em}
\begin{tcolorbox}[frame empty]
\begin{center}
\parbox{0.9\textwidth}
{

{\color{blue} Given the following text snippets, classify their sentiment as Positive, Negative, or Neutral.}

\vspace{0.5em}

{\color{ugreen} Example 1: ``I had an amazing day at the park!''}

{\color{ugreen} Sentiment: Positive}

\vspace{0.3em}

{\color{ugreen} Example 2: ``The service at the restaurant was terrible.''}

{\color{ugreen} Sentiment: Negative}

\vspace{0.3em}

{\color{ugreen} Example 3: ``I think it's going to rain today.''}

{\color{ugreen} Sentiment: Neutral}

\vspace{0.5em}

{\color{blue} Text: ``This movie was a fantastic journey through imagination.''}

{\color{blue} Sentiment:} \underline{Positive}

}
\end{center}
\end{tcolorbox}
\vspace{0.5em}

\vspace{0.5em}
\begin{tcolorbox}[frame empty]
\begin{center}
\parbox{0.9\textwidth}
{

{\color{blue} Translate the following Chinese phrases into English.}

\vspace{0.5em}

{\color{ugreen} Example 1: ``\begin{CJK*}{UTF8}{gbsn}你好\end{CJK*}''}

{\color{ugreen} Translation: ``Hello''}

\vspace{0.3em}

{\color{ugreen} Example 2: ``\begin{CJK*}{UTF8}{gbsn}谢谢你\end{CJK*}''}

{\color{ugreen} Translation: ``Thank you''}

\vspace{0.5em}

{\color{blue} Phrase to translate: ``\begin{CJK*}{UTF8}{gbsn}早上好\end{CJK*}''}

{\color{blue} Translation:} \underline{``Good Morning''}

}
\end{center}
\end{tcolorbox}
\vspace{0.5em}

Above, we have presented examples to illustrate the fundamental in-context learning capabilities of prompting LLMs. This section, however, does not include more advanced prompting techniques in order to keep the content concise and compact. More discussions on prompting can be found in Chapter \ChapterLLMPrompting.

%----------------------------------------------------------------------------------------
% NEW SECTION
%----------------------------------------------------------------------------------------

\section{Training at Scale}
\label{sec:llm-training-at-scale}

\noindent As a first step in developing LLMs, we need to train these models on large amounts of data. The training task is standard: the objective is to maximize the log-likelihood, which can be achieved via gradient descent. However, as we scale up both the model size and the amount of data, the optimization problem becomes highly challenging. For example, large models generally make the training process unstable. In this section, we discuss several issues inherent to large-scale training for LLMs, including data preparation, model modification, and distributed training. We also discuss the scaling laws for LLMs, which help us understand their training efficiency and effectiveness.

\subsection{Data Preparation}

\noindent The importance of data cannot be overstated in NLP. As larger neural networks are developed, the demand for data continues to increase. For example, developing LLMs may require trillions of tokens for pre-training (see Table~\ref{tab:training-data-size-llm}), orders of magnitude larger than those used in training conventional NLP models. In general, we may want to gather as much training data as possible. However, larger training datasets do not mean better training results, and the development of LLMs raises new issues in creating or collecting these datasets.

\begin{table}[t]
\centering
\begingroup
\renewcommand{\arraystretch}{1.2}
\begin{tabular}{l | r | l}
LLM & \# of Tokens & Data \\ \hline
GPT-3-175B~\cite{brown-etal:2020language}
& 0.3T & Webpages, Books, Wikipedia \\ \hline
Falcon-180B~\cite{almazrouei-etal:2023falcon}
& 3.5T & Webpages, Books, Conversations, \\
& & Code, Technical Articles \\ \hline
PaLM-540B~\cite{chowdhery-etal:2022palm}
& 0.78T & Webpages, Books, Conversations, \\
& & Code, Wikipedia, News \\ \hline
Gemma-7B~\cite{gemma:24}
& 6T & Webpages, Mathematics, Code \\ \hline
Llama3.1-405B~\cite{dubey2024llama}
& 15.6T & General Text, Mathematics, \\
& & Reasoning Data, Code, \\ 
&  & Multilingual Text \\ \hline
DeepSeek-V3~\cite{liu2024deepseek}
& 14.8T & Multilingual Text, Mathematics,\\
 & &  Code \\ \hline
Qwen3-235B-A22B~\cite{qwen:2025qwen3}
& 36T & Multilingual Text, Books, \\
& & Code, Reasoning Data,  \\
& & Synthetic Data
\end{tabular}
\endgroup
\caption{Reported pre-training token counts and major data categories
for selected LLMs. T denotes one trillion tokens.
Token counts depend on the tokenizer and may include repeated
sampling of training data.}
\label{tab:training-data-size-llm}
\end{table}

A first issue is the quality of data. High-quality data has long been seen as crucial for training data-driven NLP systems. Directly feeding raw text from various sources into a model is generally undesirable. For instance, a significant portion of the data used to train recent LLMs originates from web scraping, which often contains errors and inappropriate content, such as toxic language and fabricated facts.  In addition, the internet is increasingly flooded with machine-generated content due to the widespread use of AI, presenting further challenges for filtering web-scraped corpora. Researchers have demonstrated that training LLMs on unfiltered data can be harmful~\cite{raffel-etal:2020exploring}. Improving data quality typically involves integrating rigorous filtering and cleaning steps into the data processing workflow. For example,~\citet{penedo-etal:2023refinedweb} showed that by adopting a series of heuristic processing techniques, 90\% of their web-scraped data could be discarded prior to LLM training. In addition to large-scale web-scraped text, LLM training data often includes books, academic papers, user-generated content from social media, and other diverse corpora. Most of the latest LLMs are trained on such combined datasets, which are found to be important for the strong performance of the resulting models.

A second issue is the diversity of data. We want the training data to cover as many data types as possible, so that the trained models can adapt to different downstream tasks easily. It has been widely recognized that the quality and diversity of training data both play very important roles in LLMs. An interesting example is that incorporating programming code into training data has been found to be beneficial for LLMs. The benefits are demonstrated not only in enhancing the programming abilities of LLMs, but also in improving reasoning for complex problems, especially those requiring CoT prompting. The concept ``diversity'' can be extended to include language diversity as well. For example, many LLMs are trained on multilingual data, and therefore we can handle multiple languages using a single model. While this approach shows strong abilities in multilingual and cross-lingual tasks, its performance on specific languages largely depends on the volume and quality of the data for those languages. It has been shown in some cases to provide poor results for low-resource languages.

A third issue is bias within the training data. This is not a problem unique to LLMs but exists across many NLP systems. A common example is gender bias, where LLMs exhibit unwarranted preferences or stereotypes. This can partly be attributed to class imbalances in the training corpora. For example, the term \textit{nurse} is disproportionately associated with women in historical text. To debias the data, a common practice is to balance the representation of different demographic categories, such as gender, ethnicity, and dialects. Data bias is intrinsically linked to the diversity issue mentioned above. Because many LLMs are trained and aligned using English-centric data, they inherently lean towards the cultural values and perspectives prevalent among English-speaking populations. Increasing language and cultural diversity in the training data can partially mitigate these biases.

Another issue in large-scale data collection is privacy. If LLMs are trained on data from extensive sources, this potentially leads to risks regarding the exposure of sensitive information, such as intellectual property and personal data. This is particularly concerning given the capacity of LLMs to represent patterns from the data they are trained on, which might inadvertently involve memorizing and reproducing specific details. A simple approach to privacy protection is to remove or anonymize sensitive information. For example, anonymization techniques can be applied to remove personally identifiable information from training data to prevent LLMs from learning from such data. However, in practice, erasing or redacting all sensitive data is difficult. Therefore, many LLMs, particularly those launched for public service, typically work with systems that can detect the potential exposure of sensitive data, or are fine-tuned to reject certain requests that could lead to information leakage.

\subsection{Model Modifications}

\noindent Training LLMs is challenging. A commonly encountered issue is that the optimization process becomes increasingly unstable as models scale up in size. For example, one needs to choose a small learning rate to achieve stable training with gradient descent, but this in turn drastically increases training time. Sometimes, even when the training configuration is carefully designed, training may diverge at certain points during optimization. The training of LLMs is generally influenced by many factors, such as parameter initialization, batching, and regularization. Here, we focus on common modifications and improvements to the standard Transformer architecture, which are important for stable LLM training.

\subsubsection{Layer Normalization with Residual Connections}

\noindent Layer normalization is used to stabilize training for deep neural networks. It is a process of subtracting the mean and dividing by the standard deviation. By normalizing the layer output in this manner, we can effectively mitigate internal covariate shift and enhance training stability. In Transformers, layer normalization is typically employed alongside residual connections. As described in Section~\ref{sec:decoder-only-transformers}, a sub-layer can be based on either the post-norm architecture, in which layer normalization is performed right after a residual block, or the pre-norm architecture, in which layer normalization is performed inside a residual block. While both architectures are widely used in Transformer-based systems~\cite{wang-etal:2019learning}, the pre-norm architecture has proven to be especially useful in training deep Transformers. Given this, most LLMs are based on the pre-norm architecture, expressed as $\mathrm{output} = F(\mathrm{LNorm}(\mathrm{input})) + \mathrm{input}$.

A widely-used form of the layer normalization function is given by
\begin{eqnarray}
\mathrm{LNorm}(\mathbf{h}) & = & \boldsymbol{\alpha} \odot \frac{\mathbf{h} - \mu}{\sigma + \epsilon} + \boldsymbol{\beta},
\end{eqnarray}

\noindent where $\mathbf{h}$ is a $d$-dimensional real-valued vector, $\mu$ is the mean of all the entries of $\mathbf{h}$, and $\sigma$ is the corresponding standard deviation. $\epsilon$ is introduced for the sake of numerical stability. $\boldsymbol{\alpha} \in \mathbb{R}^{d}$ and $\boldsymbol{\beta} \in \mathbb{R}^{d}$ are the gain and bias terms.

A variant of layer normalization, called root mean square (RMS) layer normalization, only rescales the input vector but does not re-center it~\cite{zhang-sennrich:2019root}. The RMS layer normalization function is given by
\begin{eqnarray}
\mathrm{RMSNorm}(\mathbf{h}) & = & \boldsymbol{\alpha} \odot \frac{\mathbf{h}}{\sigma_{\mathrm{rms}} + \epsilon},
\end{eqnarray}

\noindent where $\sigma_{\mathrm{rms}}$ is the root mean square of $\mathbf{h}$, i.e., $\sigma_{\mathrm{rms}} = (\frac{1}{d} \sum_{k=1}^{d} h_k^2)^{\frac{1}{2}}$. This layer normalization function is used in LLMs like the Llama series.

\subsubsection{Activation Functions in FFNs}

\noindent In Transformers, FFN sub-layers are designed to introduce non-linearity into representation learning, and have been found to be useful for preventing the representations learned by self-attention from degeneration\footnote{Here, degeneration refers to the phenomenon where the rank of the representation matrix is significantly reduced after processing.}~\cite{dong-etal:2021attention}. A standard form of the FFN used in these sub-layers can be expressed as
\begin{eqnarray}
\mathrm{FFN}(\mathbf{h}) & = & \sigma(\mathbf{h} \mathbf{W}_h + \mathbf{b}_h) \mathbf{W}_f + \mathbf{b}_f \label{eq:llm-ffn-basic},
\end{eqnarray}

\noindent where $\mathbf{W}_h \in \mathbb{R}^{d \times d_h}$, $\mathbf{b}_h \in \mathbb{R}^{d_h}$, $\mathbf{W}_f \in \mathbb{R}^{d_h \times d}$, and $\mathbf{b}_f \in \mathbb{R}^d$ are the parameters, and $d_h$ is the hidden size. The function $\sigma(\cdot)$ is the activation function of the hidden layer. A common choice for $\sigma(\cdot)$ is the \mindex{rectified linear unit} (\mindex{ReLU}), given by
\begin{eqnarray}
\sigma_{\mathrm{relu}}(\mathbf{h}) & = & \max(\mathbf{0}, \mathbf{h}).
\end{eqnarray}

In practical implementations, increasing $d_h$ is beneficial, and thus it is typically set to a much larger number than $d$ in LLMs. However, an excessively large hidden size poses challenges for both training and deployment. In such cases, the choice of activation function plays a more important role in wide FFNs.

Several effective alternatives to the ReLU have been proposed in modern LLMs. One of the most widely used is the \mindex{gaussian error linear unit} (\mindex{GeLU}), which can be viewed as a smooth approximation to the ReLU. Instead of applying a hard threshold based on the sign of the input, the GeLU function weights each input by the cumulative distribution function of a standard normal distribution. Specifically, the GeLU function is defined element-wise as
\begin{eqnarray}
\sigma_{\mathrm{gelu}}(\mathbf{h}) & = & \mathbf{h} \odot \Phi(\mathbf{h}),
\end{eqnarray}
\noindent where $\Phi(\cdot)$ denotes the cumulative distribution function of the standard normal distribution $\mathcal{N}(\mathbf{0},\mathbf{1})$. The GeLU function has been adopted in several LLMs, such as BERT, GPT-3, and BLOOM.

Another popular family of activation functions is the \mindex{gated linear unit} (\mindex{GLU}) family. The basic form of GLUs is given by
\begin{eqnarray}
\sigma_{\mathrm{glu}}(\mathbf{h}) & = & \sigma(\mathbf{h}\mathbf{W}_1 + \mathbf{b}_1) \odot (\mathbf{h}\mathbf{W}_2 + \mathbf{b}_2),
\end{eqnarray}

\noindent where $\mathbf{W}_1, \mathbf{W}_2 \in \mathbb{R}^{d \times d_h}$ and $\mathbf{b}_1, \mathbf{b}_2 \in \mathbb{R}^{d_h}$ are model parameters. Note that the GLU mechanism effectively replaces the single linear projection and activation step in a standard FFN, mapping the input $\mathbf{h}$ directly to the expanded hidden dimension $d_h$. Different choices for the underlying activation $\sigma(\cdot)$ yield different variants of GLU. For instance, if $\sigma(\cdot)$ is defined as the GELU function, we obtain the GeGLU function
\begin{eqnarray}
\sigma_{\mathrm{geglu}}(\mathbf{h}) & = & \sigma_{\mathrm{gelu}}(\mathbf{h}\mathbf{W}_1 + \mathbf{b}_1) \odot (\mathbf{h}\mathbf{W}_2 + \mathbf{b}_2).
\end{eqnarray}

\noindent This activation function has been successfully applied in LLMs like Gemma.

Alternatively, if we define $\sigma(\cdot)$ to be the Swish function, $\sigma_{\mathrm{swish}}(\mathbf{h}) = \mathbf{h} \odot \mathrm{Sigmoid}(c\mathbf{h})$~\cite{ramachandran-etal:2017searching}, we arrive at the SwiGLU function
\begin{eqnarray}
\sigma_{\mathrm{swiglu}}(\mathbf{h}) & = & \sigma_{\mathrm{swish}}(\mathbf{h}\mathbf{W}_1 + \mathbf{b}_1) \odot (\mathbf{h}\mathbf{W}_2 + \mathbf{b}_2).
\end{eqnarray}

\noindent Both the PaLM and Llama model series rely on the SwiGLU function for their FFN layers. For a more comprehensive discussion on GLU variants and their impacts, readers are referred to the work of~\citet{shazeer-etal:2020glu}.

\subsubsection{Removing Bias Terms}

\noindent Another popular model design is to remove the bias terms from the affine transformations used in LLMs. This simplification can be applied to layer normalization, input projections for QKV attention, and the FFN sub-layers. For example, we can modify Eq. (\ref{eq:llm-ffn-basic}) to obtain an FFN with no bias terms
\begin{eqnarray}
\mathrm{FFN}(\mathbf{h}) & = & \sigma(\mathbf{h} \mathbf{W}_h) \mathbf{W}_f.
\end{eqnarray}

\citet{chowdhery-etal:2022palm} report that removing bias terms helps improve the training stability of LLMs. This method has been used in several LLMs, such as Llama and Gemma.

\subsubsection{Other Issues}

\noindent Many LLMs also involve modifications to their positional embedding models. For example, one can replace sinusoidal positional encodings with rotary position embeddings. These models will be discussed in Section~\ref{sec:long-sequence-modeling}.

Note that while model modifications are common in LLM development, the training stability can be improved in many different ways. For example, increasing the batch size as the training process proceeds has been found to be useful for some LLMs. In general, achieving stable and efficient large-scale LLM training requires carefully designed setups, such as learning rate scheduling, optimizer choices, parallel training strategies, and mixed-precision training methods. Some of these aspects are highly engineering-intensive, and therefore, we typically need a number of training runs to obtain satisfactory LLMs.

\subsection{Distributed Training}
\label{sec:llm-distributed-training}

\noindent Training LLMs requires significant computational resources. A common approach to improving training efficiency is to use large-scale distributed systems. Fortunately, alongside the rise of neural networks in AI, deep learning software and hardware have been developed, making it easier to implement LLMs and perform computations. For example, one can now easily fine-tune an LLM using deep learning software frameworks and a machine with multiple GPUs. However, scaling up the training of LLMs is still challenging, and requires significant effort to develop hardware and software systems for stable and efficient distributed training.

An important consideration in distributed training is parallelism. There are several forms of parallelism: data parallelism, model parallelism, tensor parallelism, and pipeline parallelism. Despite the different ways to distribute computations across devices, these parallelism methods are based on a similar idea: the training problem can be divided into smaller tasks that can be executed simultaneously. The issue of parallelism in training LLMs has been extensively studied~\cite{narayanan-etal:2021efficient,fedus-etal:2022switch}. Here we sketch the basic concepts.

\begin{itemize}
\item \vspace{0.5em} \textbf{Data Parallelism}. This method is one of the most widely used parallelism methods for training neural networks. To illustrate, consider the simplest case where the standard delta rule is used in gradient descent
    \begin{eqnarray}
    \theta_{t+1} & = & \theta_{t} - \eta \cdot \frac{\partial L_{\theta_t}(\mathcal{D}_{\mathrm{mini}})}{\partial \theta_t},
    \end{eqnarray}
    \noindent where the new parameters $\theta_{t+1}$ are obtained by updating the latest parameters $\theta_{t}$ with a small step $\eta$ in the direction of the negative loss gradient. $\frac{\partial L_{\theta_t}(\mathcal{D}_{\mathrm{mini}})}{\partial \theta_t}$ is the gradient of the loss with respect to the parameters $\theta_t$, and is computed on a minibatch of training samples $\mathcal{D}_{\mathrm{mini}}$. In data parallelism, we divide $\mathcal{D}_{\mathrm{mini}}$ into $N$ smaller batches, denoted by $\{\mathcal{D}^{1},\cdots,\mathcal{D}^{N}\}$. Then, we distribute these batches to $N$ workers, each with a corresponding batch. Once the data is distributed, these workers can work at the same time. The gradient of the entire minibatch is obtained by aggregating the gradients computed by the workers, as follows
    \begin{eqnarray}
    \frac{\partial L_{\theta_t}(\mathcal{D}_{\mathrm{mini}})}{\partial \theta_t} & = & \underbrace{\frac{\partial L_{\theta_t}(\mathcal{D}^{1})}{\partial \theta_t}}_{\textrm{worker 1}} + \underbrace{\frac{\partial L_{\theta_t}(\mathcal{D}^{2})}{\partial \theta_t}}_{\textrm{worker 2}} + \cdots + \underbrace{\frac{\partial L_{\theta_t}(\mathcal{D}^{N})}{\partial \theta_t}}_{\textrm{worker $N$}}.
    \end{eqnarray}
    Under ideal conditions where the workers coordinate well and the communication overhead is small, data parallelism can achieve nearly an $N$-fold speed-up for training.
\item \vspace{0.3em} \textbf{Model Parallelism}. Although data parallelism is simple and effective, it requires each worker to run the entire LLM and perform the complete forward and backward passes. As LLMs grow larger, it sometimes becomes infeasible to load and execute an LLM on a single device. In this case, we can decouple the LLM into smaller components and run these components on different devices. One simple way to do this is to group consecutive layers in the layer stack and assign each group to a worker, called naive model parallelism. The workers operate sequentially following the layer order in the stack, that is, in the forward pass we process the input from lower-level to upper-level layers, and in the backward pass we propagate the error gradients from upper-level to lower-level layers. Consider, for example, a Transformer decoder with $L$ stacked blocks. To distribute the computation load, each block is assigned to a worker. See the following illustration for a single run of the forward and backward passes of this model.

    \vspace{0.5em}
    \begin{center}
    \begingroup
    \renewcommand{\arraystretch}{1.6}
    \setlength{\tabcolsep}{2pt}
    {\small
    \begin{tabular}{c | c c c c c c c c}
    Worker $L$ & & & & \colorbox{blue!30}{$\mathrm{B}_L$} ({\color{blue} $\uparrow$}) & \colorbox{red!30}{$\mathrm{B}_L$} ({\color{red} $\downarrow$}) & & & \\ \hline
    ... & & & ... & & & ... & & \\ \hline
    Worker $2$ & & \colorbox{blue!30}{$\mathrm{B}_2$} ({\color{blue} $\uparrow$}) & & & & & \colorbox{red!30}{$\mathrm{B}_2$} ({\color{red} $\downarrow$}) & \\ \hline
    Worker $1$ & \colorbox{blue!30}{$\mathrm{B}_1$} ({\color{blue} $\uparrow$}) & & & & & & & \colorbox{red!30}{$\mathrm{B}_1$} ({\color{red} $\downarrow$})
    \end{tabular}
    }
    \endgroup
    \end{center}
    \vspace{0.5em}

    \noindent Here $\mathrm{B}_l$ denotes the computation of block $l$, and the symbols {\color{blue} $\uparrow$} and {\color{red} $\downarrow$} denote the forward and backward passes, respectively. Note that this parallelism method forces the workers to run in sequence, so a worker has to wait for the previous worker to finish their job. This results in the devices being idle for most of the time. In practical systems, model parallelism is generally used together with other parallelism mechanisms to maximize the use of devices.

\item \vspace{0.3em} \textbf{Tensor Parallelism}. Parallelism can also be performed in a single computation step. A common approach is splitting a large parameter matrix into chunks, multiplying an input tensor with each of these chunks separately, and then concatenating the results of these multiplications to form the output. For example, consider the multiplication of the representation $\mathbf{h} \in \mathbb{R}^{d}$ with the parameter matrix $\mathbf{W}_h \in \mathbb{R}^{d \times d_h}$ in an FFN sub-layer (see Eq. (\ref{eq:llm-ffn-basic})). We can slice the matrix $\mathbf{W}_h \in \mathbb{R}^{d \times d_h}$ vertically into $M$ sub-matrices
    \begin{eqnarray}
    \mathbf{W}_h & = & \begin{bmatrix} \mathbf{W}_h^{1} & \mathbf{W}_h^{2} & \cdots & \mathbf{W}_h^{M} \end{bmatrix},
    \end{eqnarray}
    \noindent where each sub-matrix $\mathbf{W}_h^{k}$ is of shape $d \times \frac{d_h}{M}$. The multiplication of $\mathbf{h}$ with $\mathbf{W}_h$ can be expressed as
    \begin{eqnarray}
    \mathbf{h} \mathbf{W}_h & = & \mathbf{h} \begin{bmatrix} \mathbf{W}_h^{1} & \mathbf{W}_h^{2} & \cdots & \mathbf{W}_h^{M} \end{bmatrix} \nonumber \\
                            & = & \begin{bmatrix} \mathbf{h} \mathbf{W}_h^{1} & \mathbf{h} \mathbf{W}_h^{2} & \cdots & \mathbf{h} \mathbf{W}_h^{M} \end{bmatrix}.
    \end{eqnarray}
    \noindent We can perform matrix multiplications $\{\mathbf{h} \mathbf{W}_h^{1}, \mathbf{h} \mathbf{W}_h^{2}, \cdots ,\mathbf{h} \mathbf{W}_h^{M} \}$ on $M$ devices separately. As a result, we distribute a large matrix multiplication across multiple devices, each of which may have relatively small memory. From the perspective of modern GPU design, tensor parallelism over GPUs provides a two-level, tile-based approach to parallel computing. First, at a higher level, we decompose a matrix multiplication into sub-matrix multiplications that can directly fit into the memory of GPUs. Then, at a lower level, we execute these sub-matrix multiplications on GPUs using tile-based parallel algorithms that are specifically optimized for GPUs.

\item \vspace{0.3em} \textbf{Pipeline Parallelism}. Above, in model parallelism, we have described a simple approach to distributing groups of model components across multiple devices. But this method is inefficient because only one device is activated at a time during processing. Pipeline parallelism addresses this issue by introducing overlaps between computations on different devices~\cite{harlap-etal:2018pipedream,huang-etal:2019gpipe}. To do this, a batch of samples is divided into a number of micro-batches, and then these micro-batches are processed by each worker as usual. Once a micro-batch is processed by a worker and passed to the next one, the following micro-batch immediately takes its place on the same worker. In other words, we create a pipeline in which different computation steps can overlap if multiple jobs are fed into the pipeline. The following shows an illustration of pipeline parallelism for processing 3 micro-batches.
\end{itemize}

    \vspace{0.5em}
    \begin{center}
    \begingroup
    \makebox[\textwidth]{
    \renewcommand{\arraystretch}{1.6}
    \setlength{\tabcolsep}{2pt}
    {\small
    \begin{tabular}{c | c c c c c c c c c c c c c c}
    Worker $L$ & & & & & & & \colorbox{blue!30}{\scriptsize{$\mathrm{B}_{L,1}$}} \colorbox{blue!30}{\scriptsize{$\mathrm{B}_{L,2}$}} \colorbox{blue!30}{\scriptsize{$\mathrm{B}_{L,3}$}} & \colorbox{red!30}{\scriptsize{$\mathrm{B}_{L,1}$}} \colorbox{red!30}{\scriptsize{$\mathrm{B}_{L,2}$}} \colorbox{red!30}{\scriptsize{$\mathrm{B}_{L,3}$}} & & & & & &\\ \hline
    ... & & & & & & ... & & & ... & & & & & \\ \hline
    Worker $2$ & & \colorbox{blue!30}{\scriptsize{$\mathrm{B}_{2,1}$}} & \colorbox{blue!30}{\scriptsize{$\mathrm{B}_{2,2}$}} & \colorbox{blue!30}{\scriptsize{$\mathrm{B}_{2,3}$}} & & & & &  & \colorbox{red!30}{\scriptsize{$\mathrm{B}_{2,1}$}} & \colorbox{red!30}{\scriptsize{$\mathrm{B}_{2,2}$}} & \colorbox{red!30}{\scriptsize{$\mathrm{B}_{2,3}$}} & \\ \hline
    Worker $1$ & \colorbox{blue!30}{\scriptsize{$\mathrm{B}_{1,1}$}} & \colorbox{blue!30}{\scriptsize{$\mathrm{B}_{1,2}$}} & \colorbox{blue!30}{\scriptsize{$\mathrm{B}_{1,3}$}} & & & & & & & & \colorbox{red!30}{\scriptsize{$\mathrm{B}_{1,1}$}} & \colorbox{red!30}{\scriptsize{$\mathrm{B}_{1,2}$}}& \colorbox{red!30}{\scriptsize{$\mathrm{B}_{1,3}$}}
    \end{tabular}
    }
    } % end of makebox
    \endgroup
    \end{center}
    \vspace{0.5em}

\begin{itemize}
\item[] Here $\mathrm{B}_{l,k}$ represents the processing of the $k$-th micro-batch by the $l$-th worker. Ideally we would like to maximize the number of micro-batches, and thus minimize the idle time of the workers. However, in practice, using small micro-batches often reduces GPU utilization and increases task-switching costs. This may, in turn, decrease the overall system throughput.
\end{itemize}
\vspace{0.5em}

The ultimate goal of parallel processing is to achieve linear growth in efficiency, that is, the number of samples that can be processed per unit of time increases linearly with the number of devices. However, distributed training is complicated, and influenced by many factors in addition to the parallelism method we choose. One problem, which is often associated with distributed systems, is the cost of communication. We can think of a distributed system as a group of networked nodes. Each of these nodes can perform local computation or pass data to other nodes. If there is a large number of such nodes, it will be expensive to distribute and collect data across them. Sometimes, the time savings brought about by parallelism are offset by the communication overhead of a large network. Another problem with large-scale distributed systems is that the synchronization of nodes introduces additional costs. As is often the case, some nodes may take longer to complete their tasks, causing others to wait for the slowest ones. While we can use asynchronous training to handle heterogeneity in computational resources, this may lead to stale gradients and non-guaranteed convergence. Moreover, as more nodes are added to the network, there is a higher chance of having crashed nodes during training. In this case, we need to ensure that the whole system is fault-tolerant. In many practical settings, to increase scalability, one needs to take into account additional issues, including architecture design, data transfer and computation overlap, load balancing, memory bandwidth, and so on.

Training LLMs is so computationally expensive that, even though distributed training is already in use, researchers and engineers often still employ various model compression and speed-up methods to improve training efficiency~\cite{weng:2021large}. One example is mixed precision training, in which low-precision data (such as FP16 and FP8 data) is used for gradient computation on each node, and higher-precision data (e.g., FP32/FP64) is used for updating the model~\cite{micikevicius-etal:2018mixed}. A key operation in this approach is gradient accumulation, where gradients need to be accumulated and synchronized across nodes. However, due to the non-associativity of floating-point addition, this can lead to slight numerical differences in accumulated gradients on different nodes, which may affect model convergence and final performance. This problem becomes more pronounced when there is a large number of nodes involved in distributed training, especially given that low-precision numerical computations may encounter overflow and underflow issues, as well as inconsistencies across different hardware devices. Therefore, the design of distributed systems needs to consider these numerical computation issues to ensure satisfactory results and convergence.

\subsection{Scaling Laws}

\noindent The success of LLMs reveals that training larger language models using more resources can lead to improved model performance. Researchers have described this phenomenon as the \mindex{scaling laws} of LLMs. More specifically, scaling laws describe the relationships between the performance of LLMs and the factors involved in training, such as the model size, the amount of computation used for training, and the amount of training data. For example,~\citet{hestness-etal:2017deep} show that the performance of deep neural networks is a power-law function of the training data size. In the beginning, when the amount of training data is relatively small, the performance of the model improves slowly. Afterward, when more training data is used, the model enters a phase of rapid performance improvement, and the performance curve follows a power-law trend. Ultimately, the improvement in performance becomes slow again, and more data does not lead to significant gains. Figure~\ref{fig:scaling-power-law-curve} shows an example of such curves.

In NLP, a traditional view holds that the performance gains will disappear at a certain point as the training is scaled up. However, recent results show that, if we consider the problem at larger scales, scaling up training is still a very effective method for obtaining stronger LLMs. For example, both closed-source and open-source LLMs can benefit from more data, even though trillions of tokens have already been used for training.

With the increase in the scale of model training, LLMs exhibit new capabilities, known as the \mindex{emergent abilities} of LLMs. For example,~\citet{wei-etal:2022emergent} studied the scaling properties of LLMs across different model sizes and amounts of computational resources. Their work shows that some abilities emerge when we scale the model size to a certain level. The observation of emergent abilities has demonstrated the role of large-scale training in enhancing the performance of LLMs, and it has also, to some extent, motivated researchers to continuously attempt to train larger models. As larger and stronger LMs continue to appear, our understanding of scaling laws continues to mature. This understanding helps researchers predict the performance of LLMs during training and estimate the computational resources required to achieve a given level of performance.

\begin{figure}[!t]
\centering
\input{./Chapter8/Figures/figure-a-curve-of-scaling-laws}
\caption{A scaling law of test error against a variable of interest (e.g., training dataset size)~\cite{hestness-etal:2017deep}. The curve of the scaling law can be divided into three phases. Initially, the test error decreases slowly when more training data is used, but this only lasts for a short period. In the second phase, the test error decreases drastically, and the curve becomes a power-law curve. After that, the error reduction slows down again in the third phase. Note that there are irreducible errors that cannot be eliminated, regardless of the amount of training data.}
\label{fig:scaling-power-law-curve}
\end{figure}

To understand how model performance scales with various factors involved in training, it is common to express model performance as a function of these factors. For example, in the simplest case, we can express the loss or error of an LLM as a function of a single variable of interest. However, there are no universal scaling laws that can describe this relationship. Instead, different functions are proposed to fit the learning curves of LLMs.

Let $x$ be the variable of interest (such as the number of model parameters) and $\mathcal{L}(x)$ be the loss of the model given $x$ (such as the cross-entropy loss on test data). The simplest form of $\mathcal{L}(x)$ is a power law
\begin{eqnarray}
\mathcal{L}(x) & = & ax^b,
\end{eqnarray}

\noindent where $a$ and $b$ are parameters that are estimated empirically. Despite its simplicity, this function has successfully interpreted the scaling behavior of language models and machine translation systems in terms of model size (denoted by $N$) and training dataset size (denoted by $D$)~\cite{gordon-etal:2021data,hestness-etal:2017deep}. For example,~\citet{kaplan-etal:2020scaling} found that the performance of their language model improves as a power law of either $N$ or $D$ after an initial transient period, and expressed these relationships using $\mathcal{L}(N) = \big( \frac{N}{8.8 \times 10^{13}} \big)^{-0.076}$ and $\mathcal{L}(D) = \big( \frac{D}{5.4 \times 10^{13}} \big)^{-0.095}$ (see Figure~\ref{fig:power-law-scaling-for-model-size-and-dataset-size}).

\begin{figure}[!t]
\centering
\input{./Chapter8/Figures/figure-model-performance-against-model-size-and-training-dataset-size}
\caption{Test loss against model size ($N$) and training dataset size ($D$) (data points are plotted for illustrative purposes). We plot test loss as a function of $N$, which is defined as $\mathcal{L}(N) = \big( \frac{N}{8.8 \times 10^{13}} \big)^{-0.076}$, and a function of $D$, which is defined as $\mathcal{L}(D) = \big( \frac{D}{5.4 \times 10^{13}} \big)^{-0.095}$ ~\cite{kaplan-etal:2020scaling}. }
\label{fig:power-law-scaling-for-model-size-and-dataset-size}
\end{figure}

An improvement to this scaling law is to add an \mindex{irreducible error} term to the power law. The form of $\mathcal{L}(x)$ is then given by

\begin{eqnarray}
\mathcal{L}(x) & = & ax^b + \epsilon_{\infty} \label{eq:scaling-laws-m2-form},
\end{eqnarray}

\noindent where $\epsilon_{\infty}$ is the irreducible error that accounts for the error due to unknown variables, which is present even as $x \to \infty$. Eq. (\ref{eq:scaling-laws-m2-form}) is one of the most widely used forms for modeling scaling laws of LLMs. For example,~\citet{rosenfeld-etal:2020constructive} developed a scaling law that involves both model scaling and dataset scaling, as follows
\begin{eqnarray}
\mathcal{L}(N,D) & = & aN^b + cD^d + \epsilon_{\infty}.
\end{eqnarray}

An example of such a formulation is the Chinchilla scaling law. It states that the test loss per token is the sum of the inverse power-law functions of $N$ and $D$, with an additional irreducible error term.~\citet{hoffmann-etal:2022training} express this scaling law as
\begin{eqnarray}
\mathcal{L}(N,D) & = & \underbrace{\frac{406.4}{N^{0.34}}}_{\textrm{model scaling}} + \underbrace{\frac{410.7}{D^{0.28}}}_{\textrm{dataset scaling}} + \underbrace{1.69}_{\textrm{irreducible error}}.
\end{eqnarray}

All the scaling laws mentioned above are based on monotonic functions. So they cannot cover functions with inflection points, such as double descent curves. In response, researchers have explored more sophisticated functions to fit the learning curves. Examples of such functions can be found in~\citet{alabdulmohsin-etal:2022revisiting} and~\citet{caballero-etal:2023broken}'s work.

The significance of scaling laws lies in providing directional guidance for LLM research: if we are still in the region of the power-law curve, using more resources to train larger models is a very promising direction. While this result compels large research groups and companies to invest heavily in computational resources to train larger models, which is very expensive, scaling laws continuously push the boundaries of AI forward. On the other hand, understanding scaling laws helps researchers make informed decisions when training LLMs. For example, given the computational resources at hand, the ultimate performance of an LLM can be estimated.

As a final note on scaling laws in this section, for LLMs, a lower test loss does not always imply better performance on all downstream tasks. To adapt LLMs, there are several steps, such as fine-tuning and prompting, that may influence the final results. Therefore, the scaling laws for different downstream tasks might vary in practice.

%----------------------------------------------------------------------------------------
% NEW SECTION
%----------------------------------------------------------------------------------------

\section{Long Sequence Modeling}
\label{sec:long-sequence-modeling}

\noindent We have already seen that, in large-scale training, more powerful language models can be developed by scaling up both data and computational resources. However, scaling can also occur along other dimensions. For instance, in many modern applications, LLMs often need to be adapted to process very long sequences. A common scenario involves pre-training an LLM on texts of standard length, and subsequently applying it to sequences that far exceed those seen during pre-training. Here, we use $\Pr(\mathbf{y}|\mathbf{x})$ to denote the conditional probability of generating text, where $\mathbf{x}$ is the context and $\mathbf{y}$ is the generated text. Broadly speaking, long sequence modeling problems can be categorized into three types:

\begin{itemize}
\item \vspace{0.5em} \textbf{Long-context text generation} (i.e., where $\mathbf{x}$ is a long sequence). For example, generating a short summary for a lengthy document.
\item \vspace{0.3em} \textbf{Long-text generation} (i.e., where $\mathbf{y}$ is a long sequence). For example, generating a comprehensive story based on a few keywords.
\item \vspace{0.3em} \textbf{Long-context and long-text generation} (i.e., where both $\mathbf{x}$ and $\mathbf{y}$ are long sequences). For example, generating a translation of a long document from Chinese to English.
\end{itemize}
\vspace{0.5em}

Recently, NLP researchers have been more interested in applying and evaluating LLMs on tasks where extremely long input texts are involved. Imagine an LLM that reads a C++ source file containing tens of thousands of lines, and outlines the functionality of the program corresponding to the source file. Such models, capable of handling extensive textual contexts, are typically called \mindex{long-context LLMs}. In this section we will restrict ourselves to long-context LLMs, but the methods discussed here can be applied to other problems.

For Transformers, dealing with long sequences is computationally expensive, as the computational cost of self-attention grows quadratically with the sequence length. This makes it infeasible to train and deploy such models for very long inputs. Two strands of research have tried to adapt Transformers to long-context language modeling.

\begin{itemize}
\item \vspace{0.5em} The first explores efficient training methods and model architectures to learn self-attention models from long-sequence data.
\item \vspace{0.3em} The other adapts pre-trained LLMs to handle long sequences with modest or no fine-tuning.
\end{itemize}
\vspace{0.5em}

Here, we will discuss the former briefly  since it can be found in general discussions of efficient Transformer architectures \cite{Tay-etal:2020efficient,xiao-and-zhu:2023introduction}. We will focus on the latter, highlighting popular methods in recent LLMs. We will also discuss the strengths and limitations of these long-sequence models.

\subsection{Hardware and Parallel Computing Optimizations}

\noindent We begin our discussion by considering improvements to standard Transformer models from the perspective of high-performance computing. Most of these improvements, though not specifically designed for LLMs, have been widely applied across various deep learning models~\cite{kim-etal:2023full}. A commonly used approach is to adopt reduced-precision data formats. For example, rather than using standard 32-bit floating-point (FP32) arithmetic, models can employ 16-bit floating-point (FP16/BF16) or even 8-bit integer (INT8) quantization. Using these low-precision data types can improve computational efficiency and memory throughput, so that longer sequences can be processed more easily. An alternative approach is to improve Transformers by using hardware-aware techniques. For example, on modern GPUs, the efficiency of Transformers can be improved by using IO-aware implementations of the self-attention function~\cite{dao-etal:2022flashattention,kwon-etal:2023efficient}.

Another way to handle long sequences is through sequence parallelism~\cite{li-etal:2023sequence,korthikanti-etal:2023reducing}. Specifically, consider the general problem of attending the query $\mathbf{q}_i$ at the position $i$ to the keys $\mathbf{K}$ and values $\mathbf{V}$. We can divide $\mathbf{K}$ along the sequence dimension (i.e., by rows) and obtain a set of sub-matrices $\{\mathbf{K}^{[1]},\cdots,\mathbf{K}^{[n_u]}\}$, each corresponding to a segment of the sequence. Similarly, we can obtain the sub-matrices of $\mathbf{V}$, denoted by $\{\mathbf{V}^{[1]},\cdots,\mathbf{V}^{[n_u]}\}$. Then, we assign each pair of $\mathbf{K}^{[u]}$ and $\mathbf{V}^{[u]}$ to a compute node (e.g., a GPU of a GPU cluster). The assigned nodes can run in parallel, thereby parallelizing the attention operation.

Recall that the output of the self-attention model can be written as
\begin{eqnarray}
\mathrm{Att}_{\mathrm{qkv}}(\mathbf{q}_i,\mathbf{K},\mathbf{V}) & = & \sum_{j=0}^{m-1} \alpha_{i,j} \mathbf{v}_j \label{eq:seq-parallel-qkv},
\end{eqnarray}

\noindent where $\alpha_{i,j}$ is the attention weight between positions $i$ and $j$. In Transformers, $\alpha_{i,j}$ is obtained by normalizing the scaled dot product between $\mathbf{q}_i$ and $\mathbf{k}_j$. Let $\beta_{i,j}$ denote the attention score between $\mathbf{q}_i$ and $\mathbf{k}_j$. We have
\begin{eqnarray}
\beta_{i,j} & = & \frac{\mathbf{q}_i \cdot \mathbf{k}_j}{\sqrt{d}}  + \mathrm{Mask}(i,j),
\end{eqnarray}

\noindent where $\mathrm{Mask}(i,j)$ is the causal mask value for the position pair $(i,j)$. Then, we define the attention weight $\alpha_{i,j}$ to be

\begin{eqnarray}
\alpha_{i,j} & = & \mathrm{Softmax}(\beta_{i,j}) \nonumber \\
             & = & \frac{\exp(\beta_{i,j})}{\sum_{j'} \exp(\beta_{i,j'})} \label{eq:seq-parallel-att-weight}.
\end{eqnarray}

On each compute node, we need to implement these equations. Given the keys and values assigned to this node, computing the numerator of the right-hand side of Eq. (\ref{eq:seq-parallel-att-weight}) (i.e., $\exp(\beta_{i,j})$) is straightforward, as all the required information is stored on the node. However, computing the denominator of the right-hand side of Eq. (\ref{eq:seq-parallel-att-weight}) involves a sum of $\exp(\beta_{i,j'})$ over all $j'$, which requires transferring data to and from other nodes. To illustrate, suppose that $\mathbf{v}_j$ and $\mathbf{k}_j$ are placed on node $u$. We can rewrite Eq. (\ref{eq:seq-parallel-att-weight}) as
\begin{eqnarray}
&   & \alpha_{i,j} \nonumber \\
& = & \frac{\overbrace{\exp(\beta_{i,j})}^{\textrm{node $u$}}}{\underbrace{\sum_{\mathbf{k}_{j'} \in \mathbf{K}^{[1]}} \exp(\beta_{i,j'}) }_{\textrm{node $1$}} + \cdots + \underbrace{\sum_{\mathbf{k}_{j'} \in \mathbf{K}^{[u]}} \exp(\beta_{i,j'}) }_{\textrm{node $u$}} + \cdots + \underbrace{\sum_{\mathbf{k}_{j'} \in \mathbf{K}^{[n_u]}} \exp(\beta_{i,j'}) }_{\textrm{node $n_u$}} } \label{eq:seq-parallel-att-weight-distributed},
\end{eqnarray}

\noindent where the notation $\mathbf{k}_{j'} \in \mathbf{K}^{[u]}$ indicates that $\mathbf{k}_{j'}$ is a row vector of $\mathbf{K}^{[u]}$. In a straightforward implementation, we first perform the summations $\{ \sum_{\mathbf{k}_{j'} \in \mathbf{K}^{[u]}} \exp(\beta_{i,j'}) \}$ separately on the corresponding nodes. Then, we collect these summation results from different nodes to combine them into a final result. This corresponds to a collective operation in the context of parallel processing. There are many efficient implementations of such operations, such as all-reduce operations. Hence the sum of all $\exp(\beta_{i,j})$ values can be computed using optimized routines in collective communication toolkits.

Given the attention weights $\{\alpha_{i,j}\}$, we then compute the attention results using Eq. (\ref{eq:seq-parallel-qkv}). The problem can be re-expressed as
\begin{eqnarray}
&   & \mathrm{Att}_{\mathrm{qkv}}(\mathbf{q}_i,\mathbf{K},\mathbf{V}) \nonumber \\
& = & \underbrace{\sum_{\mathbf{v}_{j'} \in \mathbf{V}^{[1]}} \alpha_{i,j'} \mathbf{v}_{j'} }_{\textrm{node $1$}} + \cdots + \underbrace{\sum_{\mathbf{v}_{j'} \in \mathbf{V}^{[u]}} \alpha_{i,j'} \mathbf{v}_{j'} }_{\textrm{node $u$}} + \cdots + \underbrace{\sum_{\mathbf{v}_{j'} \in \mathbf{V}^{[n_u]}} \alpha_{i,j'} \mathbf{v}_{j'} }_{\textrm{node $n_u$}} \label{eq:seq-parallel-qkv-distributed}.
\end{eqnarray}

Like Eq. (\ref{eq:seq-parallel-att-weight-distributed}), Eq. (\ref{eq:seq-parallel-qkv-distributed}) can be implemented as a reduction operation in parallel processing. First, we perform the weighted summations of values on different nodes simultaneously. Then, we collect the results from these nodes via collective operations.

Note that, although this section primarily focuses on long sequence modeling, much of the motivation for sequence parallelism comes from the distributed training methods of deep networks, as discussed in Section~\ref{sec:llm-distributed-training}. As a result, the implementation of these methods can be built on the same parallel processing library.

\subsection{Efficient Architectures}

\noindent One difficulty of applying Transformers to long sequences is that standard self-attention suffers from quadratic time and memory complexity with respect to the sequence length.  Moreover, autoregressive generation requires maintaining a \mindex{key-value cache} (or \mindex{KV cache}), whose size increases as more tokens are processed. Although the KV cache footprint grows linearly with sequence length, for extremely long contexts, this memory requirement becomes prohibitive, making standard LLM deployment impractical. As a result, the model architecture of long-context LLMs generally moves away from the standard Transformer, turning instead to the development of more efficient variants and alternatives.

One approach is to replace standard self-attention with sparse attention. This family of models is motivated by the observation that attention distributions in long sequences are often highly skewed. Thus, only a small subset of tokens significantly impacts the context of any given token. Consequently, we can prune the vast majority of attention weights, representing the mechanism in a highly compressed form. To illustrate, consider the standard self-attention formulation:
\begin{eqnarray}
\mathrm{Att}_{\mathrm{qkv}}(\mathbf{Q},\mathbf{K},\mathbf{V}) & = & \alpha(\mathbf{Q},\mathbf{K}) \mathbf{V} \label{eq:qkv-attention-attention-weights},
\end{eqnarray}

\noindent where the attention weight matrix $\alpha(\mathbf{Q},\mathbf{K}) \in \mathbb{R}^{m \times m}$ is obtained by
\begin{eqnarray}
\alpha(\mathbf{Q},\mathbf{K}) & = & \mathrm{Softmax}(\frac{\mathbf{Q} \mathbf{K}^\top}{\sqrt{d}} + \mathbf{Mask}) \nonumber \\
                              & = & \begin{bmatrix} \alpha_{0,0} & 0 & 0 & \cdots & 0 \\ \alpha_{1,0} & \alpha_{1,1} & 0 & \cdots & 0 \\ \alpha_{2,0} & \alpha_{2,1} & \alpha_{2,2} & \cdots & 0 \\ \vdots & \vdots & \vdots & \ddots & \vdots \\ \alpha_{m-1,0} & \alpha_{m-1,1} & \alpha_{m-1,2} & \cdots & \alpha_{m-1,m-1} \end{bmatrix}.
\end{eqnarray}

Each row vector $\begin{bmatrix} \alpha_{i,0} & \cdots & \alpha_{i,i} & 0 & \cdots & 0 \end{bmatrix}$ corresponds to a distribution over tokens indicating how the $i$-th token attends to others. Since language models predict the next token only based on their left-context, we normally write the output of the attention model at position $i$ as
\begin{eqnarray}
\mathrm{Att}_{\mathrm{qkv}}(\mathbf{q}_i,\mathbf{K}_{\le i},\mathbf{V}_{\le i}) & = & \begin{bmatrix} \alpha_{i,0} & \cdots & \alpha_{i,i}  \end{bmatrix} \begin{bmatrix} \mathbf{v}_0 \\ \vdots \\ \mathbf{v}_{i} \end{bmatrix} \nonumber \\
                                                                                & = & \sum_{j = 0}^{i} \alpha_{i,j} \mathbf{v}_{j},
\end{eqnarray}

\noindent where $\mathbf{K}_{\le i} = \begin{bmatrix} \mathbf{k}_0 \\ \vdots \\ \mathbf{k}_{i} \end{bmatrix}$ and $\mathbf{V}_{\le i} = \begin{bmatrix} \mathbf{v}_0 \\ \vdots \\ \mathbf{v}_{i} \end{bmatrix}$ are the keys and values up to position $i$.

In standard self-attention, the vector $\begin{bmatrix} \alpha_{i,0} & \cdots & \alpha_{i,i} \end{bmatrix}$ is assumed to be dense, meaning almost all valid entries are non-zero. In sparse attention, only a predefined subset of these entries is computed, while the remainder are explicitly ignored (forced to zero). Let $G \subseteq \{0,\cdots,i\}$ denote the index set of the active (non-zero) entries. The output of the sparse attention model at position $i$ then becomes:
\begin{eqnarray}
\mathrm{Att}_{\mathrm{sparse}}(\mathbf{q}_i,\mathbf{K}_{\le i},\mathbf{V}_{\le i}) & = & \sum_{j \in G} \alpha'_{i,j} \mathbf{v}_{j}.
\end{eqnarray}

\noindent Here $\{\alpha'_{i,j}\}$ are normalized over $G$. Hence their values are different from the original attention weights (in fact we have $\alpha'_{i,j} > \alpha_{i,j}$). The sparsity of the model is determined by the size of $G$. Sparse attention models differ in the way we define $G$. One simple approach is to define $G$ based on heuristically designed patterns. For example, a widely-used pattern involves having $G$ cover a window of tokens located near position $i$~\cite{parmar-etal:2018image}.

While sparse attention reduces the computation through the use of sparse operations, such models still have significant limitations as we need to keep the entire KV cache (i.e., $\mathbf{K}_{\le i}$ and $\mathbf{V}_{\le i}$) during inference. If the sequence is very long, storing this cache will become highly memory-intensive. To address this, we can consider a different class of attention models where the KV cache is not explicitly retained. Linear attention is one such approach~\cite{katharopoulos-etal:2020transformers}. It uses a kernel function $\phi(\cdot)$ to map each query and key into a feature space, yielding $\mathbf{q}'_i = \phi(\mathbf{q}_i)$ and $\mathbf{k}'_i = \phi(\mathbf{k}_i)$, respectively. By replacing the Softmax normalization with a linear formulation\footnote{In the new space after this transformation, the Softmax normalization can be transformed into the simple scaling normalization.}, the form of the resulting attention model is given by

\begin{eqnarray}
\mathrm{Att}_{\mathrm{qkv}}(\mathbf{q}_i,\mathbf{K}_{\le i},\mathbf{V}_{\le i}) & \approx & \mathrm{Att}_{\mathrm{linear}}(\mathbf{q}'_i,\mathbf{K}'_{\le i},\mathbf{V}_{\le i}) \nonumber \\
                                                                                & = & \frac{\mathbf{q}'_{i} \mu_i}{\mathbf{q}'_{i} \nu_i},
\end{eqnarray}

\noindent where $\mu_i$ and $\nu_i$ are variables that are computed in a recurrent manner

\begin{eqnarray}
\mu_i & = & \mu_{i-1} + {\mathbf{k}'}_{i}^\top \mathbf{v}_{i} \label{eq:linear-attention-mu-llm} \\
\nu_i & = & \nu_{i-1} + {\mathbf{k}'}_{i}^\top \label{eq:linear-attention-nu-llm}.
\end{eqnarray}

\noindent $\mu_i$ and $\nu_i$ can be seen as summaries of the history up to position $i$. A benefit of this model is that we do not need to explicitly store all past keys and values. Instead only the latest representations $\mu_i$ and $\nu_i$ are used. So the computational cost of each step is a constant, and the model can be easily extended to deal with long sequences.

In fact, this sequential approach to long sequence modeling arises naturally when we adopt the perspective of recurrent models. Such models read one token (or a small number of tokens) at a time, update the recurrent state using these inputs, and then discard them before the next token arrives. The output at each step is generated based only on the recurrent state, rather than on all the previous states. The memory footprint is determined by the recurrent state which has a fixed size. Recurrent models can be used in real-time learning scenarios where data arrives in a stream and predictions can be made at any time step. In NLP, applying recurrent models to language modeling is  one of the earliest successful attempts to learn representations of sequences. Although Transformers have been used as the foundational architecture in LLMs, recurrent models are still powerful models, especially for developing efficient LLMs. For example, recurrent models, particularly modern state space models like Mamba~\cite{gu-and-dao:2023mamba}, have seen a resurgence in language modeling and have been reconsidered as a promising alternative to Transformers.

Figure~\ref{fig:comparison-of-efficient-llm-architectures} shows a comparison of the models discussed in this subsection. Since these models, along with others not mentioned here, have been extensively discussed in related surveys~\cite{Tay-etal:2020efficient,xiao-and-zhu:2023introduction}, a more exhaustive discussion is omitted here.

\begin{figure}[!t]
\centering
\input{./Chapter8/Figures/figure-long-context-llm-architectures}
\caption{Illustrations of self-attention, sparse attention, linear attention and recurrent models. Blue boxes = cached states for producing the output at position $i$. $f(\cdot)$ = a recurrent cell.}
\label{fig:comparison-of-efficient-llm-architectures}
\end{figure}

\subsection{Cache and Memory}
\label{sec:long-context-cache-and-memory}

\noindent LLMs based on the standard Transformer architecture employ global (dense) attention mechanisms. During autoregressive inference, these models require storing the entire left-context to make predictions for future tokens. This requires a KV cache, where the representations (i.e., keys and values) of all previously generated tokens are retained. As a result, the memory footprint of this cache grows linearly as inference proceeds. In the preceding sections, we discussed methods for mitigating this context bottleneck via efficient attention algorithms, such as sparse and linear attention. A related alternative approach is to explicitly encode and compress the context via an auxiliary memory module.

\subsubsection{Fixed-size KV Cache}

A straightforward approach is to represent the keys and values using a fixed-size memory model. Suppose we have a memory $\mathrm{Mem}$ which retains the contextual information. We can write the attention operation at position $i$ in a general form
\begin{eqnarray}
\mathrm{Att}(\mathbf{q}_i,\mathrm{Mem}) & = & \mathrm{Att}_{\mathrm{qkv}}(\mathbf{q}_i,\mathbf{K}_{\le i},\mathbf{V}_{\le i}).
\end{eqnarray}

In this model, $\mathrm{Mem}$ is simply the KV cache, i.e., $\mathrm{Mem} = (\mathbf{K}_{\le i}, \mathbf{V}_{\le i})$. Thus the size of $\mathrm{Mem}$ is determined by $i$. If $\mathrm{Mem}$ is defined to have a fixed size, then the cost of performing $\mathrm{Att}(\mathbf{q}_i,\mathrm{Mem})$ will be fixed. There are several alternative ways to design $\mathrm{Mem}$.

\begin{itemize}
\item \vspace{0.5em} One of the simplest methods is to consider a fixed-size window of previous keys and values. $\mathrm{Mem}$ is therefore given by
    \begin{eqnarray}
    \mathrm{Mem} & = & (\mathbf{K}_{[i - n_c + 1,i]},\mathbf{V}_{[i - n_c + 1,i]}),
    \end{eqnarray}
    \noindent where $n_c$ denotes the size of the window. The notation $\mathbf{K}_{[i - n_c + 1,i]}$ and $\mathbf{V}_{[i - n_c + 1,i]}$ denote the keys and values over positions from $i-n_c+1$ to $i$.\footnote{More formally, we write $\mathbf{K}_{[i - n_c + 1,i]} = \begin{bmatrix} \mathbf{k}_{i - n_c + 1} \\ \vdots \\ \mathbf{k}_{i} \end{bmatrix}$ and $\mathbf{V}_{[i - n_c + 1,i]} = \begin{bmatrix} \mathbf{v}_{i - n_c + 1} \\ \vdots \\ \mathbf{v}_{i} \end{bmatrix}$. Sometimes we denote $\mathbf{K}_{[i - n_c + 1,i]}$ by $\{\mathbf{k}_{i - n_c + 1},\cdots,\mathbf{k}_{i}\}$ and $\mathbf{V}_{[i - n_c + 1,i]}$ by $\{\mathbf{v}_{i - n_c + 1},\cdots,\mathbf{v}_{i}\}$ for notational simplicity.} This model can be seen as a type of local attention model.
\item \vspace{0.5em} It is also possible to define $\mathrm{Mem}$ as a pair of summary vectors, which leads to a more compressed representation of the history. A simple way to summarize the previous keys and values is to use the moving averages of these vectors. For example, $\mathrm{Mem}$ can be defined as the unweighted moving average of the previous $n_c$ keys and values
    \begin{eqnarray}
    \mathrm{Mem} & = & \left(\frac{\sum_{j=i - n_c + 1}^{i} \mathbf{k}_{j}}{n_c} , \frac{\sum_{j=i - n_c + 1}^{i} \mathbf{v}_{j}}{n_c} \right).
    \end{eqnarray}

    Alternatively, we can use a weighted moving average
    \begin{eqnarray}
    \mathrm{Mem} & = & \left( \frac{\sum_{j=i - n_c + 1}^{i} \beta_{j - i + n_c} \mathbf{k}_{j}} {\sum_{j=1}^{n_c} \beta_j} , \frac{\sum_{j=i - n_c + 1}^{i} \beta_{j - i + n_c} \mathbf{v}_{j}}{\sum_{j=1}^{n_c} \beta_j} \right).
    \end{eqnarray}

    \noindent Here $\{\beta_1,\cdots,\beta_{n_c}\}$ are the coefficients, which can either be learned as model parameters or determined via heuristics. For example, they can be set to increasing coefficients (i.e., $\beta_{1} < \beta_{2} < \cdots < \beta_{n_c -1} < \beta_{n_c}$) in order to give larger weight to positions that are closer to $i$. We can extend the moving average to include all the positions up to $i$. This leads to the cumulative average of the keys and values, given in the form
    \begin{eqnarray}
    \mathrm{Mem} & = & \left( \frac{\sum_{j=0}^{i} \mathbf{k}_j}{i+1}, \frac{\sum_{j=0}^{i} \mathbf{v}_j}{i+1} \right).
    \end{eqnarray}

    In general, the cumulative average can be written in recursive form
    \begin{eqnarray}
    \mathrm{Mem}_{i} & = & \frac{(\mathbf{k}_i,\mathbf{v}_i) + i \cdot \mathrm{Mem}_{i-1}}{i + 1},
    \end{eqnarray}

    \noindent where $\mathrm{Mem}_{i}$ and $\mathrm{Mem}_{i-1}$ denote the cumulative averages of the current and previous positions, respectively. An advantage of this model is that we only need to store a single key-value pair during inference, rather than storing all the key-value pairs. Note that the above memory models are related to recurrent models, and more advanced techniques have been used to develop alternatives to self-attention mechanisms in Transformers~\cite{ma-etal:2023mega}.

\item \vspace{0.3em} The memory $\mathrm{Mem}$ can also be a neural network. At each step, it takes both the previous output of the memory and the current states of the model as input, and produces the new output of the memory. This neural network can be formulated as the function
    \begin{eqnarray}
    \mathrm{Mem} & = & \mathrm{Update}(S_{\mathrm{kv}}, \mathrm{Mem}_{\mathrm{pre}}) \label{eq:long-seq-mem-update}.
    \end{eqnarray}
    \noindent Here $\mathrm{Mem}$ and $\mathrm{Mem}_{\mathrm{pre}}$ represent the outputs of the memory at the current step and the previous step, respectively. $S_{\mathrm{kv}}$ is a set of key-value pairs, representing the recent states of the model. This formulation is general and allows us to develop various memory models by selecting different $\mathrm{Update}(\cdot)$ and $S_{\mathrm{kv}}$ configurations. For example, if $S_{\mathrm{kv}}$ only contains the latest key-value pair $(\mathbf{k}_i,\mathbf{v}_i)$ and $\mathrm{Update}(\cdot)$ is defined as a recurrent cell, then Eq. (\ref{eq:long-seq-mem-update}) can be expressed as an RNN-like model
    \begin{eqnarray}
    \mathrm{Mem} & = & f((\mathbf{k}_i,\mathbf{v}_i), \mathrm{Mem}_{\mathrm{pre}}),
    \end{eqnarray}

    \noindent where $f(\cdot)$ is a recurrent cell. Recurrence can also be applied to segment-level modeling for efficiency considerations. A simple approach is to divide the sequence into segments, and treat $S_{\mathrm{kv}}$ as a segment. Applying recurrent models to $\mathrm{Update}(\cdot)$ will result in memory models that operate on segments. A special case is to define $\mathrm{Update}(\cdot)$ as an FIFO function that adds $S_{\mathrm{kv}}$ into the memory and removes the oldest key-value segment from the memory, given by
    \begin{eqnarray}
    \mathrm{Mem} & = & \mathrm{FIFO}(S_{\mathrm{kv}}, \mathrm{Mem}_{\mathrm{pre}}).
    \end{eqnarray}

    \noindent Consider a memory which includes two segments, one for current segment, and one for the previous segment. In the attention operation, each position can access the history key-value pairs in the two most recent consecutive segments. While conceptually simple, this sliding-chunk local memory and its variants have been widely adopted in segment-level recurrent architectures~\cite{dai-etal:2019transformer,hutchins-etal:2022block,bulatov-etal:2022recurrent}.

\item \vspace{0.3em} The above memory models can be extended to involve multiple memories. An example of this approach is compressive Transformer~\cite{rae-etal:2020:compressive}. It employs two distinct fixed-size memories: one for modeling local context (denoted by $\mathrm{Mem}$), and the other for modeling and compressing long-term history (denoted by $\mathrm{CMem}$). The KV cache in this model is the combination of $\mathrm{Mem}$ and $\mathrm{CMem}$. The attention function can be written as
    \begin{eqnarray}
    \mathrm{Att}_{\mathrm{com}}(\mathbf{q}_i,\mathrm{Mem},\mathrm{CMem}) & = & \mathrm{Att}_{\mathrm{qkv}}(\mathbf{q}_i, [\mathrm{Mem},\mathrm{CMem}]),
    \end{eqnarray}

    \noindent where $[\mathrm{Mem},\mathrm{CMem}]$ is a combined memory of $\mathrm{Mem}$ and $\mathrm{CMem}$. Similar to standard segment-level models, the compressive Transformer processes the sequence in blocks. Each segment comprises $n_s$ consecutive tokens, and $S_{\mathrm{kv}}^{k}$ denotes the raw key-value pairs generated from the $k$-th segment. Upon the arrival of a new segment, $\mathrm{Mem}$ is updated via a FIFO queue: it appends the $n_s$ new key-value pairs from $S_{\mathrm{kv}}^{k}$ and simultaneously removes the $n_s$ oldest pairs:
    \begin{eqnarray}
    \mathrm{Mem} & = & \mathrm{FIFO}(S_{\mathrm{kv}}^{k}, \mathrm{Mem}_{\mathrm{pre}}).
    \end{eqnarray}

    \noindent Rather than discarding these ejected pairs, they are passed through a compression network. The $n_s$ raw pairs are compressed into a smaller set of $\frac{n_s}{c}$ condensed key-value pairs. Then $\mathrm{CMem}$ serves as a secondary FIFO queue, appending these $\frac{n_s}{c}$ compressed pairs and discarding its oldest $\frac{n_s}{c}$ entries to maintain fixed size. It is given by
    \begin{eqnarray}
    \mathrm{CMem}& = & \mathrm{FIFO}(C_{\mathrm{kv}}^{k}, \mathrm{CMem}_{\mathrm{pre}}),
    \end{eqnarray}

    \noindent where $C_{\mathrm{kv}}^{k}$ represents the set of compressed key-value pairs. The compressive Transformer implicitly assumes that local context should be represented explicitly with minimal  information loss, while long-range context can be more compressed.

\item \vspace{0.3em} We have already seen that both global and local contexts are useful and can be modeled using attention models. This view motivates the extension to attention models for combining both local and long-term memories~\cite{ainslie-etal:2020etc,zaheer-etal:2020big,gupta-berant:2020gmat}. A simple but widely-used approach is to involve the first few tokens of the sequence in attention, which serves as global tokens. This approach is usually applied along with other sparse attention models. An advantage of incorporating global tokens is that it serves as an attention sink that absorbs excessive attention scores, and thus stabilizes model performance when the context size is very large~\cite{xiao-etal:2024efficient}. One drawback, however, is that using a fixed-size global memory may result in information loss. When dealing with long sequences, we need to enlarge the KV cache for sufficient representations of the context, but this in turn increases the computational cost.

\end{itemize}
\vspace{0.5em}

Figure~\ref{fig:comparison-kv-cache-optimizations} shows illustrations of the above approaches. Note that, while we focus on optimization of the KV cache here, this issue is closely related to those discussed in the previous section. All of the methods we have mentioned so far can broadly be categorized as efficient attention approaches, which are widely used in various Transformer variants.

\begin{figure}[!t]
\centering
\input{./Chapter8/Figures/figure-fixed-size-memory}
\caption{Illustrations of fixed-size KV caches in LLMs. Blue boxes represent the keys and values generated during LLM inference, green boxes represent the keys and values stored or encoded in the primary memory, and orange boxes represent the keys and values stored or encoded in the compressed memory.}
\label{fig:comparison-kv-cache-optimizations}
\end{figure}

\subsubsection{Memory-based Models}

\noindent The memory modeling discussed above is based on updates to the KV cache, and the resulting models are typically referred to as \mindex{internal memories}. We now consider another family of models, called \mindex{external memories}, which operate independently to access large-scale contexts for LLMs. Many of these approaches are rooted in \mindex{memory-based methods}, a concept extensively studied in classical machine learning~\cite{Bishop:2006}. A common example is nearest-neighbor algorithms: we store context representations in a datastore and query it to find the most similar stored representations. These retrieved context representations are then incorporated to enhance the attention mechanism for the current query.

Here, we consider the \mindexraw{$k$-nearest neighbors} (\mindex{$k$-NN}) method which is one of the most popular memory-based methods. Since our focus is language modeling in this section, we define a sample in the datastore as a key-value pair corresponding to some context state. Note that ``context'' is a broad concept here, not just a sequence prefix in text generation. One might, for example, view the entire dataset as the context for predicting tokens. This allows us to retrieve the most similar context in a set of sequences, rather than a given sequence prefix. Although we will restrict ourselves to context modeling for a single sequence, in this subsection, we discuss a more general case.

Suppose we have a set of keys $\{\mathbf{k}_j\}$ with corresponding values $\{\mathbf{v}_j\}$, and suppose we store these key-value pairs in a vector database\footnote{A vector database, or vector store, is a database that provides highly optimized retrieval interfaces for finding stored vectors that closely match a query vector.}. For each query $\mathbf{q}_{i}$, we find its $k$ nearest neighbors by growing the radius of the sphere centered at $\mathbf{q}_i$ until it contains $k$ data points in $\{\mathbf{k}_j\}$. This results in a set of $k$ keys along with their corresponding values, denoted by $\mathrm{Mem}_{k\mathrm{nn}}$.  As before, we denote $\mathrm{Mem}$ as the local memory for the query, such as the KV cache of neighboring tokens. Our goal is to allow the query $\mathbf{q}_i$ to attend to both the local memory $\mathrm{Mem}$ and the long-term memory $\mathrm{Mem}_{k\mathrm{nn}}$. There are, of course, several ways to incorporate $\mathrm{Mem}$ and $\mathrm{Mem}_{k\mathrm{nn}}$ into the attention model. For example, we might simply combine them to form a single KV cache $[\mathrm{Mem},\mathrm{Mem}_{k\mathrm{nn}}]$, and attend $\mathbf{q}_i$ to $[\mathrm{Mem},\mathrm{Mem}_{k\mathrm{nn}}]$ via standard QKV attention. Or we might use $\mathrm{Mem}$ and $\mathrm{Mem}_{k\mathrm{nn}}$ in separate attention steps. An example of such an approach is the model developed by~\citet{wu-etal:2021memorizing}. It linearly combines the two types of attention, given by
\begin{eqnarray}
\mathrm{Att}(\mathbf{q}_i,\mathrm{Mem},\mathrm{Mem}_{k\mathrm{nn}}) & = & \mathbf{g} \odot \mathrm{Att}_{\mathrm{local}} + (1 - \mathbf{g}) \odot \mathrm{Att}_{k\mathrm{nn}} \\
\mathrm{Att}_{\mathrm{local}} & = & \mathrm{Att}(\mathbf{q}_i,\mathrm{Mem}) \\
\mathrm{Att}_{k\mathrm{nn}} & = & \mathrm{Att}(\mathbf{q}_i,\mathrm{Mem}_{k\mathrm{nn}}).
\end{eqnarray}

\noindent Here $\mathbf{g} \in \mathbb{R}^{d}$ is the coefficient vector, which can be the output of a learned gate.

Given the $k$-NN-based memory model described above, the remaining task is to determine which key-value pairs are retained in the datastore. For standard language modeling tasks, we consider the previously seen tokens in a sequence as the context, so we can add the keys and values of all these tokens into the datastore. In this case, the resulting $k$-NN-based attention model is essentially equivalent to a sparse attention model~\cite{gupta-etal:2021memory}.

Alternatively, we can extend the context from one sequence to a collection of sequences. For example, we might collect all key-value pairs across the sequences in a training dataset and add them to the datastore to model a larger context. Thus, LLMs can predict tokens based on a generalized context. A problem with this approach is that the computational cost can be large if many sequences are involved. Since these sequences are part of our training data, we can build and optimize an index for the vectors in the datastore before running the LLMs. As a result, the retrieval of similar vectors can be very efficient, as commonly implemented in modern vector databases.

In fact, all the aforementioned methods represent instances of the broader retrieval-augmented paradigm. Beyond directly injecting retrieved vectors into the attention mechanism, retrieval can be applied at the output distribution level. A typical application is \mindexraw{$k$-NN language modeling} (\mindexraw{$k$-NN LM})~\cite{khandelwal-etal:2020generalization}. The intuition is that similar hidden states within a Transformer network are highly predictive of similar next tokens across different contexts. In a $k$-NN LM, each item in the datastore is stored as a tuple $(\mathbf{z},w)$, where $\mathbf{z}$ represents the final hidden state of the LLM at a specific position, and $w$ represents the actual next token observed at that position. Typically, this datastore is constructed offline by passing the entire training dataset through the LLM.

During inference, given a sequence prefix, the LLM generates a current hidden state $\mathbf{h}_i$. We query the datastore using $\mathbf{h}_i$ to retrieve the $k$ closest matching states, yielding a set of tuples $\{(\mathbf{z}_1,w_1),\cdots,(\mathbf{z}_k,w_k)\}$. The target tokens $\{w_1,\cdots,w_k\}$ serve as reference outputs. To utilize these references, we convert the retrieval distances into a probability distribution over the vocabulary $V$. Let $d(\mathbf{h}_i, \mathbf{z}_j)$ be the distance metric. The probability mass assigned to a token $v \in V$ by the $k$-NN retrieval is computed by aggregating the softmax-normalized similarities of the retrieved neighbors:
\begin{eqnarray}
\mathrm{Pr}_{k\mathrm{nn}}(v|\mathbf{h}_i) & \propto & \sum_{(\mathbf{z}_j, w_j) \in \text{top-}k, \atop w_j = v} \exp\left(-d(\mathbf{h}_i, \mathbf{z}_j)\right). \label{eq:knn-lm-dist}
\end{eqnarray}

\noindent If a token $v$ does not appear among the $k$ retrieved neighbors, its probability $\mathrm{Pr}_{k\mathrm{nn}}(v|\mathbf{h}_i)$ is zero. Finally, we use a tunable hyperparameter $\lambda \in [0,1]$ to linearly interpolate between this retrieval-based distribution and the standard parametric distribution $\mathrm{Pr}_{\mathrm{lm}}(\cdot|\mathbf{h}_i)$ generated by the LLM's classification head:
\begin{eqnarray}
\mathrm{Pr}(\cdot|\mathbf{h}_i) & = & \lambda \cdot \mathrm{Pr}_{k\mathrm{nn}}(\cdot|\mathbf{h}_i) + (1 - \lambda) \cdot \mathrm{Pr}_{\mathrm{lm}}(\cdot|\mathbf{h}_i).
\end{eqnarray}

\noindent Then, as usual, we can select the next token $y$ by maximizing the probability $\mathrm{Pr}(y|\mathbf{h}_i)$.

As with information retrieval (IR) systems, the datastore can store and retrieve textual data. For example, we can store a collection of text documents in a search engine with full-text indexing, and then search it for documents that match a given text-based query. Applying IR techniques to LLMs leads to a general framework called \mindex{retrieval-augmented generation} (\mindex{RAG}). The RAG framework works as follows. We use the context $\mathbf{x}$ as the query and find the $k$ most relevant document chunks $\{\mathbf{c}_{1},\cdots,\mathbf{c}_{k}\}$ from the datastore via efficient IR techniques\footnote{In practical applications, queries are typically generated using a query generation system, which may expand it with variations of tokens and query intent.}. These search results are combined with the original context via a prompting template $g(\cdot)$\footnote{For example, the template could be:

\vspace{0.3em}
\begin{tcolorbox}[frame empty]
\begin{center}
\parbox{0.9\textwidth}
{

\hspace{4em} message = \{*$\mathbf{c}_1$*\} $\cdots$ \{*$\mathbf{c}_k$*\}

\vspace{0.3em}

\hspace{4em} input: \{*$\mathbf{x}$*\}

\vspace{0.3em}

\hspace{4em} output: \underline{\hspace{2em}}

}
\end{center}
\end{tcolorbox}

}, resulting in an augmented input for the LLM
\begin{eqnarray}
\mathbf{x}' & = & g(\mathbf{c}_{1},\cdots,\mathbf{c}_{k},\mathbf{x}).
\end{eqnarray}

\begin{figure}[!t]
\centering
\input{./Chapter8/Figures/figure-knn-models-for-llms}
\caption{Illustrations of external memories (or datastores) for language modeling.}
\label{fig:comparison-external-memories}
\end{figure}

\noindent Then, we use $\mathbf{x}'$ as the context and predict the following text using the model $\Pr(\mathbf{y}|\mathbf{x}')$. One advantage of RAG is that we need not modify the architecture of LLMs, but instead augment the input to LLMs via an additional IR system. Figure~\ref{fig:comparison-external-memories} shows a comparison of the use of different external memories in LLMs.

\subsubsection{Memory Capacity}

\noindent A memory model in LLMs, whether in the form of a simple key-value cache or an external datastore, can broadly be viewed as an encoder of contextual information. Ideally, before we say that a memory model is representative of the entire context in token prediction, we must ensure that the model can accurately reconstruct any part of that context. The standard KV cache is one such model, as it retains the complete sequence history. In this case, the model is said to possess sufficient (or unbounded) capacity for memorizing the context. In many practical applications, however, complete memorization is unnecessary. Instead, the primary goal is to enable LLMs to access salient contextual information. As a result, efficient and compressed memory models have been developed, as described in this section. Note that the longer the sequence, the more difficult it becomes for a low-capacity memory model to capture crucial contextual information. It is therefore common practice to simply scale up the memory capacity when processing long contexts.

While high-capacity memory models are generally desirable, they are computationally expensive to train and deploy. A challenging scenario occurs when tokens arrive in a continuous stream, causing the context to grow indefinitely. Developing LLMs for such streaming tasks is difficult, as it requires training Transformers on extremely long sequences. One potential solution is to use non-parametric approaches, such as retrieval-based methods. For instance, as discussed previously, we can employ a vector database to store all generated key-value pairs, and thus effectively represent the context via an external memory model. Although this approach sidesteps the inherent context-length limitations of Transformers, the construction and maintenance of external memory models lead to significant computational overhead. Furthermore, these models are typically applied to problems where the context is provided in advance and remains static during inference, and are hence unsuitable for dynamic, streaming context modeling.

In cases where the size of the context continuously grows, applying fixed-size memory models is a commonly used approach. For example, in recurrent models, a sequence of arbitrary length can be summarized into a set of hidden states, resulting in a fixed computational cost per step. While recurrent models were initially found to be not very good at handling long-distance dependencies in sequence modeling in early applications of deep learning to NLP, recent advancements have shown that their variants are now effective in modeling extremely long sequences.~\cite{bulatov-etal:2022recurrent,hutchins-etal:2022block,munkhdalai-etal:2024leave,ma-etal:2024megalodon}.

There is no general definition of memory capacity in LLMs. A simple approach might consider how much storage is used to retain contextual information. For example, memory capacity could be defined by the size of the KV cache in Transformers or the vector database used in retrieval-based methods. A related concept is model complexity. In machine learning, there are several ways to define the model complexity. One of the simplest methods is by counting the number of parameters. However, it should be emphasized that the memory models discussed here primarily serve to store information, rather than add trainable parameters. Therefore, a model with a large memory capacity is not necessarily more complex. Nevertheless, in practice determining the capacity of a memory model is not straightforward. In general, we need to consider the trade-off between maximizing the performance and controlling the memory footprint.

\subsection{Sharing across Heads and Layers}

\noindent In Transformers, the KV cache is a data structure that can be dynamically adjusted along multiple dimensions, such as heads, layers, and sequence length. For example, consider an LLM with $L$ layers. Each layer has $\tau$ attention heads, and each head produces a $d_h$-dimensional output. During inference, we store the keys and values for up to $m$ tokens. The space complexity of this caching mechanism is $O(L \cdot \tau \cdot d_h \cdot m)$.  As we have seen previously, this complexity can be reduced by caching the keys and values for fewer tokens. For example, in sliding window attention, a fixed-size window is used to cache the keys and values in local context. The sliding-window model has a space complexity of $O(L \cdot \tau \cdot d_h \cdot m_w)$, with $m_w$ being the size of the window.

In addition to reducing $m$, we can also decrease the size of the KV cache along other dimensions. A widely-used approach is to enable sharing across heads in multi-head self-attention. Recall from Section~\ref{sec:decoder-only-transformers} that multi-head self-attention uses multiple sets of queries, keys, and values (each set is called a head), each performing the QKV attention mechanism as usual. This can be expressed as
\begin{eqnarray}
\mathrm{Output} & = & \mathrm{Merge}(\mathrm{head}_1,\cdots,\mathrm{head}_{\tau}) \mathbf{W}^{\mathrm{head}},
\end{eqnarray}

\noindent where $\mathrm{head}_j \in \mathbb{R}^{d_h}$ is the output of the $j$-th head, computed using the standard QKV attention function
\begin{eqnarray}
\mathrm{head}_j = \mathrm{Att}_{\mathrm{qkv}}(\mathbf{q}_{i}^{[j]},\mathbf{K}_{\le i}^{[j]},\mathbf{V}_{\le i}^{[j]}).
\end{eqnarray}

Here, $\mathbf{q}_{i}^{[j]}$ denotes the query vector at position $i$, while
$\mathbf{K}_{\le i}^{[j]}$ and $\mathbf{V}_{\le i}^{[j]}$ denote the corresponding
key and value matrices up to position $i$, with each vector projected into the $j$-th feature subspace. So this multi-head attention model can be interpreted as performing attention on a group of feature subspaces in parallel (see Figure~\ref{fig:comparison-of-multihead-attention-models} (b)). The KV cache needs to retain the keys and values for all these heads, that is, $\{(\mathbf{K}_{\le i}^{[1]},\mathbf{V}_{\le i}^{[1]}),\cdots,(\mathbf{K}_{\le i}^{[\tau]},\mathbf{V}_{\le i}^{[\tau]})\}$.

\begin{figure}[!t]
\centering
\input{./Chapter8/Figures/figure-sharing-for-the-KV-cache}
\caption{Illustration of QKV attention based on different multi-head and sharing mechanisms. (a) = single-head attention, and (b-e) = attention with multiple heads.}
\label{fig:comparison-of-multihead-attention-models}
\end{figure}

One refinement of the multi-head attention model, called \mindex{multi-query attention} (\mindex{MQA}), is to share keys and values across heads, while maintaining distinct queries for each head~\cite{shazeer:2019fast}. In MQA, there is a single set of keys and values $(\mathbf{K}_{\le i}, \mathbf{V}_{\le i})$. In addition, there are $\tau$ queries $\{\mathbf{q}_{i}^{[1]},\cdots,\mathbf{q}_{i}^{[\tau]}\}$, each corresponding to a different head. For each head, we have
\begin{eqnarray}
\mathrm{head}_j = \mathrm{Att}_{\mathrm{qkv}}(\mathbf{q}_{i}^{[j]},\mathbf{K}_{\le i},\mathbf{V}_{\le i}).
\end{eqnarray}

\noindent Figure~\ref{fig:comparison-of-multihead-attention-models} (c) illustrates this model. By sharing keys and values, the size of the KV cache would be $O(L \cdot d_h \cdot m)$.

\mindex{Grouped query attention} (\mindex{GQA}) is a natural extension to multi-head attention and MQA~\cite{ainslie-etal:2023gqa}. In GQA, heads are divided into $n_g$ groups, each corresponding to a shared set of keys and values. Hence we have $n_g$ sets of keys and values $\{(\mathbf{K}_{\le i}^{[1]},\mathbf{V}_{\le i}^{[1]}),\cdots,(\mathbf{K}_{\le i}^{[n_g]},\mathbf{V}_{\le i}^{[n_g]})\}$. See Figure~\ref{fig:comparison-of-multihead-attention-models} (d) for an illustration. Let $g(j)$ be the group id for the $j$-th head. The GQA model can be expressed as
\begin{eqnarray}
\mathrm{head}_j = \mathrm{Att}_{\mathrm{qkv}}(\mathbf{q}_{i}^{[j]},\mathbf{K}_{\le i}^{[g(j)]},\mathbf{V}_{\le i}^{[g(j)]}).
\end{eqnarray}

\noindent The size of the KV cache of GQA is $O(L \cdot n_g \cdot d_h \cdot m)$. One benefit of GQA is that we can trade off between computational efficiency and model expressiveness by adjusting $n_g$. When $n_g = \tau$, the model becomes the standard multi-head attention model. By contrast, when $n_g = 1$, it becomes the MQA model.

Sharing can also be performed across layers. Such a method falls into the family of shared weight and shared activation methods, which have been extensively used in Transformers~\cite{dehghani-etal:2018universal,lan-etal:2020albert}. For example, one can share KV activations or attention weights across layers to reduce both computation and memory footprints~\cite{xiao-etal:2019sharing,brandon-etal:2024reducing}. Figure~\ref{fig:comparison-of-multihead-attention-models} (e) shows an illustration of this method, where a query in a layer directly accesses the KV cache of a lower-level layer.

\subsection{Position Extrapolation and Interpolation}

\noindent Since Transformer layers are order-invariant with respect to input, we need some way to encode positional information in the input tokens. To do this, it is common to add positional embeddings to token embeddings, and then feed these combined embeddings into the Transformer layer stack. In this case, the embedding at position $i$ can be expressed as
\begin{eqnarray}
\mathbf{e}_i & = & \mathbf{x}_i + \mathrm{PE}(i) \label{eq:long-context-pe-base},
\end{eqnarray}

\noindent where $\mathbf{x}_i \in \mathbb{R}^{d}$ denotes the token embedding, and $\mathrm{PE}(i) \in \mathbb{R}^{d}$ denotes the positional embedding at position $i$. Because the token embedding $\mathbf{x}_i$ is position-agnostic, the positional embedding $\mathrm{PE}(i)$ encodes the absolute or relative context of the token. A straightforward implementation is to treat $\mathrm{PE}(i)$ as a set of learnable variables, trained jointly with the rest of the model parameters. This allows the model to learn a unique vector representation for every discrete position, and thus distinguish tokens based on where they appear in a sequence.

Using learned vectors to represent absolute positions works reasonably well when the sequence lengths encountered during inference closely match those seen during training. In practical LLM pre-training, however, we must impose strict maximum length limits (e.g., $m_l$) on training sequences to maintain computational tractability, yet we often desire to apply the trained model to significantly longer sequences during downstream deployment. In such scenarios, using learned positional embeddings has obvious drawbacks, as there are no trained embeddings for positions that are not observed in the training phase.

An alternative approach to modeling positional information is to develop positional embeddings that can generalize: once trained, the embedding model can be used to handle longer sequences. Suppose that we train a positional embedding model on sequences with a maximum length of $m_l$, and we wish to apply the trained model to a sequence of length $m$ ($m \gg m_l$). If the model is only trained on a limited range of positions, then this model will simply fail to deal with new data outside that range. See Figure~\ref{fig:pe-for-a-range-of-positions} (a) for an illustration where the learned embedding model cannot model data points outside the training domain if it lacks the ability to extrapolate.

\begin{figure}[!t]
\centering
\input{./Chapter8/Figures/figure-positional-embedding-extrapolation-and-interpolation}
\caption{Illustrations of different positional embedding behaviors across a sequence length continuum. Blue points denote positions explicitly observed during training, while red points denote unseen positions encountered at inference time. In sub-figure (a), the embedding model memorizes the training domain and cannot generalize. In sub-figures (b) and (c), the model successfully generalizes to unseen lengths via extrapolation and interpolation, respectively.}
\label{fig:pe-for-a-range-of-positions}
\end{figure}

There are several approaches to making positional embedding models generalize. They can be grouped into two classes.

\begin{itemize}
\item \vspace{0.5em} \mindex{Extrapolation}. The model learned on observed data points (i.e., positions) can be directly employed to assign meaningful values to data points beyond the original range.  For instance, if a model learns a continuous functional relationship for positions $\{1, \cdots, 10\}$, it can theoretically evaluate that same function at position $15$, maintaining the correct relative distance and geometric properties with respect to earlier tokens. Figure~\ref{fig:pe-for-a-range-of-positions} (b) illustrates this concept, where a function fitted to the training domain smoothly extends to estimate values for out-of-distribution coordinates.
\item \vspace{0.3em} \mindex{Interpolation}. This approach maps a larger range of data points into the original observation range. For example, suppose we have a model designed for numbers in the range $[1,10]$. When given a new range of $[1,20]$, we can scale this down by dividing every number by 2, thereby fitting all numbers into $[1,10]$. This scaling allows us to use the model trained on the range $[1,10]$ to describe data points in the expanded range of $[1,20]$. See Figure~\ref{fig:pe-for-a-range-of-positions} (c) for an illustration of this approach.
\end{itemize}
\vspace{0.5em}

In fact, positional embeddings in many systems have achieved some level of generalization. For example, sinusoidal encoding, the most common positional embedding method, employs sine and cosine functions that can naturally extend to sequences of any length. Although this approach seems direct and simple, it does not perform well when we significantly extend the sequences for processing. In this subsection, we will discuss several alternative methods based on either extrapolation or interpolation.

\subsubsection{Attention with Learnable Biases}

\noindent One problem with Eq. (\ref{eq:long-context-pe-base}) is that the embedding model treats each token independently and therefore ignores the distance between different tokens. A common improvement to this model, known as relative positional embedding, is to consider the pairwise relationship between tokens~\cite{shaw-etal:2018self}. The general idea behind this is to compute the offset between any pair of positions and incorporate it into the self-attention model. One of the simplest forms of self-attention with relative positional embedding is given by
\begin{eqnarray}
\mathrm{Att}_{\mathrm{qkv}}(\mathbf{q}_i,\mathbf{K}_{\le i},\mathbf{V}_{\le i}) & = & \sum_{j=0}^{i} \alpha(i,j) \mathbf{v}_j \\
\alpha(i,j) & = & \mathrm{Softmax} \left(\frac{\mathbf{q}_i \mathbf{k}_{j}^\top + \color{red}{\mathrm{PE}(i,j)}}{\sqrt{d}} + \mathrm{Mask}(i,j) \right). \label{eq:attention-weight-base-T5}
\end{eqnarray}

\noindent The only difference between this model and the original self-attention model is that a bias term $\mathrm{PE}(i,j)$ is added to the query-key product in this new model. Intuitively, $\mathrm{PE}(i,j)$ can be interpreted as a distance penalty for the pair of positions $i$ and $j$. As the distance between $i$ and $j$ increases, $\mathrm{PE}(i,j)$ decreases.

$\mathrm{PE}(i,j)$ can be defined in several different ways. Here, we consider the T5 version of relative positional embedding, called the T5 bias~\cite{raffel-etal:2020exploring}. For each pair of query $\mathbf{q}_i$ and key $\mathbf{k}_j$, the offset between them is defined to be\footnote{For language modeling, a query is only allowed to attend to its left-context, and so we have $i - j \ge 0$. In the more general case of self-attention, where a token can attend to all tokens in the sequence, we may have negative offsets when $i < j$.
}
\begin{eqnarray}
d(i,j) & = & i - j.
\end{eqnarray}

A naive formulation for the bias $\mathrm{PE}(i,j)$ would assign a distinct learnable scalar for every possible discrete offset, i.e., $\mathrm{PE}(i,j) = u_{i-j}$, where $u_{i-j}$ is the parameter tied to the exact distance $i - j$. However, simply assigning a unique value to each offset will restrict the model to offsets observed during training. When $i-j$ is larger than the maximum trained offset, the model cannot generalize.

The T5 bias introduces a generalized version of this approach. Rather than allocating a unique parameter to every possible integer offset, it groups these positional offsets into distinct ``buckets,'' where all offsets mapping to the same bucket share a single learnable parameter. Specifically, given $n_b$ buckets, the allocation strategy is defined as follows:

\begin{itemize}
\item \vspace{0.5em} For the first half of the buckets (indices $0$ to $\lfloor \frac{n_b}{2} \rfloor - 1$), the mapping is one-to-one. Each offset is mapped to a unique bucket: bucket $0 \leftrightarrow$ offset $0$, bucket $1 \leftrightarrow$ offset $1$, and so on. We can express this linear mapping as $b(i - j) = i - j$.
\item \vspace{0.3em} For the remaining half of the buckets (indices $\lfloor \frac{n_b}{2} \rfloor$ to $n_b - 1$), the bucket size increases logarithmically. As a result, larger offsets are grouped into progressively wider bins. For an offset $i - j \ge \lfloor \frac{n_b}{2} \rfloor$, the bucket index is computed as:
    \begin{eqnarray}
    b(i - j) & = & \lfloor \frac{n_b}{2} \rfloor + \left\lfloor \frac{\log(i - j) - \log(\lfloor \frac{n_b}{2} \rfloor)}{\log(\mathrm{dist}_{\mathrm{max}}) - \log(\lfloor \frac{n_b}{2} \rfloor)} \cdot \left(n_b - \lfloor \frac{n_b}{2} \rfloor - 1\right) \right\rfloor,
    \end{eqnarray}
    \noindent where $\mathrm{dist}_{\mathrm{max}}$ is a hyperparameter indicating the maximum expected offset threshold.
\item \vspace{0.3em} When $i - j \ge \mathrm{dist}_{\mathrm{max}}$, the offset is clipped and placed into the final bucket $n_b - 1$. In other words, this bucket serves as a catch-all for any arbitrarily large distances extending beyond the expected sequence length.
\end{itemize}

Together, the bucketing function is defined as
{\small
\begin{eqnarray}
& & b(i - j) \nonumber \\
& = &
\begin{cases}
i - j & \text{if } 0 \le i - j < \lfloor \frac{n_b}{2} \rfloor, \\
\min\left(n_b - 1, \lfloor \frac{n_b}{2} \rfloor + \left\lfloor \frac{\log(i - j) - \log(\lfloor \frac{n_b}{2} \rfloor)}{\log(\mathrm{dist}_{\mathrm{max}}) - \log(\lfloor \frac{n_b}{2} \rfloor)} \cdot \left(n_b - \lfloor \frac{n_b}{2} \rfloor - 1\right) \right\rfloor \right) & \text{if } i - j \ge \lfloor \frac{n_b}{2} \rfloor.
\end{cases} \nonumber \\
\end{eqnarray}
}

Figure~\ref{fig:qk-distances-in-buckets-T5} shows an illustration of these buckets. We see that in the first half of the buckets, each bucket is associated with only one value of $i - j$, while in the second half, the bucket size increases as $i - j$ grows. The last bucket is designed to handle sequences of arbitrary length.

\begin{figure}[!t]
\centering
\input{./Chapter8/Figures/figure-T5-buckets}
\caption{Illustration of distributing query-key offsets into buckets in the T5 model ($n_b = 32$ and $\mathrm{dist}_{\mathrm{max}}=1024$). Boxes represent buckets. In the first half of the buckets, we use a fixed bucket size. In the second half of the buckets, we increase the bucket size logarithmically. The last bucket contains all the query-key offsets that are not covered by previous buckets.}
\label{fig:qk-distances-in-buckets-T5}
\end{figure}

All relative distances mapping to the same bucket $b(i-j)$ share an identical learnable bias scalar $u_{b(i-j)}$. Substituting $\mathrm{PE}(i,j) = u_{b(i-j)}$ back into Eq. (\ref{eq:attention-weight-base-T5}), the final attention weight computation becomes\footnote{Note that, in~\citet{raffel-etal:2020exploring}'s T5 model, the rescaling operation for the query-key product is removed. The attention weight $\alpha(i,j)$ is then given by
\begin{eqnarray}
\alpha(i,j) & = & \mathrm{Softmax}(\mathbf{q}_i \mathbf{k}_{j}^\top + u_{b(i-j)} + \mathrm{Mask}(i,j)).
\end{eqnarray}
}
\begin{eqnarray}
\alpha(i,j) & = & \mathrm{Softmax}\left( \frac{\mathbf{q}_i \mathbf{k}_{j}^\top + \color{red}{u_{b(i-j)}}}{\sqrt{d}} + \mathrm{Mask}(i,j) \right),
\end{eqnarray}

The parameters $\{u_{0},\cdots,u_{n_b-1}\}$ are learned jointly with other parameters of the model. It should be emphasized that this model can generalize to long sequences. This is because $\mathrm{PE}(i,j)$s with similar query-key offsets share the same parameter, and this sharing strategy is particularly important for achieving good generalization, given that large query-key offsets are rare in training. In practice, we often set $n_b$ to a moderate number, and thus it can help control the overfitting of positional embedding models.

\subsubsection{Attention with Fixed Biases}

\noindent Relative positional embedding models are based on a set of learned biases added to the query-key product in self-attention. An alternative approach is to assign fixed values to these biases via predefined heuristics, rather than learning them from data. One benefit of this heuristic-based approach is that it does not rely on training and thus can be directly applied to any sequence once the biases are set.

One example of such an approach is~\citet{press-etal:2022train}'s approach, called \mindex{attention with linear biases} or \mindex{ALiBi} for short. In the ALiBi approach, the bias term is defined as the negative scaled query-key offset
\begin{eqnarray}
\mathrm{PE}(i,j) & = & -\beta \cdot (i - j) \nonumber \\
                 & = & \beta \cdot (j - i) \label{eq:abili-bias},
\end{eqnarray}

\noindent where $\beta$ is the scaling factor. Adding this term to the query-key product, we obtain a new form of attention weights
\begin{eqnarray}
\alpha(i,j) & = & \mathrm{Softmax} \left(\frac{\mathbf{q}_i \mathbf{k}_{j}^\top + \color{red}{\beta \cdot (j - i)}}{\sqrt{d}} + \mathrm{Mask}(i,j) \right).
\end{eqnarray}

This model can be interpreted as imposing a linear penalty on the attention score $\mathbf{q}_i \mathbf{k}_{j}^\top$ for every step the key $j$ moves away from the query $i$. See Figure~\ref{fig:T5-and-ALiBi-biases} for a comparison of the T5 bias and the ALiBi bias.

\begin{figure}[!t]
\centering
\input{./Chapter8/Figures/figure-T5-and-Alibi-biases}
\caption{Query-key products with biases (above = the T5 bias and below = the ALiBi bias). The color scale of the biases ranges from light blue denoting small absolute values to deep blue denoting large absolute values.}
\label{fig:T5-and-ALiBi-biases}
\end{figure}

In principle, the scalar hyperparameter $\beta$ could be tuned via grid search on a validation dataset. However, \citet{press-etal:2022train} demonstrated that setting $\beta$ systematically across attention heads yields strong performance across a variety of tasks without any tuning. Specifically, for a multi-head self-attention layer comprising $n_{\mathrm{head}}$ heads, the $\beta$ values decrease geometrically. The scalar for the $k$-th head is defined as
\begin{eqnarray}
\beta_k & = & 2^{-\frac{8k}{n_{\mathrm{head}}}}. \label{eq:alibi-beta}
\end{eqnarray}

The ALiBi approach provides a simple form of relative positional embeddings. There are other similar methods for designing query-key biases using the offset $i - j$. Table~\ref{tab:query-key-biases-comparison} shows a comparison of such biases. As an aside, it is worth noting that the form of the right-hand side of Eq. (\ref{eq:abili-bias}) is similar to distance-based features used in conventional feature-based systems. For example, in statistical machine translation systems, such features are widely used to model word reordering problems, resulting in models that can generalize well across different translation tasks~\cite{Koehn:2010}.

\begin{table}[t]
\centering
\begingroup
\renewcommand{\arraystretch}{1.2}
\begin{tabular}{l | l}
Entry & Query-Key Bias ($\mathrm{PE}(i,j)$) \\ \hline
T5~\cite{raffel-etal:2020exploring} & $u_{b(i-j)}$ \\ \hline
ALiBi~\cite{press-etal:2022train} & $-\beta \cdot (\colorbox{gray!30}{$i - j$})$ \\ \hline
Kerple~\cite{chi-etal:2022kerple} & $-\beta_1 (\colorbox{gray!30}{$i - j$})^{\beta_2}$ \ \ \ (power) \\
                                  & $-\beta_1 \log(1 + \beta_2(\colorbox{gray!30}{$i - j$}))$ \ \ \ (logarithmic)\\ \hline
Sandwich~\cite{chi-etal:2023dissecting} & $\sum_{k=1}^{\bar{d}/2} \cos\big((\colorbox{gray!30}{$i - j$})/10000^{2k/\bar{d}} \big)$ \\ \hline
FIRE~\cite{li-etal:2024functional} & $f\big( \psi(\colorbox{gray!30}{$i - j$})/\psi(\max(m_{\mathrm{len}},i)) \big)$
\end{tabular}
\endgroup
\caption{Query-key biases as relative positional embeddings. $\beta$, $\beta_1$, $\beta_2$, $\bar{d}$, and $m_{\mathrm{len}}$ are hyper-parameters. In the T5 model, $b(i-j)$ denotes the bucket assigned to $i - j$. In the FIRE model, $\psi(\cdot)$ represents a monotonically increasing bounding function, such as $\psi(x) = \log(cx+1)$, and $f(\cdot)$ is an FFN.}
\label{tab:query-key-biases-comparison}
\end{table}

\subsubsection{Rotary Positional Embedding}

\noindent Similar to sinusoidal embeddings, rotary positional embeddings use predefined frequency scales~\cite{su-etal:2024roformer}. Recall that in the sinusoidal embedding model, absolute positions are represented as fixed vectors composed of sine and cosine functions with varying frequencies, which are then additively combined with token embeddings. Rotary positional embeddings, instead, model positional context by applying geometrically meaningful rotations to token embeddings in the complex domain. This yields a formulation based on multiplicative interactions:
\begin{eqnarray}
\mathbf{e}_i & = & \mathbf{x}_i R(i) \label{eq:additive-embedding-e-base},
\end{eqnarray}

\noindent where $R(i) \in \mathbb{R}^{d \times d}$ is the rotation matrix representing the rotations performed on the token embedding $\mathbf{x}_i \in \mathbb{R}^{d}$.

For simplicity, we will first consider embeddings with only two dimensions and return to a discussion of the more general formulation later. Suppose we have a 2-dimensional token embedding represented as a row vector $\mathbf{x} = \begin{bmatrix} x_1 & x_2 \end{bmatrix}$. We can represent it as a vector in a plane, originating at the origin $(0,0)$ and terminating at $(x_1, x_2)$. A counterclockwise rotation of this vector refers to an operation of moving the vector around the origin while maintaining its magnitude, as shown in Figure~\ref{fig:vector-rotation} (a). The degree of rotation is usually defined by a specific angle, denoted by $\theta$. The rotation can be expressed mathematically in the form
\begin{eqnarray}
\mathrm{Ro}(\mathbf{x},\theta) & = & \mathbf{x} R_{\theta} \nonumber \\
                               & = & \begin{bmatrix} x_1 & x_2 \end{bmatrix}
                                     \begin{bmatrix} \cos \theta & \sin \theta \\ -\sin \theta & \cos \theta \end{bmatrix} \nonumber \\
                               & = & \begin{bmatrix} \cos \theta \cdot x_1 - \sin \theta \cdot x_2 & \sin \theta \cdot x_1 + \cos \theta \cdot x_2 \end{bmatrix}.
\end{eqnarray}

\begin{figure}[!t]
\centering
\input{./Chapter8/Figures/figure-rotation-examples}
\caption{Illustrations of vector rotations in a plane. Sub-figures (a) and (b) show rotations of a vector in a single step and multiple steps, respectively. Sub-figure (c) shows the embeddings of tokens \textit{cat} and \textit{sleeping} in two different sentences. We show these sentences with a subscript affixed to each token to indicate its position. If we represent tokens as vectors, we can add positional information by rotating these vectors. This rotation preserves the ``distances'' between the vectors. For example, given that the distance between \textit{cat} and \textit{sleeping} is the same in both sentences, the angle between their embeddings also remains the same during rotation.}
\label{fig:vector-rotation}
\end{figure}

\noindent where $R_{\theta} = \begin{bmatrix} \cos \theta & \sin \theta \\ -\sin \theta & \cos \theta \end{bmatrix}$ is the rotation matrix. If two or more rotations are performed on the same vector, we can rotate the vector further. This follows from the fact that the composition of successive rotations is itself a rotation. More formally, rotating a vector by an angle $\theta$, repeated $t$ times, can be expressed as
\begin{eqnarray}
\mathrm{Ro}(\mathbf{x},t\theta) & = & \mathbf{x} R_{t \theta} \nonumber \\
                                & = & \begin{bmatrix} \cos t \theta \cdot x_1 - \sin t \theta \cdot x_2 & \sin t \theta \cdot x_1 + \cos t \theta \cdot x_2 \end{bmatrix}.
\end{eqnarray}

If we interpret $t$ as the position of a token represented by $\mathbf{x}$ in a sequence, then we will find that the above equation defines a simple positional embedding model. As shown in Figure~\ref{fig:vector-rotation} (b), we start moving the token from position 0. Each time we move one step forward, the vector is rotated by the angle $\theta$. Upon arriving at the position $t$, the representation of the token with positional context is given by $\mathrm{Ro}(\mathbf{x},t\theta)$. As the rotations do not change the magnitude of the embedding, the original ``meaning'' of the token is retained. The positional information is injected into the embedding when it gets rotated.

A common way to understand vector rotation is via complex numbers. It is easy to transform each vector $\mathbf{x} = \begin{bmatrix} x_1 & x_2 \end{bmatrix}$ in the 2D Euclidean space $\mathbb{R}^2$ to a complex number $\mathbf{x}' = x_1 + \mathtt{i} x_2$ in the complex space $\mathbb{C}$ via a bijective linear map. Then, the rotation of $\mathbf{x}$ with the angle $t\theta$ corresponds to the multiplication by $e^{\mathtt{i} t\theta}$. Given that $e^{\mathtt{i} t\theta} = \cos t \theta + \mathtt{i} \sin t \theta$, the rotation operation can be re-expressed in the form
\begin{eqnarray}
\mathbf{x} R_{t \theta} & \mapsto & \mathbf{x}' e^{\mathtt{i} t\theta} \nonumber \\
                        & =       & (x_1 + \mathtt{i} x_2) (\cos t \theta + \mathtt{i} \sin t \theta) \nonumber \\
                        & =       & \cos t \theta \cdot x_1 - \sin t \theta \cdot x_2 + \mathtt{i} (\sin t \theta \cdot x_1 + \cos t \theta \cdot x_2).
\end{eqnarray}

\noindent Here we denote the token representation $\mathbf{x}' e^{\mathtt{i} t\theta}$ by $C(\mathbf{x},t\theta)$. The inner product of the representations of the tokens at positions $t$ and $s$ can be written as
\begin{eqnarray}
\langle C(\mathbf{x},t\theta), C(\mathbf{y},s \theta) \rangle & = & (\mathbf{x}' \overline{\mathbf{y}'}) e^{\mathtt{i} (t - s) \theta} \label{eq:rope-inner-product},
\end{eqnarray}

\noindent where $\overline{\mathbf{y}'}$ is the complex conjugate of $\mathbf{y}'$. The result of this inner product involves a term $t - s$, and so it can model the offset between the two tokens.

Now we go back to representations in the 2D Euclidean space. The dot-product of $\mathrm{Ro}(\mathbf{x},t\theta)$ and $\mathrm{Ro}(\mathbf{y},s\theta)$ can be written as a function of $(t - s)\theta$
\begin{eqnarray}
\mathrm{Ro}(\mathbf{x},t\theta) [\mathrm{Ro}(\mathbf{y},s\theta)]^\top & = & \mathbf{x} R_{t \theta} [\mathbf{y} R_{s \theta}]^\top \nonumber \\
                                                                               & = & \mathbf{x} R_{t \theta} [R_{s \theta}]^\top \mathbf{y}^\top \nonumber \\
                                                                               & = & \mathbf{x} R_{(t - s) \theta} \mathbf{y}^\top.
\end{eqnarray}

\noindent Given this result, if we consider $\mathrm{Ro}(\mathbf{x},t\theta)$ and $\mathrm{Ro}(\mathbf{y},s\theta)$ as the query and the key, then the self-attention operation will implicitly involve the modeling of relative positional context.

This rotary positional embedding can be extended to multi-dimensional embeddings. For a $d$-dimensional token embedding $\mathbf{x} = \begin{bmatrix} x_1 & x_2 & \cdots & x_d \end{bmatrix}$, we can treat it as a $\frac{d}{2}$-dimensional complex vector $\mathbf{x}' = \begin{bmatrix} x'_1 & x'_2 & \cdots & x'_{d/2} \end{bmatrix} = \begin{bmatrix} x_1 + \mathtt{i} x_2 & x_3 + \mathtt{i} x_4 & \cdots & x_{d-1} + \mathtt{i} x_d \end{bmatrix}$, where each consecutive pair of items forms a complex number. Then, the rotary positional embedding in the complex space is given by
\begin{eqnarray}
C(\mathbf{x},t\theta) & = & \sum_{k=1}^{d/2} x'_k e^{\mathtt{i} t\theta_k} \vec{e}_k,
\end{eqnarray}

\noindent where $\vec{e}_k$ is the standard basis vector with a single non-zero value in the $k$-th coordinate and 0's elsewhere~\cite{biderman-etal:2021rotary}.

Although this formula involves a complicated expression, its equivalent form in the $d$-dimensional Euclidean space is relatively easy to understand. We can write it as
\begin{eqnarray}
\mathrm{Ro}(\mathbf{x},t\theta) & = & \begin{bmatrix} x_1 & x_2 & \cdots & x_d \end{bmatrix}
                                      \begin{bmatrix} R_{t\theta_1} & & & \\ & R_{t\theta_2} & & \\ & & \ddots & \\ & & & R_{t\theta_{d/2}} \end{bmatrix} \label{eq:rope-base},
\end{eqnarray}

\noindent where $R_{t \theta_k} = \begin{bmatrix} \cos t \theta_k & \sin t \theta_k \\ -\sin t \theta_k & \cos t \theta_k \end{bmatrix}$. $\theta = \begin{bmatrix} \theta_1, \cdots, \theta_{d/2}\end{bmatrix}$ are the parameters for controlling the angles of rotations in different dimensions. Typically, $\theta_k$ is set to $10000^{-\frac{2(k-1)}{d}}$, which is analogous to the setting in sinusoidal embeddings.

In a practical implementation, Eq. (\ref{eq:rope-base}) can be rewritten into a form that relies solely on the element-wise product and addition of vectors:
\begin{eqnarray}
\mathrm{Ro}(\mathbf{x},t\theta) & = & \begin{bmatrix} x_1 \\ x_2 \\ \vdots \\ x_{d-1} \\ x_{d} \end{bmatrix}^\top \odot \begin{bmatrix} \cos t\theta_1 \\ \cos t\theta_1 \\ \vdots \\ \cos t\theta_{d/2} \\ \cos t\theta_{d/2} \end{bmatrix}^\top + \begin{bmatrix} -x_2 \\ x_1 \\ \vdots \\ -x_{d} \\ x_{d-1} \end{bmatrix}^\top \odot \begin{bmatrix} \sin t\theta_1 \\ \sin t\theta_1 \\ \vdots \\ \sin t\theta_{d/2} \\ \sin t\theta_{d/2} \end{bmatrix}^\top \label{eq:rope-simple}.
\end{eqnarray}

Finally, we rewrite Eq. (\ref{eq:additive-embedding-e-base}) to obtain the form of the embedding at position $i$

\begin{eqnarray}
\mathbf{e}_i & = & \mathrm{Ro}(\mathbf{x}_i,i\theta).
\end{eqnarray}

\subsubsection{Position Interpolation}

\noindent In position interpolation, our goal is to map the positions in the new sequence to match the observed range in training. Suppose the model is trained on position indices in $[0, m_l]$. When inference requires processing a length $m > m_l$, we compress the positions in the range $[0,m]$ such that their representations fit into the original domain $[0,m_l]$.

To illustrate, consider the rotary positional embedding model described above. The embedding of each token is given by the transformation $\mathrm{Ro}(\mathbf{x}_i, i\theta)$, where the parameter vector is $\theta = \begin{bmatrix} \theta_1, \cdots, \theta_{d/2}\end{bmatrix}$. As shown in Eq. (\ref{eq:rope-simple}), $\mathrm{Ro}(\mathbf{x}_i, i\theta)$ relies on a combination of periodic functions across different feature dimensions
\begin{eqnarray}
\cos i\theta & = & \begin{bmatrix} \cos i\theta_1 & \cdots & \cos i\theta_{d/2} \end{bmatrix} \label{rope-cos-theta} \\
\sin i\theta & = & \begin{bmatrix} \sin i\theta_1 & \cdots & \sin i\theta_{d/2} \end{bmatrix} \label{rope-sin-theta},
\end{eqnarray}

\noindent $\theta_k$ follows an exponential form
\begin{eqnarray}
\theta_k & = & b^{-\frac{2(k-1)}{d}},
\end{eqnarray}

\noindent where $b$ is the base. The period of $\cos i\theta_k$ and $\sin i\theta_k$ is
\begin{eqnarray}
T_k & = & 2\pi \cdot b^{\frac{2(k-1)}{d}}.
\end{eqnarray}

The key intuition behind linear position interpolation is to adjust these periods so that the new extended positions can be encoded within the pre-trained boundaries $[0,m_l]$. The most straightforward way to achieve this is to linearly scale up $T_k$ by the extrapolation ratio $\frac{m}{m_l}$, yielding:
\begin{eqnarray}
T'_k & = & \frac{m}{m_l} \cdot 2\pi \cdot b^{\frac{2(k-1)}{d}}.
\end{eqnarray}

\noindent By doing so, all position indices in $[0, m]$ are uniformly compressed into $[0, m_l]$. This linear scaling can be implemented by dividing the position index in the embedding model by the scaling factor~\cite{chen-etal:2023extending}. The modified model with linear positional interpolation becomes
\begin{eqnarray}
\mathrm{Ro}'(\mathbf{x}_i,i\theta) & = & \mathrm{Ro} \left(\mathbf{x}_i,\frac{m_l}{m} i\theta \right).
\end{eqnarray}

Another method of positional interpolation is to scale the base $b$, a technique known as \mindex{NTK-aware scaled RoPE}\footnote{This method was first proposed in an online forum: \url{https://www.reddit.com/r/LocalLLaMA/comments/14lz7j5/ntkaware_scaled_rope_allows_llama_models_to_have/}}. Suppose the base $b$ is scaled by a factor $\lambda$. The goal is to enforce the period of this new model in the highest frequency dimension (i.e., the last dimension where $k = \frac{d}{2}$) to match that of the linear positional interpolation model. Substituting $k = \frac{d}{2}$ into the exponent yields $\frac{2(d/2-1)}{d} = \frac{d-2}{d}$. Equating the periods gives:
\begin{eqnarray}
2\pi \cdot (\lambda b)^{\frac{d-2}{d}} & = & \frac{m}{m_l} \cdot 2\pi \cdot b^{\frac{d-2}{d}}.
\end{eqnarray}

By solving this equation, we can isolate $\lambda$:
\begin{eqnarray}
\lambda^{\frac{d-2}{d}} & = & \frac{m}{m_l}.
\end{eqnarray}

Then, we have
\begin{eqnarray}
\lambda & = & \left( \frac{m}{m_l} \right)^{\frac{d}{d-2}}.
\end{eqnarray}

This approach yields a modified embedding model:
\begin{eqnarray}
\mathrm{Ro}'(\mathbf{x}_i,i\theta) & = & \mathrm{Ro}(\mathbf{x}_i,i\theta'),
\end{eqnarray}

\noindent where
\begin{eqnarray}
\theta' & = & \begin{bmatrix} (\lambda b)^{-\frac{0}{d}}, (\lambda b)^{-\frac{2}{d}}, \cdots, (\lambda b)^{-\frac{d-2}{d}}\end{bmatrix}.
\end{eqnarray}

Note that scaling the base provides a non-uniform method for scaling the periods across different dimensions of $\theta$. This method has been found to be helpful for extending LLMs to longer sequences, and several improvements have been developed~\cite{peng-etal:2024yarn,ding-etal:2024longrope}.

\subsection{Remarks}
\label{sec:long-context-llm-remarks}

\noindent In this section, we have presented a variety of methods for long-context language modeling. We close this section by discussing some interesting issues related to these methods.

\subsubsection{Need for Long Context}

\noindent One of the ultimate goals of long-context LLMs is to enable these models to precisely encode infinite context. Here, the term \textit{infinite context} refers to the ability of an LLM to continuously process incoming tokens. This motivates the development of LLMs that can handle very long context or streaming data. As discussed in Section~\ref{sec:long-context-cache-and-memory}, it is common to use fixed-size memory models to process continuously expanding context. Many such systems are based on recurrent architectures or their variants, because they are inherently suited to model sequential data where the effects of past inputs continue indefinitely. Another way to achieve infinite memory is to develop alternatives to self-attention models, for example, one can use continuous-space attention models to encode context, which removes the dependency on context length~\cite{martins-etal:2022former}.

When studying long-context LLMs, it is natural to wonder what mechanisms may explain the use of long context in language modeling. Can we compress the representation of infinite context into a relatively small model? Are all context tokens useful for predicting the next token? Can we know in advance which contextual information is critical for prediction? General answers to all these questions are not obvious, but they inspire follow-on research on explainable models, and some interesting results have been found. For example,~\citet{deletang-etal:2024language} conducted extensive experiments to show that LLMs are powerful in-context compressors. Although viewing predictive models as compression models has long been studied in machine learning, it also provides insights into our understanding of LLM scaling laws. Furthermore, \citet{pal-etal:2023future} and \citet{wu-etal:2024language} investigated whether the internal features learned up to the current computational step, even if not explicitly optimized for it, are already sufficient to accurately predict tokens multiple steps into the future. Note that the need for long-context in language modeling is highly dependent on the problem that we address. A related issue is where to apply LLMs and how to evaluate them. For example, in summarization tasks we may only need to distill and focus on a few key aspects of the text, while in retrieval-like tasks we need to ``memorize'' the entire context so that the relevant information can be accessed. We will discuss the evaluation issue later in this subsection.

\subsubsection{Pre-training or Adapting LLMs?}

The training of LLMs requires significant computational resources. Although it is straightforward to train LLMs on long sequences, the training becomes computationally prohibitive for large datasets. It is common practice to pre-train LLMs on general datasets, and then adapt them with modest fine-tuning effort. For example, LLMs with relative or rotary positional embeddings can be directly trained on large-scale data in the pre-training phase. While the resulting models may exhibit some length extrapolation capabilities in the inference phase, it may be more effective to fine-tune them on longer sequences.

Ideally, we would like to pre-train LLMs with standard Transformer architectures and adapt them to new tasks. This allows us to use many off-the-shelf LLMs and efficiently adapt them to handle long sequences. However, when new architectures are adopted, it seems inevitable that we need to train these models from scratch. This poses practical difficulties for developing long-context LLMs, as we cannot leverage well-developed, pre-trained models and must instead train them ourselves. On the other hand, fine-tuning is still an effective way to adapt LLMs with certain architectures that are different from those in pre-training. An example is models augmented with external memories. In these models, the pre-trained LLMs are fixed, and the focus is on how LLMs interact with external memory modules. In RAG, for instance, it is common to fine-tune LLMs to improve their use of retrieval-augmented inputs. Another example is to train an LLM with full attention and then replace it with sparse attention in the fine-tuning phase. The pre-trained LLM provides an initialization for the parameters of the new model, and this model is then fine-tuned as usual.

\subsubsection{Evaluating Long-context LLMs}

\noindent Evaluating long-context LLMs is important, but it is a relatively new problem in NLP. The general idea is that, if we input a long context to an LLM, then we can examine the output of the LLM to determine whether it understands the entire context and makes use of it in predicting subsequent tokens. In conventional NLP research, such evaluations are often aimed at examining the ability of NLP models in handling long-range dependencies. However, the length of contexts used in recent LLMs is much larger than those used in NLP systems a few years ago. This motivates researchers to develop new evaluation benchmarks and metrics for long-context LLMs.

One approach is to use the perplexity metric. However, despite its simplicity, this method tends to reflect the ability of LLMs to make use of local context rather than global context. It is therefore tempting to develop evaluation methods that are specific to long-context LLMs. Popular methods include various synthetic tasks where artificially generated or modified data is used to evaluate specific capabilities of long-context LLMs. In needle-in-a-haystack\footnote{\url{https://github.com/gkamradt/LLMTest_NeedleInAHaystack}} and passkey retrieval tasks~\cite{mohtashami-and-jaggi:2024random,chen-etal:2023extending}, for instance, LLMs are required to identify and extract a small, relevant piece of information from a large volume of given text. The assumption here is that an LLM with sufficient memory should remember earlier parts of the text as it processes new information. This LLM can thus pick out from the text the relevant details, which might be sparse and hidden among much irrelevant information. Alternatively, in copy memory tasks (or copy tasks for short), LLMs are required to repeat the input text or a specific segment multiple times. These tasks were initially proposed to test the extent to which recurrent models can retain and recall previously seen tokens~\cite{hochreiter-schmidhuber:1997long,arjovsky-etal:2016unitary}, and have been adopted in evaluating recent LLMs~\cite{bulatov-etal:2022recurrent,gu-and-dao:2023mamba}.

Another approach to evaluating long-context LLMs is to test them on NLP tasks that involve very long input sequences. Examples include long-document or multi-document summarization, long-document question answering, code completion, and so on. A benefit of this approach is that it can align evaluations with user expectations.

Although many methods have been developed, there is still no general way to evaluate long-context LLMs~\cite{liu-etal:2024forgetting}. One problem is that most of these methods focus on specific aspects of LLMs, rather than their fundamental ability to model very long contexts. Even though an LLM can pick out the appropriate piece of text from the input, we cannot say that it truly understands the entire context. Instead, it might just remember some important parts of the context, or even simply recall the answer via the model learned in pre-training. Moreover, the data used in many tasks is small-scale and relatively preliminary, leading to discrepancies between evaluation results and actual application performance. A more interesting issue is that the results of LLMs are influenced by many other factors and experimental setups, for example, using different prompts can lead to very different outcomes. This makes evaluation even more challenging because improvements may not solely result from better modeling of long contexts, and there is a risk of overclaiming performance gains. Nevertheless, many open questions remain in the development and evaluation of long-context LLMs. For example, these models still suffer from limitations such as restricted context length and high latency. Studying these issues is likely to prove a valuable future direction.

%----------------------------------------------------------------------------------------
% NEW SECTION
%----------------------------------------------------------------------------------------

\section{Summary}

In this chapter, we have discussed the concept of LLMs and related techniques. This can be considered a general, though not comprehensive, introduction to LLMs, laying the foundation for further discussions on more advanced topics in subsequent chapters. Furthermore, we have explored two ways to scale up LLMs. The first focuses on the large-scale pre-training of LLMs, which is crucial for developing state-of-the-art models. The second focuses on methods for adapting LLMs to long inputs, including optimizing attention models, designing more efficient and compressed KV caches, incorporating memory models, and exploring better positional embeddings.

The strength of LLMs lies in their ability to break the constraints of training NLP models for a limited number of specific tasks. Instead, LLMs learn from large amounts of text through the simple task of token prediction --- predicting the next token in a sequence given its preceding context. A general view is that, by repeating this token prediction task a large number of times, LLMs can acquire some knowledge of the world and language, which can then be applied to new tasks. As a result, LLMs can be prompted to perform any task by framing it as a task of predicting subsequent tokens given prompts. This emergent ability in language models comes from several dimensions, such as scaling up training, model size, and context size. It is undeniable that scaling laws are currently the fundamental principle adopted in developing large language models, although simply increasing model size has yet to prove sufficient for achieving AGI. These continuously scaled LLMs have been found to show capabilities in general-purpose language understanding, generation, and reasoning. More recently, it has been found that scaling compute at inference time (i.e., test-time compute) can also lead to significant improvements in complex reasoning tasks~\cite{openai:2024learning}.

Given their amazing power, LLMs have attracted considerable interest, both in terms of techniques and applications. As a result, the explosion of research interest in LLMs has also led to a vast number of new techniques and models. However, we do not attempt to provide a comprehensive literature review on all aspects of LLMs, given the rapid evolution of the field. Nevertheless, one can still gain knowledge about LLMs from general reviews~\cite{zhao-etal:2023survey,minaee-etal:2024large} or more focused discussions on specific topics~\cite{ruan-etal:2024abigsurveyofllms}.

% !Mode:: "TeX:UTF-8"
% !TEX encoding = UTF-8 Unicode

%----------------------------------------------------------------------------------------
% Natural Language Processing: Neural Networks and Large Language Models
%
% Copyright 2021-2026
% Tong Xiao (xiaotong@mail.neu.edu.cn) and Jingbo Zhu (zhujingbo@mail.neu.edu.cn)
%----------------------------------------------------------------------------------------

%----------------------------------------------------------------------------------------
%    CONFIGURATIONS
%----------------------------------------------------------------------------------------

%\setcounter{chapter}{8}

%----------------------------------------------------------------------------------------
%	CHAPTER 9
%----------------------------------------------------------------------------------------

\chapter{Prompting}

\thispagestyle{chapterurl} % add urls to the header

\noindent \textit{Prompting} refers to the process of providing an LLM with a specific input or cue to generate a desired output or perform a task. For example, if we want the LLM to translate a sentence from English to Chinese, we can prompt it as follows

\vspace{0.5em}
\begin{tcolorbox}[frame empty]
\hspace{2em} Translate the text from English to Chinese.

\vspace{0.2cm}

\hspace{2em} Text: The early bird catches the worm.

\vspace{0.2cm}

\hspace{2em} Translation: \underline{\hspace{2em}}
\end{tcolorbox}
\vspace{0.5em}

\noindent Prompting is crucial for LLMs because it directly influences how effectively these models understand and respond to user queries. A well-crafted prompt can guide an LLM to generate more accurate, relevant, and contextually appropriate responses. Furthermore, this process can be iteratively refined. By analyzing the responses of the LLM, users can adjust their prompts to align more closely with their specific needs. Given the importance of prompting in applying LLMs, prompt design has become an essential skill for users and developers working with LLMs. The growing importance of prompt design has given rise to an active research field known as \mindex{prompt engineering}, in which we design effective prompts to make better use of LLMs and enhance their practical utility in real-world applications.

An important concept related to prompting is \mindex{in-context learning}. When prompting an LLM, we can add new information to the context, such as demonstrations of problem-solving. This allows the LLM to learn from this context how to solve the problem. Here is an example of prompting LLMs with a few demonstrations of how to classify text based on sentiment polarity.

\vspace{0.5em}
\begin{tcolorbox}[frame empty]
\hspace{2em} Here are some examples of text classification.

\vspace{0.3em}

\hspace{2em} Example 1: We had a delightful dinner together. $\to$ Label: Positive

\vspace{0.2em}

\hspace{2em} Example 2: I'm frustrated with the delays. $\to$ Label: Negative

\vspace{0.3em}

\hspace{2em} What is the label for ``That comment was quite hurtful.''?

\vspace{0.2em}

\hspace{2em} Label: \underline{\hspace{2em}}
\end{tcolorbox}
\vspace{0.5em}

In-context learning is often seen as an emergent ability of LLMs that arises after pre-training. Though LLMs can be trained or tuned to perform new tasks, in-context learning provides a very efficient way to adapt these models without any training or tuning effort. Perhaps this is one of the most notable features of LLMs: they indeed learn general knowledge about the world and language during pre-training, which we can easily apply to new challenges. Moreover, in-context learning reflects the broader trend of making AI systems more generalizable and user-friendly. Instead of requiring specialized engineers to fine-tune models for every unique task, users can interact with LLMs in a more intuitive way, simply providing examples or adjusting the context as needed.

In this chapter, we focus on prompting techniques in LLMs. We begin by considering several interesting prompt designs commonly used in prompt engineering. Then, we discuss a series of refinements to these methods. Finally, we explore approaches for automating prompt design.

\vspace{0.5em}

%----------------------------------------------------------------------------------------
% NEW SECTION
%----------------------------------------------------------------------------------------

\section{General Prompt Design}

\noindent This section presents basic concepts in prompt design, along with examples of how to prompt LLMs for various NLP tasks. Since the effectiveness of prompting depends heavily on the LLMs being used, prompts often vary across different LLMs, which makes it difficult to provide a comprehensive list of prompts for all LLMs and downstream tasks. Therefore, this discussion does not focus on any specific LLM. Instead, the goal is to provide guiding principles for prompt design.

\subsection{Basics}

\noindent The term \textit{prompt} is used in many different ways. In this chapter, we define a prompt as the input text to an LLM, denoted by $\mathbf{x}$. The LLM generates an output $\mathbf{y}$ according to the conditional probability $\Pr(\mathbf{y}|\mathbf{x})$. In this generation process, the prompt serves as the condition upon which predictions are made, and it can contain any information that helps describe and solve the problem.

A prompt can be obtained using a prompt template (or template for short)~\cite{liu-etal:2023pre}. A template is a piece of text containing placeholders or variables, where each placeholder can be filled with specific information. The following are two templates for asking the LLM for weekend suggestions.

\vspace{0.5em}
\begin{tcolorbox}[frame empty]

\hspace{2em} Please give me some suggestions for a fun weekend.

\hspace{2em} \underline{\hspace{2em}}
\end{tcolorbox}
\vspace{0.5em}

\vspace{0.5em}
\begin{tcolorbox}[frame empty]

\hspace{2em} If $\{*\mathrm{premise}*\}$, what are your suggestions for a fun weekend.

\hspace{2em} \underline{\hspace{2em}}
\end{tcolorbox}
\vspace{0.5em}

In the first template, we simply instruct the LLM to return some suggestions. So the template is just a piece of text with no variables. In the second template, the variable $\{*\mathrm{premise}*\}$ needs to be specified by the user to provide a premise for making suggestions. For example, if we input
\begin{eqnarray}
\mathrm{premise} & = & \textrm{the weather is nice this weekend},
\end{eqnarray}

\noindent then we can generate a prompt

\vspace{0.5em}
\begin{tcolorbox}[frame empty]

\hspace{2em} If the weather is nice this weekend,

\hspace{2em} what would you suggest for a fun weekend?

\hspace{2em} \underline{\hspace{2em}}
\end{tcolorbox}
\vspace{0.5em}

We can also design a template with multiple variables. Here is an example in which we compare the semantic similarity between two sentences.

\vspace{0.5em}
\begin{tcolorbox}[frame empty]

\hspace{2em} Here is a sentence

\hspace{2em} $\{*\mathrm{sentence1}*\}$

\hspace{2em} Here is another sentence

\hspace{2em} $\{*\mathrm{sentence2}*\}$

\vspace{0.3em}

\hspace{2em} Compute the semantic similarity between the two sentences

\hspace{2em} \underline{\hspace{2em}}
\end{tcolorbox}
\vspace{0.5em}

A popular way to format prompts is to write each input or output in a ``name: content'' style. For example, we can describe a conversation between two people, named John and David, and use the LLM to continue the conversation. The following is a template for such prompts.

\vspace{0.5em}
\begin{tcolorbox}[frame empty]

\hspace{2em} John: $\{*\mathrm{utterance1}*\}$

\hspace{2em} David: $\{*\mathrm{utterance2}*\}$

\hspace{2em} John: $\{*\mathrm{utterance3}*\}$

\hspace{2em} David: $\{*\mathrm{utterance4}*\}$

\hspace{2em} John: $\{*\mathrm{utterance5}*\}$

\hspace{2em} David: $\{*\mathrm{utterance6}*\}$

\hspace{2em} John: $\{*\mathrm{utterance7}*\}$

\hspace{2em} David: \underline{\hspace{2em}}

\end{tcolorbox}
\vspace{0.5em}

The ``name: content'' format can be used to define the task that we want the LLM to perform. For example, given that ``Q'' and ``A'' are commonly used abbreviations for ``Question'' and ``Answer'', respectively, we can use the following template to do question-answering.

\vspace{0.5em}
\begin{tcolorbox}[frame empty]

\hspace{2em} Q: $\{*\mathrm{question}*\}$

\hspace{2em} A: \underline{\hspace{2em}}

\end{tcolorbox}
\vspace{0.5em}

This format can be used to describe more complex tasks. For instance, the following is an example of providing a specification for a translation task:

\vspace{0.5em}
\begin{tcolorbox}[frame empty]

\hspace{2em} Task: Translation

\hspace{2em} Source language: English

\hspace{2em} Target language: Chinese

\hspace{2em} Style: Formal text

\hspace{2em} Template: Translate the following sentence: $\{*\mathrm{sentence}*\}$

\hspace{2em} \underline{\hspace{2em}}

\end{tcolorbox}
\vspace{0.5em}

\noindent In practical systems, it is common to represent and store such data in key-value pairs, such as JSON\footnote{
The JSON representation is

\vspace{0.5em}
\begin{tcolorbox}[frame empty]

\{

\hspace{2em} "Task": "Translation",

\hspace{2em} "Source language": "English",

\hspace{2em} "Target language": "Chinese",

\hspace{2em} "Style": "Formal text",

\hspace{2em} "Template": "Translate the following sentence: $\{*\mathrm{sentence}*\}$",

\}

\end{tcolorbox}
\vspace{0.5em}

}.

When the problem is difficult to describe in an attribute-based manner, it is more common to instruct the LLM using a clear and detailed description. There are many ways to do this. One example is to assign a role to the LLM and provide sufficient context. The following is a template that instructs an LLM to act as an expert and answer questions from children.

\vspace{0.5em}
\begin{tcolorbox}[frame empty]

\begingroup
\setlength{\leftskip}{2em}
\setlength{\rightskip}{2em}

You are a computer scientist with extensive knowledge in the field of deep learning.

\vspace{0.3em}

Please explain the following computer-related concept to a child around 10 years old, using simple examples whenever possible.

\vspace{0.3em}

$\{*\mathrm{concept}*\}$

\vspace{0.3em}

\underline{\hspace{2em}}

\endgroup
\end{tcolorbox}
\vspace{0.5em}

\noindent The text ``You are a computer scientist ... deep learning. '' is sometimes referred to as a system prompt (or system information), and is provided to help the LLM understand the context or constraints of the task it is being asked to perform.

\subsection{In-context Learning}

\noindent Learning can occur during inference in large language models. In-context learning is one such method, where prompts involve demonstrations of problem-solving, and LLMs can learn how to solve new problems from these demonstrations. Since we do not update model parameters in this process, in-context learning can be viewed as a way to efficiently activate the knowledge learned in pre-training without additional training or fine-tuning. This enables LLMs to quickly adapt to new problems and extends the capabilities of pre-trained models without task-specific adjustments.

In-context learning can be illustrated by comparing three methods: zero-shot learning, one-shot learning, and few-shot learning. Zero-shot learning, as its name implies, does not involve a traditional ``learning'' process. Instead, it directly uses LLMs to address new problems that were not observed during training. In practice, we can iteratively adjust prompts to guide the LLMs in generating better responses, without demonstrating problem-solving steps or providing examples. Consider the following example. Suppose we want to use an LLM as an assistant that can help correct English sentences. A zero-shot learning prompt is given by

\vspace{0.5em}
\begin{tcolorbox}[frame empty]

\begingroup
\renewcommand{\arraystretch}{1.0}
\setlength{\tabcolsep}{6pt}
\begin{tabular}{r l}
{\color{gray} \footnotesize{SYSTEM}} & You are a helpful assistant, and are great at grammar correction. \\ [0.2cm]
{\color{gray} \footnotesize{USER}}  & You will be provided with a sentence in English. The task is \\
 & to output the correct sentence. \\ [0.1cm]
 & Input: She don't like going to the park. \\
 & Output: \underline{\hspace{2em}}
\end{tabular}
\endgroup

\end{tcolorbox}
\vspace{0.5em}

\noindent Here the gray words are used to indicate different fields of the prompt.

In one-shot learning, we extend this prompt by adding a demonstration of how to correct sentences, thereby allowing the LLM to learn from this newly added experience.

\vspace{0.5em}
\begin{tcolorbox}[frame empty]

\begingroup
\renewcommand{\arraystretch}{1.0}
\setlength{\tabcolsep}{6pt}
\begin{tabular}{r l}
{\color{gray} \footnotesize{SYSTEM}} & You are a helpful assistant, and are great at grammar correction. \\ [0.2cm]
{\color{gray} \footnotesize{DEMO}}  & You will be provided with a sentence in English. The task is \\
 & to output the correct sentence. \\ [0.1cm]
 & Input: There is many reasons to celebrate. \\
 & Output: There are many reasons to celebrate. \\ [0.2cm]
{\color{gray} \footnotesize{USER}}  & You will be provided with a sentence in English. The task is \\
 & to output the correct sentence. \\ [0.1cm]
 & Input: She don't like going to the park. \\
 & Output: \underline{\hspace{2em}}
\end{tabular}
\endgroup

\end{tcolorbox}
\vspace{0.5em}

Furthermore, we can add more demonstrations to enable few-shot learning.

\vspace{0.5em}
\begin{tcolorbox}[frame empty]

\begingroup
\renewcommand{\arraystretch}{1.0}
\setlength{\tabcolsep}{6pt}
\begin{tabular}{r l}
{\color{gray} \footnotesize{SYSTEM}} & You are a helpful assistant, and are great at grammar correction. \\ [0.2cm]
{\color{gray} \footnotesize{DEMO1}}  & You will be provided with a sentence in English. The task is \\
 & to output the correct sentence. \\ [0.1cm]
 & Input: There is many reasons to celebrate. \\
 & Output: There are many reasons to celebrate. \\ [0.2cm]
{\color{gray} \footnotesize{DEMO2}}  & You will be provided with a sentence in English. The task is \\
 & to output the correct sentence. \\ [0.1cm]
 & Input: Me and my friend goes to the gym every day. \\
 & Output: My friend and I go to the gym every day. \\ [0.2cm]
{\color{gray} \footnotesize{USER}}  & You will be provided with a sentence in English. The task is \\
 & to output the correct sentence. \\ [0.1cm]
 & Input: She don't like going to the park. \\
 & Output: \underline{\hspace{2em}}
\end{tabular}
\endgroup

\end{tcolorbox}
\vspace{0.5em}

In few-shot learning, we essentially provide a pattern that maps some inputs to the corresponding outputs. The LLM attempts to follow this pattern in making predictions, provided that the prompt includes a sufficient (though typically small) number of demonstrations. It is also possible to use simpler patterns to achieve this. For example, one can use the following few-shot learning prompt for translating words from Chinese to English.

\vspace{0.5em}
\begin{tcolorbox}[frame empty]

\begingroup
\renewcommand{\arraystretch}{1.0}
\setlength{\tabcolsep}{6pt}
\begin{tabular}{r r c l}
\hspace{0.5em} {\color{gray} \footnotesize{DEMO}} & \begin{CJK*}{UTF8}{gbsn} 现在 \end{CJK*} & $\to$ & now \\
& \begin{CJK*}{UTF8}{gbsn} 来 \end{CJK*} & $\to$ & come\\
& \begin{CJK*}{UTF8}{gbsn} 去 \end{CJK*} & $\to$ & go\\
& \begin{CJK*}{UTF8}{gbsn} 男孩 \end{CJK*} & $\to$ & boy\\
{\color{gray} \footnotesize{USER}} & \begin{CJK*}{UTF8}{gbsn} 女孩 \end{CJK*} & $\to$ & \underline{\hspace{2em}}\\
\end{tabular}
\endgroup

\end{tcolorbox}
\vspace{0.5em}

If the LLM is powerful enough, few-shot learning can enable it to address complex problems, such as mathematical reasoning. For example, consider the following task of summing two numbers and then dividing the sum by their product.

\vspace{0.5em}
\begin{tcolorbox}[frame empty]

\begingroup
\renewcommand{\arraystretch}{1.0}
\setlength{\tabcolsep}{6pt}
\begin{tabular}{r r c l}
\hspace{0.5em} {\color{gray} \footnotesize{DEMO}} & $12\ \ 5$ & $\to$ & $(12 + 5)/(12 \times 5) = 0.283$ \\
& $3\ \ 1$ & $\to$ & $(3 + 1)/(3 \times 1) = 1.33$ \\
& $-9\ \ 4$ & $\to$ & $(-9 + 4)/(-9 \times 4) = 0.138$ \\
& $15\ \ 15$ & $\to$ & $(15 + 15)/(15 \times 15) = 0.133$ \\
{\color{gray} \footnotesize{USER}} & $19\ \ 73$ & $\to$ & \underline{\hspace{2em}}\\
\end{tabular}
\endgroup

\end{tcolorbox}
\vspace{0.5em}

In many practical applications, the effectiveness of in-context learning relies heavily on the quality of prompts and the fundamental abilities of pre-trained LLMs. On one hand, significant prompt engineering effort is required to develop appropriate prompts that help LLMs learn more effectively from demonstrations. On the other hand, stronger LLMs can make better use of in-context learning for performing new tasks. For example, suppose we wish to use an LLM to translate words from Inuktitut to English. If the LLM lacks pre-training on Inuktitut data, its understanding of Inuktitut will be weak, and it will be difficult for the model to perform well in translation regardless of how we prompt it. In this case, we need to continue training the LLM with more Inuktitut data, rather than trying to find better prompts.

It is also interesting to explore how in-context learning emerges during pre-training and why it works during inference. One intuitive explanation is that while LLMs acquire problem-solving knowledge during pre-training, the large output space makes it difficult to select the correct prediction for novel tasks. Providing demonstrations guides the LLMs along the ``correct'' paths. Furthermore, some researchers have tried to interpret in-context learning from several different perspectives, including Bayesian inference~\cite{xie-etal:2022an}, gradient descent~\cite{dai-etal:2023can,von-etal:2023transformers}, linear regression~\cite{akyurek-etal:2023what}, meta-learning~\cite{garg-etal:2022can}, and so on.

\subsection{Prompt Engineering Strategies}

\noindent Designing prompts is highly empirical. In general, there are many ways to prompt an LLM to perform the same task, and we often need to conduct several trial-and-error iterations to find a satisfactory prompt. To write good prompts more efficiently, one can follow certain strategies. Examples of common prompting principles include:

\begin{itemize}
\item \vspace{0.5em} \textbf{Describing the task as clearly as possible}. When we apply an LLM to solve a problem, we need to provide a precise, specific, and clear description of the problem and instruct the LLM to behave as intended. This is particularly important when we want the output of the LLM to meet certain expectations. For example, suppose we are curious about climate change. A simple prompt for asking the LLM to provide some information is

    \vspace{0.5em}
    \begin{tcolorbox}[frame empty]

    \hspace{2em} Tell me about climate change.

    \hspace{2em} \underline{\hspace{2em}}

    \end{tcolorbox}
    \vspace{0.5em}

    Since this instruction is too general, the LLM may generate a response that addresses any aspect of climate change, which may not align with our specific interests. In this case, we can instead use prompts that are specific and detailed. One such example is

    \vspace{0.5em}
    \begin{tcolorbox}[frame empty]

    \begingroup
    \setlength{\leftskip}{2em}
    \setlength{\rightskip}{2em}

    Provide a detailed explanation of the causes and effects of climate change, including the impact on global temperatures, weather patterns, and sea levels. Also, discuss possible solutions and actions being taken to mitigate these effects.

    \underline{\hspace{2em}}

    \endgroup

    \end{tcolorbox}
    \vspace{0.5em}

    Now suppose we intend to explain climate change to a 10-year-old child. We can adjust the above prompt further.

    \vspace{0.5em}
    \begin{tcolorbox}[frame empty]

    \begingroup
    \setlength{\leftskip}{2em}
    \setlength{\rightskip}{2em}

    Explain the causes and effects of climate change to a 10-year-old child. Talk about how it affects the weather, sea levels, and temperatures. Also, mention some things people are doing to help. Try to explain in simple terms and do not exceed 500 words.

    \underline{\hspace{2em}}

    \endgroup

    \end{tcolorbox}
    \vspace{0.5em}

\item \vspace{0.3em} \textbf{Guiding LLMs to think}. LLMs have exhibited a surprisingly strong ability to ``think''. A common example is that well-developed LLMs have achieved impressive performance in mathematical reasoning tasks, which are considered challenging. In prompt engineering, the thinking  ability of LLMs needs to be activated through appropriate prompting, especially for problems that require significant reasoning efforts. In many cases, an LLM that is instructed to ``think'' can produce completely different results compared with instructing the same model to perform the task directly. For example, \citet{kojima-etal:2022large} found that simply appending ``Let's think step by step'' to the end of each prompt can improve the performance of LLMs on several reasoning tasks. LLMs can be prompted to ``think'' in a number of ways. One method is to instruct LLMs to generate steps for reasoning about the problem before reaching the final answer. For example, consider a task of solving mathematical problems. The following is a simple prompt for this task.

    \vspace{0.5em}
    \begin{tcolorbox}[frame empty]

    \begingroup
    \setlength{\leftskip}{2em}
    \setlength{\rightskip}{2em}

    You are a mathematician. You will be provided with a math problem. Please solve the problem.

    \underline{\hspace{2em}}

    \endgroup

    \end{tcolorbox}
    \vspace{0.5em}

    Since solving math problems requires a detailed reasoning process, LLMs would probably make mistakes if they attempted to work out the answer directly. So we can explicitly ask LLMs to follow a given reasoning process before coming to a conclusion.

    \vspace{0.5em}
    \begin{tcolorbox}[frame empty]

    \begingroup
    \setlength{\leftskip}{2em}
    \setlength{\rightskip}{2em}

    You are a mathematician. You will follow these detailed reasoning steps when solving math problems.

    \vspace{0.5em}

    Step 1: Problem Interpretation.

    The mathematician carefully listens to your query and understands the intricate details of the mathematical challenge you have presented.

    \vspace{0.5em}

    Step 2: Strategy Formulation.

    Drawing upon their extensive knowledge, the mathematician chooses the most effective strategy tailored to the type of math problem, whether it is algebra, calculus, or geometry.

    \vspace{0.5em}

    Step 3: Detailed Calculation.

    With precision and expertise, the mathematician performs the necessary calculations step by step, adhering to all mathematical principles.

    \vspace{0.5em}

    Step 4: Solution Review.

    Before providing the final answer, the mathematician meticulously checks the calculations for accuracy and offers a concise explanation or rationale for the solution.

    \vspace{0.5em}

    You will be provided with a math problem. Please solve the problem.

    \vspace{0.5em}

    $\{*\mathrm{problem}*\}$

    \vspace{0.5em}

    \underline{\hspace{2em}}

    \endgroup

    \end{tcolorbox}
    \vspace{0.5em}

    Another method to guide LLMs to ``think'' is through multiple rounds of interaction with LLMs. For example, as a first step, we can instruct LLMs to solve the problem directly

    \vspace{0.5em}
    \begin{tcolorbox}[frame empty]

    \begingroup
    \setlength{\leftskip}{2em}
    \setlength{\rightskip}{2em}

    You will be provided with a math problem. Please solve the problem.

    \vspace{0.3em}

    $\{*\mathrm{problem}*\}$

    \underline{\hspace{2em}}

    \endgroup

    \end{tcolorbox}
    \vspace{0.5em}

    Now we have an initial answer to the problem. As a second step, we prompt LLMs to evaluate the correctness of the answer and, if necessary, rework it to find a better solution.

    \vspace{0.5em}
    \begin{tcolorbox}[frame empty]

    \begingroup
    \setlength{\leftskip}{2em}
    \setlength{\rightskip}{2em}

    You will be provided with a math problem, along with a solution. Evaluate the correctness of this solution, and identify any errors if present. Then, work out your own solution.

    \vspace{0.3em}

    Problem: $\{*\mathrm{problem}*\}$

    \vspace{0.5em}

    Solution: $\{*\mathrm{solution}*\}$

    \underline{\hspace{2em}}

    \endgroup

    \end{tcolorbox}
    \vspace{0.5em}

    The prompts presented here are closely related to a long line of research on reasoning problems in LLMs. It is impossible to provide a complete discussion of all related issues because this topic covers a large family of methods. But we will see a relatively more detailed discussion on how to improve prompting through more reasoning in Section~\ref{sec:advanced-prompting}.

\item \vspace{0.3em} \textbf{Providing reference information}. As discussed in the previous section, we can include demonstrations in prompts and allow LLMs to learn in-context from these demonstrations how to perform the task. In fact, given the remarkable language understanding abilities of LLMs, we can add any type of text into the prompts and so these models can predict based on enriched contexts. In many applications, we have a variety of information that is relevant to user queries. Instead of using LLMs to make unconstrained predictions, we often want LLMs to produce outputs that are confined to the relevant text. One such example is RAG, where the relevant text for the user query is retrieved using an IR system, and we prompt LLMs to generate responses based on this provided relevant text. The following prompt shows an example.

    \vspace{0.5em}
    \begin{tcolorbox}[frame empty]

    \begingroup
    \setlength{\leftskip}{2em}
    \setlength{\rightskip}{2em}

    You are an expert that can generate answers to input queries. You have now been provided with a query and the corresponding context information. Please generate an answer based on this context information. Note that you need to provide the answer in your own words, not just copy from the context provided.

    \vspace{0.5em}

    Context information: $\{*\mathrm{IR\textrm{-}result}*\}$

    \vspace{0.3em}

    Query: $\{*\mathrm{query}*\}$

    \vspace{0.3em}

    \underline{\hspace{2em}}

    \endgroup

    \end{tcolorbox}
    \vspace{0.5em}

    If the context information is highly reliable, we can even restrict LLMs to answering using only the provided text. An example prompt is shown as follows

    \vspace{0.5em}
    \begin{tcolorbox}[frame empty]

    \begingroup
    \setlength{\leftskip}{2em}
    \setlength{\rightskip}{2em}

    You are an expert tasked with generating answers from input queries. You have been provided with a query and corresponding context information, organized in a table where each row represents a useful record. Please generate an answer using only this context information. Ensure that you provide the answer in your own words.

    \vspace{0.5em}

    Context information: $\{*\mathrm{table}*\}$

    \vspace{0.3em}

    Query: $\{*\mathrm{query}*\}$

    \vspace{0.3em}

    \underline{\hspace{2em}}

    \endgroup

    \end{tcolorbox}
    \vspace{0.5em}

    When dealing with real-world problems, we often have prior knowledge and additional information about the problems that help produce better answers. Considering such information in prompting is generally helpful in improving the result.

\item \vspace{0.3em} \textbf{Paying attention to prompt formats}. In general, the performance of LLMs is highly sensitive to the prompts we input. Sometimes a small modification to a prompt can lead to a big change in model output. An interesting example is that changing the order of sentences in a prompt may cause LLMs to generate different results. To make prompts easy to read and reduce ambiguity, it is common to format them in a way that ensures clarity. One example is that we define several fields for prompts and fill different information in each field. Another example is that we can use code-style prompts for LLMs which can understand and generate both natural language and code. See the following for a code-style prompt that performs translation where one demonstration is presented.

    \vspace{0.5em}
    \begin{tcolorbox}[frame empty]

    \begingroup
    \setlength{\leftskip}{2em}
    \setlength{\rightskip}{2em}

    [English] = [I have an apple.]

    \vspace{0.3em}

    [German] = [Ich habe einen Apfel.]

    \vspace{0.3em}

    [English] = [I have an orange.]

    \vspace{0.3em}

    [German] = \underline{\hspace{2em}}

    \endgroup

    \end{tcolorbox}
    \vspace{0.5em}

    LLMs can receive text in various formats. This allows us to use control characters, XML tags, and specific formatting to represent complex data. Furthermore, it is useful to specify how the input and output should be formatted or structured. For example, we can delimit sections of text using quotes and prompt LLMs accordingly (e.g., adding a sentence like ``the input text is delimited by double quotes'' to the prompt).

\end{itemize}
\vspace{0.5em}

Above, we have discussed only a few strategies for writing good prompts. There are, of course, many such methods, and one needs to develop their own through practice. Interested readers can refer to various online documents for more information, such as OpenAI's manual on the GPT series models\footnote{See \url{https://platform.openai.com/docs/guides/prompt-engineering/six-strategies-for-getting-better-results}.}.

\subsection{More Examples}

\noindent In this subsection, we consider more examples of prompting LLMs to perform various NLP tasks. The motivation here is not to give standard prompts for these tasks, but rather to use simple examples to illustrate how LLMs can be prompted to address NLP problems.

\subsubsection{Text Classification}

\noindent Text classification is perhaps one of the most common problems in NLP. Many tasks can be broadly categorized as assigning pre-defined labels to a given text. Here, we consider the polarity classification problem in sentiment analysis. We choose polarity classification for illustration because it is one of the most popular and well-defined text classification tasks. In a general setup for polarity classification, the goal is to categorize a given text into one of three categories: negative, positive, or neutral. Below is a simple prompt for this task (for easy reading, we highlight the task description in the prompt).

\vspace{0.5em}
\begin{tcolorbox}[frame empty]

\begingroup
\setlength{\leftskip}{2em}
\setlength{\rightskip}{2em}

\ctext[RGB]{154, 245, 175}{Analyze the polarity of the following text and classify it as positive, negative, or neutral.}

\vspace{0.7em}

Text:

The service at the restaurant was slower than expected, which was a bit frustrating.

\vspace{0.7em}

\underline{The polarity of the text can be classified as negative.}

\endgroup

\end{tcolorbox}
\vspace{0.5em}

\noindent For completeness, we show the response generated by the LLM (shown in underlined text).

Although the answer is correct, the LLM produces a descriptive sentence rather than a discrete label. The challenge arises because LLMs are inherently designed to generate text rather than assign discrete labels. Thus, classification is treated as a text generation problem.  As a result, we need another system to map the LLM's output to the label space (a process referred to as \mindex{label mapping}). For instance, we extract ``negative'' from ``The polarity of the text can be classified as negative''. This is trivial in most cases because we can identify label words via simple heuristics. But occasionally, LLMs may not express the classification results using these label words. In this case, the problem becomes more complicated, as we need a mapping from generated text to predefined labels.

One method to induce output labels from LLMs is to reframe the problem as a cloze task. For example, the following shows a cloze-like prompt for polarity classification.

\vspace{0.5em}
\begin{tcolorbox}[frame empty]

\begingroup
\setlength{\leftskip}{2em}
\setlength{\rightskip}{2em}

\ctext[RGB]{154, 245, 175}{Analyze the polarity of the following text and classify it as positive, negative, or neutral.}

\vspace{0.7em}

Text:

The service at the restaurant was slower than expected, which was a bit frustrating.

\vspace{0.7em}

The polarity of the text is \underline{negative}

\endgroup

\end{tcolorbox}
\vspace{0.5em}

\noindent We can use LLMs to complete the text and fill in the blank with the most appropriate word. Ideally, the filled word should be $\mathrm{positive}$, $\mathrm{negative}$, or $\mathrm{neutral}$. However, LLMs are not guaranteed to generate these label words. One method to address this problem is to constrain the output space to the set of label words and select the one with the highest probability. Then, the output label is given by
\begin{eqnarray}
\mathrm{label} & = & \argmax_{y \in Y} \Pr(y|\mathbf{x}),
\end{eqnarray}

\noindent where $y$ denotes the token (or word) filled in the blank, and $Y$ denotes the set of label words\\ $\{\mathrm{positive}, \mathrm{negative}, \mathrm{neutral}\}$.

Another approach to generating labels with LLMs is to constrain the output with prompts. For example, we can prompt LLMs to predict from a predefined label set. Here is an example.

\vspace{0.5em}
\begin{tcolorbox}[frame empty]

\begingroup
\setlength{\leftskip}{2em}
\setlength{\rightskip}{2em}

\ctext[RGB]{154, 245, 175}{Analyze the polarity of the following text and classify it as positive, negative, or neutral.}

\vspace{0.7em}

Text:

The service at the restaurant was slower than expected, which was a bit frustrating.

\vspace{0.7em}

What is the polarity of the text?

\vspace{0.3em}

Just answer: positive, negative, or neutral.

\vspace{0.7em}

\underline{Negative}

\endgroup

\end{tcolorbox}
\vspace{0.5em}

Sentiment analysis is a common NLP problem that has generally been well understood by LLMs through pre-training or fine-tuning. Thus we can prompt LLMs using simple instructions to perform the task. However, for new classification problems, it may be necessary to provide additional details about the task, such as the classification standards, so that the LLMs can perform correctly. To do this, we can add a more detailed description of the task and/or demonstrate classification examples in the prompts. To illustrate, consider the following example.

\vspace{0.5em}
\begin{tcolorbox}[frame empty]

\begingroup
\setlength{\leftskip}{2em}
\setlength{\rightskip}{2em}

\ctext[RGB]{154, 245, 175}{Analyze the polarity of the following text and classify it as positive, negative, or neutral. Here's what each category represents:}

\vspace{0.5em}

\ctext[RGB]{154, 245, 175}{Positive: This indicates that the text conveys a positive emotion or attitude. For example, texts expressing happiness, satisfaction, excitement, or admiration are considered positive.}

\vspace{0.3em}

\ctext[RGB]{154, 245, 175}{Negative: This refers to a text that expresses a negative emotion or attitude. It encompasses feelings of sadness, anger, frustration, or criticism.}

\vspace{0.3em}

\ctext[RGB]{154, 245, 175}{Neutral: Neutral sentiment is used to describe texts that do not exhibit clear positive or negative emotions but instead convey informational, factual, or indifferent tones.}

\vspace{0.7em}

Text:

The service at the restaurant was slower than expected, which was a bit frustrating.

\vspace{0.7em}

What is the polarity of the text?

\vspace{0.7em}

\underline{Negative}

\endgroup

\end{tcolorbox}
\vspace{0.5em}

While it seems straightforward to use LLMs for classification problems, there are still issues that have not been well addressed. For example, when dealing with a large number of categories, it remains challenging to effectively prompt LLMs. Note that if we face a very difficult classification problem and have a certain amount of labeled data, fine-tuning LLMs or adopting ``BERT + classifier''-like architectures is also desirable.

\subsubsection{Information Extraction}

\noindent Many NLP tasks can be framed as information extraction problems, which involve identifying and extracting specific pieces of information from unstructured text. Such information includes named entities, relationships, events, and other relevant data points. The goal of information extraction is to transform raw data into a structured format that can be easily analyzed and used in various downstream applications.

As information extraction covers a wide range of problems, we cannot discuss them all here. Instead, we start with the task of named entity recognition --- a task that has long been a central focus in NLP. Named entity recognition is the task of identifying and classifying named entities in text. These entities often include proper nouns and other expressions such as dates, monetary values, and percentages. First, consider a simple example of extracting person names from a given text\footnote{The text is from \url{https://edition.cnn.com/travel}}.

\vspace{0.5em}
\begin{tcolorbox}[frame empty]

\begingroup
\setlength{\leftskip}{2em}
\setlength{\rightskip}{2em}

\ctext[RGB]{154, 245, 175}{Identify all person names in the provided text.}

\vspace{0.7em}

Text:

Is the UK really doing that badly or have travel trends shifted, post-pandemic? For Tom Jenkins, CEO of the European Tourism Organisation, it's the latter. ``I think the UK is doing perfectly well but we'll see more people going to Europe,'' he says of 2024, adding that there's ``allegedly a slight plateau of American demand for the UK.''

\vspace{0.7em}

\underline{The person name identified in the provided text is:}

\vspace{0.3em}

\underline{Tom Jenkins}

\endgroup

\end{tcolorbox}
\vspace{0.5em}

Next, we can expand the scope to extract all named entities.

\vspace{0.5em}
\begin{tcolorbox}[frame empty]

\begingroup
\setlength{\leftskip}{2em}
\setlength{\rightskip}{2em}

\ctext[RGB]{154, 245, 175}{Identify and classify all named entities in the provided text into categories such as person names, locations, dates, and organizations. List each entity with its type on one line.}

\vspace{0.7em}

Text:

Is the UK really doing that badly or have travel trends shifted, post-pandemic? For Tom Jenkins, CEO of the European Tourism Organisation, it's the latter. ``I think the UK is doing perfectly well but we'll see more people going to Europe,'' he says of 2024, adding that there's ``allegedly a slight plateau of American demand for the UK.''

\vspace{0.7em}

\underline{1. Tom Jenkins - Person Name}

\vspace{0.3em}

\underline{2. European Tourism Organisation - Organization}

\vspace{0.3em}

\underline{3. UK - Location}

\vspace{0.3em}

\underline{4. Europe - Location}

\vspace{0.3em}

\underline{5. 2024 - Date}

\endgroup

\end{tcolorbox}
\vspace{0.5em}

Given these identified named entities, we can apply other information extraction tasks to the text. For example, we can identify the relationships between these named entities (call it \mindex{relation extraction}). The corresponding prompt is given below.

\vspace{0.5em}
\begin{tcolorbox}[frame empty]

\begingroup
\setlength{\leftskip}{2em}
\setlength{\rightskip}{2em}

\ctext[RGB]{154, 245, 175}{Given a text and a list of named entities identified within it, analyze and describe the relationships between these entities. Explain how each entity is contextually related to others.}

\vspace{0.7em}

Text:

Is the UK really doing that badly or have travel trends shifted, post-pandemic? For Tom Jenkins, CEO of the European Tourism Organisation, it's the latter. ``I think the UK is doing perfectly well but we'll see more people going to Europe,'' he says of 2024, adding that there's ``allegedly a slight plateau of American demand for the UK.''

\vspace{0.7em}

Named Entity:

Tom Jenkins, European Tourism Organisation, UK, Europe, 2024

\vspace{0.7em}

\uline{Tom Jenkins - European Tourism Organisation: Tom Jenkins is the CEO of the European Tourism Organisation. This indicates a professional association where Tom Jenkins likely has a significant role in directing the organization's activities, especially those related to tourism in Europe.}

\vspace{0.3em}

\uline{...}

\endgroup

\end{tcolorbox}
\vspace{0.5em}

If an LLM has been fine-tuned to follow instructions, adapting it to various information extraction tasks is relatively straightforward. For example, the following is a general prompt template for information extraction.

\vspace{0.5em}
\begin{tcolorbox}[frame empty]

\begingroup
\setlength{\leftskip}{2em}
\setlength{\rightskip}{2em}

\ctext[RGB]{154, 245, 175}{You will be provided with a text. Your task is to $\{*\mathrm{task\textrm{-}description}*\}$}

\vspace{0.7em}

Text: $\{*\mathrm{text}*\}$

\vspace{0.7em}

\underline{\hspace{2em}}

\endgroup

\end{tcolorbox}
\vspace{0.5em}

\noindent Here $\mathrm{task\textrm{-}description}$ is the instruction for the task. Examples of such instructions include

\vspace{0.3em}
\hspace{2em} \textit{extract the keywords from the text.}

\vspace{0.3em}
\hspace{2em} \textit{extract the key events from the text.}

\vspace{0.3em}
\hspace{2em} \textit{detect the coreference and anaphoric links between entities from the text.}

\vspace{0.3em}

\noindent Typically, providing a more detailed task description will yield more accurate responses from the LLM.

\subsubsection{Text Generation}

\noindent In NLP, the term \textit{text generation} is used broadly to cover a wide range of problems, many of which focus on generating new content based on an input. These problems can be roughly categorized into two classes: 1) continuing from an initial input text (call it \mindex{text completion}), and 2) converting the input text into another text (call it \mindex{text transformation}). Here, we present examples of prompts for the former and discuss the latter later.

A simple example of text completion is continuing to write a story starting from a few initial sentences.

\vspace{0.5em}
\begin{tcolorbox}[frame empty]

\begingroup
\setlength{\leftskip}{2em}
\setlength{\rightskip}{2em}

\ctext[RGB]{154, 245, 175}{You are a writer who is great at writing stories. You are provided with a few sentences to start with. Please continue the story from the provided sentences.}

\vspace{0.7em}

Start of the story:

One sunny afternoon, Lily discovered a tiny, golden key hidden under a rock in her backyard. Curious, she picked it up and wondered what it might unlock. Just then, her cat, Whiskers, began to meow excitedly, leading her towards the old oak tree at the edge of the garden.

\vspace{0.7em}

\uline{Lily followed Whiskers, her heart pounding with excitement and a touch of nervousness. The old oak tree had always seemed mysterious to her, its gnarled branches casting long shadows even on the brightest days. As they reached the tree, Whiskers darted ahead and started pawing at something buried beneath a pile of autumn leaves ...}

\endgroup

\end{tcolorbox}
\vspace{0.5em}

Likewise, we can complete a conversation given the conversational context.

\vspace{0.5em}
\begin{tcolorbox}[frame empty]

\begingroup
\setlength{\leftskip}{2em}
\setlength{\rightskip}{2em}

\ctext[RGB]{154, 245, 175}{You are provided with a conversation between a tourist and a taxi driver. Please complete the conversation.}

\vspace{0.7em}

Tourist: Could you take me to the downtown museum, please?

\vspace{0.3em}

Taxi Driver: Of course! First time in the city?

\vspace{0.3em}

Tourist: Yes, any must-see places around?

\vspace{0.3em}

Taxi Driver: Definitely check out the botanical gardens after the museum. They're nearby and quite beautiful.

\vspace{0.3em}

\uline{Tourist: Great, thanks for the tip!}

\vspace{0.3em}

\uline{Taxi Driver: No problem. Enjoy your visit!}

\endgroup

\end{tcolorbox}
\vspace{0.5em}

Beyond simple continuation, this paradigm can be extended to instruction-driven generation tasks, where the model ``completes'' a response based on user requirements. These requirements can specify the desired style, tone, length, and content. For instance, users may request a formal report, a creative story, or programming code. Below is an example of a prompt for generating a poem based on predefined constraints.

\vspace{0.5em}
\begin{tcolorbox}[frame empty]

\begingroup
\setlength{\leftskip}{2em}
\setlength{\rightskip}{2em}

\ctext[RGB]{154, 245, 175}{You are a poet. Please write a poem based on the following requirements.}

\vspace{0.7em}

Theme: Spring's rejuvenation

\vspace{0.3em}

Structure: Four-line poem with a consistent rhythm

\vspace{0.3em}

Emotion: Convey a sense of happiness and renewal through vivid spring imagery.

\vspace{1.0em}

\underline{Spring awakens beneath the golden sun}

\vspace{0.7em}

\underline{Fresh blossoms dance with morning light}

\vspace{0.3em}

\underline{Green fields whisper gentle songs of joy}

\vspace{0.3em}

\underline{New hopes arise with every spring breeze}

\endgroup

\end{tcolorbox}
\vspace{0.5em}

Furthermore, when an LLM is pre-trained or fine-tuned on a corpus encompassing both natural language and programming languages, we can prompt it to perform text-to-code generation tasks. In this case, the model simply translates natural language instructions into functional code. The following example shows how to prompt the LLM to write a Python function.

\vspace{0.5em}
\begin{tcolorbox}[frame empty]

\begingroup
\setlength{\leftskip}{2em}
\setlength{\rightskip}{2em}

\ctext[RGB]{154, 245, 175}{Please write a Python function to calculate the average of a list of numbers.}

\vspace{0.7em}

\uline{def calculate\_average(numbers):}

\hspace{2em} \uline{if numbers:\ \ \ \ \ \# Check if the list is not empty}

\hspace{2em} \hspace{2em} \uline{return sum(numbers) / len(numbers)}

\hspace{2em} \uline{else:}

\hspace{2em} \hspace{2em} \uline{return 0\ \ \ \# Return 0 if the list is empty}

\endgroup

\end{tcolorbox}
\vspace{0.5em}

\subsubsection{Text Transformation}

\noindent Typical text transformation tasks include machine translation, summarization, text style transfer, and so on. The prompts used to perform these tasks are not significantly different from the ones presented previously. Here, we present a few examples to illustrate these prompts.

\vspace{0.5em}
\begin{tcolorbox}[frame empty]

\begingroup
\setlength{\leftskip}{2em}
\setlength{\rightskip}{2em}

\ctext[RGB]{154, 245, 175}{Translate the following text from English to Spanish.}

\vspace{0.7em}

Text:
\vspace{0.3em}

The quick brown fox jumps over the lazy dog.

\vspace{0.7em}

\underline{El r{\'a}pido zorro marr{\'o}n salta sobre el perro perezoso.}

\endgroup

\end{tcolorbox}
\vspace{0.5em}

\vspace{0.5em}
\begin{tcolorbox}[frame empty]

\begingroup
\setlength{\leftskip}{2em}
\setlength{\rightskip}{2em}

\ctext[RGB]{154, 245, 175}{Summarize the following article in no more than 50 words:}

\vspace{0.7em}

Article:
\vspace{0.3em}

In recent years, urban areas have been facing significant challenges related to sustainability and environmental management. Cities are experiencing higher rates of pollution, increased traffic congestion, and greater demands on infrastructure. This growth has led to numerous environmental issues, including elevated levels of air and water pollution, increased waste production, and strained public services ...

\vspace{0.7em}

\uline{Urban areas are grappling with sustainability challenges, such as rising pollution, traffic congestion, and infrastructure demands ...}

\endgroup

\end{tcolorbox}
\vspace{0.5em}

\vspace{0.5em}
\begin{tcolorbox}[frame empty]

\begingroup
\setlength{\leftskip}{2em}
\setlength{\rightskip}{2em}

\ctext[RGB]{154, 245, 175}{Rewrite this text in a formal tone.}

\vspace{0.7em}

Text:
\vspace{0.3em}

Hey, what's up? Long time no see!

\vspace{0.7em}

\uline{Hello, how have you been? It has been quite some time since we last met!}

\endgroup

\end{tcolorbox}
\vspace{0.5em}

\subsubsection{Question Answering}

\noindent In question answering, a system is given a question and is expected to produce an appropriate answer. Prompt templates for general-purpose question answering can be:

\vspace{0.5em}
\begin{tcolorbox}[frame empty]

\begingroup
\setlength{\leftskip}{2em}
\setlength{\rightskip}{2em}

$\{*\mathrm{question}*\}$

\vspace{0.3em}

\underline{\hspace{2em}}

\endgroup

\end{tcolorbox}
\vspace{0.5em}

\vspace{0.5em}
\begin{tcolorbox}[frame empty]

\begingroup
\setlength{\leftskip}{2em}
\setlength{\rightskip}{2em}

Question: $\{*\mathrm{question}*\}$

\vspace{0.3em}

Answer: \underline{\hspace{2em}}

\endgroup

\end{tcolorbox}
\vspace{0.5em}

\vspace{0.5em}
\begin{tcolorbox}[frame empty]

\begingroup
\setlength{\leftskip}{2em}
\setlength{\rightskip}{2em}

Q: $\{*\mathrm{question}*\}$

\vspace{0.3em}

A: \underline{\hspace{2em}}

\endgroup

\end{tcolorbox}
\vspace{0.5em}

Question answering is important in NLP because many problems can be framed as question-answering tasks. In particular, many recent reasoning tasks are defined in the form of question answering. For example, in the MMLU benchmark~\cite{hendrycks-etal:2021measuring}, each example consists of a multiple-choice question, and LLMs are required to select the correct answer. The following shows an example prompt for answering a question in this dataset.

\vspace{0.5em}
\begin{tcolorbox}[frame empty]

\begingroup
\setlength{\leftskip}{2em}
\setlength{\rightskip}{2em}

$(\mathbb{Z},*)$ is a group with $a*b = a+b+1$ for all $a$, $b$ in $\mathbb{Z}$. The inverse of $a$ is

\vspace{0.5em}

(A)\ \ \ $0$

\vspace{0.3em}

(B)\ \ \ $-2$

\vspace{0.3em}

(C)\ \ \ $a-2$

\vspace{0.3em}

(D)\ \ \ $(2+a)*-1$

\vspace{0.5em}

\underline{D}

\endgroup

\end{tcolorbox}
\vspace{0.5em}

Another widely-used benchmark is the GSM8K dataset~\cite{cobbe-etal:2021training}. It consists of thousands of grade school math word problems. For each problem, we use LLMs to generate a solution in natural language. For example, the following is a prompt used to solve problems in GSM8K.

\vspace{0.5em}
\begin{tcolorbox}[frame empty]

\begingroup
\setlength{\leftskip}{0em}
\setlength{\rightskip}{2em}

\begin{tabular}{r l l}
{\color{gray} \footnotesize{DEMO}} &
\multicolumn{2}{p{0.8\textwidth}}{
Q: Jerry's two daughters play softball on different teams. They each have 8 games this season. Each team practices 4 hours for every game they play. If each game lasts for 2 hours, how many hours will Jerry spend at the field watching his daughters play and practice altogether?

\vspace{0.3em}

A: Jerry will spend $8$ games * $2$ hours per game = $\ll 8*2=16 \gg 16$ hours watching one daughter play her games. He will spend $16 * 2 = \ll 16*2=32 \gg 32$ hours watching both daughters play their games. He will spend $8$ games * $4$ hours of practice = $\ll 8*4=32 \gg 32$ hours watching one daughter practice. He will spend $32 * 2 = \ll 32*2=64 \gg 64$ hours watching both daughters practice. He will spend a total of $32$ hours watching games + $64$ hours watching practice $= \ll 32+64=96 \gg 96$ hours. \#\#\#\# $96$

\vspace{0.7em}

Q: Mary bought six apples from the store. From the apples she bought, for each that Mary ate, she planted two trees from the remaining ones. How many apples did Mary eat?

\vspace{0.3em}

A: She planted eight trees. This means she used half of that amount of apples, which is $8$ trees $/ 2$ trees/apple $= \ll 8/2=4 \gg 4$ apples. That means that she planted four of the six apples she bought, leaving only $6$ apples $-$ $4$ apples $= \ll 6-4=2 \gg 2$ apples to be eaten. \#\#\#\# $2$

\vspace{0.7em}
} \\
{\color{gray} \footnotesize{USER}} &
\multicolumn{2}{p{0.8\textwidth}}{
Q: Boris has 100 apples. Beck has 23 fewer apples than Boris. If Boris gives Beck 10 apples, how many fewer apples does Beck have than Boris now?
\vspace{0.3em}

A: \uline{Boris starts with $100$ apples and gives Beck $10$ apples, leaving him with $100$ apples $- 10$ apples $= \ll 100-10=90 \gg 90$ apples. Initially, Beck has $23$ fewer apples than Boris, giving him a starting count of $100 - 23 = \ll 100-23=77 \gg 77$ apples. After receiving $10$ apples from Boris, Beck's total comes to $77$ apples $+ 10$ apples $= \ll 77+10=87 \gg 87$ apples. Consequently, Beck now has $90 - 87 = \ll 90-87=3 \gg 3$ fewer apples than Boris. \#\#\#\# 3}
}
\end{tabular}

\endgroup

\end{tcolorbox}
\vspace{0.5em}

\noindent Here a few-shot prompt is adopted. The LLM learns from these demonstrations of problem-solution pairs not only how to solve problems but also how to format the output. For example, the final result of the calculation appears after the \#\#\#\# token, and $\ll \cdots \gg$ annotates the detailed calculation steps (called \mindex{calculation annotation})\footnote{During prediction, a calculator is used when we see $\ll \cdots \gg$. More specifically, once the LLM encounters ``$=$'' in a $\ll \cdots \gg$, then the calculator calculates the expression on the left-hand side of ``=''. This method helps reduce the calculation errors made by LLMs.}.

%----------------------------------------------------------------------------------------
% NEW SECTION
%----------------------------------------------------------------------------------------

\section{Advanced Prompting Methods}
\label{sec:advanced-prompting}

\noindent So far in this chapter, we have introduced the basic concepts related to LLM prompting and presented a number of prompts for NLP tasks. We now consider several techniques for enhancing the effectiveness of prompting.

\subsection{Chain of Thought}
\label{sec:chain-of-thought}

\noindent We have encountered the concept of \mindex{chain of thought} (\mindex{CoT}) several times in this chapter and previous ones~\cite{wei-etal:2022chain,chowdhery-etal:2022palm}. CoT methods provide a simple way to prompt LLMs to generate step-by-step reasoning for complex problems, thereby approaching tasks in a more human-like manner. Rather than coming to a conclusion directly, the CoT methods instruct LLMs to generate reasoning steps or to learn from demonstrations of detailed reasoning processes. To illustrate CoT, we consider the problem of algebraic calculation, as commonly described in the literature. Suppose we are given an algebraic problem

\vspace{0.8em}

\hspace{5em} \textit{Calculate the average of the numbers 2, 4, and 9.}

\vspace{0.8em}

\noindent We can pose this as a question to prompt an LLM.

\vspace{0.5em}
\begin{tcolorbox}[frame empty]

\begingroup
\setlength{\leftskip}{2em}
\setlength{\rightskip}{2em}

Q: Please calculate the average of the numbers $2$, $4$, and $9$.

\vspace{0.3em}

A: \underline{The answer is $6$.}

\endgroup

\end{tcolorbox}
\vspace{0.5em}

For nontrivial reasoning tasks, directly prompting an LLM may not always yield the correct answer. A simple improvement is to add demonstrations of similar problems in the prompt, and thus the LLM can learn from these demonstrations.

\vspace{0.5em}
\begin{tcolorbox}[frame empty]

\begingroup
\setlength{\leftskip}{2em}
\setlength{\rightskip}{2em}

Q: Please calculate the average of the numbers $1$, $3$, $5$, and $7$.

\vspace{0.3em}

A: The answer is $4$.

\vspace{0.7em}

Q: Please calculate the average of the numbers $2$, $4$, and $9$.

\vspace{0.3em}

A: \underline{The answer is $7$.}

\endgroup

\end{tcolorbox}
\vspace{0.5em}

The problem here is that, although we have shown a similar question-answer pair, it remains difficult for the LLM to reason out the correct answer. In CoT, LLMs not only learn from the correspondence between questions and answers but also benefit from the detailed problem-solving steps that are used to derive the answers. To do this, we can incorporate some reasoning steps into the prompt to obtain a CoT prompt.

\vspace{0.5em}
\begin{tcolorbox}[frame empty]

\begingroup
\setlength{\leftskip}{2em}
\setlength{\rightskip}{2em}

Q: Please calculate the mean square of the numbers 1, 3, 5, and 7.

\vspace{0.3em}

A: \ctext[RGB]{154, 245, 175}{Calculate the square of each number: $1^2 = 1$, $3^2=9$, $5^2=25$, and $7^2=49$. Sum the squares, $1+9+25+49=84$. There are $4$ numbers in total. Divide the sum by the number of items, $84 / 4 = 21$.} The answer is $21$.

\vspace{0.7em}

Q: Please calculate the average of the numbers $2$, $4$, and $9$.

\vspace{0.3em}

A: \underline{\ctext[RGB]{154, 245, 175}{Calculate $2 + 4 + 9$, which equals $15$. There are three numbers. Divide the}}

\underline{\ctext[RGB]{154, 245, 175}{total sum by the count, resulting in $15/3=5$.} The answer is $5$.}

\endgroup

\end{tcolorbox}
\vspace{0.5em}

\noindent Here we highlight the reasoning steps in green. By providing detailed reasoning demonstrations, the LLM is encouraged to generate intermediate steps that lead to the correct answer.

There are several benefits to using CoT prompting. First, CoT allows LLMs to decompose complex problems into smaller, sequential reasoning steps. This resembles how humans solve complex problems step by step, making it particularly effective for tasks requiring detailed, multi-step reasoning. Second, CoT makes the reasoning process more transparent and interpretable. Since the intermediate reasoning steps are visible, we can better understand how a conclusion was reached. Third, if users can follow the logic behind the reasoning process, they will be more likely to trust the predictions of an LLM. This is particularly important when applying LLMs in fields like medicine, education, and finance. Fourth, CoT is an in-context learning approach, and thus, it is applicable to most well-trained, off-the-shelf LLMs. Moreover, CoT provides efficient ways to adapt LLMs to different types of problems. It can even inspire more creative solutions by exploring various alternative reasoning paths, which might not be obvious when arriving at a conclusion directly.

The method described above requires providing one or more examples of CoT reasoning, typically called the few-shot CoT method. By contrast, the zero-shot CoT method does not require such examples. It instead prompts LLMs to reason step-by-step by incorporating specific instructions into prompts. For example, below is a zero-shot CoT prompt.

\vspace{0.5em}
\begin{tcolorbox}[frame empty]

\begingroup
\setlength{\leftskip}{2em}
\setlength{\rightskip}{2em}

Q: Please calculate the average of the numbers $2$, $4$, and $9$.

\vspace{0.3em}

A: \ctext[RGB]{154, 245, 175}{Let's think step-by-step.}

\vspace{0.3em}

\uline{We have three numbers: $2$, $4$, and $9$. Add these numbers together, $2+4+9=15$. Determine how many numbers there are, which in this case is three. The average is calculated by dividing the total sum by the number of elements. Completing the division gives $15/3 = 5$. So the answer is $5$.}

\endgroup

\end{tcolorbox}
\vspace{0.5em}

Following the instruction ``Let's think step by step'', the LLM is prompted to generate detailed reasoning steps. As discussed in \citet{kojima-etal:2022large}'s work, prompting with such instructions may result in LLMs generating only the reasoning steps without a clear conclusion. In this case, a second round of prompting can be used to extract the answer from these reasoning steps. For example, \citet{kojima-etal:2022large} create a second prompt which combines both the input and output from the first round of prompting. Using this combined input, the LLM can continue its reasoning process and then generate the correct answer. Furthermore, it is possible to prompt LLMs to reason using instructions other than ``Let's think step by step'', such as ``Let's think logically'' and ``Please show me your thinking steps first''.

While we have illustrated CoT methods using an algebraic reasoning problem, these methods can be applied to a variety of problems. Typical applications of CoT include mathematical reasoning, logical reasoning, commonsense reasoning, symbolic reasoning, code generation, etc. See Figure~\ref{fig:cot-application-examples} for more examples of applying CoT in various tasks.

\begin{figure}[!t]
\centering
\begin{tabular}{c c}
\begin{tcolorbox}[width=0.45\textwidth,height=8cm,valign=top]
\footnotesize{
Q: What home entertainment equipment requires cable?

Answer Choices: (a) radio shack (b) substation (c) television (d) cabinet

\vspace{0.2em}

A: \ctext[RGB]{154, 245, 175}{The answer must require cable. Of the above choices, only television requires cable.} So the answer is (c).

\vspace{0.5em}

Q: What type of person typically contracts illness?

Answer Choices: (a) hospital (b) head (c) sick person (d) elderly person (e) doctor's office

\vspace{0.2em}

A: \underline{\ctext[RGB]{154, 245, 175}{The answer must be a person suscepti-}} \underline{\ctext[RGB]{154, 245, 175}{ble to illness. Of the choices, "elderly per}} \underline{\ctext[RGB]{154, 245, 175}{son" is correct because they typically have}} \underline{\ctext[RGB]{154, 245, 175}{weaker immune systems.}} \underline{So the answer is (d).}
}
\end{tcolorbox}
&
\begin{tcolorbox}[width=0.45\textwidth,height=8cm,valign=top]
\footnotesize{
Q: Do hamsters provide food for any animals?

\vspace{0.2em}

A: \ctext[RGB]{154, 245, 175}{Hamsters are prey animals. Prey are food for predators. Thus, hamsters provide food for some animals.} So the answer is yes.

\vspace{0.5em}

Q: Can you use Microsoft Office without internet?

\vspace{0.2em}

A: \underline{\ctext[RGB]{154, 245, 175}{Yes, you can use Microsoft Office appli-}}

\underline{\ctext[RGB]{154, 245, 175}{cations, although features like cloud stor-}}

\underline{\ctext[RGB]{154, 245, 175}{age and online collaboration require inter-}}

\underline{\ctext[RGB]{154, 245, 175}{net.}} \underline{So the answer is yes.}
}
\end{tcolorbox}\\
\small{(a) CSQA} & \small{(b) StrategyQA} \\
\\
\begin{tcolorbox}[width=0.45\textwidth,height=8cm,valign=top]
\footnotesize{
Q: Complete the rest of the sequence, making sure that the parentheses are closed properly. Input: $[$ $\{$

\vspace{0.2em}

A: \ctext[RGB]{154, 245, 175}{Let's think step by step.}

\vspace{0.1em}

0: empty stack

\vspace{0.1em}

1: $[$ ; stack: $[$

\vspace{0.1em}

2: $\{$ ; stack: $[$ $\{$

\vspace{0.1em}

So the answer is $\}$ $]$.

\vspace{0.5em}

Q: Complete the rest of the sequence, making sure that the parentheses are closed properly. Input: $<$ $[$ $[$

\vspace{0.2em}

A: \underline{\ctext[RGB]{154, 245, 175}{Let's think step by step.}}

\vspace{0.1em}

\underline{0: empty stack}

\vspace{0.1em}

\underline{1: $<$ ; stack: $<$}

\vspace{0.1em}

\underline{2: $[$ ; stack: $<$ $[$}

\vspace{0.1em}

\underline{3: $[$ ; stack: $<$ $[$ $[$}

\vspace{0.1em}

\underline{So the answer is $]$ $]$ $>$.}
}
\end{tcolorbox}
&
\begin{tcolorbox}[width=0.45\textwidth,height=8cm,valign=top]
\footnotesize{
Q: Take the last letters of the words in "Elon Musk" and concatenate them.

\vspace{0.2em}

A: \ctext[RGB]{154, 245, 175}{The last letter of "Elon" is "n". The last letter of "Musk" is "k". Concatenating them is "nk".} The answer is nk.

\vspace{0.5em}

Q: Take the last letters of each word in "Renata Mariela Mona Kristin" and concatenate them.

\vspace{0.2em}

A: \underline{\ctext[RGB]{154, 245, 175}{The last letter of "Renata" is "a". The}}

\underline{\ctext[RGB]{154, 245, 175}{last letter of "Mariela" is "a". The last let-}}

\underline{\ctext[RGB]{154, 245, 175}{ter of "Mona" is "a". The last letter of}}

\underline{\ctext[RGB]{154, 245, 175}{"Kristin" is "n". Concatenating them is }}

\underline{\ctext[RGB]{154, 245, 175}{"aaan".}} \underline{The answer is aaan.}

}
\end{tcolorbox}\\
\small{(c) Dyck languages} & \small{(d) Last Letter Concatenation}
\end{tabular}
\caption{CoT in four different reasoning tasks, including CSQA, StrategyQA, Dyck languages, and Last Letter Concatenation. The CoT parts are highlighted in green.}
\label{fig:cot-application-examples}
\end{figure}

CoT is one of the most active research topics in prompt engineering. This has not only improved the performance of LLM prompting, but has also motivated a wide range of research on evaluating and verifying LLM reasoning capabilities. Although we have focused on the basic idea of CoT in this section, it can be improved in several ways. For example, we can consider the reasoning process as a problem of searching through many possible paths, each of which may consist of multiple intermediate states (i.e., reasoning steps). In general, we wish the search space to be well-defined and sufficiently large, so that we are more likely to find the optimal result. For this reason, an area of current LLM research is aimed at designing better structures for representing reasoning processes, allowing LLMs to tackle more complex reasoning challenges. These structures include tree-based structures~\cite{yao-etal:2024tree}, graph-based structures~\cite{besta-etal:2024graph}, etc. By using these compact representations of reasoning paths, LLMs can explore a wider range of decision-making paths, analogous to System 2 thinking\footnote{System 1 and System 2 thinking, as described by~\citet{kahneman:2011thinking}, represent two different modes of cognitive processing. System 1 is fast, automatic, intuitive, and emotional. This mode of thinking operates effortlessly and quickly, and is often what guides our daily decisions, judgments, and impressions. System 2 is slow, deliberate, and analytical. It is activated when we need to perform complex computations.}. Another line of research focuses on prompting LLMs with multi-round interactions. This involves decomposing complex problems into sub-problems, verifying and refining model outputs, employing model ensembling, and so on. Note that these methods and the issues involved are not limited to CoT. In fact, they are often used as more general approaches to improving LLMs, while CoT can be seen as a way to test the capabilities of LLMs. We will see discussions of some of these issues in the following subsections.

Before leaving our discussion of CoT, we should consider its practical limitations. One major limitation is the need for detailed, multi-step reasoning demonstrations in few-shot CoT scenarios, which may be difficult to obtain, either automatically or manually. Also, there is no standard method for breaking down complex problems into simpler problem-solving steps. This often heavily depends on the user's experience. In addition, errors in intermediate steps can also affect the accuracy of the final conclusion. For further discussion on the pros and cons of CoT, the interested reader can refer to recent surveys on this topic~\cite{chu-etal:2023survey,yu-etal:2023towards,zhang-etal:2023igniting}.

\subsection{Problem Decomposition}
\label{sec:problem-decomposition}

\noindent We have seen that LLMs can benefit from solving a complex problem by breaking it down into simpler sub-problems. Such an approach can be seen as an example of a broader paradigm known as \mindex{problem decomposition}, which has been extensively explored and discussed in psychology and computer science. From a psychological perspective, complex problem-solving refers to a process of addressing a problem using knowledge that helps overcome the barriers of the problem\footnote{A relatively formal definition can be found in \citet{frensch-and-funke:2014complex}'s book: \textit{complex problem-solving occurs to overcome barriers between a given state and a desired goal state by means of behavioral and/or cognitive, multi-step activities}.}.  There is often no clear path to solving a complex problem. However, it is often advantageous to employ strategies that decompose the problem, thereby making it easier to tackle the corresponding sub-problems with less effort. For example, consider writing a blog about the risks of AI. If we simply prompt an LLM with the instruction ``Please write a blog about the risks of AI'', the LLM may generate a blog with arbitrary structures and writing styles. A better method, instead, could be to outline the blog and provide more detailed information about each section. Consider the following prompt

\vspace{0.5em}
\begin{tcolorbox}[frame empty]

\begingroup
\setlength{\leftskip}{2em}
\setlength{\rightskip}{2em}

You are a blog writer. Please follow the provided outline below to write a blog about the risks of AI.

\vspace{0.5em}

\begin{itemize}
\item Introduction\\
Introduce AI, its relevance, and the importance of understanding its risks for youth.
\item Privacy Concerns\\
Discuss how AI might compromise personal privacy through interactions online.
\item Misinformation\\
Explore AI's role in spreading misinformation and influencing young people's decisions.
\item Cyberbullying\\
Highlight how AI tools can be utilized in cyberbullying and the impact on mental health.
\item Tips for Safe AI Use\\
Offer guidelines for responsible AI usage and promote critical thinking.
\item Conclusion\\
Recap main points and encourage proactive engagement with AI ethics.
\end{itemize}

\vspace{0.5em}

\underline{\hspace{2em}}

\endgroup

\end{tcolorbox}
\vspace{0.5em}

\noindent Here we give the title and major points for each section. Then, the LLM can use this structure to break down the writing task by filling in content for these sections. Note that the way to structure the blog can be provided by humans or even generated automatically. For example, we can use the LLM to first generate the outline, and then ask it to follow this outline to complete the writing.

In computer science, decomposing complex problems is a commonly used strategy in software and hardware system design. A well-known example is the divide-and-conquer paradigm, which is often used to design algorithms for computation problems that can be reduced to simpler, more manageable problems. For example, consider the problem of determining whether a document discusses the risks of AI. We can instruct the LLM with the following prompt.

\vspace{0.5em}
\begin{tcolorbox}[frame empty]

\begingroup
\setlength{\leftskip}{2em}
\setlength{\rightskip}{2em}

You are provided with a text. Please determine whether it discusses the risks of AI.

\vspace{0.5em}

$\{*\mathrm{document}*\}$

\vspace{0.5em}

\underline{\hspace{2em}}

\endgroup

\end{tcolorbox}
\vspace{0.5em}

If the document is long, inference can become computationally expensive. Alternatively, we can divide the document into relatively short segments and perform the same task on each segment. These segments can be processed in parallel to further reduce the computational cost. Next, we determine the relevance of each segment to the topic of AI risks. The final output is then generated using another prompt.

\vspace{0.5em}
\begin{tcolorbox}[frame empty]

\begingroup
\setlength{\leftskip}{2em}
\setlength{\rightskip}{2em}

Your task is to determine whether a text discusses the risks of AI. This text has been divided into segments, and you have obtained the relevance of each segment to the topic of AI risks. Based on this, please provide your final result.

\vspace{0.5em}

Segment 1: $\{*\mathrm{relevance\textrm{-}to\textrm{-}the\textrm{-}topic1}*\}$

\vspace{0.3em}

Segment 2: $\{*\mathrm{relevance\textrm{-}to\textrm{-}the\textrm{-}topic2}*\}$

\vspace{0.3em}

Segment 3: $\{*\mathrm{relevance\textrm{-}to\textrm{-}the\textrm{-}topic3}*\}$

\vspace{0.3em}

...

\vspace{0.5em}

\underline{\hspace{2em}}

\endgroup

\end{tcolorbox}
\vspace{0.5em}

Now let us return to a more general discussion of problem decomposition in prompting. While problem decomposition can be applied to various NLP problems, it has been more extensively discussed and tested in reasoning tasks recently. For complex reasoning tasks, we often need a multi-step reasoning path to reach a correct conclusion. We can use LLMs to perform such reasoning in three different ways. First, LLMs can directly reach the conclusion. In other words, they can predict without explicit reasoning processes, and the underlying reasoning process remains implicit and uninterpretable. Second, LLMs are prompted to generate a multi-step reasoning path that leads to the conclusion, like CoT. However, we run LLMs just once, and all intermediate steps in reasoning are generated in a single prediction. Third, we break down the original problem into a number of sub-problems, which are either addressed in separate runs of LLMs or tackled using other systems. Here we focus our attention on the third approach, which is closely related to problem decomposition. Note, however, that a more comprehensive discussion could cover all these approaches, while the first two have been discussed to some extent in this chapter.

A general framework for problem decomposition involves two elements.

\begin{itemize}
\item \vspace{0.5em} \mindex{Sub-problem Generation}. This involves decomposing the input problem into a number of sub-problems.
\item \vspace{0.3em} \mindex{Sub-problem Solving}. This involves solving each sub-problem and deriving intermediate and final conclusions through reasoning.
\end{itemize}
\vspace{0.5em}

These two issues can be modeled in different ways, leading to various problem decomposition methods. One approach is to treat them as separate steps in a two-step process. For example, consider the blog writing task described at the beginning of this subsection. In the first step, we decompose the entire problem into sub-problems all at once (i.e., outline the blog). In the second step, we solve the sub-problems either sequentially or in another order (i.e., fill in content for each section as needed). The final output of this process combines the results from solving each sub-problem. While this method is simple and straightforward, it assumes that the problem is compositional, making it more suitable for tasks like writing and code generation.

However, many real-world problems require complex reasoning. A key characteristic of such problems is that their reasoning paths are not fixed in advance and may depend on intermediate results obtained during reasoning. In such cases, it is undesirable to use fixed sub-problem generation in advance. Instead, sub-problems should be generated dynamically based on the input problem, and, if possible, generated on the fly during the reasoning process. This makes problem decomposition more challenging compared with designing divide-and-conquer algorithms. Ideally, we would like to jointly design both the systems for sub-problem generation and sub-problem solving. But a more practical and widely used approach is to adopt separate models for these tasks. A straightforward way to achieve this is to adapt an LLM for these tasks by either prompting or tuning the model.

Here we consider a method based on the above idea, called \mindex{least-to-most prompting}~\cite{zhou-etal:least}. The motivation for this method arises from the challenges of solving difficult reasoning problems --- those that cannot be addressed by simply generalizing from a few examples. For these problems, a more effective problem-solving strategy is to follow a progressive sequence of sub-problems that systematically lead to the conclusion. More specifically, in the least-to-most prompting method, sub-problem generation is performed by prompting an LLM with instructions and/or demonstrations. For example, below is a 2-shot prompt for sub-problem generation in least-to-most prompting.

\vspace{0.5em}
\begin{tcolorbox}[frame empty]

\begingroup
\setlength{\leftskip}{0em}
\setlength{\rightskip}{2em}

\begin{tabular}{r l l}
{\color{gray} \footnotesize{TASK}} & \multicolumn{2}{p{0.83\textwidth}}{
Your task is to decompose a problem into several sub-problems. You will be given a few examples to illustrate how to achieve this.
\vspace{0.5em}
} \\
{\color{gray} \footnotesize{DEMO}} &
\multicolumn{2}{p{0.83\textwidth}}{
Q: In a community, 5\% of the population are infants, 15\% are children, 40\% are adults, and 40\% are seniors. Which group makes up the largest portion of the population?

\vspace{0.3em}

A: To answer the question ``Which group makes up the largest portion of the population?'', we need to know: ``\ctext[RGB]{154, 245, 175}{How many percent are infants?}'', ``\ctext[RGB]{154, 245, 175}{How many percent are children?}'', ``\ctext[RGB]{154, 245, 175}{How many percent are adults?}'', ``\ctext[RGB]{154, 245, 175}{How many percent are seniors?}''.

\vspace{0.5em}

Q: Alice, Bob, and Charlie brought beads for their group project in their craft class. Alice has twice as many beads as Bob, and Bob has five times as many beads as Charlie. If Charlie has 6 beads, how many beads can they use for their craft project?

\vspace{0.3em}

A: To answer the question ``How many beads can they use for their craft project?'', we need to know: ``\ctext[RGB]{154, 245, 175}{How many beads does Bob have?}'', ``\ctext[RGB]{154, 245, 175}{How many beads does Alice have?}''.
\vspace{0.5em}
} \\
{\color{gray} \footnotesize{USER}} &
\multicolumn{2}{p{0.83\textwidth}}{
Q: The environmental study conducted from 2015 to 2020 revealed that the average temperature in the region increased by 2.3 degrees Celsius. What was the duration of the environmental study?

\vspace{0.3em}

A: \underline{To answer the question ``\ctext[RGB]{241, 148, 138}{What was the duration of the environmental}}

\underline{\ctext[RGB]{241, 148, 138}{study?}'', we need to know: ``\ctext[RGB]{153, 202, 240}{When did the environmental study start?}'',}

\underline{``\ctext[RGB]{237, 196, 107}{When did the environmental study end?}''.}
}
\end{tabular}

\endgroup

\end{tcolorbox}
\vspace{0.5em}

By learning from the examples, the LLM can generate two sub-problems for answering the new problem ``What was the duration of the environmental study?'' (highlighted in blue and orange). Given these sub-problems, we solve them sequentially. For each sub-problem, we take all previously-generated QA pairs as context, and then produce the answer. For the example above, we need to answer the first sub-problem by prompting the LLM, like this

\vspace{0.5em}
\begin{tcolorbox}[frame empty]

\begingroup
\setlength{\leftskip}{0em}
\setlength{\rightskip}{2em}

\begin{tabular}{r l l}
& \multicolumn{2}{p{0.78\textwidth}}{
The environmental study conducted from 2015 to 2020 revealed that the average temperature in the region increased by 2.3 degrees Celsius.
\vspace{0.5em}
} \\
{\color{gray} \footnotesize{SUB-PROB1}} &
\multicolumn{2}{p{0.78\textwidth}}{
Q: \ctext[RGB]{153, 202, 240}{When did the environmental study start?}

\vspace{0.3em}

A: \underline{The environmental study started in \ctext[RGB]{153, 202, 240}{2015}.}
}
\end{tabular}

\endgroup

\end{tcolorbox}
\vspace{0.5em}

Once we have the answer to the first sub-problem, we proceed to the second one. This time, we include both the first sub-problem and its corresponding answer in the input.

\vspace{0.5em}
\begin{tcolorbox}[frame empty]

\begingroup
\setlength{\leftskip}{0em}
\setlength{\rightskip}{2em}

\begin{tabular}{r l l}
& \multicolumn{2}{p{0.78\textwidth}}{
The environmental study conducted from 2015 to 2020 revealed that the average temperature in the region increased by 2.3 degrees Celsius.
 \vspace{0.5em}
} \\
{\color{gray} \footnotesize{SUB-PROB1}} &
\multicolumn{2}{p{0.78\textwidth}}{
Q: \ctext[RGB]{153, 202, 240}{When did the environmental study start?}

\vspace{0.2em}

A: The environmental study started in \ctext[RGB]{153, 202, 240}{2015}.
\vspace{0.5em}
}\\
{\color{gray} \footnotesize{SUB-PROB2}} &
\multicolumn{2}{p{0.78\textwidth}}{
Q: \ctext[RGB]{237, 196, 107}{When did the environmental study end?}

\vspace{0.2em}

A: \underline{The environmental study ended in \ctext[RGB]{237, 196, 107}{2020}.}
}
\end{tabular}

\endgroup

\end{tcolorbox}
\vspace{0.5em}

Finally, we use the LLM to solve the original problem given the answers to all the sub-problems.

\vspace{0.5em}
\begin{tcolorbox}[frame empty]

\begingroup
\setlength{\leftskip}{0em}
\setlength{\rightskip}{2em}

\begin{tabular}{r l l}
& \multicolumn{2}{p{0.78\textwidth}}{
The environmental study conducted from 2015 to 2020 revealed that the average temperature in the region increased by 2.3 degrees Celsius.
 \vspace{0.5em}
} \\
{\color{gray} \footnotesize{SUB-PROB1}} &
\multicolumn{2}{p{0.78\textwidth}}{
Q: \ctext[RGB]{153, 202, 240}{When did the environmental study start?}

\vspace{0.3em}

A: The environmental study started in \ctext[RGB]{153, 202, 240}{2015}.
\vspace{0.5em}
}\\
{\color{gray} \footnotesize{SUB-PROB2}} &
\multicolumn{2}{p{0.78\textwidth}}{
Q: \ctext[RGB]{237, 196, 107}{When did the environmental study end?}

\vspace{0.2em}

A: The environmental study ended in \ctext[RGB]{237, 196, 107}{2020}.
\vspace{0.5em}
}\\
{\color{gray} \footnotesize{FINAL}} &
\multicolumn{2}{p{0.78\textwidth}}{
Q: \ctext[RGB]{241, 148, 138}{What was the duration of the environmental study?}

\vspace{0.2em}

A: \underline{The duration of the environmental study was \ctext[RGB]{241, 148, 138}{5 years}}.
}
\end{tabular}

\endgroup

\end{tcolorbox}
\vspace{0.5em}

The least-to-most method offers a basic approach to prompting LLMs to generate and solve sub-problems separately. We can improve it in several ways. One simple improvement is to apply various advanced prompting techniques, which do not require changes to the problem decomposition framework. For example, we can incorporate CoT into the prompting to enhance the reasoning performance of sub-problem generation and solving.

Another improvement is to explore methods for better decomposing problems and organizing problem-solving paths. To describe these approaches, we will use the symbol $p_0$ to denote the input problem, and use the symbols $\{p_1,\cdots,p_n \}$ to denote the sub-problems corresponding to $p_0$. For least-to-most prompting, we decompose $p_0$ into $\{p_1,\cdots,p_n \}$, given by
\begin{eqnarray}
\{p_1,\cdots,p_n \} & = & G(p_0), \label{eq:least-to-most-problem-decomposition}
\end{eqnarray}

\noindent where $G(\cdot)$ denotes the function of sub-problem generation. Then, we solve the sub-problems $\{p_1,\cdots,p_n\}$ sequentially, yielding a sequence of answers $\{a_1,\cdots,a_n\}$. For answering the $i$-th sub-problem $p_i$, we include both the original problem $p_0$ and all previously generated problem-answer pairs in the context for prediction. The answer $a_i$ is given by
\begin{eqnarray}
a_i & = & S_i(p_i,\{p_0,p_{<i},a_{<i}\}), \label{eq:least-to-most-answer-i}
\end{eqnarray}

\noindent where $p_{<i}=\{p_1,\cdots,p_{i-1}\}$ and $a_{<i}=\{a_1,\cdots,a_{i-1}\}$. $S_i(\cdot)$ denotes the function that solves the sub-problem $p_i$ given the context $\{p_0,p_{<i},a_{<i}\}$. The last step is to generate the answer to the original problem $p_0$, which can be expressed in a similar manner to Eq.~(\ref{eq:least-to-most-answer-i}).
\begin{eqnarray}
a_0 & = & S_0(p_0,\{p_{\le n},a_{\le n}\}).
\end{eqnarray}

One way to refine this model is to modify the $G(\cdot)$ function so that the model can dynamically generate answers. Instead of generating all sub-problems at one time, we can generate each of them during problem-solving~\cite{dua-etal:2022successive}. To do this, we can replace Eq.~(\ref{eq:least-to-most-problem-decomposition}) with
\begin{eqnarray}
p_i & = & G_i(p_0,\{p_{<i},a_{<i}\}).
\end{eqnarray}

\noindent Hence we obtain a sub-problem generation model that operates in a step-by-step manner. At step $i$, we first prompt an LLM with the original problem $p_0$ together with the previous reasoning history $\{p_{<i},a_{<i}\}$ to generate a new sub-problem $p_i$. We then solve this sub-problem using the same or another LLM conditioned on the same context (see Eq.~(\ref{eq:least-to-most-answer-i})). This method effectively extends the reasoning capacity of LLMs by allowing them to dynamically generate and solve sub-problems in intermediate reasoning steps. As a result, the reasoning paths are not fixed in advance, and the models can choose and adapt their reasoning strategies during problem-solving.

Another way to improve the above model is to focus on developing better sub-problem solvers. In our previous discussion, we restricted $S_i(\cdot)$ to LLMs that are prompted to solve the sub-problem $p_i$. In fact, we can expand this function to any system that is capable of addressing the sub-problem. For example, $S_i(\cdot)$ could make calls to IR systems, thereby allowing us to access a broader range of data for problem-solving. Another example is using $S_i(\cdot)$ as a calculator to accurately compute results in mathematical problem-solving. If the sub-problem $p_i$ is complex and requires multiple intermediate problem-solving steps, it is also possible to further decompose $p_i$ into smaller sub-problems. For example, $S_i(\cdot)$ can be defined as a recursive program that generates and solves sub-problems. This incorporates recursion into problem-solving and allows us to address problems by iteratively decomposing them. As a result, we can define a hierarchical structure for problem-solving~\cite{khot2023:decomposed}.

A further generalization of this formulation is to view it as a reinforcement learning problem. A typical method is to model a problem-solving process as a decision making process. In each step of this process, an action is taken based on the current state. These actions can include all functions for sub-problem generation and solving (i.e., $G_i(\cdot)$ and $S_i(\cdot)$). Thus, the action sequence corresponds to a problem-solving path. Since the discussion of reinforcement learning problems is beyond the scope of this chapter, we skip the precise description of this learning task. Nevertheless, developing an agent or controller to determine when and how to generate and solve a sub-problem is also a natural choice.

In NLP, problem decomposition is related to a long line of research on multi-hop question answering~\cite{mavi-etal:2024multi}. This task requires the system to gather and combine information from multiple pieces of text to provide an accurate answer to a complex question. For example, to answer the question ``What is the capital of the country where Albert Einstein was born?'', we need to know ``Where was Albert Einstein born?'' and ``What is the capital of Germany?''. Earlier work in this area and related ones has investigated the issue of problem decomposition, though the methods might not be based on LLMs. For example, a popular method is to develop an additional neural model to generate simpler questions that address different aspects of the original question~\cite{andreas-etal:2016neural,talmor-etal:2018web,min-etal:2019multi}. This question generator can create questions in a batch or sequential manner.

Broadly speaking, problem decomposition is also related to the compositionality issue in NLP~\cite{drozdov-etal:2022compositional,press-etal:2023measuring}. For example, in semantic parsing, we map natural language sentences into structured meaning representations by breaking them down into constituent parts and understanding the sentences based on the meanings of these parts and the rules used to combine them. In early studies of this field, highly compositional sentences were considered easier test cases for evaluating systems, as it is relatively straightforward to decompose such sentences and compose the meanings of their parts. However, the task becomes much more difficult when more generalization is required for modeling compositionality in new data. In this case, we want systems to have improved abilities of \mindex{compositional generalization}. In more recent research on LLMs, this issue has been frequently discussed in compositional reasoning tasks, such as SCAN\footnote{The SCAN tasks (Simplified versions of the CommAI Navigation tasks) are designed to evaluate the ability of LLMs to perform compositional generalization~\cite{lake-etal:2018generalization}. They involve translating natural language commands into a sequence of actions. For example, a command ``jump opposite left and walk thrice'' can be translated into the action sequence ``LTURN LTURN JUMP WALK WALK WALK''.}, as it is considered an important aspect of testing the language understanding and reasoning abilities of LLMs. This also presents new tasks for developing and examining problem decomposition methods.

In LLMs, one interesting application of problem decomposition is tool use. In some cases, it is necessary to integrate external tools into LLMs to access accurate data not available during training or fine-tuning. For example, LLMs can integrate with APIs to fetch real-time data such as weather updates, stock market prices, or news feeds, enabling them to provide up-to-date responses to user queries. When using tools, LLM predictions might include markers that indicate where and how to call external APIs. This requires decomposing the problem into sub-problems, with some handled by the LLMs and others by external tools. More detailed discussions on this topic will be presented in Section~\ref{sec:rag-and-tool-use}.

\subsection{Self-refinement}
\label{sec:self-refinement}

\noindent In many cases, LLM predictions can be inaccurate or flawed. Given that current LLMs are capable of performing tasks such as refinement and correction, it is valuable to explore methods that enable these models to self-refine their outputs. Self-refinement is a common phenomenon in human cognitive processes and daily behavior. For example, when designing a product, a designer might first create a basic prototype and then refine the design after evaluation and testing to enhance user experience and functionality. This refinement process can be iterated several times until the design is satisfactory. The predict-then-refine paradigm has a long history in NLP. One early example is Brill's tagger~\cite{brill:1992simple}, where an initial POS tagging result of a sentence can be iteratively refined using a rule-based system. In the era of deep learning, a good deal of work on sequence-to-sequence problems, such as grammar correction and text rewriting, can also be viewed as examples of this paradigm.

We can prompt LLMs to do self-refinement. Consider a simple example where we first use an LLM to translate a text from Chinese to English and then use the same model to refine the translation. Here is how we can prompt the model to translate:

\vspace{0.5em}
\begin{tcolorbox}[frame empty]

\begingroup
\setlength{\leftskip}{2em}
\setlength{\rightskip}{2em}

Translate the following text from Chinese to English:

\vspace{0.5em}

Chinese:

\vspace{0.2em}

\begin{CJK*}{UTF8}{gbsn}
一系列考古发现奠定红山文化在中华文明起源研究中的重要地位。最新公布的研究成果认为，大约从距今5800年开始，中华大地上各个区域相继出现较为明显的社会分化，进入文明起源的加速阶段。
\end{CJK*}

\vspace{0.5em}

English:

\vspace{0.2em}

\uline{A series of discoveries have cemented the significant role of the Hongshan culture in studies on the origins of Chinese civilization. The latest research findings suggest that, starting from around 5800 years ago, many regions across China began to experience noticeable social differentiations, entering an accelerated phase in the origin of civilization.}

\endgroup

\end{tcolorbox}
\vspace{0.5em}

Then, we can prompt it to refine the translation:

\vfill

{\color{white} NiuTrans}

\vspace{0.5em}
\begin{tcolorbox}[frame empty]

\begingroup
\setlength{\leftskip}{2em}
\setlength{\rightskip}{2em}

You are provided with a Chinese text along with its English translation. Please refine the translation to improve its accuracy and fluency if necessary. \vfill

\vspace{0.5em}

Chinese:

\vspace{0.2em}

\begin{CJK*}{UTF8}{gbsn}
一系列考古发现奠定红山文化在中华文明起源研究中的重要地位。最新公布的研究成果认为，大约从距今5800年开始，中华大地上各个区域相继出现较为明显的社会分化，进入文明起源的加速阶段。
\end{CJK*}

\vspace{0.5em}

English:

\vspace{0.2em}

A series of discoveries have cemented the significant role of the Hongshan culture in studies on the origins of Chinese civilization. The latest research findings suggest that, starting from around 5800 years ago, many regions across China began to experience noticeable social differentiations, entering an accelerated phase in the origin of civilization.

\vspace{0.5em}

Refinement:

\vspace{0.2em}

\underline{A series of \ctext[RGB]{154, 245, 175}{archaeological} discoveries have cemented the significant role} \underline{of the Hongshan culture in studies on the origins of Chinese civilization.} \underline{The latest research findings \ctext[RGB]{154, 245, 175}{suggest} that, starting from around 5800 years} \underline{ago, \ctext[RGB]{154, 245, 175}{various} regions across China began to experience noticeable social} \underline{differentiations, entering an accelerated phase in the origin of civilization.}

\endgroup

\end{tcolorbox}
\vspace{0.5em}

\noindent Here modifications to the input translation are highlighted in green. This process can be iterative: at each step, we use the translation from the previous step as input, and prompt the LLM to produce a potentially better translation.

In the above example, the refinement of the initial model output was achieved by prompting the LLM with a simple instruction like ``Please refine it!''. However, the refinement is based solely on the instruction-following ability of the LLM, and there is no guidance or supervision on how and where to improve the model output. A more effective approach would be to consider feedback on specific aspects that need refinement. For example, we can prompt the LLM with ``Please correct all the grammatical errors in the translation'', so that the model can focus more on grammatical error correction during refinement.

A general framework of self-refinement with LLMs involves three steps~\cite{madaan-etal:2024self}.

\begin{itemize}
\item \vspace{0.5em} \textbf{Prediction}. We use an LLM to produce the initial model output.
\item \vspace{0.2em} \textbf{Feedback Collection}. We obtain feedback on the model output.
\item \vspace{0.2em} \textbf{Refinement}. We use the LLM to refine the model output based on the feedback.
\end{itemize}
\vspace{0.5em}

The last two steps can be repeated multiple times, which leads to an iterative self-refinement process. In this framework, receiving accurate and detailed feedback is critical as it guides the subsequent refinement step. In general, feedback can be obtained in various ways~\cite{fernandes-etal:2023bridging,pan-etal:2024automatically}. We can manually examine the model's predictions to identify issues. Also, we can use a feedback model to automatically generate feedback. For instance, a reward model trained on some labeled data can help evaluate the model output and return a score indicating its quality.

Since the focus of this chapter is on LLM prompting, we consider using LLMs to provide feedback. To illustrate, let us consider an example of generating good responses to user questions:

\begin{itemize}
\item \vspace{0.5em} First, we prompt an LLM to answer the input question, as usual.

\vspace{0.5em}
\begin{tcolorbox}[frame empty]

\begingroup
\setlength{\leftskip}{2em}
\setlength{\rightskip}{2em}

Generate a response to the question: ``What are some major environmental concerns today?''

\vspace{0.5em}

Response:

\vspace{0.2em}

\uline{Major environmental concerns today include climate change, air and water pollution, deforestation, loss of biodiversity, and ocean degradation.}

\endgroup

\end{tcolorbox}
\vspace{0.5em}

\item Then, using the following prompt, we prompt the same LLM again to get feedback on the generated response.

\vspace{0.5em}
\begin{tcolorbox}[frame empty]

\begingroup
\setlength{\leftskip}{2em}
\setlength{\rightskip}{2em}

You will be provided with a question and its corresponding response. Evaluate the accuracy and depth of this response, and identify any factual inaccuracies or areas where additional details could enhance understanding.

\vspace{0.5em}

Question:

\vspace{0.2em}

What are some major environmental concerns today?

\vspace{0.5em}

Response:

\vspace{0.2em}

Major environmental concerns today include climate change, air and water pollution, deforestation, loss of biodiversity, and ocean degradation.

\vspace{0.5em}

Feedback:

\vspace{0.2em}

\uline{The response provided is accurate and highlights key environmental concerns. However, it would benefit from more specific and detailed descriptions of these concerns. Here are suggestions for improvement:}

\vspace{0.2em}

\hspace{1em} \uline{1. Include specific examples of environmental issues.}

\vspace{0.1em}

\hspace{1em} \uline{2. Add information about major pollutants.}

\vspace{0.1em}

\hspace{1em} \uline{3. Specify the regions most affected.}

\endgroup

\end{tcolorbox}
\vspace{0.5em}

\item Given the feedback the LLM produces, we prompt it to refine the response.

\vspace{0.5em}
\begin{tcolorbox}[frame empty]

\begingroup
\setlength{\leftskip}{2em}
\setlength{\rightskip}{2em}

You will be provided with a question, its corresponding response, and some feedback on this response.

\vspace{0.5em}

Question:

\vspace{0.2em}

What are some major environmental concerns today?

\vspace{0.5em}

Response:

\vspace{0.2em}

Major environmental concerns today include climate change, air and water pollution, deforestation, loss of biodiversity, and ocean degradation.

\vspace{0.5em}

Feedback:

\vspace{0.2em}

The response provided is accurate and highlights key environmental concerns. However, it would benefit from more specific and detailed descriptions of these concerns. Here are suggestions for improvement:

\vspace{0.2em}

\hspace{1em} 1. Include specific examples of environmental issues.

\vspace{0.1em}

\hspace{1em} 2. Add information about major pollutants.

\vspace{0.1em}

\hspace{1em} 3. Specify the regions most affected.

\vspace{0.5em}

Now refine the response based on the provided feedback.

\vspace{0.2em}

Refined Response:

\vspace{0.2em}

\uline{Key environmental concerns include climate change, evidenced by rising global temperatures and severe weather patterns; significant air and water pollution in urban and industrial areas; rampant deforestation in regions like the Amazon, causing biodiversity loss; and ocean degradation, highlighted by coral reef bleaching and widespread overfishing.}

\endgroup

\end{tcolorbox}
\vspace{0.5em}

\end{itemize}

Ideally, if a strong LLM is adopted, we would like to have it perform all three steps without extra training. On the other hand, if we have enough labeled data for the task of interest, we can enhance the performance of the LLM using supervised learning. For example, we can fine-tune the LLM to better adapt it to refinement tasks, or alternatively, use task-specific models, which may not necessarily be based on LLMs~\cite{welleck-etal:2023generating,schick-etal:2023peer}. In a broader sense, improving LLMs for self-refinement tasks can be seen as an alignment issue. For example, it has been found that some self-correction abilities can be activated through RLHF~\cite{ganguli-etal:2023capacity}. However, discussing these issues is beyond the scope of this chapter. Further discussion can be found in Chapter \ChapterLLMTuning.

In the context of LLMs, self-refinement relates to several concepts that mimic advanced reasoning capabilities, such as the ability to self-reflect. One perspective is that endowing LLMs with self-reflection enables them to generate more accurate predictions and exhibit self-correction. This reflective capacity can be activated in various ways, such as by prompting the LLMs to engage in deliberate, step-by-step reasoning, or by providing in-context examples from which the models can learn and adjust. To illustrate, we examine the \mindex{deliberate-then-generate} (\mindex{DTG}) method proposed by \citet{li2023-etal:deliberate}, which explicitly prompts LLMs to deliberate. In the DTG framework, given an initial, potentially flawed model output, the LLM is prompted to identify the specific error types before providing an improved revision. Below is a DTG prompt template for Chinese-to-English translation tasks.

\vspace{0.5em}
\begin{tcolorbox}[frame empty]

\begingroup
\setlength{\leftskip}{2em}
\setlength{\rightskip}{2em}

Given the Chinese sentence: $\{*\mathrm{source}*\}$

\vspace{0.2em}

The English translation is: $\{*\mathrm{target}*\}$

\vspace{0.5em}

Please first detect the type of error, and then refine the translation.

\vspace{0.2em}

Error Type: \underline{\ctext[RGB]{241, 148, 138}{\ \ \ \ \ \ \ \ \ \ \ \ \ \ \ }}

\vspace{0.2em}

\underline{\ctext[RGB]{153, 202, 240}{\ \ \ \ \ \ \ \ \ \ \ \ \ \ \ }}

\endgroup

\end{tcolorbox}
\vspace{0.5em}

We aim to first predict the error type (red), and then produce a refined translation (blue). This process of deliberation is guided by the instruction ``Please first detect the type of error, and then refine the translation''. It encourages LLMs to initially engage in thoughtful analysis and then give better results. Since error type prediction and refinement are performed in a single run of the LLM, this method incorporates both steps of feedback and refinement into one process.

In the above prompts, we assume that the LLM we use is able to review the input translation and correctly identify its error types. However, this raises new difficulties as the model may not be good at finding errors in translations. This will in turn result in extra fine-tuning or prompt engineering efforts. So a simpler method is to reduce the burden of error identification and use LLMs for deliberation only. To do this, we can replace the input translation with a random translation and assign a default error type. An example of such a prompt is shown below.

\vspace{0.5em}
\begin{tcolorbox}[frame empty]

\begingroup
\setlength{\leftskip}{2em}
\setlength{\rightskip}{2em}

Given the Chinese sentence:

\vspace{0.2em}

\begin{CJK*}{UTF8}{gbsn}一系列考古发现奠定红山文化在中华文明起源研究中的重要地位。\end{CJK*}

\vspace{0.2em}

The English translation is:

\vspace{0.2em}

\ctext[RGB]{154, 245, 175}{A variety of innovative techniques have redefined the importance of modern art in contemporary cultural studies.}

\vspace{0.5em}

Please first detect the type of error, and then refine the translation.

\vspace{0.2em}

Error Type: \ctext[RGB]{241, 148, 138}{Incorrect Translation}

\vspace{0.2em}

\underline{\ctext[RGB]{153, 202, 240}{\ \ \ \ \ \ \ \ \ \ \ \ \ \ \ }}

\endgroup

\end{tcolorbox}
\vspace{0.5em}

In this example, the input translation is not generated by the LLM but is instead randomly sampled from the dataset. Thus, it is simply an incorrect translation for the source sentence, and we can set the error type accordingly. The LLM then generates a new translation by taking both the source sentence and the incorrect translation as input. The design of this prompt can also be considered as activating the learning capabilities of LLMs through ``negative evidence''~\cite{marcus-etal:1993negative}, thereby enabling them to reflect and produce better outcomes through contrastive analysis. Nevertheless, this method does not rely on any feedback and can enhance the performance of a single LLM prediction via simple prompting.

Note that while DTG is non-iterative, iterative learning and refinement are commonly used in NLP. An advantage of these iterative approaches is that they mimic human learning and problem-solving, where continuous feedback and adjustments lead to progressively improved outcomes. Iterative methods can be applied to a range of LLM prompting problems. For example, in problem decomposition, one can incorporate new sub-problems and their solutions into the context at each step, and thus LLMs can progressively approach the solution of the original problem. On the other hand, iterative methods raise several issues that are absent in non-iterative methods. For example, errors in earlier steps may negatively impact subsequent problem-solving, and determining when to stop iterating often requires additional engineering effort.

\subsection{Ensembling}

\noindent Model ensembling for text generation has been extensively discussed in the NLP literature. The idea is to combine the predictions of two or more models to generate a superior final output. This technique is directly applicable to LLMs. For example, we can collect a set of LLMs and run each of them on the same input. The final output is obtained by combining the predictions of these models.

For LLM prompting, it is also possible to improve performance by combining predictions based on different prompts. Suppose we have an LLM and a collection of prompts that address the same task. We can run this LLM with each of the prompts and then combine the predictions. For example, below are three different prompt templates for text simplification.

\vspace{0.5em}
\begin{tcolorbox}[frame empty]

\begingroup
\setlength{\leftskip}{2em}
\setlength{\rightskip}{2em}

Make this text simpler.

\vspace{0.2em}

$\{*\mathrm{text}*\}$

\underline{\hspace{2em}}

\endgroup

\end{tcolorbox}
\vspace{0.5em}

\vspace{0.5em}
\begin{tcolorbox}[frame empty]

\begingroup
\setlength{\leftskip}{2em}
\setlength{\rightskip}{2em}

Condense and simplify this text.

\vspace{0.2em}

$\{*\mathrm{text}*\}$

\underline{\hspace{2em}}

\endgroup

\end{tcolorbox}
\vspace{0.5em}

\vspace{0.5em}
\begin{tcolorbox}[frame empty]

\begingroup
\setlength{\leftskip}{2em}
\setlength{\rightskip}{2em}

Rewrite for easy reading.

\vspace{0.2em}

$\{*\mathrm{text}*\}$

\underline{\hspace{2em}}

\endgroup

\end{tcolorbox}
\vspace{0.5em}

\noindent Each of these prompts will lead to a different prediction, and we can combine these three predictions to produce the final output.

Formally, let $\{\mathbf{x}_1,\cdots,\mathbf{x}_K\}$ be $K$ prompts for performing the same task. Given an LLM $\Pr(\cdot|\cdot)$, we can find the most likely prediction for each $\mathbf{x}_i$ using $\hat{\mathbf{y}}_i= \argmax_{\mathbf{y}_i} \Pr(\mathbf{y}_i|\mathbf{x}_i)$. These predictions can be combined to form a ``new'' prediction:
\begin{eqnarray}
\hat{\mathbf{y}} & = & \mathrm{Combine}(\hat{\mathbf{y}}_1,\cdots,\hat{\mathbf{y}}_K). \label{eq:llm-combination-with-diff-prompts}
\end{eqnarray}

\noindent Here $\mathrm{Combine}(\cdot)$ is the combination model, which can be designed in several different ways. For example, we can select the best prediction by voting or by identifying the one that overlaps the most with others. Another approach is to combine token-level probability scores during decoding. Let $\hat{y}_{j}$ be the predicted token at the $j$-th step for model combination. The predicted token $\hat{y}_{j}$ is given by
\begin{eqnarray}
\hat{y}_{j} & = & \argmax_{y_j} \sum_{k=1}^{K} \log \Pr(y_j|\mathbf{x}_k,\hat{y}_1,\cdots,\hat{y}_{j-1}).
\end{eqnarray}

In ensembling for LLM prompting, it is generally advantageous to use diverse prompts so that the combination can capture a broader range of potential responses. This practice is common in ensemble learning, as diversity helps average out biases and errors that may be specific to any single model or configuration. From the Bayesian viewpoint, we can treat the prompt $\mathbf{x}$ as a latent variable, given the problem of interest, $p$. This allows the predictive distribution of $\mathbf{y}$ given $p$ to be written as the distribution $\Pr(\mathbf{y}|\mathbf{x})$ marginalized over all possible prompts
\begin{eqnarray}
\Pr(\mathbf{y}|p) & = & \int \Pr(\mathbf{y}|\mathbf{x}) \Pr(\mathbf{x}|p) d \mathbf{x}.
\end{eqnarray}

\noindent The integral computes the total probability of $\mathbf{y}$ by considering all possible values of $\mathbf{x}$, weighted by their probabilities given $p$. Here $\Pr(\mathbf{y}|\mathbf{x})$ is given by the LLM, and $\Pr(\mathbf{x}|p)$ is the prior distribution of prompts for the problem. This formulation is advantageous because the integral accounts for uncertainty in the choice of $\mathbf{x}$. As a result, the final predictive distribution $\Pr(\mathbf{y}|p)$ becomes more robust to variations and biases in individual prompts. However, computing this integral directly can be computationally infeasible due to the potentially infinite space of $\mathbf{x}$. One approach to addressing this issue is to employ methods like Monte Carlo sampling, which approximate the integral using a manageable, finite number of prompts.

While the Bayesian treatment is mathematically well-defined, it is common practice in NLP to assume a non-informative or uniform prior and focus instead on constructing a set of diverse prompts. Consequently, the output can be computed using a straightforward combination model, as described in Eq.~(\ref{eq:llm-combination-with-diff-prompts}). The issue of creating high-quality, diverse prompts has been studied in CoT and other in-context learning areas. Most of the research focuses on incorporating a variety of demonstration examples across different prompts. Here, we list some of these methods.

\begin{itemize}
\item \vspace{0.5em} Given a problem, we manually create a number of demonstrations and use different ones for different prompts.
\item \vspace{0.2em} Given a problem, we use LLMs to automatically generate demonstrations and prompts.
\item \vspace{0.2em} Given a prompt, we create different prompts by changing the order of demonstrations in the prompt.
\item \vspace{0.2em} Given a prompt, we use LLMs to generate a number of similar prompts.
\item \vspace{0.2em} Given a prompt, we transform it into other forms, e.g., translating it into other languages.
\end{itemize}
\vspace{0.5em}

Of course, in practice, we can combine these methods to achieve greater diversity. An underlying assumption here is that diverse prompts can lead to diverse model outputs. This is particularly true when dealing with relatively novel and complex problems. For stronger and more robust LLMs, the variance in output for similar prompts might not be large. In this case, the benefit of using multiple prompts can be modest.

In addition to providing diverse prompts for LLMs, another approach is to make use of the inherent variance in the outputs of LLMs. One simple way is to sample multiple outputs from the hypothesis space. This is straightforward for LLMs because their search algorithms are naturally suited to producing multiple results. For example, if beam search is used, we can easily collect all completed hypotheses in the beam to generate multiple output candidates. Also, we can modify the search algorithms to implement more powerful sampling methods and explore a wider range of high-quality hypotheses.

The next question is how to combine diverse model predictions. This leads us back to the model combination issue as described in Eq.~(\ref{eq:llm-combination-with-diff-prompts}). Here we consider the \mindex{self-consistency} method, which outputs not necessarily the highest-probability prediction, but rather the one that is most consistent with the other predictions~\cite{wang-etal:2022rationale,wang-etal:2023self}. First, an LLM is prompted with CoT as usual and generates multiple reasoning paths by sampling. Then, the frequency of each answer is counted across these reasoning paths. The final output is the answer that appears most frequently. Below are three different predictions generated by prompting an LLM with CoT.

\vspace{0.5em}
Prediction 1 (Correct):
\begin{tcolorbox}[frame empty]

\begingroup
\setlength{\leftskip}{2em}
\setlength{\rightskip}{2em}

Q: Three friends play a game where they flip a fair coin. Each friend flips the coin once. What is the probability that exactly one of them flips heads?

\vspace{0.5em}

A: Let's think step by step.

\vspace{0.2em}

\uline{Each coin flip is an independent event with a $50\%$ chance of landing heads (H) and $50\%$ tails (T). Consider the possible outcomes for three flips: HHH, HHT, HTH, HTT, THH, THT, TTH, TTT. Out of these, only HHH, HHT, HTH, and THH are relevant. We are only interested in scenarios with exactly one head: HTT, THT, and TTH. There are $3$ such outcomes out of $8$ total, giving a probability of $3/8$.}

\endgroup

\end{tcolorbox}
\vspace{0.5em}

\vspace{0.5em}
Prediction 2 (Correct):
\begin{tcolorbox}[frame empty]

\begingroup
\setlength{\leftskip}{2em}
\setlength{\rightskip}{2em}

Q: Three friends play a game where they flip a fair coin. Each friend flips the coin once. What is the probability that exactly one of them flips heads?

\vspace{0.5em}

A: Let's think step by step.

\vspace{0.2em}

\uline{Using a binomial probability formula, where $p$ (success or heads) is $0.5$, and each of the $3$ trials (flips) is independent, we look for exactly one success (head). The formula for exactly $k$ successes in $n$ trials is given by the binomial coefficient formula $\binom{n}{k} p^k (1-p)^{n-k}$. Plugging in the values, we get $\binom{3}{1} (0.5)^1 (0.5)^{3-1} = 3 \cdot 0.5 \cdot 0.25 = 0.375$.}

\endgroup

\end{tcolorbox}
\vspace{0.5em}

\vspace{0.5em}
Prediction 3 (Wrong):
\begin{tcolorbox}[frame empty]

\begingroup
\setlength{\leftskip}{2em}
\setlength{\rightskip}{2em}

Q: Three friends play a game where they flip a fair coin. Each friend flips the coin once. What is the probability that exactly one of them flips heads?

\vspace{0.5em}

A: Let's think step by step.

\vspace{0.2em}

\uline{Since each coin flip has a $50\%$ chance of landing heads and a $50\%$ chance of landing tails, and we are looking for the case where only one of the three flips results in heads, we might initially think this is simply a matter of picking which flip is the heads. There are three flips, so one might incorrectly assume that the chance of picking one specific outcome like this would be $1$ out of $3$. Thus, they might conclude that the probability of exactly one head is $1/3=33.3\%$.}

\endgroup

\end{tcolorbox}
\vspace{0.5em}

Predictions 1 and 2 correctly identify the three cases where exactly one head is flipped, both obtaining a probability of 37.5\%. The reasoning in Prediction 3 fails to account for the total number of outcomes possible with three coin flips, thus giving a wrong answer of 33.3\%. Therefore, we select 37.5\% as the final answer because it is the consensus.

Self-consistency provides a criterion for determining the best prediction in a pool of candidates. Since the prompt and the model are fixed in this method, it is not strictly a prompt ensembling method. Instead, it can be seen as an instance of output ensembling methods, also known as hypothesis selection methods, which have long been explored in NLP, particularly for text generation problems~\cite{xiao-etal:2013bagging}. In these methods, multiple outputs are generated by varying model architectures or parameters. Each output is then assigned a score by some criterion, and the outputs are re-ranked based on these scores. There are various ways to define the scoring function, such as measuring the agreement between an output and others, and using a stronger model to rescore each output\footnote{An interpretation of self-consistency is to view it as a minimum Bayes risk search process. It searches for the best output by minimizing the Bayes risk. More specifically, a risk function $R(\mathbf{y},\mathbf{y}_r)$ is defined on each pair of outputs (denoted by $(\mathbf{y},\mathbf{y}_r)$), representing the cost of replacing $\mathbf{y}$ with $\mathbf{y}_r$. Given a set of outputs $\varOmega$, the risk of an output $\mathbf{y} \in \varOmega$ is given by
\begin{eqnarray}
\mathrm{Risk}(\mathbf{y}) & = & \mathbb{E}_{\mathbf{y}_r \sim \Pr(\mathbf{y}_r|\mathbf{x})} R(\mathbf{y},\mathbf{y}_r) \nonumber \\
                          & = & \sum_{\mathbf{y}_r \in \varOmega} R(\mathbf{y},\mathbf{y}_r) \cdot \Pr(\mathbf{y}_r|\mathbf{x}).
\end{eqnarray}

}. Figure~\ref{fig:llm-ensembling-comparison} shows a comparison of different ensembling methods for LLMs.

\begin{figure}[!t]
\centering
\input{./Chapter9/Figures/figure-ensembling-methods}
\caption{Ensembling methods for LLMs. In standard model ensembling (a), multiple LLMs varying in architectures or parameters are used. Each LLM receives the same prompt and produces a prediction. These predictions are combined to generate the final prediction. In prompt ensembling (b), we have one LLM and multiple prompts. The LLM produces a prediction for each prompt, and these predictions are combined as usual. In output ensembling (c), the LLM samples multiple predictions over the prediction space given a prompt. It can be seen as a method to boost the performance of the LLM itself. Note that these ensembling methods can be combined to increase the diversity of predictions. For example, we can use both prompt ensembling and output ensembling to obtain more diverse predictions.}
\label{fig:llm-ensembling-comparison}
\end{figure}

Now, let us briefly review the methods we have discussed so far in this section, such as problem decomposition and self-refinement. It is apparent that these methods enhance decision-making by introducing more ``choices'' into the reasoning process. To some extent, they all involve evaluating and providing feedback on the results of LLMs. For example, in self-refinement, we need to offer suggestions for improving the prediction of LLMs, and in output ensembling, we select the optimal output from a pool of candidates. In this sense, these methods fall under the broader category of predict-then-verify approaches, where predictions are initially made, then verified and refined. The fundamental problem here involves verifying and evaluating the reasoning results or intermediate steps. This issue is somewhat related to the problem of training reward models in RLHF, although RLHF addresses a different aspect. In fact, the development of verifiers has been explored and implemented in reasoning with LLMs. Most work, rather than developing heuristic-based inference-time algorithms, focuses on learning verifiers in a supervised manner. A straightforward method is to train verifiers as binary classifiers, such as classifying an answer as correct or incorrect, although these verifiers are typically used as scoring models. Given a reasoning path for a problem, the verifiers can be used to score either the entire path (called outcome-based approaches)~\cite{cobbe-etal:2021training}, or each individual reasoning step (called process-based approaches)~\cite{uesato-etal:2022solving,lightman-etal:2024lets}.

\subsection{RAG and Tool Use}
\label{sec:rag-and-tool-use}

\noindent RAG is generally employed when standard LLMs lack sufficient factual accuracy or access to up-to-date knowledge. By drawing from external databases and documents, RAG can significantly improve contextual relevance and factual grounding of responses. This approach is particularly useful in scenarios that require high factual accuracy and up-to-date information, such as complex question answering.

While RAG has been introduced in previous chapters, we briefly outline its core steps here for completeness.

\begin{itemize}
\item \vspace{0.5em} We prepare a collection of texts that serves as an external knowledge source.
\item \vspace{0.3em} We retrieve relevant texts for a given query.
\item \vspace{0.3em} We input both the retrieved texts and the query into an LLM, which is then prompted to produce the final prediction.
\end{itemize}
\vspace{0.5em}

Steps 1 and 2 can be implemented by using an external information retrieval system. For example, we can store the collection of texts in a vector database and then retrieve relevant texts using vector-based search techniques. Since information retrieval is not the focus of this chapter, we will assume that such systems are available off-the-shelf and use them directly.

Here we present how to prompt LLMs to make use of retrieved texts. To illustrate, consider an example of using LLMs to answer the following question.

\begin{center}
\textit{Where will the 2028 Olympics be held?}
\end{center}

We can simply input this question into an online search engine. It will then return the relevant pieces of text found on the internet, for example,

\begin{flushleft}
\begingroup
\setlength{\leftskip}{2em}
\setlength{\rightskip}{2em}

(Wikipedia)

\textit{The 2028 Summer Olympics, officially the Games of the XXXIV Olympiad and commonly known as Los Angeles 2028 or LA28, is an upcoming international multi-sport event scheduled to take place from July 14-30, 2028, in the United States.  ...}

\vspace{0.5em}
(The Sporting News)

\textit{In 2028, Los Angeles will become the third city, following London and Paris respectively, to host three Olympics after hosting the Summer Games in 1932 and 1984. It will also be the first time the United States has hosted an Olympic Games since the 2002 Winter Games in Salt Lake City. ...}

\vspace{0.5em}

...

\endgroup
\end{flushleft}

We can use these retrieved texts as additional context, and prompt an LLM to generate a response based on these texts. Below is an example RAG prompt.

\vspace{0.5em}
\begin{tcolorbox}[frame empty]

\begingroup
\setlength{\leftskip}{2em}
\setlength{\rightskip}{2em}

Your task is to answer the following question. To help you with this, relevant texts are provided. Please base your answer on these texts.

\vspace{0.5em}

Question:

\vspace{0.2em}

Where will the 2028 Olympics be held?

\vspace{0.5em}

Relevant Text 1:

\vspace{0.2em}

The 2028 Summer Olympics, officially the Games of the XXXIV Olympiad and commonly known as Los Angeles 2028 or LA28 ...

\vspace{0.5em}

Relevant Text 2:

In 2028, Los Angeles will become the third city, following London and Paris respectively, to host three Olympics after ...

\vspace{0.2em}

...

\vspace{0.5em}

\underline{The 2028 Olympics will be held in Los Angeles.}

\endgroup

\end{tcolorbox}
\vspace{0.5em}

This prompt assumes that the provided texts are relevant to the question and expects the LLM to generate a faithful response using these texts. However, the information retrieval system may sometimes provide irrelevant or incorrect texts, which may lead the LLM to produce an incorrect answer. One straightforward way to address this issue is to improve the accuracy of the information retrieval system. Nevertheless, as with most AI systems, errors may still occur. Therefore, it is also necessary to enhance the robustness of the LLM, so that it can make reasonable predictions even when the input is inaccurate. Below is a new prompt that enables the LLM to be more faithful to the facts, and allows it to choose not to answer questions when the information provided is inaccurate.

\vspace{0.5em}
\begin{tcolorbox}[frame empty]

\begingroup
\setlength{\leftskip}{2em}
\setlength{\rightskip}{2em}

Your task is to answer the following question. To help you with this, relevant texts are provided. Please base your answer on these texts.

\vspace{0.2em}

\ctext[RGB]{154, 245, 175}{Please note that your answers need to be as accurate as possible and faithful to the facts. If the information provided is insufficient for an accurate response, you may simply output "No answer!".}

\vspace{0.5em}

Question:

\vspace{0.2em}

Where will the 2028 Olympics be held?

\vspace{0.5em}

Relevant Text 1:

\vspace{0.2em}

The 2024 Summer Olympics, officially the Games of the XXXIII Olympiad and branded as Paris 2024, were an international multi-sport event ...

\vspace{0.5em}

...

\vspace{0.5em}

\underline{\ctext[RGB]{154, 245, 175}{No answer!}}

\endgroup

\end{tcolorbox}
\vspace{0.5em}

\noindent In this example, the LLM refuses to answer because the provided information is insufficient and irrelevant to the question.

Both RAG and fine-tuning are common methods for adapting LLMs using task-specific data. Standard RAG is training-free and can be directly applied to LLMs. To further improve RAG, it is also possible to fine-tune LLMs, though this will require some training effort. For example, we can fine-tune LLMs using human-labelled data to supervise them in learning to refuse to answer. Note that, while the examples shown above seem simple, RAG is not trivial. From the prompt engineering perspective, different use cases may require different prompts, though our broader goal is to develop prompting strategies that generalize across tasks. In many cases, we need to control how much we depend on the retrieved context to make predictions. Sometimes, LLMs must generate responses strictly based on the provided texts, while at other times, they may need to generate responses using their existing knowledge if the provided texts are insufficient. There are many aspects of RAG, such as improvements to the retrieval systems, that cannot be covered in this chapter. Interested readers can refer to surveys of RAG techniques for more information~\cite{li-etal:2022survey,gao-etal:2023retrieval}.

One reason we discuss RAG here is that it can be broadly regarded as an instance of the general problem decomposition framework (see Section~\ref{sec:problem-decomposition}). RAG divides problem-solving into two steps. In the first step, we collect relevant and supporting information for a given query from various knowledge sources. In the second step, we use LLMs to generate responses based on the collected information. If we extend the concept of problem decomposition further, we will find that many tasks requiring the use of external systems or tools can be treated as similar problems. One such example is tool use in LLMs. In many applications, LLMs need to employ external databases, APIs, and even simulation tools to generate accurate responses. For example, LLMs can access real-time data from financial markets to provide financial information or access medical knowledge bases to offer personalized medical information. This integration extends the capabilities of LLMs by allowing them to interact with, and in some contexts, influence or control external systems. Consequently, LLMs can function as simple autonomous agents rather than merely text generators~\cite{franklin-and-graesser:1996agent}.

Tool use is a broad research area. Here we narrow our discussion to tasks that can be facilitated by calling external APIs to solve some of the sub-problems~\cite{parisi-etal:2022talm,gao-etal:2023pal}. Consider again the example of asking an LLM to answer ``Where will the 2028 Olympics be held?''. Suppose the LLM can access a web search tool. We can then prompt the LLM to answer the question using web search, like this:

\vspace{0.5em}
\begin{tcolorbox}[frame empty]

\begingroup
\setlength{\leftskip}{2em}
\setlength{\rightskip}{2em}

Your task is to answer the following question. You may use external tools, such as web search, to assist you.

\vspace{0.5em}

Question:

\vspace{0.2em}

Where will the 2028 Olympics be held?

\vspace{0.5em}

\underline{The information regarding this question is given as follows:}

\vspace{0.2em}

\underline{\ctext[RGB]{154, 245, 175}{\texttt{\{tool: web-search, query: "2028 Olympics"\}}}}

\vspace{0.2em}

\underline{So the answer is: Los Angeles}

\endgroup

\end{tcolorbox}
\vspace{0.5em}

\noindent Here \texttt{{tool: web-search, query: "2028 Olympics"}} indicates a request to the web search system using the query ``2028 Olympics''. The generated tool command is intercepted by an external system, which executes the corresponding tool and appends the returned result to the context. The updated context is then fed back into the LLM for subsequent prediction, and thus the model can generate the correct answer using the retrieved information.

Consider another example where we ask the LLM to solve a mathematical problem.

\vspace{0.5em}
\begin{tcolorbox}[frame empty]

\begingroup
\setlength{\leftskip}{2em}
\setlength{\rightskip}{2em}

Problem:

\vspace{0.2em}

A swimming pool needs to be filled with water. The pool measures 10 meters in length, 4 meters in width, and 2 meters in depth. Calculate the volume of the pool in cubic meters and then determine how many liters of water are needed to fill it (considering 1 cubic meter equals 1000 liters).
\vspace{0.5em}

Solution:

\underline{To solve this problem, the LLM needs to first calculate the volume of the}

\underline{pool by using the formula for the volume of a rectangular prism: $\text{Length} \times$}

\underline{$\text{Width} \times \text{Depth}.$ Therefore, the volume is $10 \, \text{m} \times 4 \, \text{m} \times 2 \, \text{m} =$ \ctext[RGB]{154, 245, 175}{\texttt{\{tool:}}}

\ctext[RGB]{154, 245, 175}{\texttt{calculator, expression: 10 * 4 * 2\}}} $\text{m}^3$. \underline{Next, to find out how}

\underline{many  liters of water are needed, the LLM multiplies the volume in cubic}

\underline{meters by 1000 (since 1 cubic meter equals 1000 liters). Thus, \ctext[RGB]{154, 245, 175}{$80$} $ \times 1000$}

\underline{$= $ \ctext[RGB]{153, 202, 240}{\texttt{\{tool: calculator, expression: 80*1000\}}} $\, \text{liters}.$}

\endgroup

\end{tcolorbox}
\vspace{0.5em}

\noindent Here the string \texttt{\{tool: calculator, expression: 10 * 4 * 2\}} triggers the invocation of a mathematical interpreter to calculate the result of the expression. Note that the result (i.e., 80) will replace \texttt{\{tool: calculator, expression: 10 * 4 * 2\}} and can be referred to in the following token predictions. For example, in the last step of problem-solving, 80 is used instead of \texttt{\{tool: calculator, expression: 10 * 4 * 2\}}.

A key difference between the tool use examples here and the previously discussed RAG examples is that in tool use, external functions can be called during inference. In contrast, in RAG, the retrieved texts are provided before the prediction process begins. However, from the language modeling perspective, they are similar: before generating the final result, we use external tools, either manually or automatically, to obtain sufficient and relevant context. A high-level interpretation of these approaches is that they both rely on an ``agent'' that can determine where and how to call external functions to generate the context necessary for prediction.

An issue with tool use is that the original LLM is not trained to generate the necessary markers for tool use. Therefore, we need to fine-tune the LLM to adapt it for these tasks~\cite{schick-etal:2024toolformer}. As this chapter focuses on prompting, we will not present the details of this fine-tuning process. To put it simply, we first need to annotate data. For each fine-tuning example, we replace parts of the output that require the use of external tools with predefined commands or markers. Then, we use this labeled data to fine-tune the parameters of the LLM as usual. As a result, the LLM can gain the ability to generate commands for calling external tools. During inference, we can execute these tool use commands in the model outputs to get assistance from external tools.

%----------------------------------------------------------------------------------------
% NEW SECTION
%----------------------------------------------------------------------------------------

\section{Learning to Prompt}

\noindent So far in this chapter, we have considered several basic prompting strategies and various refinements to them. However, all the prompts we have discussed were designed manually. This leads to a number of problems. First, designing high-quality prompts is difficult and often requires extensive experimentation to identify effective prompts. Since different LLMs may vary in their sensitivity to prompting styles, developing prompts that generalize well across models can be resource-intensive. Second, manual prompt design relies heavily on human expertise, which can limit the diversity of approaches and overlook potentially effective prompts that are not immediately obvious to humans. Third, prompts created by humans can be complex and verbose, leading to longer inputs for LLMs and higher computational costs.

In this section, we discuss techniques for automated prompting. These methods aim to automatically create, optimize, and represent prompts so that the downstream tasks can be addressed more effectively and efficiently. In particular, we consider three issues here.

\begin{itemize}
\item \vspace{0.5em} How can we automate the process of designing and optimizing prompts for LLMs?
\item \vspace{0.3em} Are there representations of prompts beyond plain text strings, and how can we learn such representations?
\item \vspace{0.3em} How can we make prompts more concise and compact, thereby reducing their complexity and length?
\end{itemize}
\vspace{0.5em}

Note that there are many settings in which we can investigate these issues. For example, prompts may be optimized for a specific LLM or designed to generalize across different LLMs. These settings can lead to different methods and application scenarios, but these methods may overlap in some ways. In the following discussion, we will cover several different scenarios and discuss the connections between various methods.

\subsection{Prompt Optimization}

\noindent Because prompt design is labor-intensive, it is desirable to use machine learning models to discover the optimal prompt for a specific task (call it \mindex{automatic prompt design} or \mindex{prompt optimization}). This approach can broadly be regarded as an instance of \mindex{automated machine learning} (\mindex{AutoML}), which aims to reduce or eliminate the need for expert-driven manual design of machine learning models. Although our focus here is on the design of prompts, prompts themselves are discrete structures. Therefore, prompt optimization is similar to the search over discrete model architectures in machine learning. One closely related area is \mindex{neural architecture search} (\mindex{NAS}), where the optimal neural networks are identified by exploring a space of possible neural networks~\cite{zoph-and-Le:2016neural,elsken-etal:2019nas}. If we consider prompt optimization as a search process, then we can describe a general prompt optimization framework involving the following components:

\begin{itemize}
\item \vspace{0.5em} \mindex{Prompt Search Space}. This defines all possible prompts that the algorithms can explore. For example, one can edit some seed prompts to generate a set of diverse candidate prompts.
\item \vspace{0.3em} \mindex{Performance Estimation}. Once a prompt is chosen, it needs to be evaluated. For example, a straightforward way is to input it to an LLM and measure its performance on a validation set.
\item \vspace{0.3em} \textbf{Search Strategy}. The search process is generally the same as that used in many AI systems. At each step, the system explores a set of promising prompts in the search space and evaluates them. This process continues as more prompts are explored. The outcome of the search is the best-performing prompt observed until the search stops.
\end{itemize}
\vspace{0.5em}

This is a very general framework, and different prompt optimization systems can vary in their design of each component. A widely used approach is to use LLMs as the basis to develop these components. Initially, a few prompts are provided. Then, the following process is iterated until a stopping criterion is met: 1) the prompts are evaluated on a validation set; 2) a candidate pool is maintained by keeping only the most promising prompts; and 3) new prompts are created by employing LLMs to infer similar prompts from this candidate pool. One benefit of this approach is that it allows us to use off-the-shelf LLMs to perform the tasks mentioned above without the need for substantial system development. To achieve this, we can prompt or fine-tune LLMs to adapt them to these tasks. Here we consider \citet{zhou-etal:2023large}'s method for illustrating LLM-based prompt optimization. It involves the following steps.

\begin{itemize}
\item \vspace{0.5em} \textbf{Initialization}. Let $C$ represent the pool of candidate prompts to be explored. The first step is to add initial prompts into $C$.  We can do this in several ways. A simple method is to create such prompts by hand for a given task. However, in many cases where humans have limited knowledge about how to write effective prompts for the task, developing prompts becomes challenging. In these cases, it is desirable to use LLMs to generate prompts. For example, we can directly instruct LLMs to produce prompts, providing them with a description of the task.

\vspace{0.5em}
\begin{tcolorbox}[frame empty]

\begingroup
\setlength{\leftskip}{2em}
\setlength{\rightskip}{2em}

You are given a task to complete using LLMs. Please write a prompt to guide the LLMs.

\vspace{0.5em}

$\{*\mathrm{task\textrm{-}description}*\}$

\vspace{0.5em}

\underline{\hspace{2em}}

\endgroup

\end{tcolorbox}
\vspace{0.5em}

    This method is straightforward, but it still requires a human-provided description of the task. An alternative method is to use LLMs to generate prompts given examples of the input and output of the task. Here is a prompt template.

\vspace{0.5em}
\begin{tcolorbox}[frame empty]

\begingroup
\setlength{\leftskip}{2em}
\setlength{\rightskip}{2em}

You are provided with several input-output pairs for a task. Please write an instruction for performing this task.

\vspace{0.5em}

Input: $\{*\mathrm{input}1*\}$\ \ \ Output: $\{*\mathrm{output}1*\}$

\vspace{0.2em}

Input: $\{*\mathrm{input}2*\}$\ \ \ Output: $\{*\mathrm{output}2*\}$

...

\vspace{0.5em}

\underline{\hspace{2em}}

\endgroup

\end{tcolorbox}
\vspace{0.5em}

In this way, LLMs can infer the corresponding instruction for the task from the provided inputs and outputs.

\item \vspace{0.3em} \textbf{Evaluation}. Once we obtain the candidate pool $C$, we need to evaluate the prompts in $C$. One method is to feed each prompt into an LLM and assess the results on the downstream task. For example, we can evaluate the output of the LLM given an input using a pre-defined metric, or alternatively, use the log-likelihood of the output as a measure of the quality of the prompt.

\item \vspace{0.3em} \textbf{Pruning}. If $C$ contains a large number of prompts, it is reasonable to prune the unpromising prompts within it, thus reducing the computational burden in subsequent steps. This is a standard pruning problem. Given the evaluation score for each prompt, a simple method is to keep only a certain percentage of the prompts and discard the rest.

\item \vspace{0.3em} \textbf{Expansion}. Expansion is a key operation in search algorithms used to explore different states in the search space. The expansion operation here can be defined as a function:
    \begin{eqnarray}
    C' & = & \mathrm{Expand}(C, f),
    \end{eqnarray}

    \noindent where $C'$ is the set of new prompts generated from $C$ using the model $f$. If we consider $f$ as an LLM, we can perform the expansion operation by instructing $f$ to generate new and relevant prompts based on $C$. Below is an example.

\vspace{0.5em}
\begin{tcolorbox}[frame empty]

\begingroup
\setlength{\leftskip}{2em}
\setlength{\rightskip}{2em}

Below is a prompt for an LLM. Please provide some new prompts to perform the same task.

\vspace{0.5em}

Input: $\{*\mathrm{prompt}*\}$

\vspace{0.5em}

\underline{\hspace{2em}}

\endgroup

\end{tcolorbox}
\vspace{0.5em}

Then, we replace $C$ with $C'$. The steps of evaluation, pruning and expansion can be repeated, and so we can gradually explore a wider range of prompts.

\end{itemize}
\vspace{0.5em}

In prompt optimization, the expansion step plays a key role, as it defines how we explore the search space, and our goal is to find optimal results with minimal effort. One improvement to this step is to treat the problem as a paraphrasing task. A simple method is to apply off-the-shelf paraphrasing systems, either based on LLMs or other models, to transform input prompts into semantically equivalent forms~\cite{jiang-etal:2020can}. Alternatively, we can define specific edit operations, such as insertions and modifications, for each token. A given prompt can be edited into new prompts by applying these operations~\cite{prasad-etal:2023grips}. Also, further evaluation and pruning can be applied to filter out low-quality prompts. In addition to framing prompt generation as a paraphrasing problem, we can improve the quality of prompts during expansion by learning from feedback~\cite{pryzant-etal:2023automatic}. This approach is somewhat related to the self-refinement issue discussed in Section~\ref{sec:self-refinement}. An LLM can be used to generate feedback on an input prompt, which is then revised based on this feedback. This feedback-and-revision cycle can be repeated multiple times until the result converges or the desired outcome is achieved.

Another approach to prompt optimization is to apply classic optimization techniques. For example, the problem can be framed as an evolutionary computation problem, where prompts are treated as candidates that evolve generation by generation as the optimization progresses~\cite{guo-etal:2024connecting}. Since many powerful optimization algorithms have been developed in related fields, they can be directly applied to this problem.

In practice, we might be tempted to use existing LLM APIs to implement the steps described above. Such an approach, however, depends heavily on the inference and in-context learning abilities of the LLMs. If these LLMs are not strong and lack adaptation to the tasks, they may introduce errors into search, for example, generating incorrect prompts during expansion. In such cases, it is preferable to train models that are better suited to the tasks. One approach in this research direction employs reinforcement learning, which has been widely used in solving discrete decision making and optimization problems. For example, \citet{deng-etal:2022rlprompt} developed a prompt generator by integrating an FFN-based adaptor into an LLM. The prompt generator is trained as a typical policy network, but only the parameters of the adaptor are updated while the remaining parameters of the model are kept unchanged. During training, the reward is obtained by testing the generated prompts using another LLM, similar to the evaluation method as discussed above. Once the training is complete, the prompt generator is then employed to generate new prompts.

Note that, in our discussion here, prompts are simply seen as sequences of tokens, and the output of prompt optimization is such a sequence. However, in a strict sense, prompts often contain multiple components, such as instructions, demonstrations, and user inputs. While our discussed approaches are mostly general, much work in prompt optimization has focused on learning better instructions for prompting. Specifically, the goal is to generate instructions that effectively guide LLMs based on a given task. Of course, the concept of prompt optimization can also be extended to learning other parts of prompts. For example, there has been substantial research interest in learning to select or generate demonstrations in CoT~\cite{liu-etal:2022makes,rubin-etal:2022learning,zhang-etal:2023automatic}. One of the differences between learning instructions and learning demonstrations is that generating high-quality demonstrations using LLMs is relatively easy and the focus of learning demonstrations is typically on how to sample appropriate demonstrations from a pool of candidates. In contrast, the difficulty in learning instructions is partly because pre-trained LLMs are not suited to predict the quality of instructions, and testing these instructions on downstream tasks is computationally expensive. This makes the optimization methods costly to apply, and exploring a wide variety of instructions poses significant challenges.

\subsection{Soft Prompts}

\noindent Although developing natural language prompts, either manually or automatically, is a straightforward and widely applied approach, it presents some problems. One problem is that natural language prompts can be complex and lengthy, resulting in significant computational burdens when processed by LLMs. In many applications, users may need to perform a task repeatedly, and inputting the same long prompt into the LLMs a large number of times is clearly inefficient. Another problem is that while prompts are typically represented as discrete token sequences (call them \mindex{hard prompts}) in regular LLM input, the LLMs encode them as continuous vectors. This raises the question of whether there are more compact and efficient ways to represent prompts.

In this subsection, we introduce the concept of \mindex{soft prompts}, which can be viewed as hidden, distributed representations of prompts. When prompting LLMs, we are concerned with communicating tasks or questions to elicit the desired responses. We can define hard prompts as explicit, predefined text sequences that users input directly into LLMs to guide the responses. In contrast, we can think of soft prompts as implicit, adaptable prompting patterns represented in the continuous hidden space of LLMs. Unlike hard prompts, which are expressed in natural language and should be understandable for humans, soft prompts are encoded in a format that is more comprehensible to the model rather than to humans. To illustrate, consider a simple prompt

\vspace{0.5em}
\begin{tcolorbox}[frame empty]

\begingroup
\setlength{\leftskip}{2em}
\setlength{\rightskip}{2em}

Translate the sentence into Chinese.

\vspace{0.5em}

Consider it done!

\vspace{0.5em}

\underline{\hspace{2em}}

\endgroup

\end{tcolorbox}
\vspace{0.5em}

\noindent Here, the instruction ``Translate the sentence into Chinese'' can be seen as a hard prompt, denoted by the token sequence $c_1 \cdots c_5$. By feeding these tokens into an LLM, they are transformed into a sequence of real-valued vectors $\mathbf{h}_1 \cdots \mathbf{h}_5$, each corresponding to a token. We can view $\mathbf{h}_1 \cdots \mathbf{h}_5$ as an intuitive illustration of soft prompts, as shown in Figure~\ref{fig:hard-and-soft-prompt-examples}.

\begin{figure}[!t]
\centering
\input{./Chapter9/Figures/figure-hard-prompt-and-soft-prompt}
\caption{Illustration of hard and soft prompts. Here  the hard prompt is the instruction we input to the LLM for performing the task. The LLM encodes this instruction as usual, and the intermediate representations corresponding to the instruction can be viewed as some sort of soft prompt.}
\label{fig:hard-and-soft-prompt-examples}
\end{figure}

While the above example shows that soft prompts can be generated by transforming hard prompts, there is not necessarily a direct correspondence between them. In fact, we do not even need to interpret soft prompts using meaningful text. Instead, soft prompts are represented as continuous vectors in the hidden space of LLMs and can be optimized directly as learnable parameters. Such a treatment allows us to explore prompting methods beyond text. As another benefit, soft prompts provide dense, learnable representations for encoding how we guide LLMs to generate specific outputs. The training and application of these representations require significantly lower computational costs than those required for processing long hard prompts. This approach would be of great practical value in LLM inference applications where the same prompt is repeatedly used.

\subsubsection{Adapting LLMs to Simplified Prompts}

\noindent One obvious way to adapt an LLM for a particular task is to simply fine-tune the model using labeled data. This leads to a variety of LLM alignment methods, such as supervised fine-tuning, which update the model parameters by aligning the responses to given prompts with supervision signals. Fine-tuned LLMs embed task-related information in model parameters, and thus these models can respond correctly when dealing with similar prompts to those in fine-tuning.

If we take this idea further, we can expect LLMs to absorb the knowledge about prompting of a task as much as possible during fine-tuning. Consequently, the prompting information is partially captured in the model parameters, and the fine-tuned LLMs can perform the task with less prompting. Here we consider a simple form of prompt, where only an instruction (denoted by $\mathbf{c}$) and a user input (denoted by $\mathbf{z}$) are included. A prompt can be expressed using the following tuple:
\begin{eqnarray}
\mathbf{x} & = & (\mathbf{c}, \mathbf{z}).
\end{eqnarray}

Given a set of prompt-response pairs $\mathcal{D}=\{(\mathbf{x},\mathbf{y})\}$, the objective of fine-tuning is to minimize the total loss incurred over this set. A popular method is to minimize the negative log-likelihood (i.e., maximize the log-likelihood) with respect to the model parameters $\theta$:
\begin{eqnarray}
\hat{\theta} & = & \argmax_{\theta} \sum_{(\mathbf{x},\mathbf{y}) \in \mathcal{D}} \log \mathrm{Pr}_{\theta}(\mathbf{y}|\mathbf{x}) \nonumber \\
             & = & \argmax_{\theta} \sum_{(\mathbf{x},\mathbf{y}) \in \mathcal{D}} \log \mathrm{Pr}_{\theta}(\mathbf{y}|\mathbf{c}, \mathbf{z}),
\end{eqnarray}

\noindent where $\mathrm{Pr}_{\theta}(\cdot|\cdot)$ is the probability predicted by an LLM with the parameters $\theta$\footnote{In practice, we initialize $\theta$ with the parameters obtained from pre-training, and then adjust $\theta$ moderately to ensure that the results after fine-tuning do not deviate too much from the pre-trained results.}.

In general, the instruction in each fine-tuning example should follow the guideline of prompt design, for example, a good instruction should be as clear as possible and provide a detailed description of the task. However, the method described in the above equation does not restrict the instruction to any particular form. This flexibility allows us to instruct LLMs in any way that we want. Consider an example where we intend to instruct LLMs to translate an English sentence into Chinese. Of course, as mentioned earlier in this chapter, we can prompt LLMs using the instruction

\vspace{1em}
\hspace{4em} \textit{Translate the following sentence from English to Chinese.}
\vspace{1em}

\noindent If we want the instruction to be simpler, we may rephrase it into a simpler form

\vspace{1em}
\hspace{4em} \textit{Translate this into Chinese.}
\vspace{1em}

\noindent We can even define the instruction as a single phrase

\vspace{1em}
\hspace{4em} \textit{Translate!}
\vspace{1em}

With fine-tuning, we can adapt LLMs to follow any of these instructions. From an efficient prompting perspective, there are computational advantages in simplifying instructions in prompting. For example, we can use simple instructions like ``Translate!'' to perform tasks that would typically require more complex and detailed instructions. This can make subsequent prompting during inference much easier. On the other hand, fine-tuning LLMs with overly simplified instructions may be harmful to the generalization of the models. Since simplified instructions can lead to a loss of information, it is more likely that the LLMs will overfit the fine-tuning data and fail to generalize beyond those instructions. This issue becomes more pronounced when both complex and simplified instructions are included during fine-tuning, since the available labeled data is often insufficient to cover a wide range of instruction variations.

An alternative way to adapt LLMs for simplified instructions is through knowledge distillation. As an example, we consider the context distillation method~\cite{snell-etal:2022learning}. The goal of this method is to learn a student model that can make use of simplified instructions from a well-trained instruction-following teacher model. Figure~\ref{fig:context-distillation} shows an illustration of this approach. The teacher model is built using a standard fine-tuning process: we first collect a certain amount of data that includes instructions, user inputs, and correct responses, and then we continue to train a pre-trained model with this dataset. For building the student model, we need to construct a new dataset $\mathcal{D}'$ where each sample is a tuple consisting of an instruction, a corresponding simplified instruction, and a user input, denoted by $\mathbf{x}' = (\mathbf{c}, \mathbf{c}', \mathbf{z})$. Knowledge distillation is performed by minimizing a loss function defined on the outputs of the teacher and student models:
\begin{eqnarray}
\hat{\theta} & = & \argmin_{\theta} \sum_{\mathbf{x}' \in \mathcal{D}'} \mathrm{Loss}(\mathrm{Pr}^{t}(\cdot|\cdot),\ \mathrm{Pr}_{\theta}^{s}(\cdot|\cdot),\ \mathbf{x}'),
\end{eqnarray}

\begin{figure}[!t]
\centering
\input{./Chapter9/Figures/figure-contex-distillation}
\caption{Illustration of context distillation~\cite{snell-etal:2022learning}. The teacher model is a standard LLM, which takes both the context and the user input as model input and produces a prediction as model output. Then, we simplify the context (e.g., simplifying the instruction in prompting) and use the student model to make predictions based on the simplified context and the user input. The student model is trained by minimizing the loss between the predictions produced by the two models.}
\label{fig:context-distillation}
\end{figure}

\noindent where $\mathrm{Pr}^{t}(\cdot|\cdot)$ denotes the pre-trained teacher model, and $\mathrm{Pr}_{\theta}^{s}(\cdot|\cdot)$ denotes the student model with the parameters $\theta$. To keep the notation simple we will write $\mathrm{Loss}(\mathrm{Pr}^{t}(\cdot|\cdot),\ \mathrm{Pr}_{\theta}^{s}(\cdot|\cdot),\ \mathbf{x}')$ as $\mathrm{Loss}$ for short. A commonly used loss is the sequence-level loss, which has the basic form:
\begin{eqnarray}
\mathrm{Loss} & = & -\sum_{\mathbf{y}} \mathrm{Pr}^{t}(\mathbf{y}|\mathbf{c},\mathbf{z}) \log \mathrm{Pr}_{\theta}^{s}(\mathbf{y}|\mathbf{c}',\mathbf{z}).
\end{eqnarray}

But this function is computationally infeasible because it requires summing over an exponentially large number of possible output sequences. A variant of this method is to train the student model using outputs generated by the teacher model. For each sample, we use the teacher model to produce an output $\hat{\mathbf{y}} = \argmax_{\mathbf{y}} \log \mathrm{Pr}^{t}(\mathbf{y}|\mathbf{c},\mathbf{z})$. Then we consider $\hat{\mathbf{y}}$ as the target for learning, and the loss function is given by
\begin{eqnarray}
\mathrm{Loss} & = & -\log \mathrm{Pr}_{\theta}^{s}(\hat{\mathbf{y}}|\mathbf{c}', \mathbf{z}). \label{eq:context-distillation-simple}
\end{eqnarray}

Alternatively, we can minimize the distances between the probability distributions generated by the two models~\cite{askell-etal:2021general}. For example, the loss function can be defined as the KL divergence between the two output distributions:
\begin{eqnarray}
\mathrm{Loss} & = & \mathrm{D}_{\mathrm{KL}}(\mathrm{P}^{t} \,\|\, \mathrm{P}_{\theta}^{s}), \label{eq:context-distillation-kl}
\end{eqnarray}

\noindent where
\begin{eqnarray}
\mathrm{P}^t & = & \mathrm{Pr}^{t}(\cdot|\mathbf{c},\mathbf{z}), \\
\mathrm{P}^s_{\theta} & = & \mathrm{Pr}_{\theta}^{s}(\cdot|\mathbf{c}', \mathbf{z}).
\end{eqnarray}

Simplifying prompts through fine-tuning or distillation does not necessarily produce explicit soft prompts in the conventional sense. However, these approaches share a common intuition: task-specific prompting behavior can gradually be absorbed into compact continuous representations within the model.

Although we have restricted ourselves to knowledge distillation for instructions, the approaches discussed here are general. By learning from the outputs of the teacher model, the knowledge in prompting can be distilled into the parameters of the student model. Therefore, the distilled model can be viewed as implicitly encoding task-specific prompting behavior in its parameters. This method can be applied to many other problems in prompt learning, such as compressing long contexts and learning soft prompts as specific components of LLMs.

\subsubsection{Learning Soft Prompts for Parameter-efficient Fine-tuning}

\noindent Full fine-tuning, which updates all model parameters, is a common method for adapting LLMs to downstream tasks. Although fine-tuning is much less computationally expensive than pre-training, it is still costly to apply in practice. This issue motivates the development of parameter-efficient fine-tuning methods, which aim to minimize the number of parameters that need to be updated.

One approach, known as \mindex{prefix fine-tuning}, prepends a sequence of trainable vectors, or prefixes, to the input of each Transformer layer~\cite{li-liang:2021prefix}. These prefixes can be thought of as soft prompts that serve as additional context to guide the behavior of the model under specific tasks. During fine-tuning, we need only to learn the prefixes for embedding task-specific knowledge. Thus, this method is efficient because it only modifies a small part of the model rather than adjusting the entire set of model parameters.

Specifically, let the input of a layer at depth $l$ be denoted by $\mathbf{H}^l = \mathbf{h}_0^l \mathbf{h}_1^l \cdots \mathbf{h}_m^l$. The output of the layer can be expressed as:
\begin{eqnarray}
\mathbf{H}^{l+1} & = & \mathrm{Layer}(\mathbf{H}^{l}).
\end{eqnarray}

In prefix fine-tuning, we extend the sequence $\mathbf{h}_0^l \mathbf{h}_1^l \cdots \mathbf{h}_m^l$ by adding a few vectors at the beginning, which we denote as $\mathbf{p}_0^l \mathbf{p}_1^l \cdots \mathbf{p}_n^l$. Hence $\mathbf{H}^l$ can be written in the form:
\begin{eqnarray}
\mathbf{H}^l = \underbrace{\mathbf{p}_0^l\ \mathbf{p}_1^l\ \cdots\ \mathbf{p}_n^l}_{\text{trainable}} \underbrace{\mathbf{h}_0^l\ \mathbf{h}_1^l\ \cdots\ \mathbf{h}_m^l}_{\text{previous layer output}}. \label{eq:prefix-finetuning-layer-input}
\end{eqnarray}

\noindent We retain only the last $m+1$ representations as the hidden states passed to the next layer:
\begin{eqnarray}
\overline{\mathbf{H}}^{l+1} & = & \mathrm{Layer}(\mathbf{H}^{l})[-m-1:\ ] \nonumber \\
                       & = & \mathbf{h}_0^{l+1}\mathbf{h}_1^{l+1}\cdots\ \mathbf{h}_m^{l+1},
\end{eqnarray}

\noindent where $[-m-1:\ ]$ denotes the slicing operation that extracts the last $m+1$ elements of a sequence. Given $\overline{\mathbf{H}}^{l+1}$, the input of the next layer can be expressed in the same form as Eq.~(\ref{eq:prefix-finetuning-layer-input}):
\begin{eqnarray}
\mathbf{H}^{l+1} & = & \mathbf{p}_0^{l+1}\ \mathbf{p}_1^{l+1}\ \cdots\ \mathbf{p}_n^{l+1}\ \overline{\mathbf{H}}^{l+1} \nonumber \\
                 & = & \mathbf{p}_0^{l+1}\ \mathbf{p}_1^{l+1}\ \cdots\ \mathbf{p}_n^{l+1} \mathbf{h}_0^{l+1}\mathbf{h}_1^{l+1}\cdots \mathbf{h}_m^{l+1}.
\end{eqnarray}

\noindent Here each $\mathbf{p}_i \in \mathbb{R}^d$ can be seen as a learnable parameter. During training, $\mathbf{p}_0^l \mathbf{p}_1^l \cdots\mathbf{p}_n^l$ are trained as usual, and the parameters of the original Transformer model are kept fixed.

Figure~\ref{fig:prefix-finetuning} shows an illustration of prefix fine-tuning for a translation task. Here, only the prefix vectors $\mathbf{p}_0^l$ and $\mathbf{p}_1^l$ are updated via gradients propagated from the output layer (i.e., the Chinese translation). By adjusting these vectors for the translation task, the model adapts accordingly. This makes $\mathbf{p}_0^l$ and $\mathbf{p}_1^l$ serve as prompts which activate the LLM to perform the task without needing explicit input prompts like ``Translate the following sentence from English to Chinese''. At test time, we prepend the optimized $\mathbf{p}_0^l$ and $\mathbf{p}_1^l$ to the layer, and the LLM will then translate the input sentence. Note that prefix fine-tuning introduces additional $L \times n \times d$ parameters, where $L$ is the number of layers, $n$ is the number of prefixes, and $d$ is the dimensionality of each prefix. However, this number is much smaller compared to the total number of parameters in the LLM, making the fine-tuning process highly efficient.

\begin{figure}[!t]
\centering
\input{./Chapter9/Figures/figure-prefix-finetuning}
\caption{Illustration of prefix fine-tuning for a translation task (Look out! $\to$ \begin{CJK*}{UTF8}{gbsn} 小心 \end{CJK*}!). For each layer, we add two prefixes $\mathbf{p}_0^l$ and $\mathbf{p}_1^l$ at the beginning. The LLM is trained to minimize the loss on the predictions given the input. During this process, only the prefixes are optimized while the rest of the parameters remain fixed. Therefore, the model can adapt to the task with only a small number of trainable parameters. At inference time, the LLM works with optimized prefixes, and can perform the task without the need for explicit hard prompts.}
\label{fig:prefix-finetuning}
\end{figure}

While prefix fine-tuning is simple, it still requires modifications to LLMs. Alternatively, separating soft prompts from the LLMs allows us to preserve the original model architecture, making it more efficient for deployment across different tasks without the need to adjust the core model. One such method is prompt tuning~\cite{lester-etal:2021power}. Like prefix fine-tuning, prompt tuning incorporates trainable vectors so that LLMs can adapt to given tasks by adjusting these vectors. However, prompt tuning differs in that it modifies only the embedding layer.

Recall that in LLMs each input token $z_i$ is represented by an embedding $\mathbf{e}_i$. These embeddings are generally learned through a token embedding model and are then used as the real inputs to the LLMs, replacing the symbolically represented tokens. In prompt tuning, a number of pseudo embeddings $\mathbf{p}_0 \cdots \mathbf{p}_n$ are added at the beginning of the token embedding sequence. So the actual input to the LLMs can be expressed as:
\begin{equation}
\underbrace{\mathbf{p}_0\ \mathbf{p}_1\ \cdots\ \mathbf{p}_n}_{\text{trainable}}\ \underbrace{\mathbf{e}_0\ \mathbf{e}_1\ \cdots\ \mathbf{e}_m}_{\text{token embeddings}}.
\end{equation}

\noindent Note that a pseudo embedding does not need to correspond to any token in natural language. Instead these embeddings can be seen as ``soft prompt embeddings'' that serve to condition the LLMs. By training soft prompt embeddings on task-specific data, they learn to interact adaptively with the token embeddings $\mathbf{e}_0 \cdots \mathbf{e}_m$ and guide the behavior of LLMs. Since prompt tuning does not change the underlying parameters of pre-trained LLMs, it is considered a lightweight and efficient method of fine-tuning, improving task-specific performance while maintaining their generalization capabilities. See Figure~\ref{fig:prefix-prompt-tuning} for an illustration of prompt tuning.

\begin{figure}[!t]
\centering
\input{./Chapter9/Figures/figure-prompt-tuning}
\caption{Illustration of prompt tuning for a translation task (Look out! $\to$ \begin{CJK*}{UTF8}{gbsn} 小心 \end{CJK*}!). Instead of using fixed textual prompts, soft prompts are learnable embeddings that are added at the beginning of the embedding sequence. During fine-tuning, only these prompt embeddings are optimized to efficiently adapt the LLM to the given task. Once optimized, the prompt embeddings are used to instruct the LLM to perform the task as new data arrives.}
\label{fig:prefix-prompt-tuning}
\end{figure}

Since $\mathbf{p}_0\ \mathbf{p}_1\ \cdots\ \mathbf{p}_n$ is itself a sequence, we can employ sequence models to better represent it. For example, a Transformer model can encode this sequence, and the resulting representation can then be used as the input to the LLM. In other words, we can develop an additional model for encoding soft prompts. Another way to improve prompting is by combining soft and hard prompts, thereby taking advantage of both types~\cite{liu-etal:2023gpt}. In the embedding sequence, we can arrange or intersperse these prompts. This would result in different prompt patterns. For example, see Figure~\ref{fig:prompt-pattern-soft-and-hard-prompts} for a simple pattern that combines both types of prompts.

\begin{figure}[!t]
\centering
\input{./Chapter9/Figures/figure-prompt-pattern-soft-and-hard-prompts}
\caption{A prompt pattern that combines soft prompts, hard prompts, and user input. Here, $c_0 \cdots c_{m'}$ denotes the hard prompt and $\mathbf{q}_0 \cdots \mathbf{q}_{m'}$ denotes the corresponding embedding sequence.}
\label{fig:prompt-pattern-soft-and-hard-prompts}
\end{figure}

Here we have considered methods for inserting soft prompts in LLMs. But we skip the details of training these soft prompts and assume that the reader is familiar with the standard supervised learning process, that is, maximizing the likelihood of the correct model output given the model input. In fact, learning soft prompts can be related to many issues in LLM fine-tuning. For example, if we consider it as a context compression problem, we can apply the knowledge distillation methods described previously. In \citet{mu-etal:2024learning}'s work, prompts are compressed and represented as a few pseudo tokens, which are appended to each input sequence. The embeddings of these pseudo tokens are optimized to mimic the predictions of a standard-prompted model. In other words, the prompting knowledge is distilled from a teacher model into the pseudo tokens.

Broadly speaking, many parameter-efficient fine-tuning methods can be thought of as learning some sort of soft prompt~\cite{lialin-etal:2023scaling}. When we fine-tune a part of an LLM for a task, this process can essentially be seen as injecting task-related prompting information into that specific part of the model. Another widely-used approach to parameter-efficient fine-tuning is to add an adaptor layer between the existing model layers. This approach allows us to fine-tune only the adaptor layer on specific tasks without altering the underlying architecture or retraining the entire model. In this sense, adaptor layers can be viewed as soft prompts that encode prompting and task-related information and interact with the original LLM to help it adapt. To summarize, Figure~\ref{fig:comparison-of-soft-prompts-in-llms} shows a comparison of different methods of using soft prompts in LLMs.

\begin{figure}[!t]
\centering
\input{./Chapter9/Figures/figure-soft-prompts-in-llms}
\caption{Illustrations of using soft prompts in LLMs. Here tunable soft prompts are shown in blue, and components whose parameters are fixed during fine-tuning are shown in gray. In sub-figure (a), soft prompts are prefixes appended to each layer of the LLM. In sub-figure (b), soft prompts are used as input embeddings for the LLM. In sub-figures (c) and (d), soft prompts are broadly treated as components of the model that are fine-tuned for task adaptation.}
\label{fig:comparison-of-soft-prompts-in-llms}
\end{figure}

\subsubsection{Learning Soft Prompts with Compression}

\noindent Another approach to learning soft prompts is to frame it as a compression problem. As a simple example, consider the problem of approximating a long context using a continuous representation~\cite{wingate-etal:2022prompt}. Suppose we have a user input $\mathbf{z}$ and its context $\mathbf{c}$ (such as long instructions and demonstrations). Now we want to learn a compressed representation of the context, denoted by $\sigma$, such that the prediction conditioned on $\mathbf{z}$ and $\sigma$
is as close as possible to that conditioned on $\mathbf{z}$ and $\mathbf{c}$. This goal can be loosely formulated as
\begin{eqnarray}
\hat{\sigma} & = & \argmin_{\sigma} s(\hat{\mathbf{y}},\hat{\mathbf{y}}_{\sigma}),
\end{eqnarray}

\noindent where $\hat{\mathbf{y}} = \argmax_{\mathbf{y}} \Pr(\mathbf{y}|\mathbf{c},\mathbf{z})$ and $\hat{\mathbf{y}}_{\sigma} = \argmax_{\mathbf{y}} \Pr(\mathbf{y}|\sigma,\mathbf{z})$ are the LLM predictions given the full context and the compressed context, respectively. The function $s(\cdot,\cdot)$ measures the discrepancy between the two predictions.

One general framework for achieving this is knowledge distillation, where $\hat{\mathbf{y}}$ and $\hat{\mathbf{y}}_{\sigma}$ can be seen as the predictions of the teacher model and the student model, respectively. This formalization links our discussion to the context distillation problem discussed earlier. The training objective can be defined analogously to Eqs.~(\ref{eq:context-distillation-simple}) and~(\ref{eq:context-distillation-kl}). For instance, a straightforward maximum likelihood objective is given by
\begin{eqnarray}
\hat{\sigma} & = & \argmax_{\sigma} \log \Pr(\hat{\mathbf{y}}|\sigma, \mathbf{z}).
\end{eqnarray}

\noindent Alternatively, we can minimize the KL divergence between the output distributions, giving
\begin{eqnarray}
\hat{\sigma} & = & \argmin_{\sigma} \mathrm{D}_{\mathrm{KL}}(\Pr(\cdot|\mathbf{c},\mathbf{z}) \,\|\, \Pr(\cdot|\sigma,\mathbf{z})).
\end{eqnarray}

\noindent The difference with the models in Eqs.~(\ref{eq:context-distillation-simple}) and~(\ref{eq:context-distillation-kl}) is that here the compressed context is represented as real-valued vectors (call them \mindex{prompt embeddings}), rather than as normal tokens. By applying the above methods, we distill the context from the token sequence $\mathbf{c}$ into the embeddings $\sigma$. Note that the teacher model $\Pr(\cdot|\mathbf{c},\mathbf{z})$ and the student model $\Pr(\cdot|\sigma,\mathbf{z})$ may not share the same architecture or model settings. In practice, we generally wish for the teacher model to be stronger, while the student model should be smaller and more efficient.

While compressing full context into continuous representations is a straightforward approach to learning soft prompts, it requires a teacher model that can deal with long input sequences. In many cases, however, the context is so long that applying an LLM is too costly or infeasible. Modeling long input sequences can fall under the broad family of efficient methods for long-context LLMs. Many techniques have been developed to address this issue. For example, one can use a fixed-size KV cache to store the past information at each step during inference. Efficient Transformer architectures and long-context LLMs have been extensively discussed in this book. For more detailed discussions of these topics, interested readers can refer to Chapter \ChapterLLM.

There are also methods specifically designed to compress long sequences into soft prompts. Here we consider \citet{chevalier-etal:2023adapting}'s method as an example. The basic idea is to gradually learn soft prompts by compressing the input sequence into a fixed-size representation. For simplicity, we use $\mathbf z$ to denote the entire input sequence to be compressed. Given a long input sequence, we first divide it into a number of segments $\{\mathbf{z}^1, \cdots, \mathbf{z}^K\}$. We then process these segments in sequence, each time generating a representation of the input we have processed so far, denoted by $\sigma^{<i+1}$. To do this, a few summary tokens ${\langle \mathrm{g}_1 \rangle, \cdots, \langle \mathrm{g}_{\kappa} \rangle}$ are introduced. At each step, we take a segment $\mathbf{z}^i = z_1^{i} \cdots z_{m_i}^{i}$, along with the previous compressed representation $\sigma^{<i}$ and the summary tokens $\langle \mathrm{g}_1 \rangle, \cdots, \langle \mathrm{g}_{\kappa} \rangle$ as input, and use an LLM to produce the corresponding hidden representation sequence at the last Transformer layer. An example of this process is illustrated in Figure~\ref{fig:compressing-context-with-summary-tokens}.

\begin{figure}[!t]
\centering
\input{./Chapter9/Figures/figure-compressing-context-with-summary-tokens}
\caption{Illustration of compressing a context segment into soft prompts ($\kappa = 2$ and $m_i = 4$). The input to the LLM includes the soft prompts from the previous step ($\sigma^{<i}_1$ and $\sigma^{<i}_2$), the tokens of the segment ($z_1, z_2, z_3,$ and $z_4$), and the summary tokens ($\langle \mathrm{g}_1 \rangle$ and $\langle \mathrm{g}_2 \rangle$). Given these, the LLM operates as usual. We then extract the outputs at the last Transformer layer that correspond to the summary tokens. These outputs can be viewed as the soft prompts that accumulated up to this segment.}
\label{fig:compressing-context-with-summary-tokens}
\end{figure}

Here $\sigma^{<i}$ is essentially a memory. The model operates in an RNN fashion. Each time we take a segment and update this memory by encoding both the previous memory state and the segment. Therefore, the $\sigma^{<i}$ produced at the last segment is a representation of the entire context sequence. The Transformer model for learning these representations can be a standard LLM but we need to fine-tune it to adapt to this context representation task.

Note that here we simply consider \textit{prompt} and \textit{context} as similar terms, even though they are not the same. Although we use the term \textit{prompt} broadly to include contextual information supplied to the model., we can often view it as a type of context. From this perspective, the methods discussed here can be applied to general text compression problems.

\subsection{Prompt Length Reduction}

While soft prompts provide dense hidden representations, they are generally not directly interpretable. This lack of interpretability makes it difficult for users to understand how prompts influence LLM outputs. Moreover, modifying soft prompts often requires additional fine-tuning, which limits their utility in dynamic environments where prompts frequently change.

One alternative approach to improving prompt efficiency is to simplify the prompt text itself. Unlike soft prompts, simplified textual prompts remain interpretable and can be modified directly by users. For example, below is a prompt for answering questions on healthcare and finance.

\vspace{0.5em}
\begin{tcolorbox}[frame empty]

\begingroup
\setlength{\leftskip}{2em}
\setlength{\rightskip}{2em}

The task involves developing a language model capable of understanding and responding to user inquiries across various domains, with a particular emphasis on healthcare and finance. Considering the broad range of potential queries, from the specifics of medical diagnoses to the nuances of financial regulations, the model must ensure a comprehensive understanding and accurate responses.

\vspace{0.5em}

Question:

\vspace{0.2em}

What are the best practices for using artificial intelligence in diagnosing cardiovascular diseases?

\vspace{0.5em}

\underline{\hspace{2em}}

\endgroup

\end{tcolorbox}
\vspace{0.5em}

We can simplify the task description by deleting the unimportant parts.

\vspace{0.5em}

\begin{center}
\parbox{0.8\linewidth}{
\textit{The task involves developing a language model capable of understanding and responding to user inquiries {\color{gray} \sout{across various domains, with a particular emphasis}} on healthcare and finance. {\color{gray} \sout{Considering the broad range of potential queries, from the specifics of medical diagnoses to the nuances of financial regulations,}} The model must ensure a comprehensive understanding and accurate responses.}
}
\end{center}

\vspace{0.5em}

We can also paraphrase it into a shorter text.

\vspace{0.5em}

\begin{center}
\parbox{0.8\linewidth}{
\textit{The task involves developing a language model focused on healthcare and finance, capable of understanding and accurately responding to a wide range of user inquiries.}
}
\end{center}

\vspace{0.5em}

This problem can be viewed as a standard NLP task --- text simplification. Therefore, the methods used for text simplification are generally applicable and not limited to prompt simplification. There are many ways to achieve this. One simple method is to define some heuristics and identify redundant words that can be eliminated without losing essential information. For example, we can examine each token in a sequence in terms of its contribution to the overall meaning and remove those that provide minimal value~\cite{li-etal:2023compressing,jiang-etal:2023llmlingua}. Another method involves framing the problem as a sequence-to-sequence task. Given labeled data for text simplification, we can train an encoder-decoder model to transform each input text into its simplified form. In addition, given that many LLMs have been fine-tuned and aligned to perform text simplification tasks, it is straightforward to use these models to simplify prompts. For example, we can prompt an LLM to simplify a prompt while satisfying constraints such as length limits.

%----------------------------------------------------------------------------------------
% NEW SECTION
%----------------------------------------------------------------------------------------

\section{Summary}

\noindent In this chapter, we have discussed a variety of topics related to LLM prompting. Our discussion has focused mainly on two aspects:

\begin{itemize}
\item \vspace{0.5em} How to design basic prompts to guide the predictions of LLMs and refine these prompts for more effective and efficient problem-solving?
\item \vspace{0.3em} How to automate the design and representation of prompts?
\end{itemize}
\vspace{0.3cm}

\noindent Solutions to these issues involve both general prompt designs and more advanced techniques, such as CoT and prompt learning, which have been explored extensively in recent research.

In NLP, prompting can be viewed as a technology that has evolved along with LLMs, and in a sense, it has opened the door to the practical application of these models in an impressive range of problem domains. In fact, if we expand the concept of prompts to some extent, it can be traced back to the early days of machine learning and NLP. For example, many NLP systems use hand-crafted features and templates to ``prompt'' specific tasks. Imagine developing a feature to indicate whether a text is formal or informal. We can feed this feature into a machine translation system to condition the translation on the type of the input text.

The widespread use of the modern concept of prompts began with the rise of large pre-trained models in the field of NLP. Initially, these models, such as BERT, were adapted to specific downstream tasks mainly through fine-tuning. However, researchers soon discovered that by designing specific ``prompts'' --- adding certain words or sentences to the input --- the models could be triggered to respond to specific tasks without extensive fine-tuning. This motivated the NLP community to develop and apply universal foundation models that can be prompted to address various tasks without changing the underlying architecture and the pre-training procedure.

Prompting approaches were first experimented with smaller models and later demonstrated impressive capabilities with large models like GPT-3, which could generate high-quality text in response to simple prompts across various tasks. As prompting technology evolved, prompt engineering emerged as a critical area of research. As discussed in this chapter, it broadly involves designing effective prompts to maximize model performance, encompassing both hand-crafted and automatically generated prompts. More recent research has explored how to enhance the effectiveness of prompting through techniques like few-shot learning, zero-shot learning, and CoT reasoning, enabling LLMs to work effectively across a wide range of scenarios. A general discussion of prompting can be very broad, and we cannot cover all details in this chapter. For more advanced techniques of prompting, the reader can refer to recent surveys. Topics include in-context learning~\cite{li:2023practical,dong-etal:2022survey}, CoT~\cite{chu-etal:2023survey,yu-etal:2023towards,zhang-etal:2023igniting}, efficient prompting~\cite{chang-etal:2024efficient}, and general prompt engineering~\cite{liu2023-etal:prompting,chen-etal:2023unleashing}.

Note that although we would ideally like to develop general prompting methods without adjusting model architectures and parameters, the results of prompting generally depend heavily on the quality and size of the given LLMs. For stronger models, such as commercialized online LLMs, simple prompts may be sufficient to instruct these models to perform tasks correctly. In this case, prompt engineering is relatively easy, though careful prompt refinement remains necessary. By contrast, for less capable or smaller-scale models, we may need to carefully design the prompts to achieve the desired results. In many cases, fine-tuning is still necessary to adapt the models to sophisticated prompting strategies.

% !Mode:: "TeX:UTF-8"
% !TEX encoding = UTF-8 Unicode

%----------------------------------------------------------------------------------------
% Natural Language Processing: Neural Networks and Large Language Models
%
% Copyright 2021-2026
% Tong Xiao (xiaotong@mail.neu.edu.cn) and Jingbo Zhu (zhujingbo@mail.neu.edu.cn)
%----------------------------------------------------------------------------------------

%----------------------------------------------------------------------------------------
%    CONFIGURATIONS
%----------------------------------------------------------------------------------------

%\setcounter{chapter}{9}

%----------------------------------------------------------------------------------------
%	CHAPTER 10
%----------------------------------------------------------------------------------------

\chapter{Alignment}

\thispagestyle{chapterurl} % add urls to the header

\noindent Alignment is not a new concept in NLP, but its meaning varies across different domains and over time. In traditional NLP, the term \textit{alignment} typically refers to the task of linking corresponding elements between two sets, such as aligning words between a Chinese sentence and an English sentence. As LLMs become increasingly important in NLP research, this term is now more broadly used to describe the process of aligning model outputs with human expectations. The problem that alignment addresses is that the output of a model may not align with the specific goals or contexts intended by users. For example, pre-trained LLMs may not be able to follow user instructions because they were not trained to do so. Another example is that LLMs may generate harmful content or perpetuate biases inherent in their training data. This poses new challenges in ensuring that LLM outputs are not only accurate and relevant, but also ethically sound and non-discriminatory.

Pre-training LLMs alone can result in a variety of alignment problems. The ultimate goal of alignment is to resolve or mitigate these issues to ensure LLMs are both helpful and safe. Since LLMs are trained on large amounts of data, one might assume that with enough high-quality data, pre-training could make LLMs sufficiently accurate and safe, potentially eliminating the need for a separate alignment phase. However, the reality is that it is nearly impossible to gather data that encompasses all tasks or adequately represents human preferences. This makes it difficult to achieve alignment through pre-training alone. At present, alignment remains a critical step in the development of LLMs.

In this chapter, we focus on alignment methods for LLMs. We begin by discussing general alignment concepts. Then, we consider two widely used approaches: \mindex{instruction alignment} and \mindex{human preference alignment}. The former employs supervised fine-tuning techniques to guide LLMs in generating outputs that adhere closely to user instructions. The latter typically relies on reinforcement learning techniques, where models are trained based on human feedback. While these methods are motivated by different goals, they are commonly used together to develop well-aligned LLMs.

%----------------------------------------------------------------------------------------
% NEW SECTION
%----------------------------------------------------------------------------------------

\section{Overview of LLM Alignment}

\noindent Alignment can be achieved in several different ways. We need different methods for LLM alignment because the problem itself is complex and multifaceted, demanding a blend of technical strategies. Here we consider three widely-used approaches to aligning LLMs.

The first approach is to fine-tune LLMs with labeled data. This approach is straightforward as it simply extends pre-training to adapt the LLM to specific tasks. An example of this is \mindex{supervised fine-tuning} (\mindex{SFT}), in which the LLM is further trained on a dataset comprising task-specific instructions paired with their expected outputs. The SFT dataset is generally much smaller compared to the original training set, but this data is highly specialized. The result of SFT is that the LLM can learn to execute tasks based on user instructions. These can be tasks previously encountered during SFT or novel but similar tasks. For example, fine-tuning the LLM on a set of question-answer pairs enables the model to respond to specific questions, even those not directly covered in the SFT dataset. This method proves particularly useful when input-output relationships are relatively easy to describe and the data is straightforward to annotate.

The second approach is to fine-tune LLMs using reward models. One difficulty in alignment is that human values and expectations are complex and hard to describe. In many cases, humans themselves struggle to articulate exactly what is ethically correct or culturally appropriate. As a result, collecting and annotating fine-tuning data is not as straightforward as it is for SFT. Moreover, alignment is not merely a data-fitting task. The limited samples annotated by humans are often insufficient to comprehensively define these complex behaviors. Instead, the model must be taught to evaluate which outputs best reflect human preferences. For example, we not only want the outputs to be technically accurate but also to align with human expectations and values. One idea is to develop a reward model that mimics human judgments. This reward model would work by rewarding the LLM whenever it generates responses that align more closely with human preferences, much like how a teacher provides feedback to a student. To obtain such a reward model, we can train a scoring function from human preference data. The trained reward model is then used as a guide to adjust and refine the LLM. This frames the LLM alignment task as a reinforcement learning task. The resulting methods, such as \mindex{reinforcement learning from human feedback} (\mindex{RLHF}), have been demonstrated to be particularly successful in adapting LLMs to adhere to the subtleties of human behavior and social norms.

The third approach is to perform alignment during inference rather than during training or fine-tuning. From this perspective, prompting in LLMs can also be seen as a training-free form of alignment. So we can dynamically adapt an LLM to various tasks at minimal cost. Another approach to inference-time alignment is to rescore the outputs of an LLM. For example, we could develop a scoring system to simulate human feedback on the outputs of the LLM (like a reward model) and prioritize those that receive more positive feedback.

The three methods mentioned above are typically used in sequence once the pre-training is complete: we first perform SFT, then RLHF, and then prompt the LLM during inference. This roughly divides the development of LLMs into two stages --- the pre-training stage and the alignment stage. Figure \ref{fig:pre-training-then-alignment} shows an illustration of this. Since prompting techniques were extensively discussed in the previous chapter, the remainder of this chapter will focus primarily on fine-tuning-based alignment methods.

\begin{figure}[!t]
\centering
\input{./Chapter10/Figures/figure-pretrain-then-align}
\caption{Schematic illustration of the pre-training-then-alignment pipeline for developing LLMs. In the pre-training stage, we train an LLM on vast amounts of data using next token prediction. Then, in the alignment stage, we align the LLM to user instructions, intents, and preferences. This includes instruction alignment, human preference alignment, and prompting.}
\label{fig:pre-training-then-alignment}
\end{figure}

%----------------------------------------------------------------------------------------
% NEW SECTION
%----------------------------------------------------------------------------------------

\section{Instruction Alignment}

\noindent One feature of LLMs is that they can follow the prompts provided by users to perform various tasks. In many applications, a prompt consists of a simple instruction and user input, and the goal is for the LLM to perform the task accurately according to that instruction. This capability is commonly referred to as instruction following. For example, below is an example prompt in which the LLM is asked to extract key points and generate a concise summary of a lengthy article.

\vspace{0.5em}
\begin{tcolorbox}[frame empty]

\begingroup
\renewcommand{\arraystretch}{1.0}
\setlength{\tabcolsep}{6pt}
\begin{tabular}{r l}
{\color{gray} \small{Instruction}} & Summarize this text in three sentences. \\ [0.2cm]
{\color{gray} \small{Input}} & Daylight Savings Time (DST) - the process of moving clocks \\
& forward by one hour in the summer - was started in Germany in \\
& 1916. During World War One it was a way to save ... \\ [0.2cm]
{\color{gray} \small{Output}} & \underline{\hspace{2em}}
\end{tabular}
\endgroup

\end{tcolorbox}
\vspace{0.5em}

\noindent This task requires the LLM to understand the instruction ``Summarize this text in three sentences'' and summarize the text accordingly. However, pre-trained base LLMs are typically trained for next-token prediction rather than for generating outputs that follow instructions. Applying a pre-trained LLM to the above example would likely cause the model to continue the article instead of summarizing it. The goal of instruction alignment (or \mindex{instruction fine-tuning}) is to tune the LLM to accurately respond to user instructions and intentions. The rest of this section will discuss some topics related to instruction alignment, including fine-tuning LLMs to follow instructions, generating or collecting instruction data, and generalizing instruction alignment.

\subsection{Supervised Fine-tuning}

\noindent One straightforward approach to adapting LLMs to follow instructions is to fine-tune these models using annotated input-output pairs \cite{ouyang-etal:2022training,wei-etal:2022finetuned}. In the context of LLM alignment, SFT is often implemented as instruction fine-tuning. Unlike standard language model training, where the loss is computed over all tokens in a sequence, SFT computes the loss only over the output tokens conditioned on the input. Let $\mathbf{x}=x_0...x_m$ be an input sequence (e.g., instruction + user input) and $\mathbf{y}=y_1...y_n$ be the corresponding output sequence. SFT data consists of such input-output pairs (denoted by $S$), where each output is a desired response corresponding to the input. For example, the following table shows examples from an SFT dataset.

\begin{center}
\begin{tabular}{l | l}
$\mathbf{x}$ (instruction + user input) & $\mathbf{y}$ (output) \\ \hline
\colorbox{gray!30}{Summarize the following article.} & $\{*\textrm{summary}*\}$ \\
Article: In recent years, solar energy has seen  & \\
unprecedented growth, becoming the fastest-growing ... & \\ \hline
\colorbox{gray!30}{Analyze the sentiment of the following review.} & Positive \\
Review: I absolutely loved the new dining experience. & \\
The food was divine and the service was impeccable. & \\ \hline
\colorbox{gray!30}{Translate the following sentence into French.} & La pratique aide \\
Sentence: practice indeed helps. & effectivement. \\ \hline
\colorbox{gray!30}{Extract the main financial figures from the following} & Revenue: \$10 million, \\
\colorbox{gray!30}{earnings report.} & Profit Margin: 15\% \\
Report: The company reported a revenue of \$10 million\\
in the first quarter with a profit margin of 15\% ... \\ \hline
\colorbox{gray!30}{Classify the following email as spam or not spam.} & Spam \\
Text: Congratulations! You've won a \$500 gift card. \\
Click here to claim now. \\ \hline
\colorbox{gray!30}{Provide a solution to the following technical issue.} & First, check for ... \\
Issue: my computer is running slow and often freezes. &
\end{tabular}
\end{center}

\noindent where the instructions are highlighted. This dataset contains instructions and the corresponding outputs for several different NLP problems, and so we can fine-tune an LLM to handle multiple tasks simultaneously.

In SFT, we aim to maximize the probability of the correct output given the input. Consider an LLM with pre-trained parameters $\hat{\theta}$. The fine-tuning objective can then be formulated as:
\begin{eqnarray}
\tilde{\theta} & = & \argmax_{\hat{\theta}^+} \sum_{(\mathbf{x},\mathbf{y}) \in \mathcal{D}} \log \mathrm{Pr}_{\hat{\theta}^+}(\mathbf{y}|\mathbf{x}), \label{eq:instruction-fine-tuning}
\end{eqnarray}

\noindent where $\tilde{\theta}$ denotes the parameters optimized via fine-tuning, and $\hat{\theta}^+$ represents an adjustment to $\hat{\theta}$. Here we will omit the superscript $+$ and use $\theta$ to represent $\hat{\theta}^+$ to keep the notation uncluttered. But the reader should keep in mind that the fine-tuning starts from the pre-trained parameters rather than randomly initialized parameters.

The objective function $\log \mathrm{Pr}_{\theta}(\mathbf{y}|\mathbf{x})$ is computed by summing the log-probabilities of the tokens in $\mathbf{y}$, conditional on the input $\mathbf{x}$ and all the previous tokens $\mathbf{y}_{<i}$:
\begin{eqnarray}
\log \mathrm{Pr}_{\theta}(\mathbf{y}|\mathbf{x}) & = & \sum_{i=1}^{n} \log \mathrm{Pr}_{\theta}(y_i|\mathbf{x},\mathbf{y}_{<i}).
\end{eqnarray}

\noindent This formulation is equivalent to minimizing the cross-entropy loss.

If we concatenate $\mathbf{x}$ and $\mathbf{y}$ as a single sequence, a more general form of language modeling is based on the joint log-probability $\log \mathrm{Pr}_{\theta}(\mathbf{x},\mathbf{y})$, that is, we minimize the loss (or maximize the log-likelihood) over all tokens of the sequence $\mathrm{seq}_{\mathbf{x},\mathbf{y}} = [\mathbf{x},\mathbf{y}]$. We can write the probability of this sequence using the chain rule
\begin{eqnarray}
\log \mathrm{Pr}_{\theta}(\mathrm{seq}_{\mathbf{x},\mathbf{y}}) & = & \log \mathrm{Pr}_{\theta}(\mathbf{x},\mathbf{y}) \nonumber \\
 & = & \underbrace{\log \mathrm{Pr}_{\theta}(\mathbf{x})}_{\text{set to $0$}} + \underbrace{\log \mathrm{Pr}_{\theta}(\mathbf{y}|\mathbf{x})}_{\text{loss computation}}. \label{eq:sft-xy-chain-rule}
\end{eqnarray}

\noindent There are two terms on the right-hand side of the equation. We can simply ignore the first term $\log \mathrm{Pr}_{\theta}(\mathbf{x})$, and focus only on the second term $\log \mathrm{Pr}_{\theta}(\mathbf{y}|\mathbf{x})$ for loss computation. As a result, the training can be implemented using standard LLMs. For the sequence $\mathrm{seq}_{\mathbf{x},\mathbf{y}}$, we first run the forward pass as usual. Then, during the backward pass, we force the loss corresponding to $\mathbf{x}$ to be zero. Figure \ref{fig:sft-forward-backward} shows an illustration of this process.

\begin{figure}[!t]
\centering
\input{./Chapter10/Figures/figure-sft-forward-and-backward}
\caption{Illustration of supervised fine-tuning for LLMs. We concatenate the input and the output into a single sequence. During the forward pass, we run the LLM as usual. During the backward pass, we compute the loss only for the output part and simply set the loss for the input part to 0.}
\label{fig:sft-forward-backward}
\end{figure}

By taking $\log \mathrm{Pr}_{\theta}(\mathrm{seq}_{\mathbf{x},\mathbf{y}})$ as the objective function, we can describe SFT using a regular form of language model training:
\begin{eqnarray}
\tilde{\theta} & = & \argmax_{\theta} \sum_{(\mathbf{x},\mathbf{y}) \in \mathcal{D}} \log \mathrm{Pr}_{\theta}(\mathrm{seq}_{\mathbf{x},\mathbf{y}}). \label{eq:sft-single-round-prediction-mle}
\end{eqnarray}

The problem we considered above is fundamentally a \mindex{single-round prediction} problem, where the LLM generates a response based on a single input without any further interaction or feedback from the user. The input is processed, and the output is generated in one go. This is typical in scenarios where a single question is asked, and a single answer is provided, with no follow-up questions or clarifications. However, in practice, we sometimes have to handle multi-round prediction problems, for example, an LLM engages in a dialogue over multiple turns. In this setting, the LLM not only generates responses based on the initial input but also incorporates subsequent inputs that might refine or expand on earlier interactions. For example, we can use the LLM to act as a healthcare assistant chatbot and have a conversation with the user, like this

\vspace{0.5em}
\begin{tcolorbox}[frame empty]

\begingroup
\renewcommand{\arraystretch}{1.0}
\setlength{\tabcolsep}{6pt}
\begin{tabular}{r l}
{\color{gray} \small{User}} & I've been feeling very tired lately.\\  [0.1cm]
{\color{gray} \small{Chatbot}} & \uline{I'm sorry to hear that. Besides feeling tired, have you noticed any}  \\
& \uline{other symptoms?} \\ [0.2cm]
{\color{gray} \small{User}} & Yes, I'm also experiencing headaches frequently.\\  [0.1cm]
{\color{gray} \small{Chatbot}} & \uline{How long have these symptoms been going on?}  \\ [0.2cm]
{\color{gray} \small{User}} & About a week now.\\  [0.1cm]
{\color{gray} \small{Chatbot}} & \uline{It might be good to check in with a healthcare professional. Would}  \\
& \uline{you like help setting up an appointment?} \\ [0.2cm]
{\color{gray} \small{User}} & Yes, please. Can it be after work hours?\\  [0.1cm]
{\color{gray} \small{Chatbot}} & \uline{Sure, I can arrange that. There are slots available next Wednesday} \\ [0.1cm]
& \uline{and Thursday after 5 PM. Which day works better for you?} \\ [0.2cm]
& ...
\end{tabular}
\endgroup

\end{tcolorbox}
\vspace{0.5em}

In this task, there are several rounds of conversation, each involving the generation of a response based on the user's request or question and the conversational history. Suppose we have $K$ rounds of conversation, denoted by $\{\mathbf{x}^1,\mathbf{y}^1,\mathbf{x}^2,\mathbf{y}^2,...,\mathbf{x}^K,\mathbf{y}^K\}$. Here $\mathbf{x}^k$ and $\mathbf{y}^k$ denote the user request and the response, respectively, for each round $k$. The log-probability of generating the response can be written as $\log \mathrm{Pr}_{\theta}(\mathbf{y}^k|\mathbf{x}^1,\mathbf{y}^1,...,\mathbf{x}^k)$. Our goal is then to maximize the sum of these log-probabilities
\begin{eqnarray}
\tilde{\theta} & = & \argmax_{\theta} \sum_{k=1}^{K} \log \mathrm{Pr}_{\theta}(\mathbf{y}^k|\mathbf{x}^1,\mathbf{y}^1,...,\mathbf{x}^k). \label{eq:sft-xy-conversation-objective-function}
\end{eqnarray}

A straightforward implementation of this involves calculating the conditional probability for each $k$. However, it requires running the LLM $K$ times, each time with an increased conversational history to make predictions. A more efficient method is to perform loss computation of all responses in a single run of the LLM. To do this, we represent the conversation as a sequence $\mathrm{seq}_{\mathbf{x}^1,\mathbf{y}^1,...,\mathbf{x}^K,\mathbf{y}^K} = [\mathbf{x}^1,\mathbf{y}^1,...,\mathbf{x}^K,\mathbf{y}^K]$ (or $\mathrm{seq}$ for short). The log-probability of this sequence is given by
\begin{eqnarray}
\log \mathrm{Pr}_{\theta}(\mathrm{seq}) & = & \log \mathrm{Pr}_{\theta}(\mathbf{x}^1,\mathbf{y}^1,...,\mathbf{x}^K,\mathbf{y}^K) \nonumber \\
                                        & = & \underbrace{\log \mathrm{Pr}_{\theta}(\mathbf{x}^1)}_{\text{set to $0$}} + \underbrace{\log \mathrm{Pr}_{\theta}(\mathbf{y}^1|\mathbf{x}^1)}_{\text{loss computation}} + \cdots + \nonumber \\
                                        &   & \underbrace{\log \mathrm{Pr}_{\theta}(\mathbf{x}^K|\mathbf{x}^1,\mathbf{y}^1,...,\mathbf{y}^{K-1})}_{\text{set to $0$}} + \nonumber \\
                                        &   & \underbrace{\log \mathrm{Pr}_{\theta}(\mathbf{y}^K|\mathbf{x}^1,\mathbf{y}^1,...,\mathbf{x}^K)}_{\text{loss computation}}. \label{eq:sft-xy-conversation-chain-rule}
\end{eqnarray}

The trick here is that we ignore the loss for generating user inputs (i.e., $\log \mathrm{Pr}_{\theta}(\mathbf{x}^1),...,\allowbreak \log \mathrm{Pr}_{\theta}(\mathbf{x}^K|\mathbf{x}^1,\mathbf{y}^1,...,\mathbf{y}^{K-1})$), as illustrated in Figure \ref{fig:sft-conversational-models}. Hence we only compute the probabilities of generating the responses given their conversational histories, in other words, Eq. (\ref{eq:sft-xy-conversation-chain-rule}) is equivalent to Eq. (\ref{eq:sft-xy-conversation-objective-function}). As with Eq. (\ref{eq:sft-single-round-prediction-mle}), the training of this multi-round prediction model can be achieved by maximizing the log likelihood over a training dataset $\mathcal{D}$:
\begin{eqnarray}
\tilde{\theta} & = & \argmax_{\theta} \sum_{\mathrm{seq} \in \mathcal{D}} \log \mathrm{Pr}_{\theta}(\mathrm{seq}).
\end{eqnarray}

\begin{figure}[!t]
\centering
\input{./Chapter10/Figures/figure-sft-multi-round-prediction}
\caption{Illustration of supervised fine-tuning for conversational models. Here the LLM acts as a chatbot to respond to each request based on the conversational history. The conversation progresses by alternating between the user and the chatbot. In SFT, we treat the entire conversation as a sequence, just like in standard LLMs, but compute the loss only for the responses of the LLM.}
\label{fig:sft-conversational-models}
\end{figure}

While implementing the SFT methods introduced above seems trivial as they are fundamentally the same as regular language model training, there are still issues that need to be considered in practice. For example,

\begin{itemize}
\item \vspace{0.5em} SFT requires labeled data. This makes SFT quite different from pre-training, where raw text is used as training data and is readily available. As in other supervised machine learning problems, data annotation and selection in SFT are not simple tasks. In general, we wish to develop SFT data that is both substantial in quantity and high in quality, and this data should be highly relevant to the tasks the LLM will perform. On the other hand, there is a need to fine-tune LLMs with less data to minimize computational and data construction costs. Often, the quality of LLMs is highly dependent on the data used in SFT. Thus, such data must be carefully developed and examined. As we will see in later subsections, SFT can be more efficient and effective through more advanced techniques for data construction.
\item \vspace{0.3em} SFT is still computationally expensive for LLMs due to their large size. As a result, maintaining and updating such models is resource-intensive. For example, applying gradient updates to billions of parameters within an LLM requires significant computational power and memory. This often requires high-performance computing environments, which are costly to operate. To address these challenges, various optimization strategies, such as pruning, quantization, and the use of more efficient training algorithms, have been explored. In particular, there has been significant interest in parameter-efficient fine-tuning methods which are designed to maintain state-of-the-art performance without the need for extensive computational resources. We have seen in Chapter \ChapterLLMPrompting\ that applying techniques like soft prompts can make the fine-tuning process more efficient. For further discussion on parameter-efficient methods, the reader can refer to related papers on this issue \cite{houlsby-etal:2019parameter,hu-etal:2022lora,han-etal:2024parameter}.
\item \vspace{0.3em} SFT can be regarded as a post-training step following pre-training. It is a separate training phase designed to preserve the advantages of the initial pre-training while incorporating new adjustments. This may seem paradoxical because updating a pre-trained LLM with further data potentially causes the model to forget some of its prior knowledge. Imagine a scenario where we have a large amount of SFT data and extensively fine-tune the LLM. In this case, the LLM could overfit the data, which in turn may reduce generalization performance or cause catastrophic forgetting. A common strategy to mitigate this issue is to employ regularization and early stopping techniques. Another practical approach is to use a smaller learning rate to gently adjust the weights of the LLM. In addition, fine-tuning with data from diverse sources and problem domains can also be beneficial. Nevertheless, in practice, the SFT step is often carefully examined and requires substantial engineering and experimental efforts to optimize.
\end{itemize}
\vspace{0.5em}

\subsection{Fine-tuning Data Construction}

\noindent Fine-tuning data is so important that a large body of recent research on LLMs has focused on developing datasets for instruction fine-tuning. As with most machine learning tasks, there are generally two approaches to constructing fine-tuning datasets: manually generated data and automatically generated data.

\subsubsection{Manually Generated Data}

\noindent One straightforward method is to recruit human annotators to create input-output pairs for the tasks of interest. Unlike data annotation in conventional NLP, such as text classification, where annotators simply assign labels to collected texts according to guidelines, creating fine-tuning data for LLMs requires more steps and effort, thus making it more challenging. Suppose we want to obtain fine-tuning data for the English-to-Chinese machine translation task. The first step is to write a prompt template to describe the task and format the problem clearly. For example,

\vspace{0.5em}
\begin{tcolorbox}[frame empty]

\begingroup
\renewcommand{\arraystretch}{1.0}
\setlength{\tabcolsep}{6pt}
\begin{tabular}{r l}
{\color{gray} \small{Instruction}} & Translate the text from English to Chinese. \\ [0.2cm]
{\color{gray} \small{User Input}} & $\{*\mathrm{text}*\}$  \\ [0.2cm]
{\color{gray} \small{Output}} & \underline{$\{*\mathrm{translation}*\}$}
\end{tabular}
\endgroup

\end{tcolorbox}
\vspace{0.5em}

Then, we collect pairs of source and target texts (i.e., English texts and their corresponding Chinese translations), and replace the variables $\{*\mathrm{text}*\}$ and $\{*\mathrm{translation}*\}$ to generate the fine-tuning samples. For example, given a pair of English and Chinese sentences

\begin{center}
\begin{tabular}{r c l}
How's the weather today? & $\to$ & \begin{CJK*}{UTF8}{gbsn} 今天天气怎么样？ \end{CJK*} \\
\multicolumn{1}{c}{$\{*\mathrm{text}*\}$} & & \multicolumn{1}{c}{$\{*\mathrm{translation}*\}$}
\end{tabular}
\end{center}

\noindent we can generate a fine-tuning sample using the prompt template, like this

\vspace{0.5em}
\begin{tcolorbox}[frame empty]

\begingroup
\renewcommand{\arraystretch}{1.0}
\setlength{\tabcolsep}{6pt}
\begin{tabular}{r l}
{\color{gray} \small{Instruction}} & Translate the text from English to Chinese. \\ [0.2cm]
{\color{gray} \small{User Input}} & How's the weather today?  \\ [0.2cm]
{\color{gray} \small{Output}} & \underline{\begin{CJK*}{UTF8}{gbsn} 今天天气怎么样？ \end{CJK*}}
\end{tabular}
\endgroup

\end{tcolorbox}
\vspace{0.5em}

\noindent That is,
\begin{eqnarray}
& \mathbf{x}: & \text{Translate the text from English to Chinese.$\backslash$n How's the weather today?} \nonumber \\ & \mathbf{y}: & \text{\begin{CJK*}{UTF8}{gbsn} 今天天气怎么样？ \end{CJK*}} \nonumber
\end{eqnarray}

\noindent We can use this $(\mathbf{x},\mathbf{y})$ pair to fine-tune the LLM, as described in the previous subsection.

One difficulty here is that there are many different ways to write prompt templates for the same task, and different people may produce prompt templates with varying qualities and complexities. Sometimes, we may write prompt templates with overly complex or verbose instructions. Sometimes, we may not even know exactly what the target task is and how to describe it. A widely-adopted strategy is to create prompt templates for existing NLP tasks, given that there have been so many well-established NLP problems and benchmarks \cite{bach-etal:2022promptsource,wang-etal:2022super,mishra-etal:2022cross}. In this case, annotators can be given the original task description and many examples. Then, they can use their own ways to express how to prompt the LLM to perform the tasks. Note that, while such a method can ease the process of creating prompts, we still need annotation frameworks and crowdsourcing systems to manage the work and conduct quality control. For example, we generally need to design annotation guidelines and a unified format for writing prompt templates, especially when many annotators are contributing to the same task. One advantage of inducing prompts from existing NLP tasks is that, once the prompt templates have been developed, it is easy to generate fine-tuning examples using the annotated samples in the original datasets. For example, given a bilingual dataset for English-to-Chinese translation, we can easily create a number of fine-tuning examples by filling the slots in the above template with the sentence pairs in this dataset.

Another approach is to directly use the naturally existing data available on the internet. A common example is to collect question-and-answer pairs from QA websites to fine-tune LLMs for open-domain QA tasks \cite{joshi-etal:2017triviaqa}. Many benchmarks in QA are constructed in this way because there are so many types of questions that it is difficult for a small group of annotators to anticipate all possible question types. Instead, using data from those websites can ensure that the LLM fine-tuning data is sufficiently large and diverse.

In addition to employing existing resources, another straightforward way to develop a fine-tuning dataset is to crowdsource the data. A simple approach is to allow users to input any question, after which responses are either manually written or automatically generated by an LLM and then reviewed and corrected by human annotators. It is thus possible to capture real user behavior and consequently gather inputs and outputs for a large number of new problems that traditional NLP tasks do not cover.

An issue related to the construction of the fine-tuning datasets is that we usually want the data to be as diverse as possible. Many studies have found that increasing the diversity of fine-tuning data can improve the robustness and generalization ability of LLMs. For this reason, there has been considerable interest in involving more diverse prompts and tasks in LLM fine-tuning datasets. We will provide further discussion on the generalization of fine-tuning in Section \ref{sec:instruction-generalization}.

\subsubsection{Automatically Generated Data}

\noindent One limitation of manual data generation is that the quality and diversity largely depend on human experience and creativity. Therefore, if we expect LLMs to generalize to diverse user instructions, relying on human-annotated data for LLM fine-tuning is often inefficient. Moreover, the coverage of such data can be limited, and the data may even contain biases introduced by the annotators themselves. An alternative approach is to generate data automatically. For example, we can collect a number of questions through crowdsourcing, and employ a well-tuned LLM to generate answers to the questions. These question-answer pairs are then used as fine-tuning samples as usual. This method, though very simple, has been extensively applied to generate large-scale fine-tuning data for LLMs.

The above approach to producing synthetic fine-tuning data is similar to methods used in NLP data augmentation. If we have an LLM, we can produce an output in response to any input. Repeating this process for different inputs allows us to create large amounts of fine-tuning data.  Such a method is particularly useful for fine-tuning new LLMs using a well-tuned LLM. However, one disadvantage of this approach is that it relies on human-crafted or collected inputs for data generation, which may turn out to be inappropriate for generalizing LLMs. In many LLM applications, a significant challenge arises from the broad range of user questions and requests, many of which are not covered in existing NLP tasks and datasets. In these cases, it becomes necessary to generate not only the predictions but also the inputs themselves.

Here we consider \mindex{self-instruct} as an example to illustrate how to generate LLM fine-tuning samples \cite{wang-etal:2023selfinstruct,honovich-etal:2023unnatural}. The idea is that we can prompt an LLM to create a new instruction by learning from other instructions. Given this instruction, the LLM can then fill in other fields (such as the user input) and generate outputs. Figure \ref{fig:self-instruct} shows a schematic illustration of self-instruct. Here we give a brief outline of the key steps involved.

\begin{figure}[!t]
\centering
\input{./Chapter10/Figures/figure-self-instruct}
\caption{Illustration of self-instruct \cite{wang-etal:2023self}. This method maintains a pool of instructions and corresponding input-output samples. Initially, the pool contains a number of hand-crafted instructions and samples. Each time, we draw a few instructions from the pool. An LLM is then prompted to generate new instructions and samples based on those drawn. Finally, the newly generated instructions and samples are filtered and added to the pool.}
\label{fig:self-instruct}
\end{figure}

\begin{itemize}
\item \vspace{0.5em} The self-instruct algorithm maintains a pool of tasks. Initially it contains a number of seed hand-crafted tasks, each with an instruction and input-output sample. As the algorithm proceeds, LLM-generated instructions and samples will be added to this pool.
\item \vspace{0.3em} At each step, a small number of instructions are drawn from the instruction pool. For example, we can randomly select a few human-written instructions and a few LLM-generated instructions to ensure diversity.
    \newpage
\item \vspace{0.3em} The selected instructions are used as in-context demonstrations for generating a new instruction. Below is an example template for prompting the LLM.

    \vspace{0.5em}
    \begin{tcolorbox}[frame empty]

    \begingroup
    \setlength{\leftskip}{2em}
    \setlength{\rightskip}{2em}

    You are provided several different instructions for performing some tasks. Please generate an instruction based on these.

    \vspace{0.2cm}

    Task 1: $\{*\mathrm{instruction}1*\}$

    \vspace{0.2cm}

    Task 2: $\{*\mathrm{instruction}2*\}$

    \vspace{0.2cm}

    Task 3: $\{*\mathrm{instruction}3*\}$

    \vspace{0.2cm}

    Task 4: $\{*\mathrm{instruction}4*\}$

    \vspace{0.2cm}

    New Task: \underline{\hspace{2em}}

    \endgroup

    \end{tcolorbox}
    \vspace{0.5em}

\item \vspace{0.3em} Given the generated instruction, the LLM is then prompted to complete the sample by filling in the remaining input fields and generating the corresponding output. Below is a prompt template.

    \vspace{0.5em}
    \begin{tcolorbox}[frame empty]

    \begingroup
    \setlength{\leftskip}{2em}
    \setlength{\rightskip}{2em}

    You are provided with a set of input-output samples, each composed of an instruction, a user input, and an output. Please generate a new sample based on these.

    \vspace{0.2cm}

    Sample 1: $\{*\mathrm{instruction}1*\}$

    \vspace{0.1cm}

    Input: $\{*\mathrm{user\text{-}input}1*\}$

    \vspace{0.1cm}

    Output: $\{*\mathrm{output}1*\}$

    \vspace{0.2cm}

    Sample 2: $\{*\mathrm{instruction}2*\}$

    \vspace{0.1cm}

    Input: $\{*\mathrm{user\text{-}input}2*\}$

    \vspace{0.1cm}

    Output: $\{*\mathrm{output}2*\}$

    \vspace{0.2cm}

    New Sample: $\{*\mathrm{new\text{-}instruction}*\}$

    \vspace{0.1cm}

    \underline{\hspace{2em}}

    \endgroup

    \end{tcolorbox}
    \vspace{0.5em}

\item \vspace{0.3em} This newly generated sample is evaluated or filtered using heuristic rules (such as filtering out samples or instructions that are similar to those already in the pool). If it passes, the sample and instruction are added to the pool.
\end{itemize}
\vspace{0.5em}

This generation process can be repeated many times to obtain a sufficient number of fine-tuning samples. Note that, above, we just show simple prompt templates for generating instruction and fine-tuning samples. Of course, we can develop better templates to generate more diverse and accurate instruction and fine-tuning samples. For example, for certain tasks like text classification, the LLM may tend to produce biased predictions, for example, most generated samples belong to a single class. In such cases, we can adjust the order of generation of different fields. More specifically, we can specify the output (i.e., the class) with some prior, and prompt the LLM to generate user input given both the instruction and the output. This method resembles \mindex{input inversion}, where the LLM generates the input based on the specified output \cite{longpre-etal:2023flan}.

Using LLM-generated instructions and fine-tuning samples has been a common method for developing LLMs, especially given that manually developing such data is so expensive that most research groups cannot afford it. In several well-tuned LLMs, their fine-tuning datasets include a certain amount of synthetic data, which has been shown to be effective \cite{ouyang-etal:2022training,taori-etal:2023alpaca,chiang-etal:2023vicuna}. There have been further studies on synthetic data generation for LLM fine-tuning. For example, one can generate more diverse instructions by introducing evolutionary algorithms \cite{xu-etal:2024wizardlm}, or use synthetic data as supervision signals in a more advanced fine-tuning process \cite{chen-etal:2024self}. More recently, there has also been considerable interest in using synthetic data in the pre-training stage \cite{gunasekar-etal:2023textbooks,allal-etal:2024cosmopedia}.

In many applications, a real-world scenario is that, given a task, we can collect or annotate a relatively small amount of fine-tuning data, for example, we can recruit experts to create questions for QA tasks in a specific domain. But the quantity and diversity of this data are in general not sufficient. In this case, we can use self-instruct techniques to generate more diverse question-answer pairs, and thus augment the fine-tuning data. This provides a way of bootstrapping the LLM starting from a seed set of fine-tuning samples. Note that using self-generated data is a common practice and has long been applied in NLP. For example, this approach has been successfully used in parsing and machine translation \cite{charniak:1997statistical,sennrich-etal:2016improving}.

\subsection{Fine-tuning with Less Data}

\noindent With the increasing prominence of instruction fine-tuning, there has been a surge in demand for large-scale, high-quality fine-tuning data. For example, the FLAN fine-tuning dataset, which is compiled from 1,836 tasks, contains 15 million samples \cite{longpre-etal:2023flan}. Fine-tuning LLMs with such large datasets is typically a computationally expensive task, especially given that updating the large number of parameters in LLMs is resource-intensive. One approach for mitigating this issue is to explore efficient model training methods, for example, one can use parameter-efficient methods to update only a small portion of the model.

Another approach is to fine-tune LLMs using only the most informative training examples. We can thus reduce the amount of data that needs to be processed while still maintaining fine-tuning performance. There are several methods to achieve this. For example, \citet{zhou-etal:2023lima} built an instruction-following dataset containing only 1,000 samples by carefully crafting the prompts and collecting samples from a variety of NLP tasks. They showed that the LLaMa 65B model fine-tuned with this dataset could be competitive with or even better than models with much more fine-tuning effort. This suggests that LLMs can be adapted to respond to diverse tasks without necessarily needing fine-tuning on all types of instruction-following data. \citet{chen-etal:2024alpagasus} developed a system based on the GPT-3.5 model to assess the quality of each instruction-following sample. Therefore, they could select high-quality samples from existing datasets, showing better fine-tuning performance with fewer fine-tuning samples. Researchers have also developed methods to either select or filter out data using heuristics \cite{zhao-etal:2024long,ge-etal:2024clustering}, or to prioritize data that more significantly influences the fine-tuning process \cite{xia-etal:2024less}. In fact, most of these methods can be seen as instances of larger families of data selection and filtering methods. And it is often the case that using higher quality (but maybe less) data is beneficial for training NLP models.

The findings in instruction fine-tuning somewhat differ from traditional views in NLP: the ability of models to handle complex problems can be activated with a small amount of annotated data, rather than requiring massive amounts of supervised data for extensive training. One possible explanation is that the ability of generating correct responses given instructions has been learned during pre-training, but the LLM may not assign high probability to such responses during inference. Fine-tuning can slightly adjust the models to get them to follow instructions, and thus require significantly less training effort than pre-training. This is closely related to what is known as the \mindex{superficial alignment hypothesis}, which suggests that learning primarily occurs during pre-training, and the subsequent fine-tuning or alignment phase does not significantly contribute to the underlying knowledge base of an LLM \cite{zhou-etal:2023lima}. Since the core abilities and knowledge of the model are already established from pre-training, effective fine-tuning for alignment with user needs can be achieved with relatively small fine-tuning effort. This implies the possibility of fine-tuning LLMs with very little data. In another direction, it may not be necessary to restrict fine-tuning to paired instruction-response data. For example, \citet{hewitt-etal:2024instruction} found that instruction-following can be implicitly achieved by fine-tuning LLMs only on responses, without corresponding instructions.

A concept related to the discussion here is sample efficiency. A machine learning method is called \mindex{sample efficient} if it can learn effectively from a small number of training examples. In this sense, instruction fine-tuning is sample efficient compared with pre-training. From the perspective of machine learning, sample-efficient methods can be seen as effective ways to sample the space of data, and are advantageous as they make optimal use of scarce data. These ideas are also relevant to reinforcement learning settings, where sample efficiency is particularly important. Therefore, sampling-based learning techniques, such as many reinforcement learning algorithms, can benefit from these sample efficient approaches. For example, in human preference alignment, we can either efficiently sample preference data via reward models \cite{liu-etal:2024statistical} or improve the sampling efficiency in policy learning \cite{wang-etal:2024esrl}.

\subsection{Instruction Generalization}
\label{sec:instruction-generalization}

In many machine learning and NLP problems, training a model to generalize is a fundamental goal. For example, in text classification, we expect our model to correctly classify new texts that were not seen during training. However, generalization poses additional challenges in instruction fine-tuning. We expect instruction-fine-tuned LLMs to not only generate appropriate responses for different inputs within a task but also to accurately perform various tasks as described by different instructions. To illustrate this issue, consider an LLM $\Pr(\mathbf{y}|\mathbf{c},\mathbf{z})$, where $\mathbf{c}$ is an instruction, $\mathbf{z}$ is a user input, and $\mathbf{y}$ is the corresponding model output (i.e., the response). Suppose that the performance of this model is evaluated in terms of a metric, written as $\mathrm{Performance}(\Pr(\mathbf{y}|\mathbf{c},\mathbf{z}))$ or $\mathrm{P}(\mathbf{c},\mathbf{z},\mathbf{y})$ for short. Informally, when we say this model can generalize within a given task (indicated by the instruction $\mathbf{c}^*$), there exists some threshold $\epsilon$ such that the average performance on new inputs is above this value:
\begin{eqnarray}
\frac{1}{|\mathcal{Z}|} \sum_{\mathbf{z}'\in \mathcal{Z}} \mathrm{P}(\mathbf{c}^*,\mathbf{z}',\mathbf{y}') > \epsilon,
\end{eqnarray}

\noindent where $\mathcal{Z}$ is the set of new inputs, and $\mathbf{z}'$ is an input, and $\mathbf{y}'$ is the model output corresponding to input $\mathbf{z}'$.

Likewise, we can say that this model can generalize across tasks if the average performance over all instruction-input pairs is above some $\epsilon$:
\begin{eqnarray}
\frac{1}{|\mathcal{D}|} \sum_{(\mathbf{c}',\mathbf{z}') \in \mathcal{D}} \mathrm{P}(\mathbf{c}',\mathbf{z}',\mathbf{y}') > \epsilon,
\end{eqnarray}

\noindent where $\mathcal{D}$ is the set of new instruction-input pairs.

Here, we need to deal with variations in two dimensions: instruction and user input. This makes the generalization problem very complex, because, intuitively, a model needs to learn from a large number of tasks and different input-output pairs associated with each task to achieve good generalization. As we have discussed several times in this book, achieving such generalization incurs much lower cost than pre-training. In general, fine-tuning LLMs with instruction-response data to some extent can lead to models eliciting instruction-following capabilities on new tasks. Nevertheless, it is typically believed that certain efforts are still needed to adapt LLMs to make them understand and execute instructions broadly.

One way to generalize instruction fine-tuning is to increase the diversity of the fine-tuning data. In earlier studies on instruction fine-tuning, researchers developed many datasets, covering a wide variety of NLP tasks and different instructions for each task \cite{wang-etal:2022super,sanh-etal:2022multitask,longpre-etal:2023flan}. By transforming these tasks into a unified format, one can fine-tune an LLM with a sufficiently large number of samples. For example, several instruction fine-tuning datasets contain over 100 NLP tasks and more than one million samples. However, these early datasets mostly focus on existing academic problems, but not tasks encountered in real-world applications. Much recent work has shifted focus to addressing new and more practical problems. For example, there has been considerable interest in constructing datasets that contain large and complex demonstrations and responses from SOTA models to real user queries \cite{wang-etal:2023far,teknium:2023openhermes}.

Perhaps the use of large and diverse fine-tuning datasets has its origins in attempts to scale LLMs in different dimensions. Indeed, scaling laws have been used broadly to motivate the development of a wide range of different instruction-fine-tuned LLMs. And it is reasonable to scale instruction fine-tuning to make an LLM follow broad instructions. From the perspective of LLM alignment, however, scaling instruction fine-tuning might not be the most efficient way to achieve generalization.

One problem is that instruction fine-tuning relies on supervised learning, which models task generalization based on explicit instruction-response mappings. However, such an approach does not capture subtle or complex human preferences (e.g., tone, style, or subjective quality) because these are hard to encode as explicit instruction-response data. Moreover, the generalization performance is bounded by the diversity and quality of the instruction-response dataset. Given these limitations, we would instead like to employ preference models as an additional fine-tuning step following instruction fine-tuning, so the LLMs can generalize further (see Section \ref{sec:human-preference-alignment}).

Another view is that some instruction-response mappings may already be learned during pre-training. However, since we often do not know exactly what data is used in the pre-training, it is hard to judge whether such mappings still need to be learned during fine-tuning. A related question is whether out-of-distribution generalization is primarily achieved during pre-training or fine-tuning. While directly answering this question is beyond the scope of this chapter, it has been shown that pre-training on large and diverse datasets is effective in improving out-of-distribution performance \cite{hendrycks-etal:2020pretrained,radford-etal:2021learning,gunasekar-etal:2023textbooks}. This raises an interesting problem: if an LLM has been well pre-trained at scale, fine-tuning may not be as essential for out-of-distribution generalization, since the model may have already encountered sufficient distributional variation. This prompts researchers to fine-tune LLMs with modest effort or to explore new methods to achieve instruction-following. As discussed in the previous sub-section, for example, instruction following can be achieved by fine-tuning on a small number of carefully selected instruction-response pairs \cite{zhou-etal:2023lima}, or even by using methods that are not explicitly designed to do so \cite{kung-and-peng:2023models}.

The above discussion provides two different strategies: one requires scaling up fine-tuning datasets for larger diversity, the other requires small but necessary fine-tuning datasets for efficient LLM adaptation. However, in practice, involving diverse instructions often helps. In many cases, we need to adapt our LLM for specific purposes. But the LLM, which has possibly  encoded broad instruction-following mappings during pre-training, might tend to behave as a general-purpose instruction executor even with modest fine-tuning. An interesting phenomenon is that when fine-tuning on math data, the resulting LLM might not specialize in math outputs. Instead, this model might respond normally to general instructions, for example, it could generate poetry if instructed to do so \cite{hewitt:2024instructionblog}. This is not a bad thing, but it shows that LLMs may not easily change their nature of following general instructions. In this case, additional adaptations with more diverse data may help adjust the way the LLM follows instructions, particularly for those tasks we aim to address.

\subsection{Weak-to-Strong Generalization}

\noindent So far we have explored a variety of instruction fine-tuning methods based on labeled data. One of the limitations of many such methods is that they require the data to be annotated by humans or generated by strong LLMs, both of which can provide accurate supervision signals. However, in many cases, the LLM we have in hand is already strong (or at least excels in specific aspects of problem solving), and thus it is not easy to find a superior model for supervision. Even for human experts, when the problem becomes complex, providing correct and detailed answers might be difficult, or sometimes infeasible. For example, when faced with an extremely long document, the experts would find it challenging to identify any inconsistencies, subtle biases, or missing key points without conducting an exhaustive and time-consuming review.

One may ask at this point: can we use weak LLMs to supervise strong LLMs? This seems to be a significant challenge, but it may reflect a future scenario where we need to supervise AI systems that are smarter than humans or existing AI systems \cite{burns-etal:2023weakblog}. The problem of using smaller, less complex models to improve the training of larger, more complex models is also called the \mindex{weak-to-strong generalization} problem. While there have not been mature approaches to weak-to-strong generalization, using smaller models to assist stronger models has indeed proven useful in several areas of LLMs.

For instruction fine-tuning, one of the simplest ways of applying weak LLMs is to use these models to generate synthetic fine-tuning data. Suppose we have a collection of inputs $X$, where each input includes an instruction and a user input if necessary. For each $\mathbf{x} \in X$, we use a weak LLM $\Pr^{w}(\cdot)$ to generate a prediction $\hat{\mathbf{y}} = \argmax_{\mathbf{y}} \Pr^{w}(\mathbf{y}|\mathbf{x})$. Then, the strong LLM $\Pr^{s}_{\theta}(\cdot)$ can be trained on these predictions (see Eq. (\ref{eq:instruction-fine-tuning})):
\begin{eqnarray}
\tilde{\theta} & = & \argmax_{\theta} \sum_{\mathbf{x} \in X} \log \mathrm{Pr}_{\theta}^{s}(\hat{\mathbf{y}}|\mathbf{x}),
\end{eqnarray}

\noindent where $\theta$ represents the model parameters.

The above form transforms the fine-tuning problem into a knowledge distillation problem, in other words, we distill knowledge from the weak model to the strong model. Consequently, we can employ various knowledge distillation methods to achieve this goal. However, explaining weak-to-strong fine-tuning from the perspective of knowledge distillation is not straightforward. A major concern is that the strong model may merely imitate or overfit the errors of the weak model and fail to generalize. For example, the fine-tuned strong model still cannot solve difficult problems that the weak model cannot accurately predict. Fortunately, preliminary experiments in this line of research have shown positive and promising results. For example, \citet{burns-etal:2023weak} found that fine-tuning GPT-4-level models with GPT-2-level supervision could improve generalization across several NLP tasks. To measure how the weak model improves the generalization of the strong model, we define the following terms:

\begin{itemize}
\item \vspace{0.5em} \mindex{Weak Performance} ($\mathrm{P}_{\mathrm{weak}}$). This is the test-set performance of the weak model, which can be regarded as the baseline performance.
\item \vspace{0.3em} \mindex{Weak-to-strong Performance} ($\mathrm{P}_{\mathrm{weak} \to \mathrm{strong}}$). This is the test-set performance of the strong model that is fine-tuned with the weak model.
\item \vspace{0.3em} \mindex{Strong Ceiling Performance} ($\mathrm{P}_{\mathrm{ceiling}}$). This is the test-set performance of the strong model that is fine-tuned with ground truth data. For example, we fine-tune the strong model with human-annotated data and take the resulting model as a ceiling.
\end{itemize}
\vspace{0.5em}

Then, the \mindex{performance gap recovered} (\mindex{PGR}) can be defined as
\begin{eqnarray}
\mathrm{PGR} & = & \max \Big\{0, \frac{\mathrm{P}_{\mathrm{weak} \to \mathrm{strong}} - \mathrm{P}_{\mathrm{weak}}}{\mathrm{P}_{\mathrm{ceiling}} - \mathrm{P}_{\mathrm{weak}}} \Big\}.
\end{eqnarray}

This metric measures how much of the gap between the ceiling performance and the weak-model performance can be recovered by the weak-to-strong model. A PGR of 1 indicates that the weak-to-strong fine-tuning can completely close the performance gap, whereas a PGR of 0 indicates no improvement. In \citet{burns-etal:2023weak}'s work, it is shown that $\mathrm{PGR}$ can be around $0.8$ on 22 NLP classification tasks. It should be noted that, while the potential of weak-to-strong fine-tuning is promising, achieving substantial weak-to-strong generalization remains a challenging goal that needs further investigation \cite{aschenbrenner:2024situational}.

Fine-tuning LLMs with weak supervision is just one choice for using small models to improve large models. Although this section primarily focuses on fine-tuning LLMs, we also mention other methods here to give a more complete discussion (see Figure \ref{fig:small-to-large-improvement} for illustrations of these methods).

\begin{figure}[!t]
\centering
\input{./Chapter10/Figures/figure-small-models-for-llms}
\caption{Illustrations of using small models to improve large models in LLMs. One approach involves using smaller models for the fine-tuning or pre-training of larger models. This includes generating synthetic data (a), incorporating auxiliary loss (b), and selecting appropriate data (c). Another approach involves combining small models and large models. This includes learning a strong model by aggregating multiple small models (d), and cascading small models with large models (e).}
\label{fig:small-to-large-improvement}
\end{figure}

\begin{itemize}
\item \vspace{0.5em} Instead of using small models to generate synthetic data, it is also straightforward to incorporate knowledge distillation loss based on these models. For example, a simple loss function that measures the difference between the small and large models can be defined as:
    \begin{eqnarray}
    \mathcal{L}_{\mathrm{kd}} & = & D_{\mathrm{KL}}(\mathrm{Pr}^{w}(\cdot|\mathbf{x})\ \|\ \mathrm{Pr}^{s}_{\theta}(\cdot|\mathbf{x})),
    \end{eqnarray}
    \noindent which encourages the strong model distribution to align with the weak model distribution. Then, we can add this loss to the original loss of language modeling, and obtain the following training objective
    \begin{eqnarray}
    \tilde{\theta} & = & \argmax_{\theta} \sum_{(\mathbf{x},\mathbf{y}) \in \mathcal{D}} \log \mathrm{Pr}_{\theta}^{s}(\mathbf{y}|\mathbf{x}) - \lambda \cdot \mathcal{L}_{\mathrm{kd}},
    \end{eqnarray}
    \noindent where $\mathcal{D}$ is the set of input and output pairs, and $\lambda$ is the interpolation coefficient. This method can be employed in either the pre-training or fine-tuning phase. We can adjust $\lambda$ to control how much the small model influences the training. For example, we can gradually decrease $\lambda$ to make the training rely more on the original language modeling loss as the large model becomes more capable.
\item \vspace{0.3em} Another approach to involving small models in LLM pre-training and fine-tuning is to use them to do data selection or filtering. Given a sequence, we can compute the likelihood or cross-entropy using a small model. These quantities can then be used as criteria for selecting or filtering data. For example, sequences with low likelihood or high cross-entropy might be excluded from the training set, as they are less aligned with the small model's learned distribution. Conversely, sequences with high likelihood or low cross-entropy can be prioritized, ensuring that the training focuses on more relevant or high-quality data.
\item \vspace{0.3em} Ensemble learning is a simple and effective way to build a strong model by combining multiple weak models. Applying this technique to LLMs is straightforward. We can aggregate distributions predicted by multiple small models or specialized submodels, and derive the final prediction from the aggregated results. This aggregation can be done using methods such as majority voting, weighted averaging, or stacking.
\item \vspace{0.3em} Small models can also be employed at inference time to improve overall efficiency. Suppose we have a large model that is slow but more accurate, and a small model that is fast but less accurate. In model cascading, the small model first processes the input data, quickly generating preliminary results. If these results meet certain pre-defined criteria, they can be directly used. However, if the initial results are not sufficiently good, the input is then passed to the larger, more accurate model to produce a better result. This approach significantly reduces computational costs and latency, as the small model can effectively handle many inputs without access to the large model.
\end{itemize}
\vspace{0.5em}

%----------------------------------------------------------------------------------------
% NEW SECTION
%----------------------------------------------------------------------------------------

\section{Human Preference Alignment: RLHF}
\label{sec:human-preference-alignment}

\noindent So far in this chapter, we have focused on fine-tuning LLMs using labeled input-output pairs. This approach allows us to adapt LLMs for instruction-following via supervised learning. In many applications, however, LLMs are required not only to follow instructions but also to act in ways that are more aligned with human values and preferences. Consider a scenario where a user asks an LLM how to hack into a computer system. If the LLM is not appropriately aligned, it may respond by providing details on how to perform this illegal activity. Instead, a more desirable response might be to advise the user against engaging in illegal activities and offer a general overview of the consequences of such actions. The difficulty in achieving this is that the ethical nuances and contextual considerations required for an LLM to respond appropriately in such scenarios are often difficult to encode explicitly in fine-tuning data. What is even more challenging is that humans themselves often cannot precisely express their own preferences.

In this section, we discuss an alternative LLM fine-tuning method, called reinforcement learning from human feedback or RLHF for short \cite{christiano-etal:2017deep,stiennon-etal:2020learning}. The basic idea behind RLHF is that LLMs can learn from comparisons of model outputs using reward models (see Figure \ref{fig:sft-vs-rlhf}). To do this, we can recruit human experts to indicate which output they prefer from a pair of model outputs. This preference data is used to train a reward model that can predict the perceived quality of LLM outputs. Once trained, the reward model provides feedback by assigning scores to new outputs that the LLM generates in response to the inputs. The LLM uses these scores to update its parameters through reinforcement learning algorithms. In the rest of this section, we will first introduce the basic concepts of reinforcement learning to facilitate the discussion, and then discuss methods for training reward models and aligning LLMs with these models.

\begin{figure}[!t]
\centering
\input{./Chapter10/Figures/figure-sft-vs-rlhf}
\caption{Supervised fine-tuning vs. reinforcement learning from human feedback. In supervised fine-tuning, we optimize the LLM by maximizing the probability of the prediction given the input. In reinforcement learning from human feedback, we first train a reward model on human preference data (on each pair of predictions, evaluators are asked to choose which one they prefer). Then, we use this reward model to supervise the LLM during fine-tuning.}
\label{fig:sft-vs-rlhf}
\end{figure}

\subsection{Basics of Reinforcement Learning}
\label{sec:rl-basics}

\noindent We begin by looking at some basic concepts of reinforcement learning. Note that the notation used here slightly differs from that used in the previous sections and chapters because we want to make our description more consistent with that in the reinforcement learning literature. Nevertheless, we will show how this notation corresponds to the language modeling notation. The reader who is already familiar with reinforcement learning techniques may skip or skim this subsection.

A general reinforcement learning framework describes how an agent interacts with a dynamic environment. This interaction is modeled as a sequence of actions taken by the agent in response to the state of the environment. At each time step, the agent observes the current state, chooses an action based on its policy, performs the action, and then receives feedback from the environment in the form of a reward and a new state. This observation-action-reward process is repeated until the agent achieves its goal.

A reinforcement learning system involves several components:

\begin{itemize}
\item \vspace{0.5em} \textbf{Agent}. This is the learner or decision-maker in reinforcement learning. In the context of LLMs, it can be seen as the LLM itself.
\item \vspace{0.3em} \textbf{Environment}. This includes everything external to the agent with which the agent interacts. But the environment in LLMs is less about a physical or virtual space and more about the framework within which the agent (e.g., an LLM) receives feedback and learns.
\item \vspace{0.3em} \textbf{State} ($s$). A state represents the current situation of the environment. Given a sequence of tokens for language modeling, a state at a time step can be viewed as the tokens we observed so far, that is, the context tokens we take to predict the next token. For example, we can define $(\mathbf{x}, \mathbf{y}_{< t})$ as the state when predicting the next token at time step $t$.
\item \vspace{0.3em} \textbf{Action} ($a$). Actions represent possible decisions the agent can make. In the context of LLMs, actions correspond to candidate tokens in the vocabulary.
\item \vspace{0.3em} \textbf{Reward} ($R$). The reward is the feedback from the environment that evaluates the success of an action. For example, $r(s,a,s')$ denotes the reward the agent receives for taking the action $a$ in state $s$ and moving to the next state $s'$. If the state-action sequence is given, we can denote the reward at the time step $t$ as $r_t = r(s_t,a_t,s_{t+1})$. Also note that if the decision-making process is deterministic, we can omit $s_{t+1}$ because it can be determined by $s_t$ and $a_t$. In such cases, we can use $r(s_t,a_t)$ as shorthand for $r(s_t,a_t,s_{t+1})$.
\item \vspace{0.3em} \textbf{Policy} ($\pi$). For an LLM, a policy is defined as the probability distribution over the tokens that the LLM predicts, given the preceding context tokens. Formally, this can be expressed as
    \begin{eqnarray}
    \pi(a|s) = \Pr(y_t|\mathbf{x}, \mathbf{y}_{< t}),
    \end{eqnarray}
    where $a$ corresponds to the token $y_t$, and $s$ corresponds to the context $(\mathbf{x}, \mathbf{y}_{< t})$. Figure \ref{fig:llm-as-policy-in-rl} illustrates how an LLM can be treated as a policy in the reinforcement learning framework.

\begin{figure}[!t]
\centering
\input{./Chapter10/Figures/figure-llm-as-policy}
\caption{LLM as policy in reinforcement learning. At each step $t$, the LLM predicts a token $y_t$ given the model input $\mathbf{x}$ and the previously generated tokens $\mathbf{y}_{<t}$. This process can be framed as a reinforcement learning problem, where $y_t$ serves as the action, $(\mathbf{x},\mathbf{y}_{<t})$ as the state, and the predicted distribution $\Pr(y_t|\mathbf{x},\mathbf{y}_{<t})$ as the policy. The reward model evaluates the generated response and assigns reward scores, which are then used to estimate value functions such as $V(s_t)$ and $Q(s_t,a_t)$ for policy optimization. In practice, reward models in RLHF are typically defined over complete responses, although the optimization objective can still be decomposed token-wise.}
\label{fig:llm-as-policy-in-rl}
\end{figure}

\item \vspace{0.3em} \textbf{Value Function} ($V$ and $Q$). A \mindex{state-value function} (or value function, for short) assesses the expected discounted return (i.e., accumulated rewards) for an agent starting from a particular state $s$ and following a specific policy $\pi$. It is defined as:
    \begin{eqnarray}
    V(s) & = & \mathbb{E} \Big[ r(s_0,a_0,s_1) + \gamma r(s_1,a_1,s_2) + \gamma^2 r(s_2,a_2,s_3) + \cdots \ \big | \ s_0 = s, \pi \Big] \nonumber \\
         & = & \mathbb{E} \Big[ r_0 + \gamma r_1 + \gamma^2 r_2 + \cdots \ \big | \ s_0 = s, \pi \Big] \nonumber \\
         & = & \mathbb{E} \Big[ \sum_{t=0}^{\infty} \gamma^{t} r_t \ \big | \ s_0 = s, \pi \Big], \label{eq:general-value-function}
    \end{eqnarray}
    \noindent where $\gamma \in [0,1]$ is the discount factor that adjusts the importance of future rewards, $s_0 = s$ indicates that the agent starts with the state $s$, and the expectation $\mathbb{E}$ is performed over all possible trajectories (i.e., state-action sequences). Similarly, an \mindex{action-value function} (or \mindex{Q-value function}) measures the expected return starting from a state $s$ taking an action $a$ and thereafter following a policy $\pi$, given by
    \begin{eqnarray}
    Q(s,a) & = & \mathbb{E} \Big[ \sum_{t=0}^{\infty} \gamma^{t} r_t \ \big | \ s_0 = s, a_0 = a, \pi \Big],
    \end{eqnarray}
    \noindent where $a_0 = a$ indicates that the action taken at the initial state is $a$.
\end{itemize}
\vspace{0.5em}

The goal of reinforcement learning is to learn a policy that maximizes the \mindex{cumulative reward} (or \mindex{return}) the agent receives over the long run. For simplicity, in the following derivations we omit the discount factor (i.e., $\gamma = 1$), unless otherwise specified.  Given a state-action sequence $\tau = \{(s_1,a_1),...,(s_T,a_T)\}$\footnote{We assume the state-action sequence begins with $s_1$ and $a_1$, rather than $s_0$ and $a_0$, to align with the notation commonly used in this chapter, where the prediction $\mathbf{y}$ typically starts from $y_1$. Of course, it is also common to denote a state-action sequence as $\{(s_0,a_0),...,(s_T,a_T)\}$ or $\{(s_0,a_0),...,(s_{T-1},a_{T-1})\}$ in the literature. But this variation in notation does not affect the discussion of the models presented here.}, the cumulative reward  over this sequence can be written as
\begin{eqnarray}
R(\tau) & = & \sum_{t=1}^{T} r_t. \label{eq:rl-sequence-reward}
\end{eqnarray}

The expectation of this cumulative reward over a space of state-action sequences is given in the form
\begin{eqnarray}
J(\theta) & = & \mathop{\mathbb{E}}_{\tau \sim \mathcal{T}} \Big[ R(\tau) \ \big | \pi_{\theta} \Big] \nonumber \\
          & = & \sum_{\tau \in \mathcal{T}} \mathrm{Pr}_{\theta}(\tau) R(\tau) \nonumber \\
          & = & \sum_{\tau \in \mathcal{T}} \mathrm{Pr}_{\theta}(\tau) \sum_{t=1}^{T} r_t, \label{eq:performance-function-basic}
\end{eqnarray}

\noindent where $\tau \sim \mathcal{T}$ indicates that $\tau$ is drawn from the state-action sequence space $\mathcal{T}$, and the subscript $\theta$ indicates the parameters of the policy. $J(\theta)$ is also called the \mindex{performance function}.

Then the training objective is to maximize $J(\theta)$:

\begin{eqnarray}
\tilde{\theta} & = & \argmax_{\theta} J(\theta). \label{eq:rl-optimize-j-theta}
\end{eqnarray}

Now, we have a simple reinforcement learning approach: 1) we sample a number of state-action sequences; then, 2) we evaluate each sequence using the performance function; then, 3) we update the model to maximize this performance function. If we take Eq. (\ref{eq:performance-function-basic}) and use gradient descent to optimize the policy, this constitutes a policy gradient method \cite{williams:1992simple}.

Note that in many NLP problems, such as machine translation, rewards are typically sparse. For instance, a reward is only received at the end of a complete sentence. This means that $r_t = 0$ for all $t < T$, and $r_t$ is non-zero only when $t = T$. Ideally, one might prefer feedback to be immediate and frequent (dense), and thus the training of the policy can be easier and more efficient. While several methods have been proposed to address sparse rewards, such as reward shaping, we will continue in our discussion to assume a sparse reward setup, where the reward is available only upon completing the prediction.

The model described in Eqs. (\ref{eq:rl-sequence-reward}-\ref{eq:rl-optimize-j-theta}) establishes a basic form of reinforcement learning, and many variants and improvements of this model have been developed. Before showing those more sophisticated models, let us take a moment to interpret the objective function $J(\theta)$ from the perspective of policy gradient. In gradient descent, we need to compute the gradient of $J(\theta)$ with respect to $\theta$:
\begin{eqnarray}
\frac{\partial J(\theta)}{\partial \theta} & = & \frac{\partial  \sum_{\tau \in \mathcal{T}} \mathrm{Pr}_{\theta}(\tau) R(\tau)}{\partial \theta} \nonumber \\
& = & \sum_{\tau \in \mathcal{T}} \frac{\partial  \mathrm{Pr}_{\theta}(\tau) }{\partial \theta}  R(\tau) \nonumber \\
& = & \sum_{\tau \in \mathcal{T}} \mathrm{Pr}_{\theta}(\tau) \frac{\partial  \mathrm{Pr}_{\theta}(\tau)/\partial \theta}{\mathrm{Pr}_{\theta}(\tau) } R(\tau) \nonumber \\
& = & \sum_{\tau \in \mathcal{T}} \mathrm{Pr}_{\theta}(\tau) \frac{\partial  \log \mathrm{Pr}_{\theta}(\tau) }{\partial \theta} R(\tau). \label{eq:rl-gradient-j-theta}
\end{eqnarray}

In practice, the expectation is often approximated using uniformly sampled trajectories from a batch (i.e., $\mathrm{Pr}_{\theta}(\tau) = 1/|\mathcal{T}|$).  In this case we can simplify Eq. (\ref{eq:rl-gradient-j-theta}) and need only consider the terms $\frac{\partial \log \mathrm{Pr}_{\theta}(\tau)}{\partial \theta}$ and $R(\tau)$:
\begin{eqnarray}
\frac{\partial J(\theta)}{\partial \theta} & = & \frac{1}{m} \sum_{\tau \in \mathcal{T}} \frac{\partial  \log \mathrm{Pr}_{\theta}(\tau)}{\partial \theta} R(\tau). \label{eq:rl-j-theta-gradient-simplified}
\end{eqnarray}

\noindent One advantage of this result is that $R(\tau)$ does not need to be differentiable, which means that we can use any type of reward function in reinforcement learning.

By treating the generation of the sequence $\tau$ as a Markov decision process, we can further derive $\frac{\partial \log \mathrm{Pr}_{\theta}(\tau)}{\partial \theta}$, and obtain:
\begin{eqnarray}
\frac{\partial \log \mathrm{Pr}_{\theta}(\tau)}{\partial \theta} & = & \frac{\partial}{\partial \theta} \log \prod_{t=1}^{T} \pi_{\theta}(a_t|s_t) \Pr(s_{t+1}|s_t,a_t) \nonumber \\
& = & \frac{\partial}{\partial \theta} \sum_{t=1}^{T} \underbrace{\log \pi_{\theta}(a_t|s_t)}_{\text{policy}} + \frac{\partial}{\partial \theta} \sum_{t=1}^{T} \underbrace{\log \Pr(s_{t+1}|s_t,a_t)}_{\text{dynamics}}, \label{eq:gradient-policy-and-dynamics}
\end{eqnarray}

\noindent where the gradient is decomposed into two parts: the policy gradient and the dynamics gradient. The policy component, $\log \pi_{\theta}(a_t|s_t)$, determines the log-probability of taking action $a_t$ given state $s_t$, and it is parameterized by $\theta$. The dynamics component, $\log \Pr(s_{t+1}|s_t,a_t)$, represents the log-probability of transitioning to state $s_{t+1}$ from state $s_t$ after taking action $a_t$. In typical reinforcement learning settings, the dynamics are not directly influenced by the policy parameters $\theta$, and thus, their derivatives are often zero. In this case, therefore, Eq. (\ref{eq:gradient-policy-and-dynamics}) can be simplified to:
\begin{eqnarray}
\frac{\partial \log \mathrm{Pr}_{\theta}(\tau)}{\partial \theta} & = & \frac{\partial}{\partial \theta} \sum_{t=1}^{T} \log \pi_{\theta}(a_t|s_t). \label{eq:rl-policy-gradient-product-of-pi}
\end{eqnarray}

\noindent In other words, we only concentrate on optimizing the policy without concerning ourselves with the underlying dynamics.

Substituting Eq. (\ref{eq:rl-policy-gradient-product-of-pi}) into Eq. (\ref{eq:rl-j-theta-gradient-simplified}), and expanding $R(\tau)$, we then obtain
\begin{eqnarray}
\frac{\partial J(\theta)}{\partial \theta} & = & \frac{1}{|\mathcal{T}|} \sum_{\tau \in \mathcal{T}} \frac{\partial}{\partial \theta} \Big( \sum_{t=1}^{T} \log \pi_{\theta}(a_t|s_t) \sum_{t=1}^{T} r_t \Big).
\end{eqnarray}

While this policy gradient approach is straightforward, it suffers from the problem that the variance of the estimated gradients can be very high, making the learning process noisy and inefficient. One reason for this high variance problem is that rewards can vary greatly across different steps or scenarios. Imagine that in a sequence of action decisions, the reward model tends to assign small rewards to good actions (e.g., $R_t = 2$) and large penalties to poor actions (e.g., $R_t = -50$). Such varying reward scales for good and poor actions can result in a very low total reward for the entire sequence, even if it includes good actions.

One simple method for reducing the variance of the gradient is to set a baseline $b$ and subtract it from $\sum_{t=1}^{T} r_t$, resulting in $\sum_{t=1}^{T} r_t - b$.\footnote{In fact, the use of a baseline $b$ does not change the variance of the total rewards $\sum_{t=1}^{T} r_t$. However, it is important to note that while introducing a baseline does not alter the overall variance of the rewards, it helps reduce the variance of the gradient estimates. In general, the operation $\sum_{t=1}^{T} r_t - b$ centers the rewards around zero (e.g., $b$ is defined as the expected value of $\sum_{t=1}^{T} r_t$), which can lead to reduced variance in the product $\sum_{t=1}^{T} \log \pi_{\theta}(a_t|s_t) (\sum_{t=1}^{T} r_t - b)$.} Here, the baseline can be interpreted as a reference point. By centering the rewards around this baseline, we remove systematic biases in the reward signal, making the updates more stable and less sensitive to extreme fluctuations in individual rewards.

This policy gradient model with a baseline can be given by
\begin{eqnarray}
\frac{\partial J(\theta)}{\partial \theta} & = & \frac{1}{|\mathcal{T}|} \sum_{\tau \in \mathcal{T}} \frac{\partial}{\partial \theta} \Big( \sum_{t=1}^{T} \log \pi_{\theta}(a_t|s_t) \Big) \Big(\sum_{t=1}^{T} r_t - b \Big) \nonumber \\
& = & \frac{1}{|\mathcal{T}|} \sum_{\tau \in \mathcal{T}} \frac{\partial}{\partial \theta} \Big[ \sum_{t=1}^{T} \log \pi_{\theta}(a_t|s_t) \Big(\sum_{k=1}^{T} r_k - b \Big) \Big] \nonumber \\
& = & \frac{1}{|\mathcal{T}|} \sum_{\tau \in \mathcal{T}} \frac{\partial}{\partial \theta} \Big[ \sum_{t=1}^{T} \log \pi_{\theta}(a_t|s_t) \Big(\sum_{k=1}^{t-1} r_k + \sum_{k=t}^{T} r_k - b \Big) \Big]. \label{eq:policy-gradient-with-baseline}
\end{eqnarray}

\noindent Here we write $\sum_{k=1}^{T} r_k$ as the sum of two terms $\sum_{k=1}^{t-1} r_k$ and $\sum_{k=t}^{T} r_k$ to distinguish between the rewards accrued before and after the action at time step $t$. Note that rewards received before time step $t$ do not depend on the action taken at time step $t$. Therefore, the term $\sum_{k=1}^{t-1} r_k$ does not contribute to the gradient and can be omitted. This leads to a simplified version of Eq. (\ref{eq:policy-gradient-with-baseline})
\begin{eqnarray}
\frac{\partial J(\theta)}{\partial \theta} & = & \frac{1}{|\mathcal{T}|} \sum_{\tau \in \mathcal{T}} \frac{\partial}{\partial \theta} \Big[ \sum_{t=1}^{T} \log \pi_{\theta}(a_t|s_t) \Big(\sum_{k=t}^{T} r_k - b \Big) \Big].
\end{eqnarray}

\noindent Also note that removing $\sum_{k=1}^{t-1} r_k$ can further reduce the variance of the gradient.

There are many ways to define the baseline $b$. Here we consider the value function of the state $s_t$, that is, the estimated value of being in state $s_t$: $V(s_t) = \mathbb{E}(r_t + r_{t+1} + \cdots + r_{T})$. Hence we have
\begin{eqnarray}
A(s_t,a_t) & = & \sum_{k=t}^{T} r_k - b \nonumber \\
           & = & \sum_{k=t}^{T} r_k - V(s_t),
\end{eqnarray}

\noindent where $\sum_{k=t}^{T} r_k $ represents the actual return received, and $V(s_t)$ represents the expected return. $A(s_t,a_t)$ (or $A_t$ for short) is called the \mindex{advantage} at time step $t$, which quantifies the relative benefit of the action $a_t$ compared to the expected value of following the policy from the state $s_t$ onward.

By using the advantage function $A(s_t,a_t)$, the gradient of $J(\theta)$ can be written in the form
\begin{eqnarray}
\frac{\partial J(\theta)}{\partial \theta} & = & \frac{1}{|\mathcal{T}|} \sum_{\tau \in \mathcal{T}} \frac{\partial}{\partial \theta} \Big( \sum_{t=1}^{T} \log \pi_{\theta}(a_t|s_t) A(s_t,a_t) \Big). \label{eq:a2c-policy-gradient}
\end{eqnarray}

This objective forms the basis of \mindex{advantage actor-critic} (\mindex{A2C}) methods in reinforcement learning \cite{mnih-etal:2016asynchronous}. In this method, the actor aims at learning a policy. It updates the policy parameters using Eq. (\ref{eq:a2c-policy-gradient}) to help focus more on actions that are likely to improve performance. The critic, on the other hand, updates its estimation of the value function, which is used to calculate the advantage function $A(s_t,a_t)$, thus serving as the evaluator of the policy being learned by the actor.

In the A2C method, $A(s_t,a_t)$ is typically expressed as the difference of the action-value function $Q(s_t,a_t)$ and the state-value function $V(s_t)$:
\begin{eqnarray}
A(s_t,a_t) & = & Q(s_t,a_t) - V(s_t). \label{eq:advantage-q-v}
\end{eqnarray}

\noindent At first glance, this model may seem challenging to develop because it requires two separate sub-models to calculate $Q(s_t,a_t)$ and $V(s_t)$, respectively. Fortunately, considering that $Q(s_t,a_t)$ can be defined as the return $r_t + V(s_{t+1})$, we can rewrite Eq. (\ref{eq:advantage-q-v}) as
\begin{eqnarray}
A(s_t,a_t) & = & r_t + V(s_{t+1}) - V(s_t),
\end{eqnarray}

\noindent or alternatively, introduce the discount factor $\gamma$ to obtain a more general form
\begin{eqnarray}
A(s_t,a_t) & = & r_t + \gamma V(s_{t+1}) - V(s_t).
\end{eqnarray}

\noindent $A(s_t,a_t) = r_t + \gamma V(s_{t+1}) - V(s_t)$ is also called the \mindex{temporal difference} (\mindex{TD}) error. What we need is to train a critic network for the value function $V(s_t)$, and then use it to compute the advantage function\footnote{The training loss for the value network (or critic network) in A2C is generally formulated as the mean squared error between the computed return $r_t + \gamma V(s_{t+1})$ and the predicted state value $V(s_t)$. Suppose that the value network is parameterized by $\omega$. The loss function is given by
\begin{eqnarray}
\mathcal{L}_v(\omega) & = & \frac{1}{M} \sum \big(r_t + \gamma V_{\omega}(s_{t+1}) - V_{\omega}(s_t) \big)^2,
\end{eqnarray}

\noindent where $M$ is the number of training samples, for example, for a sequence of $T$ tokens, we can set $M=T$.
}.

Up to this point, we have spent considerable space discussing the basics of reinforcement learning, especially on how to derive the optimization objective for the A2C method. However, reinforcement learning is a vast field, and many technical details cannot be covered here. The interested reader can refer to reinforcement learning books for more details \cite{Sutton-and-Barto:2018RL,szepesvari:2010algorithms}. Nevertheless, we now have the necessary knowledge to further discuss RLHF. In the subsequent subsections, we will return to the discussion on LLM alignment, demonstrating how to use the A2C method for aligning with human preferences.

\subsection{Training Reward Models}
\label{sec:training-reward-models}

\noindent We have shown that reward models play a very important role in the general reinforcement learning framework and form the basis for computing value functions. We now consider the problem of training these reward models.

In RLHF, a reward model is a neural network that maps a pair of input and output token sequences to a scalar reward score. Given an input $\mathbf{x}$ and an output $\mathbf{y}$, the reward can be expressed as
\begin{eqnarray}
r & = & \mathrm{Reward}(\mathbf{x},\mathbf{y}),
\end{eqnarray}

\noindent where $\mathrm{Reward}(\cdot)$ is the reward model. $r$ can be interpreted as a measure of how well the output $\mathbf{y}$ aligns with the desired behavior given the input $\mathbf{x}$. As discussed in the previous subsection, both $\mathbf{x}$ and $\mathbf{y}$ are assumed to be complete token sequences. As a result, the reward model evaluates the quality of a fully generated output conditioned on the input, rather than assigning rewards to intermediate generation steps. For example, during sequence generation, the reward is typically provided only at the final position of the output sequence $\mathbf{y}=y_1...y_n$. To keep the notation uncluttered, we will use $r(\mathbf{x},\mathbf{y})$ to denote the reward model $\mathrm{Reward}(\mathbf{x},\mathbf{y})$ from here on.

There are many ways to implement the reward model. One simple approach is to build the reward model based on a pre-trained LLM. More specifically, we can concatenate $\mathbf{x}$ and $\mathbf{y}$ to form a single token sequence $\mathrm{seq}_{\mathbf{x},\mathbf{y}} = [\mathbf{x}, \mathbf{y}]$. We run a pre-trained LLM on this sequence, as usual, and at each position, we obtain a representation from the last Transformer layer. Then, we take the representation at the last position (denoted by $\mathbf{h}_{\mathrm{last}}$) and map it to a scalar via a linear transformation:
\begin{eqnarray}
r(\mathbf{x},\mathbf{y}) & = & \mathbf{h}_{\mathrm{last}} \mathbf{W}_{r},
\end{eqnarray}

\noindent where $\mathbf{h}_{\mathrm{last}}$ is a $d$-dimensional vector, and $\mathbf{W}_{r}$ is a $d \times 1$ linear mapping matrix. This architecture of the reward model is illustrated in Figure \ref{fig:reward-model-architecture}.

\begin{figure}[!t]
\centering
\input{./Chapter10/Figures/figure-reward-model-architecture}
\caption{Architecture of a Transformer-based reward model. The main component of this model is still an LLM. We use the Transformer decoder as the sequence representation model. We take the representation at the last position of the decoder as the representation of the entire sequence $[\mathbf{x},\mathbf{y}]$. We then map this representation to a scalar through a linear transformation, which serves as the reward score for $[\mathbf{x},\mathbf{y}]$.}
\label{fig:reward-model-architecture}
\end{figure}

To train the reward model, the first step is to collect human feedback on a set of generated outputs. Given an input $\mathbf{x}$, we use the LLM to produce multiple candidate outputs $\{\mathbf{y}_1,...,\mathbf{y}_N\}$. Human feedback can be obtained in several ways:

\begin{itemize}
\item \vspace{0.5cm} \textbf{Pairwise Comparison} (\textbf{Pairwise Ranking}). Given two different outputs, human experts select which one is better.
\item \vspace{0.3cm} \textbf{Rating}. Human experts provide a score or rating to each output. This score is often a continuous or discrete numerical value, such as a score on a scale (e.g., 1-5 stars, or 1-10 points). In some cases, the rating might be binary, indicating a ``yes/no'' or ``positive/negative'' preference.
\item \vspace{0.3em} \textbf{Listwise Ranking}. Human experts are asked to rank or order the given set of possible outputs.
\end{itemize}
\vspace{0.5em}

Here we consider pairwise comparison feedback as it is one of the simplest and most common forms of human feedback used in RLHF.  In this setting, for each comparison, two outputs $(\mathbf{y}_a,\mathbf{y}_b)$ are randomly drawn from the candidate pool $\{\mathbf{y}_1,...,\mathbf{y}_N\}$. Human experts are then presented with these pairs and asked to decide which output they prefer based on criteria such as clarity, relevance, and accuracy. The human feedback can be encoded as a binary label. For example, $\mathbf{y}_a \succ \mathbf{y}_b$ indicates a preference for $\mathbf{y}_a$, and $\mathbf{y}_b \succ \mathbf{y}_a$ indicates a preference for $\mathbf{y}_b$.

One simple and widely used model for describing such pairwise comparisons is the \mindex{Bradley-Terry model} \cite{bradley-and-terry:rank}. It is a probabilistic model that estimates the probability that one item is preferred over another. Adapting this model to the notation used here, we can write the probability that $\mathbf{y}_a$ is preferred over $\mathbf{y}_b$ in the form
\begin{eqnarray}
\Pr(\mathbf{y}_a \succ \mathbf{y}_b | \mathbf{x}) & = & \frac{e^{r(\mathbf{x},\mathbf{y}_a)}}{e^{r(\mathbf{x},\mathbf{y}_a)} + e^{r(\mathbf{x},\mathbf{y}_b)}} \nonumber \\
                                                  & = & \frac{e^{r(\mathbf{x},\mathbf{y}_a)-r(\mathbf{x},\mathbf{y}_b)}}{e^{r(\mathbf{x},\mathbf{y}_a)-r(\mathbf{x},\mathbf{y}_b)}+1} \nonumber \\
                                                  & = & \mathrm{Sigmoid}(r(\mathbf{x},\mathbf{y}_a)-r(\mathbf{x},\mathbf{y}_b)).
\end{eqnarray}

When training the reward model, our objective is to maximize this preference probability. A loss function based on the Bradley-Terry model is given by
\begin{eqnarray}
\mathcal{L}_r(\phi) & = & -\mathbb{E}_{(\mathbf{x},\mathbf{y}_a,\mathbf{y}_b) \sim \mathcal{D}_r} \big[ \log \mathrm{Pr}_{\phi}(\mathbf{y}_a \succ \mathbf{y}_b | \mathbf{x}) \big], \label{eq:pairwise-reward-loss-expectation}
\end{eqnarray}

\noindent where $(\mathbf{x},\mathbf{y}_a,\mathbf{y}_b)$ is drawn from a human-annotated dataset $\mathcal{D}_r$ consisting of preference pairs of outputs and their corresponding inputs. $\phi$ represents the parameters of the reward model, which includes both the parameters of the Transformer decoder and the linear mapping matrix $\mathbf{W}_r$. In practice, assuming $(\mathbf{x},\mathbf{y}_a,\mathbf{y}_b)$ is uniformly sampled from $\mathcal{D}_r$, the expectation can be approximated by the empirical average
\begin{eqnarray}
\mathcal{L}_r(\phi) & = & -\frac{1}{|\mathcal{D}_r|} \sum_{(\mathbf{x},\mathbf{y}_a,\mathbf{y}_b) \in \mathcal{D}_r} \log \mathrm{Pr}_{\phi}(\mathbf{y}_a \succ \mathbf{y}_b | \mathbf{x}). \label{eq:pairwise-reward-loss-sum}
\end{eqnarray}

The goal of training the reward model is to find the optimal parameters $\hat{\phi}$ that minimize this loss function, given by
\begin{eqnarray}
\hat{\phi} & = & \argmin_{\phi} \mathcal{L}_r(\phi).
\end{eqnarray}

\noindent Since the reward model itself is also an LLM, we can simply use the Transformer training procedure to optimize the reward model. The difference from training a standard LLM is that we only need to replace the cross-entropy loss with the pairwise comparison loss as described in Eq. (\ref{eq:pairwise-reward-loss-sum}). After training the reward model, we can apply the trained reward model $r_{\hat{\phi}}(\cdot)$ to supervise the target LLM for alignment.

It is worth noting that although we train the reward model to perform pairwise ranking, we apply it to score each input-output pair independently during the alignment process. The pairwise ranking objective ensures that the reward model is sensitive to subtle differences between outputs, but we rely on the continuous scores produced by the reward model to guide the optimization of the LLM. An advantage of this approach is that we can choose from or combine various ranking loss functions, and still apply the resulting reward models in the same way as we have done in this subsection. This flexibility makes the reward modeling framework broadly applicable across different forms of human preference supervision.

\subsection{Training LLMs}
\label{sec:rlhf-training-llms}

\noindent Having obtained the reward model, we then train the policy (i.e., the LLM) via the A2C method. Recall from Section \ref{sec:rl-basics} that a state-action sequence or trajectory $\tau$ can be evaluated by the utility function
\begin{eqnarray}
U(\tau;\theta) & = &  \sum_{t=1}^{T} \log \pi_{\theta}(a_t|s_t) A(s_t,a_t), \label{eq:utility-basic-form}
\end{eqnarray}

\noindent where $A(s_t,a_t)$ is the advantage of taking the action $a_t$ given the state $s_t$. An estimate of $A(s_t,a_t)$ is defined as the TD error $ r_t + \gamma V(s_{t+1}) - V(s_t)$, where the value function $V(s_t)$ is trained with the reward model.

Given this utility function, the A2C-based loss function can be written in the form
\begin{eqnarray}
\mathcal{L}(\theta) & = & -\mathbb{E}_{\tau \sim \mathcal{T}} \big[U(\tau;\theta) \big] \nonumber \\
          & = & -\mathbb{E}_{\tau \sim \mathcal{T}} \big[ \sum_{t=1}^{T} \log \pi_{\theta}(a_t|s_t) A(s_t,a_t) \big],
\end{eqnarray}

\noindent where $\mathcal{T}$ is a space of state-action sequences. As usual, the goal of training the policy is to minimize this loss function
\begin{eqnarray}
\tilde{\theta} & = & \argmin_{\theta} \mathcal{L}(\theta).
\end{eqnarray}

If we map the problem back to the language modeling problem and adopt the notation from LLMs, the loss function can be written as:
\begin{eqnarray}
\mathcal{L}(\theta) & = & -\mathbb{E}_{(\mathbf{x},\mathbf{y}) \sim \mathcal{D}} \big[ U(\mathbf{x},\mathbf{y};\theta) \big],
\end{eqnarray}

\noindent where
\begin{eqnarray}
U(\mathbf{x},\mathbf{y};\theta) & = & \sum_{t=1}^{T} \log \pi_{\theta}(y_t|\mathbf{x},\mathbf{y}_{<t}) A(\mathbf{x},\mathbf{y}_{<t},y_t).
\end{eqnarray}

\noindent Here $\mathcal{D}$ is a set of input-output pairs, and $\pi_{\theta}(y_t|\mathbf{x},\mathbf{y}_{<t}) = \mathrm{Pr}_{\theta}(y_t|\mathbf{x},\mathbf{y}_{<t})$ is the LLM parameterized by $\theta$.

In general, we do not have a human annotated input-output dataset $\mathcal{D}$ in RLHF, but a dataset containing inputs only. The outputs, in this case, are typically the predictions made by the LLM. The loss function is then defined as
\begin{eqnarray}
\mathcal{L}(\theta) & = & -\mathbb{E}_{\mathbf{x} \sim \mathcal{D}} \mathbb{E}_{\mathbf{y} \sim \pi_{\theta}(\cdot|\mathbf{x})} \big[ U(\mathbf{x},\mathbf{y};\theta) \big],
\end{eqnarray}

\noindent where $\mathcal{D}$ denotes the input-only dataset, and $\mathbf{y} \sim \pi_{\theta}(\cdot|\mathbf{x})$ denotes that the output $\mathbf{y}$ is sampled by the policy $\pi_{\theta}(\cdot|\mathbf{x})$.

The above formulation provides a basic form of the A2C method for LLMs. Improved versions of this model are more commonly used in RLHF. In the following discussion, we will still use the reinforcement learning notation to simplify the presentation and will get back to the language modeling notation later.

One common improvement of policy gradient methods is to construct a surrogate objective using \mindex{importance sampling}. The resulting objective can be written as
\begin{eqnarray}
U(\tau;\theta) & = &  \sum_{t=1}^{T} \frac{\pi_{\theta}(a_t|s_t)}{\pi_{\theta_{\mathrm{ref}}}(a_t|s_t)} A(s_t,a_t).
\end{eqnarray}

\noindent Here we replace the log-probability $\log \pi_{\theta}(a_t|s_t)$ with the ratio $\frac{\pi_{\theta}(a_t|s_t)}{\pi_{\theta_{\mathrm{ref}}}(a_t|s_t)}$. $\theta_{\mathrm{ref}}$ denotes the parameters of the previous policy (such as an initial model from which we start the training). So $\frac{\pi_{\theta}(a_t|s_t)}{\pi_{\theta_{\mathrm{ref}}}(a_t|s_t)}$, also called the \mindex{ratio function}, can be interpreted as the probability ratio between the current policy $\pi_{\theta}$ and the reference policy $\pi_{\theta_{\mathrm{ref}}}$ (call it the reference policy). By using the ratio function we reweight the observed rewards based on the likelihood of the actions under the current policy versus the reference policy. When $\frac{\pi_{\theta}(a_t|s_t)}{\pi_{\theta_{\mathrm{ref}}}(a_t|s_t)} > 1$, the action $a_t$ is more favored by the current policy compared to the reference policy. By contrast, when $\frac{\pi_{\theta}(a_t|s_t)}{\pi_{\theta_{\mathrm{ref}}}(a_t|s_t)} < 1$, the action $a_t$ is less favored by the current policy\footnote{
Consider a more general case where we wish to evaluate the policy using its expected reward (also see Eq. (\ref{eq:performance-function-basic}))
\begin{eqnarray}
J(\theta) & = & \mathbb{E}_{\tau \sim \pi_{\theta}} \Big[ R(\tau) \Big].
\end{eqnarray}

\noindent Here $\tau \sim \pi_{\theta}$ means that the sequence $\tau$ is generated by the policy $\pi_{\theta}$. Alternatively, we can write $J(\theta)$ in another form
\begin{eqnarray}
J(\theta) & = & \mathbb{E}_{\tau \sim \pi_{\theta_{\mathrm{ref}}}} \Big[ \frac{\mathrm{Pr}_{\theta}(\tau)}{\mathrm{Pr}_{\theta_{\mathrm{ref}}}(\tau)} R(\tau) \Big]. \label{eq:importance-sampling-sequence-reward}
\end{eqnarray}

It is straightforward to see that the right-hand sides of these equations are essentially the same since $\mathbb{E}_{\tau \sim \pi_{\theta_{\mathrm{ref}}}} \Big[ \frac{\mathrm{Pr}_{\theta}(\tau)}{\mathrm{Pr}_{\theta_{\mathrm{ref}}}(\tau)} R(\tau) \Big] = \sum_{\tau} \mathrm{Pr}_{\theta_{\mathrm{ref}}}(\tau) \frac{\mathrm{Pr}_{\theta}(\tau)}{\mathrm{Pr}_{\theta_{\mathrm{ref}}}(\tau)} R(\tau) = \sum_{\tau} \mathrm{Pr}_{\theta}(\tau) R(\tau) = \mathbb{E}_{\tau \sim \pi_{\theta}} \Big[ R(\tau) \Big]$.

Note that this equivalence holds only when the expectation is performed over the entire sequence space. In practice, however, we often only sample a relatively small number of sequences using a policy in policy learning. As a result, the sampling method itself matters. Eq. (\ref{eq:importance-sampling-sequence-reward}) offers an interesting manner to separate the sampling and reward computation processes: we first use a baseline policy (with $\theta_{\mathrm{ref}}$) to sample a number of sequences, and then use the target policy (with $\theta$) to compute the expected reward. In this way, we separate the policy used for collecting the data, and the policy used for computing the gradient. This approach avoids the need to directly sample from the policy we are evaluating, which can be beneficial in cases where generating sequences from the target policy is expensive or difficult. In reinforcement learning, $\mathbb{E}_{\tau \sim \pi_{\theta_{\mathrm{ref}}}} \Big[ \frac{\mathrm{Pr}_{\theta}(\tau)}{\mathrm{Pr}_{\theta_{\mathrm{ref}}}(\tau)} R(\tau) \Big]$ is often called a \mindex{surrogate objective}.

Eq. (\ref{eq:importance-sampling-sequence-reward}) can also be interpreted from a policy gradient perspective. For $\mathbb{E}_{\tau \sim \pi_{\theta_{\mathrm{ref}}}} \Big[ \frac{\mathrm{Pr}_{\theta}(\tau)}{\mathrm{Pr}_{\theta_{\mathrm{ref}}}(\tau)} R(\tau) \Big]$, the gradient at $\theta=\theta_{\mathrm{ref}}$ is given by
\begin{eqnarray}
\frac{\partial}{\partial \theta} \mathbb{E}_{\tau \sim \pi_{\theta_{\mathrm{ref}}}} \Big[ \frac{\mathrm{Pr}_{\theta}(\tau)}{\mathrm{Pr}_{\theta_{\mathrm{ref}}}(\tau)} R(\tau) \Big] \Big |_{\theta = \theta_{\mathrm{ref}}} & = & \mathbb{E}_{\tau \sim \pi_{\theta_{\mathrm{ref}}}} \Big[ \frac{\partial \mathrm{Pr}_{\theta}(\tau) |_{\theta = \theta_{\mathrm{ref}}}}{\partial \theta} R(\tau) \Big].
\end{eqnarray}

The right-hand side is a standard form used in policy gradient methods, meaning that we compute the direction of the parameter update at the point $\theta=\theta_{\mathrm{ref}}$ on the optimization surface.
}.

A problem with the model presented in Eq. (\ref{eq:importance-sampling-sequence-reward}) (as well as in Eq. (\ref{eq:utility-basic-form})) is that the variance in the gradient estimates is often high, making the learning process unstable. To mitigate this issue, techniques such as clipping are often employed to bound the importance weights and prevent large updates. A clipped version of the utility function (also called the clipped surrogate objective function) is given by
\begin{eqnarray}
U_{\mathrm{clip}}(\tau;\theta) & = & \sum_{t=1}^{T} \mathrm{Clip}\left( \frac{\pi_{\theta}(a_t|s_t)}{\pi_{\theta_{\mathrm{ref}}}(a_t|s_t)}, A(s_t,a_t) \right),
\end{eqnarray}

\noindent where
\begin{eqnarray}
& & \mathrm{Clip}\left( \frac{\pi_{\theta}(a_t|s_t)}{\pi_{\theta_{\mathrm{ref}}}(a_t|s_t)}, A(s_t,a_t)\right) \nonumber \\
& = &  \min\left( \frac{\pi_{\theta}(a_t|s_t)}{\pi_{\theta_{\mathrm{ref}}}(a_t|s_t)} A(s_t,a_t),\mathrm{bound} \left(\frac{\pi_{\theta}(a_t|s_t)}{\pi_{\theta_{\mathrm{ref}}}(a_t|s_t)}, 1 - \epsilon, 1 + \epsilon \right) A(s_t,a_t) \right).
\end{eqnarray}

\noindent Here the function $\mathrm{bound}(\frac{\pi_{\theta}(a_t|s_t)}{\pi_{\theta_{\mathrm{ref}}}(a_t|s_t)}, 1 - \epsilon, 1 + \epsilon)$ constrains the ratio function to the range $[1-\epsilon, 1+\epsilon]$.

A further improvement to the above model is to consider \mindex{trust regions} in optimization \cite{schulman-etal:2015trust}. In reinforcement learning, a large update to the policy can lead to instability, where the agent may start performing worse after an update. A reasonable idea is to optimize the model in the trust region, which refers to a region around the current parameter estimate where the model is well-behaved. One approach to incorporating trust regions is to impose a constraint on the size of the policy update, ensuring that the current policy does not deviate significantly from the reference policy. In RLHF, PPO is often combined with an additional KL-style penalty that constrains the policy to remain close to the reference policy. This can be achieved by adding a penalty related to the discrepancy between the current and reference policies. A simple form of such a penalty is given by the difference in the log-probability of the sequence $\tau$ under the current policy versus the reference policy:
\begin{eqnarray}
\mathrm{Penalty} & = & \log \pi_{\theta}(\tau) - \log \pi_{\theta_{\mathrm{ref}}}(\tau).
\end{eqnarray}

\noindent In practice, this penalty can be approximated by considering only the policy probabilities and ignoring the dynamics. This gives
\begin{eqnarray}
\mathrm{Penalty} & = & \sum_{t=1}^{T} \log \pi_{\theta}(a_t|s_t) - \sum_{t=1}^{T} \log \pi_{\theta_{\mathrm{ref}}}(a_t|s_t).
\end{eqnarray}

\noindent By including this penalty in the optimization objective, we encourage the current policy to remain close to the reference policy, limiting very large updates that could destabilize learning.

We can incorporate this penalty into the clipped surrogate objective function, and obtain
\begin{eqnarray}
U_{\mathrm{ppo\text{-}clip}}(\tau;\theta) & = & U_{\mathrm{clip}}(\tau;\theta) - \beta \mathrm{Penalty},
\end{eqnarray}

\noindent where $\beta$ is the weight of the penalty. This training method is called \mindex{proximal policy optimization} (\mindex{PPO}), which is one of the most popular reinforcement learning methods used in LLMs and many other fields \cite{schulman-etal:2017proximal}.

Now we can write the objective of training LLMs in the form of PPO

\begin{eqnarray}
U_{\mathrm{ppo\text{-}clip}}(\mathbf{x},\mathbf{y};\theta) & = & U_{\mathrm{clip}}(\mathbf{x},\mathbf{y};\theta) - \beta \mathrm{Penalty} \label{eq:ppo-objective-llm},
\end{eqnarray}

\noindent where
\begin{eqnarray}
U_{\mathrm{clip}}(\mathbf{x},\mathbf{y};\theta) & = & \sum_{t=1}^{T} \mathrm{Clip}\left( \frac{\pi_{\theta}(y_t|\mathbf{x},\mathbf{y}_{<t})}{\pi_{\theta_{\mathrm{ref}}}(y_t|\mathbf{x},\mathbf{y}_{<t})}, A(\mathbf{x},\mathbf{y}_{<t},y_t)  \right), \\
\mathrm{Penalty} & = & \log \mathrm{Pr}_{\theta}(\mathbf{y}|\mathbf{x}) - \log \mathrm{Pr}_{\theta_{\mathrm{ref}}}(\mathbf{y}|\mathbf{x}) \nonumber \\
                 & = & \sum_{t=1}^{T} \log \mathrm{Pr}_{\theta}(y_t|\mathbf{x},\mathbf{y}_{<t}) - \sum_{t=1}^{T} \log \mathrm{Pr}_{\theta_{\mathrm{ref}}}(y_t|\mathbf{x},\mathbf{y}_{<t}).
\end{eqnarray}

Although the notation here appears a bit tedious, the idea of PPO is simple: we develop an objective by combining the clipped likelihood ratio of the target and reference policies with an advantage function, and then impose a penalty that ensures policy updates are not too large. The PPO-based RLHF is illustrated in Figure \ref{fig:ppo-rlhf}.

\begin{figure}[!t]
\centering
\input{./Chapter10/Figures/figure-rlhf-diagram}
\caption{Illustration of RLHF. The first step is to collect human preference data and train the reward model using this data. Once the reward model is optimized, along with the reference model, we proceed to train both the policy and the value function. At each prediction step, we compute the sum of the PPO-based loss and update the parameters of the policy. This requires access to the reward model, the reference model, and the value function at hand. At the same time, we update the parameters of the value function by minimizing the MSE loss.}
\label{fig:ppo-rlhf}
\end{figure}

To summarize, implementing RLHF requires building four models, all based on the Transformer decoder architecture.

\begin{itemize}
\item \vspace{0.5em} \textbf{Reward Model} ($r_{\phi}(\cdot)$ where $\phi$ denotes the parameters). The reward model learns from human preference data to predict the reward for each pair of input and output token sequences. It is a Transformer decoder followed by a linear layer that maps a sequence (the concatenation of the input and output) to a real-valued reward score.
\item \vspace{0.3em} \textbf{Value Model} or \textbf{Value Function} ($V_{\omega}(\cdot)$ where $\omega$ denotes the parameters). The value function receives reward scores from the reward model and is trained to predict the expected sum of rewards that can be obtained starting from a state. It is generally based on the same architecture as the reward model.
\item \vspace{0.3em} \textbf{Reference Model} ($\pi_{\theta_{\mathrm{ref}}}(\cdot) = \mathrm{Pr}_{\theta_{\mathrm{ref}}}(\cdot)$ where $\theta_{\mathrm{ref}}$ denotes the parameters). The reference model is the baseline LLM that serves as a starting point for policy training. In RLHF, it represents the previous version of the model or a model trained without human feedback. It is used to sample output sequences and contribute to the loss computation for policy training.
\item \vspace{0.3em} \textbf{Target Model} or \textbf{Policy} ($\pi_{\theta}(\cdot) = \mathrm{Pr}_{\theta}(\cdot)$ where $\theta$ denotes the parameters). This policy governs how the LLM decides the most appropriate next token given its context. It is trained under the supervision of both the reward model and the value model.
\end{itemize}
\vspace{0.5em}

In practice, these models need to be trained in a certain order. First, we need to initialize them using existing models. For example, the reward model and the value model can be initialized with a pre-trained LLM, while the reference model and the target model can be initialized with a model that has been instruction fine-tuned. Note that, at this point, the reference model is ready for use and will not be further updated. Second, we need to collect human preference data and train the reward model on this data. Third, both the value model and the policy are trained simultaneously using the reward model. At each position in an output token sequence, we update the value model by minimizing the MSE of value prediction, and the policy is updated by minimizing the PPO loss.

%----------------------------------------------------------------------------------------
% NEW SECTION
%----------------------------------------------------------------------------------------

\section{Improved Human Preference Alignment}

\noindent In the previous section, we reviewed the basic concepts of reinforcement learning and the general framework of RLHF. In this section, we will discuss some refinements of RLHF and alternative methods to achieve human preference alignment.

\subsection{Better Reward Modeling}
\label{sec:better-reward-modeling}

\noindent In Section \ref{sec:training-reward-models}, we highlighted the task of learning from human preferences as well as the use of pairwise ranking loss for training reward models. Here we consider more methods for reward modeling. Our discussion will be relatively general, and since the reward model is widely used in many reinforcement learning problems, it will be easy for us to apply the methods discussed here to RLHF and related applications.

\subsubsection{Supervision Signals}

\noindent The training of reward models can broadly be seen as a ranking problem, where the model learns to assign scores to outputs such that the resulting ranking reflects the preferences indicated by humans. There are several approaches to training reward models from a ranking perspective.

One approach is to extend pairwise ranking to listwise ranking. For each sample in a dataset, we can use the LLM to generate multiple outputs, and ask human experts to order these outputs. For example, given a set of four outputs $\{\mathbf{y}_1,\mathbf{y}_2,\mathbf{y}_3,\mathbf{y}_4\}$, one possible ordering of these outputs can be $\mathbf{y}_2 \succ \mathbf{y}_3 \succ \mathbf{y}_1 \succ \mathbf{y}_4$. A straightforward way to model such rankings is to sum the pairwise comparison losses. For example, we can define the listwise loss by accumulating the loss over all pairs of outputs:
\begin{eqnarray}
\mathcal{L}_{\mathrm{list}} & = & -\mathbb{E}_{\substack{(\mathbf{x},Y) \sim \mathcal{D}_r}} \Big[ \frac{2}{N(N-1)} \sum_{\substack{\mathbf{y}_a \in Y, \mathbf{y}_b \in Y\\ \mathbf{y}_a \succ \mathbf{y}_b}} \log \mathrm{Pr}(\mathbf{y}_a \succ \mathbf{y}_b | \mathbf{x}) \Big],
\end{eqnarray}

\noindent where $Y$ is a list of outputs, and $N$ is the number of outputs in the list. $\mathrm{Pr}(\mathbf{y}_a \succ \mathbf{y}_b | \mathbf{x})$ can be defined using the Bradley-Terry model, that is, $\Pr(\mathbf{y}_a \succ \mathbf{y}_b | \mathbf{x}) = \mathrm{Sigmoid}(r(\mathbf{x},\mathbf{y}_a)-r(\mathbf{x},\mathbf{y}_b))$. Here we omit the $\phi$ superscript on the $\Pr(\cdot)$ to keep the notation uncluttered.

An extension to the Bradley-Terry model for listwise ranking could involve a ranking mechanism that takes into account the entire list of outputs rather than just pairwise comparisons. One such model is the \mindex{Plackett-Luce model}, which generalizes the Bradley-Terry model to handle multiple items in a ranking \cite{plackett:1975analysis}. In the Plackett-Luce model, for each item in a list, we define a ``worth'' for this item that reflects its relative strength of being chosen over other items. For the reward modeling problem here, the worth of $\mathbf{y}$ in the list $Y$ can be defined as
\begin{eqnarray}
\alpha(\mathbf{y}) & = & \exp(r(\mathbf{x},\mathbf{y})).
\end{eqnarray}

\noindent Then the probability that $\mathbf{y}$ is selected as the top-ranked item from $Y$ is given by
\begin{eqnarray}
\Pr(\mathbf{y}\text{ is selected}|\mathbf{x},Y) & = & \frac{\alpha(\mathbf{y})}{\sum_{\mathbf{y}' \in Y} \alpha(\mathbf{y}')} \nonumber \\
                                                & = & \frac{\exp(r(\mathbf{x},\mathbf{y}))}{\sum_{\mathbf{y}' \in Y} \exp(r(\mathbf{x},\mathbf{y}'))}.
\end{eqnarray}

Suppose $\mathring{Y}$ is an ordered list $\mathbf{y}_{j_1} \succ \mathbf{y}_{j_2} \succ \cdots \succ \mathbf{y}_{j_N}$. The overall log-probability of this ordered list can be defined as the sum of the conditional log-probabilities at each stage of selection, given by
\begin{eqnarray}
\log \Pr(\mathring{Y} | \mathbf{x}) & = & \log \Pr(\mathbf{y}_{j_1} \succ \mathbf{y}_{j_2} \succ \cdots \succ \mathbf{y}_{j_N} | \mathbf{x}) \nonumber \\
                                    & = & \log \Pr(\mathbf{y}_{j_1} | \mathbf{x}, \{\mathbf{y}_{j_1},\mathbf{y}_{j_2},...,\mathbf{y}_{j_N}\}) + \nonumber \\
                                    &   & \log \Pr(\mathbf{y}_{j_2} | \mathbf{x}, \{\mathbf{y}_{j_2},...,\mathbf{y}_{j_N}\}) + \nonumber \\
                                    &   & \cdots + \nonumber \\
                                    &   & \log \Pr(\mathbf{y}_{j_N} | \mathbf{x}, \{\mathbf{y}_{j_N}\}) \nonumber \\
                                    & = & \sum_{k = 1}^{N} \log \Pr(\mathbf{y}_{j_k} | \mathbf{x}, \mathring{Y}_{\ge k}),
\end{eqnarray}

\noindent where $\mathring{Y}_{\ge k}$ represents the subset of the list of outputs that remain unselected at the $k$-th stage, i.e., $\mathring{Y}_{\ge k} = \{\mathbf{y}_{j_k},...,\mathbf{y}_{j_N}\}$. Given the log-probability $\log \Pr(\mathring{Y} | \mathbf{x})$, we can define the loss function based on the Plackett-Luce model by
\begin{eqnarray}
\mathcal{L}_{\mathrm{pl}} & = & -\mathbb{E}_{(\mathbf{x},\mathring{Y}) \sim \mathcal{D}_r} \big[ \log \Pr(\mathring{Y} | \mathbf{x}) \big].
\end{eqnarray}

There are also many other pairwise and listwise methods for modeling rankings, such as RankNet \cite{burges-etal:2005learning} and ListNet \cite{cao-etal:2007learning}. All these methods can be categorized into a large family of learning-to-rank approaches, and most of them are applicable to the problem of modeling human preferences. However, discussing these methods is beyond the scope of this chapter. Interested readers may refer to books on this topic for more details \cite{liu:2009learning,li2011:learning}.

In addition to pairwise and listwise ranking, pointwise methods provide an alternative way to train reward models and capture human preferences. Unlike methods that focus on the relative rankings between different outputs, pointwise methods treat each output independently. For example, human experts might assign a score to an individual output, such as a rating on a five-point scale. The objective is to adjust the reward model so that its outputs align with these scores. A simple way to achieve pointwise training is through regression techniques where the reward of each output is treated as a target variable. Let $\varphi(\mathbf{x},\mathbf{y})$ be the score assigned to $\mathbf{y}$ given $\mathbf{x}$ by humans. Pointwise reward models can be trained by minimizing a loss function, often based on mean squared error or other regression losses, between the predicted reward $r(\mathbf{x},\mathbf{y})$ and the actual human feedback $\varphi(\mathbf{x},\mathbf{y})$. For example, the loss function could be
\begin{eqnarray}
\mathcal{L}_{\mathrm{point}} & = & \mathbb{E}_{(\mathbf{x},\mathbf{y}) \sim \mathcal{D}_r} \big[ (\varphi(\mathbf{x},\mathbf{y}) - r(\mathbf{x},\mathbf{y}) )^2\big].
\end{eqnarray}

While pointwise methods are conceptually simpler and can directly guide the reward model to predict scores, they might not always be the best choice in RLHF. A problem is that these methods may struggle with high variance in human feedback, especially when different experts provide inconsistent scores for similar outputs. Because they focus on fitting to absolute scores rather than relative differences, inconsistencies in scoring can lead to poor model performance. Moreover, fitting to specific scored outputs might hurt generalization, particularly given that training data is often very limited in RLHF. In contrast, methods that consider relative preferences can promote the learning of more generalized preference patterns. Nevertheless, there are scenarios where pointwise methods might still be suitable. For example, in tasks where training data is abundant and the costs of obtaining accurate, consistent annotations are low, pointwise methods can prove effective.

In fact, to make the supervision signal for training the reward model more robust, we can also introduce additional regularization terms into training. For example, if we consider the first term $U_{\mathrm{ppo\text{-}clip}}(\mathbf{x},\mathbf{y};\theta)$ in Eq. (\ref{eq:ppo-objective-llm}) as a type of generalized reward, then the second term (i.e., the penalty term) can be viewed as a form of regularization for the reward model, except that here the goal is to train the policy rather than the reward model. Another example is that \citet{eisenstein-etal:2023helping} develop a regularization term based on the squared sum of rewards, and add it to the pairwise comparison loss in RLHF:
\begin{eqnarray}
\mathcal{L}_{\mathrm{reg}} & = & \mathcal{L}_{\mathrm{pair}} -\lambda \mathbb{E}_{(\mathbf{x},\mathbf{y}_a,\mathbf{y}_b) \sim \mathcal{D}_r} \big[ (r(\mathbf{x},\mathbf{y}_a) + r(\mathbf{x},\mathbf{y}_b) )^2 \big] \nonumber \\
& = & -\mathbb{E}_{(\mathbf{x},\mathbf{y}_a,\mathbf{y}_b) \sim \mathcal{D}_r} \big[ \log \mathrm{Pr}_{\phi}(\mathbf{y}_a \succ \mathbf{y}_b | \mathbf{x}) \big] \nonumber \\
&   & -\lambda  \mathbb{E}_{(\mathbf{x},\mathbf{y}_a,\mathbf{y}_b) \sim \mathcal{D}_r} \big[ (r(\mathbf{x},\mathbf{y}_a) + r(\mathbf{x},\mathbf{y}_b) )^2 \big],
\end{eqnarray}

\noindent where $\lambda$ is the weight of the regularization term. Optimizing with this regularization term can help mitigate the underdetermination of reward models\footnote{A model is called underdetermined if there are multiple alternative sets of parameters that can achieve the same objective.}.

\subsubsection{Sparse Rewards vs. Dense Rewards}

\noindent As discussed in Section \ref{sec:human-preference-alignment}, the rewards provided by human feedback are typically sparse: they are observed only at the end of sequences, rather than continuously throughout the generation process. Dealing with sparse rewards has long been a concern in reinforcement learning, and has been one of the challenges in many practical applications. For example, in robotics, reward functions often need to be shaped to ease optimization rather than relying only on end-of-sequence rewards. Various methods have been developed to address this issue. One common approach is reward shaping, where the original function is modified to include intermediate rewards, thereby providing more immediate feedback. Also, one can adopt curriculum learning to sequence tasks such that their complexity gradually increases. This can help models master simpler tasks first and gradually prepare them for more complex challenges. There are many such methods that can mitigate the impact of sparse rewards, such as intrinsic motivation and hindsight experience replay. Most of these methods are general and discussed extensively in the broader reinforcement learning literature, such as the seminal textbook by \citet{Sutton-and-Barto:2018RL}.

Although we do not discuss methods for mitigating sparse rewards in detail here, an interesting question arises: why are sparse rewards so successful in RLHF? Recall from Section \ref{sec:rl-basics} that the supervision signal received at each time step $t$ is not the reward for the current action, but rather some form of the accumulated rewards from $t$ until the last time step. Such supervision signals are dense over the sequence, because the reward obtained at the end of the sequence can be transferred back to that time step, regardless of which time step it is. In other words, the sparse rewards are transformed into the dense supervision signals. Furthermore, from the perspective of reward shaping, \citet{ng1999-etal:policy} show that the reward at $t$ is computed as
\begin{eqnarray}
r'(s_t,a_t,s_{t+1}) & = & r(s_t,a_t,s_{t+1}) + f(s_t,a_t,s_{t+1}), \label{eq:transformed-reward-function}
\end{eqnarray}

\noindent where $r'(\cdot)$ is the transformed reward function, $r(\cdot)$ is the original reward function, and $f(\cdot)$ is the shaping reward function. To ensure the optimality of the policy under the transformed reward function, the shaping reward function can be given in the form
\begin{eqnarray}
f(s_t,a_t,s_{t+1}) & = & \gamma \Phi(s_{t+1}) - \Phi(s_t), \label{eq:shaping-reward-function}
\end{eqnarray}

\noindent where $\Phi(s)$ is called the potential value of the state $s$. If we define $\Phi(s)$ as the value function as in Eq. (\ref{eq:general-value-function}) and substitute Eq. (\ref{eq:shaping-reward-function}) into Eq. (\ref{eq:transformed-reward-function}), we obtain
\begin{eqnarray}
r'(s_t,a_t,s_{t+1}) & = & r(s_t,a_t,s_{t+1}) + \gamma V(s_{t+1}) - V(s_t).
\end{eqnarray}

\noindent It is interesting to see that this expression is equivalent to the one-step temporal-difference error used in PPO. This relates advantage-based methods to reward shaping: the advantage is essentially a shaped reward.

On the other hand, one of the reasons for adopting end-of-sequence rewards lies in the nature of the RLHF tasks. Unlike traditional reinforcement learning environments where the agent interacts with a dynamic environment, RLHF tasks often involve complex decision-making based on linguistic or other high-level cognitive processes. These processes do not lend themselves easily to frequent and meaningful intermediate rewards because the quality and appropriateness of the actions can only be fully evaluated after observing their impact in the larger context of the entire sequence or task. In this case, the reward signals based on human feedback, though very sparse, are typically very informative and accurate. Consequently, this sparsity, together with the high informativeness and accuracy of the human feedback, can make the learning both robust and efficient.

\subsubsection{Fine-grained Rewards}
\label{sec:fine-grained-rewards}

\noindent For many applications, our objective will be more complex than merely evaluating an entire text. For example, in sentiment analysis, we often do not just determine the sentiment of a text, but need to analyze the sentiment in more detail by associating it with specific aspects of a topic discussed in the text. Consider the sentence ``The camera of the phone is excellent, but the battery life is disappointing''. In this example, we would need to separately analyze the sentiments expressed about the camera and the battery. Such analysis, known as aspect-based sentiment analysis, helps provide a finer-grained understanding of the customer review compared to general sentiment analysis.

For the problem of reward modeling, we often need to model different parts of a sequence at a similarly fine-grained level. A straightforward way to achieve this is to divide a sequence into distinct segments and compute a reward for each segment independently \cite{wu-etal:2023finegrained}. Suppose an output token sequence $\mathbf{y}$ can be divided into $n_s$ segments $\{\bar{\mathbf{y}}_1,...,\bar{\mathbf{y}}_{n_s}\}$ based on some criterion. We can then use the reward model to evaluate each of these segments. By taking the input $\mathbf{x}$, the full output $\mathbf{y}$, and the specific segment $\bar{\mathbf{y}}_k$ into account, the reward score for the $k$-th segment is given by
\begin{eqnarray}
r^k & = & r(\mathbf{x}, \mathbf{y}, \bar{\mathbf{y}}_k).
\end{eqnarray}

\noindent The total reward score for the entire output sequence is then simply the sum of these segment-level rewards
\begin{eqnarray}
r(\mathbf{x},\mathbf{y}) & = & \sum_{k=1}^{n_s} r(\mathbf{x}, \mathbf{y}, \bar{\mathbf{y}}_k).
\end{eqnarray}

\noindent This reward  $r(\mathbf{x},\mathbf{y})$ can then be used to train the policy via standard RL algorithms like PPO.

A problem with this approach is that training reward models at the segment level is not as straightforward as learning from human preferences on entire texts, as it is difficult to obtain segment-level human preference data. For rating-like problems (e.g., we rate a segment according to its level of misinformation), one simple approach is to assign a rating score to each segment and train the reward model using pointwise methods. For example, we can use a strong LLM to rate the sequences $\bar{\mathbf{y}}_1...\bar{\mathbf{y}}_{k-1}$ and $\bar{\mathbf{y}}_1...\bar{\mathbf{y}}_k$, and obtain the scores $s(\bar{\mathbf{y}}_1...\bar{\mathbf{y}}_{k-1})$ and $s(\bar{\mathbf{y}}_1...\bar{\mathbf{y}}_k)$. We can then define the score of the segment $\bar{\mathbf{y}}_{k}$ as the difference between $s(\bar{\mathbf{y}}_1...\bar{\mathbf{y}}_k)$ and $s(\bar{\mathbf{y}}_1...\bar{\mathbf{y}}_{k-1})$
\begin{eqnarray}
s(\bar{\mathbf{y}}_k) & = &s(\bar{\mathbf{y}}_1...\bar{\mathbf{y}}_k) - s(\bar{\mathbf{y}}_1...\bar{\mathbf{y}}_{k-1}).
\end{eqnarray}

Using these segment-level scores, we can train the reward model with a regression loss function
\begin{eqnarray}
\mathcal{L}_{\mathrm{rating}} & = & \mathbb{E}_{(\mathbf{x},\mathbf{y},\bar{\mathbf{y}}_k) \sim \mathcal{D}_r} \big[ (s(\bar{\mathbf{y}}_k) - r(\mathbf{x}, \mathbf{y}, \bar{\mathbf{y}}_k))^2 \big].
\end{eqnarray}

Sometimes, alignment can be treated as a classification problem, for example, we assess whether a segment has ethical issues. In this case, the segment can be labeled as ethical or unethical, either by humans or using additional classifiers. Given these discrete labels, we can train the reward model using classification loss functions. Suppose the ground truth label is $\hat{r} \in \{1,-1\}$, where $-1$ indicates an unethical segment and $1$ indicates an ethical one, and let $r(\mathbf{x}, \mathbf{y}, \bar{\mathbf{y}}_k)$ represent the scalar output produced by the reward model\footnote{To allow the reward model to output classification probabilities or logits, we can adapt the linear layer described in Section \ref{sec:training-reward-models} accordingly.}. The reward model can then be trained using the hinge loss:
\begin{eqnarray}
\mathcal{L}_{\mathrm{hinge}} & = & \mathbb{E}_{(\mathbf{x},\mathbf{y},\bar{\mathbf{y}}_k) \sim \mathcal{D}_r} \big[  \max(0, 1 - r(\mathbf{x}, \mathbf{y}, \bar{\mathbf{y}}_k) \cdot \hat{r}) \big],
\end{eqnarray}

\noindent where $\hat{r} \in \{1,-1\}$ denotes the ground truth label.

The remaining issue here is how to split $\mathbf{y}$ into segments. One approach is to define a fixed-length segmentation, where $\mathbf{y}$ is divided into equal-length chunks. However, this may not always be ideal, as the content of the sequence may not align well with fixed boundaries. An alternative approach is to segment $\mathbf{y}$ based on specific linguistic or semantic cues, such as sentence boundaries, topic shifts, or other meaningful structures in the text. Such a segmentation can be achieved by using linguistic segmentation systems or prompting LLMs to identify natural breaks in the sequence. Another approach is to use dynamic segmentation methods based on the complexity of the sequence. For example, segments could be defined where there is a significant change in the reward score, which might correspond to shifts in the task being modeled.

\subsubsection{Combination of Reward Models}

\noindent A reward model can be viewed as a proxy for the environment. Since the true environment is often too complex or unknown, developing a perfect proxy for the environment is generally not possible. As a result, over-aligning LLMs with this imperfect proxy might lead to decreased performance, known as the \mindex{overoptimization problem} \cite{stiennon-etal:2020learning,gao-etal:2023scaling}\footnote{This problem is also called \mindex{reward hacking} or \mindex{reward gaming} \cite{krakovna-etal:2020specification,skalse-etal:2022defining,pan-etal:2022theeffects}, which refers to the phenomenon where the agent attempts to trick the reward model but fails to align its actions with the true intended objectives of the task. Imagine a student who is assigned homework and is rewarded with points or praise for completing it. The student might then find ways to finish the homework with minimal effort to maximize the reward, such as copying and pasting solutions from the internet or previous assignments, rather than solving the problems themselves.}. We can also explain this through Goodhart's law, which states: \textit{when a measure becomes a target, it ceases to be a good measure} \cite{goodhart:1984problems}.

Addressing the overoptimization problem is not easy, and no fully satisfactory solution currently exists. The ideal approach might be to develop an oracle reward model that can perfectly capture the true objectives of the task and prevent the agent from ``tricking''. However, creating such a model is extremely difficult due to the complexity of the real-world environment, as well as the challenge of defining all the relevant factors that contribute to the desired outcome. Instead, a more practical approach is to combine multiple reward models, thereby reducing the mismatch between the training objective and the true objective that arises from using a single reward model \cite{coste-etal:2024reward}.

Combining a set of reward models is conceptually straightforward and can often be formulated as an ensemble learning problem. A simple yet common approach is to combine the outputs of these models to obtain a more reliable reward estimation:
\begin{eqnarray}
r_{\mathrm{combine}} & = & \sum_{k=1}^{K} w_k \cdot r_k(\mathbf{x},\mathbf{y}),
\end{eqnarray}

\noindent where $r_k(\cdot)$ is the $k$-th reward model in the ensemble, $w_k$ is the weight of $r_k(\cdot)$, and $K$ is the number of reward models. This combined reward can then be used to supervise the training of a policy. In fact, there are many ways to combine different models, for example, one can make predictions using Bayesian model averaging
or develop a fusion network to learn to combine the predictions from different models. Alternatively, one can frame this task as a multi-objective optimization problem, and use multiple reward models to train the policy simultaneously. These methods have been extensively discussed in the literature on optimization and machine learning \cite{miettinen:1999nonlinear,Bishop:2006}.

In addition to model combination methods, another important issue is how to collect or construct multiple different reward models. One of the simplest approaches is to employ ensemble learning techniques, such as developing diverse reward models from different subsets of a given dataset or from various data sources. For RLHF, it is also possible to construct reward models based on considerations of different aspects of alignment. For example, we can develop a reward model to evaluate the factual accuracy of the output and another reward model to evaluate the completeness of the output. These two models are complementary to each other, and can be combined to improve the overall evaluation of the output. Another approach is to employ different off-the-shelf LLMs as reward models. This approach is simple and practical, as many strong pretrained LLMs are readily available and we just need to use them with little or no modification. This raises an interesting question: can an LLM that aligns with other LLMs outperform those LLMs? Probably not at first glance. Intuitively, one might assume that a target LLM merely imitating other LLMs based on limited supervision would struggle to fully capture the nuanced behaviors of its supervisors. However, given the strong generalization ability of LLMs, this approach can, in fact, be quite beneficial. For example, using open-source or commercial LLMs as reward models has demonstrated strong performance in aligning LLMs, even achieving state-of-the-art results on several popular tasks \cite{lambert-etal:2024rewardbench}.

\subsection{Direct Preference Optimization}

\noindent Although learning reward models is a standard step in reinforcement learning, it makes the entire training process much more complex than standard supervised training. Training a reliable reward model is itself not an easy task, and a poorly trained reward model can greatly affect the outcome of policy learning. We now consider an alternative alignment method, called \mindex{direct preference optimization} (\mindex{DPO}), which simplifies the training framework by eliminating the need to explicitly model rewards \cite{rafailov-etal:2024direct}. This method directly optimizes the policy based on user preferences, rather than training a separate reward model. As a result, we can achieve human preference alignment in a supervised learning-like fashion. Figure \ref{fig:rlhf-vs-dpo} shows a comparison of the standard RLHF method and the DPO method.

\begin{figure}[!t]
\centering
\input{./Chapter10/Figures/figure-rlhf-vs-dpo}
\caption{Standard RLHF (PPO) vs. DPO. In RLHF, the human preference data is used to train a reward model, which is then employed in training the policy as well as the value function. In DPO, the use of human preference data is more direct, and the policy is trained on this data without the need for reward model training.}
\label{fig:rlhf-vs-dpo}
\end{figure}

Before deriving the DPO objective, let us first review the objective of policy training used in RLHF. As discussed in Section \ref{sec:rlhf-training-llms}, the policy is typically trained by optimizing a loss function with a penalty term. The DPO method assumes a simple loss function where the quality of the output $\mathbf{y}$ given the input $\mathbf{x}$ is evaluated by the reward model $r(\mathbf{x},\mathbf{y})$. The training objective is thus given by
\begin{eqnarray}
\tilde{\theta} & = & \argmin_{\theta} \mathbb{E}_{\mathbf{x} \sim \mathcal{D}} \mathbb{E}_{\mathbf{y} \sim \pi_{\theta}(\cdot|\mathbf{x})} \big[ \underbrace{-r(\mathbf{x},\mathbf{y})}_{\text{loss}} + \beta \underbrace{(\log \pi_{\theta}(\mathbf{y}|\mathbf{x}) - \log \pi_{\theta_{\mathrm{ref}}}(\mathbf{y}|\mathbf{x}) )}_{\text{penalty}} \big]. \label{eq:dpo-basic-objective}
\end{eqnarray}

\noindent Note that in this optimization problem, only the term $\pi_{\theta}(\mathbf{y}|\mathbf{x})$ depends on the target policy $\pi_{\theta}(\cdot)$. Both the reward model $r(\mathbf{x}, \mathbf{y})$ and the reference model $\pi_{\theta_{\mathrm{ref}}}(\mathbf{y}|\mathbf{x})$ are assumed to be fixed given $\mathbf{x}$ and $\mathbf{y}$. This is a strong assumption compared with PPO, but as will be shown later, it simplifies the problem and is important for deriving the DPO objective.

Since $\theta$ is the variable we want to optimize, we rearrange the right-hand side of Eq. (\ref{eq:dpo-basic-objective}) to make the dependence on $\theta$ explicit:
\begin{eqnarray}
&    & \tilde{\theta} \nonumber \\
& = & \argmin_{\theta} \mathbb{E}_{\mathbf{x} \sim \mathcal{D}} \mathbb{E}_{\mathbf{y} \sim \pi_{\theta}(\cdot|\mathbf{x})} \big[ \beta \log \pi_{\theta}(\mathbf{y}|\mathbf{x}) - \beta \log \pi_{\theta_{\mathrm{ref}}}(\mathbf{y}|\mathbf{x}) - r(\mathbf{x},\mathbf{y}) \big] \nonumber \\
& = & \argmin_{\theta} \mathbb{E}_{\mathbf{x} \sim \mathcal{D}} \mathbb{E}_{\mathbf{y} \sim \pi_{\theta}(\cdot|\mathbf{x})} \big[ \log \pi_{\theta}(\mathbf{y}|\mathbf{x}) - \big(\log \pi_{\theta_{\mathrm{ref}}}(\mathbf{y}|\mathbf{x}) + \frac{1}{\beta} r(\mathbf{x},\mathbf{y}) \big) \big] \nonumber \\
& = & \argmin_{\theta} \mathbb{E}_{\mathbf{x} \sim \mathcal{D}} \mathbb{E}_{\mathbf{y} \sim \pi_{\theta}(\cdot|\mathbf{x})} \Big[ \underbrace{\log \pi_{\theta}(\mathbf{y}|\mathbf{x})}_{\text{dependent on $\theta$}} - \underbrace{\log \Big(\pi_{\theta_{\mathrm{ref}}}(\mathbf{y}|\mathbf{x}) \exp \big(\frac{1}{\beta} r(\mathbf{x},\mathbf{y}) \big) \Big)}_{\text{not dependent on $\theta$}} \Big].
\label{eq:dpo-raw-objective-with-no-normalization}
\end{eqnarray}

This equation defines the objective function as the difference between the log-probability of the policy and another function of $\mathbf{y}$. This form of the objective function is not ideal, as we usually prefer to see the difference between two distributions, so that we can interpret this difference as some kind of divergence between the distributions. A simple idea is to convert the second term (i.e., $ \log \pi_{\theta_{\mathrm{ref}}}(\mathbf{y}|\mathbf{x}) \exp(\frac{1}{\beta} r(\mathbf{x},\mathbf{y}))$) into a log-probability distribution over the domain of $\mathbf{y}$. If we treat $ \pi_{\theta_{\mathrm{ref}}}(\mathbf{y}|\mathbf{x}) \exp(\frac{1}{\beta} r(\mathbf{x},\mathbf{y}))$ as an unnormalized probability of $y$, we can convert it into a normalized probability by dividing it by a normalization factor:
\begin{eqnarray}
Z(\mathbf{x}) & = & \sum_{\mathbf{y}} \pi_{\theta_{\mathrm{ref}}}(\mathbf{y}|\mathbf{x}) \exp \big(\frac{1}{\beta} r(\mathbf{x},\mathbf{y}) \big).
\end{eqnarray}

Hence we can define a probability distribution by
\begin{eqnarray}
\pi^*(\mathbf{y}|\mathbf{x}) & = & \frac{\pi_{\theta_{\mathrm{ref}}}(\mathbf{y}|\mathbf{x}) \exp \big(\frac{1}{\beta} r(\mathbf{x},\mathbf{y}) \big)}{Z(\mathbf{x})}.
\end{eqnarray}

We then rewrite Eq. (\ref{eq:dpo-raw-objective-with-no-normalization}) as
\begin{eqnarray}
\tilde{\theta} & = & \argmin_{\theta} \mathbb{E}_{\mathbf{x} \sim \mathcal{D}} \mathbb{E}_{\mathbf{y} \sim \pi_{\theta}(\cdot|\mathbf{x})} \Big[ \log \pi_{\theta}(\mathbf{y}|\mathbf{x}) - \log \frac{\pi_{\theta_{\mathrm{ref}}}(\mathbf{y}|\mathbf{x}) \exp \big(\frac{1}{\beta} r(\mathbf{x},\mathbf{y}) \big)}{Z(\mathbf{x})} \nonumber \\
& & \hspace{10.5em} - \log Z(\mathbf{x}) \Big] \nonumber \\
& = & \argmin_{\theta} \mathbb{E}_{\mathbf{x} \sim \mathcal{D}} \mathbb{E}_{\mathbf{y} \sim \pi_{\theta}(\cdot|\mathbf{x})} \Big[ \log \pi_{\theta}(\mathbf{y}|\mathbf{x}) - \log \pi^{*}(\mathbf{y}|\mathbf{x}) - \log Z(\mathbf{x}) \Big] \nonumber \\
& = & \argmin_{\theta} \mathbb{E}_{\mathbf{x} \sim \mathcal{D}} \bigg[ \mathbb{E}_{\mathbf{y} \sim \pi_{\theta}(\cdot|\mathbf{x})} \Big[ \log \pi_{\theta}(\mathbf{y}|\mathbf{x}) - \log \pi^{*}(\mathbf{y}|\mathbf{x}) \Big] \nonumber \\
&  & \hspace{6.5em} - \mathbb{E}_{\mathbf{y} \sim \pi_{\theta}(\cdot|\mathbf{x})} \big[ \log Z(\mathbf{x}) \big] \bigg] \nonumber \\
& = & \argmin_{\theta} \mathbb{E}_{\mathbf{x} \sim \mathcal{D}} \Big[ \underbrace{D_{\mathrm{KL}} \big(\pi_{\theta}(\cdot|\mathbf{x})\ \|\ \pi^{*}(\cdot|\mathbf{x}) \big)}_{\text{KL divergence}} - \underbrace{\log Z(\mathbf{x})}_{\text{constant wrt. $\theta$}} \Big].
\end{eqnarray}

Since $\log Z(\mathbf{x})$ is independent of $\theta$, it does not affect the result of the $\argmin_{\theta}$ operation, and can be removed from the objective. Now we obtain a new training objective which finds the optimal policy $\pi_{\theta}$ by minimizing the KL divergence between $\pi_{\theta}(\cdot|\mathbf{x})$ and $\pi^{*}(\cdot|\mathbf{x})$
\begin{eqnarray}
\tilde{\theta} & = & \argmin_{\theta} \mathbb{E}_{\mathbf{x} \sim \mathcal{D}} \Big[ D_{\mathrm{KL}} \big(\pi_{\theta}(\cdot|\mathbf{x})\ \|\ \pi^{*}(\cdot|\mathbf{x}) \big) \Big].
\end{eqnarray}

Clearly, the solution to this optimization problem is given by
\begin{eqnarray}
\pi_{\theta}(\mathbf{y}|\mathbf{x}) & = & \pi^{*}(\mathbf{y}|\mathbf{x}) \nonumber \\
& = & \frac{\pi_{\theta_{\mathrm{ref}}}(\mathbf{y}|\mathbf{x}) \exp \big(\frac{1}{\beta} r(\mathbf{x},\mathbf{y}))}{Z(\mathbf{x})}.
\end{eqnarray}

Given this equation, we can express the reward $r(\mathbf{x},\mathbf{y})$ using the target model $\pi_{\theta}(\mathbf{y}|\mathbf{x})$, the reference model $\pi_{\theta_{\mathrm{ref}}}(\mathbf{y}|\mathbf{x})$, and the normalization factor $Z(\mathbf{x})$:
\begin{eqnarray}
r(\mathbf{x},\mathbf{y}) & = & \beta \left(\log \frac{\pi_{\theta}(\mathbf{y}|\mathbf{x})}{\pi_{\theta_{\mathrm{ref}}}(\mathbf{y}|\mathbf{x})} + \log Z(\mathbf{x}) \right). \label{eq:dpo-reward-model-expression}
\end{eqnarray}

This is interesting because we initially seek to learn the policy $\pi_{\theta}(\cdot)$ using the reward model $r(\mathbf{x},\mathbf{y})$, but eventually obtain a representation of the reward model based on the policy. Given the reward model defined in Eq. (\ref{eq:dpo-reward-model-expression}), we can apply it to the Bradley-Terry model to calculate the preference probability (also see Section \ref{sec:training-reward-models}):
\begin{eqnarray}
\mathrm{Pr}_{\theta}(\mathbf{y}_a \succ \mathbf{y}_b | \mathbf{x}) & = & \mathrm{Sigmoid}(r(\mathbf{x},\mathbf{y}_a)-r(\mathbf{x},\mathbf{y}_b)) \nonumber \\
& = & \mathrm{Sigmoid}\bigg(\beta \Big(\log \frac{\pi_{\theta}(\mathbf{y}_a|\mathbf{x})}{\pi_{\theta_{\mathrm{ref}}}(\mathbf{y}_a|\mathbf{x})} + \log Z(\mathbf{x}) \Big) - \nonumber \\
&    & \hspace{4.3em} \beta \Big(\log \frac{\pi_{\theta}(\mathbf{y}_b|\mathbf{x})}{\pi_{\theta_{\mathrm{ref}}}(\mathbf{y}_b|\mathbf{x})} + \log Z(\mathbf{x}) \Big) \bigg) \nonumber \\
& = & \mathrm{Sigmoid}\bigg( \beta \log \frac{\pi_{\theta}(\mathbf{y}_a|\mathbf{x})}{\pi_{\theta_{\mathrm{ref}}}(\mathbf{y}_a|\mathbf{x})}  - \beta \log \frac{\pi_{\theta}(\mathbf{y}_b|\mathbf{x})}{\pi_{\theta_{\mathrm{ref}}}(\mathbf{y}_b|\mathbf{x})} \bigg).
\end{eqnarray}

This formula is elegant because it converts the difference in rewards into the difference in ratio functions, and we do not need to calculate the value of $Z(\mathbf{x})$. As a result, we no longer need a reward model, but only need the target policy and reference model to calculate the probability of preferences. Finally, we can train the target policy by minimizing the following DPO loss function
\begin{eqnarray}
\mathcal{L}_{\mathrm{dpo}}(\theta) & = & -\mathbb{E}_{(\mathbf{x},\mathbf{y}_a,\mathbf{y}_b) \sim \mathcal{D}_r} \big[ \log \mathrm{Pr}_{\theta}(\mathbf{y}_a \succ \mathbf{y}_b | \mathbf{x}) \big].
\end{eqnarray}

\noindent The form of this loss function is very similar to that used in training reward models in RLHF (see Eq. (\ref{eq:pairwise-reward-loss-expectation})). But it should be noted that the loss function here depends on the parameters of the policy (i.e., $\theta$) rather than the parameters of the reward model (i.e., $\phi$).

The main advantage of DPO lies in its simplicity and efficiency. The DPO objective is very straightforward --- it directly optimizes for preference-based feedback, rather than relying on separately developed reward models. Moreover, DPO is generally more sample-efficient, as it learns from a fixed dataset without the need for the computationally expensive sampling process used in PPO. This makes DPO a popular method for human preference alignment, especially when developing and applying reward models via reinforcement learning is challenging.

DPO can broadly be viewed as an \mindex{offline reinforcement learning} method, where the training data is pre-collected and fixed, and there is no exploration. In contrast, online reinforcement learning methods like PPO, which require exploring new states through interaction with the environment (using the reward model as a proxy), also have their unique advantages. One of the benefits of online reinforcement learning is that it allows the agent to continuously adapt to changes in the environment by learning from real-time feedback. This means that, unlike offline methods, online methods are not constrained by the static nature of pre-collected data and can discover new problem-solving strategies. In addition, exploration can help the agent cover a wider range of state-action pairs, thus improving generalization. This could be an important advantage for LLMs, as generalization is considered a critical aspect in applying such large models.

\subsection{Automatic Preference Data Generation}

\noindent Although learning from human preferences is an effective and popular method for aligning LLMs, annotating preference data is costly. Learning from human feedback not only suffers from limited scalability, but may also introduce bias because human feedback is inherently subjective. As a result, one can turn to AI feedback methods to address these scalability and consistency issues without the limitations associated with human annotators.

As with data generation for instruction fine-tuning, generating preference data using LLMs is straightforward. Given a set of inputs, we first use an LLM to generate pairs of outputs. Then, we prompt the LLM to determine the preferred output for each pair, given the corresponding input. Below is an example of prompting the LLM to generate a preference label for a pair of consumer service responses.

\vspace{0.5em}
\begin{tcolorbox}[frame empty]

\begingroup
\setlength{\leftskip}{2em}
\setlength{\rightskip}{2em}

Consider a customer service scenario where a customer poses a request. You will review two responses to this request. Please indicate which response is preferred. Note that a good response should be courteous, clear, and concise. It should address the customer's concern directly, provide helpful information or a solution, and maintain a positive tone.

\vspace{0.3cm}

Request:

\vspace{0.1cm}

\textit{Hello, I noticed that my order hasn't arrived yet, though it was scheduled to arrive several days ago. Could you please update me on its status? Thank you!}

\vspace{0.2cm}

Response A:

\vspace{0.1cm}

\textit{I'm very sorry for the delay and understand how disappointing this can be. We're doing our best to sort this out quickly for you.}

\vspace{0.3cm}

Response B:

\vspace{0.1cm}

\textit{Hey, stuff happens! Your package will get there when it gets there, no need to stress.}

\vspace{0.3cm}

\underline{Response A is preferred.}

\endgroup

\end{tcolorbox}
\vspace{0.5em}

Once we collect such preference labels, we can use them, along with the input and two responses, to train the reward model. We can, of course, consider providing a few demonstrations or using sophisticated prompting techniques, such as CoT, to improve labeling performance. For example, we can include in the prompt an example showing how and why one of the two responses is preferred based on a CoT rationale.

In addition to preference labels, we can also obtain the probability associated with each label \cite{lee-etal:2023rlaif}. A simple method is to extract the probabilities for the label tokens, such as ``A'' and ``B'', from the probabilities output by the LLM. We can then use the Softmax function or other normalization techniques to re-normalize these probabilities into a distribution over the labels. These probabilities of preferred labels can serve as pointwise supervision signals for training the reward model, as discussed in Section \ref{sec:better-reward-modeling}.

While scaling up AI data generation is easy, ensuring its accuracy and diversity remains crucial. Here, the data quality and diversity issues involve not only the labeling of preferences but also the inputs and outputs of the model. Therefore, we often need to use a variety of techniques to obtain large-scale, high-quality data. For example, one can generate diverse model outputs and annotations by using different LLMs, prompts, in-context demonstrations, and so on \cite{cui-etal:2024ultra}. \citet{dubois-etal:2024alpacafarm} report that the variability in pairwise preference data is important for training LLMs from either human or AI feedback.

While learning from AI feedback is highly scalable and generally more consistent than human feedback, this method is more suited to well-defined tasks where objective performance metrics are available. By contrast, learning from human feedback is more advantageous when aligning AI systems with human values, preferences, and complex real-world tasks that require understanding of subtle or subjective context. These methods can be combined to train LLMs that benefit from both human insights and the scalability of AI feedback.

\subsection{Step-by-step Alignment}

\noindent So far, our discussion of alignment has primarily focused on the use of reward models for evaluating entire input-output sequence pairs. These methods can be easily adapted to scenarios where the correctness of the output can be verified based on the final answer. For example, in the task of calculating a mathematical expression, a reward model can provide positive feedback if the answer is correct, and negative feedback if the answer is wrong. However, in many problems that require complex reasoning, simply examining the correctness of the final result is insufficient for learning. Imagine a student who is only given the final answer to a challenging math problem. Knowing whether the final answer is right or wrong does not help the student figure out where they went wrong and how to calculate the correct answer. A better approach would be to guide the student with a step-by-step breakdown of the problem-solving process and help the student understand the underlying concepts and logic behind these steps.

In Chapter \ChapterLLMPrompting, we studied CoT methods to prompt LLMs to explicitly write out intermediate steps or the reasoning process needed to reach a conclusion or solve a problem. We saw that breaking down a problem into smaller parts could make it easier to understand the solution path and increase the accuracy of the output. These methods can be naturally extended to the alignment of LLMs, that is, we supervise the model during the intermediate steps of reasoning. Consider a reasoning task where an LLM produces a sequence of reasoning steps $\mathbf{y} = \{\bar{\mathbf{y}}_1,...,\bar{\mathbf{y}}_{n_s}\}$ for the given input. The result of the reasoning is assumed to be contained in the last step $\bar{\mathbf{y}}_{n_s}$, and can be easily verified. For this reasoning problem, \citet{uesato-etal:2022solving} categorize LLM fine-tuning approaches into two classes:

\begin{itemize}
\item \vspace{0.5em} \mindex{Outcome-based Approaches}. Supervision occurs only when the end result is verified. This is a standard method for learning from human feedback we have discussed in this chapter. For example, the LLM is optimized to maximize some form of the reward $r(\mathbf{x},\mathbf{y})$.
\item \vspace{0.3em} \mindex{Process-based Approaches}. Supervision is applied to all intermediate steps in addition to the last step. To do this, we need to develop a reward model to give a supervision signal at each step, and design loss functions that can make use of such supervision signals.
\end{itemize}
\vspace{0.5em}

Figure \ref{fig:outcome-vs-process-based-approaches} shows two LLM outputs for an example math problem. Although the LLM gives the correct final answer in both cases, it makes mistakes during the problem-solving process in the second output. Outcome-based approaches overlook these mistakes and give positive feedback for the entire solution. By contrast, process-based approaches can take these mistakes into account and provide additional guidance on the detailed reasoning steps.

\begin{figure}[!t]
\centering
\input{./Chapter10/Figures/figure-reasoning-paths-for-process-based-approaches}
\caption{Two LLM responses to a math problem. In response 1, both the final result and all the reasoning steps are correct. In response 2, the final result is correct, but there are mistakes in the reasoning process (highlighted in red). For outcome-based approaches, both responses are considered correct. For process-based approaches, the mistakes in response 2 can be considered in reward modeling.}
\label{fig:outcome-vs-process-based-approaches}
\end{figure}

A key challenge for process-based approaches is that we need to get step-level feedback during a (potentially) long reasoning path. We can collect or generate reasoning paths corresponding to problems from existing datasets. Human experts then annotate each step with correctness labels. These annotations can be used to directly train LLMs or as rewards in reward modeling. However, in practice, richer annotations are often employed \cite{lightman-etal:2024lets}. In addition to the \textit{correct} and \textit{incorrect} labels, a step can also be labeled as \textit{neutral} to indicate that while the step may be technically correct, it might still be problematic within the overall reasoning process. Furthermore, to improve the efficiency of data annotation, techniques such as active learning can be employed. Identifying obvious errors usually does not significantly contribute to learning from reasoning mistakes. Instead, annotating steps that the model confidently considers correct but are actually problematic is often more effective.

Given a set of step-level annotated reasoning paths and corresponding inputs, we can train a reward model to provide feedback for supervising policy learning. The reward model can be treated as a classification model, and so its architecture can be a Transformer decoder with a Softmax layer stacked on top. At step $k$, the reward model takes both the problem description (denoted by $\mathbf{x}$) and the reasoning steps generated so far (denoted by $\bar{\mathbf{y}}_{\le k}$) as input and outputs a probability distribution over the label set $\{\text{\textit{correct}}, \text{\textit{incorrect}}\}$ or $\{\text{\textit{correct}}, \text{\textit{incorrect}}, \text{\textit{neutral}}\}$. Then the learned reward model is used to evaluate reasoning paths by assessing the correctness of each step. A simple method to model correctness is to count the number of steps that are classified as \textit{correct}, given by
\begin{eqnarray}
r(\mathbf{x},\mathbf{y}) & = & \sum_{k=1}^{n_s} \delta(\text{\textit{correct}},C(\mathbf{x},\bar{\mathbf{y}}_{\le k})),
\end{eqnarray}

\noindent where $C(\mathbf{x},\bar{\mathbf{y}}_{\le k})$ denotes the label with the maximum probability. We can also use log-probabilities of classification to define the reward of the entire path
\begin{eqnarray}
r(\mathbf{x},\mathbf{y}) & = &  \sum_{k=1}^{n_s} \log \Pr(\textit{correct}|\mathbf{x},\bar{\mathbf{y}}_{\le k}),
\end{eqnarray}

\noindent where $\Pr(\textit{correct}|\mathbf{x},\bar{\mathbf{y}}_{\le k})$ denotes the probability of the \textit{correct} label generated by the reward model. The reward score $r(\mathbf{x},\mathbf{y})$ can then be used to train the policy in RLHF as usual.

While we restrict our discussion to math problems, the approaches described here are general and can be applied to a wide variety of tasks that involve multi-step reasoning and decision-making. Moreover, we can consider various aspects when assessing the quality of a step, rather than just its correctness. For example, in dialogue systems, responses must not only be accurate but also contextually appropriate across multiple turns of conversation. If a model provides a correct response but fails to maintain coherence in the context of the ongoing dialogue, step-level feedback could help the model identify and correct such discrepancies. Also note that the process-based approaches are related to the fine-grained reward modeling approaches discussed in Section \ref{sec:fine-grained-rewards}. All these approaches essentially aim to provide more detailed supervision to LLMs by breaking their outputs into smaller, more manageable steps. However, process-based feedback focuses more on evaluating the correctness of a step based on its preceding steps, while the approaches in Section \ref{sec:fine-grained-rewards} emphasize evaluating each step independently.

The idea of aligning LLMs step by step has great application potential, especially considering the recent shift towards more complex reasoning tasks in the use of LLMs. For example, both the GPT o1 and o3 models are designed with more advanced reasoning techniques (such as long internal CoT) to solve challenging problems like scientific and mathematical reasoning \cite{openai:2024learning}. These tasks often rely on long and complex reasoning paths, and therefore, it seems essential to introduce detailed supervision signals in the reasoning process. Moreover, from a practical perspective, effective supervision on long reasoning paths not only improves reasoning performance, but also helps the model eliminate unnecessary reasoning steps, thereby reducing reasoning complexity and improving efficiency.

\subsection{Inference-time Alignment}

\noindent So far in this section, we have explored a variety of methods to align models with human preferences and annotations. However, one of the significant limitations of many such methods is that LLMs must be fine-tuned. For RLHF and its variants, training LLMs with reward models can be computationally expensive and unstable, leading to increased complexity and costs when applying these approaches. In this case, we can consider performing alignment at inference time, thus avoiding the additional complexity and effort involved.

One simple way to achieve inference-time alignment is to use the reward model to select the best one from $N$ alternative outputs generated by the LLM, a method known as \mindex{Best-of-$N$ sampling} (\mindex{BoN sampling}). We can consider BoN sampling as a form of reranking. In fact, reranking methods have been widely used in NLP tasks, such as machine translation, for a long time. They are typically applied in situations where training complex models is costly. In such cases, directly reranking the outputs allows for the incorporation of these complex models at a very low cost\footnote{Reranking methods can also help us explore what are known as model errors and search errors, although these issues are not often discussed in the context of LLMs. For example, suppose we have an old model and a new, more powerful model. We can use the new model to select the best output from the $N$-best list of the old model as the oracle output. The performance difference between the oracle output and the top-1 output of the original $N$-best list reflects the performance gain brought by the new model. If the performance gain is significant, we can say that the old model has more model errors. If the gain is small, it may indicate that the issue lies in search errors, as the best candidates were not found.}.

In the BoN sampling process, the LLM takes the input sequence $\mathbf{x}$ and generates $N$ different output sequences $\{\hat{\mathbf{y}}_1,...,\hat{\mathbf{y}}_N\}$:
\begin{eqnarray}
\{\hat{\mathbf{y}}_1,...,\hat{\mathbf{y}}_N\} & = & \mathop{\mathrm{argTopN}}_{\mathbf{y}} \left[ \Pr(\mathbf{y}|\mathbf{x}) \right],
\end{eqnarray}

\noindent where the $\mathrm{argTopN}$ operation returns the top-$N$ outputs that maximize the function $\Pr(\mathbf{y}|\mathbf{x})$. These outputs can be generated in a variety of ways, depending on the search algorithm used by the model (e.g., sampling or beam search). Once the $N$-best output candidates are generated, the reward model is used to evaluate and select the best one:
\begin{eqnarray}
\hat{\mathbf{y}}_{\mathrm{best}} & = & \argmax_{\hat{\mathbf{y}} \in \{\hat{\mathbf{y}}_1,\ldots,\hat{\mathbf{y}}_N\}} r(\mathbf{x},\hat{\mathbf{y}}).
\end{eqnarray}

It is worth noting that the result of BoN sampling is also influenced by the diversity of the $N$-best list. This is a common issue with most reranking methods. Ideally, the $N$-best output candidates should have relatively high quality but be sufficiently different from each other. In many text generation systems, the $N$-best outputs are very similar, often differing by just one or two words. The diversity issue is even more challenging in LLMs, as the $N$-best outputs generated by an LLM can differ in wording, yet their semantic meanings are often quite similar. In practice, one can adjust the model hyperparameters and/or use different LLMs to generate more diverse output candidates for reranking. Nevertheless, as with many practical systems, we need to make a trade-off between selecting high-quality candidates and ensuring sufficient variation in the generated outputs.

BoN sampling can be used for training LLMs as well. A closely related method is \mindex{rejection sampling}. In this method, we first select the ``best'' outputs from the $N$-best lists using the reward model, and then take these selected outputs to fine-tune the LLM. In this way, we can introduce human preferences into the training of LLMs using a much simpler approach compared to RLHF. Many LLMs have adopted rejection sampling for fine-tuning \cite{nakano-etal:2021webgpt,touvron-etal:2023llama2}.

%----------------------------------------------------------------------------------------
% NEW SECTION
%----------------------------------------------------------------------------------------

\section{Summary}

In this chapter, we have explored a range of techniques for aligning LLMs. In particular, we have discussed fine-tuning methods that enable LLMs to follow instructions and align them with human preferences. One of the benefits of fine-tuning LLMs is computational efficiency. Unlike pre-training based on large-scale neural network optimization, fine-tuning is a post-training step and so is less computationally expensive. Moreover, it is better suited to address problems that are not easily solved in pre-training, such as human value alignment. The widespread attention to AI alignment has also led to a surge of research papers on this topic, which makes it difficult to comprehensively cover all recent techniques. However, we have tried to provide a relatively detailed introduction to the fundamental approaches to alignment, such as instruction fine-tuning and RLHF.

While we have focused on LLM alignment techniques in this chapter, the term \textit{AI alignment} is a wide-ranging concept. It generally refers to the process of ensuring that the behavior of an AI system aligns with human values, goals, and expectations. The idea of AI alignment can be traced back to the early days of AI. A widely cited description of AI alignment comes from an article by the mathematician and computer scientist Norbert Wiener \cite{wiener:1960some}. The quote is as follows

\vspace{0.5em}
\begin{center}
\parbox{0.8\linewidth}{
\textit{If we use, to achieve our purposes, a mechanical agency with whose operation we cannot efficiently interfere ... we had better be quite sure that the purpose put into the machine is the purpose which we really desire.}
}
\end{center}
\vspace{0.5em}

At that time, alignment was not yet a central concern in AI research. But today, it greatly influences the design of various AI systems. For example, in robotics, alignment is critical to ensuring that autonomous robots safely interact with humans and their environments. In autonomous driving, cars must not only follow traffic laws but also make complex, real-time decisions that prioritize human safety, avoid accidents, and navigate ethical dilemmas.

In current AI research, alignment is often approached by developing a surrogate objective that is analogous to the true goal and steering the AI system towards this objective. However, designing the objective of AI alignment is very difficult. One reason is that human values are diverse and often context-dependent, making it difficult to distill them into a single, universally applicable objective function. Also, the complexity of real-world environments, where values and goals often conflict or evolve over time, further complicates alignment efforts. Even if we could define an appropriate objective, AI systems may find unintended ways to achieve it, leading to ``misaligned'' outcomes that still technically satisfy the objective but in a harmful or counterproductive way.

These challenges continue to motivate AI research towards more aligned systems, either through developing new mechanisms for perceiving the world or more efficient and generalizable methods to adapt these systems to given tasks. More importantly, as AI systems become more powerful and intelligent, especially given that recent advances in LLMs have shown remarkable capabilities in dealing with many challenging problems, the need for AI alignment has become more urgent. Researchers have started to be concerned with AI safety and warn the community that they need to develop and release AI systems with great caution to prevent these systems from being misaligned \cite{russell:2019human,bengio-etal:2024managing}.

% !Mode:: "TeX:UTF-8"
% !TEX encoding = UTF-8 Unicode

%----------------------------------------------------------------------------------------
% Natural Language Processing: Neural Networks and Large Language Models
%
% Copyright 2021-2026
% Tong Xiao (xiaotong@mail.neu.edu.cn) and Jingbo Zhu (zhujingbo@mail.neu.edu.cn)
%----------------------------------------------------------------------------------------

%----------------------------------------------------------------------------------------
%    CONFIGURATIONS
%----------------------------------------------------------------------------------------

%\setcounter{chapter}{10}

%----------------------------------------------------------------------------------------
%	CHAPTER 11
%----------------------------------------------------------------------------------------

\chapter{Inference}

\thispagestyle{chapterurl} % add urls to the header

Once we have pre-trained and fine-tuned an LLM, we can apply it to make predictions on new data. This process is called inference, in which the LLM computes the probabilities of possible outputs given an input, and selects the output that maximizes the probability. The inference problem is generally expressed in the following form:
\begin{eqnarray}
\hat{\mathbf{y}} & = & \argmax_{\mathbf{y}} \Pr(\mathbf{y}|\mathbf{x}), \label{eq:llm-argmax-base}
\end{eqnarray}

\noindent where $\mathbf{x}$ is the input sequence, $\mathbf{y}$ is a possible output sequence, and $\hat{\mathbf{y}}$ is the best output sequence.

This is perhaps one of the most widely used formulas in NLP, and dates back to the early days of speech recognition and machine translation systems based on probabilistic models. Although for some applications, such as predicting a single token using a very small language model, solving this optimization problem seems trivial, for most situations the computational challenges arise from both calculating $\Pr(\mathbf{y}|\mathbf{x})$ and performing the $\argmax$ operation. The problems we therefore wish to address in this chapter are: 1) computing the prediction probability efficiently given a trained LLM, and 2) devising an efficient approximate search for $\hat{\mathbf{y}}$.

At a high level, these are fundamental issues in artificial intelligence, which have been extensively studied. Thus, many well-established techniques can be directly applied, for example, one can use greedy search algorithms to implement an efficient inference system. On the other hand, model-specific optimizations, such as efficient attention models for Transformers, can be considered to further improve efficiency. But in many practical applications, we still need to make a trade-off between accuracy and efficiency, by carefully combining various techniques.

The importance of the inference problem in LLMs also lies in the fact that many application scenarios require processing extremely long sequences. Recent studies have found that incorporating additional prompts and contextual information, such as long chain-of-thought prompts, during inference can significantly improve the performance of LLMs. This provides a new approach to scaling LLMs: better results can be achieved by increasing the compute at inference time. For instance, through inference-time scaling, \citet{openai:2024learning}'s o1 and \citet{deepseek:2025r1}'s R1 systems have demonstrated impressive performance on complex reasoning and programming tasks. This, in turn, has encouraged the NLP field to focus more on the issue of efficient inference.

In this chapter, we will introduce the basic concepts and algorithms of LLM inference, including the prefill-decode framework, search (decoding) algorithms, and evaluation metrics of inference performance. We will then present methods for improving the efficiency of LLM inference, covering a range of techniques for speeding up the system and compressing the model. Finally, we will discuss inference-time scaling, which is considered an important application of inference optimization.

%----------------------------------------------------------------------------------------
% NEW SECTION
%----------------------------------------------------------------------------------------

\section{Prefilling and Decoding}

In this section, we present the prefill-decode framework, the most common approach for describing and implementing LLM inference processes. We first introduce the notation and background knowledge, and then describe the details of the framework, such as the decoding algorithms for LLM inference.

\subsection{Preliminaries}

Although we have described LLMs many times in this book, we begin by briefly defining the notation to facilitate the subsequent discussion, and to make this chapter self-contained.

\vspace{0.5em}
\begingroup
\setlength{\tabcolsep}{5pt}
\begin{tabular}{r l l}
$\mathbf{x}$: & \multicolumn{2}{p{0.75\textwidth}}{
The input token sequence. It can be viewed as the ``prompt'', which includes instructions, user inputs, and any additional context intended as input to the LLM. $\mathbf{x}$ comprises $m+1$ tokens, denoted by $x_0\cdots x_m$, where $x_0$ is the start symbol $\langle \mathrm{SOS} \rangle$.
\vspace{0.5em}
} \\
$\mathbf{y}$: & \multicolumn{2}{p{0.75\textwidth}}{
The output token sequence, also called the response to the input. $\mathbf{y}$ comprises $n$ tokens, denoted by $y_1\cdots y_n$.
\vspace{0.5em}
} \\
$\mathbf{y}_{<i}$: & \multicolumn{2}{p{0.75\textwidth}}{
The output tokens that precede position $i$, that is, $\mathbf{y}_{<i} = y_1\cdots y_{i-1}$.
\vspace{0.5em}
} \\
$\Pr(\mathbf{y}|\mathbf{x})$: & \multicolumn{2}{p{0.75\textwidth}}{
The probability of generating $\mathbf{y}$ given $\mathbf{x}$ using the LLM. If the LLM is parameterized by $\theta$, we can write it as $\Pr_{\theta}(\mathbf{y}|\mathbf{x})$.
\vspace{0.5em}
} \\
$[\mathbf{x},\mathbf{y}]$: & \multicolumn{2}{p{0.75\textwidth}}{
The concatenated token sequence of $\mathbf{x}$ and $\mathbf{y}$. That is, $[\mathbf{x},\mathbf{y}] = x_0 \cdots x_m y_1 \cdots y_n$. Occasionally, we use the notation $\mathrm{seq}_{\mathbf{x},\mathbf{y}}$ to represent $[\mathbf{x},\mathbf{y}]$.
\vspace{0.5em}
} \\
$\Pr([\mathbf{x},\mathbf{y}])$: & \multicolumn{2}{p{0.75\textwidth}}{
The probability of generating the token sequence $[\mathbf{x},\mathbf{y}]$ using the LLM.
\vspace{0.5em}
}
\end{tabular}
\endgroup
\vspace{0.5em}

As described in Eq. (\ref{eq:llm-argmax-base}), the goal of LLM inference is to maximize $\Pr(\mathbf{y}|\mathbf{x})$. Modeling this conditional probability is common in NLP. At first glance, it may appear to be a sequence-to-sequence modeling problem, where we transform a sequence into another using encoder-decoder models. However, we are not discussing sequence-to-sequence problems or encoder-decoder architectures. Instead, as discussed in earlier chapters, this modeling problem can be addressed using decoder-only models. To do this, we can interpret the log-scale probability $\log \Pr(\mathbf{y}|\mathbf{x})$ as the difference between $\log \Pr([\mathbf{x},\mathbf{y}])$ and $\log \Pr(\mathbf{x})$:
\begin{eqnarray}
\log \Pr(\mathbf{y}|\mathbf{x}) & = & \log \Pr([\mathbf{x},\mathbf{y}]) - \log \Pr(\mathbf{x}),
\end{eqnarray}

\noindent where $\log \Pr([\mathbf{x},\mathbf{y}])$ and $\log \Pr(\mathbf{x})$ can be obtained by running the LLM on the sequences $[\mathbf{x},\mathbf{y}]$ and $\mathbf{x}$, respectively. For example, we can calculate the probability of generating $\mathbf{x}$ using the chain rule:
\begin{eqnarray}
\log \Pr(\mathbf{x}) & = & \log \Pr(x_0\cdots x_m) \nonumber \\
                     & = & \log \big[ \Pr(x_0) \Pr(x_1|x_0) \cdots \Pr(x_m|x_0\cdots x_{m-1}) \big] \nonumber \\
                     & = & \underbrace{\log \Pr(x_0)}_{=0} + \sum_{j=1}^{m} \log \Pr(x_j|\mathbf{x}_{<j}) \nonumber \\
                     & = & \sum_{j=1}^{m} \log \Pr(x_j|\mathbf{x}_{<j}).
\end{eqnarray}

\noindent In other words, we calculate the token prediction log-probability at each position of $\mathbf{x}$, and sum all these log-probabilities. Note that since $x_0$ is the predefined start symbol, we omit its contribution here.

In common implementations of LLMs, however, we do not need to compute the log-probability of the input sequence, but use the LLM to directly compute the log-probability of the output sequence in the following form:
\begin{eqnarray}
\log \Pr(\mathbf{y}|\mathbf{x}) & = & \sum_{i=1}^{n} \log \Pr(y_i|\mathbf{x},\mathbf{y}_{<i}), \label{eq:llm-log-prob-y-x-sum}
\end{eqnarray}

\noindent where $\mathbf{x}$ and $\mathbf{y}_{<i}$ together form the context for predicting $y_i$. Following common notation in the literature, we write $\Pr(y_i|\mathbf{x},\mathbf{y}_{<i})$ as shorthand for $\Pr(y_i|[\mathbf{x},\mathbf{y}_{<i}])$.

\vspace{0.3em}
Now, we have two sub-problems in addressing the inference issue described in Eq. (\ref{eq:llm-argmax-base}):

\begin{itemize}
\item \vspace{0.3em} \textbf{Model Computation}: we model $\Pr(y_i|\mathbf{x},\mathbf{y}_{<i})$ and compute it in an efficient manner.
\item \vspace{0.3em} \textbf{Search}: we find the optimal (or sub-optimal) output sequence in terms of $\log \Pr(\mathbf{y}|\mathbf{x})$.
\end{itemize}
\vspace{0.3em}

The second sub-problem is a classic issue in NLP. We will show in Section~\ref{sec:llm-decoding-algorithms} that there are several well-studied algorithms that can be applied to efficiently search the space of possible output sequences. The first sub-problem requires a language model to produce a distribution over a vocabulary $V$ given a sequence of context tokens. We can do this by training a Transformer decoder, which outputs the distribution:
\begin{eqnarray}
\Pr(\cdot|\mathbf{x},\mathbf{y}_{<i}) & = & \mathrm{Softmax}(\mathbf{H} \mathbf{W}^{o})_{m+i} \label{eq:llm-token-predic-prob-softmax} \\
\mathbf{H} & = & \mathrm{Dec}([\mathbf{x},\mathbf{y}_{<i}]).
\end{eqnarray}

\noindent Here $\mathrm{Dec}(\cdot)$ produces a sequence of representations, each corresponding to a position of the input sequence. So, if we input $[\mathbf{x},\mathbf{y}_{<i}]$ to the LLM, $\mathbf{H}$ is an $i' \times d$ matrix, where $d$ is the dimensionality of each representation, and $i' = m+i$ is the number of context tokens. We can then use a Softmax layer to transform these representations into distributions of tokens. $\mathbf{W}^{o} \in \mathbb{R}^{d \times |V|}$ is the linear mapping matrix of the Softmax layer, and $\mathbf{H} \mathbf{W}^{o}$ transforms the $d$-dimensional representations in $\mathbf{H}$ into the $|V|$-dimensional representations. The use of the subscript $m+i$ indicates that the Softmax operation is applied only to the representation at position $m+i$. See Figure~\ref{fig:llm-transformer-decoder-achitecture} for an illustration of this architecture.

\begin{figure}[!t]
\centering
\input{./Chapter11/Figures/figure-llm-decoder}
\caption{The decoder-only architecture for LLMs. The decoder consists of an embedding layer and a stack of Transformer layers. In each Transformer layer, the input passes through a self-attention module followed by an FFN. The output of the decoder is a sequence of representations that are taken as input to a Softmax layer, which generates a distribution of tokens for each position.}
\label{fig:llm-transformer-decoder-achitecture}
\end{figure}

$\mathrm{Dec}(\cdot)$ is a Transformer decoding network that consists of an embedding network and a number of stacked self-attention and FFN networks. We will not discuss Transformers in detail here, as readers can easily learn about these models from the literature. However, it is worth pointing out that the difficulty of inference is in part from the use of the self-attention mechanism in Transformers. Recall that a general form of single-head self-attention is given by:
\begin{eqnarray}
\mathrm{Att}_{\mathrm{qkv}}(\mathbf{q}_{i'},\mathbf{K},\mathbf{V}) & = & \mathrm{Softmax}(\frac{\mathbf{q}_{i'} \mathbf{K}^{\top}}{\sqrt{d}}) \mathbf{V}, \label{eq:llm-single-head-qkv-attention}
\end{eqnarray}

\noindent where $\mathbf{q}_{i'} \in \mathbb{R}^{d}$ is the query at the position $i'$ (i.e., position of $y_i$ ), and $\mathbf{K}$ and $\mathbf{V} \in \mathbb{R}^{i' \times d}$ are the keys and values up to $i'$, respectively.

At each step during inference, we call the self-attention function $\mathrm{Att}_{\mathrm{qkv}}(\cdot)$, followed by an FFN, to generate a $d$-dimensional representation that integrates information from both the current token and its left context. This process is repeated through $L$ layers of self-attention and FFN, forming a stack of Transformer layers. The output of the $L$-th layer in this stack is the final representation.

Each time, the model attends position $i'$ to all previous positions, which results in $2 i'$ vector products ($i'$ times for $\mathbf{q}_{i'} \mathbf{K}^{\top}$ and $i'$ times for the product of $\mathrm{Softmax}(\frac{\mathbf{q}_{i'} \mathbf{K}^{\top}}{\sqrt{d}})$ and $\mathbf{V}$). Hence, generating a sequence of length $\mathrm{len}$ has a time complexity of $O(L \times d \times \mathrm{len}^2)$ for the self-attention network. Clearly, the inference of this model is slow for long sequences due to its quadratic time complexity with respect to sequence length. Therefore, many improvements to Transformers and alternative models have focused on efficient methods that are faster than this quadratic time complexity, such as sparse attention mechanisms and linear-time models. A detailed discussion of efficient Transformers can be found in the previous chapters, and this section will focus on the standard Transformer architecture.

Note that in self-attention, the queries, keys, and values of a layer are linear mappings from the same input (i.e., the output of the previous layer). Once a new key-value pair is generated, it is repeatedly used in subsequent inference steps. Rather than regenerating these key-value pairs during inference, a more desirable way is to store them in a structure, called the \mindex{key-value cache}, or the \mindex{KV cache}. Thus, $(\mathbf{K},\mathbf{V})$ can naturally be viewed as a KV cache. This cache is updated as follows:
\begin{eqnarray}
\mathbf{K} & = & \mathrm{Append}(\mathbf{K}, \mathbf{k}_{i'}), \label{eq:kv-cache-update-k} \\
\mathbf{V} & = & \mathrm{Append}(\mathbf{V}, \mathbf{v}_{i'}), \label{eq:kv-cache-update-v}
\end{eqnarray}

\noindent where $(\mathbf{k}_{i'}, \mathbf{v}_{i'})$ is the newly generated key-value pair at position $i'$, and $\mathrm{Append}(\mathbf{a},\mathbf{b})$ denotes a function that appends a row vector $\mathbf{b}$ to a matrix $\mathbf{a}$. Figure~\ref{fig:kv-cache-llm} shows how a Transformer decoder works with a KV cache.

\begin{figure}[!t]
\centering
\input{./Chapter11/Figures/figure-llm-kv-cache}
\caption{Illustration of the KV cache. We update the KV cache at a position, perform the attention operation, and then move to the next position to repeat the process.}
\label{fig:kv-cache-llm}
\end{figure}

Finally, the process of computing $\log \Pr(\mathbf{y}|\mathbf{x})$ is summarized as follows:

\begin{enumerate}
\item \vspace{0.5em} We concatenate $\mathbf{x}$ and $\mathbf{y}$ into a sequence $[\mathbf{x},\mathbf{y}]$. For each position $i'$ of this sequence, we perform the following steps.
\begin{enumerate}
\item \vspace{0.3em} We compute the embedding of the token at position $i'$, and feed the resulting embedding as an initial representation into the stack of Transformer layers.
\item \vspace{0.3em} In each Transformer layer, we pass the input representation through the self-attention network first and then through an FFN. In the self-attention network, the input representation is transformed into  $\mathbf{q}_{i'}$, $\mathbf{k}_{i'}$, and $\mathbf{v}_{i'}$. Then, we update the KV cache $(\mathbf{K},\mathbf{V})$ using $\mathbf{k}_{i'}$ and $\mathbf{v}_{i'}$ (see Eqs. (\ref{eq:kv-cache-update-k}-\ref{eq:kv-cache-update-v})). Then, we compute the output of the attention model by attending $\mathbf{q}_{i'}$ to $(\mathbf{K},\mathbf{V})$ (see Eq. (\ref{eq:llm-single-head-qkv-attention})).
\item If $i' > m$ (i.e., $i = i' - m \ge 1$), we take the output of the Transformer stack and compute the token prediction probability $\Pr(y_i|\mathbf{x},\mathbf{y}_{<i})$ via the Softmax layer (see Eq. (\ref{eq:llm-token-predic-prob-softmax})).
\end{enumerate}
\item When reaching the end of the sequence, we obtain $\log \Pr(\mathbf{y}|\mathbf{x})$ by summing $\log \Pr(y_i|\mathbf{x},\mathbf{y}_{<i})$ over $i \in [1,n]$ (see Eq. (\ref{eq:llm-log-prob-y-x-sum})).
\end{enumerate}
\vspace{0.5em}

\subsection{A Two-phase Framework}
\label{sec:llm-inference-prefilling-and-decoding}

As we have seen, language modeling is a standard autoregressive process, where each token is generated one at a time, conditioned on the previous tokens. For Transformers, this requires the model to maintain a KV cache that stores past representations, and allows the newly generated representation to attend to this KV cache. When viewed from the perspective of KV-cache computation, LLM inference naturally divides into two phases:

\begin{itemize}
\item \vspace{0.5em} \mindex{Prefilling}. The prefilling phase initializes the KV cache for the input sequence $\mathbf{x}$.  It is called prefilling because the model prepares and stores the key-value pairs for each token in the input before the actual token generation begins. The process of prefilling in an LLM can be expressed as:
    \begin{eqnarray}
    \mathrm{cache} & = & \mathrm{Dec}_{\mathrm{kv}}(\mathbf{x}), \label{eq:prefilling-kv-cache}
    \end{eqnarray}

    \noindent where $\mathrm{Dec}_{\mathrm{kv}}(\cdot)$ is the decoding network (i.e., the same as $\mathrm{Dec}(\cdot)$), but it returns the KV cache in self-attention instead of the output representations. $\mathrm{cache}$ is a list, given by:
    \begin{eqnarray}
    \mathrm{cache} & = & \{\mathrm{cache}^1,\cdots,\mathrm{cache}^L\},
    \end{eqnarray}

    \noindent where $\mathrm{cache}^l$ represents the key-value pairs for the $l$-th layer.

\item \vspace{0.3em} \mindex{Decoding}. The decoding phase continues generating tokens based on the KV cache, as illustrated in Figure~\ref{fig:kv-cache-llm}. When a new token is input into the decoder, we update the KV cache in each layer by adding the new key-value pair. The updated cache is then used for self-attention computation.  The token generation stops when some stopping criterion is met, such as when the generated token is the end symbol. The goal of decoding is to find the best predicted sequence, which is given by:
    \begin{eqnarray}
    \hat{\mathbf{y}} & = & \argmax_{\mathbf{y}} \Pr(\mathbf{y}|\mathrm{cache}). \label{eq:llm-argmax-kv-cache}
    \end{eqnarray}

    \noindent Although the probabilistic model is still conditioned on $\mathbf{x}$, the decoding process operates on the KV cache computed from $\mathbf{x}$. Thus we use $\Pr(\mathbf{y}|\mathrm{cache})$ instead of $\Pr(\mathbf{y}|\mathbf{x})$ to emphasize reliance on the KV cache.
\end{itemize}
\vspace{0.5em}

The prefilling and decoding processes are illustrated in Figure~\ref{fig:prefilling-and-decoding}. Note that both processes share the same underlying autoregressive model. However, as shown in Table~\ref{tab:prefilling-vs-decoding}, they differ in several aspects, which lead to very different implementations in practice.

In essence, although prefilling is still based on autoregressive token prediction, it can be considered an encoding process. This is because our goal is not to generate tokens, but to build a context representation (i.e., the KV cache) for the subsequent steps in the decoding phase. In this sense, it shares some similarities with encoder-based models such as BERT, where we encode the input sequence into a sequence of contextualized token representations. On the other hand, unlike BERT which generates bidirectional sequence representations, prefilling is based on standard language modeling tasks, and is thus unidirectional. Note that, since the entire sequence $\mathbf{x}$ is input into the model all at once, all queries can be packed together and the self-attention operation is performed on $\mathbf{x}$ in parallel. Let $\mathbf{Q}$ be the queries that are packed into one matrix. The self-attention model in prefilling can be defined as:
\begin{eqnarray}
\mathrm{Att}_{\mathrm{qkv}}(\mathbf{Q},\mathbf{K},\mathbf{V}) & = & \mathrm{Softmax}(\frac{\mathbf{Q} \mathbf{K}^{\top}}{\sqrt{d}} + \mathbf{Mask}) \mathbf{V}, \label{eq:llm-prefilling-qkv-attention}
\end{eqnarray}

\begin{figure}[!t]
\centering
\input{./Chapter11/Figures/figure-filling-and-decoding}
\vspace{-0.5cm} \caption{Illustration of the prefilling and decoding processes. In prefilling, the entire input sequence is processed together and the KV cache is filled. In decoding, the LLM generates the output sequence step by step based on the prefilled KV cache.}
\label{fig:prefilling-and-decoding}
\end{figure}

\noindent where $\mathbf{Q}, \mathbf{K}, \mathbf{V} \in \mathbb{R}^{(m+1) \times d}$. $\mathbf{Mask} \in \mathbb{R}^{(m+1) \times (m+1)}$ is a mask that ensures that each token only attends to itself and the tokens that precede it in the sequence. It is represented by setting the values in the mask corresponding to future tokens to a large negative number, for example, for the query $\mathbf{q}_{i}$ and the key $\mathbf{k}_{j}$, we set the value of the entry $(i,j)$ to $-\infty$ if $i < j$. One advantage of processing the sequence with a single self-attention computation is that we can make better use of the parallel computing capabilities of modern GPUs, and so speed up prefilling. In general, the prefilling process is considered compute-bound. This is because merging multiple computational operations into one operation reduces the number of data transfers and the performance bottleneck usually comes from the computational capacity rather than memory bandwidth.

Decoding is a standard left-to-right text generation process. The token sequence is generated autoregressively by predicting one token at a time based on the KV cache. Each newly generated token needs to attend to all previous tokens, following Eq. (\ref{eq:llm-single-head-qkv-attention}). Therefore, the decoding process is memory-bound due to its frequent access to the KV cache. The cost of decoding grows significantly as more tokens are generated. In most cases, decoding is computationally more expensive than prefilling. Note that this is not just because, in decoding, the LLM generates tokens one by one and repeatedly updates the KV cache. As we will see in the following subsection, we may need to explore multiple different token sequences during decoding, which makes the problem more complex and increases its cost further.

\begin{table}[!t]
\centering
\begingroup
\renewcommand{\arraystretch}{1.2}
\begin{tabular}{r | L{0.30\textwidth} | L{0.30\textwidth}}
& Prefilling & Decoding  \\  \hline
Goal & Set up initial context $\mathbf{x}$. & Continue generating tokens $\mathbf{y}$ after the initial input. \\ \hline
Token Visibility & Tokens in $\mathbf{x}$ are presented all at once. & Tokens in $\mathbf{y}$ are generated sequentially, that is, predicting a token depends on previously generated tokens. \\ \hline
Context Use & Build the context or encoded representation of the input. & Use the cached key-value pairs (from prefilling) to generate further tokens. \\ \hline
Resource Limitation & Compute-bound & Memory-bound \\ \hline
Computational Cost & High & Very High
\end{tabular}
\endgroup
\caption{Prefilling vs Decoding.}
\label{tab:prefilling-vs-decoding}
\end{table}

\subsection{Decoding Algorithms}
\label{sec:llm-decoding-algorithms}

So far our discussion of LLM inference has primarily focused on the model computation problem, that is, how to compute $\Pr(\mathbf{y}|\mathbf{x})$. Now we turn to the discussion of the search problem. The problem can be stated as: given an LLM $\Pr(\mathbf{y}|\mathbf{x})$, how do we efficiently generate a high-quality output sequence $\hat{\mathbf{y}}$ given the input sequence $\mathbf{x}$? Naively, we can consider all of the output sequences, compute the prediction probability for each, and then select the output sequence having the highest probability. This method can guarantee the globally optimal solution, but direct exhaustive search is impractical for LLMs as the number of possible output sequences grows exponentially with the length of $\mathbf{y}$.

In practice, various heuristic search algorithms, such as greedy search and sampling-based search, are commonly employed to approximate the solution. Each of these methods offers trade-offs between search quality and computational efficiency. The search problem, therefore, becomes a balancing act between exploration and exploitation, where the goal is to find an efficient strategy that produces high-quality outputs without exploring the entire space.

Before giving a more detailed discussion of these methods, let us first informally define what a search space is and how it is represented. In LLM inference, we define a hypothesis as a tuple of input and output sequences. Since $\mathbf{x}$ is fixed during inference, we can simply consider each hypothesis as an output sequence. The search space, denoted by $\mathcal{Y}$, is then the set of all possible hypotheses (i.e., output sequences) that the model can generate. The search problem for LLM inference can be re-expressed as:
\begin{eqnarray}
\hat{\mathbf{y}} & = & \argmax_{\mathbf{y} \in {\color{red} \mathcal{Y}}} \Pr(\mathbf{y}|\mathbf{x}). \label{eq:llm-general-decoding-objective}
\end{eqnarray}

In NLP, $\mathcal{Y}$ is commonly represented in a tree data structure to facilitate search. Figure~\ref{fig:llm-decoding-search-tree} shows an example of the search tree resulting from a small vocabulary. In this example, a node represents a prefix sequence that can be shared by many sequences. The search begins at the root of the tree, which can be regarded as the beginning of all sequences that can be generated\footnote{Here, since the predictions in LLMs are based on $\mathbf{x}$, we can think of the root as a representation of $\mathbf{x}$.}. Each child node extends the prefix of its parent node by adding one token from the vocabulary to the sequence, along with the probability of predicting the token given the prefix. This process continues as each node further branches out into additional child nodes, each representing a new possible extension of the sequence with another token. The search tree thus grows deeper and wider, representing an exponentially growing number of potential sequences as more tokens are appended. This structure allows us to efficiently traverse through possible sequences, evaluating each in terms of the log-probability accumulated over the path from the root to that node. For example, in Figure~\ref{fig:llm-decoding-search-tree}, the path from the root to the node 17 corresponds to the output sequence ``\textit{Cats are playful.}''. The prediction log-probability $\log \Pr(\mathbf{y}|\mathbf{x})$  is the sum of the log-probabilities of all the nodes on this path.

\begin{figure}[!t]
\centering
\input{./Chapter11/Figures/figure-search-tree}
\vspace{-0.5cm} \caption{A search tree for decoding. At each node, we expand the tree by considering all possible tokens, each leading to a new node representing a potential continuation of the text. Here we highlight a path through nodes 0, 3, 9, 11, and 17. The path represents the output sequence ``\textit{cats are playful.}'', whose log-probability can be computed by accumulating the log-probabilities of these nodes.}
\label{fig:llm-decoding-search-tree}
\end{figure}

In general, the search tree is organized into levels, where each level consists of all nodes at the same depth. Thus, the tree can be viewed as being expanded level by level, where each level corresponds to one decoding step. As the search progresses, new tokens are appended to these sequences, expanding them incrementally.

Let $Y_i$ be the set of the sequences that the LLM generates at step $i$. $Y_i$ can be obtained by expanding each sequence in $Y_{i-1}$ with all possible next tokens in the vocabulary $V$, given in the following recursive form:
\begin{eqnarray}
Y_i & = & Y_{i-1} \times V,
\end{eqnarray}

\noindent where $Y_{i-1} \times V$ denotes the Cartesian product of $Y_{i-1}$ and $V$ (i.e., each sequence in $Y_{i-1}$ is concatenated with each token in $V$). Note that if a sequence in $Y_{i-1}$ is complete (e.g., ending with the $\langle \mathrm{EOS} \rangle$ token), it will not be expanded any further. Let $\Psi(Y_i)$ be the set of all complete sequences in $Y_i$. Then, the search space can be expressed as:
\begin{eqnarray}
\mathcal{Y} & = & \Psi(Y_1) \bigcup \Psi(Y_2) \bigcup \cdots \bigcup \Psi(Y_{n_{\mathrm{max}}}),
\end{eqnarray}

\noindent where $n_{\mathrm{max}}$ is the maximum length of a sequence.

Most decoding algorithms follow this level-by-level search process. However, $\mathcal{Y}$ consists of an exponentially large number of sequences, and a direct search
in such a vast space is computationally infeasible. Therefore, practical decoding algorithms often rely on strategies to prune the search space and avoid exploring low-quality sequences. For example, at each decoding step, $Y_i$ can be obtained in the following way:
\begin{eqnarray}
Y_i & = & \mathrm{Prune}(Y_{i-1} \times V),
\end{eqnarray}

\noindent where $\mathrm{Prune}(\cdot)$ is a function that selectively removes sequences less likely to result in high-quality outcomes. In general, we expect that $|Y_i| \ll |Y_{i-1}| \cdot |V|$. Thus we can drastically reduce the number of sequences under consideration at each step, ensuring that the computational load does not grow exponentially with the sequence length.

Next, we will introduce these decoding algorithms. Some of them have already been discussed in sequence-to-sequence models, while others are more commonly used in LLMs.

\subsubsection{Greedy Decoding}

Greedy search (or greedy decoding) is one of the most widely used decoding methods in NLP, particularly in text generation tasks like machine translation. The idea behind greedy search is straightforward: at each step in generation, it selects the next token that has the highest prediction probability. For each sequence $\mathbf{y} = y_1\cdots y_i \in Y_{i-1} \times V$, we can evaluate it using
$\log \Pr(\mathbf{y}|\mathbf{x})$. This log-probability can be easily computed by noting that:
\begin{eqnarray}
\log \Pr(\mathbf{y}|\mathbf{x}) & = & \log \Pr(y_1\cdots y_i|\mathbf{x}) \nonumber \\
& = & \underbrace{\log \Pr(\mathbf{y}_{<i}|\mathbf{x})}_{\text{accumulated up to the parent node}} + \underbrace{\log \Pr(y_i|\mathbf{x},\mathbf{y}_{<i})}_{\text{newly computed for the current node}},
\end{eqnarray}

\noindent Here the first term is the sum of the log-probabilities of the path from the root to the parent node, which has been computed in the previous decoding steps. At step $i$, we only need to compute the second term which is the standard token prediction log-probability produced by the LLM.

The ``best'' token at step $i$ is then chosen as:
\begin{eqnarray}
y_i^{\mathrm{top}1} & = & \argmax_{y_i \in V} \log \Pr(y_1\cdots y_i|\mathbf{x}) \nonumber \\
                    & = & \argmax_{y_i \in V} \big[ \underbrace{\log \Pr(\mathbf{y}_{<i}|\mathbf{x})}_{\text{fixed wrt. $y_i$}} + \log \Pr(y_i|\mathbf{x},\mathbf{y}_{<i}) \big] \nonumber \\
                    & = & \argmax_{y_i \in V} \log \Pr(y_i|\mathbf{x},\mathbf{y}_{<i}).
\end{eqnarray}

Thus, the ``best'' sequence generated up to step $i$ is given by:
\begin{eqnarray}
\mathbf{y}^{\mathrm{top}1} & = & y_1\cdots y_{i-1} y_i^{\mathrm{top}1}.
\end{eqnarray}

Finally, $Y_i$ contains only this sequence:
\begin{eqnarray}
Y_i & = & \{\mathbf{y}^{\mathrm{top}1}\}.
\end{eqnarray}

The greedy choice in one decoding step is illustrated in Figure~\ref{fig:llm-decoding-methods} (a). Greedy search offers computational efficiency and is simple to implement. Its primary disadvantage, however, lies in its suboptimal nature --- high-quality sequences are likely pruned at early stages of decoding. Therefore, greedy search is appealing for tasks that demand speed and simplicity. For tasks that require better search results, alternative strategies such as beam search, which explores multiple potential paths simultaneously, are preferable.

\begin{figure}[!t]
\centering
\input{./Chapter11/Figures/figure-decoding-methods-one-step}
\vspace{-0.5cm} \caption{Illustrations of greedy decoding, beam decoding, top-$k$ decoding and top-$p$ decoding methods (in one decoding step).}
\label{fig:llm-decoding-methods}
\end{figure}

\subsubsection{Beam Decoding}

Beam search (or beam decoding) is a natural extension of greedy search. Instead of retaining only the single most probable sequence at each step, beam search maintains a fixed number of the best candidate sequences, known as the ``beam width'' $K$. See Figure~\ref{fig:llm-decoding-methods} (b) for an illustration.

Unlike greedy search, which only expands one parent node, beam search considers extensions from all $K$ sequences stored in $Y_{i-1}$. At step $i$, we evaluate all possible continuations by computing the full sequence log-probabilities for the set $Y_{i-1} \times V$. The top $K$ sequences with the highest accumulated probabilities are then selected to form $Y_i$:
\begin{eqnarray}
Y_i & = & \{\mathbf{y}^{\mathrm{top}1},\cdots,\mathbf{y}^{\mathrm{top}K}\} \nonumber \\
    & = & \mathop{\mathrm{argTopK}}_{\mathbf{y} \in Y_{i-1} \times V}\ \log \Pr(\mathbf{y}|\mathbf{x}), \label{eq:llm-beam-search-k-best}
\end{eqnarray}

\noindent where $\mathrm{argTopK}$ is a function that ranks the probabilities of all evaluated sequences and returns the top $K$ candidates. Adjusting the beam width $K$ allows us to balance search efficiency and accuracy. While a larger beam explores a wider search space, the performance gains often diminish. In practical LLM inference, a relatively small beam width, such as $K=2$ to $K=5$, is usually sufficient to achieve high-quality generation. Note that while beam search has historically played an important role in sequence generation tasks such as machine translation, sampling-based decoding methods are now more commonly used in open-ended LLM text generation because they produce more diverse and flexible outputs.

\subsubsection{Sampling-based Decoding}

Both greedy and beam search generate deterministic outputs, that is, given an LLM, the output of the model will always be the same every time it processes the same input. The deterministic nature of greedy and beam search ensures predictability and reliability in applications where consistent outcomes are critical, such as in formal document generation, where varying outputs could cause confusion or errors. On the other hand, one disadvantage of these methods is the lack of diversity and flexibility. For example, in creative tasks like story generation or conversational agents, generic or repetitive outputs generated by deterministic systems are often less engaging.

To introduce variation into LLM outputs, we can use sampling-based decoding methods. There are two commonly used methods.

\begin{itemize}
\item \vspace{0.5em} \textbf{Top-$k$ Sampling}. This method selects the next token from the top-$k$ most likely candidates at each step of the generation process~\cite{fan-etal:2018hierarchical}. Let $\overline{V}_i$ be the selection pool for top-$k$ sampling. We can define it as:
    \begin{eqnarray}
    \overline{V}_i & = & \{y_i^{\mathrm{top}1},\cdots,y_i^{\mathrm{top}k}\},
    \end{eqnarray}

    \noindent where $\{y_i^{\mathrm{top}1},\cdots,y_i^{\mathrm{top}k}\}$ are the top-$k$ tokens selected based on their prediction probabilities (see Eq. (\ref{eq:llm-beam-search-k-best})). Once the selection pool is determined, we renormalize the probability distribution over $\overline{V}_i$. One of the simplest ways to do this is to renormalize their probabilities:
    \begin{eqnarray}
    \overline{\Pr}(y_i|\mathbf{x},\mathbf{y}_{<i}) & = & \frac{\Pr(y_i|\mathbf{x},\mathbf{y}_{<i})}{\sum_{y_j \in \overline{V}_i} \Pr(y_j|\mathbf{x},\mathbf{y}_{<i})}. \label{eq:llm-topk-sampling-renormalization}
    \end{eqnarray}

    Alternatively, we can calculate the distribution by using the Softmax function:
    \begin{eqnarray}
    \overline{\Pr}(y_i|\mathbf{x},\mathbf{y}_{<i}) & = & \frac{\exp(u_{y_i})}{\sum_{y_j \in \overline{V}_i} \exp(u_{y_j})},
    \end{eqnarray}

    \noindent where $u_{y_i}$ is the logit for token $y_i$. Then, we sample a token $\bar{y}_i$ from this distribution:
    \begin{eqnarray}
    \bar{y}_i & \sim & \overline{\Pr}(y_i|\mathbf{x},\mathbf{y}_{<i}).
    \end{eqnarray}

    The corresponding sequence is $\bar{\mathbf{y}} = y_1\cdots y_{i-1} \bar{y}_i$, and $Y_i$ is given by:
    \begin{eqnarray}
    Y_i & = & \{\bar{\mathbf{y}}\}. \label{eq:llm-topk-sampling-yi-final}
    \end{eqnarray}

\item \vspace{0.3em} \textbf{Top-$p$ Sampling}. This sampling method, also known as \mindex{nucleus sampling}, follows a procedure similar to that of top-$k$ sampling. Instead of drawing from a fixed size candidate pool, it selects the next token from the smallest set of tokens that together have a cumulative probability higher than a predefined threshold $p$~\cite{holtzman-etal:2020the}. In this way we prevent the prediction from choosing from low-probability tokens in the long tail that could lead to incoherent or nonsensical outputs. To obtain the candidate pool in the top-$p$ sampling method, we can sort all tokens by their predicted probabilities. Then, starting with the token with the highest probability, we continue to add tokens to the candidate pool until the cumulative probability of the tokens in the pool reaches or exceeds $p$ (we denote the size of the candidate pool at this point as $k_p$). The candidate pool can then be expressed as:
    \begin{eqnarray}
    \overline{V}_i & = & \{y_i^{\mathrm{top}1},\cdots,y_i^{\mathrm{top}{k_p}}\}.
    \end{eqnarray}

    The subsequent steps, such as the renormalization of the distribution and sampling, are the same as in the top-$k$ sampling method (see Eqs.(\ref{eq:llm-topk-sampling-renormalization}-\ref{eq:llm-topk-sampling-yi-final})).
\end{itemize}
\vspace{0.5em}

See Figure~\ref{fig:llm-decoding-methods} (c-d) for illustrations of the top-$k$ and top-$p$ sampling methods. By limiting the choices to a smaller set of high-probability tokens, these methods strike a balance between randomness and coherence. They allow for more diverse outputs while still maintaining a reasonable level of relevance and fluency. However, the value of $k$ or $p$ must be carefully chosen: if $k$ or $p$ is too small, the output may still be overly deterministic (more like greedy decoding), and if $k$ or $p$ is too large, the LLM might produce degenerate outputs.

In order to further control the randomness of the token selection process, the renormalized distribution $\overline{\Pr}(\cdot)$ is typically obtained by using the Softmax function with the temperature parameter, given by:
\begin{eqnarray}
\overline{\Pr}(y_i|\mathbf{x},\mathbf{y}_{<i}) & = & \frac{\exp(u_{y_i}/\beta)}{\sum_{y_j \in \overline{V}_i} \exp(u_{y_j}/\beta)},
\end{eqnarray}

\noindent Here $\beta$ is a temperature parameter that controls the sharpness of the probability distribution derived from logits. In Figure~\ref{fig:llm-softmax-with-temperature}, we show simple examples involving distributions generated by the above function with different temperatures. When the temperature is set to a higher  value, the resulting probability distribution becomes more uniform, as the differences between the logits are diminished. This means that each token in the candidate pool has a more equal chance of being selected, leading to greater diversity in the generated output. By contrast, when the temperature is set to a lower value, the distribution becomes sharper, making the high-probability tokens even more likely to be chosen, which often results in more deterministic outputs. For example, if we set $p$ to $1$ and $\beta$ to a very small number (approaching zero), the top-$p$ sampling method will become equivalent to the greedy search method.

\begin{figure}[!t]
\centering
\input{./Chapter11/Figures/figure-softmax-with-different-temperatures}
\vspace{-0.5cm} \caption{Histogram estimates of the distributions generated by the Softmax function with different values of the temperature parameter $\beta$.}
\label{fig:llm-softmax-with-temperature}
\end{figure}

\subsubsection{Decoding with Penalty Terms}

One common improvement to decoding methods in text generation is to modify the search objective. For example, one can replace maximum a posteriori (MAP) decoding with minimum Bayes risk (MBR) decoding~\cite{kumar-etal:2004minimum}, where the focus shifts from selecting the single most probable output to choosing an output that minimizes the expected risk over a distribution of possible outputs. Here we explore methods that incorporate penalty terms into decoding. These methods offer a simple but effective way to make decoding more controllable.

Recall from Eq. (\ref{eq:llm-general-decoding-objective}) that the goal of decoding is to maximize the likelihood of the output sequence. With penalty terms, the objective is extended to include additional factors that penalize certain behaviors in the generated text. A general form of the new objective is given by:
\begin{eqnarray}
\hat{\mathbf{y}} & = & \argmax_{\mathbf{y} \in \mathcal{Y}} \big[ \log \Pr(\mathbf{y}|\mathbf{x}) - \lambda \cdot \mathrm{Penalty}(\mathbf{x},\mathbf{y}) \big], \label{eq:llm-decoding-objective-with-penalty}
\end{eqnarray}

\noindent where $\mathrm{Penalty}(\mathbf{x},\mathbf{y})$ is a function that quantifies the degree to which the generated sequence $\mathbf{y}$ violates certain constraints or exhibits undesirable behaviors given the input $\mathbf{x}$. The design of $\mathrm{Penalty}(\cdot)$ is very flexible, thus allowing us to incorporate a wide range of constraints or prior knowledge into it. Below, we present some common types of penalty functions.

\begin{itemize}
\item \vspace{0.5em} \textbf{Repetition Penalty}. A repetition penalty discourages the model from generating repetitive or redundant text. The penalty function might measure the frequency of repeated tokens or phrases in the generated sequence and impose a penalty proportional to their occurrence.
\item \vspace{0.3em} \textbf{Length Penalty}. A length penalty ensures that the generated sequence adheres to a desired length. For example, in text summarization tasks, the penalty function could penalize outputs that are too short or too long.
\item \vspace{0.3em} \textbf{Diversity Penalty}. A diversity penalty promotes variation in the generated text. For example, in beam search, we can measure the similarity between generated hypotheses, and encourage the model to explore different hypotheses.
\item \vspace{0.3em} \textbf{Constraint-based Penalty}. A constraint-based penalty enforces specific constraints related to the content or style of the generated text. For example, in machine translation, the penalty function could penalize outputs that deviate from a desired tone or terminology.
\end{itemize}
\vspace{0.5em}

In general, we can consider $\mathrm{Penalty}(\mathbf{x},\mathbf{y})$ as a function that defines the cost assigned to the generated sequence $\mathbf{y}$ given the input sequence $\mathbf{x}$. Alternatively, this function can be defined to assess the hidden states of an LLM when generating $\mathbf{y}$. For example, \citet{su-etal:2022contrastive} develop a penalty term that calculates the maximum distance between the representation of the predicted token and the representations of the previously generated tokens. Therefore, the search objective will penalize degenerate outputs, such as text with many repetitions.

The method described in Eq. (\ref{eq:llm-decoding-objective-with-penalty}) is general and can be easily adapted to different search algorithms. For example, in greedy search, we can keep the single sequence that maximizes $\log \Pr(\mathbf{y}|\mathbf{x}) - \lambda \cdot \mathrm{Penalty}(\mathbf{x},\mathbf{y})$ at each decoding step; in sampling-based search, we can rank and select the top-ranked sequences based on $\log \Pr(\mathbf{y}|\mathbf{x}) - \lambda \cdot \mathrm{Penalty}(\mathbf{x},\mathbf{y})$ to form the candidate pool.

\subsubsection{Speculative Decoding}

\mindex{Speculative decoding} stems from the concept of \mindex{speculative execution}, where a system makes educated guesses about future actions and performs them in advance. If the guess is correct, the results are immediately available, which speeds up processing. In the case of LLM inference, suppose we have two models. One is a smaller, faster model (called the draft model), and the other is the full, stronger model (called the verification model). These two models represent two ends of the spectrum in LLM inference: the draft model is efficient but not very accurate; the verification model is usually the one we want to run, but it is very slow. Given a prefix, we first use the draft model to speculatively predict a sequence of likely future tokens. This is a standard autoregressive decoding process, but it is still fast in practice due to the high efficiency of the draft model. Then, the verification model evaluates the speculated tokens in parallel. It checks whether the predicted tokens can be accepted under the verification model. Note that, since we can deal with these tokens all at once, the verification can be done in a single step for all the tokens simultaneously, rather than in a token-by-token manner. If the speculated tokens are correct, they are accepted, and the process continues with the next set of tokens. If they are incorrect, the incorrect speculations are discarded, and the verification model is used to generate new tokens.

To be more specific, let us see the speculative decoding method presented in \citet{leviathan-etal:2023fast}'s work. In this method, the draft model is a small language model, denoted by $\Pr_q(y_i|\mathbf{x},\mathbf{y}_{<i})$, while the verification model is the full LLM, denoted by $\Pr_p(y_i|\mathbf{x},\mathbf{y}_{<i})$. The goal is to use the draft model, given a prefix, to autoregressively predict up to $\tau$ tokens. The verification model is then employed to generate the last token at the point where speculative predictions are rejected. Figure~\ref{fig:speculative-decoding-one-step} illustrates one step in this decoding process. For simplicity, we describe the greedy version of speculative decoding.

\begin{figure}[!t]
\centering
\input{./Chapter11/Figures/figure-speculative-decoding}
\vspace{-0.5cm} \caption{Illustration of one step of speculative decoding. The goal is to predict as many next tokens as possible using the draft model. There are four sub-steps. Given the context, we first use the draft model to predict the next $\tau$ tokens. Then, we evaluate these predictions in parallel using the verification model. Next, we determine the maximum number of predicted tokens that can be accepted. Finally, we use the evaluation model to predict a new token following these accepted tokens.}
\label{fig:speculative-decoding-one-step}
\end{figure}

The speculative decoding algorithm can be summarized as follows.

\begin{itemize}
\item \vspace{0.5em} Given the prefix $[\mathbf{x},\mathbf{y}_{\le i}]$, we use the draft model to predict the next $\tau$ consecutive tokens, denoted by $\{\hat{y}_{i+1},\cdots,\hat{y}_{i+\tau}\}$. This is a token-by-token generation process, given by:
    \begin{eqnarray}
    \hat{y}_{i+t} & = & \argmax_{y_{i+t}} \mathrm{Pr}_q(y_{i+t}|\mathbf{x},\mathbf{y}_{\le i},\hat{y}_{i+1}\cdots\hat{y}_{i+t-1}).
    \end{eqnarray}
\item \vspace{0.3em} We evaluate $\{\hat{y}_{i+1},\cdots,\hat{y}_{i+\tau}\}$ using the verification model, that is, we compute $\{\mathrm{Pr}_p(\hat{y}_{i+1}|\mathbf{x},\mathbf{y}_{\le i})$ $,\cdots,\mathrm{Pr}_p(\hat{y}_{i+\tau}|\mathbf{x},\mathbf{y}_{\le i},\hat{y}_{i+1}\cdots\hat{y}_{i+\tau-1})\}$. Note that we can compute these probabilities in parallel, and so this verification step is efficient.
\item \vspace{0.3em} We determine the maximum number of accepted speculated tokens. To keep the notation uncluttered, we denote $\mathrm{Pr}_q(\hat{y}_{i+t}|\mathbf{x},\mathbf{y}_{\le i},\hat{y}_{i+1}\cdots\hat{y}_{i+t-1})$ and $\mathrm{Pr}_p(\hat{y}_{i+t}|\mathbf{x},\mathbf{y}_{\le i},\hat{y}_{i+1}\cdots\hat{y}_{i+t-1})$ simply by $q(\hat{y}_{i+t})$ and $p(\hat{y}_{i+t})$, respectively. We then define the acceptance rule as follows: if $q(\hat{y}_{i+t}) \le p(\hat{y}_{i+t})$, then we accept this speculation. By contrast, if $q(\hat{y}_{i+t}) > p(\hat{y}_{i+t})$, we reject this speculation with probability $1-\frac{p(\hat{y}_{i+t})}{q(\hat{y}_{i+t})}$. Starting from $\hat{y}_{i+1}$, the maximum number of accepted consecutive speculated tokens is defined as:
    \begin{eqnarray}
    n_a & = & \min \left\{t-1 | 1 \le t \le \tau, r_t > \frac{p(\hat{y}_{i+t})}{q(\hat{y}_{i+t})} \right\},
    \end{eqnarray}
    \noindent where $r_t$ is a variable drawn from the uniform distribution $U(0,1)$. If all speculative tokens are accepted, we set $n_a=\tau$.
\item Given $n_a$, we keep the speculated tokens $\{\hat{y}_{i+1},\cdots,\hat{y}_{i+n_a}\}$. We then use the verification model to make a new prediction at $i+n_a+1$:
    \begin{eqnarray}
    \bar{y}_{i+n_a+1} & = & \argmax_{y_{i+n_a+1}} \mathrm{Pr}_p(y_{i+n_a+1}|\mathbf{x},\mathbf{y}_{\le i},\hat{y}_{i+1}\cdots\hat{y}_{i+n_a}).
    \end{eqnarray}
\item Above, we have described one step of speculative decoding. The result sequence (including both the context and predicted tokens) is illustrated as follows
\begin{center}
\input{./Chapter11/Figures/figure-speculative-sequence}
\end{center}
    Once we have finished this step, we add the predicted tokens $\{\hat{y}_{i+1},\cdots,\hat{y}_{i+n_a},\bar{y}_{i+n_a+1} \}$ to the context, and repeat the above process.
\end{itemize}
\vspace{0.5em}

In practice, we usually wish to use a smaller draft model so that predicting $\{\hat{y}_{i+1},\cdots,\hat{y}_{i+n_a}\}$ would be computationally cheaper. But a very small draft model is less accurate and can result in smaller $n_a$. We therefore need to carefully select the draft model to make the trade-off between the computational efficiency and accuracy.

\subsubsection{Stopping Criteria}

Stopping criteria are a critical component of LLM inference. They typically involve rules or conditions that specify when the model should stop generating text during decoding. Most LLMs are trained to generate an end-of-sequence token (e.g., $\langle \mathrm{EOS} \rangle$ or $\langle /s \rangle$) to signal the end of the generated text. Thus, the simplest strategy is to halt the generation process when this token is produced. For beam search, which explores multiple hypotheses simultaneously, the process can continue until a given number of complete sequences have been generated.

In practical applications, it will generally be undesirable to generate very long sequences, and so we need to reduce the decoding cost and unnecessary verbosity. One commonly used stopping criterion is the maximum length of the output. The model stops generating text once it has produced a predetermined number of tokens. Alternatively, we can stop the decoding based on actual computational cost, such as the computational resources or time constraints. For example, in real-time applications like chatbots, decoding may need to stop after a certain time limit to ensure responsiveness.

Another approach is to design stopping criteria based on the behavior of LLMs. For example, decoding can be stopped if the probability of predicting the next token falls below a certain threshold. In addition to probability-based stopping, a repetition detection module can be implemented to trigger the model to stop if it begins repeating tokens or phrases beyond a predefined limit. This helps prevent redundant or incoherent outputs.

\subsection{Evaluation Metrics for LLM Inference}

Evaluating the performance of LLM inference systems involves a variety of metrics to assess how well these models meet desired standards, such as quality, robustness, usability, and efficiency. As with most NLP systems, we can evaluate LLMs using standard NLP evaluation metrics, such as perplexity and F1 score. We can also examine their robustness by testing how well they handle ambiguous or challenging inputs, including adversarial examples, perturbed inputs, and out-of-distribution data. Additionally, usability can be assessed by measuring how well the generated outputs align with user expectations in terms of fluency, coherence, relevance, and diversity. Human evaluators can rate the naturalness of the text or assess whether the responses are contextually appropriate and logically consistent. Ethical and fairness metrics can also be included to ensure LLMs avoid perpetuating biases or generating harmful content.

All of the evaluation metrics mentioned above focus primarily on assessing the quality of the outputs. Given the high cost of deploying and serving LLMs, efficiency metrics are also very important for practitioners. Below are some commonly used efficiency metrics~\cite{nvidia-nim-llm-benchmarking}:

\begin{itemize}
\item \vspace{0.5em} \textbf{Request Latency}. This metric measures the total time taken from when a request is sent to the LLM until the complete response is received. This includes the time taken for data transmission, processing by the model, and the return of the output to the user.
\item \vspace{0.3em} \textbf{Throughput}. This metric measures the number of requests or generated tokens that the system can process per unit time. Throughput is commonly reported as requests per second (RPS) or tokens per second (TPS).
\item \vspace{0.3em} \textbf{Time to First Token} (TTFT). This metric measures the time between sending a request and generating the first token of the response. If data transmission does not consume too much time, then TTFT mainly consists of the time required for the prefilling stage and the generation of the first token.
\item \vspace{0.3em} \textbf{Inter-token Latency} (ITL). This metric refers to the time taken to generate each subsequent token after the first one. It reflects the efficiency of the decoding process.
\item \vspace{0.3em} \textbf{Resource Utilization}. This involves measuring the computational resource usage (e.g., CPU and GPU utilization) and memory consumption of the model during inference.
\end{itemize}
\vspace{0.5em}

In addition to these metrics, energy efficiency and cost efficiency are practical considerations for deploying LLMs at scale. Energy efficiency measures the amount of electrical power consumed by the model during inference. Cost efficiency, on the other hand, evaluates the total expenses related to deploying and maintaining the model.

In general, choosing the right evaluation metrics depends on the specific task and application. While quality-focused metrics are essential for assessing LLMs, efficiency metrics are equally crucial for their effective deployment in real-world applications. A comprehensive evaluation framework should include both sets of metrics to accurately measure an LLM's performance and practicality.

%----------------------------------------------------------------------------------------
% NEW SECTION
%----------------------------------------------------------------------------------------

\section{Efficient Inference Techniques}

In practical applications, we often wish a system to be as efficient as possible. For LLM inference, this typically involves two types of improvements: reducing memory requirements and accelerating inference. For example, we can modify the Transformer architecture to avoid memory explosion when processing very long input sequences. Another example is that we can compress input sequences to reduce computational overhead while preserving their semantic information. In addition, techniques like quantization and pruning can be employed to further optimize memory usage and inference speed.

Efficient inference is a wide-ranging topic that overlaps with several areas of LLM research, such as architecture design and model compression. Most of these topics have been covered in previous chapters. For example, in Chapter \ChapterLLM, we discussed long-context LLMs; and in Chapter \ChapterLLMPrompting, we discussed prompt compression methods for reducing prompt length. In this section, we focus on techniques that are commonly used in LLM deployment and serving.

\subsection{Prefix Caching}

In real-world applications, it is common to cache frequent requests and their corresponding responses. When a new request hits the cache, the system can retrieve the response directly from the cache instead of recomputing the result. One straightforward implementation is a key-value datastore (e.g., a hash table) that maps input sequences to their LLM-generated output sequences. In the simplest case, we can collect frequent queries, generate their responses using the LLM, and store these query-response pairs in the datastore. This creates a basic sequence-level caching mechanism that allows the system to bypass LLM computation when the input sequence exactly matches a cached query.

A straightforward extension of the caching mechanism is to cache prefixes and their corresponding KV cache states. Given an input sequence $\mathbf{x}$ in a dataset $\mathcal{D}$, we can process it as in the standard prefilling phase. This produces a sequence of prefixes and their corresponding KV cache states:
\begin{eqnarray}
x_0\ (\text{i.e., } \mathbf{x}_{<1}) & \Rightarrow & \mathrm{cache}_{<1} \nonumber \\
x_0 x_1\ (\text{i.e., } \mathbf{x}_{<2}) & \Rightarrow & \mathrm{cache}_{<2} \nonumber \\
& \cdots & \nonumber \\
x_0 x_1 \cdots x_{m-1}\ (\text{i.e., } \mathbf{x}_{<m}) & \Rightarrow & \mathrm{cache}_{<m} \nonumber
\end{eqnarray}

\noindent where $\mathrm{cache}_{<i}$ denotes the KV cache for the prefix $\mathbf{x}_{<i}$ (see also Eq. (\ref{eq:prefilling-kv-cache})). All these mappings can be stored in the prefix cache for efficient reuse.

When processing a new sequence that shares a common prefix with a previously seen sequence in $\mathcal{D}$, we can load the corresponding cached KV states instead of recomputing them. Specifically, if a new input $\mathbf{x}'$ shares a prefix $\mathbf{x}_{<k}$ with an existing sequence (i.e., $\mathbf{x}'_{<k} = \mathbf{x}_{<k}$ for some $k \le m$), we can initialize the KV cache with $\mathrm{cache}_{<k}$ and only compute the KV states for the remaining tokens $\mathbf{x}'_{\ge k}$.

As usual, we can maintain a key-value datastore that maps frequently encountered prefixes to their precomputed KV caches. The lookup can be performed using a hash of the prefix tokens, allowing constant-time access to the cached states. Care must be taken to manage memory usage, as storing all possible prefixes may be infeasible for large datasets. Practical systems often employ least recently used (LRU) caching methods or other strategies to balance between computational savings and memory constraints.

\subsection{Batching}

Batching in LLM inference refers to processing multiple input sequences simultaneously as a group (called a batch) rather than one at a time. Because modern GPUs excel at parallel processing, batching allows them to compute multiple sequences in a single forward pass, keeping the hardware fully occupied. Therefore, when serving LLMs at scale, batching is important for improving computational efficiency and maximizing hardware utilization\footnote{See \url{https://docs.nvidia.com/deeplearning/performance/dl-performance-gpu-background/index.html\#understand-perf} for a simple evaluation.}.

To illustrate the idea of batching, Figure~\ref{fig:basic-llm-batching-methods} (a-b) shows simple examples with batch sizes of 1 and 4, respectively. When using a batch size of 1 (i.e., without batching), the GPU processes one input sequence at a time. Thus, the processing is sequential: the next sequence must wait for the current computation to finish. By contrast, when using a batch size of 4, the GPU can process four sequences simultaneously in a single forward pass. As the input sequences vary in length, we need to standardize their length using padding techniques. Here we use left padding, which adds dummy tokens to the beginnings of short sequences, so that all sequences in the batch have the same length in prefilling. For decoding, tokens are generated simultaneously for all these sequences, and generation continues until all sequences in the batch finish decoding.

\begin{figure}[!t]
\centering
\input{./Chapter11/Figures/figure-basic-beam-search-methods}
\vspace{-0.5cm} \caption{Illustrations of basic batching methods. We use a 2D layout to illustrate the batch, where each square represents a token. Red squares indicate tokens in the prefilling stage, blue squares represent tokens in the decoding stage, green squares denote padding tokens, and gray squares correspond to meaningless tokens. Subfigures (a) and (b) compare the cases where the batch sizes are 1 and 4, respectively. Subfigure (c) shows the strategy of grouping sequences with similar lengths into the same batch. Subfigure (d) illustrates the disaggregation of prefilling and decoding. In this approach, we can make better use of GPU parallelism by concatenating multiple short sequences into a single long sequence for joint processing. This allows us to maximize the number of tokens processed in a batch while minimizing the number of padding tokens. However, as a trade-off, we need to copy the KV cache to the decoding engine and reorganize it after the prefilling phase, which introduces additional data transfer overhead.}
\label{fig:basic-llm-batching-methods}
\end{figure}

The above examples imply a trade-off between throughput and latency, which is a very important consideration in designing and implementing LLM inference systems. If we choose a smaller batch size, the latency would be lower, as fewer tokens need to be processed in a single run of inference. Imagine that we have only one sequence. The result becomes available immediately after generation completes, with no additional computational overhead. However, this low-latency advantage comes at the cost of underutilizing parallel computing resources, as the parallelism of GPUs remains largely idle during sequential processing. On the other hand, if we use a larger batch, we can make better use of the parallelism, as GPUs can be occupied by large-scale matrix computations. As a result, we can process more tokens in the same period of time and the throughput is improved. However, since the result is obtained only when the last token in the batch is predicted, the latency would be higher.

In practice, we usually prefer to use a slightly larger batch, but try to fill the batch with sequences of similar lengths to reduce the number of padding tokens and improve device utilization. For example, we can group the incoming user requests in a short period of time into buckets, each of which contains sequences with similar lengths. Then, we can fill the batch with sequences in the same bucket, so that we can minimize wasted computational resources, as illustrated in Figure~\ref{fig:basic-llm-batching-methods} (c).

Another approach to implementing batching in LLMs is to disaggregate the prefilling and decoding processes~\cite{wu-etal:2023fast,patel-etal:2024splitwise,zhong-etal:2024distserve}. For example, we can perform prefilling on one GPU, and perform decoding on another GPU. One advantage of disaggregation is that we can rearrange the input sequences in the batch to better fill it, because there is no interference between prefilling and decoding. For example, we can pack multiple short sequences together, thus ensuring that the lengths of sequences in the batch are as consistent as possible, as illustrated in Figure~\ref{fig:basic-llm-batching-methods} (d). In this way, we can maximize the throughput of the prefilling phase. However, as a trade-off, we need to transfer the KV cache to the devices performing decoding, which also incurs extra communication overhead. Typically, this method requires a high-bandwidth, low-latency network to achieve optimal performance.

In this section, we will discuss several improvements to the above basic batching strategies. Most of them are based on a non-disaggregated architecture, that is, decoding and prefilling can be considered as different stages of a model executed on the same device.

\subsubsection{Scheduling}

A practical LLM inference system typically consists of two components:

\begin{itemize}
\item \vspace{0.5cm} \mindex{Scheduler}. Its primary role is to efficiently queue and dispatch tasks (i.e., input sequences) to the inference engine based on the current system load and request priorities. This often involves a variety of batching strategies that group certain requests together to maximize processing efficiency.
\item \vspace{0.3cm} \mindex{Inference Engine}. It is responsible for the actual execution of the LLMs, processing the queued requests as they come in. As discussed previously, this engine involves both prefilling and decoding processes.
\end{itemize}
\vspace{0.5em}

This architecture is illustrated in Figure~\ref{fig:llm-inference-scheduler-and-engine}. Incorporating scheduling into batch processing provides a flexible way to optimize both the system's throughput and latency, thereby achieving a better balance between them. For example, the batching methods shown in Figure~\ref{fig:basic-llm-batching-methods} (a) and (b) can be viewed as simple forms of scheduling, called \mindex{request-level scheduling}. In this strategy, once a batch is filled and sent to the engine, the processing of the entire batch cannot be interrupted. The scheduler waits for this batch to be processed before handling the next batch~\cite{timonin-etal:2022}.

\begin{figure}[!t]
\centering
\input{./Chapter11/Figures/figure-inference-scheduling}
\vspace{-0.5cm} \caption{Illustration of the LLM inference architecture involving a scheduler and an inference engine. Each time, the scheduler selects a number of user requests to form a batch and sends it to the inference engine. The scheduler can interact with the inference engine and adjust the batch at certain points during inference, such as at the beginning of batch processing and at the start of each token prediction.}
\label{fig:llm-inference-scheduler-and-engine}
\end{figure}

A more sophisticated scheduling strategy, called \mindex{iteration-level scheduling}, interacts with the inference engine at each token prediction step rather than at the sequence level. This approach allows dynamic batch adjustment during inference, as illustrated in Figure~\ref{fig:llm-inference-and-iteration-level-scheduling}. Such fine-grained control lets the system prioritize latency-sensitive requests or sequences. For instance, if an urgent request arrives at some decoding step, the scheduler can add this request to the batch so that it can be processed as early as possible. In the following subsections, we will discuss batching methods based on iteration-level scheduling.

\begin{figure}[!t]
\centering
\input{./Chapter11/Figures/figure-request-level-vs-iteration-level-scheduling}
\vspace{-0.5cm} \caption{Illustrations of request-level scheduling and iteration-level scheduling. In request-level scheduling, once a batch is created and sent to the inference engine, we cannot adjust the batch. In other words, scheduling only occurs after the processing of a batch finishes. In iteration-level scheduling, we can perform scheduling during batch processing. For example, if a new request arrives at some point during inference, we can add it to the batch and continue processing.}
\label{fig:llm-inference-and-iteration-level-scheduling}
\end{figure}

\subsubsection{Continuous Batching}

\mindex{Continuous batching} is an iteration-level scheduling method used in the Orca system~\cite{yu-ettal:2022orca}. In this method, an iteration refers to either the entire prefilling procedure or a single decoding step. For example, given an input sequence $\mathbf{x}=x_0\cdots x_m$ and an output sequence $\mathbf{y}=y_1\cdots y_n$, there are $n+1$ iterations in total: one for prefilling, and $n$ for generating the output tokens (one per token).  During scheduling, the batch can be adjusted between iterations. For example, we can either add a new input sequence to the batch, or remove a completed sequence from the batch at some iteration, even while the batch is still being processed.

The general process of continuous batching includes the following steps:

\begin{itemize}
\item \vspace{0.3cm} Initially, a batch is created with one or more input sequences, based on both the inference engine's processing capacity and the current user requests. The batch is then fed into the inference engine.

\item \vspace{0.3em} The inference engine processes the batch iteration by iteration. After each iteration, the scheduler may adjust the batch in one of the following ways:
    \begin{itemize}
    \item \vspace{0.3em} If a sequence in the batch completes generation (i.e., generates the end-of-sequence symbol), that sequence is removed from the batch.
    \item \vspace{0.3em} If a new user request arrives and the inference engine has additional processing capacity, it is added to the batch.
    \item \vspace{0.3em} If no sequences are added to or removed from the batch, the batch remains unchanged.
    \end{itemize}

\item \vspace{0.3em} The processing terminates only when all sequences have been completed and no new user requests arrive.
\end{itemize}

\vspace{0.5em}

See Figure~\ref{fig:llm-continuous-batching} for an example of continuous batching. In this example, we start with two user requests,  $\mathbf{x}_1$ and $\mathbf{x}_2$. These two sequences are packed into a batch and sent to the inference engine for processing. After the engine completes two iterations, a new user request, $\mathbf{x}_3$, arrives. At this point, the scheduler adjusts the batch by adding $\mathbf{x}_3$ to it. The inference engine then continues processing the updated batch. Note that the inference engine now processes different sequences in different ways: $\mathbf{x}_1$ and $\mathbf{x}_2$ proceed with the decoding process (i.e., predicting the next tokens), while $\mathbf{x}_3$ undergoes the prefilling process. After some time, the generation for $\mathbf{x}_2$ completes. As it happens, two more user requests, $\mathbf{x}_4$ and $\mathbf{x}_5$, arrive. The scheduler removes the completed sequence $\mathbf{x}_2$ from the batch and, considering the current load of the inference engine, adds $\mathbf{x}_4$ to the batch. However, $\mathbf{x}_5$ must wait until another sequence in the batch finishes before it can be added.

\begin{figure}[!t]
\centering
\input{./Chapter11/Figures/figure-continuous-batching}
\vspace{-0.5cm} \caption{Illustration of batch adjustment in continuous batching. Instead of fixing a batch of input sequences and processing them to completion (as in request-level batching), continuous batching dynamically updates the batch during inference. The system continuously accepts and adds new requests (e.g., $\mathbf{x}_3$ and $\mathbf{x}_4$) into the current batch as long as there is available compute capacity.}
\label{fig:llm-continuous-batching}
\end{figure}

The idea behind continuous batching is to keep the inference engine fully utilized by processing as many sequences as possible, thereby maximizing computational resource usage. A key difference between continuous batching and standard batching (see Figure~\ref{fig:basic-llm-batching-methods}) lies in the fact that, in continuous batching, prefilling and decoding can occur simultaneously across different sequences, whereas in standard batching, these two phases are performed sequentially for the entire batch. As discussed in Section~\ref{sec:llm-inference-prefilling-and-decoding}, prefilling is considered a compute-bound process, while decoding is considered a memory-bound process. The intuition behind overlapping prefilling and decoding is to reduce idle times for both computation and data transfer. Consider two mini-batches: one for prefilling and one for decoding. While the prefilling mini-batch keeps the GPUs occupied, the decoding mini-batch can perform memory transfers concurrently.

Another difference between continuous batching and standard batching is that continuous batching is prefilling-prioritized, while standard batching is decoding-prioritized~\cite{agrawal-etal:2024taming}. In continuous batching, once the inference engine has spare computational resources, the scheduler will add new requests to the batch. In other words, these newly added requests will be processed for prefilling as early as possible. This approach improves system throughput, but at the cost of increased latency, as the newly added requests extend the processing time of earlier ones. In contrast, in standard batching, once the batch is created, we must wait for the last sequence in the batch to complete before processing new requests. This ensures relatively low latency, but results in lower device utilization and system throughput.

It is important to note that the cost of continuous batching is that we need to continuously reorganize the batches, which involves rearranging the data in memory. Each time a new request is added, the scheduler needs to reassess and optimize the current batch structure. This dynamic adjustment can incur additional memory and computational overhead, especially when the batches are frequently adjusted. Therefore, while this method can improve throughput, it may also lead to increased memory fragmentation and, in some cases, introduce additional latency.

\subsubsection{PagedAttention}

PagedAttention (or paged KV caching) is a technique used in the vLLM system~\cite{kwon-etal:2023efficient}. Inspired by operating system paging, it optimizes memory usage during LLM inference --- particularly for the KV cache --- by addressing fragmented memory allocation in dynamic batching scenarios with variable-length sequences. The idea behind PagedAttention is to divide the KV cache into smaller memory pages. In this way, we do not need to store the KV cache of the full sequence in contiguous memory. Instead, the KV cache is divided into fixed-size blocks (analogous to memory pages in an operating system), which can be non-contiguously allocated in physical memory. One advantage of PagedAttention is that it enables flexible memory management, supporting dynamic sequence growth without requiring expensive reallocation or copying of large contiguous memory regions. Note that PagedAttention is not specifically designed for batching. But it indeed helps improve memory efficiency in batched inference scenarios, where memory management is more demanding and complicated.

\begin{figure}[!t]
\centering
\input{./Chapter11/Figures/figure-paged-attention}
\vspace{-0.5cm} \caption{Illustration of memory allocation in PagedAttention. There are two sequences in the batch, as illustrated in sub-figure (a). Since the memory is fragmented, the KV cache is stored in a large unused block of memory in standard self-attention (see sub-figure (b)), but the fragmented memory is not used. By contrast, in PagedAttention (see sub-figure (c)), the KV cache is divided into smaller blocks and thus fits into fragmented memory.}
\label{fig:llm-paged-attention-example}
\end{figure}

Consider a simple example of memory allocation in Figure~\ref{fig:llm-paged-attention-example} in which self-attention is performed for a batch consisting of two sequences. For each sequence, we need to attend the current token to the key-value pairs in the KV cache of this sequence, as required by self-attention. In the standard implementation of self-attention, the KV cache is stored in a contiguous block of memory, allowing us to efficiently access this contiguous memory. However, in a paged KV caching system, the KV cache is divided into smaller, fixed-size memory blocks which are not necessarily contiguous. These smaller KV cache blocks can be more effectively allocated to fragmented memory regions, thereby improving memory utilization. Another benefit of distributing chunks of the KV cache across different memory blocks is that it enables parallelization of the caching process. For example, if the input sequence is long and the memory bandwidth is sufficient, it would be beneficial to write and read the key and value vectors of different segments of the sequence in parallel across multiple memory blocks.

In general, storing contiguous data in non-contiguous regions can cause issues, for example, may introduce additional memory access overhead, which reduces I/O efficiency. However, when handling large-scale data (e.g., performing large matrix multiplications), we typically do not process all the data at once but instead divide it into smaller blocks for block-level computation. From this perspective, it is also reasonable to partition the attention computation. If the paging strategy is well designed, the additional overhead in memory access can be minimal, while the improvement in memory utilization can be significant.

\subsubsection{Chunked Prefilling}

We have seen that, in iteration-level scheduling, prefilling and decoding for different sequences can occur simultaneously. This can be seen as a prefilling-prioritized strategy which can maximize the throughput. However, one such iteration can take a long time if the input sequence is very long and the prefilling process dominates the computation. In this case, decoding for other sequences has to wait until the prefilling completes, leading to increased latency for generating output tokens. Therefore, while prefilling-prioritized strategies are effective for maximizing hardware utilization, they may introduce significant variability in token generation latency, particularly when the system is handling a mix of long and short input sequences.

A simple way to reduce decoding latency is to make computations for different sequences in the batch comparable. One such method is to divide sequences into chunks and perform prefilling chunk by chunk. This approach, often referred to as chunked prefilling, processes smaller portions of each sequence at a time, allowing the system to better balance the computational load across sequences~\cite{agrawal-etal:2023sarathi}. By choosing an appropriate chunk size, we can ensure that when prefilling and decoding overlap for two sequences, their processing within the same iteration tends to take a similar amount of time. As a result, decoding idle time is reduced and overall throughput is improved.

Figure~\ref{fig:llm-chunked-prefilling} shows an illustration of chunked prefilling in a few iterations. In this example, the batch contains two sequences. The whole prefilling process of the first sequence is divided into three prefilling steps, giving rise to the chunks denoted $P_{11}$, $P_{12}$ and $P_{13}$. Each chunk corresponds to one iteration and can thus overlap with one decoding step. In this way, during the prefilling of the first sequence, we can perform three decoding steps, rather than only a single decoding step, as is the case in standard iteration-level scheduling. As a result, the idle time of the decoding process is reduced, and the output tokens can be generated earlier.

Chunked Prefilling improves decoding efficiency by overlapping prefilling and decoding, but at the cost of additional memory overhead and scheduling complexity. In standard prefilling, we process the whole input sequence once, building the KV cache in one go. By contrast, in chunked prefilling, each chunk needs a separate forward pass to compute its attention outputs and update the KV cache. As a result, we need to maintain the KV cache of early chunks while processing later chunks. This also compromises the parallelism of completing the prefilling for the entire sequence in a single pass. In practice, it is usually possible to balance throughput and latency by choosing an appropriate chunk size.

\begin{figure}[!t]
\centering
\input{./Chapter11/Figures/figure-chunked-prefilling}
\vspace{-0.5cm} \caption{Comparison of simple iteration-level scheduling and chunked prefilling. $P_{x y}$ denotes the $y$-th prefilling step for sequence $x$, and $D_{x y}$ denotes the $y$-th decoding step for sequence $x$. In simple iteration-level scheduling (or prefilling-prioritized scheduling), since prefilling is treated as a single iteration, $D_{22}$ has to wait for the completion of the prefilling of sequence 1. In chunked prefilling, the prefilling process can be divided into multiple steps. Thus, $D_{22}$ can execute during prefilling for sequence 1 (i.e., during $P_{12}$).}
\label{fig:llm-chunked-prefilling}
\end{figure}

It is worth noting that the methods discussed in this subsection can broadly be categorized as priority-based scheduling methods. In these methods, we can give priority to certain requests, or to certain prefilling or decoding steps, so that system resources are allocated in a way that better aligns with specific performance goals. As presented above, for example, we may prioritize decoding over prefilling to minimize token generation latency, or prioritize prefilling over decoding to maximize overall throughput in batch-processing scenarios. Practitioners can design custom priority policies for specific needs and operational constraints in real-world applications, such as request deadlines and importance levels defined by users.

\subsection{Parallelization}

Parallelization is a widely used approach for scaling LLM inference, especially for large-scale deployments. In Chapter \ChapterPretraining, we have discussed several common strategies for parallelizing LLM pre-training, such as model parallelism, tensor parallelism, and pipeline parallelism. We have also discussed efficient architectures that are easy to deploy in distributed computing systems. For example, in MoE models, we assign different experts to different devices\footnote{In LLMs, the experts are typically FFN modules. So each expert is a part of the FFN component in the Transformer architecture.}. Only the active experts for a given input are executed, which significantly improves computational efficiency while maintaining model quality. Many of these methods can be directly applied to LLM inference with minimal modifications.

However, applying these parallelization techniques to inference poses new challenges compared to pre-training. These issues become especially pronounced in real-time or low-latency inference scenarios, where load imbalance across devices and communication overhead can significantly impact performance. For example, unlike pre-training, where batches can be prepared in advance, inference must handle variable-length sequences in real time. This makes it harder to maintain optimal device utilization and complicates scheduling across heterogeneous computational resources. A related issue is load balancing. When a large number of requests arrive in a short period of time, the system must efficiently distribute workloads across available devices. For example, real-world requests typically exhibit highly variable computational demands due to differences in task types and prompt lengths. Such variability renders simple static load balancing approaches ineffective, and so we need to use finer-grained strategies that can adapt to runtime conditions. The problem becomes even more complicated when we deploy the system on heterogeneous hardware and there are strict latency constraints.

In the development of LLMs, parallelization is closely related to LLM serving. Building a high-quality LLM serving system is not a simple task --- it typically requires combining multiple techniques, such as architectural design, workload distribution, and LLM-specific hardware/software optimizations. As such, LLM serving constitutes a broad topic that often demands substantial engineering expertise. Here, we will not go into the details of LLM serving. For related concepts and techniques, readers may refer to relevant open-source systems (such as vLLM\footnote{\url{https://github.com/vllm-project/vllm}}, TensorRT-LLM\footnote{\url{https://github.com/NVIDIA/TensorRT-LLM}} and TGI\footnote{\url{https://github.com/huggingface/text-generation-inference}}) and papers~\cite{pope-etal:2023efficiently,li-etal:2024llm}.

\subsection{Remarks}

We have considered many methods for improving the efficiency of LLMs in this and previous chapters. Although these approaches address different issues, most of them essentially explore trade-offs between various performance factors. One important trade-off is between inference speed and accuracy. For example, techniques like quantization, pruning, and knowledge distillation can significantly reduce computational overhead and latency but may slightly degrade model performance. Conversely, using higher numerical precision or larger models enhances accuracy but at the cost of slower inference and higher resource demands.

Another important consideration in LLM inference is the memory-compute trade-off. As in computer system design, we need to consider the balance between memory usage and computation required to generate the output. In particular, storing intermediate results such as KV caches during inference can significantly reduce redundant computation, but at the expense of increased memory usage. In KV caching, storing past attention states avoids recomputation of self-attention over previous tokens, thereby reducing compute time per token. However, as the number of tokens grows, so does the memory footprint of the KV cache, especially when processing very long sequences or multiple sequences in parallel. In response, various techniques have been developed to reduce memory consumption by partially recomputing intermediate states. For instance, chunked or windowed attention limits the attention span to a recent subset of tokens, and thus reduces the KV cache size. However, this often requires additional compute if past information must be reprocessed.

Note that considering the memory-compute trade-off is a very general principle. It can be extended beyond attention mechanisms and Transformers to other components in system design. An example is the choice of data precision. Using lower-precision formats such as FP16 or INT8 can reduce both memory usage and memory bandwidth requirements, effectively alleviating pressure on the memory subsystem. However, lower precision may lead to numerical instability or slight accuracy degradation, requiring careful calibration or retraining. Thus, this trade-off can also be seen as a memory-compute-accuracy triangle, where improvements in one dimension may come at the expense of another.

Beyond speed, accuracy, and memory, several other dimensions also influence LLM inference efficiency. Some of these dimensions have been discussed in this chapter, while others have not. Here we outline them as follows.

\begin{itemize}
\item \vspace{0.5em} \textbf{Throughput vs. Latency}: In large-scale multi-user LLM serving scenarios, we often aim to maximize system throughput. For example, as discussed in this section, we can batch multiple requests together to increase the number of tokens processed at the same time. However, batching increases waiting time and may lead to higher per-request latency, especially for short or interactive requests. By contrast, optimizing for low latency often requires serving requests individually or in smaller batches, which underutilizes hardware resources and reduces throughput. Achieving a good balance depends on the quality-of-service requirements and user interaction patterns.
\item \vspace{0.3cm} \textbf{Generalization vs. Specialization}: General-purpose LLMs are trained to perform a wide range of tasks with a single set of parameters. While flexible, they may be less efficient or accurate for specific tasks. Specialized models can yield better performance and lower inference costs for targeted applications. However, maintaining multiple specialized models increases system complexity and storage requirements. The trade-off between maintaining a single general model versus multiple specialized models is an important system-level design choice.
\item \vspace{0.3em} \textbf{Energy Efficiency vs. Performance}: High-performance inference often requires running large models at high throughput on powerful accelerators, which consumes considerable energy. This may be problematic for edge deployments or energy-sensitive environments. Techniques like model compression can improve energy efficiency, but usually with some degradation in output quality or increase in latency. Energy constraints thus introduce another important dimension in optimizing LLM inference.
\end{itemize}
\vspace{0.5em}

%----------------------------------------------------------------------------------------
% NEW SECTION
%----------------------------------------------------------------------------------------

\section{Inference-time Scaling}

\begin{figure}[!t]
\centering
\includegraphics[scale=0.23]{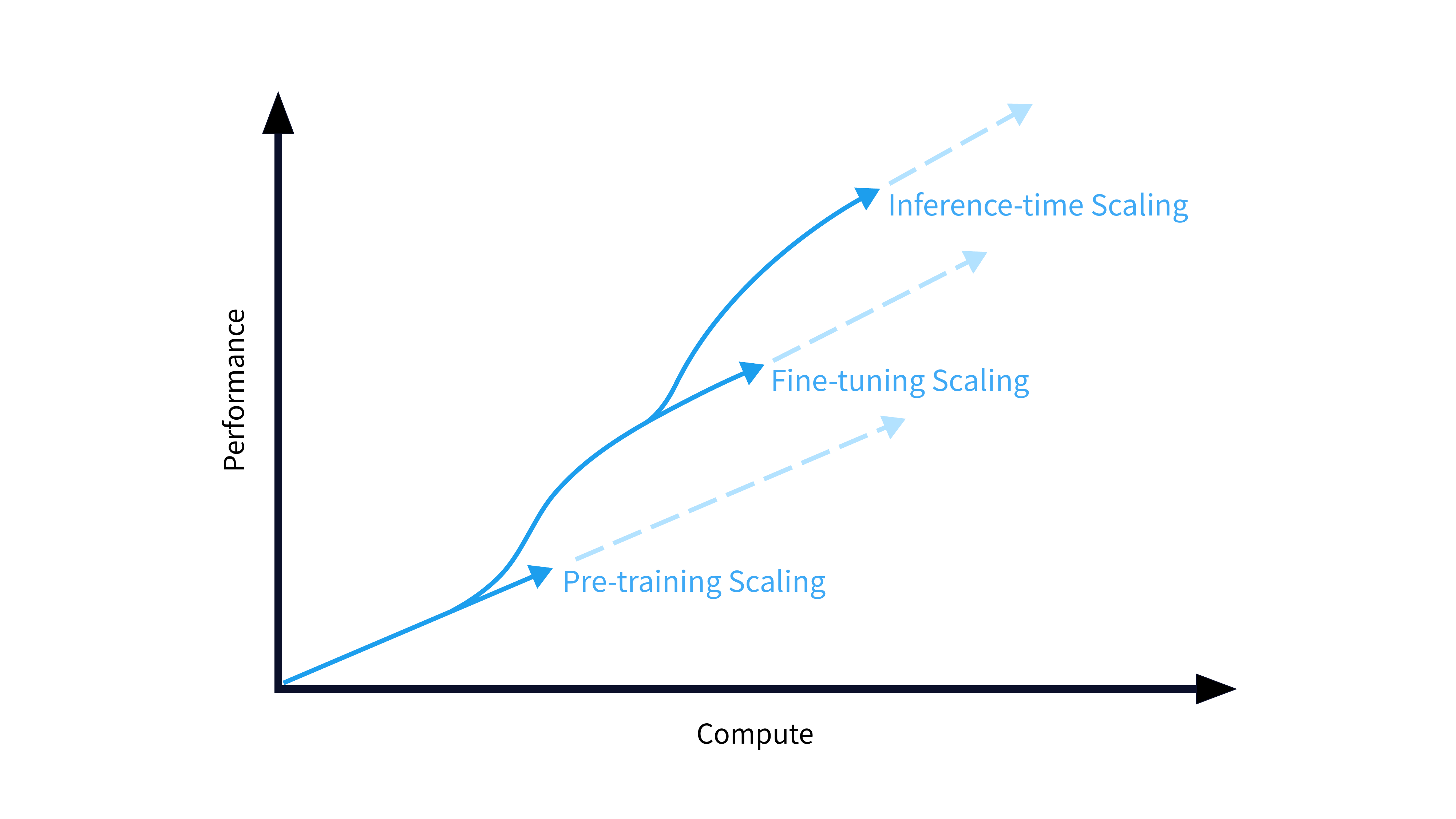}
\vspace{-0.5cm} \caption{Scaling for pre-training, fine-tuning and inference stages~\cite{briski:2025how}.}
\label{fig:scaling-laws-for-llm-development}
\end{figure}

Scaling laws can be considered one of the fundamental principles guiding the development of LLMs. In previous chapters, we discussed several times that scaling up training data, model size, and compute can effectively improve LLM performance. In fact, scaling laws also apply to downstream stages such as fine-tuning and inference (see Figure~\ref{fig:scaling-laws-for-llm-development}). Here we consider \mindex{inference-time scaling}, which has been widely employed by recent LLMs to solve challenging problems, such as complex math problems~\cite{snell-etal:2025scaling}. Unlike scaling during pre-training and fine-tuning, which focuses on improving LLMs via parameter updates, inference-time scaling improves model performance during inference without further training. These methods scale inference in different ways, such as ensembling multiple model outputs, increasing context length, adopting more aggressive decoding algorithms, and using external tools to extend model capabilities.

While inference-time scaling is wide-ranging, in this section we consider those methods that incorporate more compute into inference (called inference-time compute scaling). Here is a list of inference-time (test-time) compute scaling methods, organized by category:

\begin{itemize}
    \item \vspace{0.5em} \textbf{Context Scaling}. It involves scaling the input or context to improve generation (or potentially scale the output).
    \item \vspace{0.3em} \textbf{Search Scaling}. It involves increasing computational effort during decoding.
    \item \vspace{0.3em} \textbf{Output Ensembling}. It involves combining multiple model outputs.
    \item \vspace{0.3em} \textbf{Generating and Verifying Thinking Paths}. It involves guiding LLMs to generate and verify thinking paths for solving complex reasoning problems.
    \end{itemize}
    \vspace{0.5em}

We will describe these methods in the following subsections.

\subsection{Context Scaling}

Context scaling improves LLM performance by extending the input to the model. A straightforward approach is to incorporate more relevant context during inference so that the model can condition its predictions on more prior information. One example is few-shot prompting. It augments the context with multiple input-output examples, and so the model can learn task behavior implicitly from these examples without parameter updates. On top of few-shot prompting, we can use chain-of-thought prompting to encourage the model to produce intermediate reasoning steps before producing the final answer. Note that chain-of-thought prompting is one of the most important methods for improving reasoning performance. By explicitly providing intermediate steps in problem-solving, we can prompt the model to break down complex tasks into simpler sub-tasks, which is found to be very beneficial for generating accurate and interpretable outputs.

Beyond extending the prompt with examples or reasoning steps, another approach to context scaling involves dynamically incorporating external knowledge. This is often achieved using RAG. RAG systems first retrieve relevant document snippets from a large collection of documents or a database based on the current input. These retrieved pieces of information are then added to the context provided to the LLM. This essentially expands the context to include timely or specialized external knowledge. By doing so, the model grounds its responses in specific knowledge found in the external source. The LLM can thus generate responses that are not only relevant to the input but also factually accurate and up-to-date.

However, as the context grows, these methods often suffer from the constraints of a finite context window. While model architectures and techniques (like efficient attention models) are continually evolving to support longer contexts, processing extremely long inputs still poses challenges. Increased computational cost is one factor. More importantly, when the context window becomes very large, the model might struggle to attend effectively to the most relevant information (e.g., the ``lost in the middle'' phenomenon). Therefore, effective context scaling is not just about adding more information, but also about strategically selecting, structuring, and presenting the most pertinent information within the model's processing capabilities.

Here we omit the detailed discussion of these methods, as they have already been covered in previous chapters. See Chapters \ChapterLLM\ and \ChapterLLMPrompting\ for more details, including prompting, RAG, and long-sequence modeling methods.

\subsection{Search Scaling}

In LLMs, decoding is a search process that aims to efficiently find the best output sequence given the input sequence. Search scaling (or decoding scaling) typically involves two aspects: scaling the output length and scaling the search space.

Scaling the output length refers to increasing the number of tokens generated during inference. This is especially important in tasks that require long-form generation, such as story writing. More recently, generating outputs with long thinking paths has shown strong performance in math problem solving and code generation. For example, encouraging the model to generate long thinking paths before producing the final answers has been found to be very beneficial in performing complex reasoning. This idea has been widely used in developing recent LLMs for reasoning, such as \citet{openai:2024learning}'s o1 and \citet{deepseek:2025r1}'s R1. We will discuss output-length scaling in more detail in Section~\ref{sec:generating-long-thinking-paths}.

Scaling the search space, on the other hand, refers to expanding the set of candidate output sequences considered during search, so that higher-quality outputs can be found. As discussed in Section~\ref{sec:llm-decoding-algorithms}, a simple example is that in beam search increasing the beam width allows more candidate sequences to be explored at each decoding step. This increases the chance of discovering better outputs, especially in tasks where the optimal solution is not immediately apparent from local decisions.

In addition to decoding algorithm adjustments, it is also possible to explore compact structures to encode a large number of outputs. For example, we can construct and navigate a tree or graph of reasoning steps~\cite{yao-etal:2024tree}. In this paradigm, each node represents a partial solution or intermediate step, and edges represent transitions between reasoning states. Such structured search enables the model to consider multiple paths simultaneously. Another related direction is Monte Carlo tree search-inspired decoding, where the model stochastically explores and scores different paths based on learned heuristics or external reward models.

Search scaling is a very general idea, and it is often implicitly involved in the design of search procedures that exploit search structure, heuristics, and model uncertainty. Many of the above methods have been discussed previously, though they were not originally developed with scaling as their primary goal. However, search scaling inherently comes with computational costs. Increasing beam width, for instance, directly translates to higher memory usage and longer inference times. In practice, there is often a point of diminishing returns, where further expansion of the search space yields marginal improvements in output quality at a significant computational expense. Therefore, an effective strategy often involves finding an optimal balance between scaling and computational feasibility.

\subsection{Output Ensembling}

If multiple model outputs are available, it is often beneficial to combine them in order to reduce the impact of individual errors and produce a more reliable final output. Each model might capture different aspects of the underlying data distribution or possess unique strengths and weaknesses. By ensembling, we can average out the noise or random errors present in individual predictions, leading to a more stable and reliable outcome. In LLM ensembling, one of the simplest approaches is to average the probability distributions over the next token from each model, and select the next token using this averaged distribution. Or, if we regard the problem as a discrete decision-making task, majority voting can be employed. More sophisticated methods might involve re-ranking candidate outputs generated by different models based on a separate scoring function or even using a meta-learner to intelligently combine the predictions.

Output ensembling requires additional inference-time computation, such as running multiple models or sampling multiple outputs. This not only increases the latency of inference but also leads to additional system complexity. But the quality of outputs does not continue to improve indefinitely as more models are added. In some cases, the benefits of output ensembling may diminish as the number of component models in the ensemble exceeds a certain threshold. Instead, the benefits of ensembling are generally greater when the individual models are diverse (i.e., they make different errors), even if there are a relatively small number of component models. Therefore, if multiple models are available, diversity among the ensemble members is also desirable.

Output ensembling illustrates that inference-time scaling is not limited to increasing model size or output length. By allocating additional computation to generate and aggregate multiple candidate outputs, ensembling can improve robustness and output quality. As with other forms of inference-time scaling, these gains must be balanced against the additional computational cost.

\subsection{Thinking-Path Generation and Verification}
\label{sec:generating-long-thinking-paths}

So far, we have viewed inference-time scaling as a general class of methods for scaling various aspects of inference, such as sequence length, model size, and/or search strategies. In fact, one successful application is the use of inference-time scaling to enhance the reasoning capabilities of LLMs. As we have seen, chain-of-thought prompting improves the reasoning performance of LLMs by encouraging the generation of intermediate reasoning steps before arriving at a final answer. However, reasoning problems are often so complicated that we cannot obtain high-quality solutions by providing simple chain-of-thought prompts. For example, when solving a math problem, we typically need to reason through a sequence of steps. At each step, we need to work out some intermediate result, verify it, and then determine what to do next. The reasoning path is not a fixed pattern but a dynamically generated thinking process that often involves trial and error, backtracking, and self-correction. This requires more sophisticated prompting strategies or search algorithms to navigate such complex reasoning. In this subsection, we focus on inference-time scaling methods that go beyond simple chain-of-thought to address complex reasoning problems more effectively.

At a high level, methods for scaling the reasoning of LLMs can be categorized into two classes:

\begin{itemize}
\item \vspace{0.5em} \textbf{Training-free Methods}. These methods aim to improve reasoning capabilities without updating the pre-trained model parameters. Instead, they focus on techniques applied during inference, such as sophisticated prompting strategies (e.g., chain-of-thought) and algorithmic control over the reasoning process (e.g., search).
\item \vspace{0.3em} \textbf{Training-based Methods}. These methods involve further training or fine-tuning the model parameters to explicitly improve reasoning abilities, such as supervised fine-tuning on datasets with reasoning examples (e.g., math problems with step-by-step solutions).
\end{itemize}
\vspace{0.5em}

In the following, we first discuss training-free methods, and then training-based methods.

\subsubsection{Solution-level Search with Verifiers}

Given an input sequence (e.g., a math problem), there are many possible output sequences (e.g., solutions to the problem). If we have a model to evaluate or verify each solution, we can select the best one. This is the fundamental principle behind methods like best-of-$N$ sampling, where multiple outputs are generated, and the optimal result is picked based on some  selection mechanism. Such a selection process can be viewed as a search problem, which involves two components:

\begin{itemize}
\item \vspace{0.5em} \textbf{Search Algorithm}. This defines the strategy used to explore the space of possible output sequences (solutions) and generate a set of candidates. It can range from simple independent sampling to more sophisticated search techniques as discussed in Section~\ref{sec:llm-decoding-algorithms}.
\item \vspace{0.3em} \textbf{Verifier}.  This is a model or function responsible for evaluating the quality, correctness, or utility of each candidate solution generated by the search algorithm. It provides a score, a probability, or a judgment that allows the system to select the best among the candidates. The verifier can be another LLM, or even a set of predefined rules or heuristics.
\end{itemize}
\vspace{0.5em}

Given an input problem $\mathbf{x}$, we can represent an output solution $\mathbf{y}$ as a sequence of reasoning steps:
\begin{eqnarray}
\mathbf{y} & = & (a_1,a_2,\cdots,a_{n_r}),
\end{eqnarray}

\noindent where $a_i$ is the $i$-th reasoning step, and $a_{n_r}$ is the last step which contains the answer to the problem. See Figure~\ref{fig:multi-step-reasoning-example} for an example of a multi-step reasoning path.

\begin{figure}[!t]
\centering
\input{./Chapter11/Figures/figure-reasoning-path}
\vspace{-0.5cm} \caption{Illustration of multi-step reasoning. This example is from the PRM800K dataset~\cite{lightman-etal:2024lets}. Given a math problem, the LLM is prompted to generate a thinking path (or reasoning path) consisting of several reasoning steps. Each step addresses a sub-problem based on the results of the previous steps. The answer to the original problem is contained in the last step.}
\label{fig:multi-step-reasoning-example}
\end{figure}

The search algorithm can generate a set of candidate solutions:
\begin{eqnarray}
\mathcal{D}_c & = & \{\mathbf{y}_1,\cdots,\mathbf{y}_K\}.
\end{eqnarray}

Then, we can use a verifier, which evaluates each solution using the function $V(\mathbf{y})$, to score the candidates in $\mathcal{D}_c$. The final output is the best candidate selected by the verifier:
\begin{eqnarray}
\hat{\mathbf{y}} & = & \argmax_{\mathbf{y} \in \mathcal{D}_c} V(\mathbf{y}).
\end{eqnarray}

Although verifying the entire reasoning path is possible, a simpler alternative is to verify only the final reasoning step. In this way the verifier function $V(\mathbf{y})$ is simplified to depend solely on the final answer contained within $a_{n_r}$. This can be achieved in various ways, depending on the nature of the problem and the expected answer format.

\begin{itemize}
\item \vspace{0.5em} For some math and coding problems, we can use off-the-shelf tools as verifiers. Examples include proof checkers for mathematical theorems, interpreters or compilers for code execution, and unit test systems for verifying program correctness against predefined test cases.
\item \vspace{0.3em} If there is labeled data for evaluating the answer, such as human preference data, we can train a reward model on such data. The learned reward model is then used as the verifier which assigns a scalar score to each candidate answer.
\item \vspace{0.3em} If there are no existing systems or suitable reward models, we can use another LLM to act as the verifier. This LLM is prompted to assess the quality of the candidate answer. It could potentially be a more capable model, or the same LLM used with a specific ``evaluator'' prompt.
\item \vspace{0.3em} Alternatively, simpler heuristic-based verifiers can be designed. A commonly used approach is to employ majority voting, where the most frequently occurring answer among a set of candidates is selected.
\end{itemize}
\vspace{0.3em}

Using these verifiers, we can evaluate a set of candidate solutions generated during search. One simple strategy, which is often referred to as \mindex{parallel scaling}~\cite{brown-etal:2024large,snell-etal:2024scaling}, involves generating $K$ candidate solutions by running the base LLM $K$ times independently.  In this process, we can adjust the temperature in sampling to control the diversity in the outputs. The verifier then assesses each of these $K$ complete solutions, and the one with the highest score is selected as the final output. This is conceptually very similar to best-of-$N$ sampling, which in previous chapters we primarily described as a method of selecting the best one from a set of sampled outputs using a reward model.

Another approach is \mindex{sequential scaling}, which builds a sequence of solutions incrementally~\cite{gou-etal:2024critic,zhang-etal:2024small}. It starts with an initial solution generated by the LLM with prompting. Then, we use a verifier (often the same LLM) to evaluate the solution. This can be seen as a critique stage. The output of this stage is some form of feedback, such as textual critiques pinpointing errors or suggesting improvements, numerical scores reflecting solution quality, or even a revised plan or intermediate step to guide the next generation. This feedback, along with the original problem and the current solution, is then used to prompt the LLM to generate a potentially improved solution. This can be seen as a refine stage. This critique-refine cycle can be repeated, forming an iterative loop:
\begin{eqnarray}
\mathbf{y}_{k+1} & = & \mathrm{Refine}(\mathbf{x}, \mathbf{y}_k, \mathrm{Feedback}(\mathbf{y}_k)),
\end{eqnarray}

\noindent where $\mathrm{Feedback}(\mathbf{y}_k)$ represents the feedback from the verifier. The $\mathrm{Refine}(\cdot)$ function generates the improved solution $\mathbf{y}_{k+1}$ by prompting the LLM with the original problem $\mathbf{x}$, the previous solution $\mathbf{y}_k$, and this feedback. The process can be repeated for $K$ iterations, or until the solution quality, as assessed by the verifier, converges to a satisfactory level. This iterative framework, where a solution is progressively improved through cycles of generation, evaluation (critique), and revision, is precisely what constitutes self-refinement~\cite{shinn-etal:2023reflexion,madaan-etal:2024self}. In such scenarios, the role of the verifier is not just to pick the best complete solution from a static set, but to actively guide the generation process itself.

See Figure~\ref{fig:parallel-and-sequential-scaling} for illustrations of parallel scaling and sequential scaling. Note that there are other ways to perform search and obtain different sets of candidate solutions. One alternative method is to organize the search as a tree structure. This approach, often referred to as tree search, provides a more structured way to explore the space of possible reasoning paths. In solution-level search, each node of the tree represents a complete solution. During search, we need to expand a node into a set of child nodes, representing new solutions that can be considered in verification. The expansion process typically involves taking an existing solution (the parent node) and using the LLM to generate variations or alternative solutions.

\begin{figure}[!t]
\centering
\input{./Chapter11/Figures/figure-parallel-scaling-and-sequential-scaling}
\vspace{-0.5cm} \caption{Illustrations of parallel scaling and sequential scaling. In parallel scaling, we obtain multiple solutions by running the LLM several times independently. In sequential scaling, the LLM generates an initial solution. Then, we use the LLM to refine it iteratively, with each refinement yielding a new, possibly better solution.}
\label{fig:parallel-and-sequential-scaling}
\end{figure}

\subsubsection{Step-level Search with Verifiers}

While the methods discussed above primarily focus on generating complete solutions before final selection, the search process can also be integrated more deeply into the step-by-step generation of the reasoning path itself. This leads to approaches that perform step-level search with verifiers, where guidance or pruning occurs at intermediate reasoning steps $\{a_1,\cdots,a_{i}\}$ rather than only after a full solution $\mathbf{y}$ is formed.

Such fine-grained control is particularly beneficial for complex reasoning problems where a single incorrect intermediate step can render the entire subsequent reasoning chain invalid. By evaluating or guiding the generation at each intermediate step, the LLM can explore the reasoning space more effectively, potentially pruning unpromising paths early or allocating more resources to explore more plausible ones.

Step-level search with verifiers can also be modeled as a tree search problem. In this paradigm, each node (or state) corresponds to a partial reasoning path, $\mathbf{a}_{\le i} = (a_1,\cdots,a_i)$, representing the sequence of $i$ reasoning steps taken so far (i.e., a path from the root node to the current node). The objective of the search process is to explore the underlying state space, starting from an initial empty path, to find a complete path that constitutes a correct solution. Note that we use $\mathbf{a}_{\le i}$ here to represent a partial reasoning path instead of $\mathbf{y}_{\le i}$. While this makes notation a bit inconsistent with that used for representing complete solutions ($\mathbf{y}$) or full paths in solution-level search, it serves to highlight the focus on individual actions or steps.

The core components of step-level search with verifiers are:

\begin{itemize}
\item \vspace{0.5em} \textbf{Node Representation}. A node is a partial reasoning path $\mathbf{a}_{\le i} = (a_1,\cdots,a_i)$. The root node is an empty path, and terminal nodes are complete reasoning paths.
\item \vspace{0.3em} \textbf{Node Expansion}. Given a current partial path $\mathbf{a}_{\le i}$, the LLM is used to generate one or more candidate next reasoning steps $\{a_{i+1}^{(1)},\cdots,a_{i+1}^{(M)}\}$. Each candidate step, when appended to $\mathbf{a}_{\le i}$, forms a new potential partial path $\mathbf{a}_{\le i + 1} = (a_1,\cdots,a_i,a_{i+1}^{(j)})$.
\item \vspace{0.3em} \textbf{Verification}. The verifier $V(\cdot)$ evaluates the quality of a newly generated step in the context of the current partial path $\mathbf{a}_{\le i} = (a_1,\cdots,a_i)$ and the original problem $\mathbf{x}$. As with solution-level verification, step-level verifiers might output a numerical score, a categorical label, or textual feedback.
\item \vspace{0.3em} \textbf{Search}. This governs how the search space is explored. Based on the evaluations from the verifier, the search strategy decides which partial paths to extend further, which to prune, and the order of exploration.
\end{itemize}
\vspace{0.5em}

This step-by-step verification allows for dynamic adjustments to the reasoning process. If a step $a_{i+1}$ is deemed incorrect or unpromising by $V(\cdot)$, the search algorithm can backtrack and explore alternative steps from $\mathbf{a}_{\le i}$, or even from an earlier node $\mathbf{a}_{\le i'}$ (where $i' < i$). Conversely, if a step is highly rated, resources can be focused on extending that path. See Figure~\ref{fig:step-level-search-with-verifiers} for an illustration of step-level search with verifiers.

\begin{figure}[!t]
\centering
\input{./Chapter11/Figures/figure-step-level-search-with-verifiers}
\vspace{-0.5cm} \caption{Illustration of step-level search with verifiers. $a_{i}^{(j)}$ = the $j$-th candidate for the $i$-th reasoning step, {\color{blue} $\boxtimes$} = candidate pruned by the LLM's output probability, and {\color{ugreen} $\boxtimes$} = candidate pruned by the verifier. Given the input problem as the root node, we expand the tree by generating multiple reasoning steps at each expansion. Each candidate can be pruned by either likelihood (as in standard decoding) or step-level verification. The unpruned candidates are then expanded to generate further reasoning steps. The process is iterated until a complete reasoning chain leading to a final answer is generated, or until a predefined search limit is reached.}
\label{fig:step-level-search-with-verifiers}
\end{figure}

Clearly, this search framework is very similar to that used in decoding methods for LLMs, as discussed in Section~\ref{sec:llm-decoding-algorithms}. For example, beam search maintains a set of $K$ most promising partial sequences at each generation step. This is a form of step-level search where the ``verifier'' is implicitly the LLM's own probability model, and the ``search'' is the pruning mechanism to maintain the beam size.

However, step-level search with explicit verifiers, as described here, is different from standard decoding methods. One key difference is that the verifier can be a much more sophisticated component than just the raw output probabilities of the generative LLM. The design of step-level verifiers follows a design similar to solution-level verification. A step-level verifier might be a language model that assesses the quality of an individual reasoning step within the context of the preceding path. This LLM can even be fine-tuned to enhance its verification capability. Alternatively, for domains with well-defined rules, it could be a symbolic engine or a set of programmatic checks. Furthermore, verifiers can be designed to predict the future utility or likelihood of success given the current partial path, drawing inspiration from value functions in reinforcement learning. Human expertise can also be incorporated to provide judgments on critical steps, especially in high-stakes scenarios.

One example of such a step-level verifier, particularly when using human feedback to assess intermediate progress, is the \mindex{process reward model} (\mindex{PRM}).  A PRM is typically a separate language model trained to output a scalar reward for each reasoning step $a_{i'}$ within a partial path $\mathbf{a}_{\le i}$. It provides a more direct and fine-grained supervisory signal compared to \mindex{outcome reward models} (\mindex{ORMs}) which only evaluate the final solution. However, the development of PRMs relies on step-level human annotations, such as preferences on different next steps. Collecting supervision for each intermediate step is considerably more labor-intensive and requires greater cognitive effort from human annotators than simply labeling final outcomes.

One alternative approach to developing training data for step-level verification is to use LLMs to generate such annotations automatically. For example, we can take a strong LLM, referred to as a teacher model, and prompt it to first generate a complete reasoning path for a given problem. Then, at each intermediate step within this path, we can prompt the same teacher LLM (or another capable LLM) to generate several alternative candidate next steps in addition to the one it originally chose. The teacher LLM can then be prompted again to evaluate these alternatives. These evaluation results (e.g., correct vs. incorrect) can then serve as data annotations. Alternatively, the generalization capabilities of PRMs can be leveraged. We can train a PRM on tasks where step-level verification is easier and then generalize this PRM to other tasks with little or no additional training.

Note that step-level verification also comes with its own problems. Frequent verification, especially if using an LLM as the verifier, can substantially increase computational costs and latency. The design of effective step-level verifiers is non-trivial itself. An inaccurate verifier might prematurely discard good reasoning paths or fail to identify flawed ones, thereby misleading the search. This makes the development of such systems more difficult.

\subsubsection{Encouraging Long Thinking}

So far in this subsection, most of the methods are implicitly based on a simple idea: generating longer reasoning paths can help. In addition to CoT and search with verification, we can consider alternative methods to achieve this. For example, we can prompt the LLM by explicitly asking for extended deliberation. Beyond direct prompting, we can also make modifications to the decoding process itself, such as adjusting token limits or applying penalties for short outputs. Another approach is to employ multi-stage generation schemes where the model incrementally builds upon its reasoning.

\subsubsection{Training-based Scaling}

As well as considering inference-time scaling methods without training, we also wish to consider methods that can improve intrinsic reasoning capabilities of LLMs by modifying their parameters through further training. While such training-based scaling methods typically require additional training cost and computational resources, they embed stronger reasoning skills directly into the model parameters, which in turn can lead to improved reasoning performance. We can even combine them with training-free methods to further improve inference-time scaling.

Although our discussion here is restricted to reasoning problems, methods for training-based scaling are common. Most of them have been discussed in Chapter \ChapterLLMTuning. Here, we will briefly describe how these methods can be applied to improving inference-time scaling for reasoning problems.

\begin{itemize}
\item \vspace{0.5em} \textbf{Fine-tuning on Reasoning Data}. One of the most direct ways to enhance reasoning is by fine-tuning pre-trained LLMs on datasets specifically curated for reasoning tasks. These datasets can range from simple input-output pairs to more structured formats that include step-by-step reasoning processes. Typical examples include datasets of math word problems, logical deduction exercises, or code generation with explanations. By training on such data, the model learns from common reasoning patterns, and thus can generate detailed and coherent reasoning paths at test time.
\item \vspace{0.3em} \textbf{Reinforcement Learning for Reasoning}. If we regard a verifier as a reward model, we can see that the methods discussed in the previous subsection are a direct application of the reward model to reasoning problems, though they are training-free. Of course, we can apply this reward model to LLM fine-tuning. This follows a standard paradigm of reinforcement learning. Given a reward model, the LLM, acting as a policy, is fine-tuned using reinforcement learning algorithms. The LLM generates reasoning steps or full solutions, receives feedback (rewards) from the reward model, and updates its parameters to produce outputs that maximize these rewards. This process aligns the LLM outputs with high-quality reasoning behavior, thereby encouraging the LLM to generate more reliable reasoning paths. Another key problem is the training of the reward model. Generally, this reward model could be an outcome reward model that evaluates the correctness or quality of the final answer, or a process reward model that assesses the quality of each intermediate reasoning step, as discussed in the context of step-level verifiers. In some cases, we can even develop a reward model based on simple rules, such as giving bonuses to longer outputs.
\item \vspace{0.3em} \textbf{Knowledge Distillation for Reasoning}. In this approach, a smaller, more efficient student LLM is trained to mimic the reasoning outputs or internal representations of a larger, more capable teacher LLM. The teacher model might generate detailed reasoning steps for a variety of problems. The student model then learns to reproduce these high-quality reasoning demonstrations. This strategy makes stronger reasoning capabilities more accessible by deploying them in smaller models that are less computationally expensive at inference time.
\item \vspace{0.3em} \textbf{Iterative Refinement}. Training-based scaling can also involve iterative refinement. For example, an LLM can generate solutions to a set of problems. These solutions and their reasoning paths are then verified, either by humans or automatic verifiers. The correct reasoning paths are subsequently added to the training data, and the LLM is further fine-tuned on this augmented dataset. This creates a cycle where the LLM progressively improves its reasoning capabilities through repeated generation, critique, and learning.
\end{itemize}
\vspace{0.5em}

The primary advantage of these training-based scaling methods is that they endow the LLM with stronger inherent reasoning skills. This directly contributes to improved inference-time scaling in several ways: it can lead to more efficient inference, as the LLM might require less extensive search or fewer generation samples to arrive at a correct solution. Moreover, the base quality of generated steps or solutions is higher. Therefore, a well-trained LLM might generalize its learned reasoning abilities to novel problems more effectively than an LLM relying solely on in-context learning or training-free inference schemes.

On the other hand, training-based approaches also present challenges, compared to the training-free counterparts. The creation of high-quality, large-scale training datasets for reasoning can be expensive and labor-intensive. The fine-tuning process itself, particularly for the largest LLMs or when using RL, can be computationally intensive and require substantial engineering effort. There is also the risk of the model overfitting to the specific types of problems or reasoning styles present in the training data, potentially limiting its performance on out-of-distribution tasks.

%----------------------------------------------------------------------------------------
% NEW SECTION
%----------------------------------------------------------------------------------------

\section{Summary}

In this chapter, we have discussed inference for LLMs. We have presented the prefilling-decoding framework and related decoding algorithms for LLM inference. Then, we have described several techniques for efficient inference. We have also discussed inference-time scaling, which is one of the most important approaches to improving LLM reasoning.

Inference over sequential data has long been a fundamental topic in AI~\cite{wozengraft-and-reiffen:1961,viterbi:1967error,forney-etal:1972maximum}. In the context of NLP, this line of work dates back to the very early days of speech recognition and statistical machine translation~\cite{Koehn:2010}, where researchers faced the challenge of efficiently searching vast hypothesis spaces to find the most probable output sequence. Techniques like beam search and various pruning strategies were developed then to make this computationally tractable. At that time, models were relatively weak, and much of the research focused on developing powerful search algorithms to reduce search errors. These foundational ideas continue to influence modern approaches.

As we enter the era dominated by deep learning methods, models based on deep neural networks have become extremely powerful. Even with very simple search algorithms, these models can achieve excellent results. In this context, inference no longer seems as ``important'' as it once was, and research attention has gradually shifted toward model architectures, training methods, and scaling up models.

However, this trend has recently shifted. With the rise of LLMs, inference has once again attracted significant attention. This renewed focus is primarily reflected in two aspects:

\begin{itemize}
\item \vspace{0.5em} The inference cost for LLMs is very high. For example, efficiently deploying LLMs in high-concurrency, low-latency scenarios remains a challenging problem, making inference efficiency critically important. In this context, efficient architecture designs, optimized search algorithms, and various inference optimization strategies hold substantial practical significance.
\item \vspace{0.3em} Input and output sequence lengths have significantly increased in complex tasks. Especially in tasks like mathematical reasoning, the growth of sequence lengths further highlights the importance of inference efficiency. Moreover, scaling the inference process has recently proven to be an effective way to improve the reasoning capabilities of models. Therefore, achieving efficient inference scaling is emerging as a particularly promising research direction.
\end{itemize}
\vspace{0.5em}

Inference is now a wide-ranging topic that encompasses many techniques. It involves not only the development of model architectures and decoding algorithms, but is increasingly shaped by the intricate engineering and sophisticated systems-level optimizations required to deploy LLMs effectively and efficiently. Many of these techniques are beyond the scope of NLP or a specific AI area. Instead, the frontier of LLM inference optimization now extends deeply into domains traditionally considered core computer science and engineering.

% !Mode:: "TeX:UTF-8"
% !TEX encoding = UTF-8 Unicode

%----------------------------------------------------------------------------------------
% Natural Language Processing: Neural Networks and Large Language Models
%
% Copyright 2021-2026
% Tong Xiao (xiaotong@mail.neu.edu.cn) and Jingbo Zhu (zhujingbo@mail.neu.edu.cn)
%----------------------------------------------------------------------------------------

%----------------------------------------------------------------------------------------
%    CONFIGURATIONS
%----------------------------------------------------------------------------------------

%\setcounter{chapter}{11}

%----------------------------------------------------------------------------------------
%	CHAPTER 12
%----------------------------------------------------------------------------------------

\chapter{Reasoning}

\thispagestyle{chapterurl} % add urls to the header

In the previous chapters, we considered LLMs for general-purpose applications, where models generate responses based on input queries. These models can naturally handle a wide variety of tasks, such as machine translation and question answering. Therefore, many NLP problems can be framed as text generation tasks and addressed using LLMs. Beyond simple text generation, a natural next step is developing models capable of deeper thought to solve more complex problems. In particular, we wish that LLMs can generate intermediate thinking steps and thus exhibit reasoning abilities.

Reasoning has long been considered one of the most challenging problems in AI. Although the NLP community has attempted to address it using methods such as parsing sentences into logical forms, these efforts mainly relied on building reasoning models based on a series of pipeline modules, which are generally prone to cascading errors. In contrast, LLMs provide a more general framework for reasoning where the thinking path can be explicitly expressed as text and so we can enable LLMs to reason during text generation. While reasoning can be viewed as an emergent ability that arises from scaling LLMs, it requires specific strategies to invoke this capability, including CoT prompting, inference-time scaling, and training LLMs to reason.

In this chapter, we will discuss methods for improving the reasoning capabilities of LLMs. We begin by introducing basic concepts related to reasoning. Next, we present techniques for developing reasoning models. Since prompting and CoT methods have been extensively discussed in the previous chapters, here we focus on two lines of research: 1) scaling inference-time compute and 2) learning to reason. Finally, we summarize the chapter with some remarks.

%----------------------------------------------------------------------------------------
% NEW SECTION
%----------------------------------------------------------------------------------------

\section{Background and Concepts}

Here we define the reasoning problem and discuss related concepts.

\subsection{What is Reasoning?}

Historically, artificial intelligence research has categorized reasoning into three primary logical forms: \textbf{deduction} (deriving specific conclusions from known premises, such as mathematical proofs), \textbf{induction} (inferring general rules or patterns from specific examples), and \textbf{abduction} (finding the most plausible explanation for a set of observations)~\cite{peirce1931:collected}. For a long time, mainstream machine learning and NLP research primarily focused on inductive tasks. For instance, classic NLP applications like text classification and named entity recognition can be viewed as inductive tasks: models are trained to generalize mapping functions from large annotated datasets.

However, with the development of LLMs, researchers are attempting to address more difficult reasoning problems, such as complex mathematics, agentic planning, and scientific discovery, which often require rigorous deductive and abductive reasoning. Modern LLM research has revealed that the standard pre-training objective (i.e., next-token prediction over massive text corpora) is powerful enough to help models internalize knowledge of language and the world, including latent reasoning capabilities. The main challenge, however, is that reasoning is not explicitly modeled during the pre-training phase. As a result, when faced with a complex problem, LLMs attempt to output the final answer directly, rather than executing a step-by-step logical derivation.

For instance, consider a simple word problem:

\begin{quote}
\textit{If John's age is twice Mary's age, and Mary will be 10 years old in 3 years, how old is John now?}
\end{quote}

A base LLM might output an incorrect guess (e.g., 20) by simply associating the keywords ``twice'' and ``10''. In contrast, a reasoning-capable LLM might explicitly generate the logical chain before answering:

\begin{quote}
\textit{Mary is currently $10 - 3 = 7$ years old. Therefore, John is $7 \times 2 = 14$ years old.}
\end{quote}

To solve complex deductive and abductive tasks using LLMs, these latent reasoning abilities must be explicitly activated, whether through prompting techniques such as CoT or through advanced post-training paradigms such as reinforcement learning.

To better understand the reasoning problem, we can view LLM capabilities using the dual-process theory in cognitive psychology: System 1 and System 2 thinking~\cite{kahneman:2011thinking}.

\begin{itemize}
\item \vspace{0.5cm} \textbf{System 1 (Intuitive Thinking):} In human cognition, System 1 operates automatically and quickly, with little or no effort. In the context of LLMs, System 1 thinking corresponds to the paradigm where a model directly outputs the final answer in a single pass. This mode is efficient and generally sufficient for inductive tasks, such as simple question answering. However, System 1 thinking is prone to logical leaps and hallucinations when confronted with complex multi-step problems.
\item \vspace{0.3cm} \textbf{System 2 (Deliberative Thinking):} This mode is slow and conscious, requiring step-by-step logical derivation. For LLMs, System 2 thinking is engaged when the model is directed to allocate additional computational resources at inference time to explicitly generate intermediate reasoning steps. By representing the thought process as text, the LLM breaks down complex tasks, verifies intermediate conclusions, and maintains a coherent logical chain before arriving at the final answer.
\end{itemize}
\vspace{0.3cm}

For LLM reasoning, the core objective is to activate the System 2 capability. To understand this, we must look beyond traditional symbolic logic. In classic AI, reasoning is often defined as the application of hard-coded logical rules over structured knowledge bases. For LLMs, however, reasoning is an intrinsic text generation process. It can be defined as the ability to generate a coherent sequence of intermediate steps to reach the final answer. In a sense, reasoning in LLMs is a mechanism to dynamically allocate more inference compute to harder problems. Ultimately, reasoning for an LLM is the capacity to encode latent System 2 thinking into a sequence of thinking steps (reasoning steps).

\subsection{Why Reasoning Matters}

Reasoning is important in NLP research because of its application to several challenging domains:

\begin{itemize}
\item \vspace{0.5cm} \textbf{Mathematics:} Solving math problems requires strict logic. A model cannot reliably solve a problem by simply associating the numbers and keywords in the question with answers seen in similar examples. Instead, it needs to apply mathematical rules step by step to derive the correct result.
\item \vspace{0.2cm} \textbf{Coding:} Writing software is much harder than simply predicting the next word. The model must generate code based on specific user requirements and strict code logic. Especially when tackling complex problems, translating abstract requirements into functional code requires step-by-step reasoning.
\item \vspace{0.2cm} \textbf{Agentic Planning:} When an LLM is used as an autonomous agent, it has to break down big, complex goals into small, actionable steps. It also needs to plan ahead and adjust its actions if things go wrong.
\item \vspace{0.2cm} \textbf{Scientific Discovery:} Helping scientists requires looking at data, coming up with new hypotheses, and testing them carefully. This relies heavily on step-by-step logical deduction and abduction.
\end{itemize}
\vspace{0.3cm}

Beyond solving harder problems, generating reasoning steps offers practical advantages. Most importantly, it makes the model interpretable. Because the model writes down its thought process as text, humans can read it and check the logic. They can even find exactly where a mistake was made. This makes the system much easier to verify and trust. In addition, reasoning step-by-step helps reduce hallucinations. Instead of making wild guesses when faced with a hard question, the model grounds its final answer in its own logical steps, which makes the output much more reliable.

\subsection{From LLMs to Reasoning Models}

Before LLMs took over, reasoning in AI was mostly about strict rules and logic. Researchers used methods like symbolic AI and semantic parsing to translate human language into intended forms, such as knowledge graphs or database queries. While the logic in these classic systems was very clear and easy to track, the systems themselves were fragile. If they encountered a word they did not know, or if one small rule failed, the whole process would crash.

When LLMs arrived, the reasoning problem was reframed as a text generation task and could be better addressed using reasoning-oriented techniques. The development of reasoning models in the LLM era can be divided into three stages: directly guessing answers, being prompted to think, and finally, learning how to think.

Early language models were trained mainly to predict the next word. When asked a question, they tried to output the answer right away. This fast guessing worked well for simple questions or writing tasks. But for complex math or logic problems, it did not work well. The models often made mistakes or hallucinated because they did not have a way to pause and work through the problem step-by-step.

The first breakthrough came when researchers realized that LLMs already possessed hidden reasoning abilities: they just needed to be awakened. By simply using CoT prompting~\cite{wei-etal:2022chain} --- asking the model to ``think step by step'' --- the model was forced to write down its thought process before giving an answer. However, this method still relied heavily on human hints and did not actually teach the model the underlying rules of logic.

The bigger game-changer recently is using reinforcement learning to train models for reasoning. Instead of just teaching models to generate text, reinforcement learning methods train these models through trial and error. When a model finds the right logical path and gets the correct answer, it receives a reward. Over time, models like OpenAI's o1~\cite{openai:2024learning} and DeepSeek-R1~\cite{deepseek:2025r1} have shown impressive reasoning abilities, such as exploring different ideas, double-checking their own steps, and fixing mistakes on their own.

These self-correction and exploration abilities lead to an important research direction: inference-time scaling \cite{snell-etal:2024scaling}. The idea is straightforward: by allocating more computational power during generation, the model can think more deeply to solve harder problems. Ideally, a model would adjust its thinking time based on how difficult a problem actually is. However, many current reasoning models take a simpler approach --- they just apply a long thinking process to all queries.

Taking a broader view, scaling compute at inference time benefits reasoning in general. Even if an LLM has not been specifically trained to reason, we can still boost its problem-solving abilities by simply giving it more compute to ``think''. For example, we can ask the model to generate multiple different reasoning paths and vote on the best one, or we can use search algorithms to help it find the correct logical steps.

Reasoning models also motivate the recent development of agents based on LLMs \cite{yao-etal:2023react,wang-etal:2024survey}. For an AI agent to complete tasks on its own, it needs reasoning to figure out how to reach a goal step by step. Because real-world tasks rarely go perfectly, the agent also has to realize when it makes a mistake and know how to fix it. All of these require the agent to have good reasoning abilities.

The rapid development of reasoning models also presents interesting but challenging research directions \cite{li-etal:2025system,chen-etal:2026towards}. One direction focuses on efficiency. For example, if models learn to dynamically adjust their thinking time, they will not waste compute on simple questions, but will only slow down for the hard ones. Another direction is expanding reasoning beyond subjects with strict right-or-wrong answers, like math or coding, to handle open-ended real-world situations. In addition, enabling models to reason across different types of data, such as images, audio, and physical environments, is a key step toward building more advanced AI systems.

%----------------------------------------------------------------------------------------
% NEW SECTION
%----------------------------------------------------------------------------------------

\section{Reasoning via Prompting}

Prompting is the simplest method to elicit reasoning from LLMs. Because it does not require any modifications to model architectures or parameters, we can simply use well-crafted input texts to stimulate the LLM into outputting the reasoning process. One of the most widely used prompting methods is chain-of-thought (CoT) prompting. The core idea is to encourage the model to generate intermediate logical derivations before producing the final answer. This is typically achieved by providing a few step-by-step examples in the prompt, or by simply appending a direct instruction like ``Let's think step by step'' to the query. Below is an example of CoT prompting, where the CoT instruction is highlighted.

\vspace{0.5em}
\begin{tcolorbox}[frame empty]

\begingroup
\setlength{\leftskip}{2em}
\setlength{\rightskip}{2em}

Q: Tom bought 3 pairs of shoes for $\$40$ each and 2 shirts for $\$20$ each. If he uses a $10\%$ discount coupon on the entire purchase, how much does he pay in total?

\vspace{0.3em}

A: \ctext[RGB]{154, 245, 175}{Let's think step-by-step.}

\vspace{0.3em}

\uline{First, calculate the total cost of the shoes: $3$ pairs multiplied by $\$40$ equals $\$120$. Next, calculate the total cost of the shirts: $2$ shirts multiplied by $\$20$ equals $\$40$. The total cost before the discount is $\$120 + \$40 = \$160$. The discount is $10\%$ of $\$160$, which is $\$16$. Finally, subtract the discount from the total cost: $\$160 - \$16 = \$144$. So the answer is $\$144$.}

\endgroup

\end{tcolorbox}
\vspace{0.5em}

In this example, the LLM produces a reasoning path (underlined text) which contains both the reasoning process and the final answer. More formally, given a CoT prompt $\mathbf{x}$\footnote{Here a CoT prompt consists of both the user input and the CoT instruction.}, the LLM generates the most likely output by maximizing the language modeling score, like this
\begin{eqnarray}
\hat{\mathbf{y}} & = & \argmax_{\mathbf{y}} \Pr(\mathbf{y} | \mathbf{x}).
\end{eqnarray}

\noindent Here $\Pr(\mathbf{y} | \mathbf{x})$ is the text generation probability as in standard language models. Each output is a token sequence $\mathbf{y}=y_0...y_n$. It represents a reasoning path consisting of a sequence of reasoning steps:
\begin{eqnarray}
\mathbf{y} & = & \{\bar{\mathbf{y}}_1,...,\bar{\mathbf{y}}_{n_p}\}. \label{eq:reasoning-path-notation}
\end{eqnarray}

\noindent Here $n_p$ is the number of reasoning steps. Each step $\bar{\mathbf{y}}_i$ is a sub-sequence of $\mathbf{y}$, which represents a single logical deduction or intermediate calculation. Note that there are many strategies to divide a reasoning path into discrete steps, such as segmenting by individual mathematical operations. For simplicity, in this example, we treat each sentence as an individual reasoning step. Hence, we have

\[
\begin{array}{rl}
\bar{\mathbf{y}}_1: & \text{First, calculate the total cost of the shoes: }\nonumber \\
& \text{$3$ pairs multiplied by $\$40$ equals $\$120$.} \nonumber \\
\bar{\mathbf{y}}_2: & \text{Next, calculate the total cost of the shirts: } \nonumber \\
& \text{$2$ shirts multiplied by $\$20$ equals $\$40$.} \nonumber \\
\bar{\mathbf{y}}_3: & \text{The total cost before the discount is $\$120 + \$40 = \$160$.} \nonumber \\
\bar{\mathbf{y}}_4: & \text{The discount is $10\%$ of $\$160$, which is $\$16$.} \nonumber \\
\bar{\mathbf{y}}_5: & \text{Finally, subtract the discount from the total cost: $\$160 - \$16 = \$144$.} \nonumber \\
\bar{\mathbf{y}}_6: & \text{So the answer is $\$144$.} \nonumber
\end{array}
\]

Given such a reasoning path, we can extract the answer from its last step. There are many ways to improve this simple CoT-based reasoning approach. For instance, techniques like self-consistency sample multiple reasoning paths and take a majority vote to identify the most reliable answer \cite{wang-etal:2023self}. Other strategies include problem decomposition to break complex queries into manageable sub-tasks \cite{zhou-etal:least}, self-reflection methods that prompt the model to critique and correct its own logical errors \cite{shinn-etal:2023reflexion,madaan-etal:2024self}, and deliberate reasoning frameworks that efficiently explore multiple branching reasoning paths \cite{yao-etal:2024tree}.

As various CoT-based prompting techniques have been presented previously, we will not go into detail here. Readers are encouraged to review the techniques covered in Chapter \ChapterLLMPrompting.

%----------------------------------------------------------------------------------------
% NEW SECTION
%----------------------------------------------------------------------------------------

\section{Inference-time Reasoning}
\label{sec:inference-time-reasoning}

Inference-time reasoning is closely related to scaling during inference, which has been introduced in Chapter \ChapterLLMInference. In this section, we extend the previous discussion and focus more on improving the reasoning performance of LLMs.

\subsection{Inference-Time Compute Scaling}

The success of LLMs heavily depends on scaling up computing power. A common way to improve these models is training-time scaling. By simply using more training data, larger model sizes, and more compute, researchers have built very powerful LLMs \cite{kaplan-etal:2020scaling,hoffmann-etal:2022training}.

However, just scaling up the training process is not enough for complex reasoning tasks. Unlike simple text generation, reasoning needs step-by-step logic and planning. It is impossible to show a model all possible reasoning paths during training. If we only rely on training to solve hard math or logic problems, we would need a very large amount of data. Ideally, we want the model to use simple logical steps to construct its own reasoning path on the fly. Therefore, it is more appealing to activate  such reasoning abilities during inference.

Because of this, a natural idea is to scale compute during the inference phase. Thus, the LLM can spend more time exploring different reasoning steps before giving the final answer. A clear benefit of this approach is that it allows for a much more flexible and efficient use of computing resources. For example, for complex math or logic problems, the LLM can allocate more time and power, and thus ``think'' longer. In many cases, a smaller model that is given extra time to reason during inference can even outperform a much larger model that is forced to give an immediate answer. We summarize the differences between training-time scaling and inference-time scaling in Table \ref{tab:scaling-comparison}.

\begin{table}[t]
\centering
\begingroup
\renewcommand{\arraystretch}{1.3}
\begin{tabular}{R{0.20\linewidth} | L{0.33\linewidth} | L{0.33\linewidth}}
Feature & Training-Time Scaling & Inference-Time Scaling \\
\hline
How it Works & Making the model bigger and feeding it more data during the initial learning phase. & Keeping the model size fixed, but giving it more compute to "think" (e.g., exploring different paths) before answering. \\ \hline
Cost to Train & Extremely high. Very expensive and hard to fix if something goes wrong. & Low to medium. We may pay to train smaller models to check its answers. \\ \hline
Cost to Use & Predictable. The cost to generate a token is mostly based on the model's size and the length of the context. & Flexible but can get very high. The deeper the model ``thinks'', the more expensive it gets.\\ \hline
Hardware Needs (Inference) & Needs high memory bandwidth for fast data reading and writing. & Needs the same high bandwidth, plus large memory capacity to explore multiple long reasoning paths. \\ \hline
Data Needed & Massive amounts of general text. & High-quality, step-by-step examples (like math proofs) and a clear way to score if an answer is right or wrong.
\end{tabular}
\caption{Comparison of training-time and inference-time scaling}
\label{tab:scaling-comparison}
\endgroup
\end{table}

We can view inference-time scaling as a problem of \textbf{search scaling}. When we give the model extra compute to ``think'', we are essentially allowing it to explore a larger space of possible logical steps, and to have a much higher chance of finding the optimal solution. This  leads to three key questions:

\begin{itemize}
\item \vspace{0.5em} How to generate different reasoning paths?
\item \vspace{0.3em} How to search through possible steps to find the best one?
\item \vspace{0.3em} How to check if the intermediate steps and final answers are correct?
\end{itemize}
\vspace{0.5em}

In the next subsections, we will address these questions by discussing three topics: sampling multiple reasoning paths (Section~\ref{sec:sampling-reasoning-paths}), search strategies for reasoning (Section~\ref{sec:search-strategies-for-reasoning}), and verification (Section~\ref{sec:reasoning-verification}).

\subsection{Sampling Multiple Reasoning Paths}
\label{sec:sampling-reasoning-paths}

The most straightforward way to scale compute during inference is to generate not just one, but multiple reasoning paths for a single input. By generating multiple different paths, we give the LLM more chances to find the correct logic and the corresponding answer.

To achieve this, it is common practice to introduce randomness into the text generation process. A typical method is to use temperature sampling, combined with top-$k$ or nucleus (top-$p$) sampling. In standard auto-regressive generation, at each position $i$, the LLM outputs a probability distribution $\Pr(\cdot | \mathbf{x}, \mathbf{y}_{< i})$ over the vocabulary $V$ for the next token $y_i$ based on its logits $u_{y_i}$. Temperature sampling modifies this distribution by scaling the logits with a temperature parameter $T > 0$ before applying the Softmax function:
\begin{eqnarray}
\Pr(y_i | \mathbf{x}, \mathbf{y}_{< i}) & = & \frac{\exp(u_{y_i} / T)}{\sum_{y'_i \in V} \exp(u_{y'_i} / T)}.
\end{eqnarray}

\noindent A higher temperature flattens the distribution, thereby encouraging the model to explore a wider variety of reasoning steps. However, unconstrained randomness can lead to degenerate or illogical text by occasionally sampling from the ``long tail'' of low-probability tokens. To maintain the quality of the generated paths, temperature sampling is typically combined with vocabulary truncation methods:

\begin{itemize}
\item \vspace{0.5em} \textbf{Top-$k$ sampling} restricts the sampling pool to only the $k$ most likely next tokens, redistributing the probability mass among them.
\item \vspace{0.3em} \textbf{Nucleus (top-$p$) sampling} uses a dynamic truncation strategy. It selects the smallest set of top-ranked tokens whose cumulative probability exceeds a threshold $p$.
\end{itemize}
\vspace{0.5em}

By applying these stochastic decoding strategies, given an input $\mathbf{x}$, the LLM can sample a set of diverse output sequences (i.e., candidate reasoning paths):
\begin{eqnarray}
\mathcal{Y}^{\mathrm{top}} & = & \{\hat{\mathbf{y}}^1, \dots, \hat{\mathbf{y}}^N\}.
\end{eqnarray}

\noindent Here $N$ represents the inference-time compute budget. The larger $N$ is, the more compute we spend exploring different reasoning paths. Once we have this diverse set of paths, the next problem is how to select the best one. There are two widely used methods for this: Best-of-$N$ and self-consistency.

\subsubsection{Best-of-$N$ (BoN)}
We have previously introduced Best-of-$N$ sampling in Section \ChapterLLMTuning\ as a method for inference-time alignment. In that context, we used a reward model $r(\mathbf{x}, \hat{\mathbf{y}})$ to select the output that best aligns with human preferences.

For reasoning tasks, the concept of BoN sampling remains the same, but the goal shifts. Instead of a preference reward model, the ``judge'' is usually a verifier (which we can denote as a general scoring function $v$) designed to evaluate how logically sound a reasoning path is. The verifier assigns a score $v(\mathbf{x}, \hat{\mathbf{y}})$ to each sampled path, and we select the best reasoning path $\hat{\mathbf{y}}_{\mathrm{best}}$ that maximizes the score
\begin{eqnarray}
\hat{\mathbf{y}}_{\mathrm{best}} & = & \argmax_{\hat{\mathbf{y}} \in \{\hat{\mathbf{y}}^1, \dots, \hat{\mathbf{y}}^N\}} v(\mathbf{x}, \hat{\mathbf{y}}).
\end{eqnarray}

\noindent The final answer is then extracted from this highest-scoring path.  To understand the BoN process, let us consider an intuitive analogy. Imagine there is a classroom of $N$ students solving a math problem	. A teacher (the verifier) grades all their scratchpads, and then picks the student with the highest score.

While BoN sampling is powerful, its effectiveness depends on the quality of the verifier. If the verifier is weak, it can easily be fooled by flawed logic that accidentally arrives at the correct final answer (i.e., reward hacking). The design of verifiers will be discussed in more detail in Section~\ref{sec:reasoning-verification}.

\subsubsection{Self-Consistency}
An alternative approach to selecting the best path via a learned verifier is self-consistency. Self-consistency relies on the intuition that in complex reasoning tasks, there are often multiple different logical paths that lead to the same correct answer, whereas incorrect reasoning paths are likely to be diverse and lead to random errors. Instead of scoring the reasoning process, self-consistency directly marginalizes out the reasoning paths by taking a majority vote over the final answers. Formally, let $A$ be the set of all possible final answers. For each sampled path $\hat{\mathbf{y}}^k$, we extract its final predicted answer $\hat{a}^k = \mathrm{Extract}(\hat{\mathbf{y}}^k)$. The self-consistency strategy selects the answer $\hat{a}_{\mathrm{sc}}$ that appears most frequently among the $N$ sampled paths:
\begin{eqnarray}
\hat{a}_{\mathrm{sc}} & = & \argmax_{a \in A} \sum_{k=1}^N \mathbbm{1}(\mathrm{Extract}(\hat{\mathbf{y}}^k) = a),
\end{eqnarray}

\noindent where $\mathbbm{1}(\cdot)$ is the indicator function.

\subsubsection{Balancing Diversity and Quality in Sampling}

When sampling multiple reasoning paths for inference-time compute scaling, we need to seek a trade-off between diversity and quality. In reasoning tasks, we often desire a diverse set of sampled paths, which increases the chance of containing the correct logic. A common heuristic is increasing the sampling temperature. By making the distribution flatter, the sampling process avoids concentrating only on a few high-probability paths. As a result, lower-probability paths are more likely to be chosen, and the model can explore a wider variety of reasoning paths within the $N$ samples. However, as temperature increases, although the diversity of the samples improves, the average logical quality of individual paths may degrade.

To achieve a balance between diversity and quality, a straightforward method is to modify the search objective during the inference stage by introducing an explicit penalty or bonus term. In this setup, diversity can be mathematically modeled as a function that is combined with the original generation probability. More formally, the modified search objective can be written as
\begin{eqnarray}
\mathrm{score}(\mathbf{x},\mathbf{y}) & = & \log \Pr(\mathbf{y}|\mathbf{x}) + \lambda D(\mathbf{y},\mathcal{Y}^{\mathrm{top}}). \label{diversity-quality-inference}
\end{eqnarray}

\noindent Here $\log \Pr(\mathbf{y} | \mathbf{x})$ favors outputs with higher probabilities, while $D(\mathbf{y},\mathcal{Y}^{\mathrm{top}})$ serves as the penalty or bonus term that encourages diversity relative to the sampled paths $\mathcal{Y}^{\mathrm{top}}$. The hyperparameter $\lambda$ controls the strength of this trade-off.
The design of $D(\cdot)$ can be quite flexible. For example, a simple penalty might decrease the scores of tokens or $n$-grams that have already appeared frequently in $\mathcal{Y}^{\mathrm{top}}$. Conversely, a bonus function could be designed to reward distinct semantic differences. Thus, the LLM can produce new reasoning structures rather than minor rephrasings of the same idea.

From a broader perspective, this sampling dilemma is also related to the exploration--exploitation trade-off studied in reinforcement learning \cite{Sutton-and-Barto:2018RL}. In RL-based tuning, an LLM can be viewed as a policy $\pi_\theta$ that generates a reasoning path $\mathbf{y}$ for a question $\mathbf{x}$. A general regularized objective can be written as
\begin{equation}
\max_{\pi_\theta} \mathbb{E}_{\mathbf{y}\sim\pi_\theta} \left[R(\mathbf{x},\mathbf{y})\right] - \beta D_{\mathrm{KL}} \left( \pi_\theta(\cdot|\mathbf{x}) \,\|\,  \pi_{\mathrm{ref}}(\cdot|\mathbf{x}) \right) + \alpha \mathrm{Entropy} \left( \pi_\theta(\cdot|\mathbf{x}) \right),
\label{diversity-quality-rl}
\end{equation}

\noindent where $R(\mathbf{x},\mathbf{y})$ measures the quality of the generated reasoning path. The KL-divergence term constrains the learned policy to remain close to the reference policy. Its main purpose is to prevent excessive policy drift, and thus reduce undesirable behaviors such as reward hacking or degradation of language quality. By contrast, the entropy term explicitly encourages the policy to maintain a less concentrated distribution and therefore promotes exploration during training.

There is a conceptual connection between the RL objective and the diversity--quality trade-off at inference time, although the two objectives operate in different spaces. At inference time, the search procedure controls how broadly it explores reasoning paths generated by a fixed model. In RL-based tuning, the KL regularization controls how far the learned policy is allowed to move away from the reference policy. A larger KL coefficient keeps the policy close to the reference model, whereas a smaller coefficient permits the optimization process to explore a broader region of the policy space in search of higher-reward behaviors. Such policy changes may enable the model to generate reasoning paths that are unlikely under the reference policy and may therefore lead to a wider range of outputs.

Both inference-time search and KL-regularized RL involve a balance between conservative behavior and broader exploration. The difference is that inference-time methods explore different outputs under a fixed model, whereas RL searches the policy space by updating the model parameters. In either case, allowing broader exploration may discover better and more varied reasoning behaviors, but excessive exploration may reduce quality or stability.

\subsection{Search Strategies for Reasoning}
\label{sec:search-strategies-for-reasoning}

In practice, finding the optimal reasoning path can be directly cast as a classic search problem: we start from an initial state, take an action to generate a reasoning step, expand the current state into a new set of states, and repeat this process until a termination condition is met. Since we have already introduced several classic search algorithms (such as tree search and beam search) in Chapter \ChapterLLMInference, we will not revisit the details of these algorithms here. Instead, we focus on how to map the search for reasoning paths onto these classic search frameworks.

To apply a search algorithm to LLM reasoning, we need to define the standard components of a search problem:

\begin{itemize}
\item \vspace{0.5em} \textbf{Search State}: For LLM reasoning, the state is the context the model has built so far. Specifically, at step $t$, the state $s_t$ consists of the original query $\mathbf{x}$ combined with the partial reasoning path generated up to this point: $s_t = (\mathbf{x}, \bar{\mathbf{y}}_1, \dots, \bar{\mathbf{y}}_t)$.
\item \vspace{0.3em} \textbf{Action}: An action is the generation of the next reasoning step (text segment) $\bar{\mathbf{y}}_{t+1}$.
\item \vspace{0.3em} \textbf{Transition Model}: Taking an action simply appends the new text segment to the existing state. This produces the next state: $s_{t+1} = (s_t, \bar{\mathbf{y}}_{t+1}) = (\mathbf{x}, \bar{\mathbf{y}}_1, \dots, \bar{\mathbf{y}}_t, \bar{\mathbf{y}}_{t+1})$.
\item \vspace{0.3em} \textbf{State Evaluation Function}: This is a scoring function that assesses the logical correctness of the current state toward the final answer. For example, to evaluate search states, we can use hard-coded heuristics, or a trained neural verifier.
\item \vspace{0.3em} \textbf{Termination Condition}: The criteria for stopping the search. This occurs when the model explicitly generates a final answer sequence, or when the search reaches a predefined computational budget (e.g., a maximum depth of explored nodes).
\end{itemize}
\vspace{0.5em}

Once the reasoning task is framed as a search problem, various search algorithms can be applied to explore the reasoning space efficiently. For instance, simple tree search methods, such as breadth-first search and depth-first search, can be used to model the LLM's reasoning process. In these methods, the model explores multiple candidate thoughts at each step and evaluates them to decide whether to continue down a path or backtrack when a dead end is reached. For tasks with higher uncertainty or delayed rewards, Monte Carlo tree search is often employed. This method helps the model balance exploring new reasoning paths and exploiting known steps. Sometimes it even integrates external environmental feedback and self-reflection to refine the search \cite{yao-etal:2024tree,hao-etal:2023reasoning,zhou-etal:2024language}.

To address the redundancy of tree structures, where multiple different reasoning paths might converge on the same intermediate conclusion, some methods model the reasoning space as a graph (such as DAG) \cite{zhang-etal:2025cumulative}. In this way, the model can build more efficient search space structures, and capture the non-linear interdependencies of logical deduction. Beyond external search algorithms, it is also possible to internalize search dynamics directly into the parameters of LLMs. For example, by training LLMs on the execution traces of heuristic search algorithms like A*, they can learn to implicitly generate the search process and optimal plans without relying on an external control script \cite{lehnert-etal:2025beyond}.

While scaling inference-time compute through search is a powerful strategy, the relationship between search effort and reasoning performance is not simply linear. Intuitively, we might expect that giving a model more time to explore will consistently yield better answers. However, in practice, the performance gains quickly plateau. One major reason is the inherent capability ceiling of the LLM: if the model lacks the knowledge to solve a problem, no amount of search will create the correct logic out of thin air. In addition, as search effort increases, the diversity of the generated paths drops, meaning the model wastes compute exploring redundant ideas that offer little new value.

\subsection{Verification}
\label{sec:reasoning-verification}

Next, we discuss the verification problem in reasoning. Note that while we focus on inference-time methods in this section, many verifiers used in reasoning models still require training. Therefore, we also present some details on training these models in the following discussion.

\subsubsection{Process Reward Models}

There are two main paradigms for verifying reasoning results: \mindex{outcome reward models} (ORMs) and \mindex{process reward models} (PRMs) \cite{uesato-etal:2022solving,lightman-etal:2024lets}. An ORM evaluates only the final answer of a reasoning path. While easier to train, ORMs are often insufficient for reasoning tasks because they suffer from ``false positives''. To illustrate this, consider a simple math problem:

\vspace{0.5em}
\begin{quote}
\textit{Solve for $x$: $2x + 4 = 10$.}
\end{quote}

\vspace{0.5em}
A model might generate the following reasoning path:
\vspace{0.5em}

\begin{quote}
\textit{$2x + 4 = 10$ means $x$ must be half of $10$, which is $5$.}

\textit{But $5 - 2 = 3$.}

\textit{Therefore, $x = 3$.}
\end{quote}
\vspace{0.5em}

Although the final answer (i.e., $3$) is correct, the logical steps are completely flawed. If we only use an ORM, this path might receive a high score, and accidentally reinforce incorrect logic. To address this, we can use PRMs to evaluate the correctness of each individual step in the reasoning path. Given a reasoning path $\mathbf{y} = \{\bar{\mathbf{y}}_1, \dots, \bar{\mathbf{y}}_{n_p}\}$, a PRM assigns a correctness score $r_t$ to the current step $\bar{\mathbf{y}}_t$, conditioned on the input $\mathbf{x}$ and all previous steps $\bar{\mathbf{y}}_{<t}$. We can define the step-level verification function as
\begin{eqnarray}
r_t & = & v_{\mathrm{prm}}(\bar{\mathbf{y}}_t | \mathbf{x}, \bar{\mathbf{y}}_{<t}).
\end{eqnarray}

\noindent Here $v_{\mathrm{prm}}(\cdot)$ is a neural network that is trained to predict the correctness score. For example, we can attach a simple classification head (such as a linear layer followed by a sigmoid activation function) to the output representation of an existing LLM.

To determine the overall quality of the entire reasoning path, we can aggregate these step-level scores. A simple approach is to take the minimum score across all steps. In other words, we only consider the reasoning correct if its ``weakest link'' is logically sound. The verification score is then given by
\begin{eqnarray}
v(\mathbf{x}, \mathbf{y}) & = & \min_{1 \le t \le n_p} v_{\mathrm{prm}}(\bar{\mathbf{y}}_t | \mathbf{x}, \bar{\mathbf{y}}_{<t}).
\end{eqnarray}

\noindent Alternatively, we can define the verification score as the average of the step-level scores
\begin{eqnarray}
v(\mathbf{x}, \mathbf{y}) & = & \frac{1}{n_p}\sum_{1 \le t \le n_p} v_{\mathrm{prm}}(\bar{\mathbf{y}}_t | \mathbf{x}, \bar{\mathbf{y}}_{<t}).
\end{eqnarray}

To train a verification model, we need an annotated dataset. Unlike training an ORM, which only requires knowing the final correct answer at the very end of the path, training a PRM requires step-level supervision. There are two commonly used ways to construct a PRM training dataset:

\begin{itemize}
\item \vspace{0.5em} \textbf{Human Annotation:} Human experts manually solve problems and label each intermediate step generated by an LLM as ``correct'' or ``incorrect''. While this method can yield accurate verifiers, it is expensive and difficult to scale, especially for advanced mathematics or coding tasks that require domain experts.
\item \vspace{0.3em} \textbf{Automated Verification via Rollouts:} We can also use automated methods to estimate the value of a step \cite{wang-etal:2024math}. For instance, if we have a dataset with known final answers (like math competitions), we can use \mindex{Monte Carlo rollouts}. From an intermediate step $\bar{\mathbf{y}}_t$, we prompt the LLM to continue generating multiple completed paths. If a high percentage of these paths eventually reach the correct final ground-truth answer $a^*$, we can assume that the intermediate step $\bar{\mathbf{y}}_t$ is on the right track. Formally, the estimated value $\hat{r}_t$ of a step can be defined as the expected probability of reaching the correct answer:
    \begin{eqnarray}
    \hat{r}_t & = & \mathbb{E}_{\mathbf{y}_{>t} \sim \pi_\theta} [ \mathbbm{1}(\mathrm{Extract}(\mathbf{y}_{>t}) = a^*) ].
    \end{eqnarray}
    \noindent Here $\mathbf{y}_{>t}$ is the remaining path from $\bar{\mathbf{y}}_t$, and $\mathrm{Extract}(\mathbf{y}_{>t})$ represents the answer extracted from the path $\mathbf{y}_{>t}$. Once these values are estimated through rollouts, a PRM can be trained to predict $\hat{r}_t$.
\end{itemize}\vspace{0.5em}

\subsubsection{Generative Verifiers}

All the verifiers mentioned above require additional architecture modification and training effort, for example, we need to train a classification head to output a numerical score. Given the strong text generation ability of LLMs, another natural approach is to directly prompt LLMs to evaluate the reasoning steps and generate verification feedback in natural language. This approach is known as \textbf{generative verification} \cite{zheng-etal:2023judging}.

Instead of modifying the model architecture to output a single scalar value, we formulate the verification task as a standard text generation problem. By feeding the query and the intermediate steps back into the LLM, we can ask it to serve as a judge. This means that the model acts like a teacher writing comments in the margins of a test paper, rather than just assigning a final grade.

To understand how this works, let us look at a simple example. Suppose the model is solving a system of equations:
\vspace{0.5em}
\begin{quote}
\textit{Problem: Solve for $x$ and $y$ given $2x + y = 10$ and $x - y = 2$.}
\end{quote}
\vspace{0.5em}

We assume that the model has already generated two correct steps:
\vspace{0.5em}
\begin{quote}
\textit{From the second equation, we can express $x$ in terms of $y$: $x = y + 2$.}

\textit{Substitute $x$ into the first equation: $2(y + 2) + y = 10$.}
\end{quote}
\vspace{0.5em}

Now it is attempting a third step which contains a calculation error:

\vspace{0.5em}
\begin{quote}
\textit{Expand the equation to get $2y + 2 + y = 10$, which simplifies to $3y + 2 = 10$.}
\end{quote}
\vspace{0.5em}

We can combine the original query, steps 1 and 2, and the current step 3 with a verification prompt (e.g., ``\textit{Given the previous steps, check if the logic in the current step is correct and explain your conclusion.}''). The generative verifier might then output a textual critique like this:
\vspace{0.5em}
\begin{quote}
\textit{Critique: Incorrect. In the current step, when distributing the $2$ over $(y + 2)$ from Step 2, you must multiply both terms inside the parentheses by $2$. It should be $2y + 4$, not $2y + 2$. The correct expansion is $2y + 4 + y = 10$, which simplifies to $3y + 4 = 10$.}
\end{quote}
\vspace{0.5em}

To formalize this generative verification process, we assume that the original input $\mathbf{x}$ consists of a query $\mathbf{z}$ and an instruction $\mathbf{c}$ that asks the model to verify the step based on the context. We can concatenate the query $\mathbf{z}$, the reasoning history $\bar{\mathbf{y}}_{\le t}$, and the instruction $\mathbf{c}$ into a long text. Then, we feed this text as input to an LLM, and obtain a natural language critique $\mathbf{p}_t$:
\begin{eqnarray}
\mathbf{p}_t & \sim & \Pr(\cdot | \mathbf{z}, \bar{\mathbf{y}}_{\le t}, \mathbf{c}).
\end{eqnarray}

\noindent Here $\Pr(\cdot)$ represents the probability distribution modeled by an existing LLM. Thus we can directly use strong LLMs to give accurate verification without training. Figure \ref{fig:pre-generative-verfication} compares this generative approach with traditional discriminative verifiers.

\begin{figure}[!t]
\centering
\input{./Chapter12/Figures/figure-generative-verfication}
\caption{Comparison between discriminative and generative verification approaches. Subfigure (a) shows a discriminative verifier which attaches a classification head to the last hidden layer of the LLM to output a numerical probability of correctness. Subfigure (b) shows a generative verifier that prompts the LLM with the query $\mathbf{z}$, reasoning history $\bar{\mathbf{y}}_{\le t}$, and an instruction $\mathbf{c}$ to generate a natural language critique instead of a scalar value.}
\label{fig:pre-generative-verfication}
\end{figure}

While natural language critiques are useful for humans, some search algorithms (such as tree search) may require numerical scores to rank different reasoning paths. To extract a continuous score from a generative verifier, a common method is to measure the probability of the model outputting an indicator token (such as ``Correct'' or ``Incorrect'') as the very first word of its critique. Therefore, the numerical verification score can be defined as:
\begin{eqnarray}
v(\mathbf{z}, \bar{\mathbf{y}}_{\le t}, \mathbf{c}) & = & \Pr(y_{\mathrm{anchor}} = \text{``Correct''} | \mathbf{z}, \bar{\mathbf{y}}_{\le t}, \mathbf{c}).
\end{eqnarray}

\noindent Here $y_{\mathrm{anchor}}$ is the indicator token that is typically extracted from the output text using heuristics.

Generative verifiers have several advantages over traditional scalar reward models. First, they provide interpretability for problem solving. Because the verifier explains its reasoning in text, developers can diagnose why a reasoning path is accepted or not. Second, they enable self-correction. Since the text critique often explicitly states how to fix a mistake (like providing the correct expansion $2y + 4$ in our example), the reasoning system can directly integrate this feedback to correct its current state and continue searching, rather than simply discarding the entire path. Third, they require minimal modification of the architecture. As long as current LLMs have strong reasoning capabilities, the exact same model can be used as both the problem solver and the verifier.

\subsubsection{Formal Symbolic Systems as Verifiers}

While neural verifiers are widely used, they share a common weakness: they are still neural networks. This means they can also hallucinate or make logical mistakes. For tasks that require strict precision, such as complex mathematics or software engineering, we can also use formal symbolic systems as verifiers. For example, in code optimization tasks, when a model attempts to optimize intermediate representation code (such as LLVM-IR), a formal equivalency checker (such as Alive2) can automatically verify if the newly generated code maintains the exact same semantics as the original.

Formal systems strictly follow pre-defined mathematical and logical rules, and so their judgments are 100\% reliable. To use a formal system as a verifier, the LLM must generate its reasoning steps as executable code or formal mathematical statements rather than just plain natural language. For example, when solving a math problem, the LLM might write a small Python script to perform a specific intermediate calculation. The system then executes this code using an external tool. The verification process is very straightforward:

\begin{itemize}
\item \vspace{0.5em} If the code runs without errors and produces a valid output, the reasoning step is verified as correct.
\item \vspace{0.3em} If the code throws an error (such as a syntax error or a math violation), the reasoning step is verified as incorrect.
\end{itemize}
\vspace{0.5em}

Formally, let an intermediate step $\bar{\mathbf{y}}_t$ be a generated block of code or formal logic, and let $E$ represent the formal execution environment. The verification score can be defined as:
\begin{eqnarray}
v(\mathbf{x}, \bar{\mathbf{y}}_{<t}, \bar{\mathbf{y}}_t) & = & \begin{cases}
1, & \text{if } E(\bar{\mathbf{y}}_t) \text{  returns the correct result,} \\
0, & \text{otherwise.}
\end{cases}
\end{eqnarray}

This approach has a clear advantage: it eliminates the ``false positives'' caused by verifier hallucinations. Furthermore, when the execution fails, the formal system usually generates an error message. This error message can be fed back into the LLM to help the model understand what went wrong and correct its logic in the next attempt.

However, using formal symbolic systems also comes with limitations. First, it requires the LLM to have strong coding or formalization skills, as translating abstract reasoning into strict code is difficult. Second, while formal systems are perfect for math and coding, they are not suitable for open-ended, real-world tasks (like writing an essay or planning a trip) where reasoning cannot be easily mapped to rigid logical rules.

%----------------------------------------------------------------------------------------
% NEW SECTION
%----------------------------------------------------------------------------------------

\section{Learning to Reason}

Improving the reasoning capabilities of LLMs is challenging, and therefore generally requires a combination of multiple methods. As discussed in the previous section, inference-time scaling aims to activate the ``thinking'' process of existing LLMs during generation. Alternatively, a natural idea is to internalize these reasoning abilities directly into the model parameters during training. In this section, we will focus on this ``learning to reason'' paradigm.

Reasoning data can be incorporated into any training phase. For example, during pre-training, reasoning paths can simply be added as normal text for standard next-token prediction. Also, we can consider reasoning-based training in the post-training phase. For example, during supervised fine-tuning, models can be fine-tuned on high-quality instances of step-by-step problem-solving. This is commonly achieved by training the model on datasets with CoT trajectories. Furthermore, reinforcement learning can be applied to further optimize these reasoning processes. Here we focus on incorporating reasoning-based supervision in LLM alignment. In the following subsections, we will discuss supervised reasoning data, process supervision, reinforcement learning techniques, and strategies for scaling reasoning capabilities.

\subsection{Supervised Reasoning Data}

Formally, in supervised fine-tuning, an LLM is trained on a dataset of input-output pairs $\mathcal{D} = \{(\mathbf{x}, \mathbf{y})\}$. The standard training objective is
\begin{eqnarray}
\hat{\theta} & = & \argmax_{\theta} \sum_{(\mathbf{x}, \mathbf{y}) \in \mathcal{D}} \log \mathrm{Pr}_{\theta} (\mathbf{y}|\mathbf{x}),
\end{eqnarray}

\noindent where $\mathrm{Pr}_{\theta}(\cdot)$ denotes the model parameterized by $\theta$.

If $\mathbf{y}$ is simply the final conclusion, the model learns to predict the answer without explicitly generating the corresponding reasoning path. For reasoning-based training/tuning, we instead redefine $\mathbf{y}$ as a reasoning path. As shown in Eq. (\ref{eq:reasoning-path-notation}), we can represent the output as a reasoning path consisting of a sequence of reasoning steps $\mathbf{y} = \{\bar{\mathbf{y}}_1,...,\bar{\mathbf{y}}_{n_p}\}$. Thus, we can rewrite the above objective as
\begin{eqnarray}
\hat{\theta} & = & \argmax_{\theta} \sum_{(\mathbf{x}, \{\bar{\mathbf{y}}_1,...,\bar{\mathbf{y}}_{n_p}\}) \in \mathcal{D}} \log \mathrm{Pr}_{\theta} (\bar{\mathbf{y}}_1,...,\bar{\mathbf{y}}_{n_p}|\mathbf{x}).
\end{eqnarray}

\noindent In this way, the model is explicitly provided with reasoning details and is trained to generate a complete reasoning path, including the final conclusion.

In practice, to separate the core reasoning process from other parts of the output, reasoning models often use special formatting tokens. A widely adopted approach is to wrap the core reasoning steps inside \texttt{<think>} and \texttt{</think>} tags. When an LLM generates text using this format, the reasoning steps written inside the tags are referred to as a \mindex{CoT trace} (or simply a \mindex{reasoning trace}). We can think of this CoT trace as the scratchpad of the LLM. Just like a student taking a math exam who writes down intermediate calculations on scrap paper before writing the final answer on the official exam sheet, the LLM uses the \texttt{<think>} block to ``show its work''.

For example, a training instance formatted with these tags might look like this:

\vspace{0.5em}
\begin{tcolorbox}[frame empty]
\begingroup
\setlength{\leftskip}{2em}
\setlength{\rightskip}{2em}

What is $15\%$ of $80$?

\vspace{0.3em}

\texttt{<think>}

To calculate this easily, I can break $15\%$ down into $10\%$ and $5\%$.

First, $10\%$ of $80$ is $8$.

Next, $5\%$ is half of $10\%$, which is $4$.

Adding them together: $8 + 4 = 12$.

\texttt{</think>}

The answer is 12.

\endgroup
\end{tcolorbox}
\vspace{0.5em}

Here the query is ``What is $15\%$ of $80$?''. There are five steps in the output, where steps 1-4 represent the CoT trace and step 5 represents the conclusion. Both the \texttt{<think>} and \texttt{</think>} tags are valid tokens, and so the model will learn to generate them to indicate the core reasoning steps. On the user end, we can remove such tags to show a clean output. Note that in practical applications, the output might be very complex and include content not related to reasoning. Therefore, introducing \texttt{<think>} and \texttt{</think>} tags can help the model concentrate on solving the reasoning problem.

Reasoning data can also be represented using various other formats. One straightforward alternative is a natural language step-by-step format. Instead of using special tokens, the data relies on text prefixes, such as ``Step 1:'', ``Step 2:'', or simply ``Thought:'' followed later by ``Answer:''. Furthermore, reasoning data can be structured using role-based message separation. In modern chat templates, instead of mixing the reasoning and the answer in a single message, the reasoning trace is assigned to an internal role (e.g., a ``thought'' or ``system\_scratchpad'' role), while the final conclusion is assigned to the standard ``assistant'' role. This separates the internal thinking of the LLM from the user-facing response at the system level.

\subsection{Process Supervision}

As we discussed in Section \ref{sec:reasoning-verification}, PRMs are powerful tools for verifying reasoning steps during inference. A natural extension is to use it to guide the training of LLMs through process supervision.

In standard outcome supervision (such as supervised fine-tuning), the model is only evaluated on the final outcome of the entire reasoning path $\mathbf{y}$. If the model generates a sequence of five steps and makes a logical error in the third step, an outcome-based training approach might simply penalize the entire sequence. The model is left to guess which exact step caused the failure. This is known as the \textbf{credit assignment problem} \cite{minsky:1961steps,sutton:1984temporal}.

Process supervision solves this by applying training signals at the step level. Instead of a single reward for the whole path, the model receives feedback for every individual step $\bar{\mathbf{y}}_t$. For example, imagine teaching a student to solve a complex algebra equation. If you only look at the final number they write down and say ``Wrong'', the student does not know if they misunderstood the formula, made a simple addition error, or forgot a minus sign. Process supervision is like a teacher pointing specifically to line 3 and saying ``Your logic broke down here''.

Process supervision is typically framed as a reinforcement learning problem. To apply reinforcement learning to process supervision, we generally follow a training loop of three steps:

\begin{itemize}
\item \textbf{Rollout}: The LLM receives an input $\mathbf{x}$ and generates a full reasoning path $\mathbf{y} = \{\bar{\mathbf{y}}_1, \dots, \bar{\mathbf{y}}_{n_p}\}$. The rollout process can be repeated multiple times to generate a set of reasoning paths $\mathcal{Y}$.
\item \textbf{Verification}: A reward model (i.e., a PRM checking intermediate steps) evaluates each path $\mathbf{y}$ and returns a reward score $R(\mathbf{x}, \mathbf{y})$.
\item \textbf{Update}: A reinforcement learning algorithm uses these rewards to update the model parameters $\theta$.
\end{itemize}

A simple training objective can be formalized as maximizing the expected reward
\begin{eqnarray}
J(\theta) & = & \mathbb{E}_{\mathbf{y} \sim \mathrm{Pr}_{\theta}(\cdot|\mathbf{x})} \left[ R(\mathbf{x}, \mathbf{y}) \right]. \label{eq:simple-rl-objective-for-reasoning}
\end{eqnarray}

\noindent We can modify this objective to have more robust training, such as introducing the advantage function into the objective as in PPO.

While this looks like a standard reinforcement learning workflow used in general LLM alignment (as described in Chapter \ChapterLLMTuning), there is a fundamental difference in the nature of the reward. In typical LLM alignment tasks, the reward model usually evaluates the entire generated text and returns a sparse reward at the very end. The reinforcement learning algorithm then struggles with the credit assignment problem, as it has to  guess which tokens or segments actually contributed to that final score.

In contrast, reasoning unfolds step-by-step. Therefore, the verification score, which we will denote as $v_t = v(\mathbf{x}, \bar{\mathbf{y}}_{\le t})$ for simplicity, is inherently a step-level signal. This characteristic makes reasoning tasks well-suited for modern reinforcement learning. Because $v_t$ is generated step by step, it naturally transforms a sparse-reward language generation task into a dense-reward environment.

The main technical question then becomes how to properly transform these natural step-level scores $v_t$ into the formal training signals (such as structured step rewards $r_t$ or advantages $A_t$) required by the algorithms. Depending on the output format of the PRM and the reinforcement learning framework, we consider the following three methods.

\subsubsection{Training with Binary Verification Scores}

In some cases, the PRM acts as a strict rule checker (such as a math calculator or a code compiler). It simply returns a binary score: $v_t \in \{0, 1\}$. A score of $1$ means that the logic at step $t$ is correct given the previous steps, and $0$ means that a logical error occurs at this step.

When dealing with binary scores, the idea is straightforward: a single wrong step invalidates the entire reasoning path. If we want to assign a sequence-level reward $R(\mathbf{x}, \mathbf{y})$ for the whole path, we can simply multiply all the step scores together
\begin{eqnarray}
R(\mathbf{x}, \mathbf{y}) & = & \prod_{t=1}^{n_p} v_t.
\end{eqnarray}

\noindent Hence the model only gets a positive reward if the verification score of every single step is $1$.

Alternatively, if the reinforcement learning algorithm supports step-level rewards (denoted as $r_t$), we can penalize the model exactly where the failure happens. For example, we can give a reward of $1$ as long as the steps remain correct ($v_t = 1$). The moment the model makes a mistake ($v_t = 0$), we assign a negative reward and stop the generation.

\subsubsection{Training with Real-valued Verification Scores}

In most neural PRMs, the verifier outputs a continuous real number $v_t \in \mathbb{R}$. This could be a probability from $0$ to $1$ (indicating the confidence that the reasoning path is correct) or an unnormalized scalar value.

To apply continuous verification scores, the most common technique is \textbf{reward shaping}. Instead of using absolute scores, we define the immediate step reward $r_t$ as the marginal difference between the current step and the previous step:
\begin{eqnarray}
r_t & = & v_t - v_{t-1}.
\end{eqnarray}

This method is very intuitive. Imagine a PRM estimating the probability of getting the final answer right. If Step 1 has a confidence of $0.4$, and Step 2 contains a brilliant deduction that raises the confidence to $0.9$, the reward for Step 2 is $r_2 = 0.9 - 0.4 = +0.5$. The model is rewarded for that specific logical leap. If Step 3 introduces a calculation error and the confidence drops to $0.1$, the reward is $r_3 = 0.1 - 0.9 = -0.8$.

When we sum these shaped rewards across the entire sequence, we have a ``telescoping sum'' that forms our sequence-level reward $R(\mathbf{x}, \mathbf{y})$:
\begin{eqnarray}
R(\mathbf{x}, \mathbf{y}) & = & \sum_{t=1}^{n_p} r_t \nonumber \\
& = & (v_1 - v_0) + (v_2 - v_1) + \dots + (v_{n_p} - v_{n_p-1}) \nonumber \\
& = & v_{n_p} - v_0.
\end{eqnarray}

\noindent Here, $v_0$ is generally assumed to be 0. In this derivation, the intermediate scores cancel out. This means that the overall sequence reward remains consistent with the evaluation of the final step.

To understand how this step-by-step dense reward actually trains the model, we need to connect it to the reinforcement learning objective. As is shown in Eq. (\ref{eq:simple-rl-objective-for-reasoning}), our goal is to maximize the expected total reward over all possible generated paths. To make training efficient and have proper credit assignment, we typically use the principle of causality: an action taken at step $t$ can only affect the rewards received after this step. Therefore, rather than evaluating a step using the entire sequence reward, we evaluate it using the future return $R_t$, which is defined as the sum of rewards from step $t$ to the end of the sequence:
\begin{eqnarray}
R_t & = & \sum_{k=t}^{n_p} r_k \nonumber \\
& = & v_{n_p} - v_{t-1}.
\end{eqnarray}

Finally, we can train the model using the surrogate loss function
\begin{eqnarray}
\mathcal{L}(\theta) & = & - \sum_{t=1}^{n_p} \log \mathrm{Pr}_{\theta}(\bar{\mathbf{y}}_t | \mathbf{x}, \bar{\mathbf{y}}_{<t}) \cdot R_t.
\end{eqnarray}

This loss function provides a step-level feedback mechanism. For any generated reasoning path, the model looks at the future return $R_t$ at each step $t$. If $R_t$ is positive, the loss function pushes the model to increase the probability $\log \mathrm{Pr}_\theta(\cdot)$ of generating this step. If $R_t$ is negative, it serves as a penalty, which pushes the model to avoid generating this step in the future.

\subsubsection{Incorporating Verification into the Advantage Function}

Many reinforcement learning algorithms (such as PPO) optimize the model using the advantage function $A_t$, which measures how much better an action was compared to what was expected. Instead of treating verification scores as raw rewards, we can incorporate them into the advantage calculation.

One method is to introduce step-level verification into the temporal difference error. If the PRM is trained to estimate the expected future reward, we can treat it as the value baseline in standard actor-critic reinforcement learning. We calculate the advantage $A_t$ using the temporal difference error
\begin{eqnarray}
A_t & = & \tilde{r}_{t} + \gamma v_t - v_{t-1},
\end{eqnarray}

\noindent where $\tilde{r}_{t}$ is any external reward from the environment (usually $0$ during intermediate thinking steps) and $\gamma$ is the discount factor.

Another method is to improve the design of the group relative advantage. In \mindex{group relative policy optimization} (\mindex{GRPO}), we dispense with absolute verification scores \cite{shao-etal:2024deepseekmath}. Instead, we compute the advantage by comparing a group of reasoning paths. For an input $\mathbf{x}$, the LLM generates a group of different reasoning paths. A verifier assigns a score $v_t^{(i)}$ to the $t$-th step of the $i$-th path. The advantage $A_t^{(i)}$ is calculated by accumulating the standardized scores from step $t$ to the end of the path, as follows
\begin{eqnarray}
A_t^{(i)} & = & \sum_{k = t}^{n_p^{(i)}} \tilde{v}_k^{(i)},
\end{eqnarray}

\noindent where $n_p^{(i)}$ is the total number of steps in the $i$-th path, and $\tilde{v}_k^{(i)}$ is the standardized score, defined as:
\begin{eqnarray}
\tilde{v}_k^{(i)} & = & \frac{v_k^{(i)} - \mu}{\sigma},
\end{eqnarray}

\noindent where $\mu$ and $\sigma$ are the mean and standard deviation of the scores across all steps of the paths in the group. For example, if a model explores 5 different ways to solve an equation, and path \#3 executes a step that scores significantly higher than the average of all steps in these paths, path \#3 receives a positive advantage. Here, the group average $\mu$ serves as the baseline, which removes the need to train and run a separate value network.

Once the advantage $A_t$ (or $A_t^{(i)}$) is calculated using either of these methods, it can be used as the scaling weight in the loss function. A simplified version of the advantage-based loss function can be written as\footnote{Note that advanced algorithms like PPO and GRPO usually enhance this objective by adding a clipping mechanism (restricting the probability ratio between the new and old policies) to prevent the model parameters from changing too drastically in a single update. However, the core mathematical principle remains the same.}:
\begin{eqnarray}
\mathcal{L}(\theta) & = & - \sum_{t=1}^{n_p} \log \mathrm{Pr}_{\theta}(\bar{\mathbf{y}}_t | \mathbf{x}, \bar{\mathbf{y}}_{<t}) \cdot A_t.
\end{eqnarray}

Note that the benefit of using the advantage $A_t$ over the raw return $R_t$ is variance reduction. By subtracting a baseline, whether it is an estimated value $v_{t-1}$ or a group average $\mu$, the advantage function provides a relative performance measure. It tells the model not just whether a reasoning step is logically sound, but whether it is better than expected. This relative feedback prevents the model from being overwhelmed by queries that are too easy or too hard.

\subsection{On-Policy Distillation for Reasoning}

While process supervision provides step-by-step reward signals to guide reasoning, it still relies on a scalar value to tell the model how well it is doing. To provide an even richer training signal, we can employ \mindex{on-policy distillation} (\mindex{OPD}). This method is generally used in post-training of LLMs, and aligns well with the goal of step-based supervision: making the training signals as dense as possible.

Standard knowledge distillation can be viewed as a kind of \mindex{off-policy distillation}. A smaller ``student'' model learns by mimicking the reasoning paths generated by a stronger ``teacher'' model. The student is trained on a fixed dataset of these pre-recorded paths. However, reasoning is a long and fragile process. If the student makes a small mistake or chooses a slightly different valid approach during actual generation, it enters a state it has never seen in the training data. Because the student does not know how to recover from its mistakes, the reasoning process may collapse. This phenomenon is known as exposure bias or distribution shift.

The core idea of OPD is that the student model learns from its own generated trajectories, rather than from trajectories sampled from the teacher model. Formally, let the teacher model be $\mathrm{Pr}^{t}(\cdot)$ and the student model (parameterized with $\theta$) be $\mathrm{Pr}_{\theta}^{s}(\cdot)$. The typical procedure of OPD consists of two steps:

\begin{itemize}
\item \vspace{0.5em} \textbf{Student Rollout}. Given an input $\mathbf{x}$, the student model uses its current parameters $\theta$ to generate a reasoning path $\mathbf{y} = \{\bar{\mathbf{y}}_1, \dots, \bar{\mathbf{y}}_{n_p}\}$. Because the student generates this path itself, it naturally explores its own reasoning behaviors. This includes making slight deviations, taking alternative approaches, or even making mistakes that a teacher would never make.
\item \vspace{0.3em} \textbf{Teacher Guidance}. We then feed this student-generated path to the teacher model. At each step $t$, instead of providing a scalar reward, the teacher model calculates the probability distribution for the next step $\mathrm{Pr}^{t}(\cdot | \mathbf{x}, \bar{\mathbf{y}}_{<t})$.
\end{itemize}
\vspace{0.5em}

Ideally, the student model is updated to align its step-level probability distribution with the distribution generated by the teacher model. The objective is to minimize the expected KL divergence between the two distributions over the reasoning paths generated by the student model:
\begin{eqnarray}\mathcal{L}_{\mathrm{step}}(\theta) & = & \mathbb{E}_{\mathbf{y} \sim \mathrm{Pr}_{\theta}^{s}(\cdot|\mathbf{x})} \left[ \sum_{t=1}^{n_p} D_{\mathrm{KL}} \left( \mathrm{Pr}_{\theta}^{s}(\cdot | \mathbf{x}, \bar{\mathbf{y}}_{<t}) \,\|\, \mathrm{Pr}^{t}(\cdot | \mathbf{x}, \bar{\mathbf{y}}_{<t}) \right) \right].
\end{eqnarray}

\noindent However, calculating this step-level distribution is computationally intractable in practice. Because a reasoning step $\bar{\mathbf{y}}_t$ typically consists of multiple tokens, the probability distribution over all possible multi-token steps resides in an exponentially large space.

To address this, we fall back to the standard OPD approach, which operates at the token level. By treating the generated reasoning path as a token sequence $\mathbf{y} = \{y_1,\dots, y_n\}$, the objective is reformulated to compute the KL divergence token by token over the vocabulary:
\begin{eqnarray}
\mathcal{L}_{\mathrm{token}}(\theta) & = & \mathbb{E}_{\mathbf{y} \sim \mathrm{Pr}_{\theta}^{s}(\cdot|\mathbf{x})} \left[ \sum_{i=1}^{n} D_{\mathrm{KL}} \left( \mathrm{Pr}_{\theta}^{s}(\cdot | \mathbf{x}, \mathbf{y}_{<i}) \,\|\, \mathrm{Pr}^{t}(\cdot | \mathbf{x}, \mathbf{y}_{<i}) \right) \right].
\end{eqnarray}

\noindent This objective is then approximated using sampling. For a given input $\mathbf{x}$, we simply perform a rollout to generate one (or a small batch of) reasoning paths from the student model, and calculate the token-level KL divergence on these sampled rollouts.

Figure \ref{fig:comparision-of-sft-ps-opd} compares OPD with supervised fine-tuning and process supervision. To further understand the difference among these methods, imagine a student trying to solve a complex geometry problem:

\begin{figure}[!t]
\centering
\input{./Chapter12/Figures/figure-comparision-of-sft-ps-opd}
\caption{Comparison of supervised fine-tuning, process supervision, and on-policy distillation methods. In supervised fine-tuning (a), the student model is trained off-policy to maximize the likelihood of golden target outputs provided by a human or a teacher model. In process supervision (b), the student model generates its own reasoning path, and a verifier evaluates these self-generated steps to compute advantages. In on-policy distillation (c), the student also generates its own path, but instead of scalar rewards, a teacher model provides a dense, token-level probability distribution at each step. The student is updated by minimizing the KL divergence between its own predicted distribution and the teacher's distribution.}
\label{fig:comparision-of-sft-ps-opd}
\end{figure}

\begin{itemize}
\item \vspace{0.5em} \textbf{Supervised Fine-tuning}. This is like giving the student a textbook filled with the correct solutions written by the teacher to memorize. If the student decides to draw an auxiliary line differently than the textbook did, the textbook offers no help on how to proceed from that new starting point.
\item \vspace{0.5em} \textbf{Process Supervision} (RL with PRMs). This is like having a strict grader who checks the student's work step-by-step and says, ``+1 point for step 1'', and ``-1 point for step 2''. It tells the student where they went wrong, but not how to fix it or what the correct next step should be.
\item \textbf{On-Policy Distillation}. This is like having an expert tutor sitting right next to the student. When the student draws that unusual auxiliary line (a state not in the standard textbook), the tutor looks at the student's current state and says, ``Okay, since you drew the line here, your next best step is to calculate this specific angle.''
\end{itemize}

Because the supervision of OPD comes in the form of a full probability distribution over the vocabulary, the student receives a very dense and rich learning signal. It learns not only what the single best next step is, but also the probabilities of other valid alternatives. This rich feedback is well suited for reasoning because it teaches the student model how to recover from deviations and maintain logical consistency across the entire output sequence, rather than just treating a rationale as an isolated component. However, applying on-policy distillation comes with a high computational cost. The process requires running a large teacher model alongside the student during training to calculate probability distributions in real time. This is more expensive than standard off-policy distillation, where the outputs of the teacher model are only generated once and stored in a static dataset.

\subsection{Scaling Training for Reasoning}

Scaling is a core strategy in the era of LLMs. When dealing with complex reasoning tasks, scaling the training process is a natural idea. Here, we briefly discuss several key aspects of scaling training for reasoning LLMs: large-scale reinforcement learning, synthetic data generation, and long-context scaling.

\subsubsection{Large-scale Reinforcement Learning}

Reinforcement learning has become a primary approach to improving the ``System 2'' reasoning abilities of LLMs. For example, the success of many powerful reasoning models comes from scaling up reinforcement learning rather than using supervised fine-tuning only. An interesting phenomenon observed during large-scale reinforcement learning-based training is the emergence of the ``aha moment''~\cite{deepseek:2025r1}. Without explicit human annotation, LLMs learn to pause, recognize when they have made a mistake, or even backtrack to try a different approach. These abilities are difficult to elicit when relying solely on limited labeled data and standard supervised training. As training compute scales, the model can allocate longer CoT traces to harder problems, and thus discover how to think deeply on its own~\cite{snell-etal:2024scaling}.

Recent large-scale empirical studies have demonstrated that the relationship between reinforcement learning compute and reasoning performance can be modeled by an S-shaped curve \cite{khatri-etal:2025art,tan-etal:2026scaling}. A key finding from this scaling law is that different training choices impact the curve differently. For instance, increasing the base capabilities of the LLM (e.g., using larger architectures or MoEs) and expanding the context window length can increase the performance limit. Conversely, algorithmic improvements, such as changing the loss function or adjusting advantage normalization, mainly make better use of the available compute budget but do not significantly improve the final performance.

Scaling reinforcement learning is difficult. First, the computational cost is very high. For every training sample, the system generally performs a rollout (running the model for inference) and then calls an extra reward model or another LLM to evaluate the generated output. When the number of training samples increases, this cost becomes significantly higher than that of standard supervised fine-tuning. This computational burden also makes running experiments and trial-and-error much more difficult. Furthermore, reinforcement learning methods are generally complex and require very careful tuning to achieve good results. For example, they are sensitive to hyperparameter settings and training data. As a result, reinforcement learning methods are relatively fragile. When training on a massive scale, ensuring that the system runs stably (for instance, making sure the training loss decreases steadily) is not just a machine learning problem, but also an engineering challenge.

\subsubsection{More Synthetic Data for Reasoning}

Because high-quality human annotations for complex multi-step reasoning are scarce, scaling training data often relies on synthetic data generation. Instead of manually writing step-by-step proofs, researchers use strong LLMs to generate diverse reasoning trajectories, which are then filtered and curated to train new models~\cite{zelikman-etal:2022star,mukherjee-etal:2023orca}.

To generate synthetic data for reasoning, a common approach is to sample multiple reasoning paths from teacher models for a given seed question~\cite{wang-etal:2023self}. We can then apply verifiers or execution environments to evaluate these paths, and retain only those that arrive at the correct final answer through sound logical steps. Furthermore, models can be used to generate variations of existing problems (e.g., changing numerical values or structural constraints in math and coding tasks) to expand the distribution of training data.

\subsubsection{Scaling Context Length}

As reasoning models tackle increasingly complex problems, the length of reasoning paths has become a primary bottleneck. The demand for large context windows in reasoning arises from several sources. First, to elicit deep System 2 thinking, the input prompt itself often requires extensive context. This includes detailed logical constraints, complex environment rules, or multiple long few-shot CoT examples. Second, the generated reasoning process for a complex task can easily span thousands or even tens of thousands of tokens before arriving at the final conclusion. Third, recent advances in LLM agents have also increased the demand for reasoning. Agentic reasoning often extends over many rounds of interaction, creating an even greater need for long context windows.

This demand has directly driven the LLM industry to develop models that support massive context windows. For instance, many LLMs now support context lengths of up to 128K tokens, and more advanced models, such as Anthropic's Claude 3.5~\cite{anthropic:2024claude35} and Google's Gemini 1.5~\cite{gemini:2024gemini}, support hundreds of thousands of tokens, or even one million tokens. The interested reader may refer to Chapter \ChapterLLM\ for a more detailed discussion on long-context LLMs.

However, handling long contexts for reasoning leads to new challenges that go beyond general context compression. In tasks like summarizing a long novel or extracting key points from documents, an LLM can afford to compress information and lose minor details. In contrast, reasoning requires strict, high-precision logical continuity. Every intermediate step is a critical link in a heavily interdependent chain. If the model ``forgets'' a specific assumption defined in the prompt, or loses track of a variable calculated in step 2 while executing step 50, the entire derivation collapses. Therefore, scaling context for reasoning is not just about expanding the KV cache to store more tokens. It requires the model to achieve reliable retrieval across long distances and to maintain strict logical consistency.

%----------------------------------------------------------------------------------------
% NEW SECTION
%----------------------------------------------------------------------------------------

\section{Evaluating Reasoning LLMs}

In this section, we discuss how to evaluate reasoning LLMs. This task is similar to evaluating standard LLMs \cite{chang-etal:2024survey}. However, reasoning problems are so complex that we often need more detailed evaluation and other special considerations.

\subsection{Outcome-based vs. Process-based Evaluation}

When evaluating how well an LLM reasons, the easiest and most common method is \textbf{outcome-based evaluation}. In this approach, we only look at the final answer produced by the model. If the final answer matches the ground truth, the model is considered correct. This method is widely used in popular reasoning benchmarks, such as GSM8K \cite{cobbe-etal:2021training} and MATH \cite{hendrycks-etal:2measuring}. However, outcome-based evaluation has the ``right answer for the wrong reason'' problem, also known as a false positive. A model might generate flawed logical steps but accidentally arrive at the correct final number. If we only check the final outcome, we might incorrectly conclude that the model possesses strong reasoning capabilities when it just got lucky.

For more detailed evaluation, we can use \textbf{process-based evaluation}. Instead of just checking the final answer, we can examine the entire reasoning path step by step \cite{wang-etal:2024math,xia-etal:2025evaluating}. A model is only considered correct if both its intermediate logic and its final conclusion are sound. This is an ideal method but is very difficult to implement in practice. The main challenge is that step-level ``gold data'' is scarce and expensive to create. Furthermore, for any complex problem, there is an exponentially large number of valid reasoning paths. It is infeasible to manually list all correct intermediate steps for every path a model might take.

To make process-based evaluation more practical, we can narrow the scope to evaluate local reasoning decisions. One approach is to reduce the problem to a simple next-step evaluation task: given a correct sequence of previous steps, we check whether the reasoning model can correctly choose or generate the next logical step. To do this, it is straightforward to use PRMs to evaluate the quality of each step generated by the LLM. This method is also related to reward model evaluation \cite{lambert-etal:2025rewardbench}. Instead of examining the output of the reasoning model, we assess the quality of the PRMs. Furthermore, we can consider the fundamental capabilities of reasoning models, such as their ability to identify errors in a given reasoning path. This can also be viewed as a test of their meta-reasoning abilities: we ask the model not just to reason, but to evaluate the reasoning process itself \cite{zeng-etal:2024mr}.

In practice, process-based evaluation is not based on a single standard. We can judge the quality of a reasoning step from different angles, such as logical consistency, coherence, and objectivity \cite{lee-and-hockenmaier:2025evaluating}. However, while process-based evaluation is much more detailed, it comes with a much higher annotation cost, and is often not robust enough. Therefore, most real-world research still relies on the simpler outcome-based evaluation.

\subsection{Evaluation Metrics}

Here we list some metrics used for evaluating reasoning LLMs.

\begin{itemize}
\item \vspace{0.5em} \textbf{Exact Match and Accuracy:} This is the simplest and most widely used metric for tasks with definitive final answers, such as math word problems. Suppose we have a dataset $\mathcal{D}$ consisting of $N$ problems. For the $i$-th problem, we extract the final answer $\hat{a}_i$ from the generated reasoning path and compare it to the ground truth $a_i^*$. The accuracy is defined as:
\begin{eqnarray}
\text{Accuracy} & = & \frac{1}{N} \sum_{i=1}^N \mathbbm{1}(\hat{a}_i = a^*_i).
\end{eqnarray}

\noindent In practice, the exact match method (checking if the strings are perfectly identical) can be too strict, for example, we reject ``$0.5$'' when the ground truth is ``$1/2$''. Therefore, researchers usually apply rule-based scripts to normalize the answers into a canonical format before comparison.

\item \vspace{0.3em} \textbf{Pass@$k$:} Because LLMs are probabilistic, a single decoding attempt might fail in reasoning due to a minor error. Pass@$k$ measures the probability that the model can find at least one correct answer if it is allowed to generate $k$ independent reasoning paths \cite{chen-etal:2021evaluating}. This metric is especially popular in code generation and complex math tasks. For the $i$-th problem in $\mathcal{D}$, if we sample $N_s \ge k$ paths and $C_i$ of them contain the correct answer, the Pass@$k$ score is defined as:
\begin{eqnarray}
\text{Pass}@k & = &  \frac{1}{N} \sum_{i=1}^N \left( 1 - \frac{\binom{N_s-C_i}{k}}{\binom{N_s}{k}} \right).
\end{eqnarray}

\noindent To generate multiple different reasoning paths, we can use the sampling methods described in Section \ref{sec:sampling-reasoning-paths}.

\item \vspace{0.3em} \textbf{Cons@$k$ (Consensus@$k$):} Unlike Pass@$k$, which requires an oracle to verify whether each of the $k$ sampled answers is correct, Cons@$k$ constructs a consensus answer without identifying the correctness of individual samples. Specifically, we generate $k$ reasoning paths, extract their final answers, and select the most frequent answer through majority voting \cite{wang-etal:2023self}. The consensus answer $\hat{a}_i^{\text{maj}}$ for the $i$-th problem is defined as:
\begin{eqnarray}
\hat{a}_i^{\text{maj}} & = & \argmax_{a_i} \sum_{j=1}^k \mathbbm{1}(\hat{a}_i^j = a_i),
\end{eqnarray}

\noindent where $\hat{a}_i^j$ is the final answer extracted from the $j$-th reasoning path. The final accuracy is then obtained by comparing $\hat{a}_{\text{maj}}$ with the ground truth:
\begin{eqnarray}
\text{Cons}@k & = & \frac{1}{N} \sum_{i=1}^N \mathbbm{1}(\hat{a}_i^{\text{maj}} = a^*_i).
\end{eqnarray}

Cons@$k$ provides a practical evaluation approach for real-world reasoning systems, as it measures whether a model can consistently arrive at the correct answer. Compared with Pass@$k$, which evaluates whether at least one sampled response is correct, Cons@$k$ emphasizes the reliability of reasoning by favoring answers that are consistently supported across multiple sampled reasoning paths.

\item \vspace{0.3em} \textbf{LLM-as-a-Judge:} For open-ended reasoning tasks, such as agentic planning or complex open-domain questions, rule-based exact matching often fails. In such cases, we can employ LLMs as automated judges. For instance, a judge LLM can be prompted to evaluate the generated intermediate steps and score them based on predefined criteria. Alternatively, LLMs can be used to compare two reasoning models. The judge is presented with the outputs from both models for the same query and decides which reasoning path is superior, or if the result is a tie. The win rate then reflects the percentage of times a specific model is preferred.
\end{itemize}
\vspace{0.5em}

\subsection{Open Issues}

One issue is data contamination. As language models are trained on increasingly massive datasets collected from the internet, it becomes very difficult to keep test data out of the training process. If a model has already seen the questions from a reasoning benchmark (like MATH or GSM8K) during its pre-training phase, it might simply be reciting a memorized solution instead of actually reasoning on the fly. This makes it hard to determine whether high benchmark scores reflect true problem-solving abilities or just good memory. To address this, the community is encouraged to explore new benchmarks, where problems are updated continuously, as well as private hold-out sets that are never released to the public.

Another issue is robustness. Current reasoning models are sometimes fragile. A model might solve a complex math problem perfectly, but fail completely if the exact same question is rephrased slightly or if irrelevant details are added to the prompt. This sensitivity suggests that models sometimes rely on superficial text patterns rather than truly understanding the underlying logic. For more robust reasoning, we need LLMs to easily adapt to new or even more challenging tasks.

Moreover, LLMs may generate hallucinated or unfaithful reasoning paths. For example, they might invent fake mathematical rules to reach a conclusion. In some cases, the generated text might look like a perfect logical deduction, but the model actually arrived at the answer via a hidden mechanism (such as a lucky guess or a memorized shortcut). This unfaithfulness highlights a gap between the text the model outputs and its actual internal representations. In reinforcement learning setups, this is often related to the reward hacking problem. If models are only rewarded for the final answer, they often learn to cheat the reward system with flawed logic. Therefore, it is valuable to introduce step-by-step supervision in RL, such as process supervision, to check if each intermediate step is truly correct.

The fourth issue is benchmark saturation. The rapid progress of reasoning LLMs means that standard evaluation benchmarks are quickly becoming outdated. Models are now achieving near-perfect scores on tests that were considered extremely difficult just a year or two ago. This benchmark saturation makes it difficult to compare new models or measure real progress in the field. Therefore, researchers need to develop harder tests that evaluate true logical derivation on novel problems. Note that human-level reasoning is often open-ended and interactive. An alternative approach to testing reasoning models might be to use them in autonomous agents, which can explore, use external tools, and recover from their own errors.

%----------------------------------------------------------------------------------------
% NEW SECTION
%----------------------------------------------------------------------------------------

\section{Remarks}

We now discuss some interesting issues related to reasoning LLMs.

\subsection{External Reasoning vs. Internal Reasoning}

In previous discussions, we assumed that the reasoning process can be well captured by a text-based reasoning path (e.g., CoT). This raises a fundamental question: Does the generated CoT text actually reflect how the model reasons? This inquiry leads us to examine the distinction between externalized reasoning and the true internal ``thinking'' processes.

When we prompt an LLM to think step by step, we are forcing it to externalize its computation into discrete tokens. However, we cannot simply view this text sequence as the actual reasoning process. On one hand, an LLM might generate a detailed and logically sound CoT, but the actual decision-making might have already occurred implicitly within the hidden representations of the very first few tokens. On the other hand, a model might produce a flawed or nonsensical CoT, yet still output the correct final answer, suggesting that its internal representations are tracking the task differently from what the generated text implies.

Researchers have demonstrated that CoT explanations can be deceptive \cite{turpin-etal:2023language,chen-etal:2025reasoning}. For example, LLMs may have bias towards generating a logical CoT to reach an incorrect answer, though the correct answer has been mentioned in the prompt. To uncover how LLMs actually think, researchers are turning to neuron activation analysis and mechanistic interpretability. The goal is to understand what happens inside the hidden layers of an LLM during the generation process. For example, in \citet{frasertaliente-etal:2026natural}'s work, natural language auto-encoders are used to convert the activations into natural language text. A natural language auto-encoder consists of an activation verbalizer and an activation reconstructor, which are used to directly map internal residual stream activations into human-readable text and back. In this way, we can decode these activations to read the thoughts of LLMs.

Another line of research focuses on reasoning without text, called \mindex{latent reasoning} or \mindex{implicit reasoning}. In standard reasoning, as we defined earlier, a reasoning path consists of a sequence of text steps. In latent reasoning, this process happens internally before the model speaks. This is achieved either by replacing discrete tokens with continuous vectors or by generating actual text words in the background that are kept hidden from the user. Here we present some of these methods.

\begin{itemize}
\item \vspace{0.5cm} One of the simplest ideas is the use of ``pause tokens''. As is standard in autoregressive generation, a language model gets one step of internal calculation for every word it outputs. Interestingly, it is discovered that if we feed the model dummy tokens (called pause tokens), the model gets extra computational time to process information in its hidden layers before it has to output the final answer \cite{goyal-etal:2024think}. This shows that LLMs could perform silent computations or latent thinking without generating real words.
\item \vspace{0.3cm} Another approach uses knowledge distillation to compress explicit reasoning into hidden states. For instance, the implicit chain of thought method trains a student model to directly mimic the hidden internal states of a larger teacher model that is thinking out loud \cite{deng-etal:2023implicit}. As a result, the student model learns to skip generating the text. Alternatively, the LLM can also learn from itself using self-distillation techniques \cite{shen-etal:2025codi}. We can force the latent thinking process to match the final hidden states of the text-based reasoning. In both methods, latent thinking is embodied as a continuous mathematical space: instead of generating a step-by-step sequence of readable text, the model computes the reasoning steps using hidden vector-based representations.
\item \vspace{0.3cm} An alternative method to reason in continuous latent spaces is to use the \mindex{chain of continuous thought} (\mindex{COCONUT}) paradigm \cite{hao-etal:2025training}. In standard text generation, the output of the final layer is converted into a discrete token, which is then fed back into the first layer for the next step. This process forces the model to compress its internal computation into a single token. COCONUT bypasses this step during the reasoning phase. It takes the output vector from the last layer and feeds it directly back into the input layer as a \mindex{continuous thought}. By repeating this process, the model builds a chain of continuous vectors before it starts generating the final text answer. Because continuous vectors can hold more information than discrete tokens, the model can explore multiple reasoning paths at the same time without committing to a specific token choice early on. This approach demonstrates that LLMs do not have to rely on human language to solve complex problems. By keeping the reasoning process in a continuous space, they can compute answers more efficiently.
\item \vspace{0.3cm} Instead of performing reasoning in a continuous space, we can also use normal text tokens to represent internal thoughts. For example, in the Quiet-STaR method \cite{zelikman-etal:2024quietstar}, the model is trained to think quietly, right in the middle of reading or writing normal text. The idea is simple: before the model predicts the next token, it generates a short sequence of hidden ``thought'' tokens. You can think of this as an inner monologue or a quick mental scratchpad that the user never sees. After generating these hidden thoughts, the LLM uses them to help guess the actual next word. If these thoughts help the LLM predict the correct word, the LLM is rewarded and learns that this way of thinking is useful. If the thoughts are useless or confusing, the LLM learns to discard them.
\end{itemize}
\vspace{0.5cm}

\subsection{Memorization vs. Reasoning}

When discussing LLMs, a question often arises: Is the model truly ``thinking'', or does it simply have a big memory and is reciting a similar answer it saw in its training data? This leads to the discussion on comparing reasoning and memorization in LLMs.

Simply put, the difference between the two can be summarized as follows:

\begin{itemize}
\item \vspace{0.5cm} Memorization is looking backward. The model adjusts its parameters to learn patterns from the training data. When faced with a similar input, the model simply replays the learned patterns.
\item \vspace{0.3cm} Reasoning is moving forward. Instead of searching for a ready-made answer, the model applies general rules and logic to build a chain of thought step by step. The core of reasoning is generalization, even when facing a completely new problem that never appeared in the training set, the model can solve it using logical tools.
\end{itemize}
\vspace{0.5cm}

In early deep learning research, this boundary was not always clear. For example, \citet{zhang-etal:2021understand} point out that deep neural networks have massive memorization capacity, easily memorizing noisy data with completely random labels. For LLMs, given their huge model sizes, they can not only memorize Wikipedia but also details hidden inside code repositories.

This memorization ability often disguises itself as reasoning. For example, if we prompt an LLM with a classic algorithm problem but tweak the numbers slightly, the LLM might rely on its memory inertia and output the original answer anyway, thus making a basic mistake. This is often referred to as the ``stochastic parrots'' phenomenon~\cite{bender-etal:2021dangers}, which means that the LLM is blindly mimicking and stitching together memorized text without truly understanding the underlying logic.

To better see the difference between memorization and reasoning, we can look at two different dimensions.

\subsubsection{Grokking and Generalization}

When exploring how models transition from memorization to true understanding, researchers has made an interesting discovery called ``grokking''~\cite{power-etal:2022grokking}. They train a small Transformer model to learn binary operations in abstract algebra (such as modular addition). They notice a very interesting phenomenon: early in training, the model quickly reached 100\% accuracy on the training set, but its accuracy on the test set was almost zero. At this stage, the model is clearly just memorizing equations by rote. However, if they do not stop training and let the model continue looping over the data long past the point of overfitting, something surprising happens. Suddenly, the test accuracy spikes to 100\% in a very short time. The model seems to experience an ``aha moment''.

Further studies show that before this ``aha moment'', the model parameters are messily memorizing specific equations (like $2 + 3 = 5$)~\cite{nanda-etal:2023progress}. But when grokking occurs, the network internally develops a general mathematical representation using trigonometric functions and Fourier transforms. It has found a universal algorithm that can explain all the data.

This experiment tells us that reasoning does not appear out of nowhere. It often emerges as a high-level generalization state when memorization hits a bottleneck, or after extensive training, where the network organizes its parameters to find the most universal explanation.

\subsubsection{The Cost Trade-off}

From the perspective of information theory and computational efficiency, LLMs naturally balance memorization and reasoning based on cost. On one hand, memorization requires more parameter capacity. To memorize every fact in the world, the model size would need to grow infinitely. On the other hand, reasoning requires more computational steps at inference time. As long as a model memorizes the general rules of addition and multiplication, it can solve many math problems by taking such computational steps during generation.

Therefore, LLMs usually take the path of least resistance. For common facts, retrieving the answer directly from parameters (or memory) is the fastest way. But for rare, multi-step tasks, using System 2 thinking is a more suitable way.

This also explains why techniques like RAG have become popular. The essence of RAG is to offload the burden of memory to an external database, and thus we can free up the parameter space for reasoning. By separating knowledge retrieval from logical deduction, the LLM no longer needs to memorize ever-changing external facts. Instead, it knows how to look things up and process them.

\subsection{Can LLMs Truly Reason?}

The question of whether LLMs can truly reason is one of the most debated topics in AI today. When we see an LLM perfectly solve a math problem or write a complex schedule, it is very tempting to believe that it understands the world just as a human does. However, to answer this question, we must look at how these models actually work.

Essentially, LLMs work by predicting the next word. Because these models have processed massive amounts of text, they have completely mastered the rules, grammar, and structure of human language. In cognitive science, this is known as \mindex{formal linguistic competence} \cite{chomsky:1965aspects,mahowald-etal:2024dissociating}. They know exactly how words should be logically ordered. However, they struggle with functional linguistic competence, which is the ability to use language to understand and navigate the real world \cite{kambhampati-etal:2024llms}. Unlike humans, who learn by physically interacting with their environment, LLMs only know the underlying patterns of text. Because they are not grounded in the real world, they lack a true understanding of physical space and strict cause-and-effect \cite{mahowald-etal:2024dissociating}.

This limitation becomes obvious when we test LLMs on planning tasks \cite{valmeekam-etal:2023planbench}. For example, for classical planning problems, such as moving colored blocks on a table into a specific stack, the LLM can often solve the problem if it uses common words they had seen in their training data. But if the names of the blocks and the actions are replaced with nonsense words, the performance dropped drastically \cite{kambhampati-etal:2024can}. A true reasoning system would not care about the names, because the underlying mathematical logic is exactly the same. The LLM fails because its familiar text pattern is broken.

Furthermore, techniques like CoT prompting force the model to output a step-by-step process. This shows some behavior of slow and deliberate human thought. But in reality, the LLM is still just moving strictly forward, and predicts what a logical explanation usually looks like. If the LLM makes a mistake or holds a biased assumption, it will often confidently generate a flawed explanation simply to justify its wrong answer \cite{chen-etal:2025reasoning}.

Ultimately, to answer whether LLMs can ``truly'' reason, we first need to define what true reasoning actually means. If we expect LLMs to think exactly like humans, it is clear from the discussion above that there are some differences. However, if we view reasoning purely as a functional process, the answer changes. Reasoning can be defined simply as the ability to take a complex problem, break it down, apply logical rules, and arrive at a correct conclusion. When an LLM correctly solves a math equation step by step, or when it writes a logical piece of code to process data, it is executing a valid logical chain. Even if this ability emerges from training large-scale neural networks rather than human-like cognition, the outcome is still successful problem-solving. As LLMs grow larger and learn from more data, they move beyond simply memorizing answers and start capturing the underlying rules of logic.

%----------------------------------------------------------------------------------------
% NEW SECTION
%----------------------------------------------------------------------------------------

\section{Summary}

In this chapter, we have presented how to apply LLMs to reasoning, which is considered one of the most challenging tasks in NLP. In particular, we have discussed methods of improving reasoning abilities in the training and inference phases. We have also studied evaluation methods for reasoning LLMs, as well as some open problems.

The finding that LLMs can explicitly map out their reasoning steps has inspired new research directions and practical applications. Instead of viewing reasoning only as a way to answer isolated questions, current research treats reasoning as a tool that drives more complex and autonomous behaviors.

First, reasoning has been closely related to the concept of ``long thinking'', which fundamentally changes how computational power is used in LLMs. As we discussed earlier in Section \ref{sec:inference-time-reasoning}, by giving models more time and computational power to think, they can solve much harder problems. This brings a big change to how LLMs work: we now need to figure out how to best distribute computing power while the model is generating an answer, rather than just during its initial training. Solving this at the system level is beyond the scope of this book. However, building highly efficient reasoning models that know how much time to spend thinking is a very promising and practical research direction.

Second, the ability to reason has been the basis of agentic reasoning, where LLMs stop being passive answering machines and become active participants in an environment. In standard text generation, reasoning is a straight line from prompt to answer. Agentic reasoning, by contrast, operates in a continuous loop of thinking, acting, and observing \cite{yao-etal:2023react}. However, this loop leads to big challenges. The issue is that LLMs can easily get lost or stuck in the loop. They might repeat the same mistakes, make bad plans, or not know how to recover from failures when an action fails \cite{shinn-etal:2023reflexion}. Keeping the model on track over a long time is still very hard.

Third, reasoning directly enables tool-augmented reasoning, which solves the limitations of LLMs, such as their inability to perform precise math or access real-time facts. Instead of forcing the LLM to guess an answer, we can train the LLM to recognize its own limits and write code \cite{gao-etal:2023pal} or call external APIs \cite{schick-etal:2024toolformer} to do the heavy lifting. However, the main problem is teaching the LLM when it needs a tool and which one to pick out of many \cite{qin-etal:2024toolllm}. Also, if a tool breaks or returns weird data, the LLM must be smart enough to catch the error and try again, rather than just trusting the bad result.

\cleardoublepage

\renewcommand{\sectionmark}[1]{\markright{#1}}

\addcontentsline{toc}{chapter}{Bibliography}

\pagestyle{fancy}
\fancyhf{}
\fancyhead[LE,RO]{\thepage}
\fancyhead[RE,LO]{Bibliography}
\renewcommand{\headrulewidth}{0.4pt}

\bibliographystyle{book-bib-style} % Give more information to each entry of the bib
\bibliography{bibliography}
\cleardoublepage

\pagestyle{fancy}
\fancyhf{}
\fancyhead[LE,RO]{\thepage}
\fancyhead[RE,LO]{Index}
\renewcommand{\headrulewidth}{0.4pt}

\appendix
\pretocmd{\chapter}{\pagenumbering{arabic}
\renewcommand*{\thepage}{\thechapter.\arabic{page}}
}{}{}

\printindex % Show index

\end{document}